\PassOptionsToPackage{dvipsnames,table}{xcolor} 
\documentclass[11pt, a4paper, logo, copyright, nonumbering]{map}

\usepackage{CJKutf8}
\usepackage{array}      
\usepackage{fontawesome}

\usepackage{tikz}
\usepackage{multicol}
\usepackage{multirow}
\usepackage{caption}
\usepackage{longtable}
\usepackage{subcaption}
\usepackage{afterpage}
\usepackage{tcolorbox}  

\usepackage{url}
\usepackage{makecell}
\usepackage{xspace}

\usepackage{tocloft}   

\usepackage{natbib}

\usepackage{xargs}  

\usepackage{todonotes}  

\usepackage{cleveref}

\usepackage{dsfont}

\usepackage{svg}
\usepackage{lscape}
\usepackage{hhline}
\usepackage{float}
\definecolor{boxcolor}{HTML}{d92523} 
\definecolor{bulbcolor}{HTML}{e3b87f} 

\newcommand{\benchmark}{\textit{SuperGPQA}\xspace}
\arrayrulecolor[HTML]{660000} 
\newcommand{\listmodelsname}{\normalsize{List of Model Score Tables}}
\newlistof{models}{csf}{\listmodelsname}

\newcommand{\centeredlinks}[5]{%
    \begin{center}
        \textbf{\hyperlink{#1}{\textcolor{#5}{#2}}}
        \quad 
        {\textcolor{#5}\textbar} 
        \quad 
        \textbf{\hyperlink{#3}{\textcolor{#5}{#4}}}
    \end{center}
}

\hypersetup{
    colorlinks=true,            
    linkcolor=blue,             
    filecolor=magenta,          
    urlcolor=cyan,              
    citecolor=purple,             
    pdftitle={Overleaf Example},
    pdfpagemode=FullScreen,
    }
 
\renewcommand{\today}{}
\newcommandx{\info}[2][1=]{\todo[linecolor=red,backgroundcolor=red!25,bordercolor=red,#1]{#2}}
\newgeometry{top=2cm,bottom=2.4cm} 

\title{\centering \benchmark: Scaling LLM Evaluation across \\285 Graduate Disciplines}

\author{
\textbf{M-A-P} \\
ByteDance Seed, 2077.AI \\
\url{https://supergpqa.github.io/}
}

\begin{abstract}

Large language models (LLMs) have demonstrated remarkable proficiency in mainstream academic disciplines such as mathematics, physics, and computer science. However, human knowledge encompasses over 200 specialized disciplines, far exceeding the scope of existing benchmarks. The capabilities of LLMs in many of these specialized fields—particularly in light industry, agriculture, and service-oriented disciplines—remain inadequately evaluated.
To address this gap, we present \textbf{\benchmark}, a comprehensive benchmark that evaluates graduate-level knowledge and reasoning capabilities across 285 disciplines. Our benchmark employs a novel Human-LLM collaborative filtering mechanism to eliminate trivial or ambiguous questions through iterative refinement based on both LLM responses and expert feedback.
Our experimental results reveal significant room for improvement in the performance of current state-of-the-art LLMs across diverse knowledge domains ( \emph{e.g.}, the reasoning-focused model DeepSeek-R1 achieved the highest accuracy of 61.82\% on \benchmark), highlighting the considerable gap between current model capabilities and artificial general intelligence. Additionally, we present comprehensive insights from our management of a large-scale annotation process, involving over 80 expert annotators and an interactive Human-LLM collaborative system, offering valuable methodological guidance for future research initiatives of comparable scope.
\end{abstract}

\tcbuselibrary{skins, breakable}  
\begin{document}
\begin{CJK*}{UTF8}{gbsn}

\maketitle
\begin{figure}[ht]
\vspace{8pt}
\begin{center}
\includegraphics[width=1.0\linewidth]{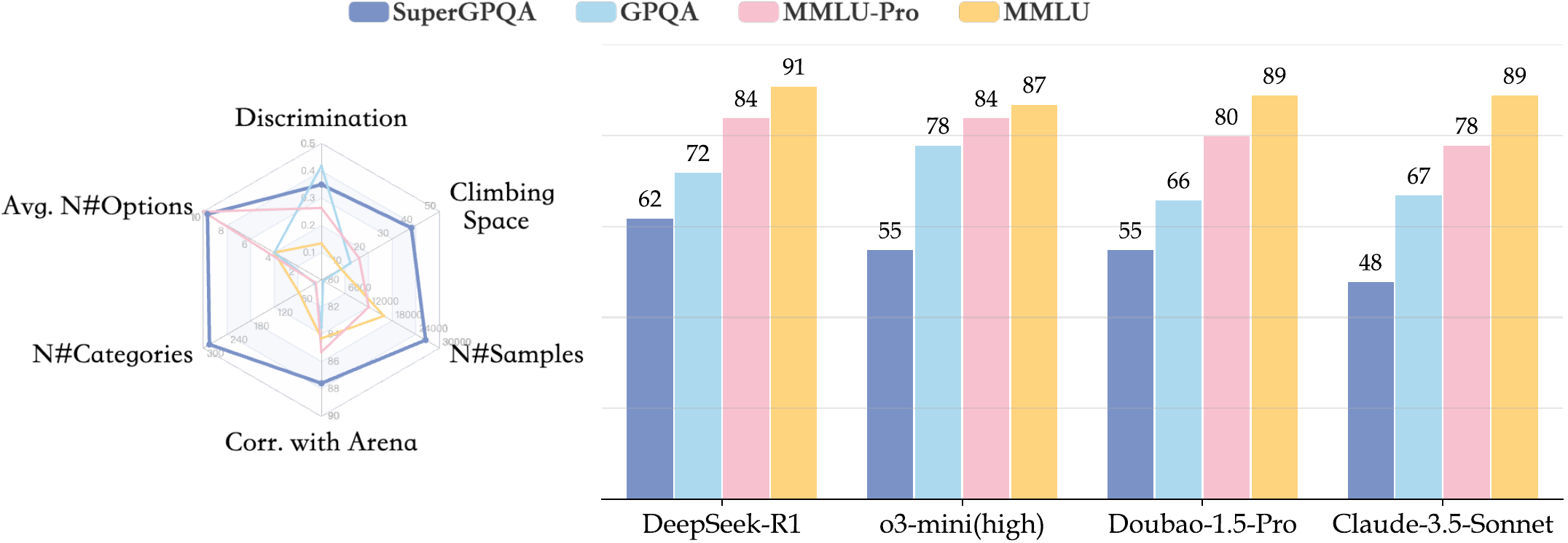}
\end{center}
\captionsetup{font={small}}
\vspace{-10pt}
\caption{
\textbf{Benchmark Comparison.} 
\textbf{Left}: Radar Chart. 
\textbf{Discrimination:} The degree of distinction between different models (detailed in Sec.~\ref{sec:discrimination}).
\textbf{Climbing Space:} The remaining improvement space for the SOTA models. 
\textbf{Corr. with Arena:} Correlation with Chatbot Arena Elo scores.
\textbf{Right}: Performance comparison of SOTA models across different benchmarks.
} 
\label{Fig: Comparison benchmarks }
\end{figure}

\newpage
\newgeometry{top=3cm,bottom=3cm} 


\hypertarget{toc}{}
\tableofcontents

\newpage

\section{Introduction}
Large language models (\textbf{LLMs}) have greatly changed human life.
LLMs are seen as the next technological singularity and proved to surpass human performance in many areas~\citep{phan2025hle}, significantly improving work efficiency.
However, the measurement of LLMs' accessibility on various real-world professionalism remains an unresolved issue, especially in the long-tailed fields with less attention, such as light industry, agriculture, and
various service-related disciplines.
Many popular benchmarks, such as MMLU~\citep{hendrycks2020measuring}, GPQA~\citep{rein2023gpqa}, and MMLU-pro~\citep{wang2024mmlupro}, evaluate LLMs' abilities across different fields, while mainly focus on common fields like mathematics, physics, chemistry, biology, and law, limiting these benchmarks' practical significance on many real-world professionalisms.
These benchmarks fail to cover the diverse and long-tail knowledge accumulated by humans.
Moreover, large language models have achieved very high scores on these benchmarks, making them lose their value as challenging frontiers.

To address the gap, we introduce \textbf{\benchmark}, a comprehensive evaluation at the boundaries of human knowledge covering the evaluation of 285 graduate-level disciplines' knowledge and reasoning capacities.
\benchmark provides at least 50 questions for each graduate-level disciplines to guarantee its accessibility on various real-world professionalism.
\benchmark is developed by a large-scale human-LLM collaboration system, with crowd-sourcing annotators, experts, and state-of-the-art (\textbf{SOTA}) LLMs participating in, and then verified by a rigorous 3-stage quality inspection process, to guarantee its reliability.
Moreover, \benchmark is qualified as a challenging frontier for SOTA reasoning LLMs, instruct LLMs, and base LLMs, 
where the best LLMs (e.g., o1 and Deepseek-R1) only achieve a score of around 60.

For building \benchmark, we propose a large-scale human-LLM collaboration system and share the valuable lessons learned in the paper.
We divide the annotation system of \benchmark into three major stages: \textbf{Source Screening}, \textbf{Transcription}, and \textbf{Quality Inspection}.
\textbf{During the source screening stage}, expert annotators collect credible resources of different disciplines' questions to guarantee the reliability and difficulty of the raw questions.
\textbf{During the transcription stage}, crowd-sourcing annotators are asked to revise or translate the raw questions to multiple-choice questions, generate complementary confusion options, and estimate the difficulty and reliability of these questions.
Crowd-sourcing annotators estimate the difficulty and reliability of candidate questions based on both expert judgments and the accuracy of LLMs' responses during the annotation process.
Crowd-sourcing annotators, rigorous real-time plagiarism checks with existing candidate questions, and a robust filtering system based on SOTA LLMs are adopted in the transcription stage to reduce the waste of funding and expert manpower.
\textbf{During the quality inspection stage}, we adopt a rigorous three-stage quality inspection process:
\begin{itemize}
    \item We select suspicious candidate questions based on a checklist of LLMs' responses. 
    \item Expert annotators review the suspicious candidate questions with unrestricted access to the web and revise these questions.
    \item The easy questions are further tailored based on the accuracy of LLMs' responses to guarantee the discrimination of \benchmark.
\end{itemize}

We share several major insights based on the evaluation results of \benchmark:

\begin{itemize}
 \item  \textbf{Reasoning capacities matter}. The reasoning models (e.g., DeepSeek-R1, o1-2024-12-17) achieve the best performance in \textbf{\benchmark}. 
    \item  \textbf{Instruction tuning is very helpful}. 
    For example,
    the results (47.40, 40.75) of DeepSeek-V3 and Qwen2.5-72B-Instruct are better than the results (32.14, 34.33) of DeepSeek-V3-Base and Qwen2.5-72B a lot, respectively.
    \item \textbf{More powerful LLMs lead to more balanced results}.
    On different difficulties. the results of simple, middle, and hard splits of DeepSeek-R1 are 63.59, 63.63, and 56.87. In contrast, the results of easy, middle, and hard splits of Qwen2.5-14B-Instruct are 44.82, 37.90, and 19.97.

    \item \textbf{Models are better in newer versions.}
    For example,  the results of GPT-4o-2024-11-20,  GPT-4o-2024-08-06, and GPT-4o-2024-05-13 are 44.40, 41.64, and 39.76, respectively. 
\end{itemize}

\section{Data Collection}\label{sec:data_collection}
\begin{figure*}[ht]
\begin{center}
\includegraphics[width=\linewidth]{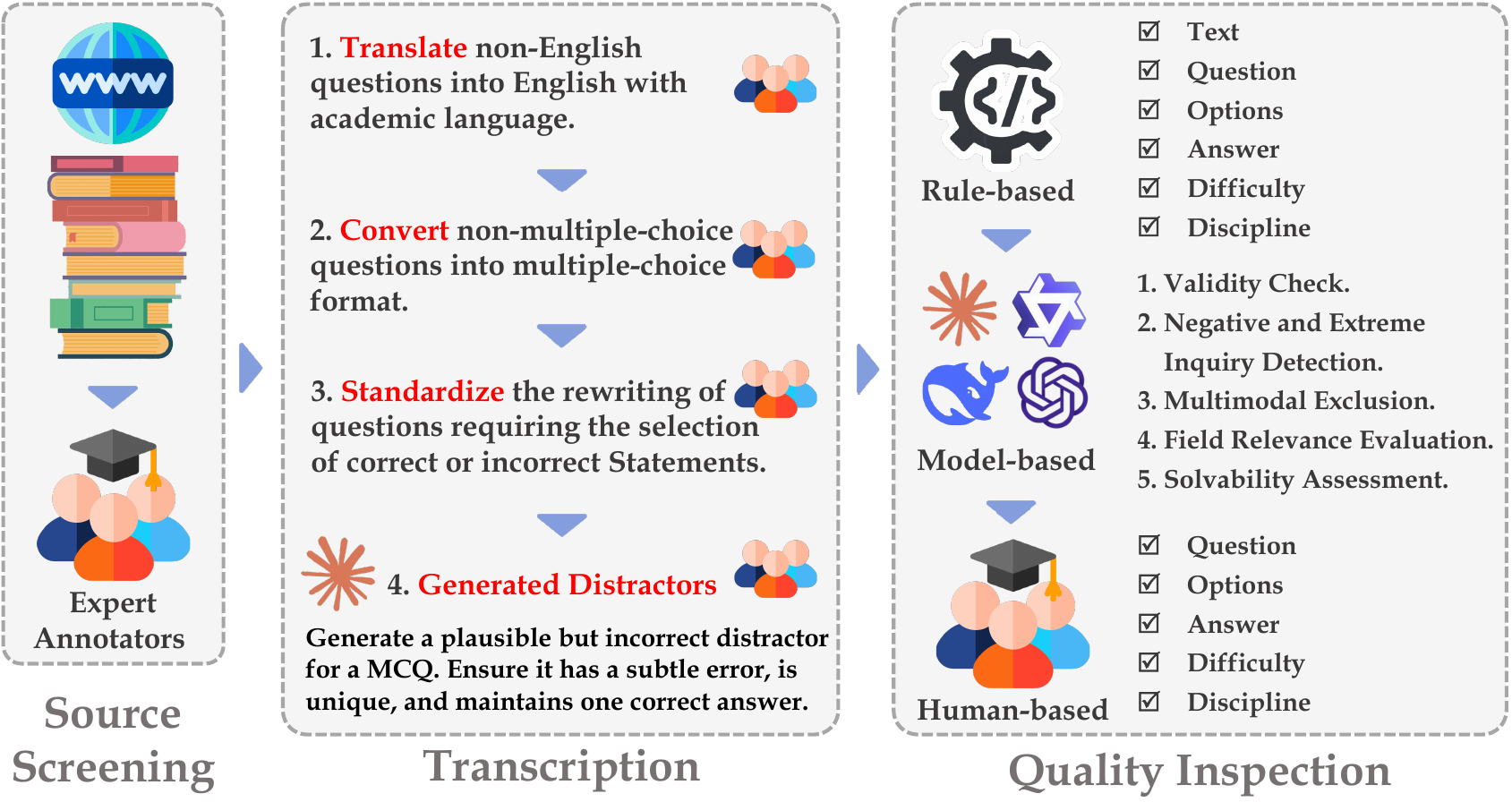}
\end{center}
\caption{Data Collection Process of \benchmark.}
\label{Fig: Data Collection}
\end{figure*}

We solicit difficult questions from well-educated experts and crowd-sourcing annotators, where we consider experts as individuals having or pursuing a PhD, as in GPQA~\citep{rein2023gpqa}, and crowd-sourcing annotators as undergraduates and master students from top-tier Chinese universities, i.e.mainly from Tsinghua University, Peking University, Zhejiang University, Beihang University, and Chinese Academy of Sciences.
\autoref{Fig: Data Collection} shows the three major stages of \benchmark's data collection pipeline: Source Screening, Transcription, and Quality Inspection, which are separately detailed in \autoref{Sec: Source-Screening}, \autoref{Sec: Transcription}, and \autoref{Sec: Quality-Inspection}.
First, experts select credible resources of different
disciplines’ questions.
Second, crowd-sourcing annotators revise the raw questions from credible resources to candidate questions.
Finally, a rigorous human-LLM collaboration quality inspection process is adopted to select difficult and reliable questions from candidate questions.

\subsection{Source Screening}
\label{Sec: Source-Screening}

\begin{tcolorbox}[colframe=boxcolor,colback=white,boxrule=0.5mm,arc=0mm]
{\Large \textcolor{bulbcolor}{\faLightbulbO}}  \textbf{Lessons:}
\begin{itemize}
    \item Crowd-Sourcing Annotators are not capable of collecting credible resources for multiple-choice question annotation with high expertise requirements.
    \item The questions and answers (\textbf{QAs}) on exercise websites are not always reliable, even sometimes these QAs are claimed verified.
    \item Multiple-choice questions modified from calculation and reasoning problems usually are more discriminatory than the original multi-choice questions available online.
\end{itemize}
\end{tcolorbox}

During the source screening stage, only expert annotators are allowed to collect credible resources of different disciplines’ questions to guarantee the reliability and difficulty of the raw questions.
In the early stage of collecting candidate questions of \benchmark, we trust crowd-sourcing annotators to collect credible resources themselves.
However, the candidate questions based on the resources found by the crowd-sourcing annotators are always judged too easy or unreliable by expert annotators.
As a result, a significant portion of early funding is wasted on ineffective questions annotated by crowd-sourcing annotators.
Additionally, we point out that QAs on exercise websites are not always reliable. 

In the early stage of \benchmark collection process, some annotators, even expert annotators, trust exercise websites that serve as corroboration of their reasoning process and answers.
In the subsequent quality inspection stage, this proved to be a costly mistake, leading us to spend a significant amount of time and cost correcting erroneous answers derived from online exercise websites.
Furthermore, we find that many SOTA LLMs, such as GPT-4o, o1-mini, and Gemini-flash, exhibit a high frequency of consistency in both process and answers with the erroneous processes and answers from several online exercise websites. 
The observations reveal that \textbf{the reliability of the solutions provided by online exercise websites is limited and there is a significant risk of data leakage}.

Expert annotators are asked to provide raw questions from credible resources with screenshots for further annotation in the source screening stage.
The screenshot greatly eases the workload of quality inspection.
We observe that the efficiency of quality inspection is greatly improved with the provided source screenshot.
The priority order for selecting original questions is:
\begin{itemize}
    \item Example problems with solutions from textbooks.
    \item Calculation and reasoning-needed questions with solutions from websites.
    \item Reasoning-needed multiple-choice questions with solutions from websites.
    \item General multiple-choice questions with solutions from websites.
    \item Questions only with answers but deemed correct by expert annotators.
\end{itemize}
A sampled list of credible resources certified by expert annotators is provided in 
\autoref{Sec: Benchmark List} for reference.

\subsection{Transcription}
\label{Sec: Transcription}

\begin{figure*}[ht]
\begin{center}
\includegraphics[width=\linewidth]{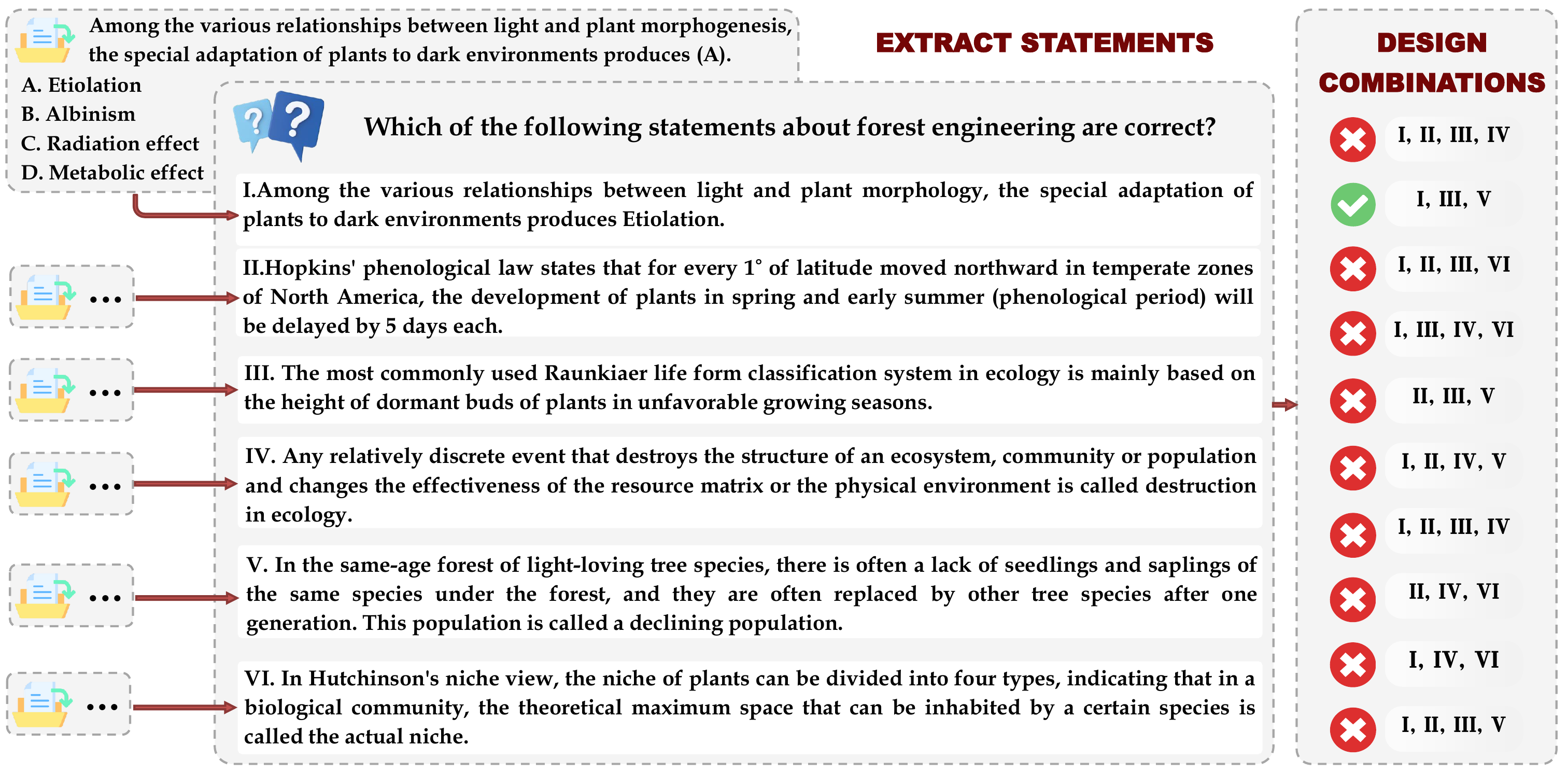}
\end{center}
\caption{Rewriting Samples of Questions Requiring the Selection of Correct or Incorrect Options.}
\label{Fig: rewriting process}
\end{figure*}

\begin{tcolorbox}[colframe=boxcolor,colback=white,boxrule=0.5mm,arc=0mm]
{\Large \textcolor{bulbcolor}{\faLightbulbO}}  \textbf{Lessons:}
\begin{itemize}
    \item Crowd-sourcing annotators have low accuracy in judging generated distractors. For question types like selecting correct or incorrect options, it is easy to generate flawed distractors, requiring unified rewriting at this stage.
\end{itemize}
\end{tcolorbox}
During the transcription stage, crowd-sourcing annotators are asked to revise the original questions into candidate questions.
Specifically,  the following operations are performed:
\begin{itemize}
    \item Translate non-English questions into English with academic language.
    \item Convert non-multiple-choice questions into multiple-choice format.
    \item Standardize the rewriting of questions requiring the selection of correct or incorrect statements, as shown in \autoref{Fig: rewriting process}.
    \item Include region-specific information where necessary, such as specifying the country for laws mentioned in the questions, except for universally accepted rules.
\end{itemize}
The questions requiring the selection of correct or incorrect statements must be standardized as shown in \autoref{Fig: rewriting process}.
Because we notice that even SOTA LLMs, e.g. Claude-3.5-Sonnet, GPT-4o-0806, suffer from generating correct suitable confounders for questions requiring the selection of correct or incorrect statements.
The full transcription tutorial is provided in \autoref{appendix: annotation tutorial}.

\subsection{Quality Inspection}
\label{Sec: Quality-Inspection}

\begin{tcolorbox}[colframe=boxcolor,colback=white,boxrule=0.5mm,arc=0mm]
{\Large \textcolor{bulbcolor}{\faLightbulbO}}  \textbf{Lessons:}
\begin{itemize}
    \item Questions where LLMs choose the same incorrect option are highly suspicious.
    \item Cases where multiple or all SOTA LLMs make the same error often indicate that the LLMs have memorized explanations from incorrect exercise websites, based on SOTA LLMs' responses.
\end{itemize}
\end{tcolorbox}
We refer to the data quality inspection and filtering methods used in LIME and MMLU-Redux~\cite{zhu2024lime,gema2024we,wu2024comparativestudyreasoningpatterns}, etc. The quality inspection process consists of three substages: \textbf{Rule-based Quality Inspection}, \textbf{LLM-based Quality Inspection}, and \textbf{Expert-based Quality Inspection}.

Candidate questions with clear formatting issues are identified and filtered out by rule-based quality inspection. 
The full checklist of rule-based quality inspection refers to \autoref{Appendix: Rule-Based Check}.
We then adopt several SOTA LLMs to generate responses and additional tags to these reserved candidate questions.
LLM-based quality inspection includes validity checks, negative and extreme inquiry detection, multimodal exclusion, field relevance evaluation, completeness assessment, and discrimination tagging based on SOTA LLMs' responses.
It provides an estimation of not only the correctness but also the discrimination of the candidate questions.
Finally, we ask the expert annotators to re-annotate the suspicious candidate questions.
The checklist for selecting suspicious candidate questions refer to \autoref{Appendix: LLM-Based Screening}.
We adopt GPT-4o-2024-08-06, Gemini-2.0-flash, Doubao-1.5-pro-32k-250115, Claude-3.5-Sonnet, DeepSeek-R1, QwQ, Qwen-2.5-72B-Instruct as SOTA LLMs for selecting suspicious candidate questions.
During the re-annotation process, following the rules stated in GPQA~\citep{rein2023gpqa}, expert annotators are asked to review and solve the given candidate questions with unrestricted access to the web.
They spend over 30 minutes on each candidate question according to the post-annotation interview.
The tutorial for manual quality review refers to \autoref{Appendix: Manual Quality Review}.

\section{Statistics}

\begin{table}[!ht]
    \centering
\small
\resizebox{\textwidth}{!}{%
\begin{tabular}{@{}lcccccccccc@{}}
\toprule
\multirow{2}{*}{Discipline} & \multirow{2}{*}{\#Num.}  & \multicolumn{3}{c}{Question \#Tokens} & \multicolumn{3}{c}{Answer \#Tokens } & \multicolumn{2}{c}{Options} &\multirow{2}{*}{Cal. Rate} \\
\cmidrule(lr){3-5} \cmidrule(lr){6-8} \cmidrule(lr){9-10} 
& & Max & Min & Avg & Max & Min & Avg & \#Num. &  \#Tokens \\
\midrule
Engineering & 7892 & 715 & 4 & 67.26 & 352 & 1 & 13.86 & 9.76 & 13.67 & 55.40\% \\
Medicine & 2755 & 447 & 4 & 39.06 & 89 & 1 & 7.57 & 9.67 & 7.38 & 3.34\% \\
Science & 9838 & 623 & 5 & 69.77 & 372 & 1 & 16.89 & 9.71 & 16.75 &66.34\% \\
Philosophy & 347 & 573 & 5 & 44.63 & 71 & 1 & 8.93 & 9.63 & 8.42 & 0.29\%\\
Military Science & 205 & 336 & 7 & 30.03 & 50 & 1 & 6.76 & 9.29 & 6.40 & 3.41\%\\
Economics & 314 & 314 & 5 & 36.15 & 98 & 1 & 7.85 & 9.86 & 7.12 & 20.50\%\\
Management & 501 & 308 & 7 & 41.87 & 217 & 1 & 7.77 & 9.72 & 6.62 & 3.79\% \\
Sociology & 143 & 107 & 6 & 25.31 & 75 & 1 & 7.34 & 9.87 & 6.91 & 2.10\% \\
Literature & 1676 & 869 & 5 & 35.80 & 88 & 1 & 6.82 & 8.78 & 6.75 & 0.60\% \\
History & 674 & 539 & 6 & 28.38 & 125 & 1 & 5.32 & 9.64 & 5.49 & 1.04\% \\
Agriculture & 485 & 331 & 7 & 27.25 & 33 & 1 & 5.32 & 9.91 & 5.24 & 1.44\% \\
Law & 656 & 560 & 6 & 66.38 & 89 & 1 & 12.17 & 9.73 & 11.45 &0.46\%\\
Education & 484 & 173 & 7 & 23.35 & 37 & 1 & 5.74 & 9.78 & 5.39 & 0.41\%\\
\hline
Overall & 26529 & 869 & 4 & 58.42 & 372 & 1 & 12.86 & 9.67 & 12.64 &42.33\%\\
\bottomrule
\end{tabular}}
\caption{Statistics of \benchmark. Tokens are calculated with Tiktoken using \texttt{\href{https://huggingface.co/BEE-spoke-data/cl100k_base}{cl100k\_base}} encoding.}
\label{tab:overall_stats}
\end{table}

\newtcolorbox[auto counter, number within=section]{mybox}[2][]{%
    colframe=white,  
    colback=white,  
    boxrule=0mm,    
    width=\textwidth, 
    title= ,        
    sharp corners,  
    valign=center,  
    boxsep=0mm,
    outer arc=0mm,
    arc=0mm,
    left=2mm,      
    right=2mm,     
    top=0mm,       
    bottom=0mm,    
    #1,              
    before=\vspace{0mm},  
    after=\vspace{0mm},   
}

{
\linespread{1}
\setlength{\tabcolsep}{0pt}  
\renewcommand{\arraystretch}{0}
\begin{table}[p]
\centering
\resizebox{1\textwidth}{!}{
\begin{tabular}{>{\raggedright\arraybackslash}p{3.5cm}>{\raggedright\arraybackslash}p{17cm}}
\toprule
\textbf{Discipline} & \textbf{Field : Subfield} \\
\hline
\begin{mybox}[colback=teal!80, coltext=white, height=2.5cm]{}\textbf{Agronomy(485)}\end{mybox} &\begin{minipage}[t]{17cm}
\begin{mybox}[colback=teal!20, coltext=black, height=2.5cm]{}
\textbf{\textcolor{teal!80!black}{Animal Husbandry(103)}} : Animal Nut. \& Feed Sci.; Animal Rear. \& Breed.\\\textbf{\textcolor{teal!80!black}{Aquaculture(56)}} : Aquacult.\\\textbf{\textcolor{teal!80!black}{Crop Science(145)}} : Crop Sci.\\\textbf{\textcolor{teal!80!black}{Forestry(131)}} : Forest Cult. \& Gen. Breed.; Landsc. Plants \& Orn. Hort.\\\textbf{\textcolor{teal!80!black}{Veterinary Medicine(50)}} : Vet. Med.\end{mybox}
\end{minipage}\\
\begin{mybox}[colback=red!60!white!40, coltext=white, height=1.5cm]{}\textbf{Economics(873)}\end{mybox} &\begin{minipage}[t]{17cm}
\begin{mybox}[colback=pink!20, coltext=black, height=1.5cm]{}
\textbf{\textcolor{pink!80!black}{Applied Economics(723)}} : Econ. Stats.; Fin.; Indus. Econ.; Int. Trade; Labor Econ.; Nat. \& Def. Econ.; Pub. Fin.; Quant. Econ.\\\textbf{\textcolor{pink!80!black}{Theoretical Economics(150)}} : Econ. Hist.; Pol. Econ.; West. Econ.\end{mybox}
\end{minipage}\\
\begin{mybox}[colback=cyan!80, coltext=white, height=1.5cm]{}\textbf{Education(484)}\end{mybox} &\begin{minipage}[t]{17cm}
\begin{mybox}[colback=cyan!20, coltext=black, height=1.5cm]{}
\textbf{\textcolor{cyan!80!black}{Education(247)}} : Edu. Tech. \& Prin.; Presch. Edu.; Spec. Edu.; Theory of Curric. \& Instr.\\\textbf{\textcolor{cyan!80!black}{Physical Education(150)}} : Phys. Edu. \& Train.; Sports Hum. \& Socio.; Sports Sci. \& Med.\\\textbf{\textcolor{cyan!80!black}{Psychology(87)}} : Psychol.\end{mybox}
\end{minipage}\\
\begin{mybox}[colback=blue!80!black!40, coltext=white, height=22cm]{}\textbf{Engineering(7892)}\end{mybox} &\begin{minipage}[t]{17cm}
\begin{mybox}[colback=blue!20, coltext=black, height=22cm]{}
\textbf{\textcolor{blue!60!black!60}{Aeronautical and Astronautical Science and Technology(119)}} : Aeronaut. \& Astronaut. Sci. \& Tech.\\\textbf{\textcolor{blue!60!black!60}{Agricultural Engineering(104)}} : Agric. Environ. \& Soil-Water Eng.; Agric. Mech. Eng.\\\textbf{\textcolor{blue!60!black!60}{Architecture(162)}} : Arch. Design \& Theory; Arch. Hist.; Urban Plan. \& Design\\\textbf{\textcolor{blue!60!black!60}{Chemical Engineering and Technology(410)}} : Chem. Transport Eng.; Elem. of Chem. React. Eng.; Fluid Flow \& Heat Transfer in Chem. Eng.; Mass Trans. \& Sep. Process in Chem. Eng.\\\textbf{\textcolor{blue!60!black!60}{Civil Engineering(358)}} : Bridge \& Tunnel Eng.; Geotech. Eng.; Struct. Eng.; Urban Infra. Eng.\\\textbf{\textcolor{blue!60!black!60}{Computer Science and Technology(763)}} : Adv. Prog. Lang.; Comp. Arch.; Comp. Net.; Comp. Soft. \& Theory; Data Struc.; Databases; Formal Lang.; Oper. Sys.; Pattern Recog.; Princip. of Comp. Org.\\\textbf{\textcolor{blue!60!black!60}{Control Science and Engineering(190)}} : Control Theory \& Eng.; Guid. Nav. \& Control; Oper. Res. \& Cyber.\\\textbf{\textcolor{blue!60!black!60}{Electrical Engineering(556)}} : Elect. Theory \& New Tech.; High Volt. \& Insul. Tech.; Power Elec. \& Elec. Drives; Power Sys. \& Autom.\\\textbf{\textcolor{blue!60!black!60}{Electronic Science and Technology(246)}} : Circuits \& Sys.; Electromag. Field \& Microwave Tech.; Microelect. \& Solid-State Elec.\\\textbf{\textcolor{blue!60!black!60}{Environmental Science and Engineering(189)}} : Environ. Eng.; Environ. Sci.; Environ. \& Res. Protect.\\\textbf{\textcolor{blue!60!black!60}{Food Science and Engineering(109)}} : Food Biochem.; Food Proc. \& Stor. Eng.\\\textbf{\textcolor{blue!60!black!60}{Forestry Engineering(100)}} : Forest Eng.; Wood Sci. \& Tech.\\\textbf{\textcolor{blue!60!black!60}{Geological Resources and Geological Engineering(50)}} : Geol. Res. \& Geol. Eng.\\\textbf{\textcolor{blue!60!black!60}{Hydraulic Engineering(218)}} : Hydraul. \& Hydrol.; Water Cons. \& Hydropower Eng.\\\textbf{\textcolor{blue!60!black!60}{Information and Communication Engineering(504)}} : Antenna \& Radio Comm.; Comm. Prin.; Comm. \& Info. Sys.; Optical Fiber Comm.; Signal \& Info. Proc.\\\textbf{\textcolor{blue!60!black!60}{Instrument Science and Technology(50)}} : Instr. Sci. \& Tech.\\\textbf{\textcolor{blue!60!black!60}{Materials Science and Engineering(289)}} : Mater. Phys. \& Chem.; Mater. Proc. Eng.\\\textbf{\textcolor{blue!60!black!60}{Mechanical Engineering(176)}} : Manuf. Autom.; Mechatron. Eng.\\\textbf{\textcolor{blue!60!black!60}{Mechanics(908)}} : Fund. of Dyn. \& Control; Rigid Body Mech.; Solid Mech.; Theor. Fluid Mech.; Theor. Mech.\\\textbf{\textcolor{blue!60!black!60}{Metallurgical Engineering(255)}} : Iron \& Steel Metall.; Non-fer. Metall.; Phys. Chem. of Metall. Proc.; Princip. of Metall.\\\textbf{\textcolor{blue!60!black!60}{Mining Engineering(100)}} : Mineral Proc. Eng.; Mining \& Safety Eng.\\\textbf{\textcolor{blue!60!black!60}{Naval Architecture and Ocean Engineering(138)}} : Marine Eng.; Ship Mech. \& Design Prin.\\\textbf{\textcolor{blue!60!black!60}{Nuclear Science and Technology(107)}} : Nuc. Energy \& React. Tech.; Radiation Prot. \& Nuclear Tech. Appl.\\\textbf{\textcolor{blue!60!black!60}{Optical Engineering(376)}} : Applied Opt.; Laser Tech.; Optoelect. Tech.; Theor. Opt.\\\textbf{\textcolor{blue!60!black!60}{Petroleum and Natural Gas Engineering(112)}} : Oil \& Gas Field Dev. \& Stor. \& Trans. Eng.; Poromech. \& Res. Phys.\\\textbf{\textcolor{blue!60!black!60}{Power Engineering and Engineering Thermophysics(684)}} : Eng. Fluid Mech.; Eng. Thermophys.; Fluid Mach. \& Eng.; Heat Trans.; Internal Comb. Eng.; Power Mach. \& Eng.; Refrig. \& Cryogen. Eng.; Thermal Energy Eng.\\\textbf{\textcolor{blue!60!black!60}{Surveying and Mapping Science and Technology(168)}} : Carto. \& Geo. Info. Eng.; Dig. Survey. \& Remote Sens. Appl.; Geodesy \& Survey. Eng.\\\textbf{\textcolor{blue!60!black!60}{Textile Science and Engineering(100)}} : Text. Chem. \& Dyeing Eng.; Text. Mater. Sci.\\\textbf{\textcolor{blue!60!black!60}{Transportation Engineering(251)}} : Road \& Rail. Eng.; Traffic Info. Eng. \& Control; Transp. Plan. \& Manag.; Vehicle Oper. Eng.\\\textbf{\textcolor{blue!60!black!60}{Weapon Science and Technology(100)}} : Mil. Chem. \& Pyro.; Weapon Syst. Sci. \& Eng.\end{mybox}
\end{minipage}\\
\begin{mybox}[colback=brown!60!white, coltext=white, height=0.8cm]{}\textbf{History(674)}\end{mybox} &\begin{minipage}[t]{17cm}
\begin{mybox}[colback=brown!20, coltext=black, height=0.8cm]{}
\textbf{\textcolor{brown!80!black!60}{History(674)}} : Archaeol. \& Museol.; Hist. Geo.; World Hist.\end{mybox}
\end{minipage}\\
\bottomrule
\end{tabular}
}
\caption{The Disciplinary Categories of \benchmark (1/2).}
\label{tab:discipline_1}
\end{table}
}

{
\linespread{1}
\setlength{\tabcolsep}{0pt}  
\renewcommand{\arraystretch}{0}
\begin{table}[p]
\centering
\resizebox{1\textwidth}{!}{
\begin{tabular}{>{\raggedright\arraybackslash}p{3.5cm}>{\raggedright\arraybackslash}p{17cm}}
\toprule
\textbf{Discipline} & \textbf{Field : Subfield} \\
\hline
\begin{mybox}[colback=red!80!black!60, coltext=white, height=1.5cm]{}\textbf{Law(656)}\end{mybox} &\begin{minipage}[t]{17cm}
\begin{mybox}[colback=red!20, coltext=black, height=1.5cm]{}
\textbf{\textcolor{red!80!black!60}{Law(591)}} : Civil \& Comm. Law; Const. \& Admin. Law; Contract Law; Crim. Law; Int. Law; Law \& Soc. Gov.; Legal Theory \& Hist.; Mil. Law; Proced. Law\\\textbf{\textcolor{red!80!black!60}{Political Science(65)}} : Pol. Sci.\end{mybox}
\end{minipage}\\
\begin{mybox}[colback=purple!80!black!60, coltext=white, height=3.8cm]{}\textbf{Literature \\and Arts(1676)}\end{mybox} &\begin{minipage}[t]{17cm}
\begin{mybox}[colback=purple!20, coltext=black, height=3.8cm]{}
\textbf{\textcolor{purple!80!black!60}{Art Studies(603)}} : Broad. \& TV Art; Dance Stud.; Design Arts; Drama \& Opera Stud.; Film Stud.; Fine Arts\\\textbf{\textcolor{purple!80!black!60}{Journalism and Communication(207)}} : Comm. \& Broad.; Hist. \& Theory of Jour. \& Media Mngmt.; Jour. \& News Prac.\\\textbf{\textcolor{purple!80!black!60}{Language and Literature(440)}} : Class. Chinese Lit.; Fr. Lang. \& Lit.; Ling. \& Appl. Ling.; Lit. Hist.; Lit. Theory; Mod. \& Cont. Chinese Lit.; Phil. \& Bib.; Russ. Lang. \& Lit.\\\textbf{\textcolor{purple!80!black!60}{Musicology(426)}} : Comp.; Harm.; Instr. \& Perf.; Music Hist., Ed. \& Tech.; Music Forms \& Anal.; Pitch \& Scales\end{mybox}
\end{minipage}\\
\begin{mybox}[colback=orange!80!black!60, coltext=white, height=3cm]{}\textbf{Management(501)}\end{mybox} &\begin{minipage}[t]{17cm}
\begin{mybox}[colback=orange!20, coltext=black, height=3cm]{}
\textbf{\textcolor{orange!80!black!60}{Business Administration(142)}} : Bus. \& Acct. Mngmt.; Tour. Mngmt. \& Tech. Econ. Mngmt.\\\textbf{\textcolor{orange!80!black!60}{Library, Information and Archival Management(150)}} : Info. Mngmt. Sci.; Info. Mngmt. \& Comm.; Lib. \& Arch. Sci.\\\textbf{\textcolor{orange!80!black!60}{Management Science and Engineering(58)}} : Mngmt. Sci. \& Eng.\\\textbf{\textcolor{orange!80!black!60}{Public Administration(151)}} : Ed. Econ., Mngmt. \& Soc. Sec.; Land Res. Mngmt. \& Admin. Mngmt.; Soc. Med. \& Health Mngmt.\end{mybox}
\end{minipage}\\
\begin{mybox}[colback=yellow!80!black, coltext=white, height=6cm]{}\textbf{Medicine(2755)}\end{mybox} &\begin{minipage}[t]{17cm}
\begin{mybox}[colback=yellow!20, coltext=black, height=6cm]{}
\textbf{\textcolor{yellow!80!black}{Basic Medicine(567)}} : For. Med.; Hum. Anat. \& Hist.-Emb.; Immun.; Path. Biol.; Pathol. \& Pathophys.; Rad. Med.\\\textbf{\textcolor{yellow!80!black}{Clinical Medicine(1218)}} : Anesth.; Clin. Lab. Diagn.; Derm. \& Ven.; Emerg. Med.; Geriat. Med.; Imag. \& Nucl. Med.; Intern. Med.; Neurol.; Nurs. \& Rehabil. Med.; Obst. \& Gyneco.; Oncol.; Ophth.; Oto. \& Rhinol.; Pediatr.; Psych. \& Ment. Health; Surg.\\\textbf{\textcolor{yellow!80!black}{Pharmacy(278)}} : Medic. Chem.; Microbiol. \& Biochem. Pharm.; Pharm. Anal.; Pharmaceut.; Pharmacol.\\\textbf{\textcolor{yellow!80!black}{Public Health and Preventive Medicine(292)}} : Epidemiol. \& Health Stats.; Health Tox. \& Envir. Health; Matern., Child \& Adol. Health; Nutr. \& Food Hyg.\\\textbf{\textcolor{yellow!80!black}{Stomatology(132)}} : Basic Stom.; Clin. Stom.\\\textbf{\textcolor{yellow!80!black}{Traditional Chinese Medicine(268)}} : Trad. Chin. Health Pres.; Trad. Chin. Med. Theory; Trad. Chin. Pharm.\end{mybox}
\end{minipage}\\
\begin{mybox}[colback=magenta!80!black!60, coltext=white, height=1cm]{}\textbf{Military \\Science(205)}\end{mybox} &\begin{minipage}[t]{17cm}
\begin{mybox}[colback=magenta!20, coltext=black, height=1cm]{}
\textbf{\textcolor{magenta!80!black!60}{Military Science(205)}} : Mil. Command \& Info. Systems; Mil. Logistics \& Equip.; Mil. Mngmt.; Mil. Thought \& Hist.\end{mybox}
\end{minipage}\\
\begin{mybox}[colback=lime!80!black!60, coltext=white, height=0.8cm]{}\textbf{Philosophy(347)}\end{mybox} &\begin{minipage}[t]{17cm}
\begin{mybox}[colback=lime!20, coltext=black, height=0.8cm]{}
\textbf{\textcolor{lime!80!black!60}{Philosophy(347)}} : Ethics; Logic; Phil. Aesth.; Phil. of Sci. \& Tech.; Relig. Stud.\end{mybox}
\end{minipage}\\
\begin{mybox}[colback=green!60!black!60, coltext=white, height=10.5cm]{}\textbf{Science(9838)}\end{mybox} &\begin{minipage}[t]{17cm}
\begin{mybox}[colback=green!20!white!60, coltext=black, height=10.5cm]{}
\textbf{\textcolor{green!80!black!60}{Astronomy(405)}} : Astron. Obs. \& Tech.; Astrophys.; Cosmology; Solar Sys. Sci.; Stell. \& Interst. Evol.\\\textbf{\textcolor{green!80!black!60}{Atmospheric Science(203)}} : Atm. Phys. \& Envir.; Dyn. Meteorol.; Meteorol.\\\textbf{\textcolor{green!80!black!60}{Biology(1120)}} : Biochem. \& Mol. Biol.; Biophys.; Botany; Cell Biol.; Ecol.; Genet.; Microbiol.; Physiol.; Zool.\\\textbf{\textcolor{green!80!black!60}{Chemistry(1769)}} : Analyt. Chem.; Electrochem.; Inorg. Chem.; Org. Chem.; Phys. Chem.; Polym. Chem. \& Phys.; Radiochem.\\\textbf{\textcolor{green!80!black!60}{Geography(133)}} : Hum. Geogr.; Phys. Geogr.\\\textbf{\textcolor{green!80!black!60}{Geology(341)}} : Geochem.; Miner., Petrol. \& Econ. Geol.; Paleontol. \& Stratig.; Prin. of Seism. Expl.; Struct. Geol.\\\textbf{\textcolor{green!80!black!60}{Geophysics(100)}} : Solid Earth Geophys.; Space Phys.\\\textbf{\textcolor{green!80!black!60}{Mathematics(2622)}} : Adv. Algebra; Combinat. Math.; Comput. Math.; Crypt.; Discr. Math.; Func. of Complex Vars.; Func. of Real Vars.; Fund. Math.; Fuzzy Math.; Geo. \& Topol.; Graph Theory; Group Theory; Math. Anal.; Num. Theory; Num. Anal.; Ord. Diff. Eq.; Poly. \& Ser. Exp.; Prob. \& Stats.; Spec. Num. Theory; Stoch. Proc.\\\textbf{\textcolor{green!80!black!60}{Oceanography(200)}} : Hydrogeol.; Marine Biol.; Marine Chem.; Underwater Acou.\\\textbf{\textcolor{green!80!black!60}{Physical Oceanography(50)}} : Phys. Oceanogr.\\\textbf{\textcolor{green!80!black!60}{Physics(2845)}} : Acou.; Atom. \& Mol. Phys.; Electrodyn.; Fluid Phys.; Part. \& Nucl. Phys.; Polym. Phys.; Quant. Mech.; Relativity; Semicond. Phys.; Solid State Phys.; Stat. Mech.; Subatom. \& Atom. Phys.; Thermodyn.; Thermo. \& Stat. Phys.\\\textbf{\textcolor{green!80!black!60}{Systems Science(50)}} : Sys. Sci.\end{mybox}
\end{minipage}\\
\begin{mybox}[colback=olive!80!black!60, coltext=white, height=1cm]{}\textbf{Sociology(143)}\end{mybox} &\begin{minipage}[t]{17cm}
\begin{mybox}[colback=olive!20, coltext=black, height=1cm]{}
\textbf{\textcolor{olive!80!black}{Sociology(143)}} : Demo. \& Anthrop.; Soc. \& Folklore Studies\end{mybox}
\end{minipage}\\

\bottomrule
\end{tabular}
}
\caption{The Disciplinary Categories of \benchmark (2/2).}
\label{tab:discipline_2}
\end{table}
}

We propose \benchmark, designed as a comprehensive benchmark to probe the upper bounds of state-of-the-art Large Language Models' capabilities. With 26,529 questions spanning 13 disciplines, 72 fields, and 285 subfields, it substantially surpasses existing benchmarks in both scale and taxonomic depth. Compared to similar ``hard'' benchmarks such as GPQA (448 questions) and MMLU-Pro (12,032 questions), \benchmark not only contains a larger question pool but also features a more challenging format with an average of 9.67 options per question, significantly higher than the conventional 4-option format (\emph{e.g.}, MMLU).

\paragraph{Comprehensiveness and Discrimination.} As revealed in \autoref{tab:overall_stats}, the distribution of questions across disciplines reveals a notable concentration in STEM fields, with Science (9,838 questions), Engineering (7,892 questions), and Medicine (2,755 questions) collectively accounting for 77.2\% of the benchmark. While this distribution might appear uneven at first glance, it emerges from our rigorous question collection and validation process. During the data annotation phase, we source a comparable number of reference books for 285 subfields (the subfields are detailed in \autoref{tab:discipline_1} and \autoref{tab:discipline_2}). Note that we also use some open-source datasets for supplementation of \benchmark in \autoref{Sec: Benchmark List}. However, the STEM disciplines yielded more questions meeting our stringent quality and difficulty criteria aforementioned in \autoref{sec:data_collection}, where the questions and options are filtered through rigorous rule-based, model-based and human-based pipeline. This natural emergence of STEM-heavy distribution aligns with the benchmark's goal of probing LLMs' upper-bound capabilities in complex reasoning tasks.
Despite the relatively smaller representation of non-STEM disciplines (\emph{e.g.}, Philosophy: 347, Literature: 1,676, History: 674 questions), our experiments demonstrate that these subsets effectively discriminate various SOTA LLMs' performance levels (detailed in \autoref{sec:results}). This once again validates the discriminative power of our benchmark across all domains, regardless of sample size.

\paragraph{Difficulty.} The difficulty distribution across disciplines 
(\autoref{tab:discipline_difficulty_distribution})
reveals varying levels of complexity. In STEM fields, we observe a more balanced distribution of difficulty levels. For instance, Engineering questions are distributed as 31.1\% hard, 43.9\% middle, and 25.0\% easy, while Science shows 42.8\% hard, 42.0\% middle, and 15.2\% easy. Non-STEM disciplines generally show a different pattern, with a higher proportion of easy and middle-difficulty questions. Notably, 42.33\% of all questions require mathematical calculations or formal reasoning, with Science (66.34\%) and Engineering (55.40\%) showing the highest calculation rates.

\begin{figure*}[ht]
\begin{center}
\includegraphics[width=0.9\linewidth]{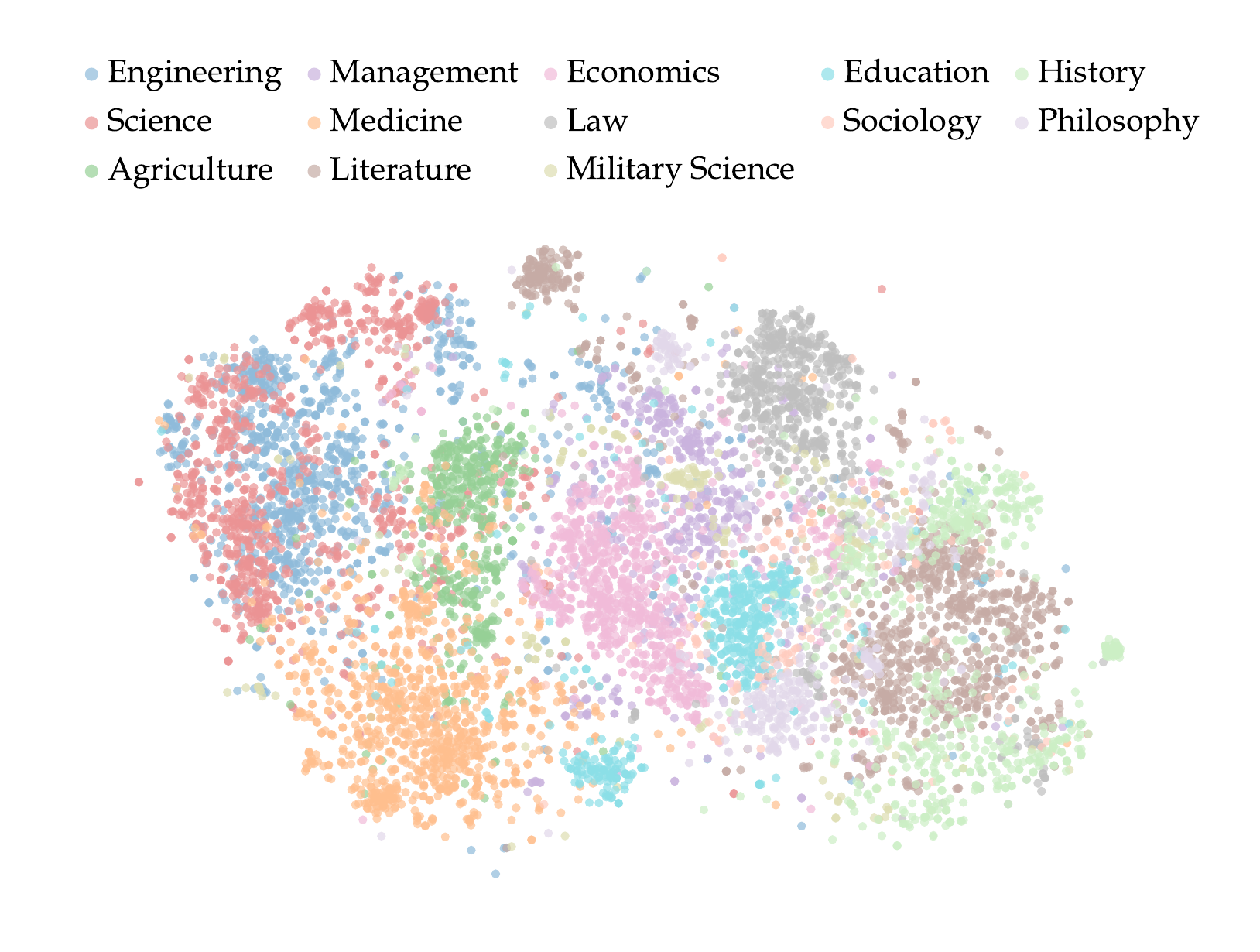}
\end{center}
\caption{The Visualization of the Text Embeddings of Question-Answer Pairs Across Disciplines.}
\label{fig:tsne}
\end{figure*}
\begin{table}[ht]
\centering
\scriptsize
\resizebox{\textwidth}{!}{

\begin{tabular}{@{}l|ccccccccccccc@{}}
\toprule
Difficulty & Agro. & Econ. & Edu. & Eng. & Hist. & Law & Lit. \& Arts & Mgmt. & Med. & Military Sci. & Phil. & Sci. & Socio. \\
\midrule
Hard (\#N) & 7 & 47 & 1 & 2458 & 3 & 57 & 12 & 6 & 217 & 4 & 27 & 4210 & 1 \\
Hard (\%) & 1.4 & 5.4 & 0.2 & 31.1 & 0.4 & 8.7 & 0.7 & 1.2 & 7.9 & 2.0 & 7.8 & 42.8 & 0.7 \\
\midrule
Middle (\#N) & 219 & 565 & 179 & 3462 & 180 & 343 & 496 & 236 & 1629 & 78 & 183 & 4133 & 45 \\
Middle (\%) & 45.2 & 64.7 & 37.0 & 43.9 & 26.7 & 52.3 & 29.6 & 47.1 & 59.1 & 38.0 & 52.7 & 42.0 & 31.5 \\
\midrule
Easy (\#N) & 259 & 261 & 304 & 1972 & 491 & 256 & 1168 & 259 & 909 & 123 & 137 & 1495 & 97 \\
Easy (\%) & 53.4 & 29.9 & 62.8 & 25.0 & 72.8 & 39.0 & 69.7 & 51.7 & 33.0 & 60.0 & 39.5 & 15.2 & 67.8 \\
\midrule
Total & 485 & 873 & 484 & 7892 & 674 & 656 & 1676 & 501 & 2755 & 205 & 347 & 9838 & 143 \\
\bottomrule
\end{tabular}}
\caption{Distribution of Difficulty Levels Across Disciplines. ``\#N'' denotes the number of samples.
}
\label{tab:discipline_difficulty_distribution}
\end{table}

\begin{figure}[ht]
\centering
    \centering
    \includegraphics[width=0.7\columnwidth, trim={0 0 -5cm 0}, clip]{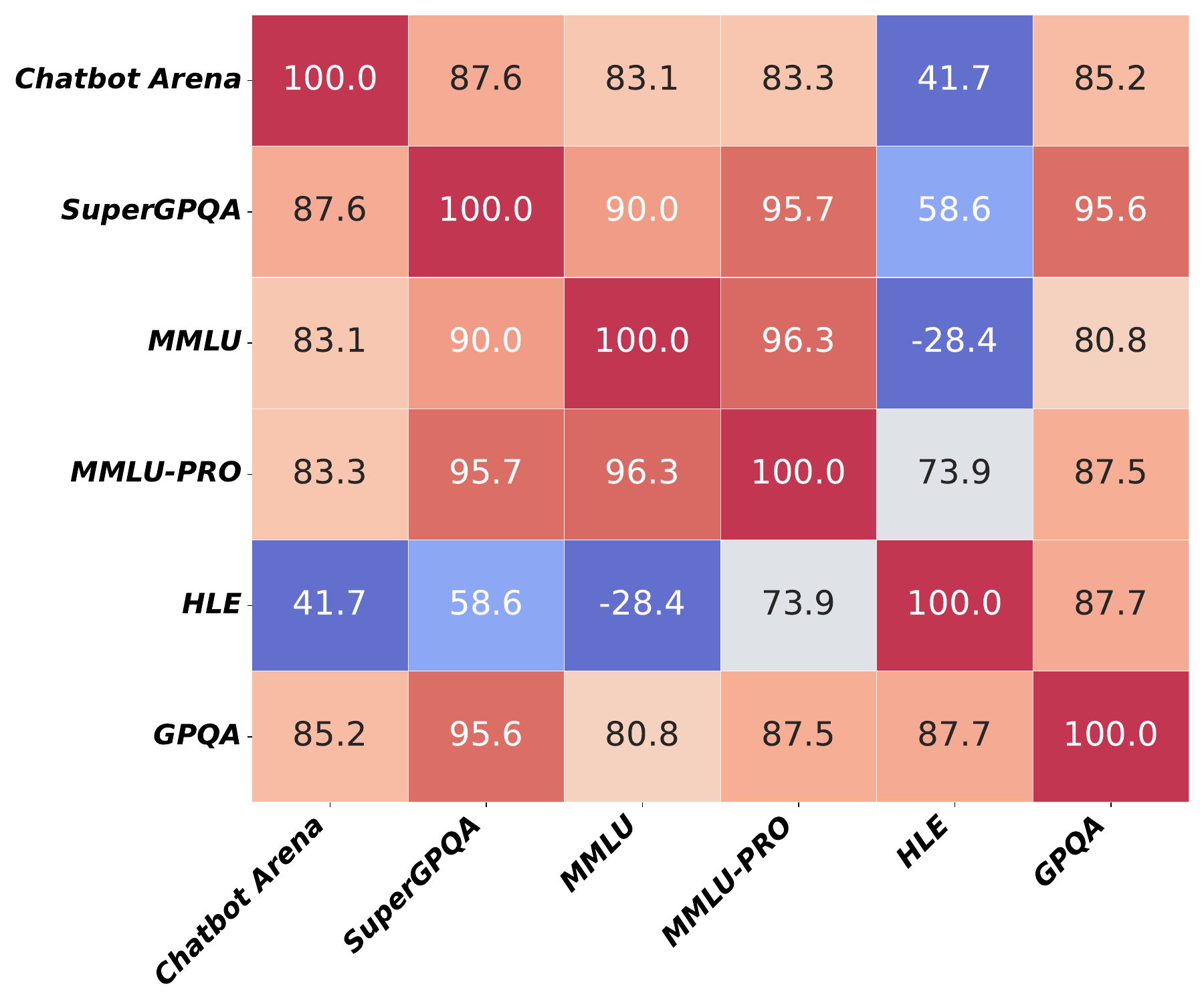}
    \caption{The Correlation Coefficients between Different Benchmarks. For MMLU, we record the results under the 5-shot setting.}
    \label{fig:correlation_benchmark}
\end{figure}

\paragraph{Sentence Length.} Question and answer length analysis reveals substantial variation across disciplines. The average question length is 58.42 tokens, with Literature questions showing the highest maximum length (869 tokens) and Engineering questions averaging 67.26 tokens. The ``answer'' options maintain a length pattern similar to the general options across different disciplines, averaging 12.86 tokens per option. Such a consistent length distribution across options aligns with one of the requirements of error option curation, which tries to confuse the models by providing confusing options in similar lengths.

\begin{figure}[ht]
    \centering
    \includegraphics[width=0.8\columnwidth, trim={0 0 -4cm 0}, clip]{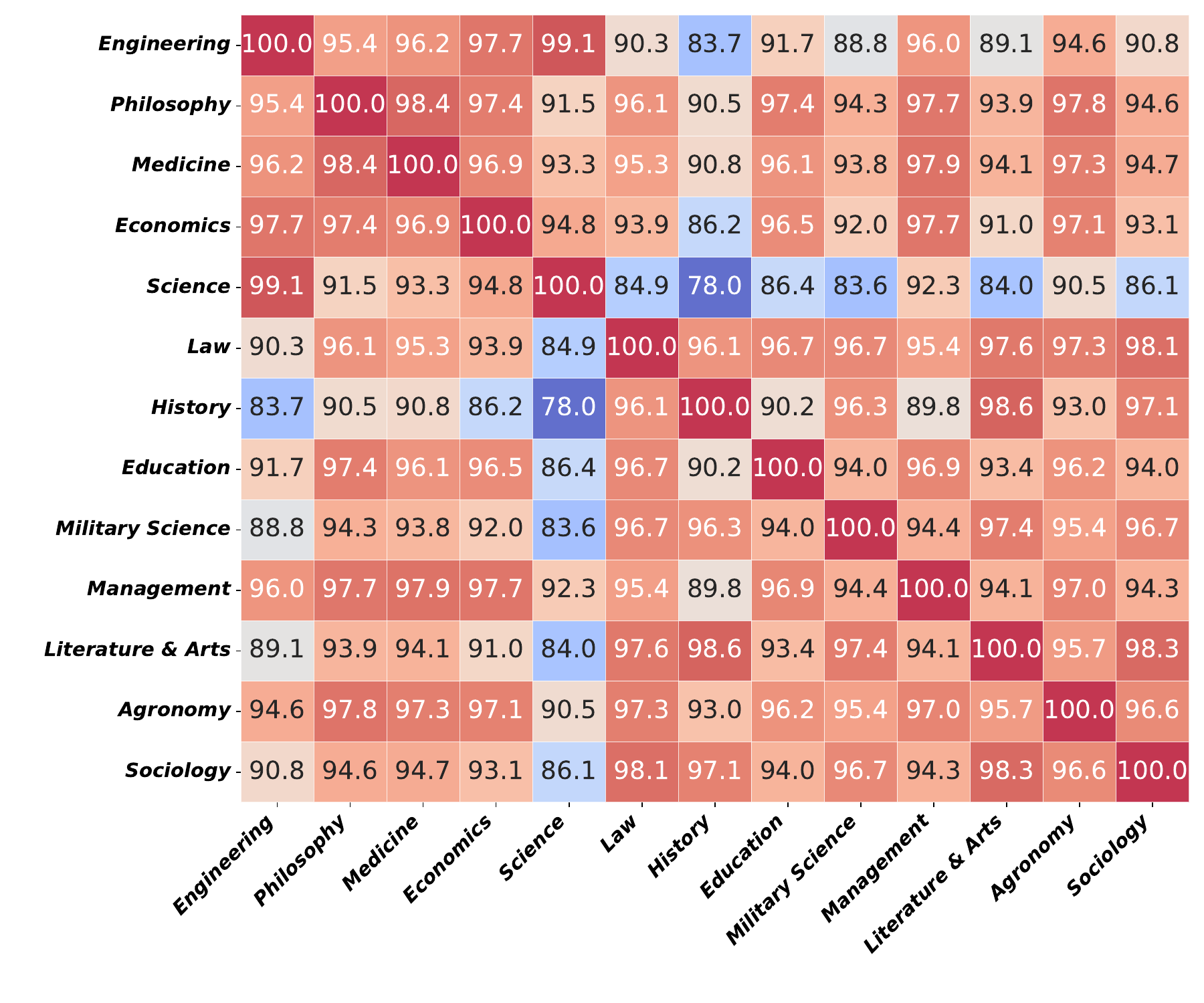}
    \caption{The Correlation Between Disciplines in \benchmark.}
    \label{fig:correlation_subclass}
\end{figure}

\paragraph{Semantic Visualization.} As shown in \autoref{fig:tsne}, we employ t-SNE visualization of question-answer pair embeddings to visualize the distribution \benchmark \footnote{We use the \texttt{gte-large-en-v1.5}~\citep{li2023towards,zhang2024mgte} encoding model and set the t-SNE parameters as: perplexity 100, learning rate 500, and 1000 iteration. For clearer visualization, we randomly select maximum 1000 samples from each disciplines.} and to further show the comprehensiveness of it.
The resulting visualization demonstrates clear clustering patterns across disciplines while maintaining substantial overlap in conceptual spaces.
The Engineering and Science demonstrate the highest degree of embedding overlap, suggesting strong semantic similarities in their Q\&A patterns. The humanities cluster (Literature, Philosophy, History) shows diffuse boundaries but maintains distinct centroids. The Military Science exhibits relatively isolated embedding patterns, indicating unique domain-specific language.
This analysis shows aligning conclusions from \autoref{fig:correlation_subclass} and reveals that the semantic structure of the \benchmark pairs reflects both the traditional organization of academic disciplines and their natural intellectual relationships, effectively capturing both domain-specific knowledge and cross-disciplinary connections.

\section{Experiments}

\subsection{Baseline Models}
We evaluate 6 reasoning models (o3-mini has three modes), 28 chat models and 17 base models on \benchmark, which includes closed-source models, open-source models, and fully open-source models. The reasoning models include Deepseek-R1 and Deepseek-R1-Zero~\citep{guo2025deepseek}, o1 and o1-mini~\citep{openai2024o1}, QwQ~\citep{qwq-32b-preview}, and o3-mini~\citep{openai2023o3mini} series models. The chat models include closed-source models such as Doubao-1.5-pro, Qwen-max, Claude-3.5~\citep{claude35addendum}, Gemini~\citep{geminiteam2024gemini15unlockingmultimodal}, GPT-4o~\citep{openai2024gpt4o}, Yi-Lightning~\citep{wake2024yi}, and open-source models like MiniMax-Text-01~\citep{li2025minimax}, Qwen2.5~\citep{yang2024qwen2} series, Llama-3.1~\citep{dubey2024llama} series, Mistral~\citep{jiang2023mistral} and Mixtral~\citep{jiang2024mixtral} series, Gemma-2 ~\citep{team2024gemma}series, Yi-1.5~\citep{Young2024YiOF} series, Phi4~\citep{phi-4}, and Granite-3.1~\citep{granite2024granite}. Additionally, there are fully open-source models like MAP-Neo~\citep{zhang2024mapneo} and OLMo-2~\citep{olmo20242}. According to our test results, these models still show a significant gap when compared to industry standards level. The base models include Qwen2.5~\citep{yang2024qwen2} series, Deepseek-V3~\citep{liu2024deepseek}, Yi-1.5~\citep{Young2024YiOF} series, Llama-3.1~\citep{dubey2024llama} series, Gemma-2~\citep{team2024gemma} series, and Mistral~\citep{jiang2023mistral} and Mixtral~\citep{jiang2024mixtral} series. Reasoning models and chat models are evaluated using a zero-shot approach, while base models are assessed using a five-shot evaluation.  Specifically, the five-shot evaluation for base models follow a similar methodology to MMLU-Pro. The specific prompts employed for both the zero-shot and five-shot evaluations are detailed in \autoref{appendix:Evaluation Prompt}. For all main results, the temperature is set to 0. The maximum number of new tokens is set to 32K for reasoning models, while for all other models, it is set to 4K. More model results can be found in \autoref{appendix: all results}, and all detailed results are provided in \autoref{appendix: model score}.

\subsection{Main Results}\label{sec:results}

We present the performances of the baselines on different levels, difficulties (\autoref{performance1}) and disciplines (\autoref{performance2}).
\definecolor{color11}{rgb}{1, 0.8, 0.8}  
\definecolor{color12}{rgb}{1, 0.9, 0.9}  
\definecolor{color21}{RGB}{255, 224, 127}  
\definecolor{color22}{RGB}{255, 239, 179}  
\definecolor{color31}{RGB}{198, 230, 195}  
\definecolor{color32}{RGB}{224, 239, 225}  

{
\linespread{1.1}
\begin{table}[p]
\small
\centering
\setlength{\tabcolsep}{3pt} 
\resizebox{\textwidth}{!}{%
\setlength{\tabcolsep}{6.6pt}
\begin{tabular}{p{4.5cm}<{\raggedright\arraybackslash}*{7}{p{1.6cm}<{\centering\arraybackslash}}}
\toprule
\textbf{Model} & \textbf{Overall} & \textbf{Overall} & \textbf{Overall} & \textbf{Overall} & \textbf{Easy} & \textbf{Middle} & \textbf{Hard}\\
& \textbf{(sample)} & \textbf{(subfield)} & \textbf{(field)} & \textbf{(discipline)} & \textbf{(sample)} & \textbf{(sample)} & \textbf{(sample)}\\
\midrule
\rowcolor{color11}
\multicolumn{8}{c}{\textbf{\textit{Reasoning Models}}}\\
\midrule
\rowcolor{color12}
DeepSeek-R1 &\boxed{61.82} & \boxed{62.61} & \boxed{61.23} & \textbf{59.95} & \underline{63.59} & \boxed{63.63} &\boxed{56.87} \\
\rowcolor{color12}
o1-2024-12-17 &\textbf{60.24} & \underline{61.25} & \underline{59.94} & \underline{59.44} & \textbf{64.40} & \underline{61.44} &\underline{53.67} \\
\rowcolor{color12}
DeepSeek-R1-Zero &\textbf{60.24} & \textbf{61.62} & \textbf{60.95} & \boxed{60.99} & \boxed{65.06} & \textbf{62.61} &50.99 \\
\rowcolor{color12}
o3-mini-2025-01-31-high &\underline{55.22} & 54.94 & 52.11 & 48.32 & 53.05 & 56.09 &\textbf{56.16} \\
\rowcolor{color12}
o3-mini-2025-01-31-medium &52.69 & 52.66 & 49.95 & 46.07 & 51.30 & 53.79 &52.37 \\
\rowcolor{color12}
o3-mini-2025-01-31-low &48.03 & 48.51 & 45.89 & 42.63 & 48.80 & 50.21 &43.53 \\
\rowcolor{color12}
o1-mini-2024-09-12 &45.22 & 45.46 & 42.53 & 39.33 & 46.77 & 47.34 &40.00 \\
\rowcolor{color12}
QwQ &43.59 & 44.40 & 43.19 & 41.63 & 46.46 & 47.40 &34.07 \\
\midrule
\rowcolor{color21}
\multicolumn{8}{c}{\textbf{\textit{Chat Models}}}\\
\midrule
\rowcolor{color22}
Doubao-1.5-pro-32k-250115 &\boxed{55.09} & \boxed{56.55} & \boxed{55.62} & \boxed{54.39} & \underline{57.70} & \boxed{60.15} &\boxed{43.80} \\
\rowcolor{color22}
Doubao-1.5-pro-32k-241225 &\textbf{50.93} & \underline{52.41} & \underline{51.76} & 51.24 & 53.54 & \textbf{56.56} &\underline{38.70} \\
\rowcolor{color22}
Qwen-max-2025-01-25 &\underline{50.08} & \textbf{52.75} & \textbf{52.47} & \underline{51.65} & \textbf{58.16} & \underline{54.95} &33.09 \\
\rowcolor{color22}
Claude-3-5-sonnet-20241022 &48.16 & 51.38 & 51.23 & \textbf{53.15} & \boxed{59.04} & 51.91 &29.99 \\
\rowcolor{color22}
Gemini-2.0-flash &47.73 & 48.70 & 47.80 & 46.10 & 53.06 & 49.56 &\textbf{38.84} \\
\rowcolor{color22}
DeepSeek-V3 &47.40 & 49.10 & 48.31 & 47.35 & 55.63 & 50.11 &33.86 \\
\rowcolor{color22}
MiniMax-Text-01 &45.11 & 47.46 & 46.97 & 47.06 & 54.51 & 48.60 &28.98 \\
\rowcolor{color22}
GPT-4o-2024-11-20 &44.40 & 47.62 & 47.50 & 48.84 & 56.84 & 48.75 &23.50 \\
\rowcolor{color22}
Llama-3.1-405B-Instruct &43.14 & 46.43 & 45.83 & 47.35 & 56.06 & 46.31 &23.70 \\
\rowcolor{color22}
GPT-4o-2024-08-06 &41.64 & 44.79 & 44.91 & 46.29 & 55.22 & 45.11 &20.98 \\
\rowcolor{color22}
Qwen2.5-72B-Instruct &40.75 & 43.66 & 43.32 & 42.10 & 48.84 & 45.42 &24.10 \\
\rowcolor{color22}
Mistral-Large-Instruct-2411 &40.65 & 43.38 & 43.13 & 43.37 & 52.92 & 43.28 &22.81 \\
\rowcolor{color22}
Qwen-max-2024-09-19 &39.96 & 42.93 & 42.16 & 41.62 & 50.23 & 43.63 &22.60 \\
\rowcolor{color22}
Qwen2.5-32B-Instruct &38.76 & 41.18 & 40.40 & 39.43 & 47.42 & 43.05 &22.13 \\
\rowcolor{color22}
Llama-3.3-70B-Instruct &37.69 & 40.56 & 40.15 & 41.12 & 49.68 & 40.68 &19.55 \\
\rowcolor{color22}
Phi-4 &37.65 & 39.59 & 38.61 & 37.66 & 45.43 & 40.91 &23.69 \\
\rowcolor{color22}
Qwen2.5-14B-Instruct &35.15 & 37.72 & 37.41 & 36.07 & 44.82 & 37.90 &19.97 \\
\rowcolor{color22}
Llama-3.1-70B-Instruct &34.86 & 38.94 & 39.18 & 40.57 & 48.22 & 37.85 &15.22 \\
\rowcolor{color22}
Yi-Lightning &33.42 & 36.57 & 36.45 & 36.92 & 43.38 & 35.32 &19.35 \\
\rowcolor{color22}
Mixtral-8x22B-Instruct-v0.1 &29.23 & 32.14 & 32.28 & 32.82 & 42.52 & 29.73 &13.82 \\
\rowcolor{color22}
Qwen2.5-7B-Instruct &28.78 & 30.78 & 30.37 & 30.63 & 37.77 & 30.98 &15.23 \\
\rowcolor{color22}
Gemma-2-27B-it &27.43 & 30.50 & 30.42 & 31.30 & 40.90 & 27.45 &12.64 \\
\rowcolor{color22}
Qwen2.5-3B-Instruct &23.31 & 25.45 & 25.86 & 25.57 & 33.10 & 23.50 &12.24 \\
\rowcolor{color22}
Granite-3.1-8B-instruct &20.83 & 22.85 & 22.92 & 22.26 & 29.48 & 19.79 &13.09 \\
\rowcolor{color22}
Qwen2.5-1.5B-Instruct &18.82 & 20.91 & 20.75 & 22.11 & 27.41 & 18.19 &10.45 \\
\rowcolor{color22}
OLMo-2-1124-13B-Instruct &18.66 & 20.46 & 20.60 & 21.80 & 27.10 & 17.85 &10.74 \\
\rowcolor{color22}
MAP-Neo-7B-Instruct-v0.1 &17.05 & 18.52 & 18.42 & 18.70 & 23.26 & 16.62 &10.95 \\
\rowcolor{color22}
OLMo-2-1124-7B-Instruct &16.81 & 18.08 & 18.57 & 18.85 & 22.80 & 15.82 &11.90 \\
\midrule
\rowcolor{color31}
\multicolumn{8}{c}{\textbf{\textit{Base Models}}}\\
\midrule
\rowcolor{color32}
Qwen2.5-72B & \boxed{34.33} & \boxed{38.08} & \boxed{38.70} & \boxed{39.54} & \boxed{46.20} & \boxed{38.12} &\textbf{15.01} \\
\rowcolor{color32}
Qwen2.5-32B & \textbf{33.16} & \textbf{36.52} & \textbf{37.33} & \textbf{38.29} & \textbf{45.12} & \textbf{36.58} &14.34 \\
\rowcolor{color32}
DeepSeek-V3-Base & \underline{32.14} & \underline{34.79} & \underline{34.58} & \underline{34.71} & 41.28 & \underline{34.50} &\boxed{18.20} \\
\rowcolor{color32}
Qwen2.5-14B & 30.19 & 33.33 & 34.14 & 34.54 & \underline{42.27} & 31.44 &\underline{14.85} \\
\rowcolor{color32}
Yi-1.5-34B & 27.62 & 30.78 & 31.03 & 32.55 & 39.68 & 27.95 &13.86 \\
\rowcolor{color32}
Llama-3.1-70B & 27.22 & 30.52 & 31.28 & 32.55 & 40.78 & 26.95 &12.78 \\
\rowcolor{color32}
Qwen2.5-7B & 25.36 & 28.19 & 28.73 & 29.60 & 36.58 & 25.94 &12.10 \\
\rowcolor{color32}
Llama-3.1-405B & 25.23 & 28.09 & 28.33 & 30.15 & 37.58 & 25.12 &11.86 \\
\rowcolor{color32}
Gemma-2-27B & 24.49 & 27.35 & 27.96 & 28.58 & 36.26 & 24.07 &12.27 \\
\rowcolor{color32}
Mixtral-8x22B-v0.1 & 22.41 & 24.71 & 25.04 & 25.02 & 32.78 & 21.67 &12.26 \\
\rowcolor{color32}
\rowcolor{color32}
Qwen2.5-3B & 20.14 & 22.81 & 23.30 & 24.42 & 30.42 & 19.81 &9.40 \\
\rowcolor{color32}
Mistral-7B-v0.3 & 19.48 & 21.50 & 21.81 & 22.27 & 27.62 & 18.65 &11.96 \\
\rowcolor{color32}
Qwen2.5-1.5B & 17.17 & 19.31 & 19.80 & 21.35 & 24.52 & 16.79 &9.74 \\
\rowcolor{color32}
OLMo-2-1124-13B & 16.07 & 18.75 & 19.82 & 21.37 & 27.24 & 14.41 &6.57 \\
\rowcolor{color32}
MAP-Neo-7B & 15.76 & 17.48 & 18.26 & 19.54 & 22.86 & 14.64 &9.83 \\
\rowcolor{color32}
Granite-3.1-8B-Base & 15.69 & 16.98 & 16.79 & 16.65 & 20.40 & 15.65 &10.60 \\
\rowcolor{color32}
OLMo-2-1124-7B & 15.15 & 17.62 & 18.30 & 19.60 & 24.43 & 13.83 &7.15 \\

\bottomrule
\end{tabular}
}
\captionsetup{font=footnotesize}
\caption{\textbf{Performance on \benchmark.} 
LLMs are scored sample-wise, subfield-wise, field-wise, and discipline-wise levels to ensure fair assessment despite imbalanced question distributions. The highest, the second-best and the third-best scores are shown in  \boxed{box}, \textbf{bold} and \underline{underlined}, respectively.}
\label{performance1}
\end{table}
}

\definecolor{color11}{rgb}{1, 0.8, 0.8}  
\definecolor{color12}{rgb}{1, 0.9, 0.9}  
\definecolor{color21}{RGB}{255, 224, 127}  
\definecolor{color22}{RGB}{255, 239, 179}  
\definecolor{color31}{RGB}{198, 230, 195}  
\definecolor{color32}{RGB}{224, 239, 225}  

{
\linespread{1.1}
\begin{table}[p]
\small
\centering
\setlength{\tabcolsep}{2pt} 

\resizebox{\textwidth}{!}{%
\begin{tabular}{lccccccccccccc}
\toprule
\textbf{Model} & \textbf{Agr.} & \textbf{Econ.} & \textbf{Edu.} & \textbf{Eng.} & \textbf{Hist.} & \textbf{Law} & \textbf{Lit \& Arts} & \textbf{Mgt.} & \textbf{Med.} & \textbf{Mil Sci.} & \textbf{Phil.} & \textbf{Sci.} & \textbf{Soc.} \\
\midrule
\rowcolor{color11}
\multicolumn{14}{c}{\textbf{\textit{Reasoning Models}}}\\
\midrule
\rowcolor{color12}
DeepSeek-R1& \boxed{54.43}& \textbf{66.09}& \textbf{54.75}& \boxed{63.10}& \underline{55.19}& \textbf{65.24}& \underline{52.45}& \underline{57.09}& \underline{59.93}& \underline{57.07}& \textbf{63.11}& \boxed{63.69}& \boxed{67.13} \\
\rowcolor{color12}
DeepSeek-R1-Zero& \textbf{53.81}& \boxed{66.44}& \boxed{60.54}& \textbf{60.28}& \textbf{58.61}& \boxed{66.77}& \boxed{56.86}& \boxed{59.68}& \textbf{60.65}& \textbf{58.54}& \boxed{63.69}& 59.93& \boxed{67.13} \\
\rowcolor{color12}
o1-2024-12-17& \underline{50.93}& \underline{61.17}& \underline{53.31}& \underline{59.17}& \boxed{60.98}& \underline{63.87}& \textbf{55.79}& \textbf{57.29}& \boxed{62.25}& \boxed{60.49}& \underline{61.38}& \textbf{61.76}& \textbf{64.34} \\
\rowcolor{color12}
o3-mini-2025-01-31-high& 41.03& 54.07& 46.28& 56.41& 36.80& 45.73& 39.44& 51.50& 51.72& 46.34& 51.01& \underline{61.72}& \underline{46.15} \\
\rowcolor{color12}
o3-mini-2025-01-31-medium& 40.62& 51.32& 41.74& 53.83& 35.01& 43.60& 37.11& 48.50& 50.34& 45.85& 48.13& 58.79& 44.06 \\
\rowcolor{color12}
o3-mini-2025-01-31-low& 37.32& 45.25& 38.43& 48.20& 32.94& 39.79& 36.22& 43.71& 48.09& 41.46& 44.09& 53.24& 45.45 \\
\rowcolor{color12}
o1-mini-2024-09-12& 34.02& 45.02& 36.16& 45.41& 26.26& 35.98& 32.22& 40.52& 44.32& 39.51& 39.48& 51.09& 41.26 \\
\rowcolor{color12}
QwQ& 38.14& 47.77& 45.25& 43.37& 29.53& 45.88& 32.82& 41.32& 43.88& 40.00& 43.52& 46.33& 43.36 \\
\midrule
\rowcolor{color21}
\multicolumn{14}{c}{\textbf{\textit{Chat Models}}}\\
\midrule
\rowcolor{color22}
Doubao-1.5-pro-32k-250115& \boxed{50.93}& \boxed{65.06}& \textbf{55.58}& \boxed{55.60}& 42.88& \textbf{58.84}& 42.30& \textbf{53.29}& \boxed{59.13}& \textbf{54.15}& \boxed{61.96}& \boxed{55.54}& 51.75 \\
\rowcolor{color22}
Doubao-1.5-pro-32k-241225& \textbf{47.42}& \textbf{60.14}& \underline{54.75}& \underline{50.76}& 37.54& 54.12& 38.31& \underline{52.69}& \underline{54.99}& \underline{53.66}& \textbf{57.06}& \textbf{51.57}& 53.15 \\
\rowcolor{color22}
claude-3-5-sonnet-20241022& \underline{47.01}& 56.59& 53.72& 47.57& \boxed{53.56}& \boxed{60.21}& \boxed{50.42}& 51.30& 49.26& \boxed{59.51}& 52.45& 45.02& \boxed{64.34} \\
\rowcolor{color22}
qwen-max-2025-01-25& 44.33& \underline{57.50}& \boxed{56.40}& \textbf{50.81}& 44.81& 54.12& 44.93& \boxed{54.69}& \textbf{56.37}& 49.27& \underline{56.20}& 47.51& \underline{54.55} \\
\rowcolor{color22}
Llama-3.1-405B-Instruct& 43.09& 49.71& 45.45& 41.89& 50.00& \underline{55.34}& 43.20& 47.70& 49.04& 52.20& 48.41& 39.80& 49.65 \\
\rowcolor{color22}
gpt-4o-2024-11-20& 42.27& 46.74& 50.00& 42.83& \textbf{52.52}& 53.81& \textbf{46.72}& 47.31& 52.52& 52.20& 52.74& 40.67& \underline{54.55} \\
\rowcolor{color22}
gpt-4o-2024-05-13& 41.44& 40.78& 44.21& 37.47& 50.74& 52.90& 45.11& 41.32& 48.82& 50.73& 45.24& 35.41& 53.85 \\
\rowcolor{color22}
DeepSeek-V3& 41.24& 49.48& 43.80& 47.21& 47.18& 51.07& 45.23& 48.50& 46.10& 48.29& 43.23& 48.30& \textbf{55.94} \\
\rowcolor{color22}
Qwen2.5-72B-Instruct& 41.24& 46.62& 44.42& 41.12& 30.71& 45.88& 34.90& 42.32& 45.74& 45.37& 43.52& 39.33& 46.15 \\
\rowcolor{color22}
MiniMax-Text-01& 39.79& 49.60& 47.52& 44.88& 45.25& 54.27& 43.02& 48.90& 47.08& 48.29& 45.53& 43.81& 53.85 \\
\rowcolor{color22}
gpt-4o-2024-08-06& 39.59& 40.78& 44.42& 38.84& \underline{50.89}& 54.12& \underline{46.30}& 45.51& 50.49& 50.73& 47.84& 38.41& 53.85 \\
\rowcolor{color22}
Mistral-Large-Instruct-2411& 38.76& 42.27& 42.15& 39.66& 44.96& 47.87& 41.11& 43.31& 43.23& 48.78& 42.07& 39.24& 50.35 \\
\rowcolor{color22}
gemini-2.0-flash& 38.56& 45.93& 42.36& 48.37& 45.25& 49.24& 43.68& 41.32& 43.77& 51.22& 45.53& \underline{50.20}& 53.85 \\
\rowcolor{color22}
Llama-3.1-70B-Instruct& 37.11& 41.12& 41.94& 33.88& 33.09& 45.88& 35.08& 46.11& 44.21& 47.80& 42.94& 30.04& 48.25 \\
\rowcolor{color22}
Llama-3.3-70B-Instruct& 36.70& 40.55& 40.29& 36.44& 41.25& 44.66& 38.54& 45.51& 43.77& 46.34& 42.94& 34.96& 42.66 \\
\rowcolor{color22}
qwen-max-2024-09-19& 36.49& 45.02& 45.66& 39.84& 35.16& 44.05& 39.08& 45.11& 43.41& 47.32& 43.23& 38.24& 38.46 \\
\rowcolor{color22}
Qwen2.5-32B-Instruct& 36.49& 43.07& 45.45& 38.93& 30.71& 41.92& 32.76& 39.52& 42.21& 44.88& 39.19& 38.24& 39.16 \\
\rowcolor{color22}
Qwen2.5-14B-Instruct& 36.08& 37.69& 39.26& 35.87& 26.41& 37.04& 31.44& 41.52& 36.91& 38.05& 38.04& 34.20& 36.36 \\
\rowcolor{color22}
Yi-Lighting& 33.81& 39.18& 35.95& 32.53& 36.05& 37.35& 36.34& 38.12& 36.95& 42.44& 37.75& 30.85& 42.66 \\
\rowcolor{color22}
Phi-4& 32.78& 40.21& 39.26& 37.27& 30.27& 41.46& 31.62& 39.12& 37.79& 39.02& 40.63& 38.87& 41.26 \\
\rowcolor{color22}
Gemma-2-27B-it& 31.13& 29.32& 30.99& 26.72& 24.78& 35.06& 29.42& 34.93& 31.00& 39.02& 35.45& 24.81& 34.27 \\
\rowcolor{color22}
Qwen2.5-7B-Instruct& 29.28& 34.59& 35.33& 28.38& 22.85& 32.62& 24.28& 33.33& 32.60& 32.68& 34.58& 27.54& 30.07 \\
\rowcolor{color22}
Mixtral-8x22B-Instruct-v0.1& 28.25& 33.45& 33.68& 29.02& 30.86& 34.60& 30.55& 35.13& 30.53& 44.39& 32.85& 26.95& 36.36 \\
\rowcolor{color22}
Qwen2.5-3B-Instruct& 25.36& 26.35& 30.99& 22.83& 18.55& 25.61& 22.32& 30.94& 26.82& 27.32& 27.09& 21.64& 26.57 \\
\rowcolor{color22}
Granite-3.1-8B-instruct& 24.74& 20.16& 23.76& 21.60& 16.32& 24.70& 20.53& 25.15& 20.58& 27.80& 21.33& 19.70& 23.08 \\
\rowcolor{color22}
Qwen2.5-1.5B-Instruct& 22.27& 22.57& 27.27& 17.64& 16.47& 26.37& 17.84& 24.95& 22.58& 25.37& 27.67& 16.85& 19.58 \\
\rowcolor{color22}
OLMo-2-1124-13B-Instruct& 20.82& 22.68& 25.21& 18.18& 16.91& 22.10& 19.57& 21.96& 21.81& 25.85& 27.38& 16.39& 24.48 \\
\rowcolor{color22}
Yi-1.5-6B-Chat& 20.41& 23.83& 25.62& 18.50& 13.95& 25.61& 18.79& 24.15& 21.89& 29.27& 25.36& 17.60& 23.78 \\
\rowcolor{color22}
MAP-Neo-7B-Instruct-v0.1& 17.94& 17.64& 21.69& 16.79& 13.50& 21.19& 16.35& 20.96& 18.91& 25.85& 21.61& 15.99& 14.69 \\
\rowcolor{color22}
Qwen2.5-0.5B-Instruct& 16.91& 16.49& 17.98& 9.77& 10.09& 13.87& 12.89& 15.97& 14.12& 13.17& 12.68& 8.55& 12.59 \\
\midrule
\rowcolor{color31}
\multicolumn{14}{c}{\textbf{\textit{Base Models}}}\\
\midrule
\rowcolor{color32}
Qwen2.5-32B& \boxed{38.76}& \textbf{40.89}& \boxed{45.87}& \textbf{32.70}& 28.49& \textbf{39.94}& 30.91& \textbf{40.72}& \textbf{38.84}& \boxed{47.80}& \boxed{43.52}& \textbf{29.45}& \boxed{39.86} \\
\rowcolor{color32}
Qwen2.5-72B& \textbf{36.29}& \boxed{44.33}& \textbf{45.25}& \boxed{33.39}& \underline{31.31}& \boxed{44.21}& \boxed{35.50}& \boxed{43.51}& \boxed{43.12}& \textbf{46.34}& \textbf{42.94}& \underline{29.38}& \textbf{38.46} \\
\rowcolor{color32}
Qwen2.5-14B& \underline{34.02}& 34.82& \underline{40.08}& 30.26& 28.49& 36.89& 28.94& \underline{40.32}& 34.37& 40.00& 37.75& 26.68& 36.36 \\
\rowcolor{color32}
Llama-3.1-405B& 29.28& 27.95& 31.20& 23.61& \boxed{34.12}& 34.30& 30.79& 34.13& 29.47& 39.51& 31.70& 21.47& 24.48 \\
\rowcolor{color32}
Llama-3.1-70B& 28.87& 29.67& 36.36& 26.19& \textbf{32.94}& 33.38& \underline{32.64}& 36.73& 30.96& \underline{40.98}& 36.89& 23.30& 34.27 \\
\rowcolor{color32}
Gemma-2-27B& 28.87& 27.26& 30.79& 24.09& 22.85& 28.96& 27.51& 32.73& 27.80& 37.56& 31.70& 21.38& 30.07 \\
\rowcolor{color32}
DeepSeek-V3-Base& 28.66& 35.17& 39.05& \underline{31.87}& 30.12& \underline{37.20}& \textbf{33.65}& 37.33& \underline{34.70}& 37.07& \underline{38.62}& \boxed{30.08}& \underline{37.76} \\
\rowcolor{color32}
Qwen2.5-7B& 28.25& 31.96& 38.02& 24.89& 22.55& 30.34& 25.06& 28.54& 30.56& 31.71& 36.60& 22.04& 34.27 \\
\rowcolor{color32}
Yi-1.5-34B& 28.25& \underline{35.85}& 37.60& 27.32& 24.63& 34.45& 30.25& 36.13& 31.18& 40.49& 35.45& 23.80& \underline{37.76} \\
\rowcolor{color32}
Mixtral-8x7B-v0.1& 28.04& 24.97& 29.34& 20.67& 23.89& 30.79& 27.27& 30.74& 25.05& 37.56& 30.84& 17.87& 28.67 \\
\rowcolor{color32}
Gemma-2-9B& 27.22& 26.23& 30.99& 21.79& 21.81& 26.83& 23.81& 28.94& 26.06& 32.68& 27.09& 20.01& 27.97 \\
\rowcolor{color32}
Mistral-7B-v0.3& 25.98& 20.96& 25.62& 19.02& 14.84& 24.24& 20.11& 23.55& 22.61& 30.24& 25.94& 17.47& 18.88 \\
\rowcolor{color32}
Mixtral-8x22B-v0.1& 23.92& 23.94& 27.27& 22.02& 20.33& 27.59& 24.64& 30.14& 22.50& 28.78& 24.50& 20.96& 28.67 \\
\rowcolor{color32}
Qwen2.5-3B& 22.47& 23.83& 28.31& 19.74& 18.40& 26.37& 21.60& 28.14& 25.41& 30.73& 29.39& 16.54& 26.57 \\
\rowcolor{color32}
MAP-Neo-7B& 20.21& 18.10& 23.35& 15.43& 15.28& 21.34& 18.79& 21.76& 17.10& 24.88& 22.19& 13.16& 22.38 \\
\rowcolor{color32}
OLMo-2-1124-13B& 19.59& 20.27& 26.24& 15.21& 19.73& 22.41& 20.88& 26.15& 19.96& 25.85& 26.51& 11.93& 23.08 \\
\rowcolor{color32}
Qwen2.5-1.5B& 19.18& 20.73& 28.10& 16.50& 15.43& 21.95& 17.72& 23.35& 20.87& 24.88& 25.65& 14.48& 28.67 \\
\rowcolor{color32}
Granite-3.1-8B-Base& 15.46& 17.87& 15.08& 16.26& 11.72& 18.29& 13.84& 14.57& 14.81& 20.49& 20.17& 15.44& 22.38 \\
\rowcolor{color32}
Qwen2.5-0.5B& 15.05& 12.37& 16.32& 10.30& 11.42& 12.96& 14.92& 11.78& 12.23& 16.59& 15.56& 8.74& 13.99 \\

\bottomrule
\end{tabular}
}
\captionsetup{font=footnotesize}
\caption{\textbf{Performance on \benchmark.} 
We present LLMs' performance on different disciplines. The highest, the second-best and the third-best scores are shown in  \boxed{box}, \textbf{bold} and \underline{underlined}, respectively.}
\label{performance2}
\end{table}
}

\paragraph{Overall Results.} In general, the top-performed reasoning models (e.g., DeepSeek-R1, o1-2024-12-17) achieve the best overall performance in \benchmark. The effectiveness of instruction tuning to improve the performances is once again verified in the benchmark. For instance, the results (47.40, 40.75) of DeepSeek-V3 and Qwen2.5-72B-Instruct are significantly better than the results (32.14, 34.33) of DeepSeek-V3-Base and Qwen2.5-72B, respectively.
More powerful LLMs achieve more balanced results on different difficulties. For example, the results of simple, middle and hard splits of DeepSeek-R1 are 63.59, 63.63 and 56.87. In contrast, the results of simple, middle and hard splits of Qwen2.5-14B-Instruct are 44.82, 37.90 and 19.97.

\paragraph{Observations for Reasoning Models.} 
Surprisingly, the performance gap between DeepSeek-R1 and DeepSeek-R1-Zero is relatively small. In \autoref{performance2}, the DeepSeek-R1 only beats DeepSeek-R1-Zero in two disciplines (i.e., Science and Engineering). In the other disciplines, the DeepSeek-R1-Zero is slightly better than DeepSeek-R1, which leaves the optimal training paradigm of the reason models an open question.
From the performances across different dimensions, compared to the o1-2024-12-17 model, the newer o1-mini and o3-mini models show decreasing scores except in science and engineering, suggesting a potential data leakage.
Moreover, for different versions of o3-mini (o3-mini-2025-01-31-high, o3-mini-2025-01-31-medium, o3-mini-2025-01-31-low), we observe obvious performance gaps for different difficulties.

\paragraph{Advantages from Pre-training Corpus.}
The newer versions of proprietary LLMs achieve significant improvements on \benchmark, considering the incremental growing of qwen-max series (2024-09-19$\rightarrow$2025-01-25: 39.96 $\rightarrow$50.08) and GPT-4o series (2024-05-13 $\rightarrow$ 2024-08-06 $\rightarrow$ 2024-11-20: 39.76 $\rightarrow$ 41.64 $\rightarrow$ 44.40). Such incremental performances following the chronological order suggest that the developers of proprietary models highly value the incorporation of long-tailed knowledge.
Moreover, we conjecture that the LLMs from Chinese firms (e.g., Qwen and Doubao) generally show superior performances is partially because their data collection pipelines are more aligned to \benchmark, i.e., a considerable ratio of the references are translated from Chinese textbooks.

\paragraph{Progress of Open-source.} 
The \benchmark also reveals a pessimistic open-source progress from the research community.
In the dimension of pre-training corpus curation, the fully open-sourced LLMs (e.g., MAP-Neo-7B and OLMo-2-1124-13B) perform similarly and lag behind to other non fully open ones in similar sizes (e.g., Qwen2.5-7B). 
It can also be observed that, compared to the proprietary models, most of the open-weight LLMs except for the DeepSeek-R1 series are not satisfactory in our benchmark, especially the hard questions.

\paragraph{Difficulty-Specific Capabilities.}  
The difficulty stratification in \benchmark reveals distinct capability patterns between reasoning-focused and knowledge-oriented LLMs. As shown in \autoref{performance1}, the \textbf{hard} split specifically challenges models' reasoning capacities, while \textbf{easy} and \textbf{middle} splits better reflect factual knowledge mastery. For instance, the o3-mini series exhibits lower scores than Doubao-1.5-pro-32k-250115 on easy and middle splits, yet surpasses it significantly on hard questions. This dichotomy suggests that:  
\begin{itemize}
    \item \textbf{Chat-oriented LLMs} (e.g., Doubao series) excel at knowledge recall for common professional questions but struggle with complex reasoning in long-tail domains.  
    \item \textbf{Reasoning-specialized models} demonstrate superior performance on hard questions through enhanced logical processing, despite potential compromises in broad knowledge coverage.  
\end{itemize}
This differentiation validates \benchmark's design rationale -- using difficulty levels as diagnostic tools to dissect complementary capabilities in modern LLMs.

\subsection{Further Analysis}


\begin{figure}[t]
    \centering
    \begin{subfigure}[b]{0.45\textwidth}
        \centering
        \includegraphics[width=\textwidth]{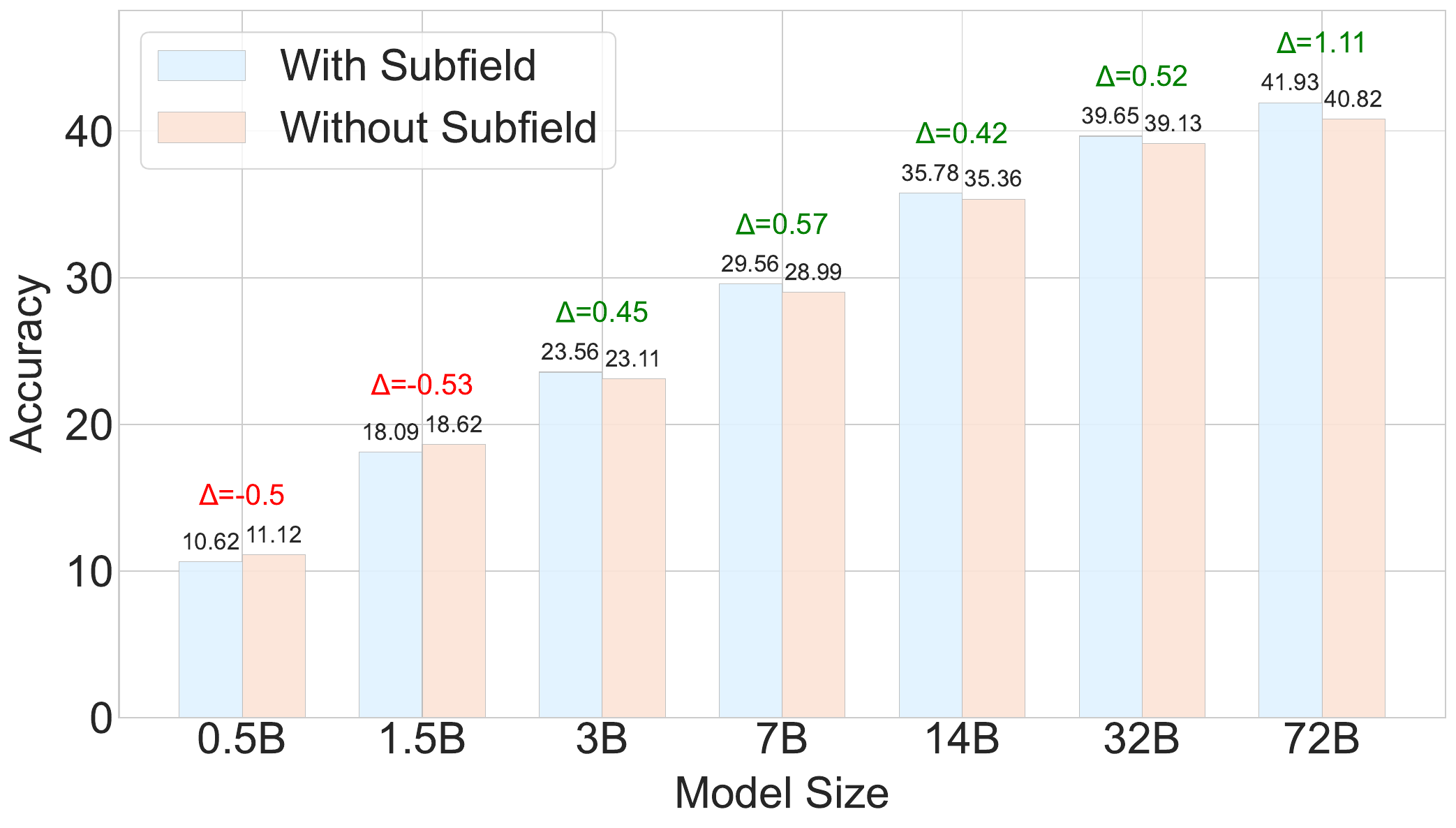}
        \caption{Impact of subfield information.}
        \label{fig:subfield}
    \end{subfigure}
    \hfill
    \begin{subfigure}[b]{0.45\textwidth}
        \centering
        \includegraphics[width=\textwidth]{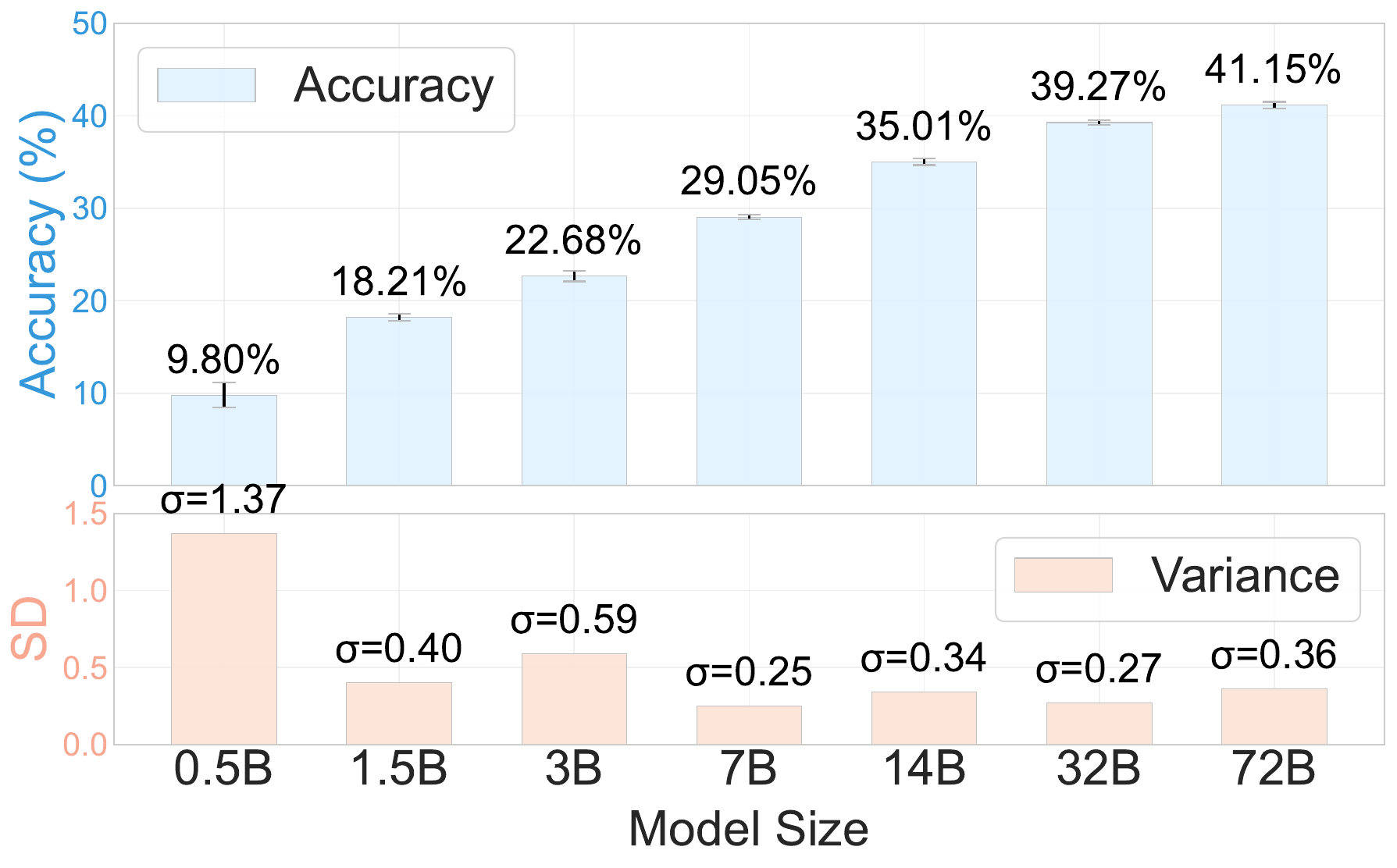}
        \caption{Model robustness analysis.}
        \label{fig:robustness}
    \end{subfigure}
    \caption{(a) Accuracy comparison of Qwen2.5 models (0.5B–72B) with and without subfield information in prompts. Larger models benefit more from additional context. 
    (b) Robustness evaluation across 24 semantically equivalent prompts, indicating larger models exhibit higher stability with lower variance.}
    \label{fig:subfield_robustness}
\end{figure}

\paragraph{Effect of Subfield Information in Prompts.}
To investigate the impact of subfield information on model performance, we conduct zero-shot evaluations under two conditions: (1) zero-shot-with-subfield, where the prompt includes a description of the problem's subfield, and (2) zero-shot-without-subfield, where no such information is provided. The prompts of zero-shot-with-subfield are shown in \autoref{appendix:Impact of Subfield Information}.
We evaluate Qwen2.5-Instruct models ranging from 0.5B to 72B parameters across these two settings.  
\textbf{\autoref{fig:subfield} show that incorporating subfield information generally leads to improved performance, particularly for larger models}. 
For example, Qwen2.5-72B-Instruct achieves an accuracy of 41.93\% in the zero-shot-with-subfield setting, compared to 40.82\% without subfield annotations. 
Similarly, Qwen2.5-32B-Instruct improves from 39.13\% to 39.65\%, and Qwen2.5-14B-Instruct sees a minor increase from 35.36\% to 35.78\%, \textbf{suggesting that additional contextual information helps larger models refine their reasoning by narrowing down the relevant domain}.  
However, for smaller models like Qwen2.5-0.5B-Instruct and Qwen2.5-1.5B-Instruct, the introduction of subfield information does not lead to a noticeable gain in accuracy. Qwen2.5-0.5B-Instruct performs slightly worse with subfield annotations (10.62\% vs. 11.12\%), while Qwen2.5-1.5B-Instruct shows near-identical performance (18.09\% vs. 18.62\%), \textbf{indicating that smaller models may lack the capacity to leverage fine-grained domain-specific cues effectively, relying more on general knowledge retrieval rather than contextual domain disambiguation}.

\paragraph{Robustness Analysis.}
Benchmark evaluations can be significantly influenced by slight variations in prompts, leading to inconsistencies in model ranking and overall assessment reliability. 
To investigate this phenomenon, we conduct robustness experimentacross Qwen2.5-Instruct models (0.5B$\sim$72B), employing 24 distinct yet semantically equivalent prompts in a zero-shot setting, using the same evaluation parameters as in the main results.
Specifically, we employ 4 types of initial prompts and 6 types of question formats, resulting in a combination of 24 different prompt styles to verify the robustness of our bench.
The combination of initial prompt 1 and question format 1 is the default prompt for our evaluation. The detailed prompts are shown in the \autoref{appendix:robustness}.
\textbf{\autoref{fig:robustness} shows that larger models exhibit increased robustness, as a steady improvement in accuracy from 9.80\% of Qwen2.5-0.5B-Instruct to 41.15\% of Qwen2.5-72B-Instruct while variance decreases ($\sigma = 1.37$ for the smallest model, dropping to $\sigma \approx 0.27 \sim 0.36$ in the largest models). }
This demonstrates the our evaluation framework is more stable with \textbf{a maximum standard deviation SD of 1.37\%}, that mitigates prompt-induced instability and provides a more reliable basis for model assessment.

\begin{figure}[t]
    \centering
    \begin{subfigure}[b]{0.45\textwidth}
        \centering
        \includegraphics[width=\textwidth]{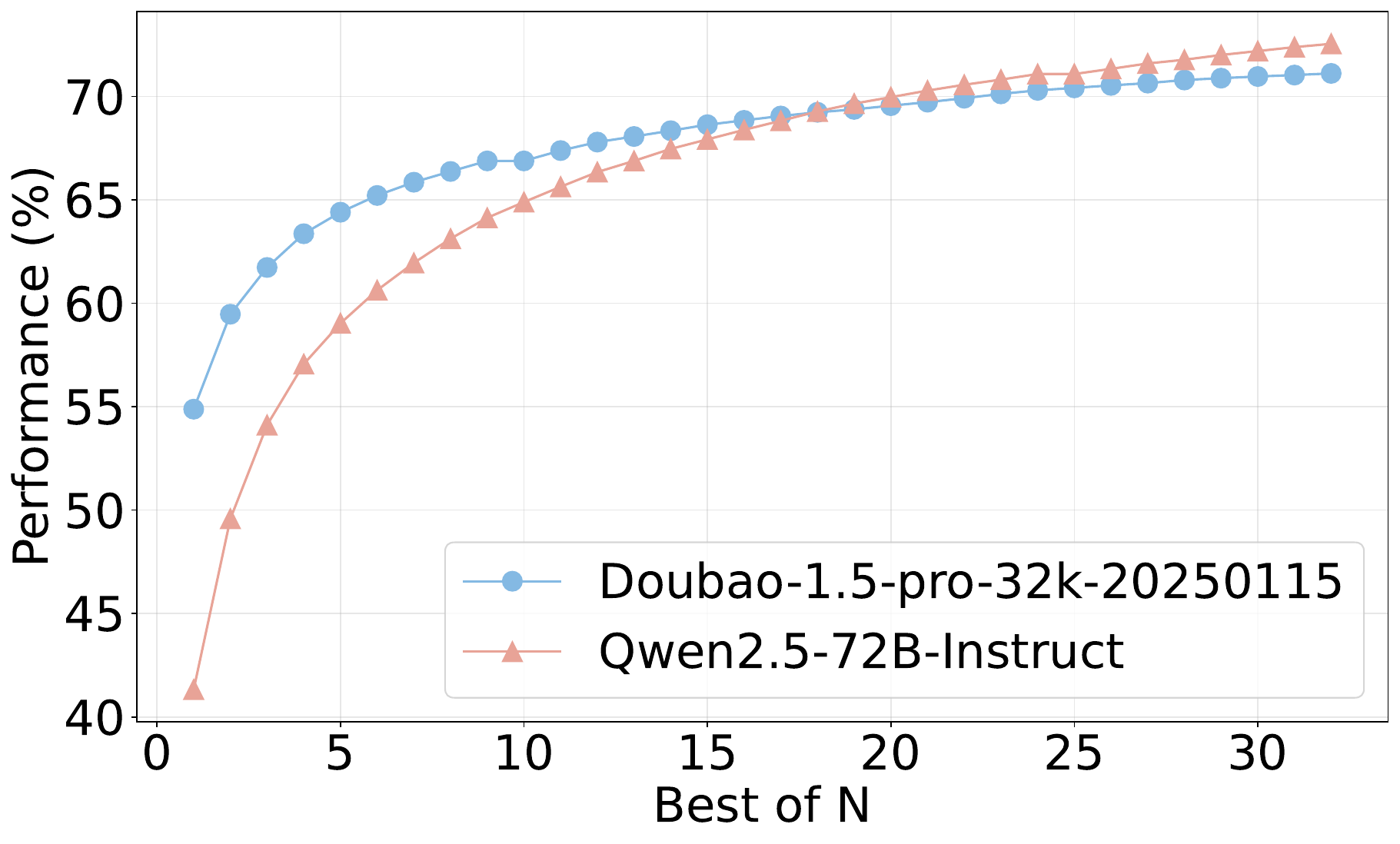}
        \caption{Best of N (BoN) performance comparison}
        \label{fig:bon}
    \end{subfigure}
    \hfill
    \begin{subfigure}[b]{0.45\textwidth}
        \centering
        \includegraphics[width=\textwidth]{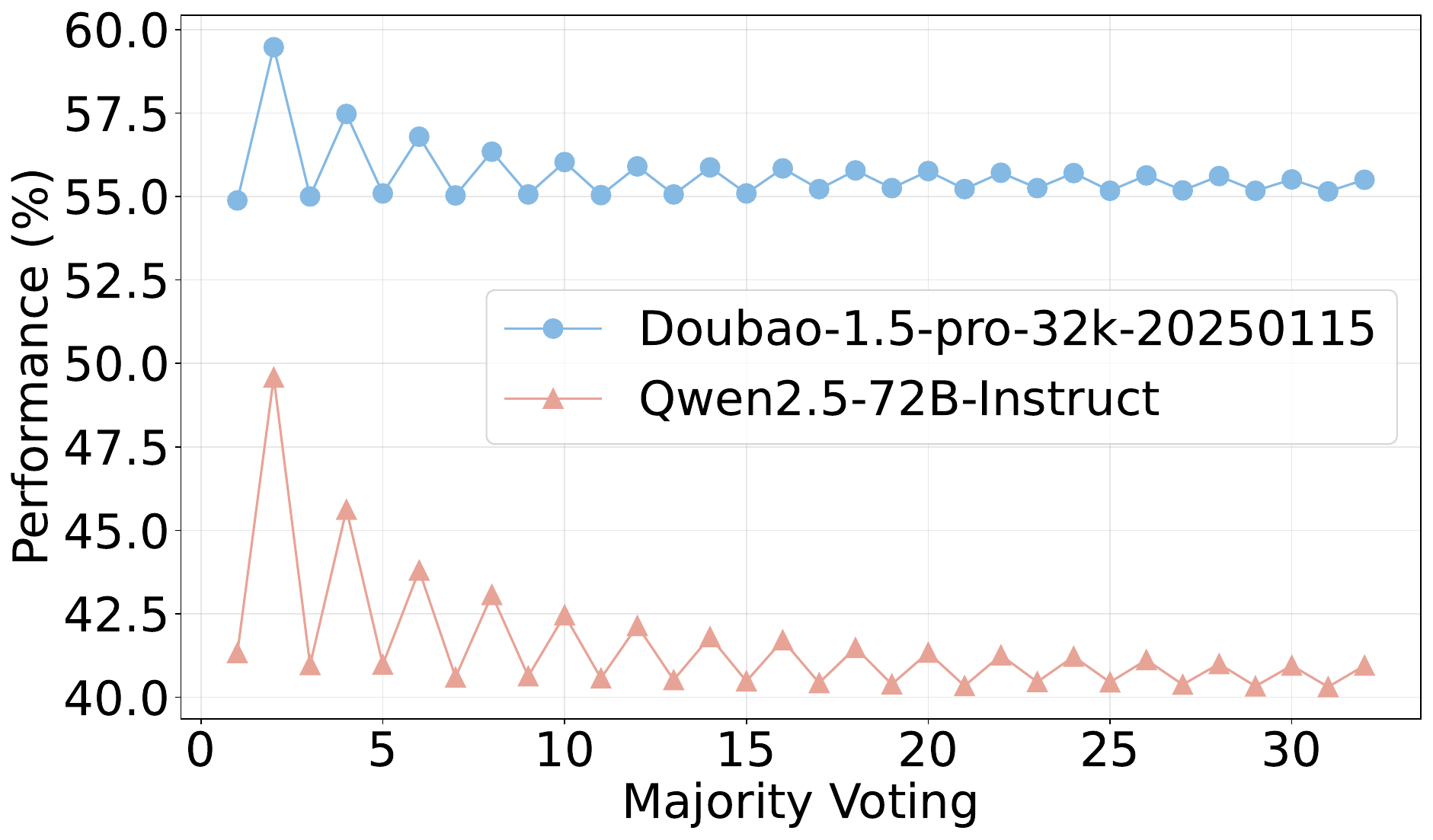}
        \caption{Majority Voting performance comparison }
        \label{fig:Majority_Voting}
    \end{subfigure}
    \caption{Performance comparison of Qwen2.5-72B-Instruct and Doubao-1.5-pro-32k-20250115 under two ensembling strategies: (a) Best of N (BoN) and (b) Majority Voting. BoN shows that Qwen2.5-72B-Instruct benefits more from increased sampling, whereas Majority Voting favors Doubao-1.5-pro-32k-20250115 due to its more consistent output distribution. }
    \label{fig:bon_Majority_Voting}
\end{figure}

\paragraph{Best of N (BoN) Analysis.}
BoN (Best of N) is a strategy that selects the highest-quality response from N independent generations which utilize stochastic sampling to improve overall performance. 
In our experiments, we evaluate Qwen2.5-72B-Instruct and Doubao-1.5-pro-32k-20250115 under BoN settings ranging from $N=1$ to $N=32$, with results presented in \autoref{fig:bon}.
\textbf{Qwen2.5-72B-Instruct exhibits a steeper BoN curve compared to Doubao-1.5-pro-32k-20250115}, suggesting that Qwen benefits more significantly from multiple sampling attempts, likely due to a higher variance in response quality. 
Conversely, \textbf{Doubao-1.5-pro-32k-20250115 shows stronger initial performance but a more gradual BoN gain}, implying that its response distribution is more consistent but less opportunistic in leveraging multiple trials.
Notably, \textbf{while Doubao maintains a lead in early BoN values ($N \leq 15$), Qwen2.5-72B-Instruct surpasses it around Bo24 and continues to outperform at Bo32}, indicating that for scenarios where extensive sampling is feasible, Qwen2.5-72B-Instruct demonstrates a greater ability to exploit high-quality outputs.

\paragraph{Majority Voting Analysis.}\footnote{Both the BoN and Majority Voting analyses are conducted using the same set of inference results, generated with temperature = 0.7 and repeated for 32 independent runs.}
Majority Voting is a strategy that selects the most frequently generated response from multiple independent runs. We evaluate Majority Voting using Qwen2.5-72B-Instruct and Doubao-1.5-pro-32k-20250115 shown in \autoref{fig:Majority_Voting}. When multiple options receive the same highest number of votes, the answer is considered correct if the correct option is among them.
We show that \textbf{Doubao-1.5-pro-32k-20250115 consistently outperforms Qwen2.5-72B-Instruct across all voting sizes, exhibiting a stable performance around 55-57\%}. 
In contrast, \textbf{Qwen2.5-72B-Instruct demonstrates fluctuations, particularly for lower N, with performance largely remaining in the 40-45\% range}, indicating that Doubao generates more consistent responses across independent runs.




\subsection{Analysis of Disciplinary Discrimination Power}
\label{sec:discrimination}
\definecolor{softblue}{rgb}{0.88, 0.95, 1.0} 
\definecolor{softred}{HTML}{FCE4D6}

To systematically evaluate the discrimination power across disciplines, we employ two complementary analytical approaches: \textbf{descriptive statistics} and \textbf{discrimination indices analysis}. The descriptive statistics approach examines the distribution characteristics of model performance within each discipline through three key metrics:
\begin{itemize}
\item \textbf{Mean Accuracy}: Reflects the overall difficulty level of the discipline.
\item \textbf{Standard Deviation (SD)}: Measures the dispersion of model performance.
\item \textbf{Coefficient of Variation (CV)}: Normalizes the standard deviation by mean accuracy, enabling cross-discipline comparison.
\end{itemize}

The discrimination indices analysis complements this by comparing performance extremes through:
\begin{itemize}
\item \textbf{High-Low Group Difference ($\Delta$)}: Calculates the mean accuracy gap between the top 3 and bottom 3 models in each discipline.
\end{itemize}

~\autoref{tab:full_analysis} presents complete results with group comparisons across all 13 disciplines, highlighting key patterns through color coding.

\begin{table}[htbp]
\centering
\small
\begin{tabular}{@{}lcccccc@{}}
\toprule
Discipline & \multicolumn{3}{c}{Descriptive Statistics} & \multicolumn{3}{c}{Discrimination Indices Analysis} \\
\cmidrule(lr){2-4} \cmidrule(lr){5-7}
 &\textbf{Mean Acc}. & \textbf{SD} & \textbf{CV} & \textbf{High} &\textbf{Low} & $\Delta$ \\
\midrule
\rowcolor{softblue}
Engineering & 53.93 & 5.75 & 0.107 & 60.85 & 47.72 & 13.13 \\
Philosophy & 55.56 & 7.33 & 0.132 & 62.92 & 46.59 & 16.33 \\
Medicine & 54.42 & 6.44 & 0.118 & 60.94 & 46.38 & 14.57 \\
Economics & 58.25 & 6.96 & 0.120 & 65.86 & 49.83 & 16.04 \\
Science & 54.52 & 6.86 & 0.126 & 62.39 & 46.94 & 15.45 \\
\rowcolor{softred}
Law & 56.92 & 7.17 & 0.126 & 65.29 & 48.68 & 16.62 \\
\rowcolor{softred}
History & 48.28 & 8.45 & 0.175 & 58.26 & 39.07 & 19.19 \\
Education & 52.15 & 5.94 & 0.114 & 57.51 & 44.15 & 13.36 \\
\rowcolor{softblue}
Military Science & 53.85 & 4.99 & 0.093 & 59.51 & 47.97 & 11.55 \\
\rowcolor{softblue}
Management & 52.73 & 5.21 & 0.099 & 58.02 & 47.04 & 10.98 \\
Literature \& Arts & 46.94 & 6.58 & 0.140 & 55.03 & 40.02 & 15.02 \\
Agronomy & 46.97 & 5.58 & 0.119 & 53.06 & 40.27 & 12.78 \\
Sociology & 57.83 & 7.33 & 0.127 & 66.20 & 50.35 & 15.85 \\
\bottomrule
\end{tabular}
\caption{Comprehensive Discrimination Analysis with Several Key Evaluation Metrics (\textbf{Mean Acc.}: Mean Accuracy, \textbf{SD}: Standard Deviation, \textbf{CV}: Coefficient of Variation, $\Delta$: High-Low Group Difference).}
\label{tab:full_analysis}
\end{table}

Our systematic analysis reveals two distinct patterns in disciplinary discrimination power:
\begin{itemize}
\item \textbf{High-discrimination disciplines}: History (SD=8.45, CV=0.175, $\Delta$=19.19) and Law (SD=7.17, CV=0.126, $\Delta$=16.62) demonstrate the strongest differentiation capacity, indicating models exhibit substantially varied performance in these domains.

\item \textbf{Low-discrimination disciplines}: Military Science (SD=4.99, CV=0.093, $\Delta$=11.55), Engineering (SD=5.75, CV=0.107, $\Delta$=13.13), and Management (SD=5.21, CV=0.099, $\Delta$=10.98) exhibit performance convergence among top models.
\end{itemize}

The observed dichotomy between humanities and STEM disciplines emerges from fundamental differences in knowledge representation. The heightened discrimination in humanities (History CV=0.175) likely originates from:
\begin{itemize}
\item Context-dependent reasoning requiring real-world knowledge synthesis.
\item Cultural nuance interpretation demands.
\item Ethical judgment variance in open-ended scenarios.
\end{itemize}

Conversely, the performance convergence in STEM fields (Engineering CV=0.107) reflects:
\begin{itemize}
\item Standardized problem-solving patterns in technical domains.
\item Mathematical consistency in training corpora.
\item Concentrated optimization efforts by model developers.
\end{itemize}

This finding validates our experimental design hypothesis: when evaluating top-performing models (per-discipline top 10 selection), humanities disciplines better reveal capability differences due to their complexity beyond pattern recognition, while STEM metrics approach performance ceilings. Our results emphasize the critical need for comprehensive cross-domain evaluation frameworks to fully capture models' heterogeneous capabilities beyond technical domains.

\section{Related Work}

\subsection{Large Language Models}

The landscape of natural language processing has been transformed by recent breakthroughs in Large Language Models (LLMs)~\citep{zhang2024mapneo,Young2024YiOF}. The introduction of GPT-3 marked a significant milestone, showcasing its ability to interpret tasks and examples from textual inputs with minimal prior training. 
Recently, the latest generation of LLMs (e.g., GPT-4~\citep{gpt4}, Claude-3.5~\footnote{\url{https://www.anthropic.com/news/claude-3-5-sonnet}}, Gemini~\citep{gemini1}, and Llama-3~\citep{llama3modelcard}), have exhibited remarkable progress in sophisticated reasoning across diverse fields.
To comprehensively evaluate and challenge the expanding capabilities of these advanced AI systems, we present \benchmark. This novel benchmark is specifically crafted to probe the knowledge boundaries of existing LLMs.


\subsection{LLM Benchmarks}

Recently, the development of the Large Language Model (LLM) has been transformed by the introduction of various benchmarks~\citep{cobbe2021training,Wang2024MTUBenchAM,bai2024mt}. Notable examples include GLUE~\citep{DBLP:conf/iclr/WangSMHLB19} and its successor SuperGLUE~\citep{DBLP:conf/nips/WangPNSMHLB19}, which have been instrumental in propelling advancements in language comprehension tasks. These foundational benchmarks paved the way for more specialized assessments, such as MMLU~\citep{hendrycks2020measuring}, HotpotQA \citep{DBLP:conf/emnlp/Yang0ZBCSM18}, BigBench~\citep{DBLP:journals/corr/abs-2206-04615}, HellaSwag~\citep{DBLP:conf/acl/ZellersHBFC19},  CommonsenseQA~\citep{DBLP:conf/naacl/TalmorHLB19}, KOR-Bench~\citep{ma2024korbenchbenchmarkinglanguagemodels},
SimpleQA~\citep{simpleqa} and Chinese SimpleQA~\citep{csimpleqa}. These newer benchmarks have expanded the evaluation scope to encompass content generation, knowledge understanding, and complex reasoning abilities.
While numerous benchmarks have been developed to evaluate LLMs’ capabilities and alignment with human values, these have often focused narrowly on performance within singular
tasks or domains. 
To enable a more comprehensive LLM assessment,
we propose \benchmark to scale the LLM evaluation to 285 graduate-level disciplines and provide a comprehensive and fine-grained analysis of foundation models.

\section{Contributions and Acknowledgements}

Multimodal Art Projection (M-A-P) is a non-profit open-source AI research community, ran by donation.
The community members are working on research topics in a wide range of spectrum, including but not limited to the pre-training paradigm of foundation models, large-scale data collection and processing, and the derived applications on coding, reasoning and music generation.

Our team members contribute to the development of \benchmark from the following perspectives: 
\begin{multicols}{2}
\begin{itemize}
    \item Data Annotation Management
    \item Data Annotation
    \item Data Quality Inspection
    \item Model Evaluation
    \item Result Analysis
    \item Paper Writing
\end{itemize}
\end{multicols}

\textbf{Leading Authors}
\begin{multicols}{2}
    \begin{itemize}
        \item Xinrun Du, M-A-P
        \item Yifan Yao, M-A-P
        \item Kaijing Ma, M-A-P
        \item Bingli Wang, SAU
        \item Tianyu Zheng, M-A-P, Tiktok
                \item King Zhu, M-A-P, OPPO
        \item Minghao Liu, 2077.AI
    \end{itemize}
\end{multicols}

\textbf{Outstanding Contributors}
\begin{multicols}{2}
    \begin{itemize}
        \item Yiming Liang, M-A-P, CASIA
        \item Xiaolong Jin, Purdue University
        \item Zhenlin Wei, HEU
        \item Chujie Zheng, Tsinghua University
    \end{itemize}
\end{multicols}

\textbf{Core Contributors (Alphabet Order)}
\begin{multicols}{2}
    \begin{itemize}
        \item Kaixin Deng, CDUT
        \item Shawn Gavin, M-A-P
        \item Shian Jia, Zhejiang University
        \item Sichao Jiang, Zhejiang University
        \item Qinrui Li, UCSB
        \item Rui Li, Peking University
        \item Sirun Li, Peking University
        \item Yizhi Li, The University of Manchester
        \item Yunwen Li, CUHK-Shenzhen
        \item Yiyan Liao, Peking University
        \item David Ma, M-A-P
        \item Yuansheng Ni, M-A-P
        \item Haoran Que, Zhipu
        \item Qiyao Wang, DUT
        \item Zekun Moore Wang, M-A-P, Beihang University
        \item Zhoufutu Wen, ByteDance.Inc
        \item Siwei Wu, The University of Manchester
        \item Tyshawn Hsing, M-A-P
        \item Ming Xu, NJUPT
        \item Zhenzhu Yang, M-A-P, CUGB
        \item Junting Zhou, Peking University
    \end{itemize}
\end{multicols}

\textbf{Contributors (Alphabet Order)}
\begin{multicols}{2}
    \begin{itemize}
        \item Yuelin Bai, M-A-P
        \item Xingyuan Bu, Alibaba.Inc
        \item Chenglin Cai, 01.AI
        \item Liang Chen, Peking University
        \item Yifan Chen, ByteDance.Inc
        \item Chengtuo Cheng, Abaka.AI
        \item Tianhao Cheng, Fudan University
        \item Keyi Ding, Hangzhou Dianzi University
        \item Siming Huang, The University of Melbourne
        \item Yun Huang, NUS
        \item Yaoru Li, Zhejiang University
        \item Yizhe Li, Zhejiang University
        \item Zhaoqun Li, Zhejiang University
        \item Tianhao Liang, Zhejiang University
        \item Chengdong Lin, Hangzhou Dianzi University
        \item Hongquan Lin, University of Science and Technology of China
        \item Yinghao Ma, Queen Mary University of London
        \item Tianyang Pang, ByteDance.Inc
        \item Zhongyuan Peng, Alibaba.Inc
        \item Zifan Peng, HKUST-Guangzhou
        \item Qige Qi, ByteDance.Inc
        \item Shi Qiu, Peking University
        \item Xingwei Qu, The University of Manchester
        \item Shanghaoran Quan, Beihang University
        \item Yizhou Tan, Harvard University
        \item Chenqing Wang, 2077.AI
        \item Hao Wang, Beihang University
        \item Yiya Wang, Peking University
        \item Yubo Wang, University of Waterloo
        \item Zili Wang
        \item Jiajun Xu, Meta
        \item Kexin Yang
        \item Ruibin Yuan, HKUST
        \item Yuanhao Yue, Fudan University
        \item Tianyang Zhan, ByteDance.Inc
        \item Chun Zhang, ByteDance.Inc
        \item Jinyang Zhang, Zhejiang University
        \item Xingjian Zhang, Princeton University
        \item Xiyue Zhang, Peking University
        \item Yue Zhang, ByteDance.Inc
        \item Yongchi Zhao, Alibaba.Inc
        \item Xiangyu Zheng, Fudan University
        \item Chenghua Zhong, USTB
    \end{itemize}
\end{multicols}

\textbf{Organization and Sponsor Committee (Alphabet Order)}
\begin{multicols}{2}
    \begin{itemize}
        \item Meng Cao, MBZUAI
        \item Yang Gao, Nanjing University
        \item Zhoujun Li, Beihang University
        \item Dayiheng Liu
        \item Qian Liu, Tiktok
        \item Tianyu Liu        
        \item Shiwen Ni, SIAT-CAS
        \item Junran Peng, USTB
        \item Yujia Qin, ByteDance.Inc
        \item Wenbo Su
        \item Guoyin Wang, ByteDance.Inc
        \item Shi Wang, ICT-CAS
        \item Jian Yang, Beihang University
        \item Min Yang, SIAT-CAS
        \item Xiang Yue, M-A-P
        \item Zhaoxiang Zhang, CASIA
        \item Wangchunshu Zhou, OPPO
    \end{itemize}
\end{multicols}

\textbf{Corresponding Authors}
\begin{multicols}{2}
    \begin{itemize}
        \item Jiaheng Liu, M-A-P, Nanjing University
        \item Qunshu Lin, Abaka.AI
        \item Wenhao Huang, M-A-P, ByteDance.Inc
        \item Ge Zhang, M-A-P, ByteDance.Inc
    \end{itemize}
\end{multicols}
\newpage

\bibliography{main.bib}

\newpage
\appendix
\section{Difficulty-Stratified Samples}
\newtcolorbox[auto counter, number within=section]{easybox}[2][]{%
  colback=white, 
  colframe=orange!90!black, 
  width=\textwidth,
  arc=2mm, 
  boxrule=0.5mm, 
  title={\normalsize\faLeaf\hspace{0.5em}#2},
  breakable, 
  fonttitle=\bfseries\Large, 
  fontupper=\small
  #1
}

\newtcolorbox[auto counter, number within=section]{middlebox}[2][]{%
  colback=white, 
  colframe=magenta!40!red!80!black, 
  width=\textwidth,
  arc=2mm, 
  boxrule=0.5mm, 
  title={\normalsize\faSearchPlus\hspace{0.5em}#2},
  breakable, 
  fonttitle=\bfseries\Large, 
  fontupper=\small
  #1
}

\newtcolorbox[auto counter, number within=section]{hardbox}[2][]{%
  colback=white, 
  colframe=red!50!black, 
  width=\textwidth,
  arc=2mm, 
  boxrule=0.5mm, 
  title={\normalsize\faTrophy\hspace{0.5em}#2},
  breakable, 
  fonttitle=\bfseries\Large, 
  fontupper=\small
  #1
}

\begin{easybox}{Easy Sample}
\textbf{Question:}
\begin{enumerate}[leftmargin=0.5cm, label={}]  
\item Which of the following statements about dance studies are correct?\begin{enumerate}[label=\arabic*.]
    \item According to Danto's definition, context is an art world with modern aspects.
    \item ``La Bayadère'' is a ballet created during the French July Revolution.
    \item The ballet ``Sylvia'' is a dance drama created during the Paris Commune period in 1871.
    \item Korean court dance, when calculated according to temporal principles, does not belong to secondary civilization.
\end{enumerate}
\end{enumerate}

\textbf{Options:}
\begin{enumerate}[leftmargin=0.5cm, label={}]  
\item A) $1,3$
\item B) $1,4$
\item C) 1,2,4
\item D) 2,3 
\item E) 1,2,3
\item F) 3 
\item G) 4
\item H) 3,4 
\item I) 1,2,3,4 
\item J) 2,4 
\end{enumerate}

\textbf{Answer:} $3,4$

\textbf{Answer letter:} H

\textbf{Discipline:} Literature and Arts

\textbf{Field:} Art Studies

\textbf{Subfield:} Dance Studies

\textbf{Difficulty:} easy
\end{easybox}

\begin{middlebox}{Middle Sample}
\textbf{Question:}
\begin{enumerate}[leftmargin=0.5cm, label={}]  
\item A deck of playing cards has 52 cards, and each shuffle changes the order of the cards, which is a permutation. If each shuffle strictly follows the operation below: first divide the cards into two equal parts, then interlace the cards from each part alternately, so that the first and last cards remain in their original positions while all other cards are rearranged. Express this permutation as a product of disjoint cycles, write out the cyclic structure of this permutation, and determine the minimum number of shuffles needed to restore the deck to its original order.
\end{enumerate}

\textbf{Options:}
\begin{enumerate}[leftmargin=0.5cm, label={}]  
\item A) $(1^2, 2, 8^6), 8$
\item B) $(1^2, 5, 5^6), 6$
\item C) $(1^3, 1, 8^5), 15$
\item D) $(1^1, 4, 7^6), 11$
\item E) $(1^2, 4, 6^4), 12$
\item F) $(2^2, 1, 9^5), 9$
\item G) $(1^2, 3, 7^5), 10$
\item H) $(2^1, 2, 7^6), 7$
\item I) $(2^2, 3, 6^5), 5$
\item J) $(1^2, 6, 4^6), 4$
\end{enumerate}

\textbf{Answer:}$(1^2, 2, 8^6), 8$

\textbf{Answer letter:} A

\textbf{Discipline:} Science

\textbf{Field:} Mathematics

\textbf{Subfield:} Group Theory

\textbf{Difficulty:} middle

\end{middlebox}

\begin{hardbox}{Hard Sample}

\textbf{Question:}
\begin{enumerate}[leftmargin=0.5cm, label={}]  
    \item A certain transmitter has a transmission power of 10 W, a carrier frequency of 900 MHz, a transmission antenna gain \( G_{\mathrm{T}}=2 \), and a receiving antenna gain \( G_{\mathrm{R}}=3 \). Calculate the output power of the receiver and the path loss at a distance of 10 km from the transmitter in free space.
\end{enumerate}

\textbf{Options:}
\begin{enumerate}[leftmargin=1cm, label=\Alph*)]  
    \item \( P_{\mathrm{R}}=R_{\mathrm{T}} G_{\mathrm{T}} G_{\mathrm{R}} \left( \frac{\lambda} {3 \pi d} \right)^{2}=2. \, 5 5 \times1 0^{-1 2} \, \, \, \mathrm{W} \) \\
          \( L={\frac{P_{\mathrm{T}}} {P_{\mathrm{R}}}}=5. \, 1 9 \times1 0^{1 1} \, ( \mathrm{~ B I I ~ 1 4 2. ~ 4 3 ~ d B} ) \)
    \item \( P_{\mathrm{R}}=R_{\mathrm{T}} G_{\mathrm{T}} G_{\mathrm{R}} \left( \frac{\lambda} {4 \pi d} \right)^{2}=5. \, 0 7 \times1 0^{-1 1} \, \, \, \mathrm{W} \) \\
          \( L={\frac{P_{\mathrm{T}}} {P_{\mathrm{R}}}}=1. \, 9 7 \times1 0^{1 0} \, ( \mathrm{~ B I I ~ 1 0 1. ~ 9 5 ~ d B} ) \)
    \item \( P_{\mathrm{R}}=R_{\mathrm{T}} G_{\mathrm{T}} G_{\mathrm{R}} \left( \frac{\lambda} {4 \pi d} \right)^{2}=1. \, 7 4 \times1 0^{-1 0} \, \, \, \mathrm{W} \) \\
          \( L={\frac{P_{\mathrm{T}}} {P_{\mathrm{R}}}}=4. \, 8 1 \times1 0^{9} \, ( \mathrm{~ B I I ~ 9 8. ~ 4 8 ~ d B} ) \)
    \item \( P_{\mathrm{R}}=R_{\mathrm{T}} G_{\mathrm{T}} G_{\mathrm{R}} \left( \frac{\lambda} {5 \pi d} \right)^{2}=8. \, 4 0 \times1 0^{-1 0} \, \, \, \mathrm{W} \) \\
          \( L={\frac{P_{\mathrm{T}}} {P_{\mathrm{R}}}}=3. \, 9 8 \times1 0^{9} \, ( \mathrm{~ B I I ~ 9 6. ~ 0 7 ~ d B} ) \)
    \item \( P_{\mathrm{R}}=R_{\mathrm{T}} G_{\mathrm{T}} G_{\mathrm{R}} \left( \frac{\lambda} {4 \pi d} \right)^{2}=2. \, 9 3 \times1 0^{-1 0} \, \, \, \mathrm{W} \) \\
          \( L={\frac{P_{\mathrm{T}}} {P_{\mathrm{R}}}}=6. \, 2 9 \times1 0^{9} \, ( \mathrm{~ B I I ~ 9 9. ~ 9 0 ~ d B} ) \)
    \item \( P_{\mathrm{R}}=R_{\mathrm{T}} G_{\mathrm{T}} G_{\mathrm{R}} \left( \frac{\lambda} {3 \pi d} \right)^{2}=6. \, 8 3 \times1 0^{-1 0} \, \, \, \mathrm{W} \) \\
          \( L={\frac{P_{\mathrm{T}}} {P_{\mathrm{R}}}}=1. \, 4 6 \times1 0^{1 0} \, ( \mathrm{~ B I I ~ 8 9. ~  35 ~ d B} ) \)
    \item \( P_{\mathrm{R}}=R_{\mathrm{T}} G_{\mathrm{T}} G_{\mathrm{R}} \left( \frac{\lambda} {4 \pi d} \right)^{3}=9. \, 1 1 \times1 0^{-1 1} \, \, \, \mathrm{W} \) \\
          \( L={\frac{P_{\mathrm{T}}} {P_{\mathrm{R}}}}=3. \, 6 4 \times1 0^{1 0} \, ( \mathrm{~ B I I ~ 1 0 4. ~ 7 8 ~ d B} ) \)
    \item \( P_{\mathrm{R}}=R_{\mathrm{T}} G_{\mathrm{T}} G_{\mathrm{R}} \left( \frac{\lambda} {4 \pi d} \right)^{1}=3. \, 1 2 \times1 0^{-0 9} \, \, \, \mathrm{W} \) \\
          \( L={\frac{P_{\mathrm{T}}} {P_{\mathrm{R}}}}=9. \, 2 4 \times1 0^{9} \, ( \mathrm{~ B I I ~ 9 9. ~ 6 3 ~ d B} ) \)
    \item \( P_{\mathrm{R}}=R_{\mathrm{T}} G_{\mathrm{T}} G_{\mathrm{R}} \left( \frac{\lambda} {4 \pi d} \right)^{2}=4. \, 2 2 \times1 0^{-1 0} \, \, \, \mathrm{W} \) \\
          \( L={\frac{P_{\mathrm{T}}} {P_{\mathrm{R}}}}=2. \, 3 7 \times1 0^{1 0} \, ( \mathrm{~ B I I ~ 1 0 3. ~ 7 5 ~ d B} ) \)
    \item \( P_{\mathrm{R}}=R_{\mathrm{T}} G_{\mathrm{T}} G_{\mathrm{R}} \left( \frac{\lambda} {6 \pi d} \right)^{2}=7. \, 9 5 \times1 0^{-1 1} \, \, \, \mathrm{W} \) \\
          \( L={\frac{P_{\mathrm{T}}} {P_{\mathrm{R}}}}=2. \, 5 0 \times1 0^{1 0} \, ( \mathrm{~ B I I ~ 1 0 2. ~ 2 0 ~ d B} ) \)
\end{enumerate}

\textbf{Answer:}
\[
P_{\mathrm{R}}=R_{\mathrm{T}} G_{\mathrm{T}} G_{\mathrm{R}} \left( \frac{\lambda} {4 \pi d} \right)^{2}=4. \, 2 2 \times1 0^{-1 0} \, \, \, \mathrm{W}
\]
\[
L={\frac{P_{\mathrm{T}}} {P_{\mathrm{R}}}}=2. \, 3 7 \times1 0^{1 0} \, ( \mathrm{~ B I I ~ 1 0 3. ~ 7 5 ~ d B} )
\]

\textbf{Answer letter:} I

\textbf{Discipline:} Engineering

\textbf{Field:} Information and Communication Engineering

\textbf{Subfield:} Communication Principles

\textbf{Difficulty:} hard

\end{hardbox}

\newtcolorbox[auto counter, number within=section]{methodbox}[2][]{%
  colback=white, 
  colframe=teal!80!green!80!black,  
  width=\textwidth,
  arc=2mm, 
  boxrule=0.5mm, 
  title={\normalsize\faWrench\hspace{0.5em}#2}, 
  breakable,
  fonttitle=\bfseries\Large, 
  fontupper=\small, 
  #1
}

\section{Annotation Tutorial}
\label{appendix: annotation tutorial}
\subsection{Material Requirements}
The specific requirements for material selection are as follows:
\begin{itemize}
    \item Ensure the selected materials are accurate, with correct answers.
    \item Ensure the materials cover a variety of knowledge points and are free from regional bias.
    \item The selected materials should be correctly stated and involve a certain level of academic reasoning, with a difficulty appropriate for graduate-level study.
    \item Avoid using materials that rely on images as conditions.
    \item Verify that the materials comply with copyright requirements.
\end{itemize}
\subsection{Annotation Methods}
During the annotation process, the following four methods are primarily used:  \textbf{1) Original Transcription Method}, \textbf{2) Non-Choice Conversion Method}, and \textbf{3) Statement Combination Method}.
\subsubsection{Original Transcription Method}
\begin{methodbox}{Original Transcription}
\subsection*{Description:}
\begin{itemize}
    \item Directly transcribe the original multiple-choice questions, ensuring the question maintains its original meaning and correctness. Additional distractors are included to increase the difficulty of the question or enhance its discriminative power.
\end{itemize}
\subsection*{Operations:}
\begin{enumerate}
    \item \textbf{Material Review:}
    \begin{itemize}
        \item Ensure that the materials meet the aforementioned requirements, including accuracy, diversity, neutrality (free from regional bias), non-image-based content, and compliance with copyright regulations.
    \end{itemize}
    \item \textbf{Question Transcription:}
    \begin{itemize}
        \item Use OCR tools to recognize the text in the materials or directly paste the original content. Transcribe the question content word-for-word, ensuring no key information is omitted, with particular attention to the accurate transcription of formulas.
    \end{itemize}
    \item \textbf{Option Transcription:}
    \begin{itemize}
        \item Use OCR tools to recognize the text in the materials or directly paste the original content. Transcribe the options word-for-word, ensuring no key information is omitted, with particular attention to the accurate transcription of formulas.
    \end{itemize}
    \item \textbf{Answer Identification:}
    \begin{itemize}
        \item Clearly mark the correct answer within the list of options, ensuring its accuracy and uniqueness.
    \end{itemize}
    \item \textbf{Distractor Addition:}
    \begin{itemize}
        \item Add distractors while maintaining the quality of the options (avoiding meaningless or excess correct answers) until the annotator deems the level of confusion sufficient. The number of options should be between 4 and 10, with more options being preferred to increase the difficulty and discriminatory power of the question.
    \end{itemize}
    \item \textbf{Field Completion:}
    \begin{itemize}
        \item Complete the category fields ("discipline","field","subfield") and select the appropriate difficulty level to ensure the integrity and accuracy of the question information.
    \end{itemize}
\end{enumerate}
\end{methodbox}

\subsubsection{Non-Choice Conversion Method}
\begin{methodbox}{Non-Choice Conversion}
\subsection*{Description:}
\begin{itemize}
    \item Convert non-multiple-choice questions (such as calculation questions, fill-in-the-blank questions, etc.) into multiple-choice format, ensuring that the conditions of the question are complete and the final result is accurate while generating reasonable distractors based on the analysis steps or calculation process.
\end{itemize}
\subsection*{Operations:}
\begin{enumerate}
    \item \textbf{Material Review:}
    \begin{itemize}
        \item Ensure that the materials meet the aforementioned requirements, including accuracy, diversity, neutrality (free from regional bias), non-image-based content, and compliance with copyright regulations.
        \item Ensure that the conditions of the question are complete and do not cause ambiguity. Add supplementary explanations if necessary.
        \item Ensure that the answer does not contain any essential explanatory process that is unsuitable for adaptation or transcription.
    \end{itemize}
    \item \textbf{Question Transcription:}
    \begin{itemize}
        \item Use OCR tools to recognize the text in the material or directly paste the original text. Transcribe the content of the original question word by word, ensuring that all numbers, formulas, or condition information in the question are accurate and complete to avoid ambiguity or incorrect answers.
        \item Add supplementary information if the question requires prior conditions from other questions.
    \end{itemize}
    \item \textbf{Answer Transcription:}
    \begin{itemize}
        \item Identify and select the correct result for the appropriate question based on the answer analysis process.
        item Use OCR tools to recognize results in the material and confirm that all numerical and formula information in the answer is accurate.
    \end{itemize}
    \item \textbf{Distractor Addition:}
    \begin{itemize}
        \item On the basis of the correct answer, consider setting common calculation errors (such as rounding errors, sign mistakes, digit errors, etc.) as distractors.
        \item For numerical results, the setting of distractors should consider reasonable ranges of calculation errors to ensure differentiation between options.
    \end{itemize}
    \item \textbf{Answer Identification:}
    \begin{itemize}
        \item Clearly mark the correct answer within the list of options, ensuring its accuracy and uniqueness.
    \end{itemize}
    \item \textbf{Field Completion:}
    \begin{itemize}
        \item Complete the category fields ("discipline","field","subfield") and select the appropriate difficulty level to ensure the integrity and accuracy of the question information.
    \end{itemize}
\end{enumerate}
\end{methodbox}
\subsubsection{Statement Combination Method}
\begin{methodbox}{Statement Combination}
\subsection*{Description:}
\begin{itemize}
    \item By integrating stated expressions related to multiple concepts or knowledge points, a multi-level, multi-perspective multiple-choice question is constructed to increase the comprehensiveness and depth of examination of the topic.
\end{itemize}
\subsection*{Operations:}
\begin{enumerate}
    \item \textbf{Statement Extraction:}
    \begin{itemize}
        \item Extract core concepts, definitions, relationships between concepts, application cases, and common misconceptions from textbooks and related learning resources, ensuring that the statements are representative and comprehensive.
        \item Extract important statements from multiple-choice questions, covering multiple knowledge points to avoid being limited to a single definition or formula, thus enhancing the overall comprehensiveness of the questions.
        \item Ensure that the extracted statements are accurate and concise, avoiding redundancy or vague expressions. The content should cover both correct and incorrect situations to provide a foundation for subsequent adaptation.
    \end{itemize}
    \item \textbf{Statement Adaptation:}
    \begin{itemize}
        \item Adapt the extracted statements to include both correct and incorrect versions.
        \item Incorrect statements should be somewhat misleading, avoiding obvious or easily dismissible errors.
        \item Number the adapted statements (e.g., I, II, III, etc.). Arrange the statements in a reasonable order, avoiding bias towards any direction (e.g., always placing correct statements first).
    \end{itemize}
    \item \textbf{Question Design:}
    \begin{itemize}
        \item Depending on the need, questions may limit the scope or conditions being tested.
        \item "Which of the following statements about [knowledge point] is correct?"
        \item "Which of the following statements about [knowledge point] is incorrect?"
    \end{itemize}
    \item \textbf{Combination Design:}
    \begin{itemize}
        \item Combine the numbered statements to create options, such as I and II; II and III; III, VI, and VII.
        \item Ensure that there is exactly one correct answer among the options.
    \end{itemize}
    \item \textbf{Answer Identification:}
    \begin{itemize}
        \item Clearly mark the correct answer within the list of options, ensuring its accuracy and uniqueness.
    \end{itemize}
    \item \textbf{Field Completion:}
    \begin{itemize}
        \item Complete the category fields ("discipline","field","subfield") and select the appropriate difficulty level to ensure the integrity and accuracy of the question information.
    \end{itemize}
\end{enumerate}
\end{methodbox}
\subsubsection{Confusion-options Generation}
During the annotation process, we use large models to assist in generating distractors. We select Claude-3.5, GPT-4, Doubao, and Qwen2.5-72B-Instruct, and follow the prompt below to generate confusion options. In the annotation process, a model is randomly selected to generate a confusion option, which is then double-checked by the annotator. Finally, the confirmed confusion option is used as the final option.

\begin{methodbox}{Prompt}
You are an expert in creating multiple-choice questions. Your task is to generate plausible but incorrect distractors for a given question that only has one correct answer. You are skilled at introducing subtle yet distinct mathematical errors to the existing answer options, making the distractor look reasonable but still wrong.

\vspace{0.5em}  
\textbf{Input Description:}
\begin{itemize}
    \item You will be given:
    \begin{itemize}
        \item A question stem.
        \item A set of answer options (no numbering).
        \item The correct answer (no numbering).
    \end{itemize}
\end{itemize}

\vspace{0.5em}  
\textbf{Guidelines:}
\begin{enumerate}
    \item \textbf{Generate Distractor:}
    \begin{itemize}
        \item The distractor must be incorrect (not the correct answer).
        \item The distractor should introduce a subtle mathematical error while maintaining the formula structure.
        \item It must be \textbf{distinct} from all existing options, including the correct answer.
        \item Avoid any \textbf{repetition} or overlap with the existing answer options in terms of value or meaning.
        \item \textbf{Plausibility}: The distractor should look reasonable and appear to be a potential answer, but still be wrong.
        \item Ensure the question still has exactly \textbf{one correct answer} after adding the distractor.
    \end{itemize}
    
    \item \textbf{Uniqueness Check:}
    \begin{itemize}
        \item \textbf{Thoroughly check} all existing options (including the correct one) to ensure the distractor is unique.
        \item If the distractor matches any existing option, regenerate it.
        \item The distractor must \textbf{not} create ambiguity; the correct answer must remain the sole valid choice.
    \end{itemize}
    
    \item \textbf{Formatting:}
    \begin{itemize}
        \item Ensure the distractor matches the format of the existing options (e.g., fractions, exponents, etc.).
        \item Avoid formatting inconsistencies such as misplaced symbols or spaces.
    \end{itemize}
\end{enumerate}

\vspace{0.5em}  
\textbf{Output Format:}
\begin{verbatim}
<distractor> your generated distractor here </distractor>
\end{verbatim}
\noindent Do not include explanations or extra information.  
\noindent Do not include numbering.

\vspace{0.5em}  
\textbf{Input:}
\begin{verbatim}
    Question: "{}",
    Options: {},
    Correct Answer: "{}"
\end{verbatim}
\noindent \textbf{Output:}
\end{methodbox}
\newtcolorbox[auto counter, number within=section]{purposebox}[2][]{%
  colback=white, 
  colframe=blue!50!black, 
  width=\textwidth,
  arc=2mm, 
  boxrule=0.5mm, 
  title={\normalsize\faCompass\hspace{0.5em}#2},
  breakable, 
  fonttitle=\bfseries\Large, 
  fontupper=\small
  #1
}

\newtcolorbox[auto counter, number within=section]{promptbox}[2][]{%
  colback=white, 
  colframe=purple!70!blue!80!black,  
  width=\textwidth,
  arc=2mm, 
  boxrule=0.5mm, 
  title={\normalsize\faInfoCircle\hspace{0.5em}#2},
  breakable,
  fonttitle=\bfseries\Large, 
  fontupper=\small
  #1
}
\section{Data Filtering and Manual Review Process Details}

In the data processing workflow, we employ a three-stage filtering and review mechanism to ensure data quality:

\begin{enumerate}
    \item \textbf{Rule-Based Pre-Check~\autoref{Appendix: Rule-Based Check}}: Data undergoes initial screening based on predefined code rules, quickly identifying and eliminating clearly invalid data points, thereby enhancing efficiency and reducing subsequent workload.
    
    \item \textbf{LLM-Based Quality Inspection~\autoref{Appendix: LLM-Based Screening}}: Large language models are utilized to perform in-depth quality assessments of the data, identifying potential errors and inconsistencies, thereby improving the accuracy and completeness of the data.
    
    \item \textbf{Manual Quality Review~\autoref{Appendix: Manual Quality Review}}: Experienced data analysts conduct a final review to ensure the high quality and usability of the data, correcting potential misjudgments and incorporating domain knowledge for a comprehensive evaluation.
\end{enumerate}

\subsection{Rule-Based Pre-Check}\label{Appendix: Rule-Based Check}

\begin{table}[ht]
\centering
\resizebox{\textwidth}{!}{
\begin{tabular}{l p{16cm}}
    \toprule 
    \textbf{Category} & \textbf{Rules} \\
    \midrule
    \multirow{4}{*}{\raisebox{-3ex}{\raisebox{-1ex}{\includegraphics[width=0.5cm]{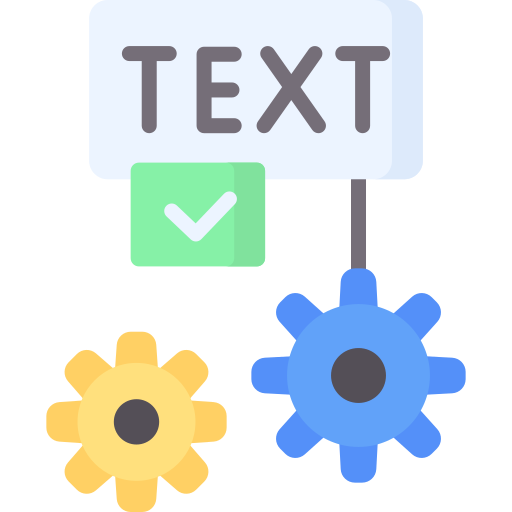}} \quad \textbf{Text}}}
    & Replace full-width punctuation with half-width punctuation. \\
    \cmidrule(lr){2-2}
    & Remove unnecessary whitespace characters (e.g., spaces, newline characters, tab spaces). \\
    \cmidrule(lr){2-2}
    & Replace common escape characters. \\
    \cmidrule(lr){2-2}
     & Add missing escape characters. \\
    \midrule
    \multirow{8}{*}{\raisebox{-15ex}{\raisebox{-1ex}{\includegraphics[width=0.5cm]{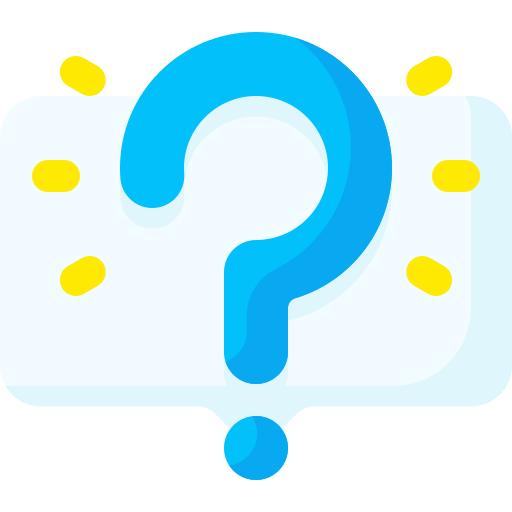}}\quad\textbf{Question}}} 
    & Not Empty or Placeholder. \\
    \cmidrule(lr){2-2}
    & Ensure the question field is in string format to avoid errors. \\
    \cmidrule(lr){2-2}
    & Text length \textgreater \, 5.  \\
    \cmidrule(lr){2-2}
    & Text Standardization. \\
    \cmidrule(lr){2-2}
    & Perplexity \(\leq\) 100, calculated using the \textit{Qwen2.5-0.5B-Instruct} model. \\
    \cmidrule(lr){2-2}
    & No Semantic Duplicates: For a given question \(q_i\), the cosine similarity with each existing question \(q_j\) :  
    \[ \text{CosineSimilarity}(q_i, q_j) = \frac{\mathbf{e_i} \cdot \mathbf{e_j}}{\|\mathbf{e_i}\| \|\mathbf{e_j}\|}<0.90 \]  where \(\mathbf{e_i}\) and \(\mathbf{e_j}\) are embeddings generated by the model SentenceTransformer("all-MiniLM-L6-v2"), and \textit{Faiss} is used for efficient similarity search. \\
    \midrule
    \multirow{5}{*}{\raisebox{-8ex}{\raisebox{-1ex}{\includegraphics[width=0.5cm]{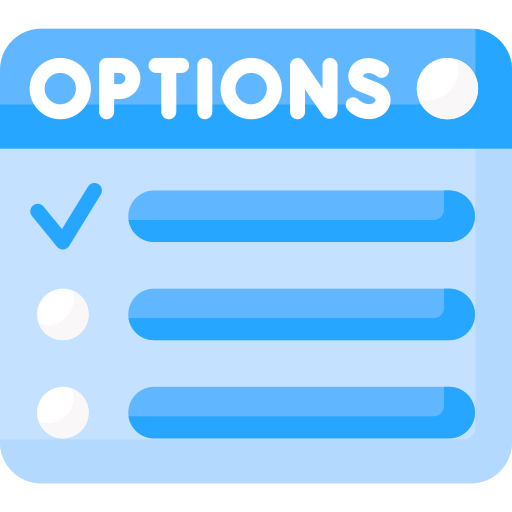}}\quad\textbf{Options}}}    
    & 4 \(\leq\) Option count \(\leq\) 10. \\
    \cmidrule(lr){2-2}
    & No empty or meaningless options (e.g., no empty strings, no only whitespace characters, no common placeholders like 'none', 'null'). \\
    \cmidrule(lr){2-2}
    & No duplicate options. \\
    \cmidrule(lr){2-2}
    & Remove built-in option prefixes, using regular expressions(\texttt{r'\^{}([A-Ja-j])[\textbackslash.\textbackslash、\textbackslash s]'} ,
\texttt{r'\^{}\textbackslash(([A-Ja-j])\textbackslash)'} ,
\texttt{r'\^{}([A-Ja-j])\textbackslash)'} ). \\
    \cmidrule(lr){2-2}
    & Text Standardization for each option. \\
    \midrule
    \multirow{3}{*}{\raisebox{-3ex}{\raisebox{-1ex}{\includegraphics[width=0.5cm]{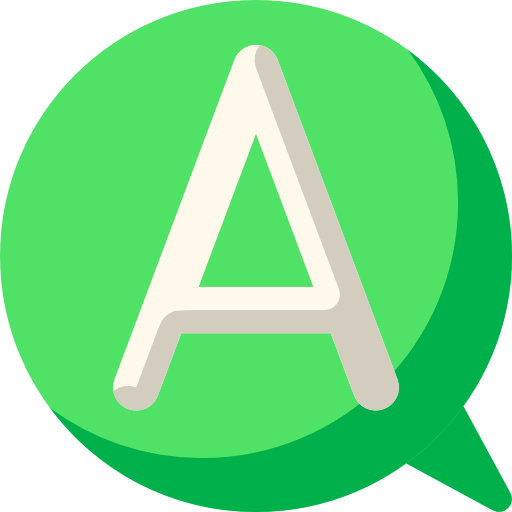}}\quad\textbf{Answer}}}   
    & Not Empty or Placeholder. \\
    \cmidrule(lr){2-2}
    & Within options list. \\
    \cmidrule(lr){2-2}
    & Text Standardization. \\
    \midrule
    \raisebox{-0.8ex}{\includegraphics[width=0.5cm]{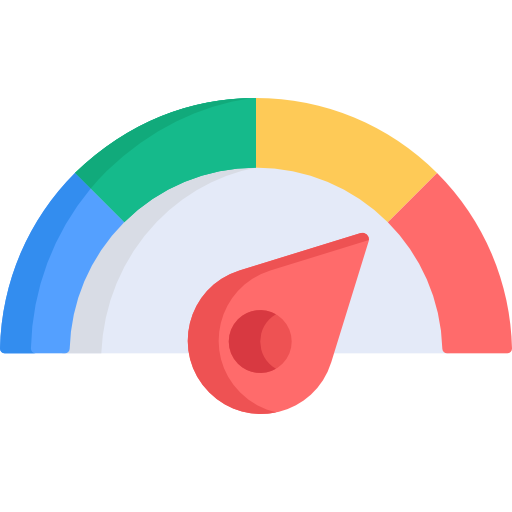}}\quad\textbf{Difficulty}    
    & Ensure difficulty exists. \\
    \midrule
    \raisebox{-0.8ex}{\includegraphics[width=0.5cm]{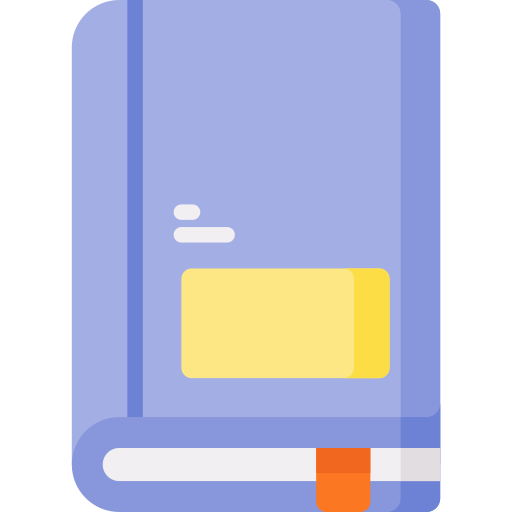}}\quad\textbf{Discipline}  
    & Ensure discipline, field, and subfield exist. \\
    \bottomrule
\end{tabular}
}
\caption{Detailed Criteria for Rule-Based Pre-Check.}
\label{tab: rule}
\end{table}

\autoref{tab: rule} presents the rules for data filtering and pre-checking, including text normalization, validation of the consistency of questions and options, integrity checks of answers, and requirements for the completeness of metadata such as difficulty, discipline, and others.

\newpage
\subsection{LLM-Based Quality Inspection}

This section outlines the quality inspection process of LLMs in data processing, including:
\textbf{1) Validity Check},
\textbf{2) Negative and Extreme Inquiry Detection}, 
\textbf{3) Multimodal Exclusion},  
\textbf{4) Field Relevance Evaluation},  
\textbf{5) Solvability Assessment}.

\label{Appendix: LLM-Based Screening}
\subsubsection{Validity Check}

\begin{purposebox}{Purpose}
\begin{itemize}
    \item Validate the question's correctness, ensuring it follows the standards, can be answered with the given text, and the options are complete and reasonable.
\end{itemize}
\end{purposebox}

\begin{promptbox}{Prompt}
You are a master of format checking, skilled at verifying whether multiple-choice questions in JSON format conform to the specifications.

\vspace{0.5em}  
You will receive an input in JSON format containing a question, options, and the answer, for example:

\begin{verbatim}
    {
    question: "Question content",
    options: ['Option content', 'Option content', 
              'Option content', 'Option content'],
    answer: "Answer content"
    }
\end{verbatim}

\vspace{0.5em}  
Please determine whether the text in the JSON represents a complete and solvable multiple-choice question. It must meet the following requirements:
\begin{enumerate}
\item The question must explicitly pose a \textbf{specific problem} or \textbf{ask a clear question} that can be answered or calculated. If the question does not clearly ask something (e.g., if it is vague, incomplete, or does not directly ask for a specific answer), it should be considered \textbf{invalid}. \\A valid question should directly require an answer, such as asking for a numerical value or selection (e.g., "What is the value of X?", "Which of the following is correct?", etc.). Questions that lack a clear problem or don't ask for a specific answer must be marked invalid.
\item The question, options, and answer must be fully defined. If any part is missing or unclear (e.g., if the answer does not match any of the listed options), the question should be deemed invalid. Note that the number of options is not a factor in determining completeness—having only one option is acceptable as long as the content is complete and coherent.
\item The question, options, and answer must be directly relevant to each other. The options should not reference each other, and there should be no circular dependencies or inter-referencing between options (e.g., “Option A is true because Option B is false”). Each option must stand independently, with no cross-references to other options.
\item Negative phrasing such as “The following options are incorrect” or “None of the above” is not allowed in the question. The question should use positive phrasing (e.g., “Which of the following is correct?”) to avoid confusion. If negative phrasing is detected in the question, it should be deemed invalid.
\item The question must be in a valid multiple-choice format. If the question is not suitable for a multiple-choice format (e.g., it is a free-form answer question like a problem or essay), it should be deemed invalid. Ensure that the question is designed specifically for multiple-choice answering, rather than being a question that requires an open-ended response.
\end{enumerate}

\vspace{0.5em}  
Please return the result in the following JSON format:

\begin{verbatim}
    {
    "is_valid": true/false
    }
\end{verbatim}
\vspace{-0.5em}  
Ensure that the output is valid JSON format and do not return any unrelated information.

Input:
\begin{verbatim}
    {
    question: "{}",
    options: {},
    answer: "{}"
    }
\end{verbatim}
Output: 
\end{promptbox}

\subsubsection{Negative and Extreme Inquiry Detection}
\begin{purposebox}{Purpose}
\begin{itemize}
    \item Identify negative questions that may appear in the options, such as 'Which of the following options is incorrect?' or 'Which of the following is not correct?'.
    \item Also, identify questions that use "most" to indicate a higher degree of likelihood or preference, such as 'Which of the following is most likely to occur?' or 'Which option is most appropriate?'.
    \item These types of questions are prone to having distractors that meet the conditions.
\end{itemize}
\end{purposebox}

\begin{promptbox}{Prompt 1}
Please evaluate whether the following text is a valid and well-formed multiple-choice question, adhering to the following criteria:
\begin{enumerate}
\item The question must not be a negation (e.g., "Which of the following is NOT...?" or "What is not...?").
\item The question should avoid vague or ambiguous phrasing, such as "Which of the following is the best/worst...?" or other uncertain expressions.
\item The answer choices should not be all affirmations or all negations, such as "All of the above" or "None of the above," as these are considered inappropriate.
\end{enumerate}

Please return the result in the following JSON format:

\begin{verbatim}
    {
    "is_valid": true/false
    }
\end{verbatim}

Ensure that the output is valid JSON and do not return any unrelated information.

Example input 1:

\begin{verbatim}
    {
    question: Which of the following procedure is not done in CHC?,
    options: ["Aboion", "Blood transfusion", "Caesaran section", 
            "Urine microscopy and culture sensitivity"],
    answer: Urine microscopy and culture sensitivity
    }
\end{verbatim}

Example output 1:

\begin{verbatim}
    {
    "is_valid": false
    }
\end{verbatim}
Example input 2:

\begin{verbatim}
    {
    question: Most common viral cause of acquired aqueductal stenosis 
    is?,
    options: ["Rubella", "Mumps", "Toxoplasma", "Enterovirus"],
    answer: Mumps
    }
\end{verbatim}
Example output 2:

\begin{verbatim}
    {
    "is_valid": false
    }
\end{verbatim}
Example input 3:

\begin{verbatim}
    {
    question: Acute cerebral edema occurs at high altitude due to,
    options: ["apillary blood pressure",     
              "Cerebral arteriolar dilation", 
              "Hypoxic damage to capillary 
               walls leading to capillary 
               leak",
              "All of the above"],
    answer: All of the above
    }
\end{verbatim}
Example output 3:

\begin{verbatim}
    {
    "is_valid": false
    }
\end{verbatim}
Example input 4:

\begin{verbatim}
    {
    question: Rideal Walker test is used to determine the efficiency 
    of the,
    options: ["Disinfectant", 
              "Moist heat sterilisation",       
              "Antibiotics", 
              "Dry heat sterilization"],
    answer: Disinfectant
    }
\end{verbatim}
Example output 4:

\begin{verbatim}
    {
    "is_valid": true
    }
\end{verbatim}

Input:
\begin{verbatim}
    {
    "question": {},
    "options": {},
    "answer": {}
    }
\end{verbatim}

Output:

\end{promptbox}

\begin{promptbox}{Prompt 2}
\subsection*{Input Description:}
You will be given:
\begin{itemize}
    \item A question statement (Question).
\end{itemize}

\subsection*{Guidelines:}
\begin{enumerate}
    \item Analyze the final part of the question (the actual question being asked) to check for specific phrases:
    \begin{itemize}
        \item \texttt{"incorrect"} in a context like \texttt{"which of the following is incorrect"}
        \item \texttt{"most"} in a context like \texttt{"which is most likely"}
    \end{itemize}
    
    \item If either of these phrases appears in the question's final asking part, output \texttt{true}, followed by a brief explanation of why the phrase is present.
    
    \item If neither phrase is found in the question's final part, output \texttt{false}, followed by a brief explanation.
\end{enumerate}

\subsection*{Output Format:}
Provide a response with:
\begin{itemize}
    \item \texttt{true} or \texttt{false} as the first part of the output
    \item A brief explanation that justifies your answer
\end{itemize}

\subsection*{Example Input:}
\textbf{Question:} \\
"A new technology is introduced in a manufacturing plant. Which of the following is most likely to occur?"

\subsection*{Example Output:}
\begin{verbatim}
{
    "is_valid": true/false,
}
\end{verbatim}
\textbf{Explanation:} Brief explanation of why the phrase is present.

\subsection*{Question:}
\begin{verbatim}
{}
\end{verbatim}

\subsection*{Output:}
\begin{verbatim}
{
    "is_valid": true/false,
}
\end{verbatim}

\end{promptbox}

\subsubsection{Multimodal Exclusion}
\begin{purposebox}{Purpose}
\begin{itemize}
    \item Check if the question relies on multimodal image information, and remove questions involving image information.
\end{itemize}
\end{purposebox}

\begin{promptbox}{Prompt}
You will be given a question or problem statement. Please determine if it strictly requires visual/image input to be solved.

\vspace{0.5em}  
Only output `true` if the question absolutely cannot be solved without an image (e.g. "What color is the car in this image?", "Describe the graph shown").

\vspace{0.5em}  
For all other cases where the question could potentially be answered with just text, output `false` (e.g. math problems, logic puzzles, verbal descriptions).

\vspace{0.5em}  
Output format:
\begin{itemize}
\item If visual input is strictly required: `true` + brief reason
\item Otherwise: `false` + brief reason
\end{itemize}

Question:
\begin{verbatim}
{}
\end{verbatim}
\end{promptbox}

\subsubsection{Field Relevance Evaluation}
\begin{purposebox}{Purpose}
\begin{itemize}
\item Assess whether the classification labels are appropriate.
\item Evaluate the classification level by level:
\begin{itemize}[label=\textbullet]
\item First, check if the "discipline" is relevant to the question.
\item If the "discipline" is appropriate, assess whether the "field" is relevant.
\item If the "field" is suitable, evaluate whether the "subfield" is relevant.
\item If the "subfield" is highly relevant to the question, the entire classification ("discipline," "field," and "subfield") is considered appropriate, even if there are minor mismatches in the "discipline" or "field."
\item If no levels meet the above criteria, or if any level other than "subfield" is deemed inappropriate and the "subfield" is not highly relevant, the classification is considered not relevant.
\end{itemize}
\end{itemize}
\end{purposebox}

\begin{promptbox}{Prompt}
You are an expert in hierarchical classification. Based on the given question and the provided three-level classification structure (\texttt{discipline}, \texttt{field}, \texttt{subfield}), evaluate whether the classification assigned to the question is appropriate.

\subsection*{Classification Structure:}
\begin{verbatim}
{
    "discipline": "{}",
    "field": "{}",
    "subfield": "{}"
}
\end{verbatim}

\subsection*{Question:}
\begin{verbatim}
{}
\end{verbatim}

\subsection*{Instructions:}
\begin{enumerate}
    \item Evaluate the classification level by level:
    \begin{itemize}
        \item First, check if the \texttt{discipline} is relevant to the question.
        \item If \texttt{discipline} is appropriate, evaluate if the \texttt{field} is relevant.
        \item If \texttt{field} is appropriate, evaluate if the \texttt{subfield} is relevant.
    \end{itemize}
    
    \item Special rule:
    \begin{itemize}
        \item If the \texttt{subfield} is found to be \textbf{highly relevant} to the question, the entire classification (\texttt{discipline}, \texttt{field}, and \texttt{subfield}) is considered appropriate, regardless of minor mismatches in \texttt{discipline} or \texttt{field}.
    \end{itemize}
    
    \item If no levels meet the above criteria, or if any level other than \texttt{subfield} is deemed inappropriate without \texttt{subfield} being highly relevant, the classification is considered not relevant.
    
    \item Output \textbf{strictly and exclusively} in the following JSON format and don't output any explanation, just output JSON:
    \begin{verbatim}
    {
        "is_relevant": true/false
    }
    \end{verbatim}
\end{enumerate}
\end{promptbox}

\subsubsection{Completeness Assessment}

\begin{purposebox}{Purpose}
\begin{itemize}
    \item Determine if the question is solvable, considering the following two dimensions:
    \begin{itemize}[label=\textbullet]
        \item Confidence: The level of confidence in the answer, categorized as "High," "Medium," or "Low."
        \item Missing Information: If an accurate answer cannot be provided, assess whether it is due to missing information (e.g., diagrams, formulas, known conditions) that makes the problem unsolvable, or if it is due to other reasons (e.g., high difficulty or uncertainty).
    \end{itemize}
\end{itemize}
\end{purposebox}

\begin{promptbox}{Prompt}
Please review and solve the following problems:

\begin{enumerate}
    \item Attempt to solve the problem and assess your confidence level in the answer.
    
    \item If you cannot provide an accurate answer, evaluate whether it is due to missing information (e.g., diagrams, formulas, known conditions) that makes the problem unsolvable, or if it's due to other reasons (e.g., high difficulty or uncertainty in solving).
    
    \item Output format:
    
    \begin{verbatim}
{
    "final_answer_letter": "A",
    "confidence": "High",
    "missing_info": false
}
    \end{verbatim}
\end{enumerate}

\subsection*{Rules:}
\begin{itemize}
    \item If the problem is missing a "diagram" (such as a geometry question requiring visual input), mark it as \texttt{missing\_info: true} and explain that the absence of the diagram or visual representation makes the question unsolvable.
    
    \item If the problem contains sufficient information but is difficult (e.g., requiring complex reasoning or having multiple possible solutions), mark \texttt{missing\_info: false} and assign a confidence level based on the model's reasoning process:
    \begin{itemize}
        \item \texttt{"High"}: If the problem is simple and straightforward, with a clear solution.
        \item \texttt{"Medium"}: If the problem involves multiple steps or requires further reasoning.
        \item \texttt{"Low"}: If the problem is complex, involves extensive information, or the model is uncertain about the solution.
    \end{itemize}
    
    \item If the problem is missing key necessary information (such as numbers, formulas, or critical conditions), mark it as \texttt{missing\_info: true}.
\end{itemize}

\subsection*{Output Format:}
\begin{verbatim}
    # The final answer option letter (e.g., A, B, C...)
    final_answer_letter: "A" 
    # Confidence level (High, Medium, Low)
    confidence: "High" 
    # Whether key information is missing, boolean (true or false)
    missing_info: false 
\end{verbatim}

\subsection*{Examples:}
\begin{itemize}
    \item \textbf{Question:} "Calculate the difference between A and B."
    \begin{itemize}
        \item \texttt{missing\_info: true}
        \item \texttt{explanation:} "Missing the values of A and B, so it cannot be calculated."
        \item \texttt{answer:} "A"
        \item \texttt{confidence:} "Low"
    \end{itemize}
    
    \item \textbf{Question:} "Given a geometric shape, calculate its area."
    \begin{itemize}
        \item \texttt{missing\_info: true}
        \item \texttt{explanation:} "The diagram is missing, so it cannot be solved."
        \item \texttt{answer:} "B"
        \item \texttt{confidence:} "Low"
    \end{itemize}
    
    \item \textbf{Question:} "Given the radius of a circle as 5, calculate the area of the circle."
    \begin{itemize}
        \item \texttt{missing\_info: false}
        \item \texttt{explanation:} "The problem is complete and solvable, high confidence."
        \item \texttt{answer:} "A"
        \item \texttt{confidence:} "High"
    \end{itemize}
\end{itemize}

\subsection*{Question:}
\begin{verbatim}
{}
\end{verbatim}

\subsection*{Output:}
\begin{verbatim}
    final_answer_letter: "A"
    confidence: "High"
    missing_info: false
\end{verbatim}

\end{promptbox}

\subsection{Manual Quality Review}
\label{Appendix: Manual Quality Review}

\autoref{tab: human} outlines the manual review process, focusing on ensuring consistency, clarity, accuracy, global perspective, and appropriate difficulty, while validating the correctness and uniqueness of answers and proper categorization.

\begin{table}[H]
\centering
\resizebox{\textwidth}{!}{
\begin{tabular}{llp{12cm}}
    \toprule 
    \textbf{Category} & \textbf{Focus} & \textbf{Description} \\
    \midrule 
    \multirow{13}{*}{\raisebox{-10ex}{\raisebox{-1ex}{\includegraphics[width=0.5cm]{pic/review/question.png}} \quad \textbf{Question}}} & \multirow{1}{*}{Source Consistency} & Align question source with annotations. \\
    \cmidrule(lr){3-3} 
    \cmidrule(lr){2-3} 
    & \multirow{2}{*}{Condition Completeness} & Base answers solely on provided text. \\
    \cmidrule(lr){3-3} 
    & & Avoid needing additional conditions like charts or legal clauses. \\
    \cmidrule(lr){2-3} 
    & \multirow{3}{*}{Question Clarity} & Be clear and specific. \\
    \cmidrule(lr){3-3} 
    & & Have only one correct answer. \\
    \cmidrule(lr){3-3} 
    & & Avoid vague or open-ended questions. \\
    \cmidrule(lr){2-3} 
    & \multirow{3}{*}{Expression Accuracy} & Write in English. \\
    \cmidrule(lr){3-3} 
    & & Use precise terminology. \\
    \cmidrule(lr){3-3} 
    & & Accurately translate specialized terms. \\
    \cmidrule(lr){2-3} 
    & \multirow{2}{*}{Formula and Numerical Accuracy} & Write formulas in LaTeX. \\
    \cmidrule(lr){3-3} 
    & & Align key formulas and numerical details. \\
    \cmidrule(lr){2-3} 
    & \multirow{2}{*}{Global Perspective} & Frame from a global perspective. \\
    \cmidrule(lr){3-3} 
    & & Avoid regional bias. \\
    \midrule 
    \multirow{6}{*}{\raisebox{-5ex}{\raisebox{-1ex}{\includegraphics[width=0.5cm]{pic/review/options.png}} \quad \textbf{Options}}} & \multirow{1}{*}{Expression Accuracy} & Accurately translate proper nouns and terminology. \\
    \cmidrule(lr){3-3} 
    \cmidrule(lr){2-3} 
    & \multirow{2}{*}{Formula and Numerical Accuracy} & Ensure formula and numerical accuracy. \\
    \cmidrule(lr){3-3} 
    & & Correctly format in LaTeX. \\
    \cmidrule(lr){2-3} 
    & \multirow{3}{*}{Distractor Relevance} & Make incorrect options meaningful but not correct. \\
    \cmidrule(lr){3-3} 
    & & Distinguish from correct answers. \\
    \cmidrule(lr){3-3} 
    & & Match correct options in length and format. \\
    \midrule 
    \multirow{4}{*}{\raisebox{-3ex}{\raisebox{-1ex}{\includegraphics[width=0.5cm]{pic/review/answer.png}} \quad \textbf{Answer}}} & \multirow{2}{*}{Answer Correctness} & Ensure the answer is in the options. \\
    \cmidrule(lr){3-3} 
    & & Accurately answers the question. \\
    \cmidrule(lr){2-3} 
    & \multirow{2}{*}{Answer Uniqueness} & Ensure the answer is unique. \\
    \cmidrule(lr){3-3} 
    & & Avoid multiple valid answers due to vague wording or incorrect options. \\
    \midrule 
    \multirow{2}{*}{\raisebox{-1ex}{\raisebox{-1ex}{\includegraphics[width=0.5cm]{pic/review/speedometer.png}} \quad \textbf{Difficulty}}} & \multirow{1}{*}{Difficulty Matching} & Match difficulty to graduate-level. \\
    \cmidrule(lr){3-3} 
    \cmidrule(lr){2-3} 
    & \multirow{1}{*}{Difficulty Reasonableness} & Ensure difficulty is reasonable based on complexity and presented options. \\
    \midrule 
    \multirow{3}{*}{\raisebox{-2ex}{\raisebox{-1ex}{\includegraphics[width=0.5cm]{pic/review/book.png}} \quad \textbf{Discipline}}} & \multirow{1}{*}{Category Accuracy} & Categorize correctly with accurate primary and secondary subjects. \\
    \cmidrule(lr){2-3} 
    & \multirow{2}{*}{Category Validity} & Correctly fill category fields. \\
    \cmidrule(lr){3-3} 
    & & No missing or unnecessary fields. \\
    \bottomrule 
\end{tabular}
}
\caption{Detailed Guidelines for Manual Quality Review.}
\label{tab: human}
\end{table}

\vspace{-5mm}
\newtcolorbox[auto counter, number within=section]{questionbox}[2][]{%
  colback=white, 
  colframe=blue!60!black,  
  width=\textwidth,
  arc=2mm, 
  boxrule=0.5mm, 
  title={\normalsize\faLeaf\hspace{0.5em}#2},
  breakable, 
  fonttitle=\bfseries\Large, 
  fontupper=\small
  #1
}

\newtcolorbox[auto counter, number within=section]{errorbox}[2][]{%
    colback=white,
    colframe=red!50!black,  
    width=\textwidth,
     arc=2mm, 
    boxrule=0.5mm, 
    title={\normalsize\faExclamationTriangle\hspace{0.5em}#2}, 
    breakable, 
    fonttitle=\bfseries\Large, 
    fontupper=\small
    #1
}

\subsection{Reasons for Failing Quality Inspection}

\begin{questionbox}{Question Block 1}
\noindent \textbf{Question:} Consider a resistor made from a hollow cylinder of carbon as shown below. The inner radius of the cylinder is $R_i=0.2$mm and the outer radius is $R_o=0.3$mm. The length of the resistor is $L=0.9$mm. The resistivity of the carbon is $\rho=3.5 \times 10^{-5} \ \Omega \cdot m$. What is the resistance?

\bigskip
\noindent \textbf{Options:}
\begin{itemize}
    \item 3.0
    \item 1.0
    \item 5.0
    \item 2.0
    \item 1.5
    \item 2.8
    \item 3.5
    \item 4.0
    \item 2.5
    \item 1.8
\end{itemize}

\noindent \textbf{Answer:} 2.5
\end{questionbox}

\begin{errorbox}{Error Analysis Block 1}
\noindent \textbf{Error Type:} Condition Setting Defect

\noindent \textbf{Error Analysis:} The solution to the correction problem requires a diagram as a condition; without the diagram, the problem cannot be solved. Therefore, this question lacks condition setting and cannot be answered.
\end{errorbox}

\begin{questionbox}{Question Block 2}
\noindent \textbf{Question:} Which of the following descriptions about the function of gastrointestinal motility is incorrect?

\bigskip
\noindent \textbf{Options:}
\begin{itemize}
    \item Controlling the release of bile from the gallbladder
    \item Facilitating the absorption of nutrients into the bloodstream
    \item Assisting in the immune response of the gut
    \item Regulating the pace of digestion
    \item Stimulating enzyme production in the pancreas
    \item Breaking down large food molecules into smaller molecules
    \item Mixing food with digestive juices
    \item Propelling food through the digestive tract
    \item Coordinating muscle contractions in the digestive system
    \item Maintaining the pH balance in the stomach
\end{itemize}

\noindent \textbf{Answer:} Breaking down large food molecules into smaller molecules
\end{questionbox}

\begin{errorbox}{Error Analysis Block 2}
\noindent \textbf{Error Type:} Incorrect option construction

\noindent \textbf{Error Analysis:} Options such as stimulating enzyme production in the pancreas, maintaining the pH balance in the stomach, and several other choices are correct answers to the question. Questions involving terms like "incorrect" or "most" require careful construction of distractors, but the reviewer fails to ensure the uniqueness of the correct answer.
\end{errorbox}

\begin{questionbox}{Question Block 3}
\noindent \textbf{Question:} Ross claims that we learn of our prima facie duties:

\bigskip
\noindent \textbf{Options:}
\begin{itemize}
    \item from the moral judgments we make in various situations.
    \item by seeing the prima facie rightness of particular acts, and then apprehending general principles.
    \item from the explicit moral instruction we receive as children.
    \item through societal norms and cultural values.
    \item from religious teachings or scriptures.
    \item by observing the consequences of our actions.
    \item by intuitively understanding moral obligations.
    \item by proving them philosophically.
    \item by apprehending general principles, and then inferring the prima facie rightness of particular acts.
    \item through legal regulations and laws.
\end{itemize}

\noindent \textbf{Answer:} by seeing the prima facie rightness of particular acts, and then apprehending general principles
\end{questionbox}

\begin{errorbox}{Error Analysis Block 3}
\noindent \textbf{Error Type:} Missing Contextual Information

\noindent \textbf{Error Analysis:} The problem does not provide sufficient contextual information, such as Ross's background or theories, to allow for proper comprehension and answer selection.
\end{errorbox}

\begin{questionbox}{Question Block 4}
\noindent \textbf{Question:} About the I-type system tracking the step signal, what's the steady-state error between them?

\bigskip
\noindent \textbf{Options:}
\begin{itemize}
    \item 10
    \item infinity
    \item 2
    \item -0.5
    \item 1
    \item 0.5
    \item undefined
    \item 0
    \item -1
    \item 100
\end{itemize}

\noindent \textbf{Answer:} 0
\end{questionbox}

\begin{errorbox}{Error Analysis Block 4}
\noindent \textbf{Error Type:} Distractor Quality Issue

\noindent \textbf{Error Analysis:} Some options, such as "infinity" and "undefined," may not be relevant or may be poorly constructed. These distractors may not effectively contribute to creating meaningful confusion.
\end{errorbox}

\begin{questionbox}{Question Block 5}
\noindent \textbf{Question:} A cube of iron was heated to 70°C and transferred to a beaker containing 100g of water at 20°C. The final temperature of the water and the iron was 23°C. What is the heat capacity?

\bigskip
\noindent \textbf{Options:}
\begin{itemize}
    \item $0.310 \ K^{-1} \ g^{-1}$
    \item $0.740 \ K^{-1} \ g^{-1}$
    \item $0.582 \ K^{-1} \ g^{-1}$
    \item $1.200 \ K^{-1} \ g^{-1}$
    \item $0.453 \ K^{-1} \ g^{-1}$
    \item $0.235 \ K^{-1} \ g^{-1}$
    \item $1.004 \ K^{-1} \ g^{-1}$
    \item $0.505 \ K^{-1} \ g^{-1}$
    \item $0.875 \ K^{-1} \ g^{-1}$
    \item $0.690 \ K^{-1} \ g^{-1}$
\end{itemize}

\noindent \textbf{Answer:} $0.453 \ K^{-1} \ g^{-1}$
\end{questionbox}

\begin{errorbox}{Error Analysis Block 5}
\noindent \textbf{Error Type:} Condition Setting Defect

\noindent \textbf{Error Analysis:} The problem lacks critical information about the mass of the iron cube, which makes it unsolvable.
\end{errorbox}

\begin{questionbox}{Question Block 6}
\noindent \textbf{Question:} The Central Committee of the Communist Party of China and the Central Government are striving to enhance our country’s military security functions. In order to achieve this, they should further promote the governance of the military according to law and strict discipline, focusing on the crucial ( ).

\bigskip
\noindent \textbf{Options:}
\begin{itemize}
    \item Checks and Balances on the Exercise of Power
    \item Procedures and Control of Power
    \item Maintenance and Supervision of Power
    \item Regulation and Oversight of Authority
    \item Coordination and Monitoring of Authority
    \item Systems and Regulation of Governance
    \item Security and Management of Authority
    \item Power Operation Management
    \item Power Operation Guarantee
    \item Power Operation Coordination
\end{itemize}

\noindent \textbf{Answer:} Checks and Balances on the Exercise of Power
\end{questionbox}

\begin{errorbox}{Error Analysis Block 6}
\noindent \textbf{Error Type:} Region-Specific Context Missing

\noindent \textbf{Error Analysis:} The question has a clear regional context (China) that should be accounted for in item selection or modification. Additionally, the item could be better adapted or refined to enhance clarity and relevance.
\end{errorbox}

\begin{questionbox}{Question Block 7}
\noindent \textbf{Question:} Use the example below to answer the question that follows.  
Identify the type of dissonance found above the asterisk in the given voice-leading example.

\bigskip
\noindent \textbf{Options:}
\begin{itemize}
    \item Neighbor tone
    \item Appoggiatura
    \item Suspension
    \item Passing tone
\end{itemize}

\noindent \textbf{Answer:} Passing tone
\end{questionbox}

\begin{errorbox}{Error Analysis Block 7}
\noindent \textbf{Error Type:} Missing Contextual Information

\noindent \textbf{Error Analysis:} The question lacks the necessary example (voice-leading notation) and cannot be answered.
\end{errorbox}

\begin{questionbox}{Question Block 8}
\noindent \textbf{Question:} Suppose an electric field $\mathbf{E}(x, y, z)$ has the form  
$$  
E_{x} = ax, \quad E_{y} = 0, \quad E_{z} = 0  
$$  
where $a$ is a constant. What is the charge density? How do you account for the fact that the field points in a particular direction, when the charge density is uniform?

\bigskip
\noindent \textbf{Options:}
\begin{itemize}
    \item $\epsilon_{0} x$
    \item $\epsilon_{0} y$
    \item $x a \epsilon$
    \item $\epsilon_{0} a$
    \item $x \epsilon_{0}$
    \item $\epsilon_{z} a$
    \item $\epsilon_{0} z$
    \item $a \epsilon_{y}$
    \item $\epsilon a_{y}$
    \item $\epsilon_{x} a$
\end{itemize}

\noindent \textbf{Answer:} $\epsilon_{0} a$
\end{questionbox}

\begin{errorbox}{Error Analysis Block 8}
\noindent \textbf{Error Type:} Incomplete Question and Answer Construction

\noindent \textbf{Error Analysis:} The question contains two sub-questions, but the answer and options only address one, causing ambiguity.
\end{errorbox}

\section{Quantitative Overview of Disciplines at Three Levels}

{
\renewcommand{\arraystretch}{1
}
\begin{table}[H]
\centering
\resizebox{1\textwidth}{!}{
\begin{tabular}{>{\raggedright\arraybackslash}p{3cm}>{\raggedright\arraybackslash}p{6cm}>{\raggedright\arraybackslash}p{10cm}|>{\raggedright\arraybackslash}p{3cm}>{\raggedright\arraybackslash}p{6cm}>{\raggedright\arraybackslash}p{10cm}}
\toprule
\textbf{Discipline} & \textbf{Field} & \textbf{Subfield} & \textbf{Discipline} & \textbf{Field} & \textbf{Subfield} \\
\midrule
\cellcolor{teal!20} \textbf{Agronomy(485)}  &  \cellcolor{teal!10} Animal Husbandry(103) &  \cellcolor{teal!5} Animal Nutrition and Feed Science(52) &  \cellcolor{blue!20}  \textbf{Engineering(7892)} &  \cellcolor{blue!10} Geological Resources and Geological Engineering(50) &  \cellcolor{blue!5} Geological Resources and Geological Engineering(50) \\
\cline{3-3}
\cline{5-6}
\cellcolor{teal!20} & \cellcolor{teal!10} &  \cellcolor{teal!5} Animal Rearing and Breeding(51) &  \cellcolor{blue!20} & \cellcolor{blue!10} Hydraulic Engineering(218) &  \cellcolor{blue!5} Hydraulics and Hydrology(136) \\
\cline{2-3}
\cline{6-6}
\cellcolor{teal!20} &  \cellcolor{teal!10} Aquaculture(56) &  \cellcolor{teal!5} Aquaculture(56) &  \cellcolor{blue!20}   &  \cellcolor{blue!10} &  \cellcolor{blue!5} Water conservancy and Hydropower Engineering(82) \\
\cline{2-3}
\cline{5-6}
\cellcolor{teal!20} &  \cellcolor{teal!10} Crop Science(145) &  \cellcolor{teal!5} Crop Science(145) &  \cellcolor{blue!20}  &  \cellcolor{blue!10} Information and Communication Engineering(504) &  \cellcolor{blue!5} Antenna and Radio Communication(114) \\
\cline{2-3}
\cline{6-6}
\cellcolor{teal!20} &  \cellcolor{teal!10} Forestry(131) &  \cellcolor{teal!5} Forest Cultivation and Genetic Breeding(81) &  \cellcolor{blue!20}   &  \cellcolor{blue!10} &  \cellcolor{blue!5} Communication Principles(116) \\
\cline{3-3}
\cline{6-6}
\cellcolor{teal!20} & \cellcolor{teal!10} &  \cellcolor{teal!5} Landscape Plants and Ornamental Horticulture(50) &  \cellcolor{blue!20} & \cellcolor{blue!10} &  \cellcolor{blue!5} Communication and Information Systems(69) \\
\cline{2-3}
\cline{6-6}
\cellcolor{teal!20} &  \cellcolor{teal!10} Veterinary Medicine(50) &  \cellcolor{teal!5} Veterinary Medicine(50) &  \cellcolor{blue!20}   &  \cellcolor{blue!10} &  \cellcolor{blue!5} Optical Fiber Communication(50) \\
\hhline{---~~~}
\cline{6-6}
\cellcolor{pink!20} \textbf{Economics(873)}  & \cellcolor{pink!10} Applied Economics(723) &  \cellcolor{pink!5} Economic Statistics(50) &  \cellcolor{blue!20} & \cellcolor{blue!10} &  \cellcolor{blue!5} Signal and Information Processing(155) \\
\cline{3-3}
\cline{5-6}
\cellcolor{pink!20} & \cellcolor{pink!10} &  \cellcolor{pink!5} Finance(176) &  \cellcolor{blue!20} & \cellcolor{blue!10} Instrument Science and Technology(50) &  \cellcolor{blue!5} Instrument Science and Technology(50) \\
\cline{3-3}
\cline{5-6}
\cellcolor{pink!20} & \cellcolor{pink!10} &  \cellcolor{pink!5} Industrial Economics(84) &  \cellcolor{blue!20} & \cellcolor{blue!10} Materials Science and Engineering(289) &  \cellcolor{blue!5} Materials Physics and Chemistry(199) \\
\cline{3-3}
\cline{6-6}
\cellcolor{pink!20} & \cellcolor{pink!10} &  \cellcolor{pink!5} International Trade(69) &  \cellcolor{blue!20} & \cellcolor{blue!10} &  \cellcolor{blue!5} Materials Processing Engineering(90) \\
\cline{3-3}
\cline{5-6}
\cellcolor{pink!20} & \cellcolor{pink!10} &  \cellcolor{pink!5} Labor Economics(74) &  \cellcolor{blue!20} & \cellcolor{blue!10} Mechanical Engineering(176) &  \cellcolor{blue!5} Manufacturing Automation(126) \\
\cline{3-3}
\cline{6-6}
\cellcolor{pink!20} & \cellcolor{pink!10} &  \cellcolor{pink!5} National and Defense Economics(50) &  \cellcolor{blue!20} & \cellcolor{blue!10} &  \cellcolor{blue!5} Mechatronic Engineering(50) \\
\cline{3-3}
\cline{5-6}
\cellcolor{pink!20} & \cellcolor{pink!10} &  \cellcolor{pink!5} Public Finance(120) &  \cellcolor{blue!20} & \cellcolor{blue!10} Mechanics(908) &  \cellcolor{blue!5} Fundamentals of Dynamics and Control(347) \\
\cline{3-3}
\cline{6-6}
\cellcolor{pink!20} & \cellcolor{pink!10} &  \cellcolor{pink!5} Quantitative Economics(100) &  \cellcolor{blue!20} & \cellcolor{blue!10} &  \cellcolor{blue!5} Rigid Body Mechanics(173) \\
\cline{2-3}
\cline{6-6}
\cellcolor{pink!20} &  \cellcolor{pink!10} Theoretical Economics(150) &  \cellcolor{pink!5} Economic History(50) &  \cellcolor{blue!20}   &  \cellcolor{blue!10} &  \cellcolor{blue!5} Solid Mechanics(93) \\
\cline{3-3}
\cline{6-6}
\cellcolor{pink!20} & \cellcolor{pink!10} &  \cellcolor{pink!5} Political Economy(50) &  \cellcolor{blue!20} & \cellcolor{blue!10} &  \cellcolor{blue!5} Theoretical Fluid Mechanics(140) \\
\cline{3-3}
\cline{6-6}
\cellcolor{pink!20} & \cellcolor{pink!10} &  \cellcolor{pink!5} Western Economics(50) &  \cellcolor{blue!20} & \cellcolor{blue!10} &  \cellcolor{blue!5} Theoretical Mechanics(155) \\
\hhline{---~~~}
\cline{5-6}
\cellcolor{cyan!20} \textbf{Education(484)}  &  \cellcolor{cyan!10} Education(247) & \cellcolor{cyan!5}  Educational Technology and Principles(96) &  \cellcolor{blue!20}  &  \cellcolor{blue!10} Metallurgical Engineering(255) &  \cellcolor{blue!5} Iron and Steel Metallurgy(105) \\
\cline{3-3}
\cline{6-6}
\cellcolor{cyan!20} & \cellcolor{cyan!10} &  \cellcolor{cyan!5} Preschool Education(50) &  \cellcolor{blue!20} & \cellcolor{blue!10} &  \cellcolor{blue!5} Non-ferrous Metallurgy(50) \\
\cline{3-3}
\cline{6-6}
\cellcolor{cyan!20} & \cellcolor{cyan!10} &  \cellcolor{cyan!5} Special Education(50) &  \cellcolor{blue!20} & \cellcolor{blue!10} &  \cellcolor{blue!5} Physical Chemistry of Metallurgical Process(50) \\
\cline{3-3}
\cline{6-6}
\cellcolor{cyan!20} & \cellcolor{cyan!10} &  \cellcolor{cyan!5} Theory of Curriculum and Instruction(51) &  \cellcolor{blue!20} & \cellcolor{blue!10} &  \cellcolor{blue!5} Principles of Metallurgy(50) \\
\cline{2-3}
\cline{5-6}
\cellcolor{cyan!20} &  \cellcolor{cyan!10} Physical Education(150) &  \cellcolor{cyan!5} Physical Education and Training(50) &  \cellcolor{blue!20}  &  \cellcolor{blue!10} Mining Engineering(100) &  \cellcolor{blue!5} Mineral Processing Engineering(50) \\
\cline{3-3}
\cline{6-6}
\cellcolor{cyan!20} & \cellcolor{cyan!10} &  \cellcolor{cyan!5} Sports Humanities and Sociology(50) &  \cellcolor{blue!20} & \cellcolor{blue!10} &  \cellcolor{blue!5} Mining and Safety Engineering(50) \\
\cline{3-3}
\cline{5-6}
\cellcolor{cyan!20} & \cellcolor{cyan!10} &  \cellcolor{cyan!5} Sports Science and Medicine(50) &  \cellcolor{blue!20} & \cellcolor{blue!10} Naval Architecture and Ocean Engineering(138) &  \cellcolor{blue!5} Marine Engineering(50) \\
\cline{2-3}
\cline{6-6}
\cellcolor{cyan!20} &  \cellcolor{cyan!10} Psychology(87) &  \cellcolor{cyan!5} Psychology(87) &  \cellcolor{blue!20}   &  \cellcolor{blue!10} &  \cellcolor{blue!5} Ship Mechanics and Design Principles(88) \\
\hhline{---~~~}
\cline{5-6}
\cellcolor{blue!20} \textbf{Engineering(7892)}  &  \cellcolor{blue!10} Aeronautical and Astronautical Science and Technology(119) & \cellcolor{blue!5}  Aeronautical and Astronautical Science and Technology(119) &  \cellcolor{blue!20}  &  \cellcolor{blue!10} Nuclear Science and Technology(107) &  \cellcolor{blue!5} Nuclear Energy and Reactor Technology(57) \\
\cline{2-3}
\cline{6-6}
\cellcolor{blue!20} &  \cellcolor{blue!10} Agricultural Engineering(104) &  \cellcolor{blue!5} Agricultural Environment and Soil-Water Engineering(54) &  \cellcolor{blue!20}   &  \cellcolor{blue!10} &  \cellcolor{blue!5} Radiation Protection and Nuclear Technology Applications(50) \\
\cline{3-3}
\cline{5-6}
\cellcolor{blue!20} & \cellcolor{blue!10} &  \cellcolor{blue!5} Agricultural Mechanization Engineering(50) &  \cellcolor{blue!20} & \cellcolor{blue!10} Optical Engineering(376) &  \cellcolor{blue!5} Applied Optics(124) \\
\cline{2-3}
\cline{6-6}
\cellcolor{blue!20} &  \cellcolor{blue!10} Architecture(162) &  \cellcolor{blue!5} Architectural Design and Theory(50) &  \cellcolor{blue!20}   &  \cellcolor{blue!10} &  \cellcolor{blue!5} Laser Technology(50) \\
\cline{3-3}
\cline{6-6}
\cellcolor{blue!20} & \cellcolor{blue!10} &  \cellcolor{blue!5} Architectural History(62) &  \cellcolor{blue!20} & \cellcolor{blue!10} &  \cellcolor{blue!5} Optoelectronic Technology(67) \\
\cline{3-3}
\cline{6-6}
\cellcolor{blue!20} & \cellcolor{blue!10} &  \cellcolor{blue!5} Urban Planning and Design(50) &  \cellcolor{blue!20} & \cellcolor{blue!10} &  \cellcolor{blue!5} Theoretical Optics(135) \\
\cline{2-3}
\cline{5-6}
\cellcolor{blue!20} &  \cellcolor{blue!10} Chemical Engineering and Technology(410) &  \cellcolor{blue!5} Chemical Transport Engineering(50) &  \cellcolor{blue!20}  &  \cellcolor{blue!10} Petroleum and Natural Gas Engineering(112) &  \cellcolor{blue!5} Oil and Gas Field Development and Storage \& Transportation Engineering(62) \\
\cline{3-3}
\cline{6-6}
\cellcolor{blue!20} & \cellcolor{blue!10} &  \cellcolor{blue!5} Elements of Chemical Reaction Engineering(107) &  \cellcolor{blue!20} & \cellcolor{blue!10} &  \cellcolor{blue!5} Poromechanics and Reservoir Physics(50) \\
\cline{3-3}
\cline{5-6}
\cellcolor{blue!20} & \cellcolor{blue!10} &  \cellcolor{blue!5} Fluid Flow and Heat Transfer in Chemical Engineering(107) &  \cellcolor{blue!20} & \cellcolor{blue!10} Power Engineering and Engineering Thermophysics(684) &  \cellcolor{blue!5} Engineering Fluid Mechanics(84) \\
\cline{3-3}
\cline{6-6}
\cellcolor{blue!20} & \cellcolor{blue!10} &  \cellcolor{blue!5} Mass Transport and Separation Process in Chemical Engineering(146) &  \cellcolor{blue!20} & \cellcolor{blue!10} &  \cellcolor{blue!5} Engineering Thermophysics(146) \\
\cline{2-3}
\cline{6-6}
\cellcolor{blue!20} &  \cellcolor{blue!10} Civil Engineering(358) &  \cellcolor{blue!5} Bridge and Tunnel Engineering(51) &  \cellcolor{blue!20}   &  \cellcolor{blue!10} &  \cellcolor{blue!5} Fluid Machinery and Engineering(64) \\
\cline{3-3}
\cline{6-6}
\cellcolor{blue!20} & \cellcolor{blue!10} &  \cellcolor{blue!5} Geotechnical Engineering(179) &  \cellcolor{blue!20} & \cellcolor{blue!10} &  \cellcolor{blue!5} Heat Transfer(129) \\
\cline{3-3}
\cline{6-6}
\cellcolor{blue!20} & \cellcolor{blue!10} &  \cellcolor{blue!5} Structural Engineering(57) &  \cellcolor{blue!20} & \cellcolor{blue!10} &  \cellcolor{blue!5} Internal Combustion Engineering(50) \\
\cline{3-3}
\cline{6-6}
\cellcolor{blue!20} & \cellcolor{blue!10} &  \cellcolor{blue!5} Urban Infrastructure Engineering(71) &  \cellcolor{blue!20} & \cellcolor{blue!10} &  \cellcolor{blue!5} Power Machinery and Engineering(52) \\
\cline{2-3}
\cline{6-6}
\cellcolor{blue!20} &  \cellcolor{blue!10} Computer Science and Technology(763) &  \cellcolor{blue!5} Advanced Programming Languages(50) &  \cellcolor{blue!20}   &  \cellcolor{blue!10} &  \cellcolor{blue!5} Refrigeration and Cryogenic Engineering(50) \\
\cline{3-3}
\cline{6-6}
\cellcolor{blue!20} & \cellcolor{blue!10} &  \cellcolor{blue!5} Computer Architecture(50) &  \cellcolor{blue!20} & \cellcolor{blue!10} &  \cellcolor{blue!5} Thermal Energy Engineering(109) \\
\cline{3-3}
\cline{5-6}
\cellcolor{blue!20} & \cellcolor{blue!10} &  \cellcolor{blue!5} Computer Networks(59) &  \cellcolor{blue!20} & \cellcolor{blue!10} Surveying and Mapping Science and Technology(168) &  \cellcolor{blue!5} Cartography and Geographic Information Engineering(50) \\
\cline{3-3}
\cline{6-6}
\cellcolor{blue!20} & \cellcolor{blue!10} &  \cellcolor{blue!5} Computer Software and Theory(79) &  \cellcolor{blue!20} & \cellcolor{blue!10} &  \cellcolor{blue!5} Digital Surveying and Remote Sensing Applications(68) \\
\cline{3-3}
\cline{6-6}
\cellcolor{blue!20} & \cellcolor{blue!10} &  \cellcolor{blue!5} Data Structures(99) &  \cellcolor{blue!20} & \cellcolor{blue!10} &  \cellcolor{blue!5} Geodesy and Surveying Engineering(50) \\
\cline{3-3}
\cline{5-6}
\cellcolor{blue!20} & \cellcolor{blue!10} &  \cellcolor{blue!5} Databases(157) &  \cellcolor{blue!20} & \cellcolor{blue!10} Textile Science and Engineering(100) &  \cellcolor{blue!5} Textile Chemistry and Dyeing Engineering(50) \\
\cline{3-3}
\cline{6-6}
\cellcolor{blue!20} & \cellcolor{blue!10} &  \cellcolor{blue!5} Formal Languages(52) &  \cellcolor{blue!20} & \cellcolor{blue!10} &  \cellcolor{blue!5} Textile Materials Science(50) \\
\cline{3-3}
\cline{5-6}
\cellcolor{blue!20} & \cellcolor{blue!10} &  \cellcolor{blue!5} Operating Systems(85) &  \cellcolor{blue!20} & \cellcolor{blue!10} Transportation Engineering(251) &  \cellcolor{blue!5} Road and Railway Engineering(50) \\
\cline{3-3}
\cline{6-6}
\cellcolor{blue!20} & \cellcolor{blue!10} &  \cellcolor{blue!5} Pattern Recognition(50) &  \cellcolor{blue!20} & \cellcolor{blue!10} &  \cellcolor{blue!5} Traffic Information Engineering and Control(65) \\
\cline{3-3}
\cline{6-6}
\cellcolor{blue!20} & \cellcolor{blue!10} &  \cellcolor{blue!5} Principles of Computer Organization(82) &  \cellcolor{blue!20} & \cellcolor{blue!10} &  \cellcolor{blue!5} Transportation Planning and Management(86) \\
\cline{2-3}
\cline{6-6}
\cellcolor{blue!20} &  \cellcolor{blue!10} Control Science and Engineering(190) &  \cellcolor{blue!5} Control Theory and Control Engineering(89) &  \cellcolor{blue!20}   &  \cellcolor{blue!10} &  \cellcolor{blue!5} Vehicle Operation Engineering(50) \\
\cline{3-3}
\cline{5-6}
\cellcolor{blue!20} & \cellcolor{blue!10} &  \cellcolor{blue!5} Guidance, Navigation and Control(51) &  \cellcolor{blue!20} & \cellcolor{blue!10} Weapon Science and Technology(100) &  \cellcolor{blue!5} Military Chemistry and Pyrotechnics(50) \\
\cline{3-3}
\cline{6-6}
\cellcolor{blue!20} & \cellcolor{blue!10} &  \cellcolor{blue!5} Operations Research and Cybernetics(50) &  \cellcolor{blue!20} & \cellcolor{blue!10} &  \cellcolor{blue!5} Weapon Systems Science and Engineering(50) \\
\cline{2-3}
\hhline{~~~---}
\cellcolor{blue!20} &  \cellcolor{blue!10} Electrical Engineering(556) &  \cellcolor{blue!5} Electrical Theory and New Technologies(190) &  \cellcolor{brown!20}  \textbf{History(674)} &  \cellcolor{brown!10} History(674) &  \cellcolor{brown!5} Archaeology and Museology(104) \\
\cline{3-3}
\cline{6-6}
\cellcolor{blue!20} & \cellcolor{blue!10} &  \cellcolor{blue!5} High Voltage and Insulation Technology(88) &  \cellcolor{brown!20} & \cellcolor{brown!10} &  \cellcolor{brown!5} Historical Geography(142) \\
\cline{3-3}
\cline{6-6}
\cellcolor{blue!20} & \cellcolor{blue!10} &  \cellcolor{blue!5} Power Electronics and Electrical Drives(182) &  \cellcolor{brown!20} & \cellcolor{brown!10} &  \cellcolor{brown!5} World History(428) \\
\cline{3-3}
\hhline{~~~---}
\cellcolor{blue!20} & \cellcolor{blue!10} &  \cellcolor{blue!5} Power Systems and Automation(96) &  \cellcolor{red!20}  \textbf{Law(656)} &  \cellcolor{red!10} Law(591) &  \cellcolor{red!5}  Civil and Commercial Law(118) \\
\cline{2-3}
\cline{6-6}
\cellcolor{blue!20} &  \cellcolor{blue!10} Electronic Science and Technology(246) &  \cellcolor{blue!5} Circuits and Systems(108) &  \cellcolor{red!20}   &  \cellcolor{red!10} &  \cellcolor{red!5} Constitutional and Administrative Law(50) \\
\cline{3-3}
\cline{6-6}
\cellcolor{blue!20} & \cellcolor{blue!10} &  \cellcolor{blue!5} Electromagnetic Field and Microwave Technology(88) &  \cellcolor{red!20} & \cellcolor{red!10} &  \cellcolor{red!5} Contract Law(50) \\
\cline{3-3}
\cline{6-6}
\cellcolor{blue!20} & \cellcolor{blue!10} &  \cellcolor{blue!5} Microelectronics and Solid-State Electronics(50) &  \cellcolor{red!20} & \cellcolor{red!10} &  \cellcolor{red!5} Criminal Law(116) \\
\cline{2-3}
\cline{6-6}
\cellcolor{blue!20} &  \cellcolor{blue!10} Environmental Science and Engineering(189) &  \cellcolor{blue!5} Environmental Engineering(89) &  \cellcolor{red!20}   &  \cellcolor{red!10} &  \cellcolor{red!5} International Law(57) \\
\cline{3-3}
\cline{6-6}
\cellcolor{blue!20} & \cellcolor{blue!10} &  \cellcolor{blue!5} Environmental Science(50) &  \cellcolor{red!20} & \cellcolor{red!10} &  \cellcolor{red!5} Law and Social Governance(50) \\
\cline{3-3}
\cline{6-6}
\cellcolor{blue!20} & \cellcolor{blue!10} &  \cellcolor{blue!5} Environmental and Resource Protection(50) &  \cellcolor{red!20} & \cellcolor{red!10} &  \cellcolor{red!5} Legal Theory and Legal History(50) \\
\cline{2-3}
\cline{6-6}
\cellcolor{blue!20} &  \cellcolor{blue!10} Food Science and Engineering(109) &  \cellcolor{blue!5} Food Biochemistry(50) &  \cellcolor{red!20}   &  \cellcolor{red!10} &  \cellcolor{red!5} Military Law(50) \\
\cline{3-3}
\cline{6-6}
\cellcolor{blue!20} & \cellcolor{blue!10} &  \cellcolor{blue!5} Food Processing and Storage Engineering(59) &  \cellcolor{red!20} & \cellcolor{red!10} &  \cellcolor{red!5} Procedural Law(50) \\
\cline{2-3}
\cline{5-6}
\cellcolor{blue!20} &  \cellcolor{blue!10} Forestry Engineering(100) &  \cellcolor{blue!5} Forest Engineering(50) &  \cellcolor{red!20}  &  \cellcolor{red!10} Political Science(65) &  \cellcolor{red!5} Political Science(65) \\
\cline{3-3}
\hhline{~~~---}
\cellcolor{blue!20} & \cellcolor{blue!10} &  \cellcolor{blue!5} Wood Science and Technology(50) &  \cellcolor{gray!20}  \textbf{-} &  \cellcolor{gray!10} - &  \cellcolor{gray!5}  - \\
\bottomrule
\end{tabular}
}
\caption{Comprehensive Overview of Three Levels of Disciplines and Their Sample Sizes.}
\end{table}

\begin{table}[H]
\centering
\ContinuedFloat
\resizebox{1\textwidth}{!}{
\begin{tabular}{>{\raggedright\arraybackslash}p{3cm}>{\raggedright\arraybackslash}p{6cm}>{\raggedright\arraybackslash}p{10cm}|>{\raggedright\arraybackslash}p{3cm}>{\raggedright\arraybackslash}p{6cm}>{\raggedright\arraybackslash}p{10cm}}
\toprule
\textbf{Discipline} & \textbf{Field} & \textbf{Subfield} & \textbf{Discipline} & \textbf{Field} & \textbf{Subfield} \\
\midrule
\cellcolor{purple!20} \textbf{Literature and Arts(1676)}  &  \cellcolor{purple!10} Art Studies(603) &  \cellcolor{purple!5} Broadcasting and Television Art(50) &  \cellcolor{lime!20}  \textbf{Philosophy(347)} &  \cellcolor{lime!10} Philosophy(347) &  \cellcolor{lime!5} Religious Studies(50) \\
\cline{3-3}
\hhline{~~~---}
\cellcolor{purple!20} & \cellcolor{purple!10} &  \cellcolor{purple!5} Dance Studies(50) &  \cellcolor{green!20}  \textbf{Science(9838)} &  \cellcolor{green!10} Astronomy(405) &  \cellcolor{green!5}  Astronomical Observation and Technology(100) \\
\cline{3-3}
\cline{6-6}
\cellcolor{purple!20} & \cellcolor{purple!10} &  \cellcolor{purple!5} Design Arts(62) &  \cellcolor{green!20} & \cellcolor{green!10} &  \cellcolor{green!5} Astrophysics(69) \\
\cline{3-3}
\cline{6-6}
\cellcolor{purple!20} & \cellcolor{purple!10} &  \cellcolor{purple!5} Drama and Opera Studies(82) &  \cellcolor{green!20} & \cellcolor{green!10} &  \cellcolor{green!5} Cosmology(78) \\
\cline{3-3}
\cline{6-6}
\cellcolor{purple!20} & \cellcolor{purple!10} &  \cellcolor{purple!5} Film Studies(158) &  \cellcolor{green!20} & \cellcolor{green!10} &  \cellcolor{green!5} Solar System Science(88) \\
\cline{3-3}
\cline{6-6}
\cellcolor{purple!20} & \cellcolor{purple!10} &  \cellcolor{purple!5} Fine Arts(201) &  \cellcolor{green!20} & \cellcolor{green!10} &  \cellcolor{green!5} Stellar and Interstellar Evolution(70) \\
\cline{2-3}
\cline{5-6}
\cellcolor{purple!20} &  \cellcolor{purple!10} Journalism and Communication(207) &  \cellcolor{purple!5} Communication and Broadcasting(58) &  \cellcolor{green!20}  &  \cellcolor{green!10} Atmospheric Science(203) &  \cellcolor{green!5} Atmospheric Physics and Atmospheric Environment(69) \\
\cline{3-3}
\cline{6-6}
\cellcolor{purple!20} & \cellcolor{purple!10} &  \cellcolor{purple!5} History and Theory of Journalism and Media Management(59) &  \cellcolor{green!20} & \cellcolor{green!10} &  \cellcolor{green!5} Dynamic Meteorology(50) \\
\cline{3-3}
\cline{6-6}
\cellcolor{purple!20} & \cellcolor{purple!10} &  \cellcolor{purple!5} Journalism and News Practice(90) &  \cellcolor{green!20} & \cellcolor{green!10} &  \cellcolor{green!5} Meteorology(84) \\
\cline{2-3}
\cline{5-6}
\cellcolor{purple!20} &  \cellcolor{purple!10} Language and Literature(440) &  \cellcolor{purple!5} Classical Chinese Literature(50) &  \cellcolor{green!20}  &  \cellcolor{green!10} Biology(1120) &  \cellcolor{green!5} Biochemistry and Molecular Biology(99) \\
\cline{3-3}
\cline{6-6}
\cellcolor{purple!20} & \cellcolor{purple!10} &  \cellcolor{purple!5} French Language and Literature(50) &  \cellcolor{green!20} & \cellcolor{green!10} &  \cellcolor{green!5} Biophysics(50) \\
\cline{3-3}
\cline{6-6}
\cellcolor{purple!20} & \cellcolor{purple!10} &  \cellcolor{purple!5} Linguistics and Applied Linguistics(50) &  \cellcolor{green!20} & \cellcolor{green!10} &  \cellcolor{green!5} Botany(166) \\
\cline{3-3}
\cline{6-6}
\cellcolor{purple!20} & \cellcolor{purple!10} &  \cellcolor{purple!5} Literary History(69) &  \cellcolor{green!20} & \cellcolor{green!10} &  \cellcolor{green!5} Cell Biology(87) \\
\cline{3-3}
\cline{6-6}
\cellcolor{purple!20} & \cellcolor{purple!10} &  \cellcolor{purple!5} Literary Theory(64) &  \cellcolor{green!20} & \cellcolor{green!10} &  \cellcolor{green!5} Ecology(144) \\
\cline{3-3}
\cline{6-6}
\cellcolor{purple!20} & \cellcolor{purple!10} &  \cellcolor{purple!5} Modern and Contemporary Chinese Literature(50) &  \cellcolor{green!20} & \cellcolor{green!10} &  \cellcolor{green!5} Genetics(184) \\
\cline{3-3}
\cline{6-6}
\cellcolor{purple!20} & \cellcolor{purple!10} &  \cellcolor{purple!5} Philology and Bibliography(50) &  \cellcolor{green!20} & \cellcolor{green!10} &  \cellcolor{green!5} Microbiology(161) \\
\cline{3-3}
\cline{6-6}
\cellcolor{purple!20} & \cellcolor{purple!10} &  \cellcolor{purple!5} Russian Language and Literature(57) &  \cellcolor{green!20} & \cellcolor{green!10} &  \cellcolor{green!5} Physiology(120) \\
\cline{2-3}
\cline{6-6}
\cellcolor{purple!20} &  \cellcolor{purple!10} Musicology(426) &  \cellcolor{purple!5} Composition(50) &  \cellcolor{green!20}   &  \cellcolor{green!10} &  \cellcolor{green!5} Zoology(109) \\
\cline{3-3}
\cline{5-6}
\cellcolor{purple!20} & \cellcolor{purple!10} &  \cellcolor{purple!5} Harmony(60) &  \cellcolor{green!20} & \cellcolor{green!10} Chemistry(1769) &  \cellcolor{green!5} Analytical Chemistry(396) \\
\cline{3-3}
\cline{6-6}
\cellcolor{purple!20} & \cellcolor{purple!10} &  \cellcolor{purple!5} Instrumentation and Performance(65) &  \cellcolor{green!20} & \cellcolor{green!10} &  \cellcolor{green!5} Electrochemistry(222) \\
\cline{3-3}
\cline{6-6}
\cellcolor{purple!20} & \cellcolor{purple!10} &  \cellcolor{purple!5} Music History, Education, and Technology(145) &  \cellcolor{green!20} & \cellcolor{green!10} &  \cellcolor{green!5} Inorganic Chemistry(138) \\
\cline{3-3}
\cline{6-6}
\cellcolor{purple!20} & \cellcolor{purple!10} &  \cellcolor{purple!5} Musical Forms and Analysis(50) &  \cellcolor{green!20} & \cellcolor{green!10} &  \cellcolor{green!5} Organic Chemistry(171) \\
\cline{3-3}
\cline{6-6}
\cellcolor{purple!20} & \cellcolor{purple!10} &  \cellcolor{purple!5} Pitch and Scales(56) &  \cellcolor{green!20} & \cellcolor{green!10} &  \cellcolor{green!5} Physical Chemistry(706) \\
\hhline{---~~~}
\cline{6-6}
\cellcolor{orange!20} \textbf{Management(501)}  & \cellcolor{orange!10} Business Administration(142) &  \cellcolor{orange!5} Business and Accounting Management(92) &  \cellcolor{green!20} & \cellcolor{green!10} &  \cellcolor{green!5} Polymer Chemistry and Physics(76) \\
\cline{3-3}
\cline{6-6}
\cellcolor{orange!20} & \cellcolor{orange!10} &  \cellcolor{orange!5} Tourism Management and Technological Economics Management(50) &  \cellcolor{green!20} & \cellcolor{green!10} &  \cellcolor{green!5} Radiochemistry(60) \\
\cline{2-3}
\cline{5-6}
\cellcolor{orange!20} &  \cellcolor{orange!10} Library, Information and Archival Management(150) &  \cellcolor{orange!5} Information Management Science(50) &  \cellcolor{green!20}  &  \cellcolor{green!10} Geography(133) &  \cellcolor{green!5} Human Geography(50) \\
\cline{3-3}
\cline{6-6}
\cellcolor{orange!20} & \cellcolor{orange!10} &  \cellcolor{orange!5} Information Management and Communication(50) &  \cellcolor{green!20} & \cellcolor{green!10} &  \cellcolor{green!5} Physical Geography(83) \\
\cline{3-3}
\cline{5-6}
\cellcolor{orange!20} & \cellcolor{orange!10} &  \cellcolor{orange!5} Library and Archival Science(50) &  \cellcolor{green!20} & \cellcolor{green!10} Geology(341) &  \cellcolor{green!5} Geochemistry(50) \\
\cline{2-3}
\cline{6-6}
\cellcolor{orange!20} &  \cellcolor{orange!10} Management Science and Engineering(58) &  \cellcolor{orange!5} Management Science and Engineering(58) &  \cellcolor{green!20}   &  \cellcolor{green!10} &  \cellcolor{green!5} Mineralogy, Petrology, and Economic Geology(121) \\
\cline{2-3}
\cline{6-6}
\cellcolor{orange!20} &  \cellcolor{orange!10} Public Administration(151) &  \cellcolor{orange!5} Education Economics, Management and Social Security(50) &  \cellcolor{green!20}   &  \cellcolor{green!10} &  \cellcolor{green!5} Paleontology and Stratigraphy(50) \\
\cline{3-3}
\cline{6-6}
\cellcolor{orange!20} & \cellcolor{orange!10} &  \cellcolor{orange!5} Land Resource Management and Administrative Management(51) &  \cellcolor{green!20} & \cellcolor{green!10} &  \cellcolor{green!5} Principles of Seismic Exploration(50) \\
\cline{3-3}
\cline{6-6}
\cellcolor{orange!20} & \cellcolor{orange!10} &  \cellcolor{orange!5} Social Medicine and Health Management(50) &  \cellcolor{green!20} & \cellcolor{green!10} &  \cellcolor{green!5} Structural Geology(70) \\
\hhline{---~~~}
\cline{5-6}
\cellcolor{yellow!20} \textbf{Medicine(2755)}  &  \cellcolor{yellow!10} Basic Medicine(567) & \cellcolor{yellow!5}  Forensic Medicine(50) &  \cellcolor{green!20}  &  \cellcolor{green!10} Geophysics(100) &  \cellcolor{green!5} Solid Earth Geophysics(50) \\
\cline{3-3}
\cline{6-6}
\cellcolor{yellow!20} & \cellcolor{yellow!10} &  \cellcolor{yellow!5} Human Anatomy and Histology-Embryology(189) &  \cellcolor{green!20} & \cellcolor{green!10} &  \cellcolor{green!5} Space physics(50) \\
\cline{3-3}
\cline{5-6}
\cellcolor{yellow!20} & \cellcolor{yellow!10} &  \cellcolor{yellow!5} Immunology(69) &  \cellcolor{green!20} & \cellcolor{green!10} Mathematics(2622) &  \cellcolor{green!5} Advanced Algebra(202) \\
\cline{3-3}
\cline{6-6}
\cellcolor{yellow!20} & \cellcolor{yellow!10} &  \cellcolor{yellow!5} Pathogen Biology(94) &  \cellcolor{green!20} & \cellcolor{green!10} &  \cellcolor{green!5} Combinatorial Mathematics(191) \\
\cline{3-3}
\cline{6-6}
\cellcolor{yellow!20} & \cellcolor{yellow!10} &  \cellcolor{yellow!5} Pathology and Pathophysiology(115) &  \cellcolor{green!20} & \cellcolor{green!10} &  \cellcolor{green!5} Computational Mathematics(50) \\
\cline{3-3}
\cline{6-6}
\cellcolor{yellow!20} & \cellcolor{yellow!10} &  \cellcolor{yellow!5} Radiation Medicine(50) &  \cellcolor{green!20} & \cellcolor{green!10} &  \cellcolor{green!5} Cryptography(50) \\
\cline{2-3}
\cline{6-6}
\cellcolor{yellow!20} &  \cellcolor{yellow!10} Clinical Medicine(1218) &  \cellcolor{yellow!5} Anesthesiology(50) &  \cellcolor{green!20}   &  \cellcolor{green!10} &  \cellcolor{green!5} Discrete Mathematics(89) \\
\cline{3-3}
\cline{6-6}
\cellcolor{yellow!20} & \cellcolor{yellow!10} &  \cellcolor{yellow!5} Clinical Laboratory Diagnostics(50) &  \cellcolor{green!20} & \cellcolor{green!10} &  \cellcolor{green!5} Functions of Complex Variables(122) \\
\cline{3-3}
\cline{6-6}
\cellcolor{yellow!20} & \cellcolor{yellow!10} &  \cellcolor{yellow!5} Dermatology and Venereology(66) &  \cellcolor{green!20} & \cellcolor{green!10} &  \cellcolor{green!5} Functions of Real Variables(67) \\
\cline{3-3}
\cline{6-6}
\cellcolor{yellow!20} & \cellcolor{yellow!10} &  \cellcolor{yellow!5} Emergency Medicine(66) &  \cellcolor{green!20} & \cellcolor{green!10} &  \cellcolor{green!5} Fundamental Mathematics(197) \\
\cline{3-3}
\cline{6-6}
\cellcolor{yellow!20} & \cellcolor{yellow!10} &  \cellcolor{yellow!5} Geriatric Medicine(50) &  \cellcolor{green!20} & \cellcolor{green!10} &  \cellcolor{green!5} Fuzzy Mathematics(50) \\
\cline{3-3}
\cline{6-6}
\cellcolor{yellow!20} & \cellcolor{yellow!10} &  \cellcolor{yellow!5} Imaging and Nuclear Medicine(67) &  \cellcolor{green!20} & \cellcolor{green!10} &  \cellcolor{green!5} Geometry and Topology(265) \\
\cline{3-3}
\cline{6-6}
\cellcolor{yellow!20} & \cellcolor{yellow!10} &  \cellcolor{yellow!5} Internal Medicine(202) &  \cellcolor{green!20} & \cellcolor{green!10} &  \cellcolor{green!5} Graph Theory(51) \\
\cline{3-3}
\cline{6-6}
\cellcolor{yellow!20} & \cellcolor{yellow!10} &  \cellcolor{yellow!5} Neurology(189) &  \cellcolor{green!20} & \cellcolor{green!10} &  \cellcolor{green!5} Group Theory(50) \\
\cline{3-3}
\cline{6-6}
\cellcolor{yellow!20} & \cellcolor{yellow!10} &  \cellcolor{yellow!5} Nursing and Rehabilitation Medicine(50) &  \cellcolor{green!20} & \cellcolor{green!10} &  \cellcolor{green!5} Mathematical Analysis(288) \\
\cline{3-3}
\cline{6-6}
\cellcolor{yellow!20} & \cellcolor{yellow!10} &  \cellcolor{yellow!5} Obstetrics and Gynecology(102) &  \cellcolor{green!20} & \cellcolor{green!10} &  \cellcolor{green!5} Number Theory(220) \\
\cline{3-3}
\cline{6-6}
\cellcolor{yellow!20} & \cellcolor{yellow!10} &  \cellcolor{yellow!5} Oncology(58) &  \cellcolor{green!20} & \cellcolor{green!10} &  \cellcolor{green!5} Numerical Analysis(67) \\
\cline{3-3}
\cline{6-6}
\cellcolor{yellow!20} & \cellcolor{yellow!10} &  \cellcolor{yellow!5} Ophthalmology(50) &  \cellcolor{green!20} & \cellcolor{green!10} &  \cellcolor{green!5} Ordinary Differential Equations(160) \\
\cline{3-3}
\cline{6-6}
\cellcolor{yellow!20} & \cellcolor{yellow!10} &  \cellcolor{yellow!5} Otorhinolaryngology(50) &  \cellcolor{green!20} & \cellcolor{green!10} &  \cellcolor{green!5} Polynomials and Series Expansions(179) \\
\cline{3-3}
\cline{6-6}
\cellcolor{yellow!20} & \cellcolor{yellow!10} &  \cellcolor{yellow!5} Pediatrics(50) &  \cellcolor{green!20} & \cellcolor{green!10} &  \cellcolor{green!5} Probability and Statistics(224) \\
\cline{3-3}
\cline{6-6}
\cellcolor{yellow!20} & \cellcolor{yellow!10} &  \cellcolor{yellow!5} Psychiatry and Mental Health(50) &  \cellcolor{green!20} & \cellcolor{green!10} &  \cellcolor{green!5} Special Number Theory(50) \\
\cline{3-3}
\cline{6-6}
\cellcolor{yellow!20} & \cellcolor{yellow!10} &  \cellcolor{yellow!5} Surgery(68) &  \cellcolor{green!20} & \cellcolor{green!10} &  \cellcolor{green!5} Stochastic Processes(50) \\
\cline{2-3}
\cline{5-6}
\cellcolor{yellow!20} &  \cellcolor{yellow!10} Pharmacy(278) &  \cellcolor{yellow!5} Medicinal Chemistry(50) &  \cellcolor{green!20}  &  \cellcolor{green!10} Oceanography(200) &  \cellcolor{green!5} Hydrogeology(50) \\
\cline{3-3}
\cline{6-6}
\cellcolor{yellow!20} & \cellcolor{yellow!10} &  \cellcolor{yellow!5} Microbiology and Biochemical Pharmacy(50) &  \cellcolor{green!20} & \cellcolor{green!10} &  \cellcolor{green!5} Marine Biology(50) \\
\cline{3-3}
\cline{6-6}
\cellcolor{yellow!20} & \cellcolor{yellow!10} &  \cellcolor{yellow!5} Pharmaceutical Analysis(50) &  \cellcolor{green!20} & \cellcolor{green!10} &  \cellcolor{green!5} Marine Chemistry(50) \\
\cline{3-3}
\cline{6-6}
\cellcolor{yellow!20} & \cellcolor{yellow!10} &  \cellcolor{yellow!5} Pharmaceutics(73) &  \cellcolor{green!20} & \cellcolor{green!10} &  \cellcolor{green!5} Underwater Acoustics(50) \\
\cline{3-3}
\cline{5-6}
\cellcolor{yellow!20} & \cellcolor{yellow!10} &  \cellcolor{yellow!5} Pharmacology(55) &  \cellcolor{green!20} & \cellcolor{green!10} Physical Oceanography(50) &  \cellcolor{green!5} Physical Oceanography(50) \\
\cline{2-3}
\cline{5-6}
\cellcolor{yellow!20} &  \cellcolor{yellow!10} Public Health and Preventive Medicine(292) &  \cellcolor{yellow!5} Epidemiology and Health Statistics(83) &  \cellcolor{green!20}  &  \cellcolor{green!10} Physics(2845) &  \cellcolor{green!5} Acoustics(245) \\
\cline{3-3}
\cline{6-6}
\cellcolor{yellow!20} & \cellcolor{yellow!10} &  \cellcolor{yellow!5} Health Toxicology and Environmental Health(80) &  \cellcolor{green!20} & \cellcolor{green!10} &  \cellcolor{green!5} Atomic and Molecular Physics(210) \\
\cline{3-3}
\cline{6-6}
\cellcolor{yellow!20} & \cellcolor{yellow!10} &  \cellcolor{yellow!5} Maternal, Child and Adolescent Health(79) &  \cellcolor{green!20} & \cellcolor{green!10} &  \cellcolor{green!5} Electrodynamics(583) \\
\cline{3-3}
\cline{6-6}
\cellcolor{yellow!20} & \cellcolor{yellow!10} &  \cellcolor{yellow!5} Nutrition and Food Hygiene(50) &  \cellcolor{green!20} & \cellcolor{green!10} &  \cellcolor{green!5} Fluid Physics(134) \\
\cline{2-3}
\cline{6-6}
\cellcolor{yellow!20} &  \cellcolor{yellow!10} Stomatology(132) &  \cellcolor{yellow!5} Basic Stomatology(53) &  \cellcolor{green!20}   &  \cellcolor{green!10} &  \cellcolor{green!5} Particle and Nuclear Physics(225) \\
\cline{3-3}
\cline{6-6}
\cellcolor{yellow!20} & \cellcolor{yellow!10} &  \cellcolor{yellow!5} Clinical Stomatology(79) &  \cellcolor{green!20} & \cellcolor{green!10} &  \cellcolor{green!5} Polymer Physics(109) \\
\cline{2-3}
\cline{6-6}
\cellcolor{yellow!20} &  \cellcolor{yellow!10} Traditional Chinese Medicine(268) &  \cellcolor{yellow!5} Traditional Chinese Health Preservation(59) &  \cellcolor{green!20}   &  \cellcolor{green!10} &  \cellcolor{green!5} Quantum Mechanics(206) \\
\cline{3-3}
\cline{6-6}
\cellcolor{yellow!20} & \cellcolor{yellow!10} &  \cellcolor{yellow!5} Traditional Chinese Medicine Theory(90) &  \cellcolor{green!20} & \cellcolor{green!10} &  \cellcolor{green!5} Relativity(79) \\
\cline{3-3}
\cline{6-6}
\cellcolor{yellow!20} & \cellcolor{yellow!10} &  \cellcolor{yellow!5} Traditional Chinese Pharmacy(119) &  \cellcolor{green!20} & \cellcolor{green!10} &  \cellcolor{green!5} Semiconductor Physics(61) \\
\hhline{---~~~}
\cline{6-6}
\cellcolor{magenta!20} \textbf{Military Science(205)}  & \cellcolor{magenta!10} Military Science(205) &  \cellcolor{magenta!5} Military Command and Information Systems(50) &  \cellcolor{green!20} & \cellcolor{green!10} &  \cellcolor{green!5} Solid State Physics(147) \\
\cline{3-3}
\cline{6-6}
\cellcolor{magenta!20} & \cellcolor{magenta!10} &  \cellcolor{magenta!5} Military Logistics and Equipment(54) &  \cellcolor{green!20} & \cellcolor{green!10} &  \cellcolor{green!5} Statistical Mechanics(50) \\
\cline{3-3}
\cline{6-6}
\cellcolor{magenta!20} & \cellcolor{magenta!10} &  \cellcolor{magenta!5} Military Management(50) &  \cellcolor{green!20} & \cellcolor{green!10} &  \cellcolor{green!5} Subatomic and Atomic Physics(51) \\
\cline{3-3}
\cline{6-6}
\cellcolor{magenta!20} & \cellcolor{magenta!10} &  \cellcolor{magenta!5} Military Thought and History(51) &  \cellcolor{green!20} & \cellcolor{green!10} &  \cellcolor{green!5} Thermodynamics(215) \\
\hhline{---~~~}
\cline{6-6}
\cellcolor{lime!20} \textbf{Philosophy(347)}  & \cellcolor{lime!10} Philosophy(347) &  \cellcolor{lime!5} Ethics(68) &  \cellcolor{green!20} & \cellcolor{green!10} &  \cellcolor{green!5} Thermodynamics and Statistical Physics(530) \\
\cline{3-3}
\cline{5-6}
\cellcolor{lime!20} & \cellcolor{lime!10} &  \cellcolor{lime!5} Logic(70) &  \cellcolor{green!20} & \cellcolor{green!10} Systems Science(50) &  \cellcolor{green!5} Systems Science(50) \\
\cline{3-3}
\hhline{~~~---}
\cellcolor{lime!20} & \cellcolor{lime!10} &  \cellcolor{lime!5} Philosophical Aesthetics(109) &  \cellcolor{olive!20}  \textbf{Sociology(143)} &  \cellcolor{olive!10} Sociology(143) &  \cellcolor{olive!5}  Demography and Anthropology(50) \\
\cline{3-3}
\cline{6-6}
\cellcolor{lime!20} & \cellcolor{lime!10} &  \cellcolor{lime!5} Philosophy of Science and Technology(50) &  \cellcolor{olive!20} & \cellcolor{olive!10} &  \cellcolor{olive!5} Social and Folklore Studies(93) \\
\bottomrule
\end{tabular}
}
\caption{Continued: Comprehensive Overview of Three Levels of Disciplines and Their Sample Sizes.}
\label{tab:detailed_discipline_2}
\end{table}
}
\section{Detailed Dataset Statistics}
\begin{figure}[H]
\small
    \centering
    \begin{subfigure}[b]{0.30\textwidth}
        \centering
        \includegraphics[width=\textwidth]{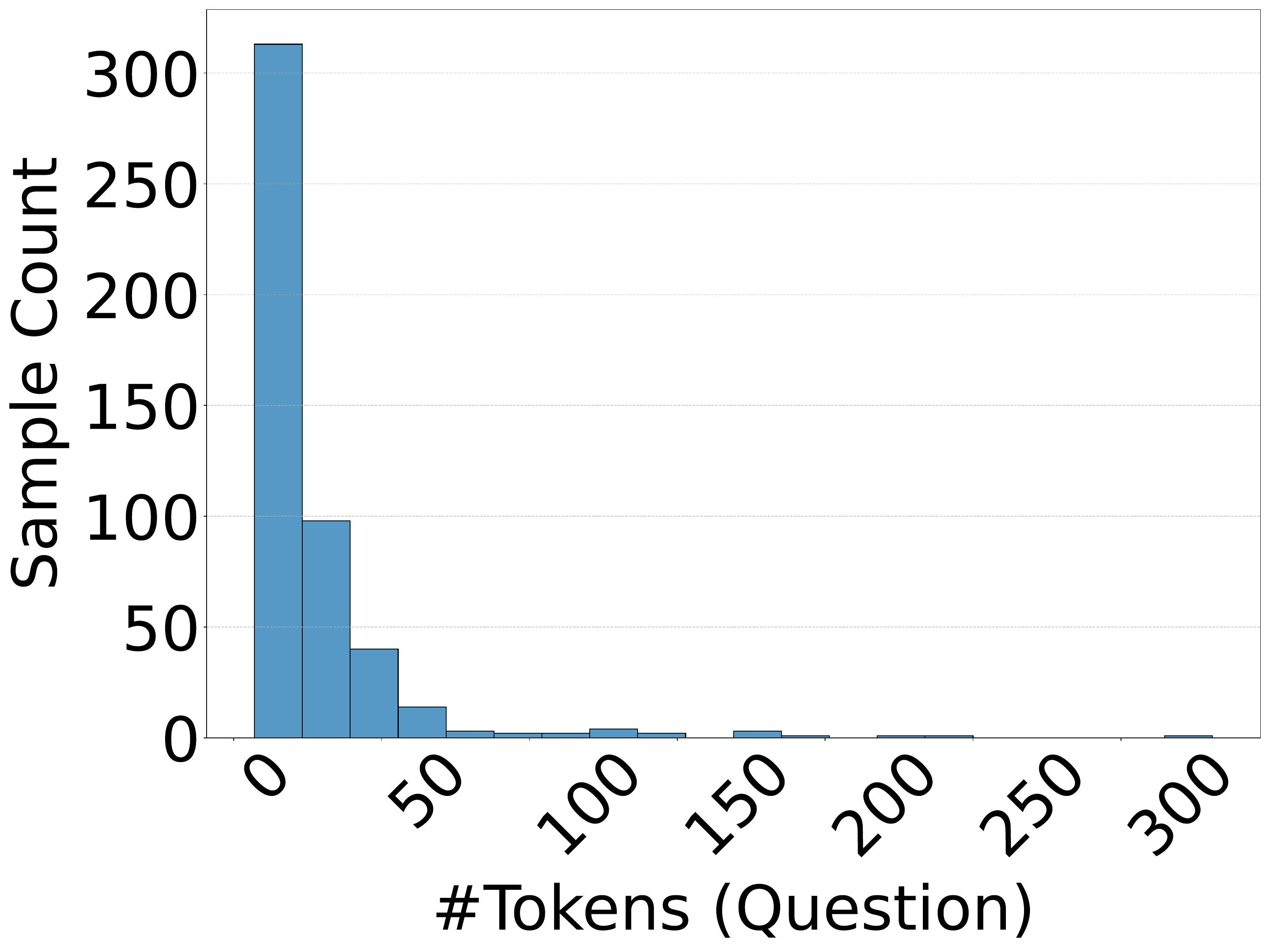}
        \caption{Agriculture.}
    \end{subfigure}
    \hfill
    \begin{subfigure}[b]{0.30\textwidth}
        \centering
        \includegraphics[width=\textwidth]{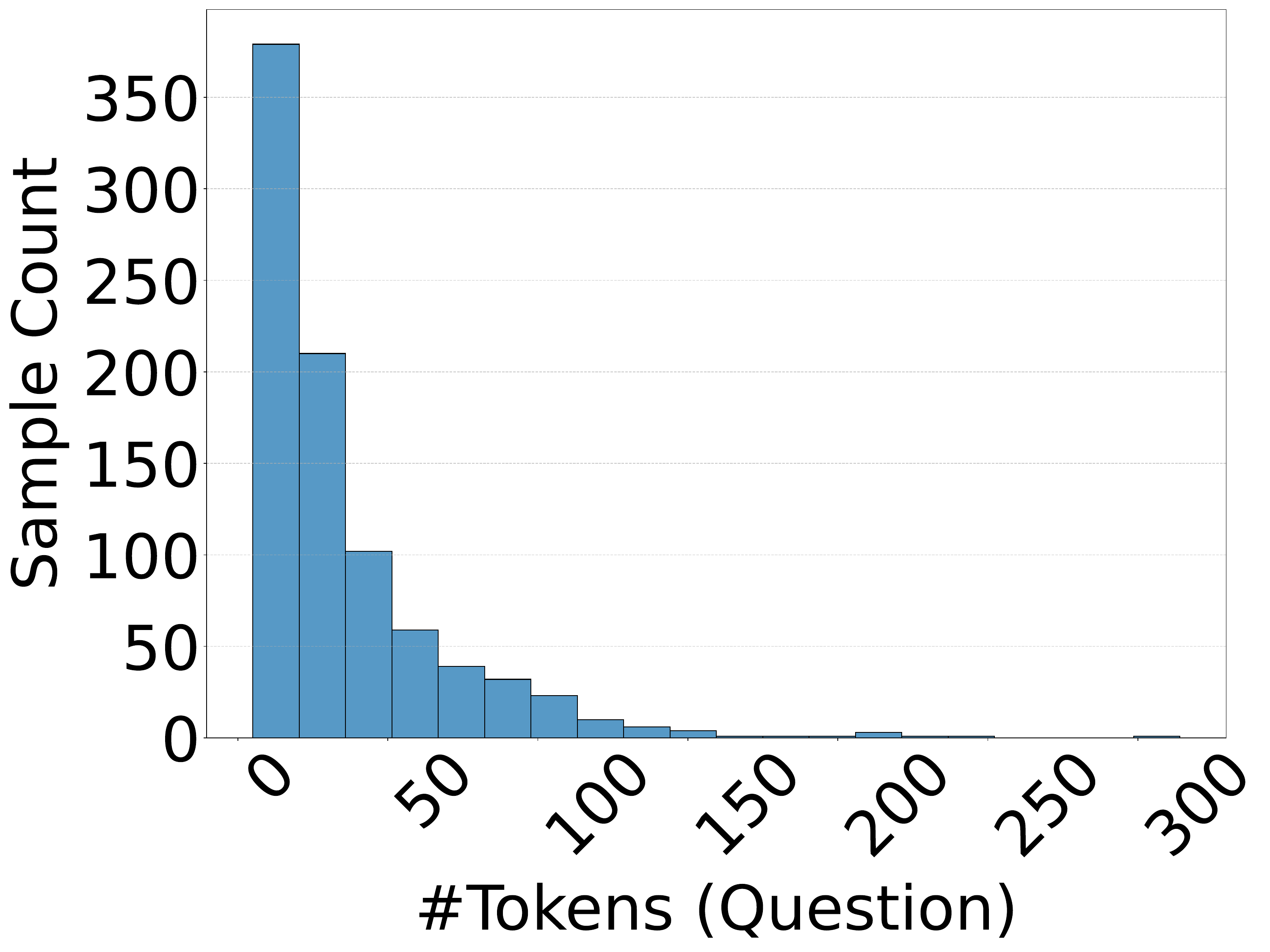}
        \caption{Economics.}
    \end{subfigure}
    \hfill
    \begin{subfigure}[b]{0.30\textwidth}
        \centering
        \includegraphics[width=\textwidth]{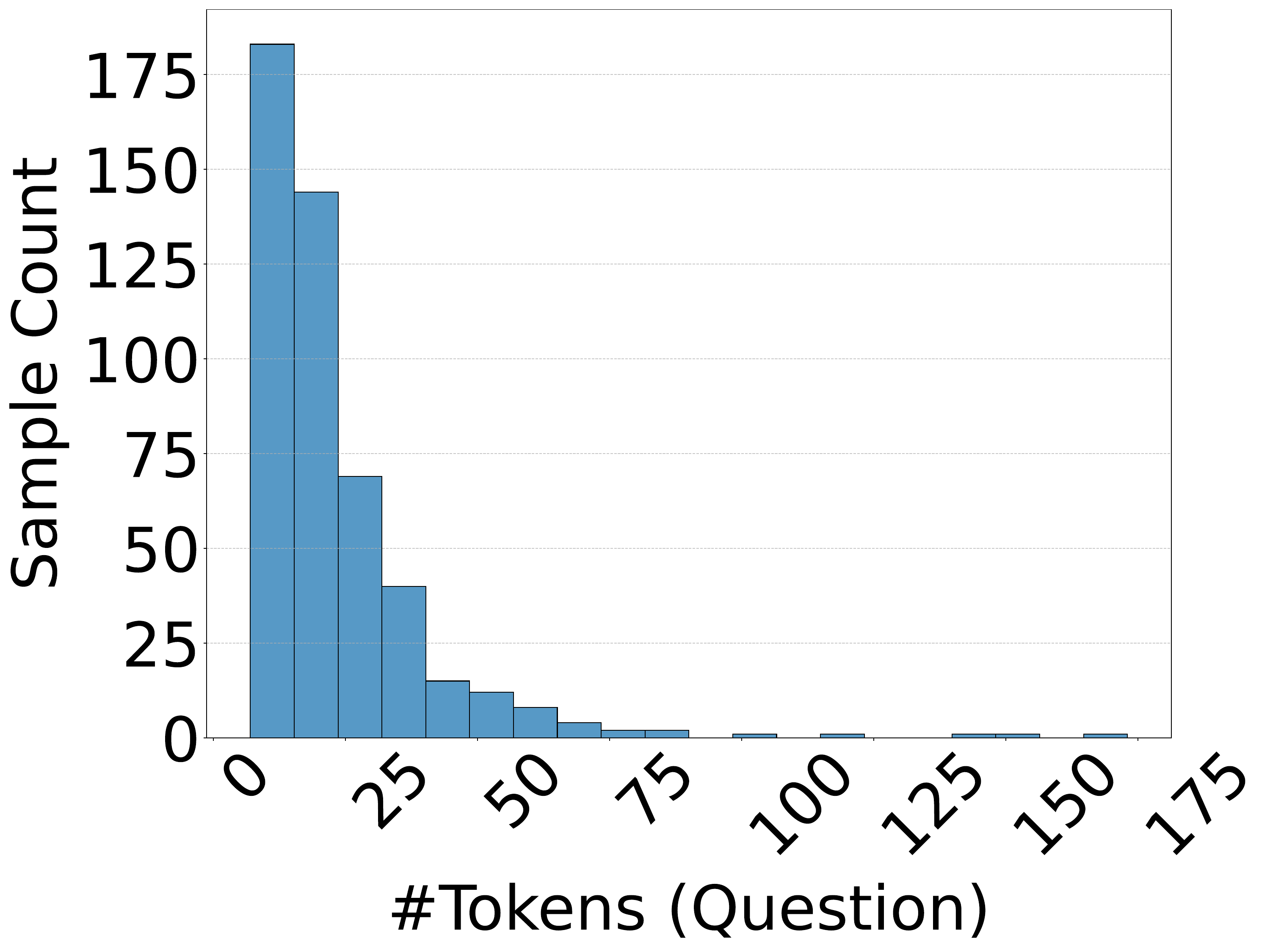}
        \caption{Education.}
    \end{subfigure}
    \\
    \begin{subfigure}[b]{0.30\textwidth}
        \centering
        \includegraphics[width=\textwidth]{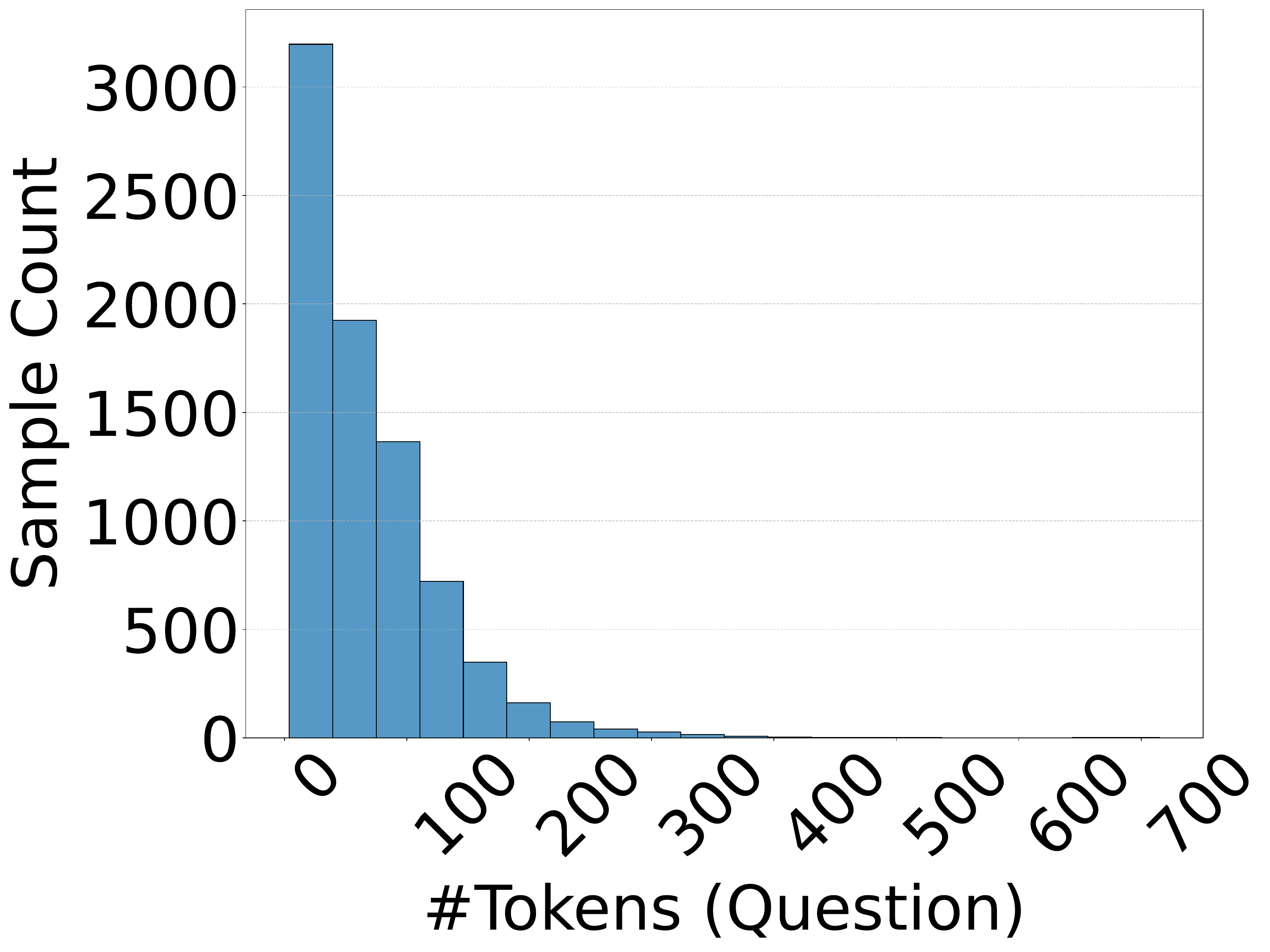}
        \caption{Engineering.}
    \end{subfigure}
    \hfill
    \begin{subfigure}[b]{0.30\textwidth}
        \centering
        \includegraphics[width=\textwidth]{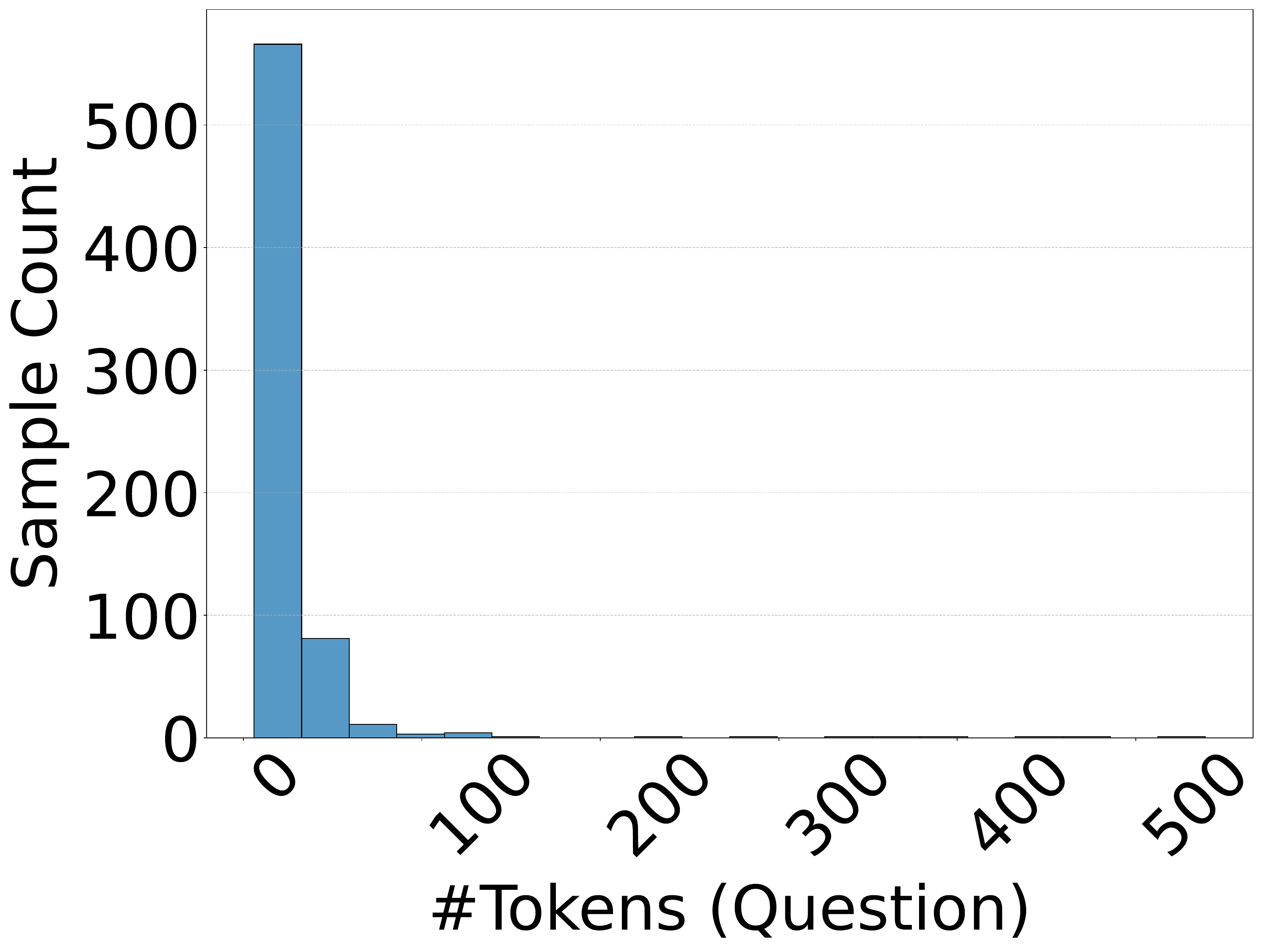}
        \caption{History.}
    \end{subfigure}
    \hfill
    \begin{subfigure}[b]{0.30\textwidth}
        \centering
        \includegraphics[width=\textwidth]{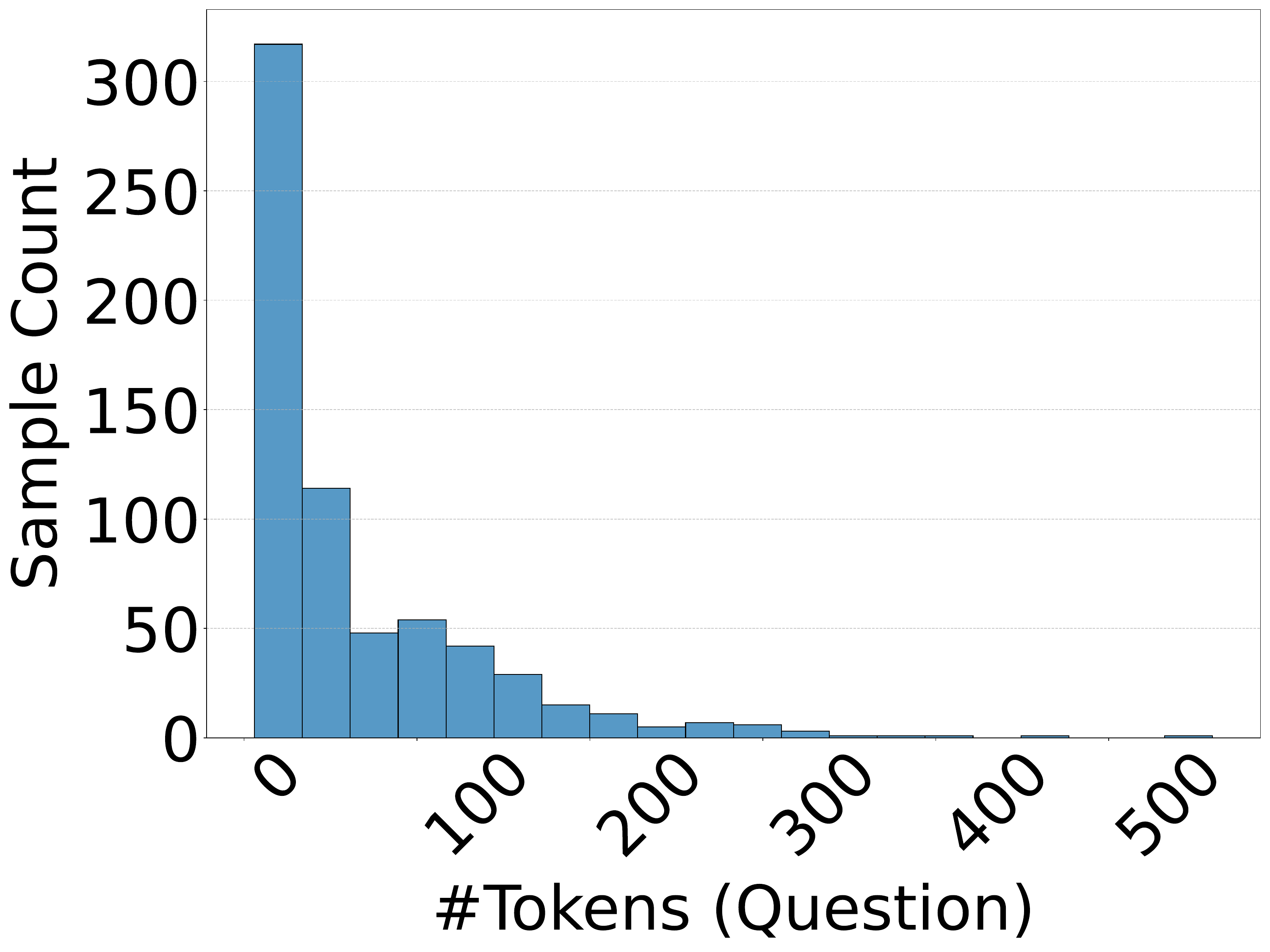}
        \caption{Law.}
    \end{subfigure}
    \\
    \begin{subfigure}[b]{0.30\textwidth}
        \centering
        \includegraphics[width=\textwidth]{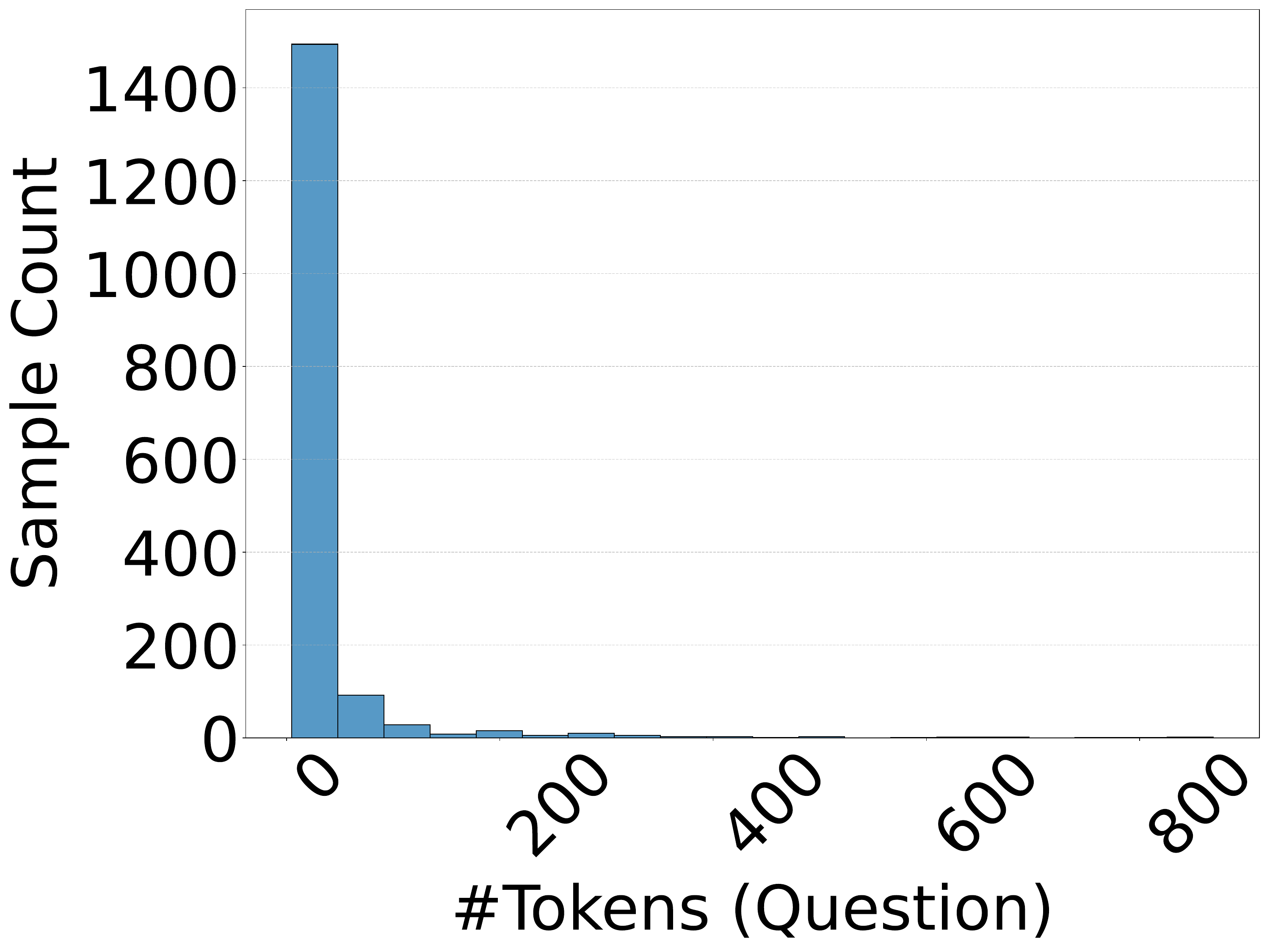}
        \caption{Literature.}
    \end{subfigure}
    \hfill
    \begin{subfigure}[b]{0.30\textwidth}
        \centering
        \includegraphics[width=\textwidth]{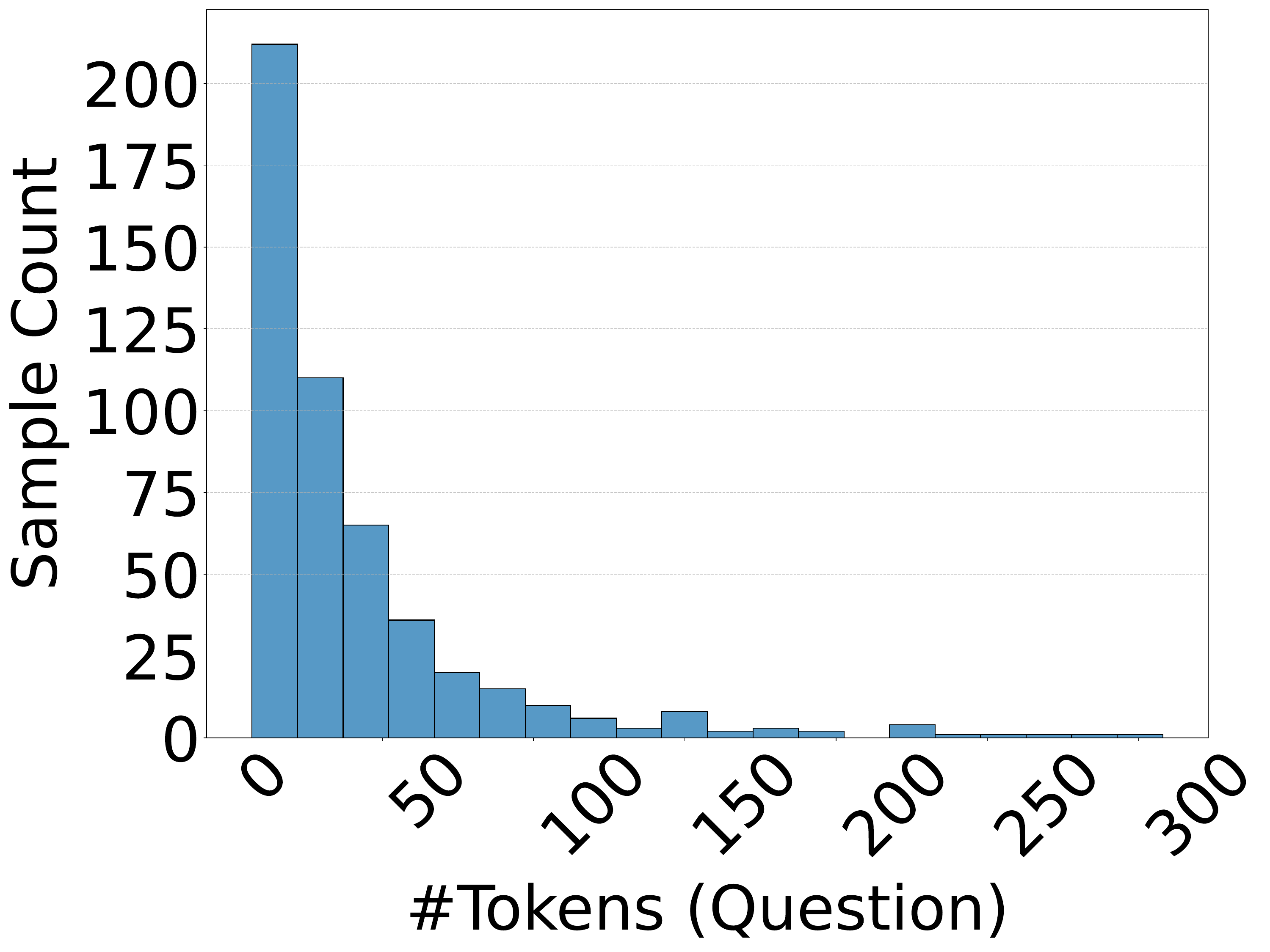}
        \caption{Management.}
    \end{subfigure}
    \hfill
    \begin{subfigure}[b]{0.30\textwidth}
        \centering
        \includegraphics[width=\textwidth]{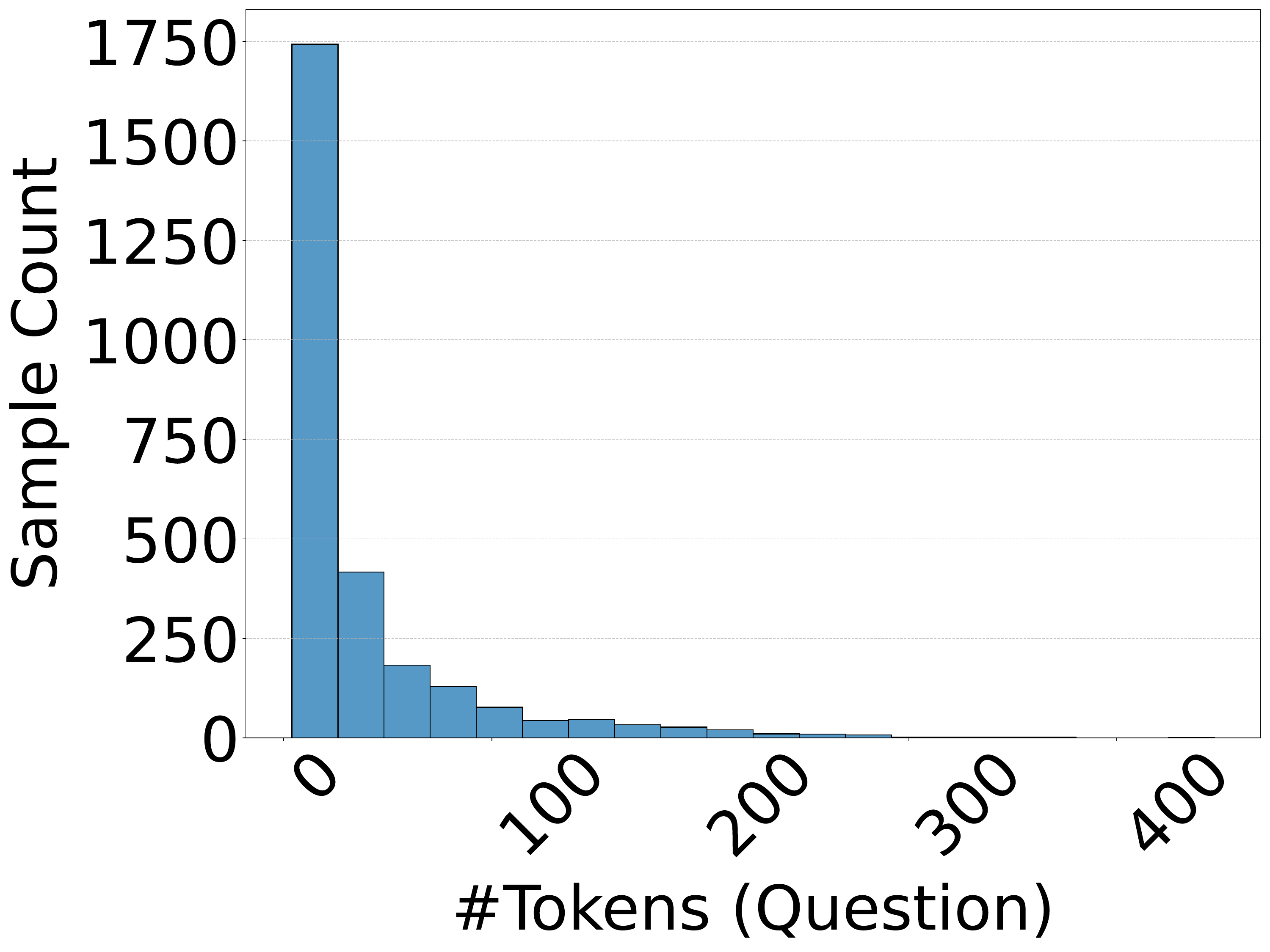}
        \caption{Medicine.}
    \end{subfigure}
    \\
        \begin{subfigure}[b]{0.30\textwidth}
        \centering
        \includegraphics[width=\textwidth]{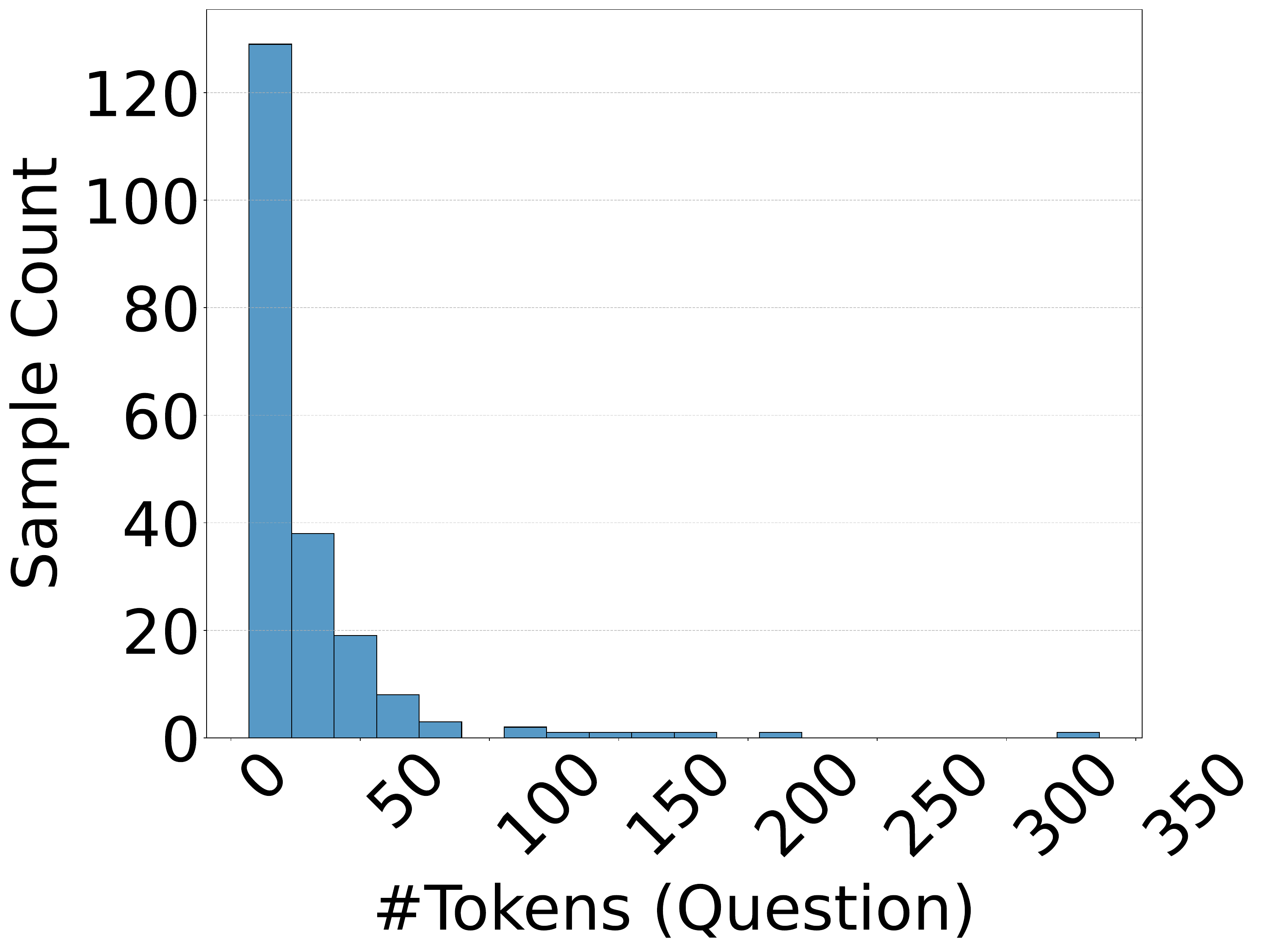}
        \caption{Military Science.}
    \end{subfigure}
    \hfill
    \begin{subfigure}[b]{0.30\textwidth}
        \centering
        \includegraphics[width=\textwidth]{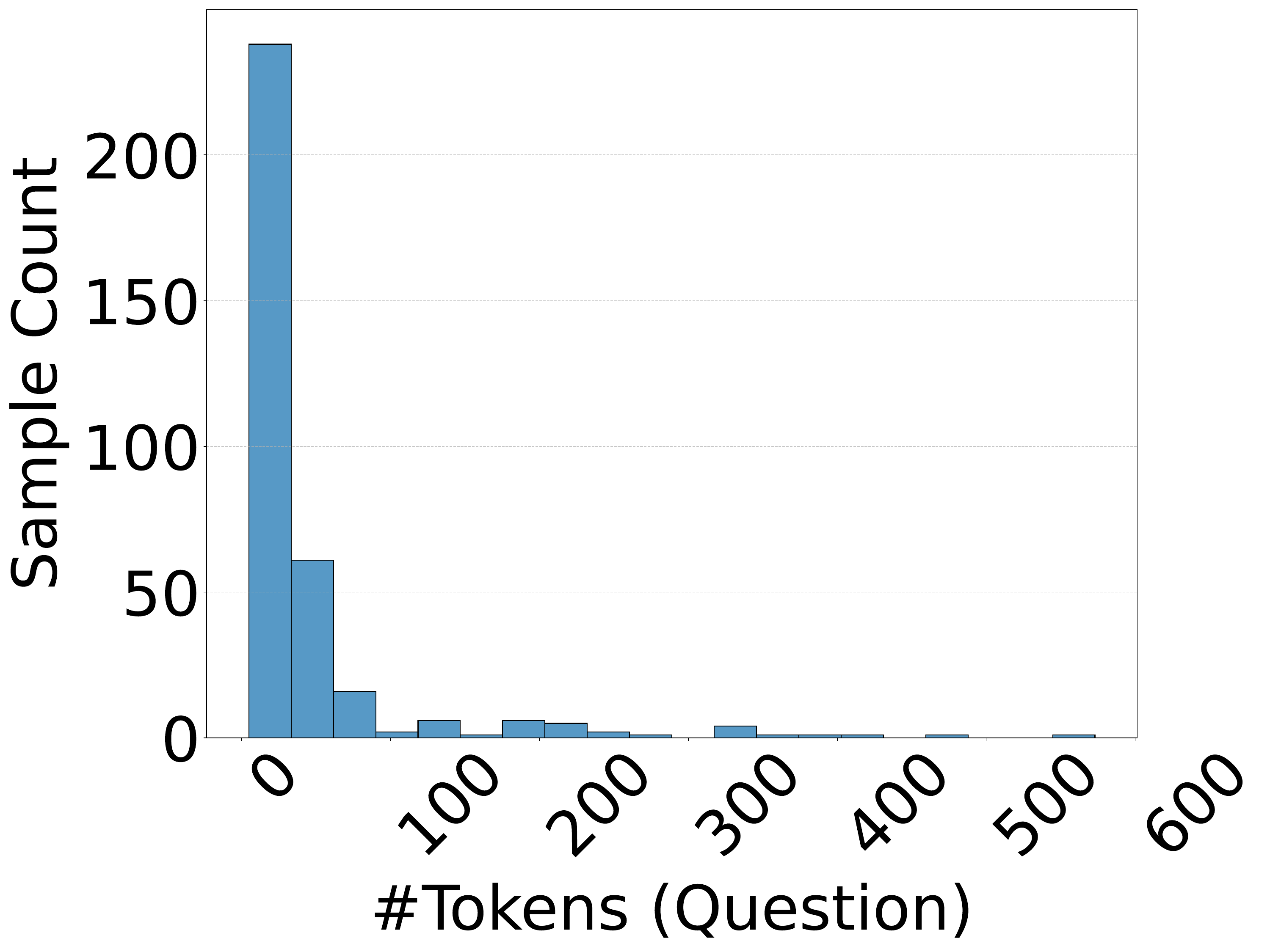}
        \caption{Philosophy.}
    \end{subfigure}
    \hfill
    \begin{subfigure}[b]{0.30\textwidth}
        \centering
        \includegraphics[width=\textwidth]{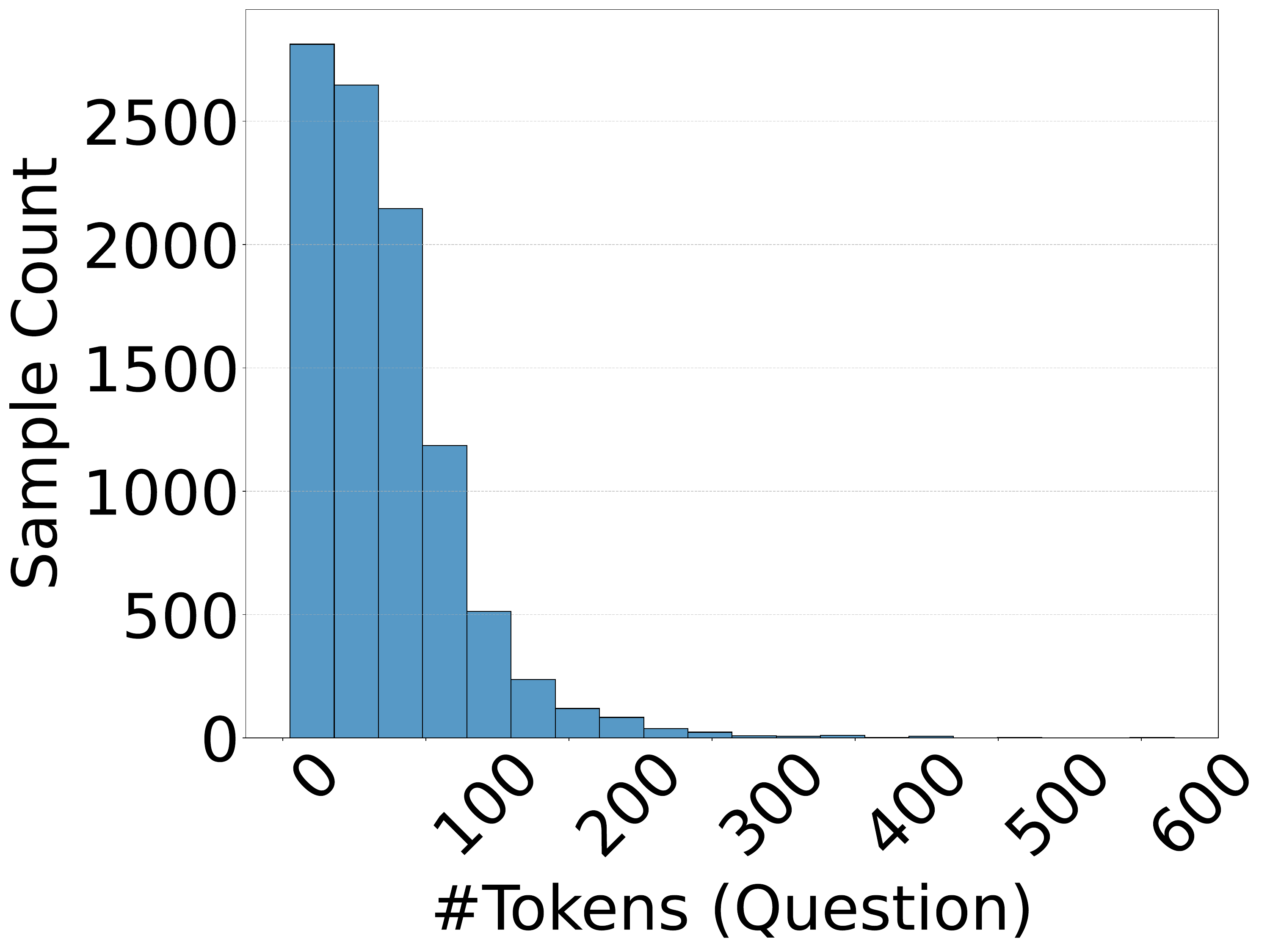}
        \caption{Science.}
    \end{subfigure}
    \\
    \begin{subfigure}[b]{0.30\textwidth}
        \centering
        \includegraphics[width=\textwidth]{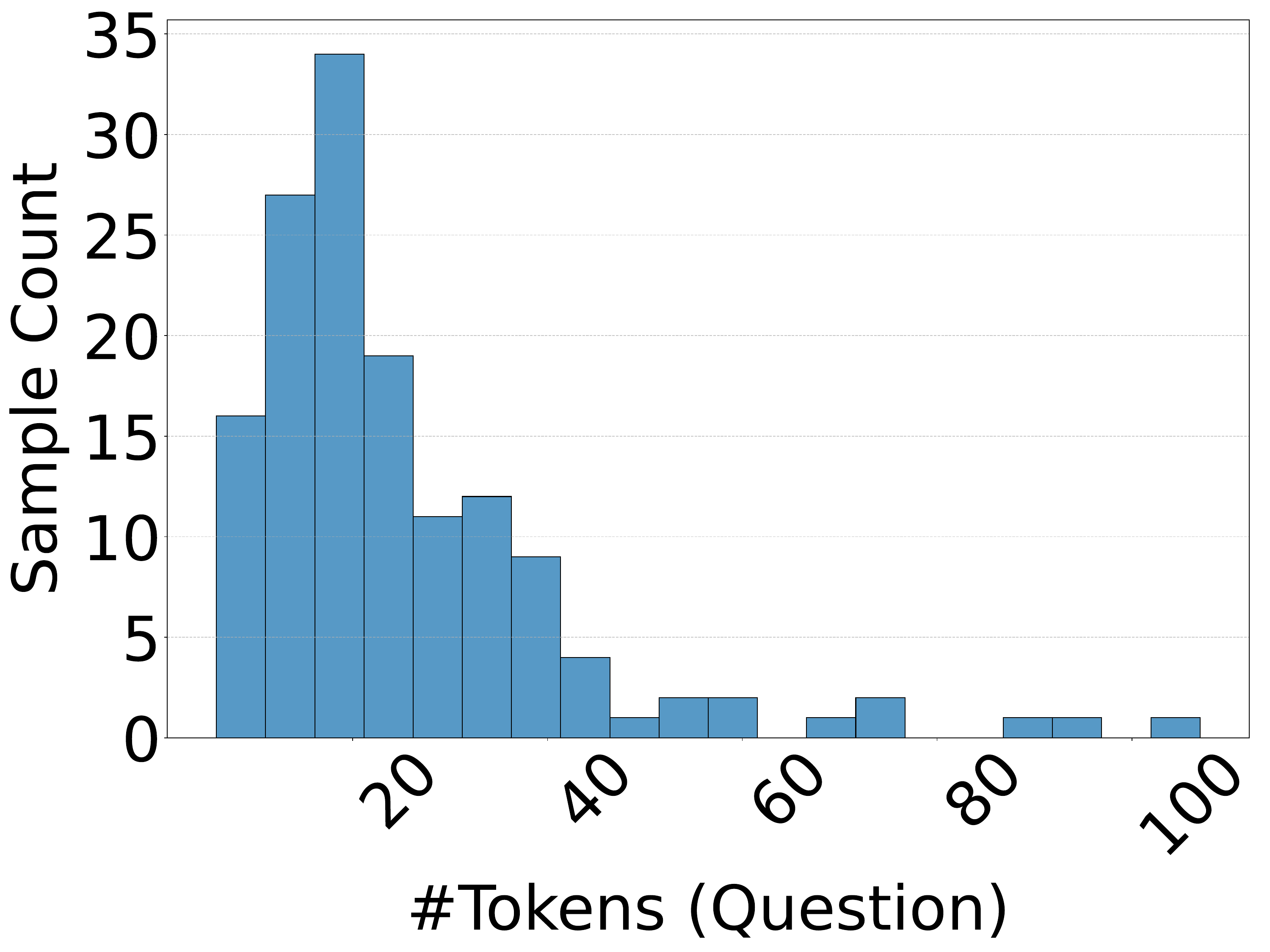}
        \caption{Sociology.}
    \end{subfigure}

    \caption{Question Length Distribution Across Disciplines.}
    \label{fig:question_lengths}
\end{figure}

\begin{figure}[H]
    \centering
    \begin{subfigure}[b]{0.30\textwidth}
        \centering
        \includegraphics[width=\textwidth]{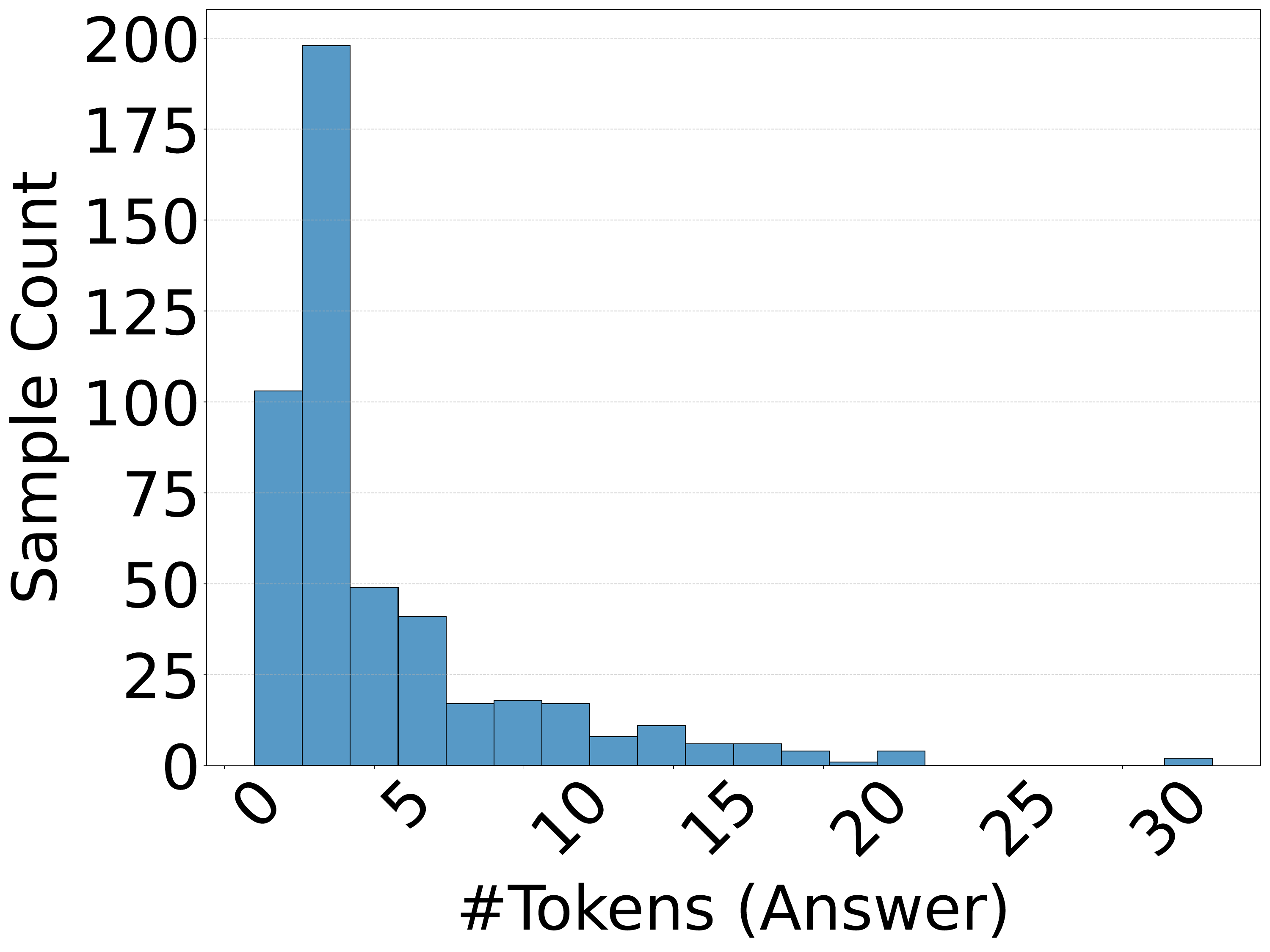}
        \caption{Agriculture.}
    \end{subfigure}
    \begin{subfigure}[b]{0.30\textwidth}
        \centering
        \includegraphics[width=\textwidth]{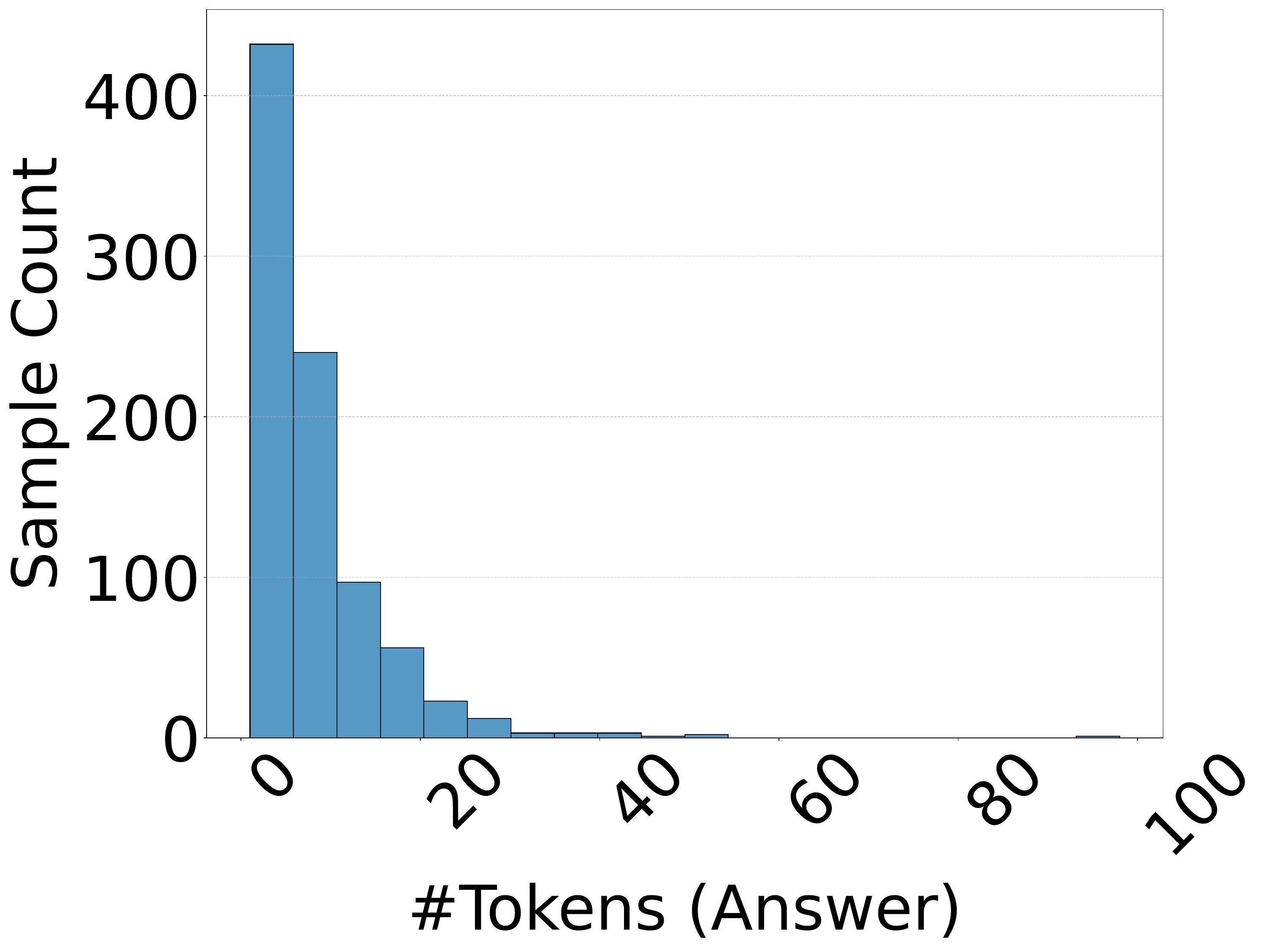}
        \caption{Economics.}
    \end{subfigure}
    \begin{subfigure}[b]{0.30\textwidth}
        \centering
        \includegraphics[width=\textwidth]{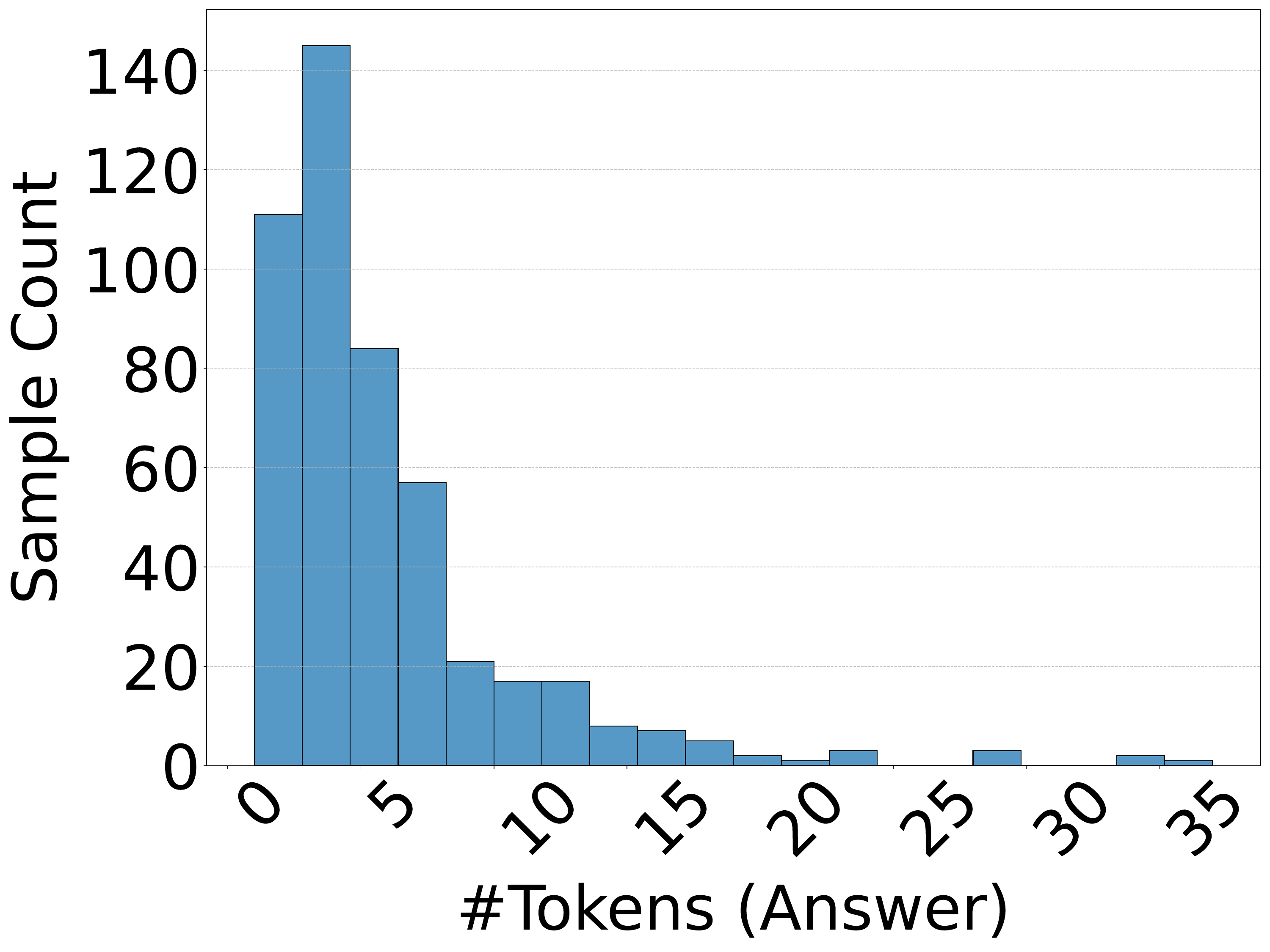}
        \caption{Education.}
    \end{subfigure}
    \\
    \begin{subfigure}[b]{0.30\textwidth}
        \centering
        \includegraphics[width=\textwidth]{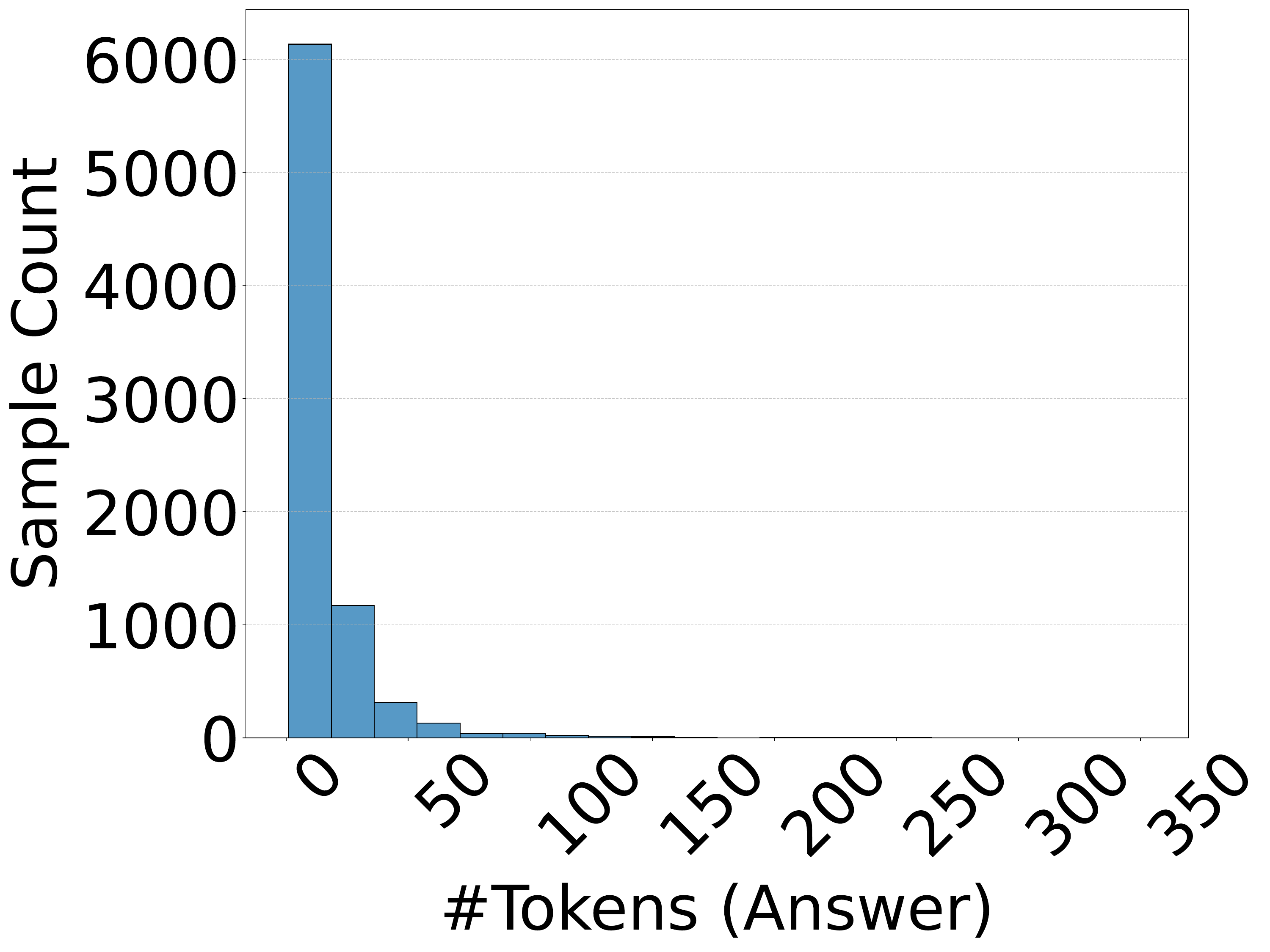}
        \caption{Engineering.}
    \end{subfigure}
    \begin{subfigure}[b]{0.30\textwidth}
        \centering
        \includegraphics[width=\textwidth]{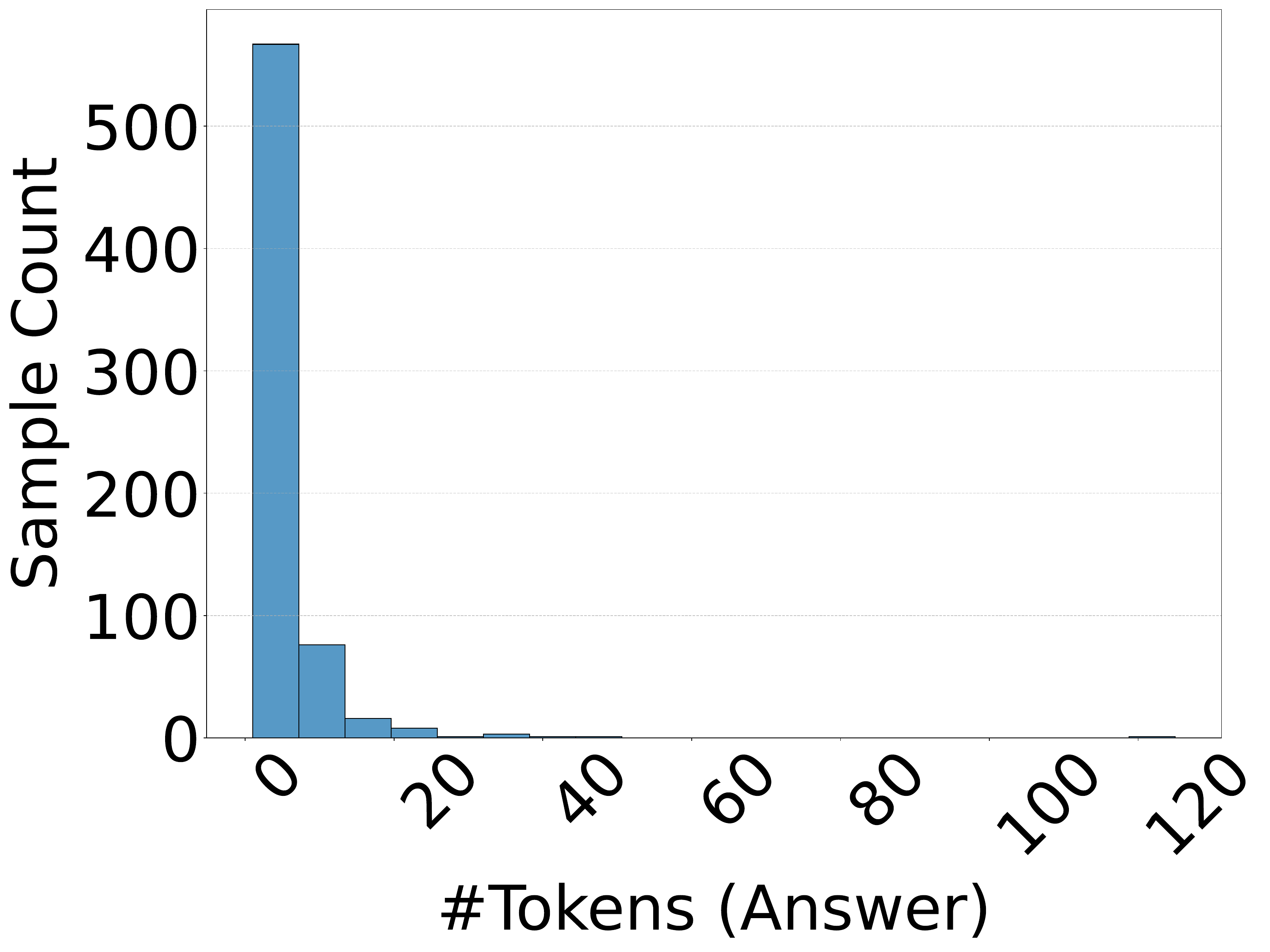}
        \caption{History.}
    \end{subfigure}
    \begin{subfigure}[b]{0.3\textwidth}
        \centering
        \includegraphics[width=\textwidth]{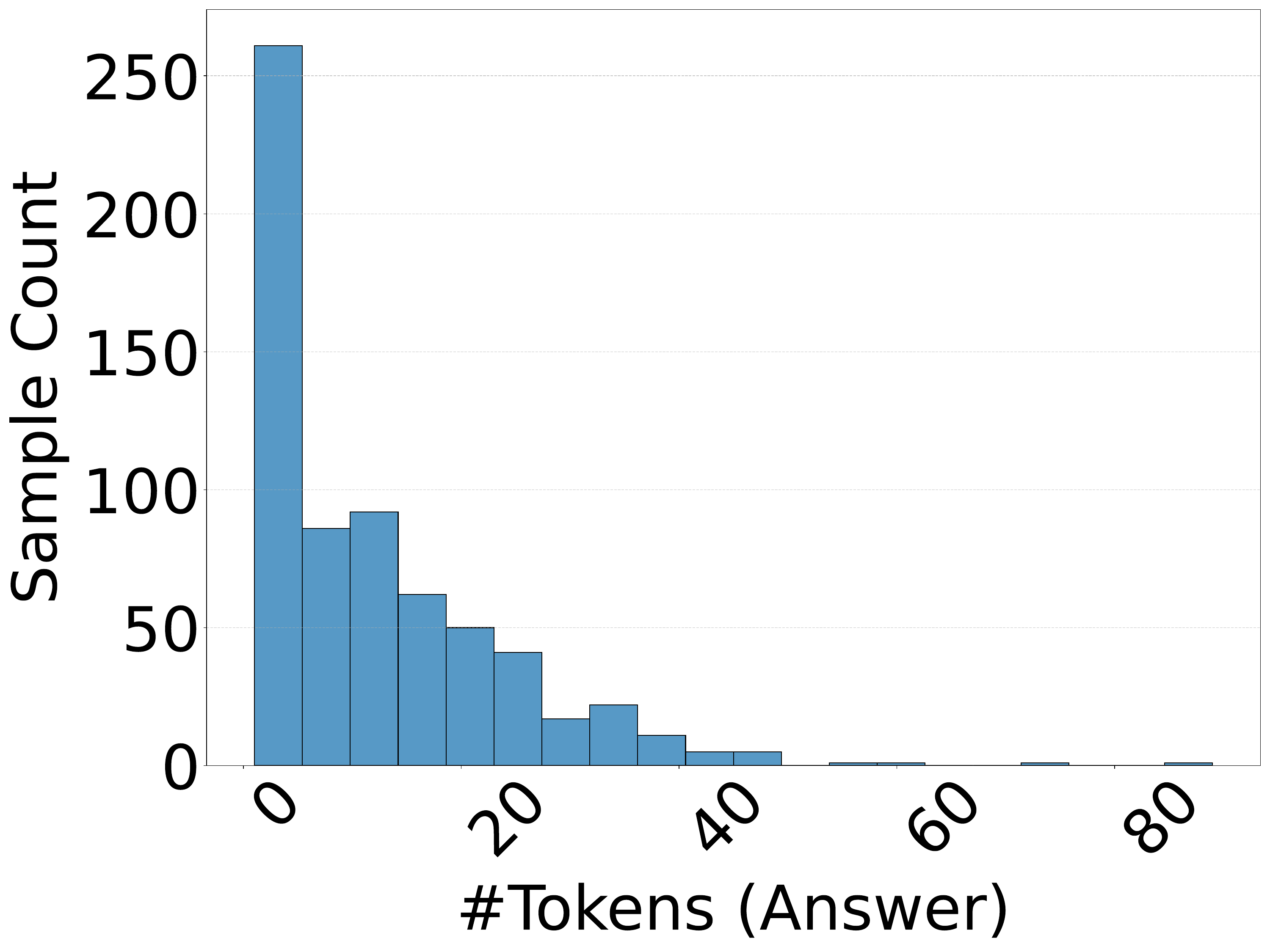}
        \caption{Law.}
    \end{subfigure}
    \\
    \begin{subfigure}[b]{0.30\textwidth}
        \centering
        \includegraphics[width=\textwidth]{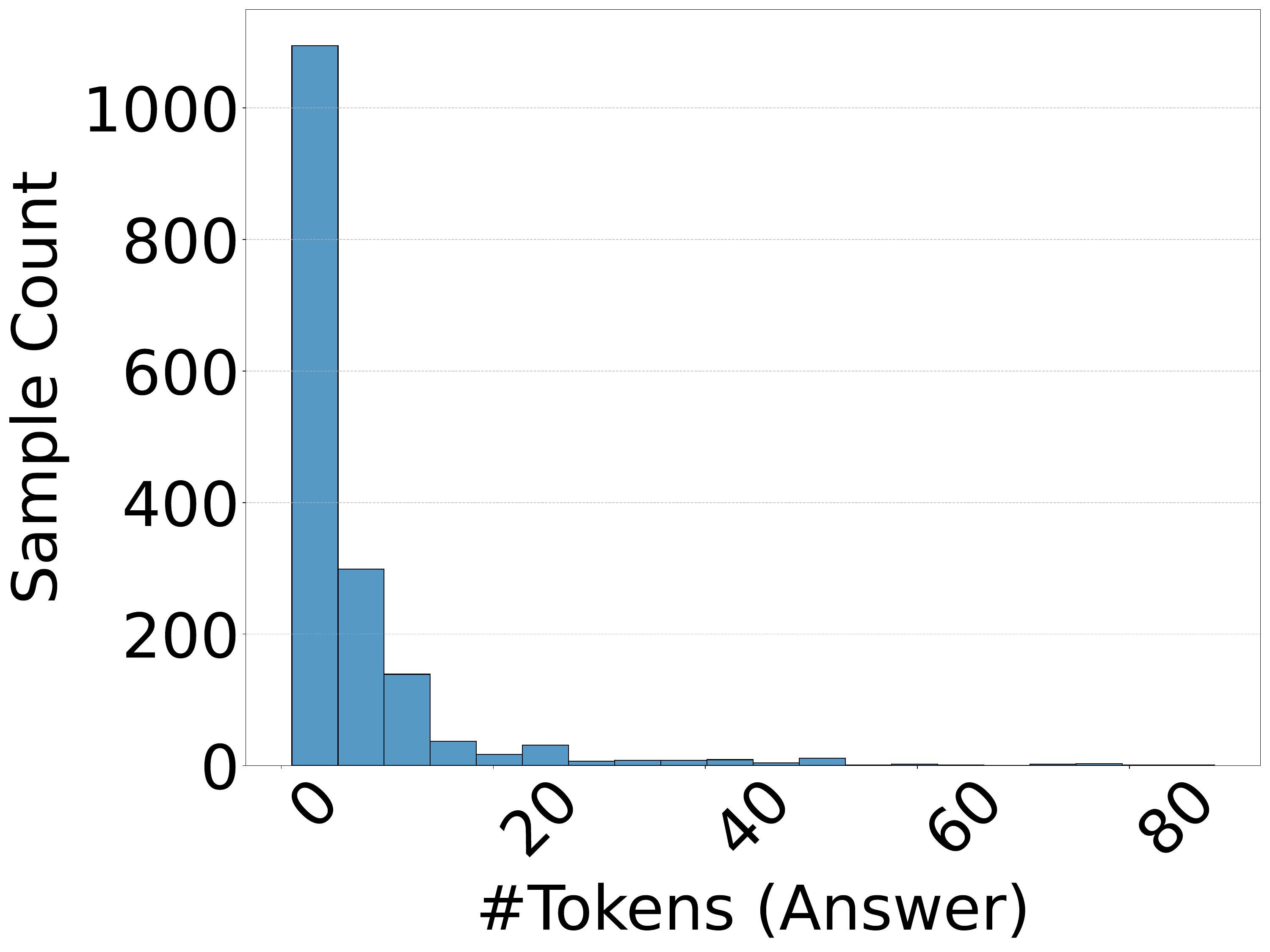}
        \caption{Literature.}
    \end{subfigure}
    \begin{subfigure}[b]{0.30\textwidth}
        \centering
        \includegraphics[width=\textwidth]{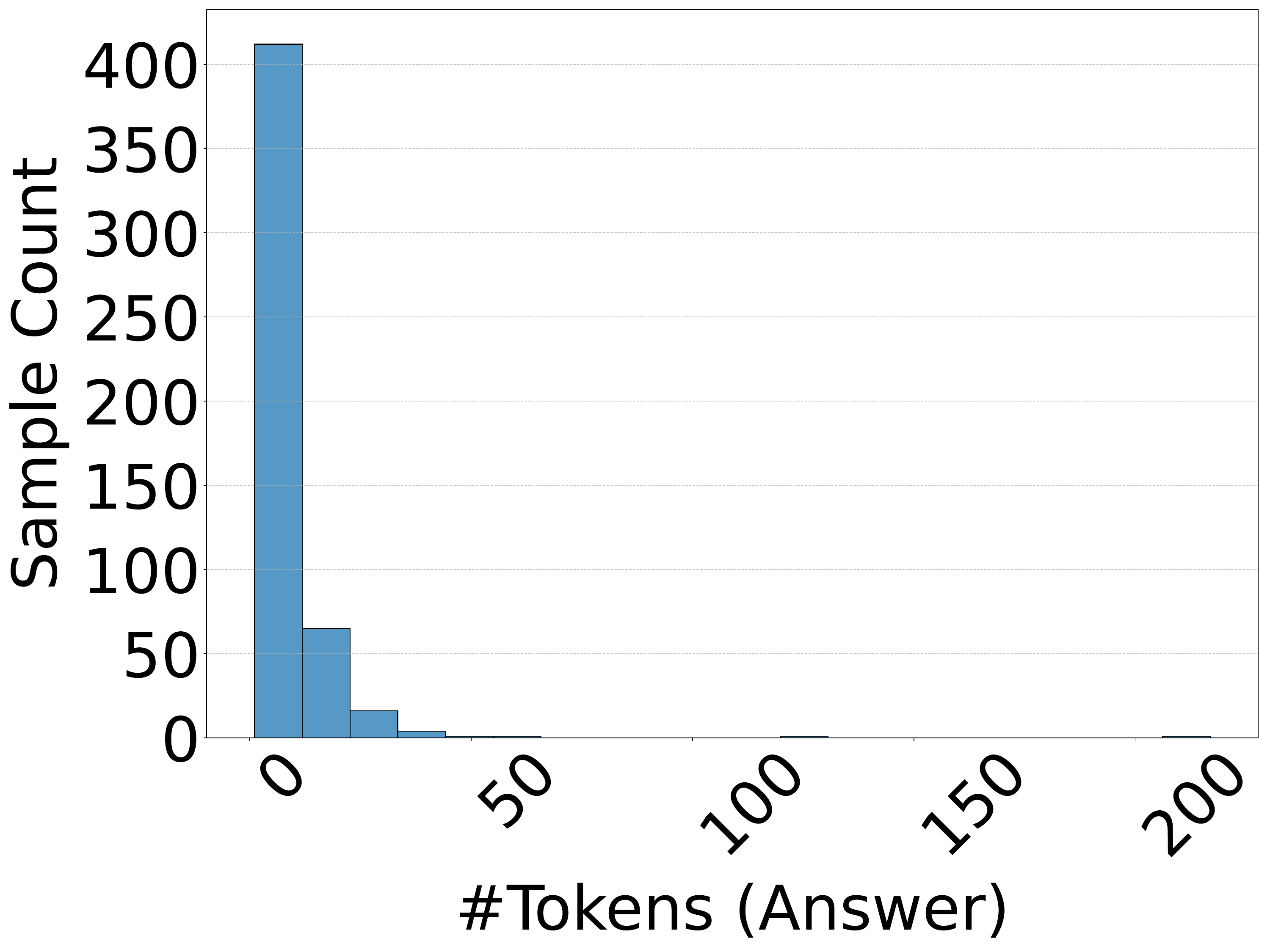}
        \caption{Management.}
    \end{subfigure}
    \begin{subfigure}[b]{0.30\textwidth}
        \centering
        \includegraphics[width=\textwidth]{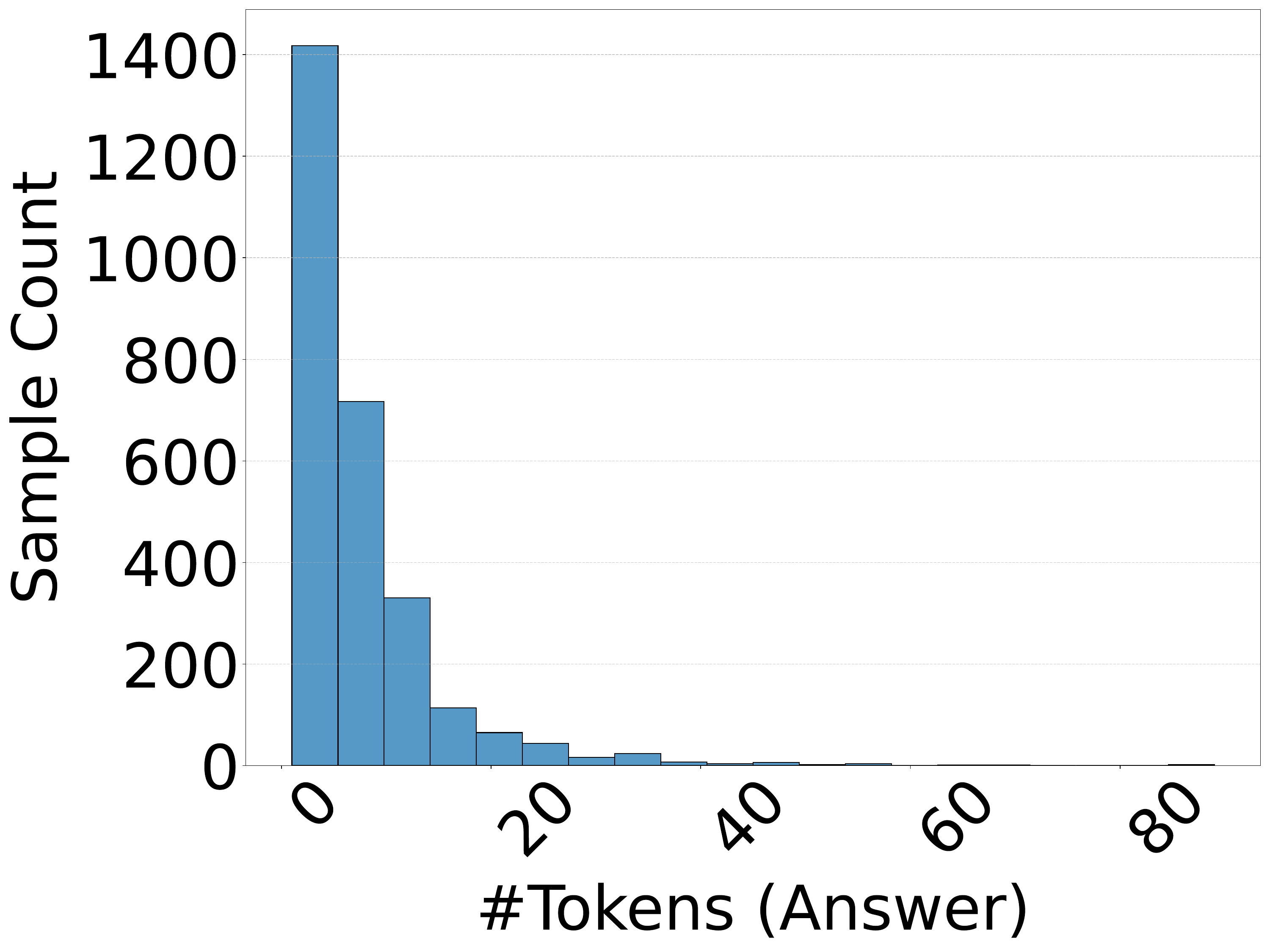}
        \caption{Medicine.}
    \end{subfigure}
    \\
        \begin{subfigure}[b]{0.30\textwidth}
        \centering
        \includegraphics[width=\textwidth]{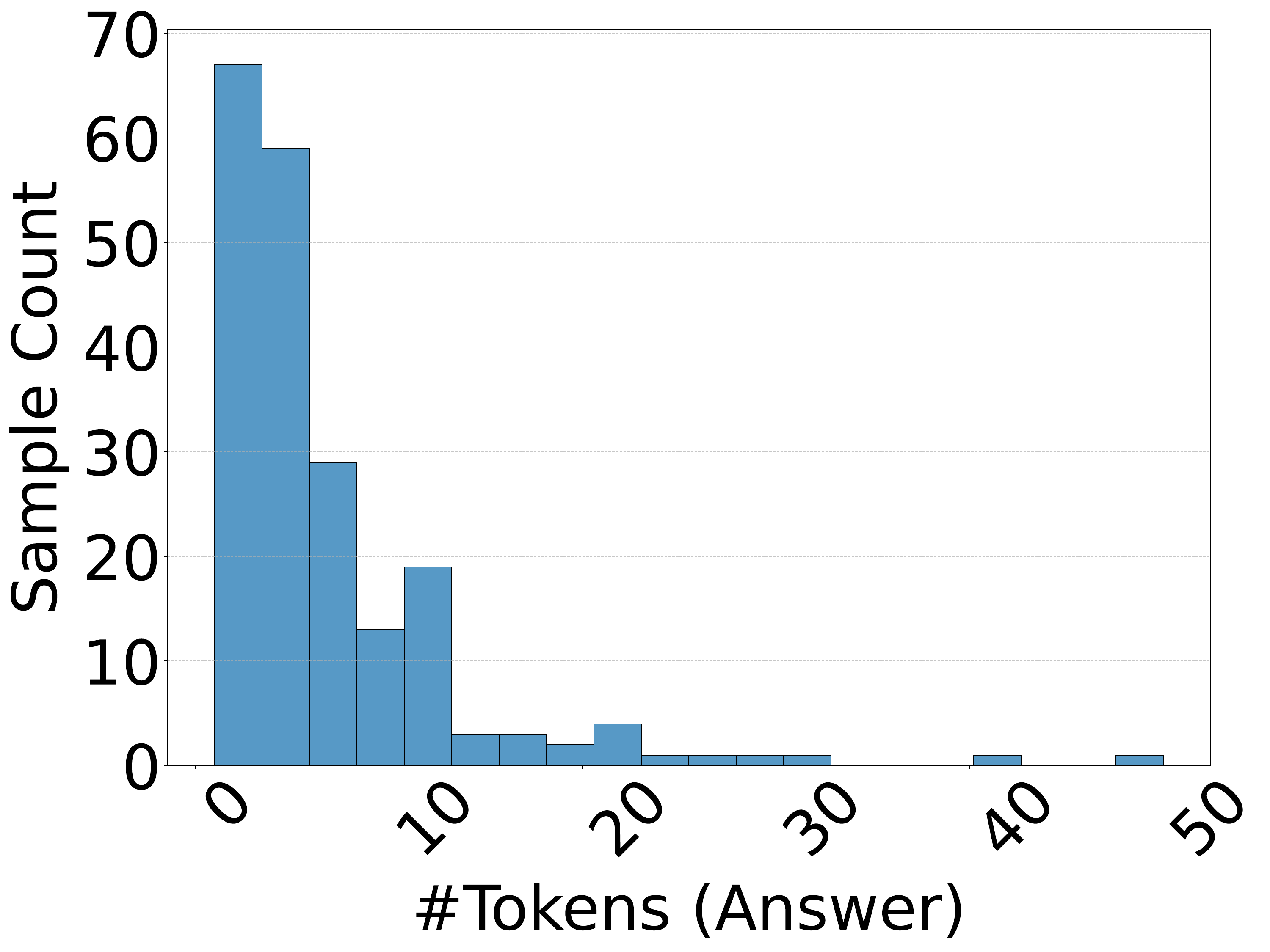}
        \caption{Military Science.}
    \end{subfigure}
    \begin{subfigure}[b]{0.30\textwidth}
        \centering
        \includegraphics[width=\textwidth]{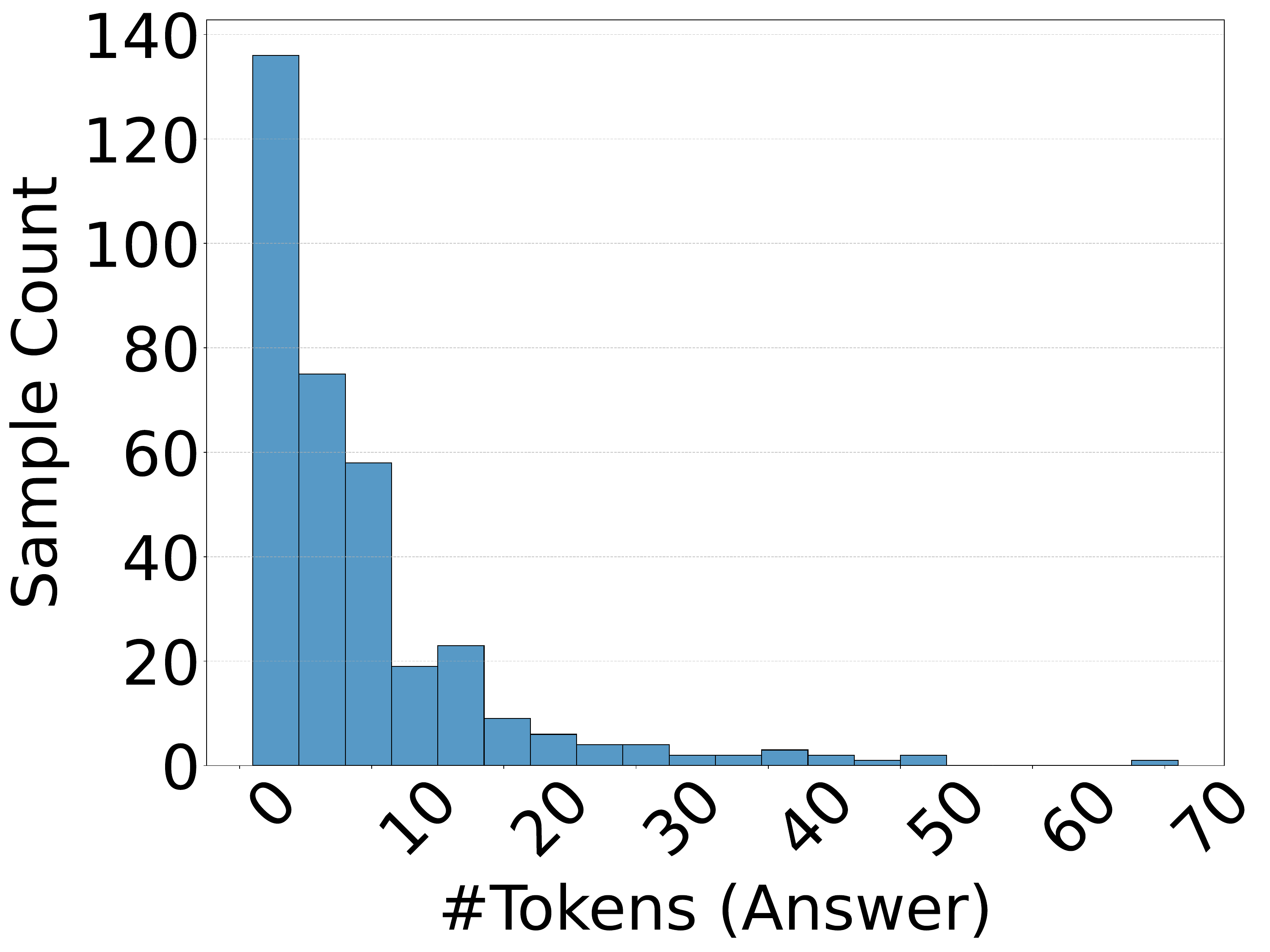}
        \caption{Philosophy.}
    \end{subfigure}
    \begin{subfigure}[b]{0.30\textwidth}
        \centering
        \includegraphics[width=\textwidth]{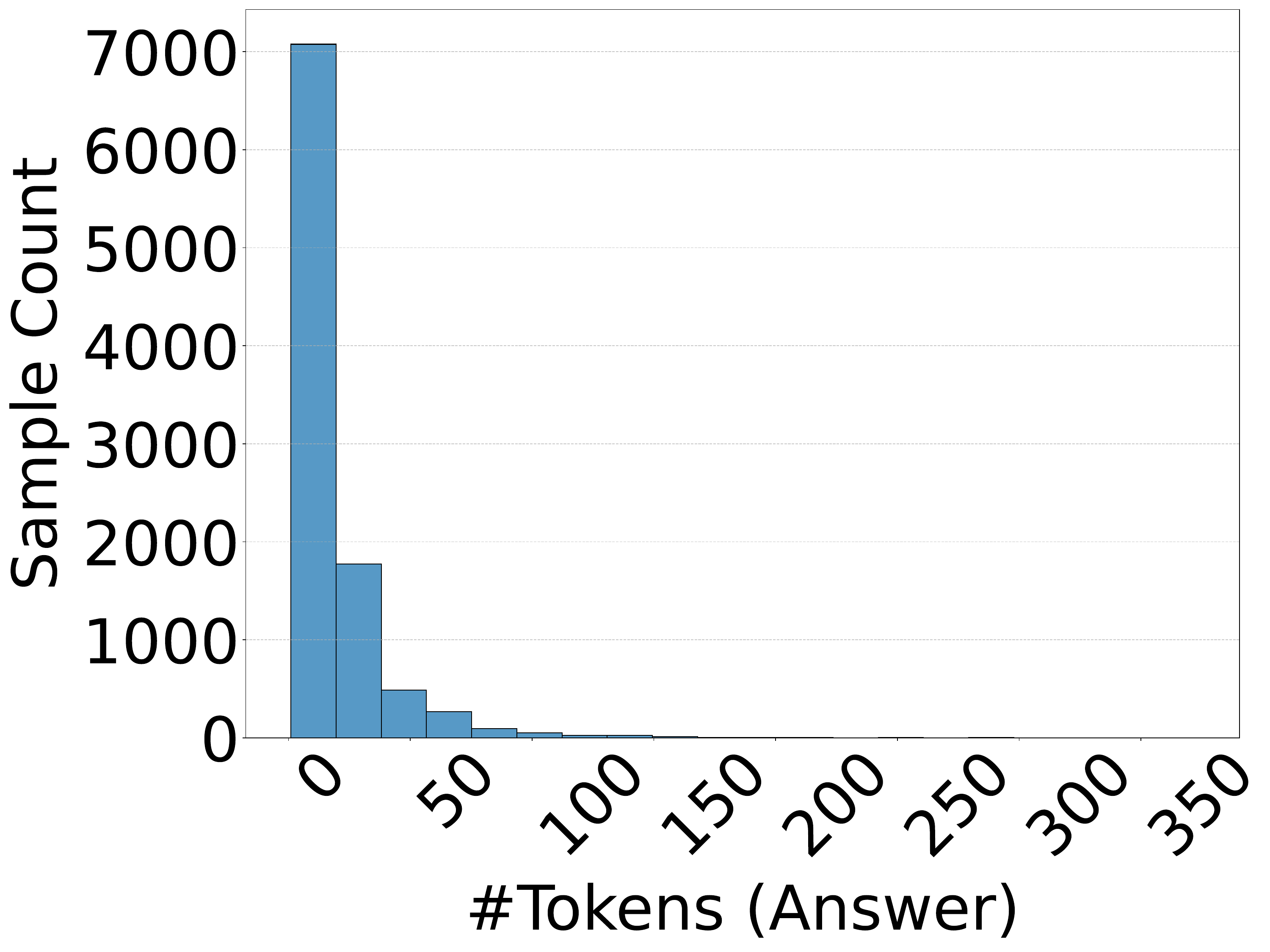}
        \caption{Science.}
    \end{subfigure}
    \\
    \begin{subfigure}[b]{0.30\textwidth}
        \centering
        \includegraphics[width=\textwidth]{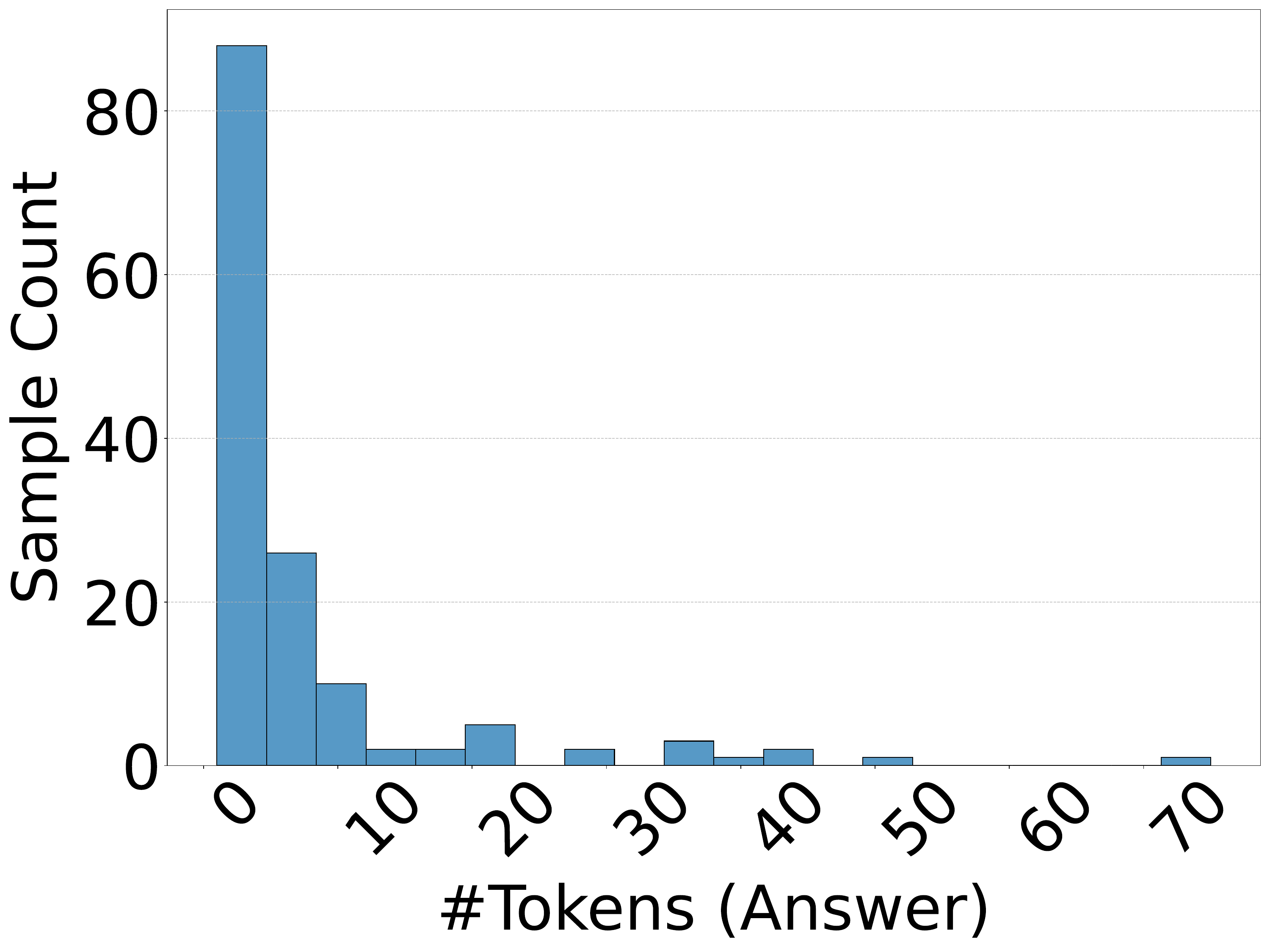}
        \caption{Sociology.}
    \end{subfigure}
    \caption{Answer Length Distribution Across Disciplines.}
    \label{fig:answer_lengths}
\end{figure}

\begin{figure}[H]
    \centering
    \begin{subfigure}[b]{0.30\textwidth}
        \centering
        \includegraphics[width=\textwidth]{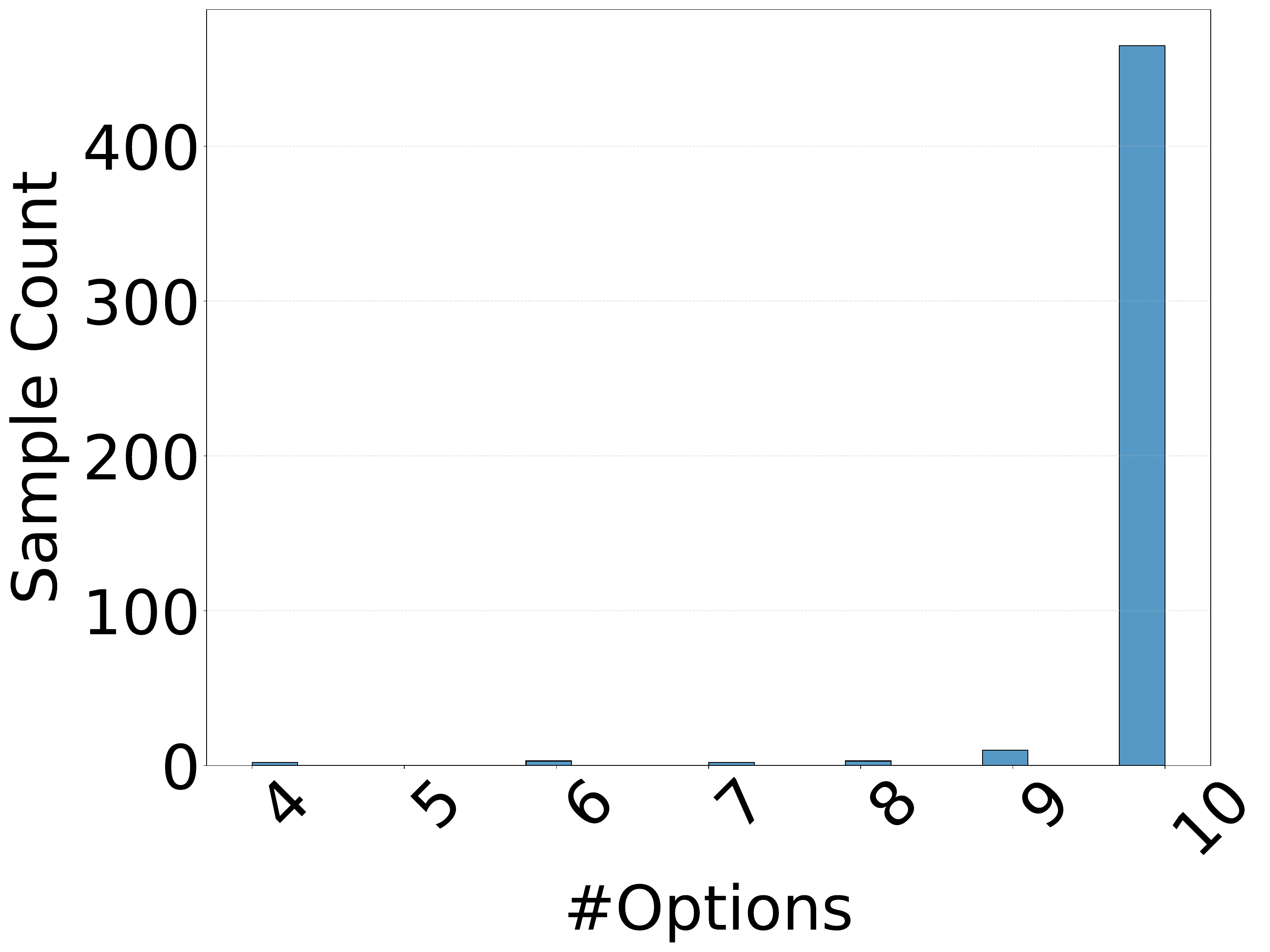}
        \caption{Agriculture.}
    \end{subfigure}
    \begin{subfigure}[b]{0.30\textwidth}
        \centering
        \includegraphics[width=\textwidth]{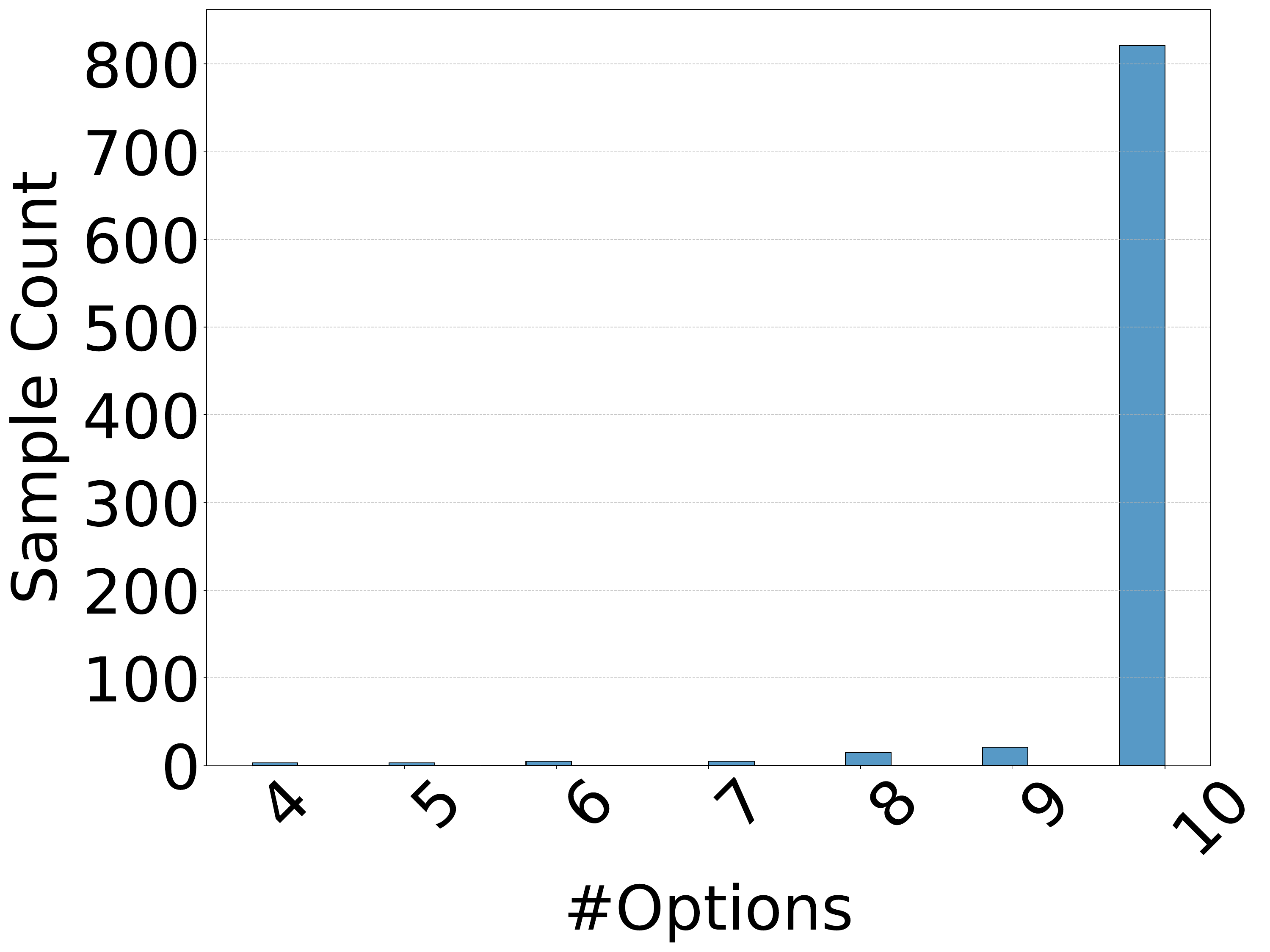}
        \caption{Economics.}
    \end{subfigure}
    \begin{subfigure}[b]{0.30\textwidth}
        \centering
        \includegraphics[width=\textwidth]{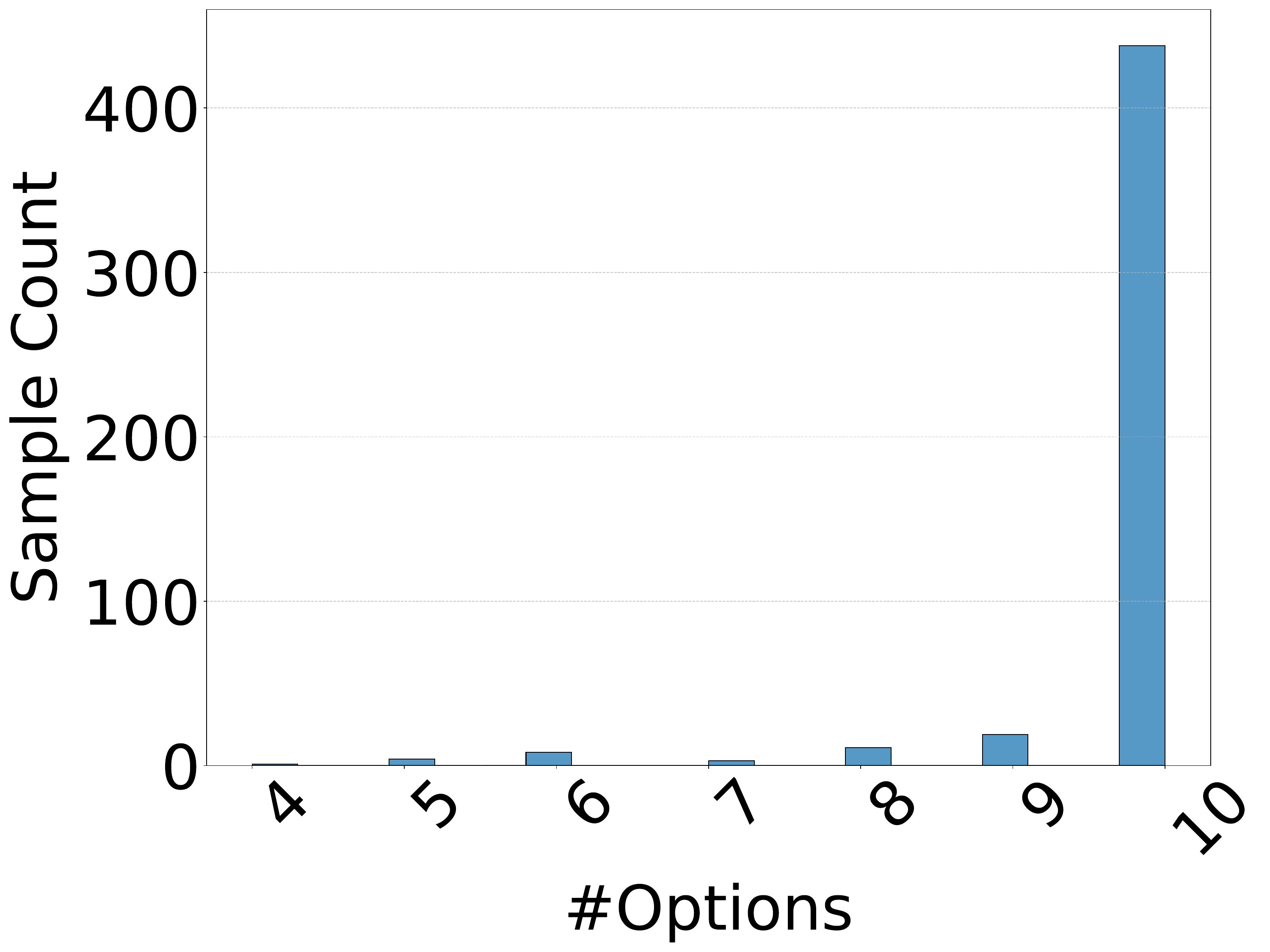}
        \caption{Education.}
    \end{subfigure}
    \\
    \begin{subfigure}[b]{0.30\textwidth}
        \centering
        \includegraphics[width=\textwidth]{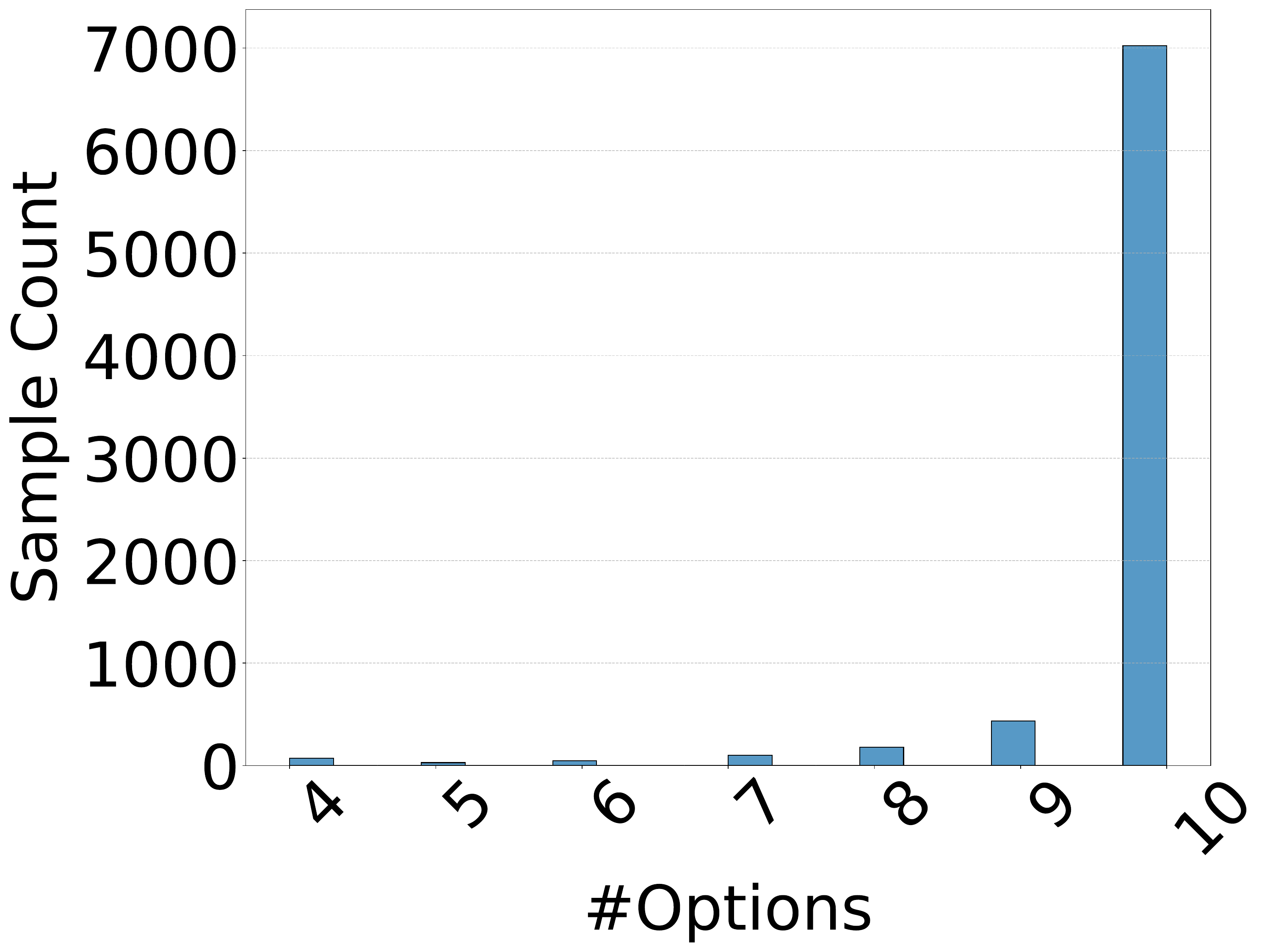}
        \caption{Engineering.}
    \end{subfigure}
    \begin{subfigure}[b]{0.30\textwidth}
        \centering
        \includegraphics[width=\textwidth]{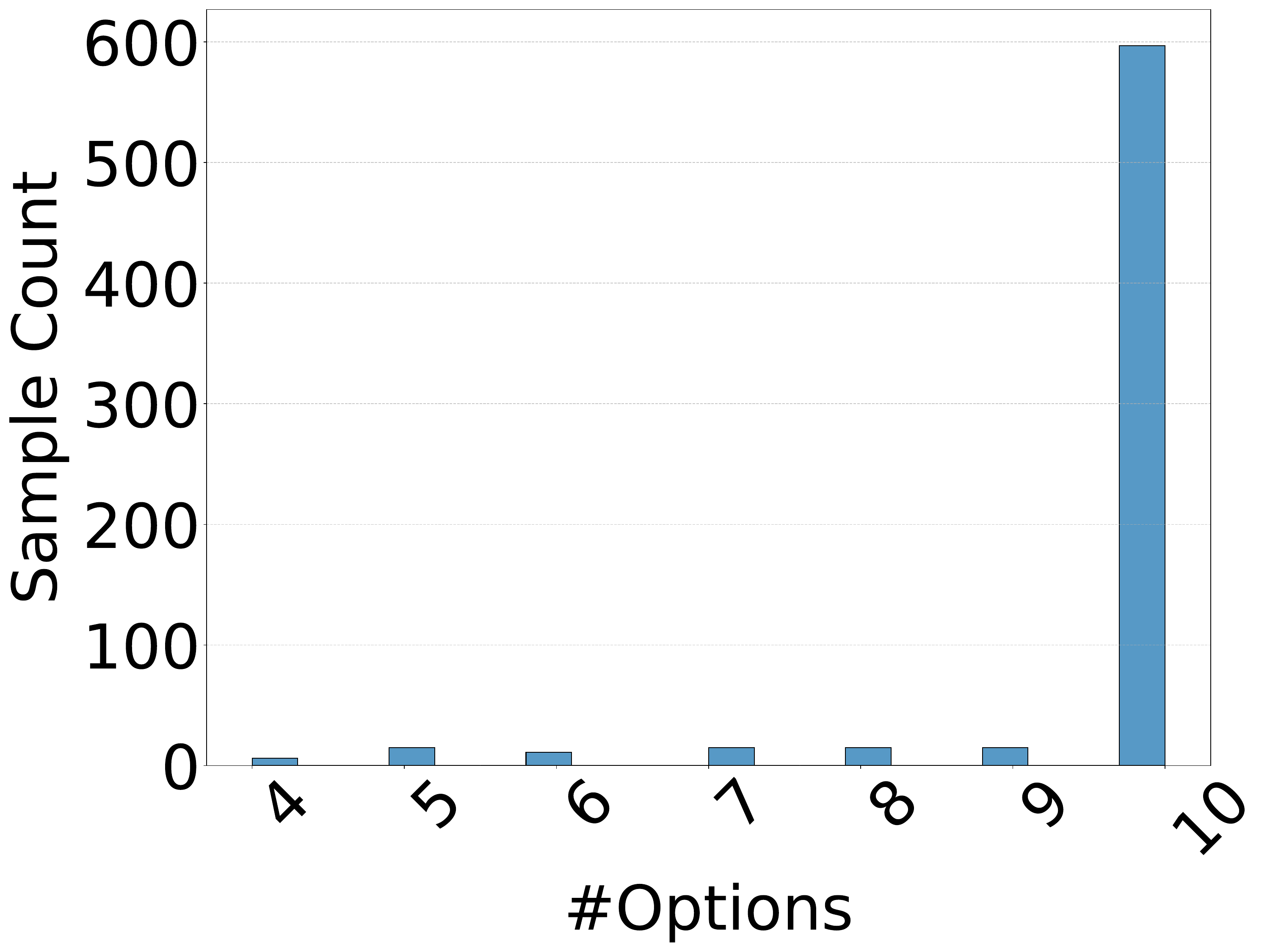}
        \caption{History.}
    \end{subfigure}
    \begin{subfigure}[b]{0.30\textwidth}
        \centering
        \includegraphics[width=\textwidth]{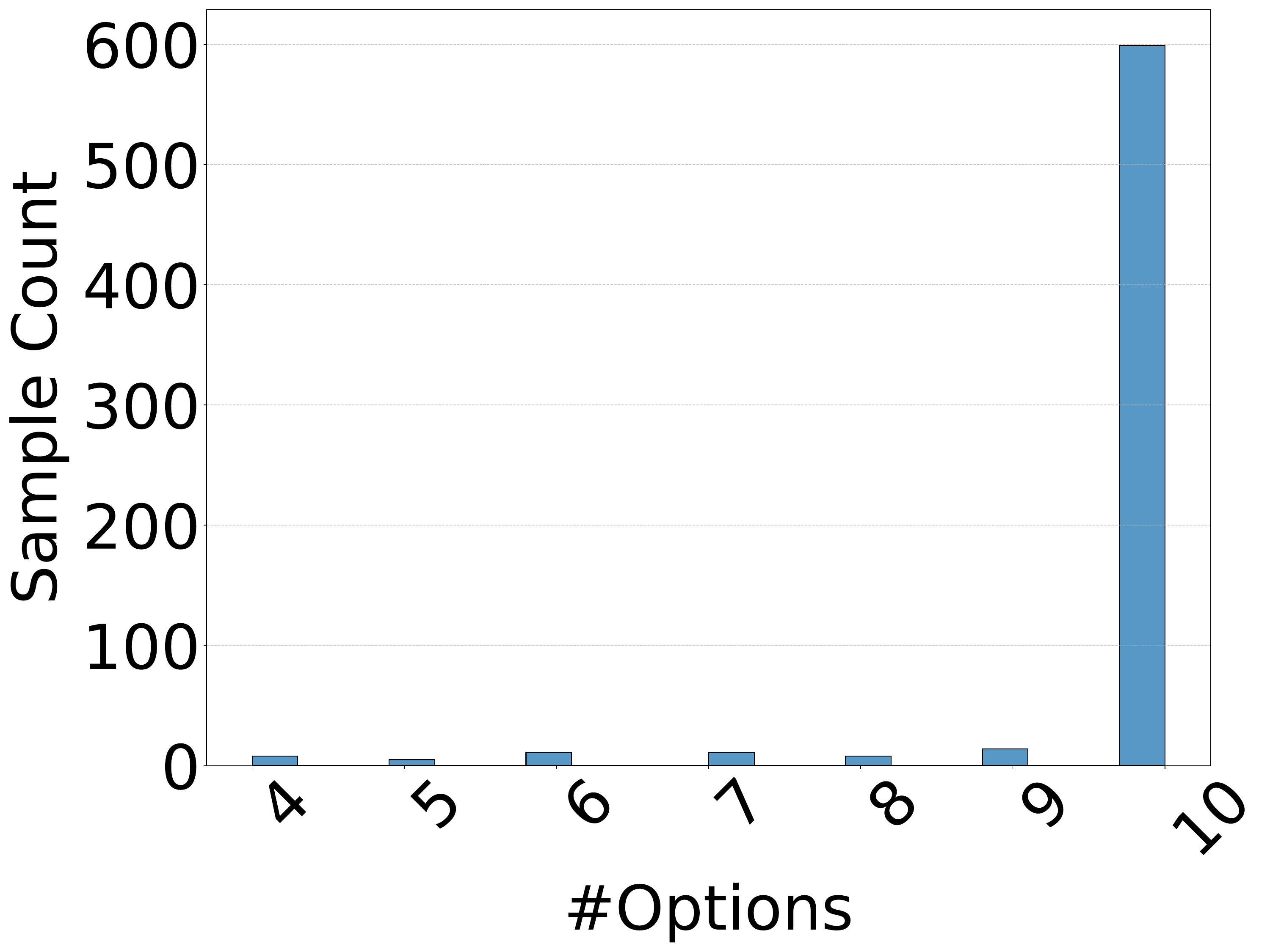}
        \caption{Law.}
    \end{subfigure}
    \\
    \begin{subfigure}[b]{0.30\textwidth}
        \centering
        \includegraphics[width=\textwidth]{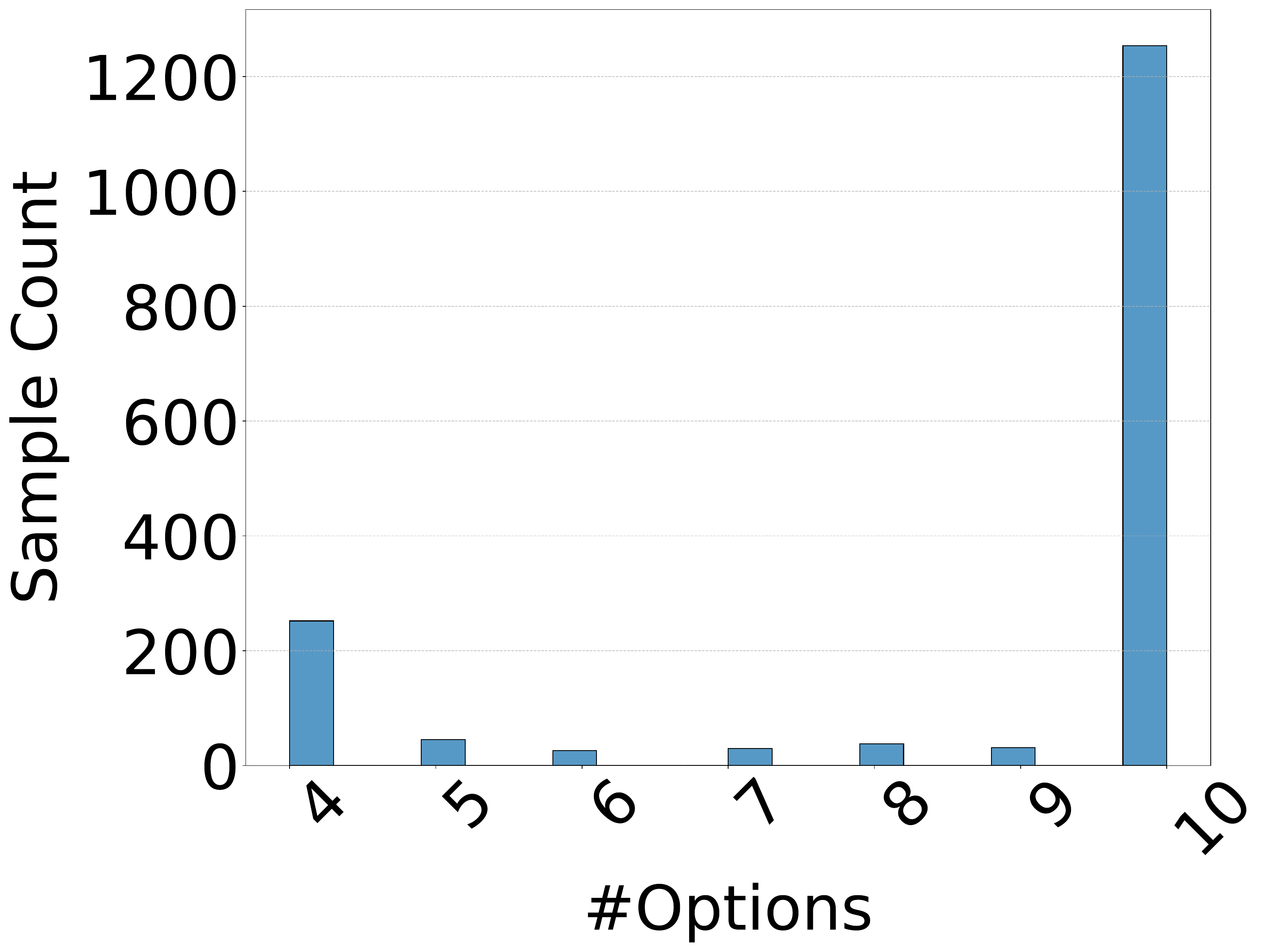}
        \caption{Literature.}
    \end{subfigure}
    \begin{subfigure}[b]{0.30\textwidth}
        \centering
        \includegraphics[width=\textwidth]{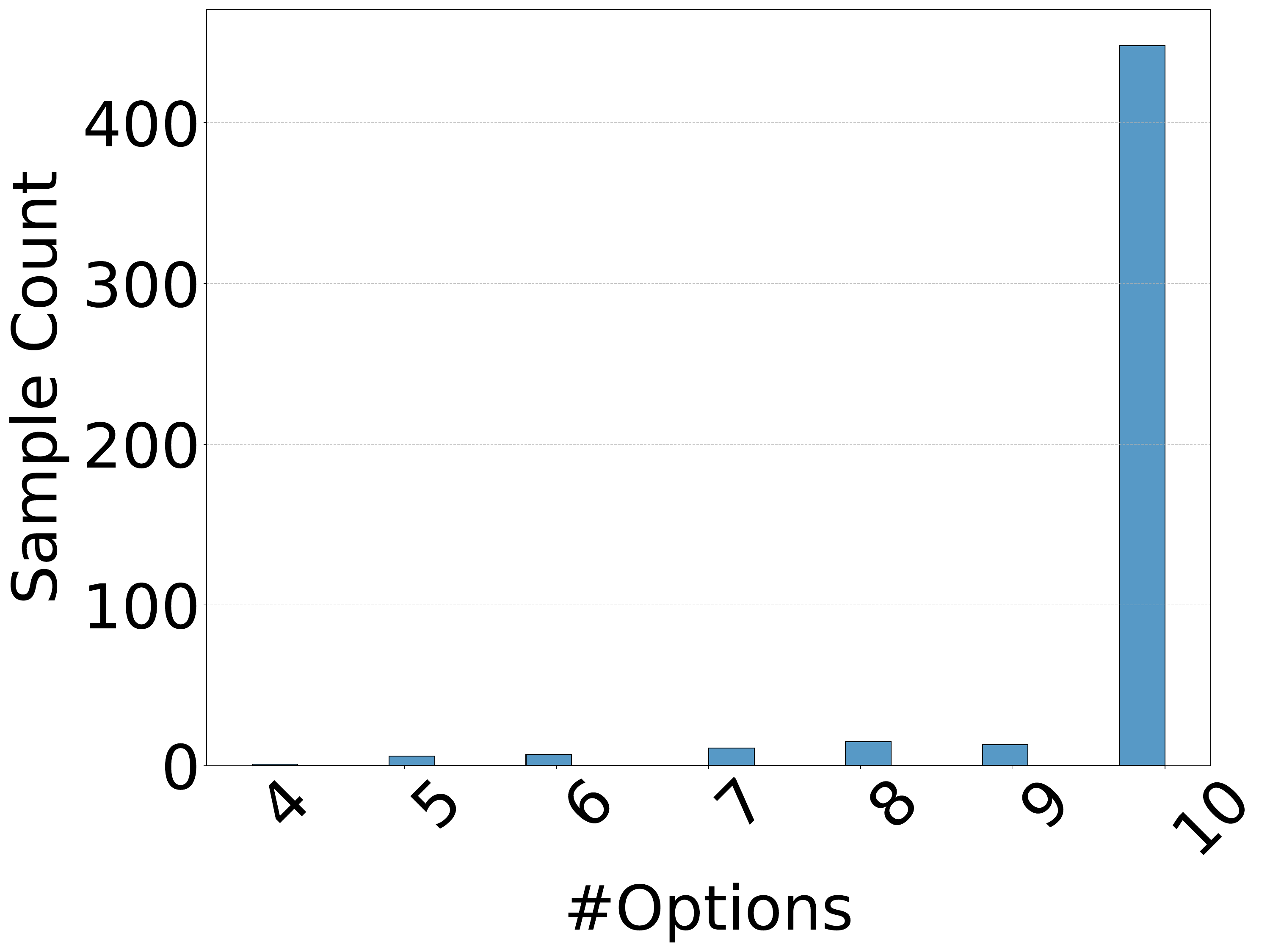}
        \caption{Management.}
    \end{subfigure}
    \begin{subfigure}[b]{0.30\textwidth}
        \centering
        \includegraphics[width=\textwidth]{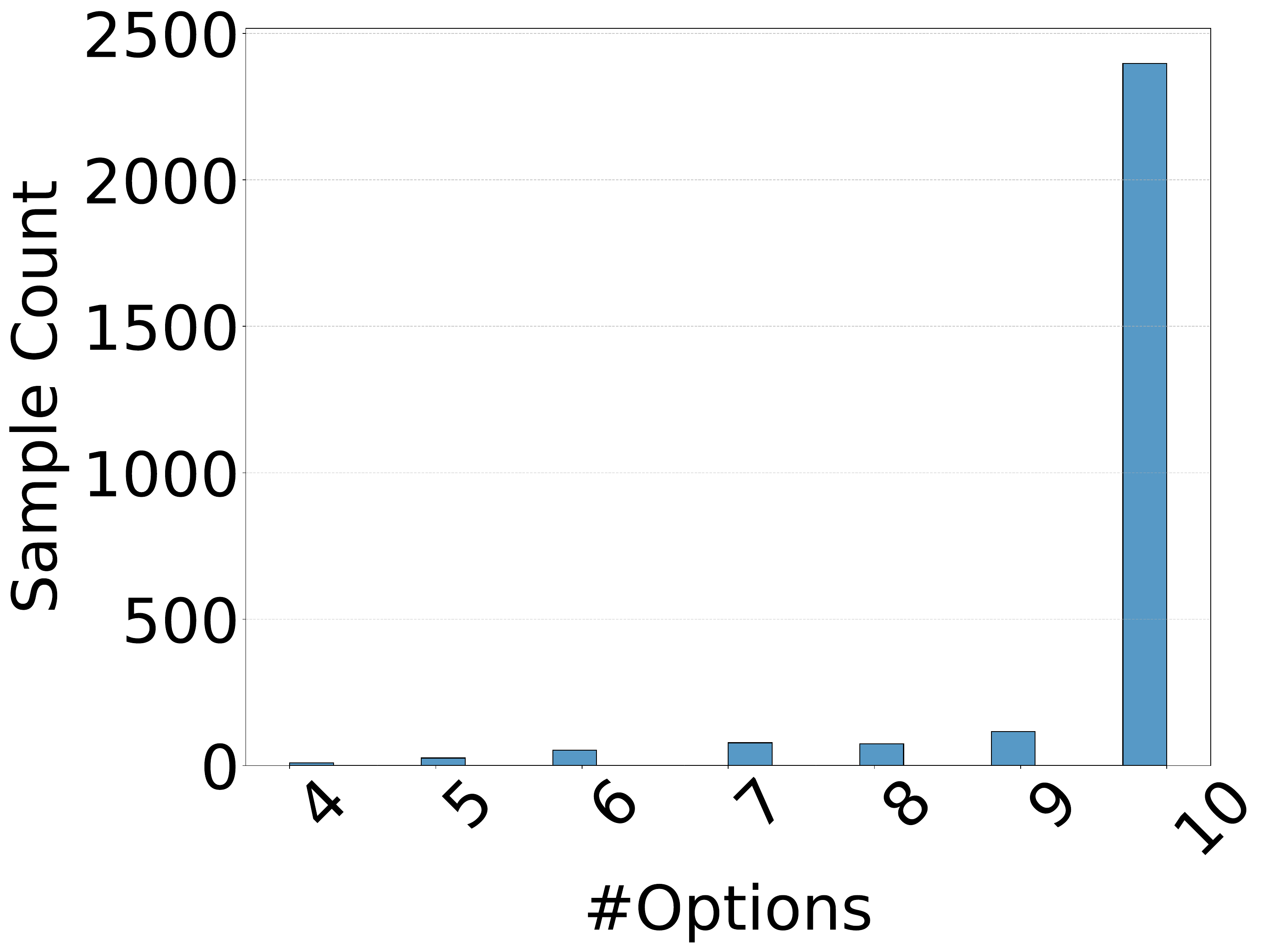}
        \caption{Medicine.}
    \end{subfigure}
    \\
        \begin{subfigure}[b]{0.30\textwidth}
        \centering
        \includegraphics[width=\textwidth]{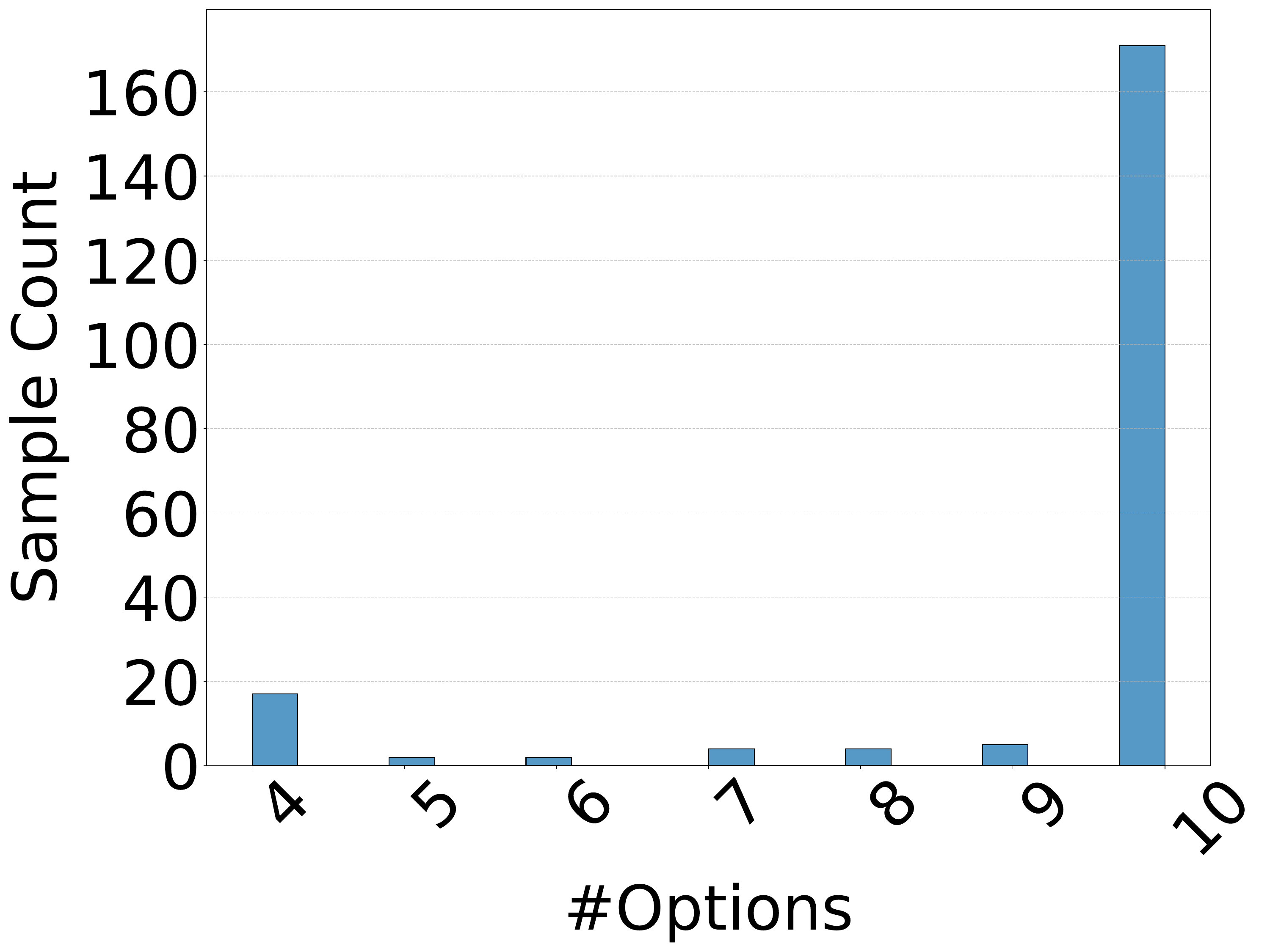}
        \caption{Military Science.}
    \end{subfigure}
    \begin{subfigure}[b]{0.30\textwidth}
        \centering
        \includegraphics[width=\textwidth]{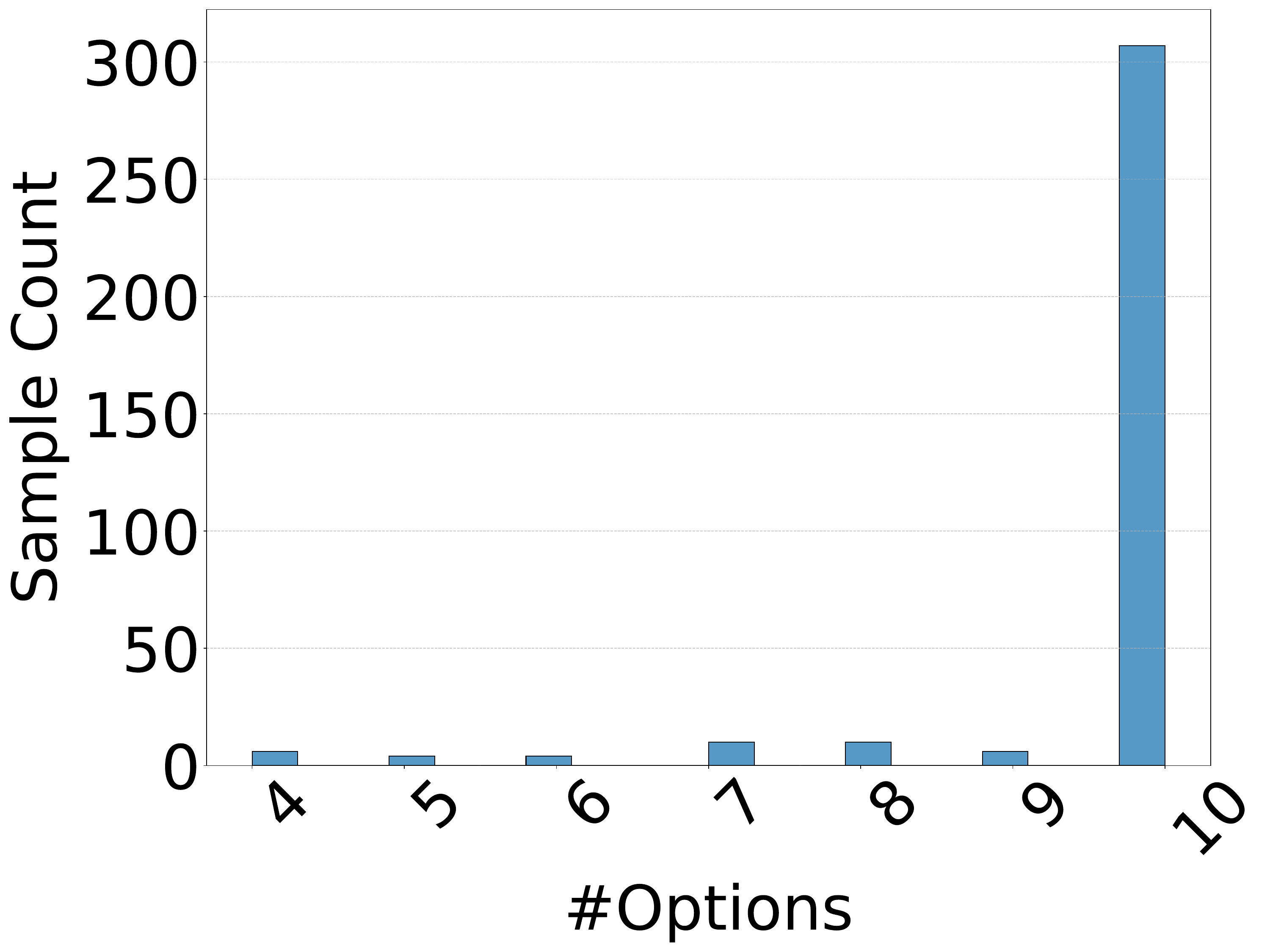}
        \caption{Philosophy.}
    \end{subfigure}
    \begin{subfigure}[b]{0.30\textwidth}
        \centering
        \includegraphics[width=\textwidth]{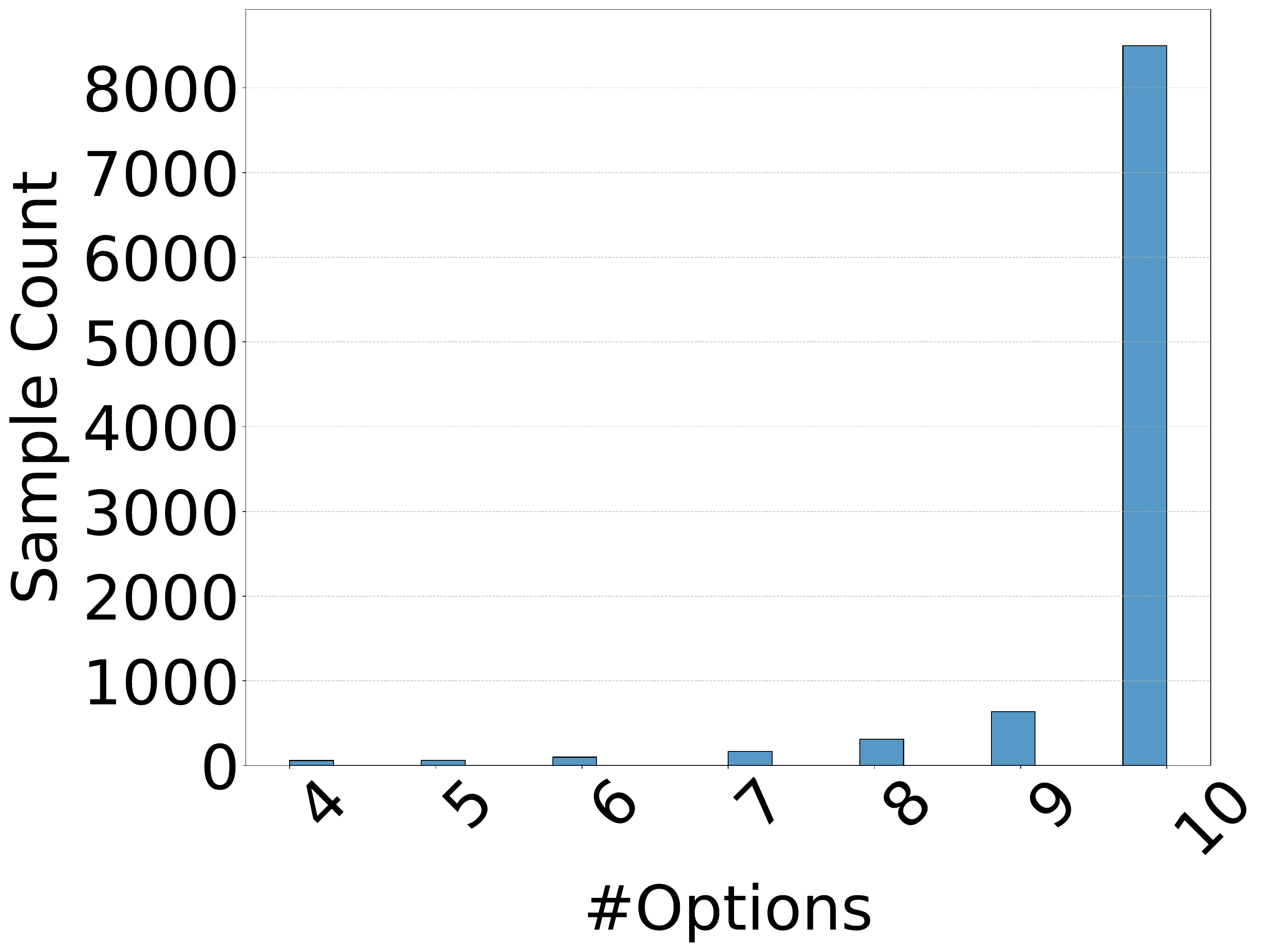}
        \caption{Science.}
    \end{subfigure}
    \\
    \begin{subfigure}[b]{0.30\textwidth}
        \centering
        \includegraphics[width=\textwidth]{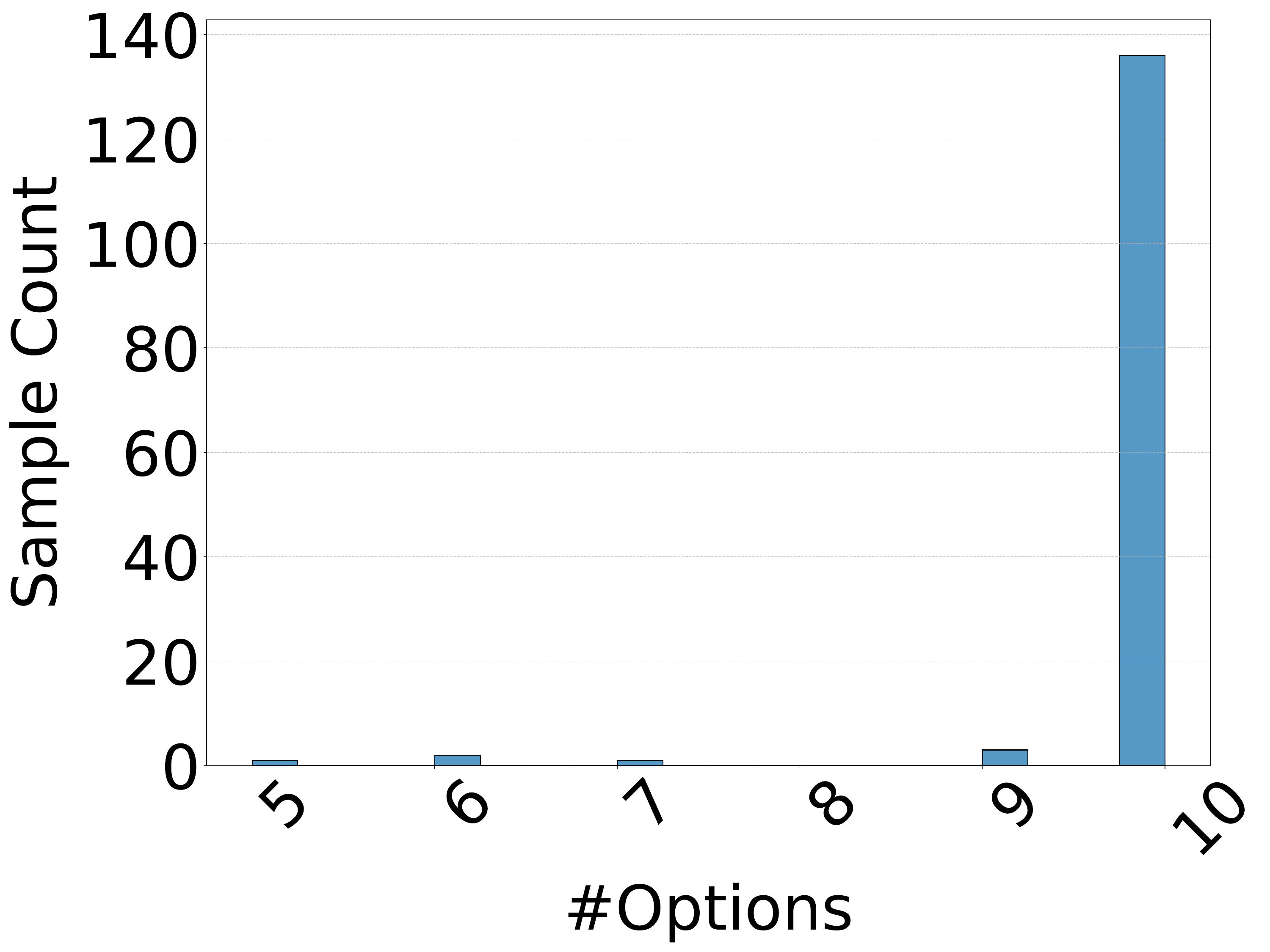}
        \caption{Sociology.}
    \end{subfigure}
    \caption{Option Count Distribution Across Disciplines.}
    \label{fig:option_counts}
\end{figure}

\section{Evaluation Prompt}
\label{appendix:Evaluation Prompt}
\subsection{Zero-shot Prompt}

\begin{promptbox}{Zero-shot}
Answer the following multiple-choice question. There is only one correct answer. The last line of
your response should be in the format ’Answer: \$LETTER (without quotes), where LETTER is
one of A, B, C, D, E, F, G, H, I, or J.
\begin{verbatim}
{Question}
\end{verbatim}
\end{promptbox}

\begin{promptbox}{Question format example:}
Question: A microwave oven is connected to an outlet, 120 V, and draws a current of 2 amps. At what rate is energy being used by the microwave oven? 
\\
A) 240 W\\
B) 120 W\\
C) 10 W\\
D) 480 W\\
E) 360 W\\
F) 200 W\\
G) 30 W\\
H) 150 W\\
I) 60 W\\
J) 300 W\\
\end{promptbox}

\subsection{Five-shot Prompt}
\begin{promptbox}{Five-shot}
Answer the following multiple-choice question. There is only one correct answer. The last line of
your response should be in the format ’Answer: \$LETTER (without quotes), where LETTER is
one of A, B, C, D, E, F, G, H, I, or J.\\

Question: A refracting telescope consists of two converging lenses separated by 100 cm. The eye-piece lens has a focal length of 20 cm. The angular magnification of the telescope is ( ).
\\
A) 10
\\
B) 40
\\
C) 6
\\
D) 25
\\
E) 15
\\
F) 50
\\
G) 30
\\
H) 4
\\
I) 5
\\
J) 20
\\
Answer: Let's think step by step. In a refracting telescope, if both lenses are converging, the focus of both lenses must be between the two lenses, and thus the focal lengths of the two lenses must add up to their separation. Since the focal length of one lens is 20 cm, the focal length of the other must be 80 cm. The magnification is the ratio of these two focal lengths, or 4.\\
Answer: H.\\

Question: Say the pupil of your eye has a diameter of 5 mm and you have a telescope with an aperture of 50 cm. How much more light can the telescope gather than your eye? 
\\
A) 1000 times more
\\
B) 50 times more
\\
C) 5000 times more
\\
D) 500 times more
\\
E) 10000 times more
\\
F) 20000 times more
\\
G) 2000 times more
\\
H) 100 times more
\\
I) 10 times more
\\
J) N/A
\\
Answer: Let's think step by step. The amount of light a telescope can gather compared to the human eye is proportional to the area of its apertures. The area of a circle is given by the formula $A = \pi \left(\frac{{D}}{{2}}\right)^2$, where $D$ is the diameter. Therefore, the relative light-gathering power is calculated as:
    \[
    \frac{{\left(\frac{{50 \text{{ cm}}}}{{2}}\right)^2}}{{\left(\frac{{5 \text{{ mm}}}}{{2}}\right)^2}} = \frac{{\left(\frac{{50 \text{{ cm}}}}{{0.1 \text{{ cm}}}}\right)^2}}{{\left(\frac{{5 \text{{ mm}}}}{{0.1 \text{{ cm}}}}\right)^2}} = \frac{{500^2}}{{5^2}} = 10000.
    \]
Answer: E.\\

Question: Where do most short-period comets come from and how do we know? \\
A) The Kuiper belt; short period comets tend to be in the plane of the solar system like the Kuiper belt. \\
B) The asteroid belt; short period comets tend to come from random directions indicating a spherical distribution of comets called the asteroid belt. \\
C) The asteroid belt; short period comets tend to be in the plane of the solar system just like the asteroid belt.\\ 
D) The Oort cloud; short period comets have orbital periods similar to asteroids like Vesta and are found in the plane of the solar system just like the Oort cloud.\\ 
E) The Oort Cloud; short period comets tend to come from random directions indicating a spherical distribution of comets called the Oort Cloud. \\
F) The Oort cloud; short period comets tend to be in the plane of the solar system just like the Oort cloud. \\
G) The asteroid belt; short period comets have orbital periods similar to asteroids like Vesta and are found in the plane of the solar system just like the asteroid belt.\\ 
Answer: Let's think step by step. Most short-period comets originate from the Kuiper belt. This is deduced from the observation that these comets tend to follow orbits that lie in the plane of the solar system, similar to the distribution of objects in the Kuiper belt itself. Thus, the alignment of these cometary orbits with the ecliptic plane points to their Kuiper belt origin.\\
Answer: A.\\

Question: Colors in a soap bubble result from light ( ). \\
A) dispersion \\
B) deflection \\
C) refraction \\
D) reflection \\
E) interference \\
F) converted to a different frequency \\
G) polarization \\
H) absorption \\
I) diffraction \\
J) transmission \\
Answer: 
Let's think step by step.The colorful patterns observed in a soap bubble are caused by the phenomenon of light interference. This occurs when light waves bounce between the two surfaces of the soap film, combining constructively or destructively based on their phase differences and the varying thickness of the film. These interactions result in vibrant color patterns due to variations in the intensity of different wavelengths of light.
\\
Answer: E.\\

Question: A microwave oven is connected to an outlet, 120 V, and draws a current of 2 amps. At what rate is energy being used by the microwave oven? \\
A) 240 W\\
B) 120 W\\
C) 10 W\\
D) 480 W\\
E) 360 W\\
F) 200 W\\
G) 30 W\\
H) 150 W\\
I) 60 W\\
J) 300 W\\
Answer: Let's think step by step. The rate of energy usage, known as power, in an electrical circuit is calculated by the product of voltage and current. For a microwave oven connected to a 120 V outlet and drawing a current of 2 amps, the power consumption can be calculated as follows:
    \[
    \text{{Power}} = \text{{Voltage}} \times \text{{Current}} = 120 \, \text{{V}} \times 2 \, \text{{A}} = 240 \, \text{{W}}.
    \]
    Therefore, the microwave oven uses energy at a rate of 240 watts.\\
Answer: A.
\begin{verbatim}
{Question}

Answer: Let's think step by step. 
\end{verbatim}
\end{promptbox}

\section{Further Experiment Analysis}

\subsection{Impact of Subfield Information}
\label{appendix:Impact of Subfield Information}

We conduct zero-shot evaluations under two conditions to investigate the impact of subfield information. 
1) zero-shot-with-subfield, where the prompt includes a description of the problem’s subfield, and (2)zero-shot-without-subfield, where no subfield information is provided.The prompts employed for the zero-shot-with-subfield evaluation are presented below.

\begin{promptbox}{Zero-shot-with-subfield}
Answer the following multiple choice question about 
\begin{verbatim}
{subfield}
\end{verbatim}
There is only one correct answer. The last line of your response should be in the format 'Answer: \$LETTER' (without quotes), where LETTER is one of A, B, C, D, E, F, G, H, I, or J.
\end{promptbox}
\vspace{0.5em}  
\begin{promptbox}{Question format example:}

The common-mode rejection ratio of the first stage amplification circuit in a three-op-amp differential circuit is determined by ( ).  

A) the absolute value of the difference in the common-mode rejection ratio of A1 and A2 themselves  
\\
B) all of the above  
\\
C) the average of A1 and A2's common-mode rejection ratios  
\\
D) the sum of A1 and A2's common-mode rejection ratios  
\\
E) the product of A1 and A2's common-mode rejection ratios  
\\
F) the square root of the product of A1 and A2's common-mode rejection ratios  
\\
G) the size of A2's common-mode rejection ratio  
\\
H) the size of A1's common-mode rejection ratio  
\\
I) The difference in the common-mode rejection ratio of A1 and A2 themselves  
\\
J) input resistance  
\end{promptbox}

\subsection{Robustness of the Evaluation}
\label{appendix:robustness}
We employ 4 types of initial prompts and 6 types of question formats, resulting in a
combination of 24 different prompt styles to verify the robustness of our bench. The 4 initial prompts and 6 types of question formats are listed below.

\begin{promptbox}{Initial Prompt 1}
Answer the following multiple choice question. There is only one correct answer. The last line of your response should be in the format 'Answer: \$LETTER' (without quotes), where LETTER is one of A, B, C, D, E, F, G, H, I, or J.
\end{promptbox}

\begin{promptbox}{Initial Prompt 2}
You are a helpful assistant. Answer the given multiple-choice question. Only one option is correct. The last line of your response should be in the format 'The correct answer is: \$LETTER', where LETTER is one of A, B, C, D, E, F, G, H, I, or J.

\end{promptbox}

\begin{promptbox}{Initial Prompt 3}
Select the correct answer for the following multiple-choice question. There is only one valid choice. The last line of your response should be in the format 'Answer: \$LETTER' (without quotes), where LETTER is one of A, B, C, D, E, F, G, H, I, or J.

\end{promptbox}

\begin{promptbox}{Initial Prompt 4}
Review the following multiple-choice question and choose the one correct answer. Ensure that your response concludes with a line exactly formatted as 'The correct answer is: \$LETTER', where LETTER represents one of A, B, C, D, E, F, G, H, I, or J.
\end{promptbox}

\begin{promptbox}{Question Format Example 1}
The common-mode rejection ratio of the first stage amplification circuit in a three-op-amp differential circuit is determined by ( ).  
\\
A) the absolute value of the difference in the common-mode rejection ratio of A1 and A2 themselves  
\\
B) all of the above  
\\
C) the average of A1 and A2's common-mode rejection ratios  
\\
D) the sum of A1 and A2's common-mode rejection ratios  
\\
E) the product of A1 and A2's common-mode rejection ratios  
\\
F) the square root of the product of A1 and A2's common-mode rejection ratios  
\\
G) the size of A2's common-mode rejection ratio  
\\
H) the size of A1's common-mode rejection ratio  
\\
I) The difference in the common-mode rejection ratio of A1 and A2 themselves  
\\
J) input resistance  
\end{promptbox}

\begin{promptbox}{Question Format Example 2}
The common-mode rejection ratio of the first stage amplification circuit in a three-op-amp differential circuit is determined by ( ).  
\\
A. the absolute value of the difference in the common-mode rejection ratio of A1 and A2 themselves  
\\
B. all of the above  
\\
C. the average of A1 and A2's common-mode rejection ratios  
\\
D. the sum of A1 and A2's common-mode rejection ratios  
\\
E. the product of A1 and A2's common-mode rejection ratios  
\\
F. the square root of the product of A1 and A2's common-mode rejection ratios  
\\
G. the size of A2's common-mode rejection ratio  
\\
H. the size of A1's common-mode rejection ratio  
\\
I. The difference in the common-mode rejection ratio of A1 and A2 themselves  
\\
J. input resistance  
Your response: 

\end{promptbox}

\begin{promptbox}{Question Format Example 3}
Question:  
\\
The common-mode rejection ratio of the first stage amplification circuit in a three-op-amp differential circuit is determined by ( ).  
\\
Options:  
\\
A) the absolute value of the difference in the common-mode rejection ratio of A1 and A2 themselves  
\\
B) all of the above  
\\
C) the average of A1 and A2's common-mode rejection ratios  
\\
D) the sum of A1 and A2's common-mode rejection ratios  
\\
E) the product of A1 and A2's common-mode rejection ratios  
\\
F) the square root of the product of A1 and A2's common-mode 
rejection ratios  
\\
G) the size of A2's common-mode rejection ratio  
\\
H) the size of A1's common-mode rejection ratio  
\\
I) The difference in the common-mode rejection ratio of A1 and A2 themselves  
\\
J) input resistance  
Please begin answering.  

\end{promptbox}

\begin{promptbox}{Question Format Example 4}
Q: The common-mode rejection ratio of the first stage amplification circuit in a three-op-amp differential circuit is determined by ( ).  
(A) the absolute value of the difference in the common-mode rejection ratio of A1 and A2 themselves (B) all of the above (C) the average of A1 and A2's common-mode rejection ratios (D) the sum of A1 and A2's common-mode rejection ratios (E) the product of A1 and A2's common-mode rejection ratios (F) the square root of the product of A1 and A2's common-mode rejection ratios (G) the size of A2's common-mode rejection ratio (H) the size of A1's common-mode rejection ratio (I) The difference in the common-mode rejection ratio of A1 and A2 themselves (J) input resistance

\end{promptbox}

\begin{promptbox}{Question Format Example 5}
**Question**:  
The common-mode rejection ratio of the first stage amplification circuit in a three-op-amp differential circuit is determined by ( ).  
\\
**Options**:  
\\
(A) the absolute value of the difference in the common-mode rejection ratio of A1 and A2 themselves  
\\
(B) all of the above  
\\
(C) the average of A1 and A2's common-mode rejection ratios  
\\
(D) the sum of A1 and A2's common-mode rejection ratios  
\\
(E) the product of A1 and A2's common-mode rejection ratios  
\\
(F) the square root of the product of A1 and A2's common-mode rejection ratios  
\\
(G) the size of A2's common-mode rejection ratio  
\\
(H) the size of A1's common-mode rejection ratio  
\\
(I) The difference in the common-mode rejection ratio of A1 and A2 themselves  
\\
(J) input resistance  

\end{promptbox}

\begin{promptbox}{Question Format Example 6}

Question: The common-mode rejection ratio of the first stage amplification circuit in a three-op-amp differential circuit is determined by ( ).  
\\
Options:  
\\
A: the absolute value of the difference in the common-mode rejection ratio of A1 and A2 themselves  
\\
B: all of the above  
\\
C: the average of A1 and A2's common-mode rejection ratios  
\\
D: the sum of A1 and A2's common-mode rejection ratios  
\\
E: the product of A1 and A2's common-mode rejection ratios  
\\
F: the square root of the product of A1 and A2's common-mode rejection ratios  
\\
G: the size of A2's common-mode rejection ratio  
\\
H: the size of A1's common-mode rejection ratio  
\\
I: The difference in the common-mode rejection ratio of A1 and A2 themselves  
\\
J: input resistance  

\end{promptbox}

\begin{CJK}{UTF8}{gbsn}


\clearpage
\hypertarget{benchmark}{}
\section{Benchmarks for Data Expansion }
\label{Sec: Benchmark List}
\begin{longtable}{@{}p{4cm}p{9cm}p{2cm}p{1cm}@{}}

\caption{Benchmark List for Data Supplementation in SuperGPQA Annotation.}
\label{tab:benchmark}\\

\toprule
\textbf{Benchmark Name} & \textbf{Title} & \textbf{Year} \\ 
\midrule
\endfirsthead

LawBench & LawBench: Benchmarking Legal Knowledge of Large Language Models~\cite{fei2023lawbench} & 2023 \\ \midrule
MedMCQA & MedMCQA: A Large-scale Multi-Subject Multi-Choice Dataset for Medical domain Question Answering~\cite{pmlr-v174-pal22a} & 2022 \\ \midrule
MedQA & What Disease does this Patient Have? A Large-scale Open Domain Question Answering Dataset from Medical Exams~\cite{jin2020disease} & 2020 \\ \midrule
MMLU-Pro & MMLU-Pro: A More Robust and Challenging Multi-Task Language Understanding Benchmark~\cite{wang2024mmlupro} & 2024 \\ \midrule
MMLU-CF & MMLU-CF: A Contamination-free Multi-task Language Understanding Benchmark~\cite{zhao2024mmlucfcontaminationfreemultitasklanguage} & 2024 \\ \midrule
ShoppingMMLU & Shopping MMLU: A Massive Multi-Task Online Shopping Benchmark for Large Language Models~\cite{jin2024shopping} & 2024 \\ \midrule
UTMath & UTMath: Math Evaluation with Unit Test via Reasoning-to-Coding Thoughts~\cite{yang2024utmath} & 2024 \\ \midrule
MusicTheoryBench & ChatMusician: Understanding and Generating Music Intrinsically with LLM~\cite{yuan2024chatmusician} & 2024 \\ \midrule
Omni-Math & Omni-MATH: A Universal Olympiad Level Mathematic Benchmark For Large Language Models~\cite{gao2024omnimathuniversalolympiadlevel} & 2024 \\ \midrule
U-MATH & U-MATH: A University-Level Benchmark for Evaluating Mathematical Skills in LLMs~\cite{chernyshev2024umath} & 2024 \\ \midrule
\newpage
\midrule
Putnam-AXIOM & Putnam-AXIOM: A Functional and Static Benchmark for Measuring Higher Level Mathematical Reasoning~\cite{fronsdal2024putnamaxiom} & 2024 \\
\midrule

Short-form Factuality & Measuring short-form factuality in large language models~\cite{wei2024measuringshortformfactualitylarge} & 2024 \\ \midrule
Chinese SimpleQA & Chinese SimpleQA: A Chinese Factuality Evaluation for Large Language Models~\cite{he2024chinesesimpleqachinesefactuality} & 2024 \\ \midrule
AIME-AOPS & AIME Problems and Solutions~\cite{aopsAIME} & 2024 \\ \midrule
AIMO Validation AIME & AIMO Validation AIME: Internal Validation Set for AIMO Progress Prize~\cite{aimoValidationAIME} & 2024 \\ 

\bottomrule
\end{longtable} 
\centeredlinks{benchmark}{Back to Section Start}{toc}{Back to Table of Contents}{blue}


\end{CJK}

\section{More Comprehensive Analysis of Baseline Performances}
\label{appendix: all results}
In addition to the models discussed in the \autoref{sec:results},  
 \autoref{performance1_all} and \autoref{performance2_all} provide a more comprehensive analysis of baseline performances, including evaluations of models from the Yi-1.5 (6B, 9B), OLMo-2 (7B), Mistral (Small), granite-3.1 (2B), Gemma-2 (2B, 9B), Llama-3.1 (8B), Mixtral (8x7B), and K2~\citep{liu2025llm360k2building65b}.
\definecolor{color11}{rgb}{1, 0.8, 0.8}  
\definecolor{color12}{rgb}{1, 0.9, 0.9}  
\definecolor{color21}{RGB}{255, 224, 127}  
\definecolor{color22}{RGB}{255, 239, 179}  
\definecolor{color31}{RGB}{198, 230, 195}  
\definecolor{color32}{RGB}{224, 239, 225}  

{
\linespread{1}
\begin{table}[H]
\scriptsize
\centering
\resizebox{\textwidth}{!}{%
\begin{tabular}{p{4cm}<{\raggedright\arraybackslash}*{7}{p{2cm}<{\centering\arraybackslash}}}
\toprule
\textbf{Model} & \textbf{Overall} & \textbf{Overall} & \textbf{Overall} & \textbf{Overall} & \textbf{Easy} & \textbf{Middle} & \textbf{Hard}\\
& \textbf{(sample)} & \textbf{(subfield)} & \textbf{(field)} & \textbf{(discipline)} & \textbf{(sample)} & \textbf{(sample)} & \textbf{(sample)}\\
\midrule
\rowcolor{color11}
\multicolumn{8}{c}{\textbf{\textit{Reasoning Models}}}\\
\midrule
\rowcolor{color12}
DeepSeek-R1 &\boxed{61.82} & \boxed{62.61} & \boxed{61.23} & \textbf{59.95} & \underline{63.59} & \boxed{63.63} &\boxed{56.87} \\
\rowcolor{color12}
o1-2024-12-17 &\textbf{60.24} & \underline{61.25} & \underline{59.94} & \underline{59.44} & \textbf{64.40} & \underline{61.44} &\underline{53.67} \\
\rowcolor{color12}
DeepSeek-R1-Zero &\textbf{60.24} & \textbf{61.62} & \textbf{60.95} & \boxed{60.99} & \boxed{65.06} & \textbf{62.61} &50.99 \\
\rowcolor{color12}
o3-mini-2025-01-31-high &\underline{55.22} & 54.94 & 52.11 & 48.32 & 53.05 & 56.09 &\textbf{56.16} \\
\rowcolor{color12}
o3-mini-2025-01-31-medium &52.69 & 52.66 & 49.95 & 46.07 & 51.30 & 53.79 &52.37 \\
\rowcolor{color12}
o3-mini-2025-01-31-low &48.03 & 48.51 & 45.89 & 42.63 & 48.80 & 50.21 &43.53 \\
\rowcolor{color12}
o1-mini-2024-09-12 &45.22 & 45.46 & 42.53 & 39.33 & 46.77 & 47.34 &40.00 \\
\rowcolor{color12}
QwQ &43.59 & 44.40 & 43.19 & 41.63 & 46.46 & 47.40 &34.07 \\
\midrule
\rowcolor{color21}
\multicolumn{8}{c}{\textbf{\textit{Chat Models}}}\\
\midrule
\rowcolor{color22}
Doubao-1.5-pro-32k-250115 &\boxed{55.09} & \boxed{56.55} & \boxed{55.62} & \boxed{54.39} & \underline{57.70} & \boxed{60.15} &\boxed{43.80} \\
\rowcolor{color22}
Doubao-1.5-pro-32k-241225 &\textbf{50.93} & \underline{52.41} & \underline{51.76} & 51.24 & 53.54 & \textbf{56.56} &\underline{38.70} \\
\rowcolor{color22}
qwen-max-2025-01-25 &\underline{50.08} & \textbf{52.75} & \textbf{52.47} & \underline{51.65} & \textbf{58.16} & \underline{54.95} &33.09 \\
\rowcolor{color22}
claude-3-5-sonnet-20241022 &48.16 & 51.38 & 51.23 & \textbf{53.15} & \boxed{59.04} & 51.91 &29.99 \\
\rowcolor{color22}
gemini-2.0-flash &47.73 & 48.70 & 47.80 & 46.10 & 53.06 & 49.56 &\textbf{38.84} \\
\rowcolor{color22}
DeepSeek-V3 &47.40 & 49.10 & 48.31 & 47.35 & 55.63 & 50.11 &33.86 \\
\rowcolor{color22}
MiniMax-Text-01 &45.11 & 47.46 & 46.97 & 47.06 & 54.51 & 48.60 &28.98 \\
\rowcolor{color22}
gpt-4o-2024-11-20 &44.40 & 47.62 & 47.50 & 48.84 & 56.84 & 48.75 &23.50 \\
\rowcolor{color22}
Llama-3.1-405B-Instruct &43.14 & 46.43 & 45.83 & 47.35 & 56.06 & 46.31 &23.70 \\
\rowcolor{color22}
gpt-4o-2024-08-06 &41.64 & 44.79 & 44.91 & 46.29 & 55.22 & 45.11 &20.98 \\
\rowcolor{color22}
Qwen2.5-72B-Instruct &40.75 & 43.66 & 43.32 & 42.10 & 48.84 & 45.42 &24.10 \\
\rowcolor{color22}
Mistral-Large-Instruct-2411 &40.65 & 43.38 & 43.13 & 43.37 & 52.92 & 43.28 &22.81 \\
\rowcolor{color22}
qwen-max-2024-09-19 &39.96 & 42.93 & 42.16 & 41.62 & 50.23 & 43.63 &22.60 \\
\rowcolor{color22}
gpt-4o-2024-05-13 &39.76 & 43.19 & 43.13 & 45.23 & 53.37 & 42.38 &20.45 \\
\rowcolor{color22}
Qwen2.5-32B-Instruct &38.76 & 41.18 & 40.40 & 39.43 & 47.42 & 43.05 &22.13 \\
\rowcolor{color22}
Llama-3.3-70B-Instruct &37.69 & 40.56 & 40.15 & 41.12 & 49.68 & 40.68 &19.55 \\
\rowcolor{color22}
phi-4 &37.65 & 39.59 & 38.61 & 37.66 & 45.43 & 40.91 &23.69 \\
\rowcolor{color22}
Qwen2.5-14B-Instruct &35.15 & 37.72 & 37.41 & 36.07 & 44.82 & 37.90 &19.97 \\
\rowcolor{color22}
Llama-3.1-70B-Instruct &34.86 & 38.94 & 39.18 & 40.57 & 48.22 & 37.85 &15.22 \\
\rowcolor{color22}
Yi-Lightning &33.42 & 36.57 & 36.45 & 36.92 & 43.38 & 35.32 &19.35 \\
\rowcolor{color22}
Mixtral-8x22B-Instruct-v0.1 &29.23 & 32.14 & 32.28 & 32.82 & 42.52 & 29.73 &13.82 \\
\rowcolor{color22}
Qwen2.5-7B-Instruct &28.78 & 30.78 & 30.37 & 30.63 & 37.77 & 30.98 &15.23 \\
\rowcolor{color22}
gemma-2-27b-it &27.43 & 30.50 & 30.42 & 31.30 & 40.90 & 27.45 &12.64 \\
\rowcolor{color22}
Yi-1.5-34B-Chat &26.03 & 28.81 & 28.84 & 29.08 & 36.99 & 26.74 &12.81 \\
\rowcolor{color22}
Mistral-Small-Instruct-2409 &25.89 & 28.46 & 28.69 & 28.59 & 37.93 & 25.89 &12.70 \\
\rowcolor{color22}
gemma-2-9b-it &24.04 & 26.89 & 27.06 & 27.74 & 37.81 & 23.05 &10.60 \\
\rowcolor{color22}
Qwen2.5-3B-Instruct &23.31 & 25.45 & 25.86 & 25.57 & 33.10 & 23.50 &12.24 \\
\rowcolor{color22}
Yi-1.5-9B-Chat &23.17 & 25.32 & 25.65 & 26.07 & 32.75 & 24.05 &11.19 \\
\rowcolor{color22}
K2-Chat &22.47 & 24.59 & 24.61 & 24.61 & 30.58 & 21.95 &14.44 \\
\rowcolor{color22}
Mixtral-8x7B-Instruct-v0.1 &22.10 & 24.76 & 24.73 & 26.36 & 34.19 & 20.52 &11.49 \\
\rowcolor{color22}
granite-3.1-8b-instruct &20.83 & 22.85 & 22.92 & 22.26 & 29.48 & 19.79 &13.09 \\
\rowcolor{color22}
Llama-3.1-8B-Instruct &20.50 & 24.07 & 24.52 & 26.12 & 32.82 & 20.37 &7.22 \\
\rowcolor{color22}
Yi-1.5-6B-Chat &19.24 & 21.32 & 21.39 & 22.21 & 27.90 & 18.83 &10.44 \\
\rowcolor{color22}
Qwen2.5-1.5B-Instruct &18.82 & 20.91 & 20.75 & 22.11 & 27.41 & 18.19 &10.45 \\
\rowcolor{color22}
OLMo-2-1124-13B-Instruct &18.66 & 20.46 & 20.60 & 21.80 & 27.10 & 17.85 &10.74 \\
\rowcolor{color22}
gemma-2-2b-it &18.61 & 19.91 & 19.97 & 20.50 & 26.40 & 16.95 &12.85 \\
\rowcolor{color22}
granite-3.1-2b-instruct &17.92 & 19.02 & 19.11 & 19.58 & 23.94 & 17.58 &11.87 \\
\rowcolor{color22}
Mistral-7B-Instruct-v0.3 &17.82 & 19.64 & 19.65 & 20.09 & 26.37 & 16.64 &10.41 \\
\rowcolor{color22}
MAP-Neo-7B-Instruct-v0.1 &17.05 & 18.52 & 18.42 & 18.70 & 23.26 & 16.62 &10.95 \\
\rowcolor{color22}
OLMo-2-1124-7B-Instruct &16.81 & 18.08 & 18.57 & 18.85 & 22.80 & 15.82 &11.90 \\
\rowcolor{color22}
Qwen2.5-0.5B-Instruct &10.77 & 11.92 & 12.47 & 13.47 & 14.90 & 10.88 &6.07 \\
\midrule
\rowcolor{color31}
\multicolumn{8}{c}{\textbf{\textit{Base Models}}}\\
\midrule
\rowcolor{color32}
Qwen2.5-72B & \boxed{34.33} & \boxed{38.08} & \boxed{38.70} & \boxed{39.54} & \boxed{46.20} & \boxed{38.12} &\textbf{15.01} \\
\rowcolor{color32}
Qwen2.5-32B & \textbf{33.16} & \textbf{36.52} & \textbf{37.33} & \textbf{38.29} & \textbf{45.12} & \textbf{36.58} &14.34 \\
\rowcolor{color32}
DeepSeek-V3-Base & \underline{32.14} & \underline{34.79} & \underline{34.58} & \underline{34.71} & 41.28 & \underline{34.50} &\boxed{18.20} \\
\rowcolor{color32}
Qwen2.5-14B & 30.19 & 33.33 & 34.14 & 34.54 & \underline{42.27} & 31.44 &\underline{14.85} \\
\rowcolor{color32}
Yi-1.5-34B & 27.62 & 30.78 & 31.03 & 32.55 & 39.68 & 27.95 &13.86 \\
\rowcolor{color32}
Llama-3.1-70B & 27.22 & 30.52 & 31.28 & 32.55 & 40.78 & 26.95 &12.78 \\
\rowcolor{color32}
Qwen2.5-7B & 25.36 & 28.19 & 28.73 & 29.60 & 36.58 & 25.94 &12.10 \\
\rowcolor{color32}
Llama-3.1-405B & 25.23 & 28.09 & 28.33 & 30.15 & 37.58 & 25.12 &11.86 \\
\rowcolor{color32}
gemma-2-27b & 24.49 & 27.35 & 27.96 & 28.58 & 36.26 & 24.07 &12.27 \\
\rowcolor{color32}
Yi-1.5-9B & 23.10 & 25.40 & 25.69 & 26.17 & 32.52 & 22.98 &12.96 \\
\rowcolor{color32}
gemma-2-9b & 22.56 & 25.19 & 25.47 & 26.26 & 33.88 & 21.34 &12.20 \\
\rowcolor{color32}
Mixtral-8x22B-v0.1 & 22.41 & 24.71 & 25.04 & 25.02 & 32.78 & 21.67 &12.26 \\
\rowcolor{color32}
Mixtral-8x7B-v0.1 & 21.76 & 24.92 & 25.45 & 27.36 & 33.66 & 20.32 &11.13 \\
\rowcolor{color32}
K2 & 20.92 & 23.62 & 23.88 & 24.65 & 30.97 & 20.01 &11.40 \\
\rowcolor{color32}
Yi-1.5-6B & 20.20 & 22.10 & 22.56 & 23.44 & 28.37 & 19.64 &12.20 \\
\rowcolor{color32}
Qwen2.5-3B & 20.14 & 22.81 & 23.30 & 24.42 & 30.42 & 19.81 &9.40 \\
\rowcolor{color32}
Llama-3.1-8B & 19.93 & 22.42 & 22.77 & 23.87 & 30.33 & 18.94 &10.16 \\
\rowcolor{color32}
Mistral-7B-v0.3 & 19.48 & 21.50 & 21.81 & 22.27 & 27.62 & 18.65 &11.96 \\
\rowcolor{color32}
Qwen2.5-1.5B & 17.17 & 19.31 & 19.80 & 21.35 & 24.52 & 16.79 &9.74 \\
\rowcolor{color32}
granite-3.1-2b-base & 16.18 & 17.90 & 17.91 & 18.09 & 23.13 & 14.91 &10.67 \\
\rowcolor{color32}
OLMo-2-1124-13B & 16.07 & 18.75 & 19.82 & 21.37 & 27.24 & 14.41 &6.57 \\
\rowcolor{color32}
MAP-Neo-7B & 15.76 & 17.48 & 18.26 & 19.54 & 22.86 & 14.64 &9.83 \\
\rowcolor{color32}
granite-3.1-8b-base & 15.69 & 16.98 & 16.79 & 16.65 & 20.40 & 15.65 &10.60 \\
\rowcolor{color32}
OLMo-2-1124-7B & 15.15 & 17.62 & 18.30 & 19.60 & 24.43 & 13.83 &7.15 \\
\rowcolor{color32}
gemma-2-2b & 13.57 & 14.98 & 15.43 & 16.21 & 19.12 & 13.47 &7.63 \\
\rowcolor{color32}
Qwen2.5-0.5B & 10.74 & 11.88 & 12.09 & 13.25 & 14.41 & 10.78 &6.65 \\

\bottomrule
\end{tabular}
}
\captionsetup{font=footnotesize}
\caption{\textbf{Detailed Performance Overview on \benchmark - Pivot Table 1.} 
LLMs are scored sample-wise, subfield-wise, field-wise, and discipline-wise levels to ensure fair assessment despite imbalanced question counts. 
The columns Easy(sample), Middle(sample), and Hard(sample) represent average scores according to difficulty. 
The highest score in each column is indicated with a \boxed{box}; the second-best score is in \textbf{bold}, and the third-best score is \underline{underlined}.}
\label{performance1_all}
\end{table}
}

\definecolor{color11}{rgb}{1, 0.8, 0.8}  
\definecolor{color12}{rgb}{1, 0.9, 0.9}  
\definecolor{color21}{RGB}{255, 224, 127}  
\definecolor{color22}{RGB}{255, 239, 179}  
\definecolor{color31}{RGB}{198, 230, 195}  
\definecolor{color32}{RGB}{224, 239, 225}  

{
\linespread{1}
\begin{table}[H]
\footnotesize
\centering
\resizebox{\textwidth}{!}{%
\begin{tabular}{lccccccccccccc}
\toprule
\textbf{Model} & \textbf{Agr.} & \textbf{Econ.} & \textbf{Edu.} & \textbf{Eng.} & \textbf{Hist.} & \textbf{Law} & \textbf{Lit \& Arts} & \textbf{Mgt.} & \textbf{Med.} & \textbf{Mil Sci.} & \textbf{Phil.} & \textbf{Sci.} & \textbf{Soc.} \\
\midrule
\rowcolor{color11}
\multicolumn{14}{c}{\textbf{\textit{Reasoning Models}}}\\
\midrule
\rowcolor{color12}
DeepSeek-R1& \boxed{54.43}& \textbf{66.09}& \textbf{54.75}& \boxed{63.10}& \underline{55.19}& \textbf{65.24}& \underline{52.45}& \underline{57.09}& \underline{59.93}& \underline{57.07}& \textbf{63.11}& \boxed{63.69}& \boxed{67.13} \\
\rowcolor{color12}
DeepSeek-R1-Zero& \textbf{53.81}& \boxed{66.44}& \boxed{60.54}& \textbf{60.28}& \textbf{58.61}& \boxed{66.77}& \boxed{56.86}& \boxed{59.68}& \textbf{60.65}& \textbf{58.54}& \boxed{63.69}& 59.93& \boxed{67.13} \\
\rowcolor{color12}
o1-2024-12-17& \underline{50.93}& \underline{61.17}& \underline{53.31}& \underline{59.17}& \boxed{60.98}& \underline{63.87}& \textbf{55.79}& \textbf{57.29}& \boxed{62.25}& \boxed{60.49}& \underline{61.38}& \textbf{61.76}& \textbf{64.34} \\
\rowcolor{color12}
o3-mini-2025-01-31-high& 41.03& 54.07& 46.28& 56.41& 36.80& 45.73& 39.44& 51.50& 51.72& 46.34& 51.01& \underline{61.72}& \underline{46.15} \\
\rowcolor{color12}
o3-mini-2025-01-31-medium& 40.62& 51.32& 41.74& 53.83& 35.01& 43.60& 37.11& 48.50& 50.34& 45.85& 48.13& 58.79& 44.06 \\
\rowcolor{color12}
o3-mini-2025-01-31-low& 37.32& 45.25& 38.43& 48.20& 32.94& 39.79& 36.22& 43.71& 48.09& 41.46& 44.09& 53.24& 45.45 \\
\bottomrule
\end{tabular}
}
\captionsetup{font=footnotesize}
\caption{\textbf{Detailed Performance Overview on \benchmark - Pivot Table 2.} 
The table presents LLMs' performance on different disciplines. The highest score in each column is indicated with a \boxed{box}; the second-best score is in \textbf{bold}, and the third-best score is \underline{underlined}.}
\label{performance2_all}
\end{table}
}

\clearpage
\hypertarget{top5}{}
\section{Ranking of the Top Five Models in Each of the 285 Subfields}
{
\setlength{\tabcolsep}{2pt}  
\footnotesize
\begin{longtable}{p{4.2cm}>{\centering\arraybackslash} p{0.8cm}|p{4.2cm} >{\centering\arraybackslash} p{0.8cm}|p{4.2cm} >{\centering\arraybackslash} p{0.8cm}}
\caption{Detailed Listing of the Top Five Models in Each of the 285 Subfields.}
\label{table:label} \\ 
\hline
\textbf{Model} & \textbf{Score} & \textbf{Model} & \textbf{Score} & \textbf{Model} & \textbf{Score}\\
\hline
\endfirsthead

\multicolumn{6}{c}{Continued}\\
\endhead
\multicolumn{2}{p{5.15cm}|}{\cellcolor{teal!10} \centering Animal Nutrition and Feed Science} & \multicolumn{2}{p{5.15cm}|}{\cellcolor{blue!10} \centering Applied Optics} & \multicolumn{2}{p{5.15cm}}{\cellcolor{yellow!10} \centering Pharmaceutics}\\
\hline
\cellcolor{teal!5} DeepSeek-R1 & \cellcolor{teal!2}50.00 & \cellcolor{blue!5} DeepSeek-R1 & \cellcolor{blue!2} 58.06 & \cellcolor{yellow!5} DeepSeek-R1-Zero & \cellcolor{yellow!2} 75.34\\
\cellcolor{teal!5} DeepSeek-R1-Zero & \cellcolor{teal!2}50.00 & \cellcolor{blue!5} o3-mini-2025-01-31-high & \cellcolor{blue!2} 57.26 & \cellcolor{yellow!5} qwen-max-2025-01-25 & \cellcolor{yellow!2} 69.86\\
\cellcolor{teal!5} claude-3-5-sonnet-20241022 & \cellcolor{teal!2}48.08 & \cellcolor{blue!5} o1-2024-12-17 & \cellcolor{blue!2} 56.45 & \cellcolor{yellow!5} Doubao-1.5-pro-32k-250115 & \cellcolor{yellow!2} 68.49\\
\cellcolor{teal!5} o3-mini-2025-01-31-medium & \cellcolor{teal!2}48.08 & \cellcolor{blue!5} o3-mini-2025-01-31-medium & \cellcolor{blue!2} 53.23 & \cellcolor{yellow!5} Doubao-1.5-pro-32k-241225 & \cellcolor{yellow!2} 67.12\\
\cellcolor{teal!5} Qwen2.5-72B-Instruct & \cellcolor{teal!2}48.08 & \cellcolor{blue!5} o3-mini-2025-01-31-low & \cellcolor{blue!2} 51.61 & \cellcolor{yellow!5} DeepSeek-R1 & \cellcolor{yellow!2} 64.38\\
\hline
\multicolumn{2}{p{5.15cm}|}{\cellcolor{teal!10} \centering Animal Rearing and Breeding} & \multicolumn{2}{p{5.15cm}|}{\cellcolor{blue!10} \centering Laser Technology} & \multicolumn{2}{p{5.15cm}}{\cellcolor{yellow!10} \centering Pharmacology}\\
\hline
\cellcolor{teal!5} Doubao-1.5-pro-32k-250115 & \cellcolor{teal!2}49.02 & \cellcolor{blue!5} DeepSeek-R1 & \cellcolor{blue!2} 82.00 & \cellcolor{yellow!5} DeepSeek-R1-Zero & \cellcolor{yellow!2} 56.36\\
\hline

\end{longtable}
}
\centeredlinks{top5}{Back to Section Start}{toc}{Back to Table of Contents}{blue}
\newpage
\section{Detailed Scores of Each Discipline for All Evaluated Models}
\label{appendix: model score}
\hypertarget{listofmodels}{}
\listofmodels
\hyperlink{toc}{Back to Table of Contents}

\newpage
\vspace{-0.5cm}
\begin{table}[t]
\refstepcounter{models}%
\addcontentsline{csf}{models}{\protect\numberline{\themodels}DeepSeek-R1}
\centering
\begin{subtable}[t]{1\textwidth}
\centering
\includegraphics[width=\textwidth]{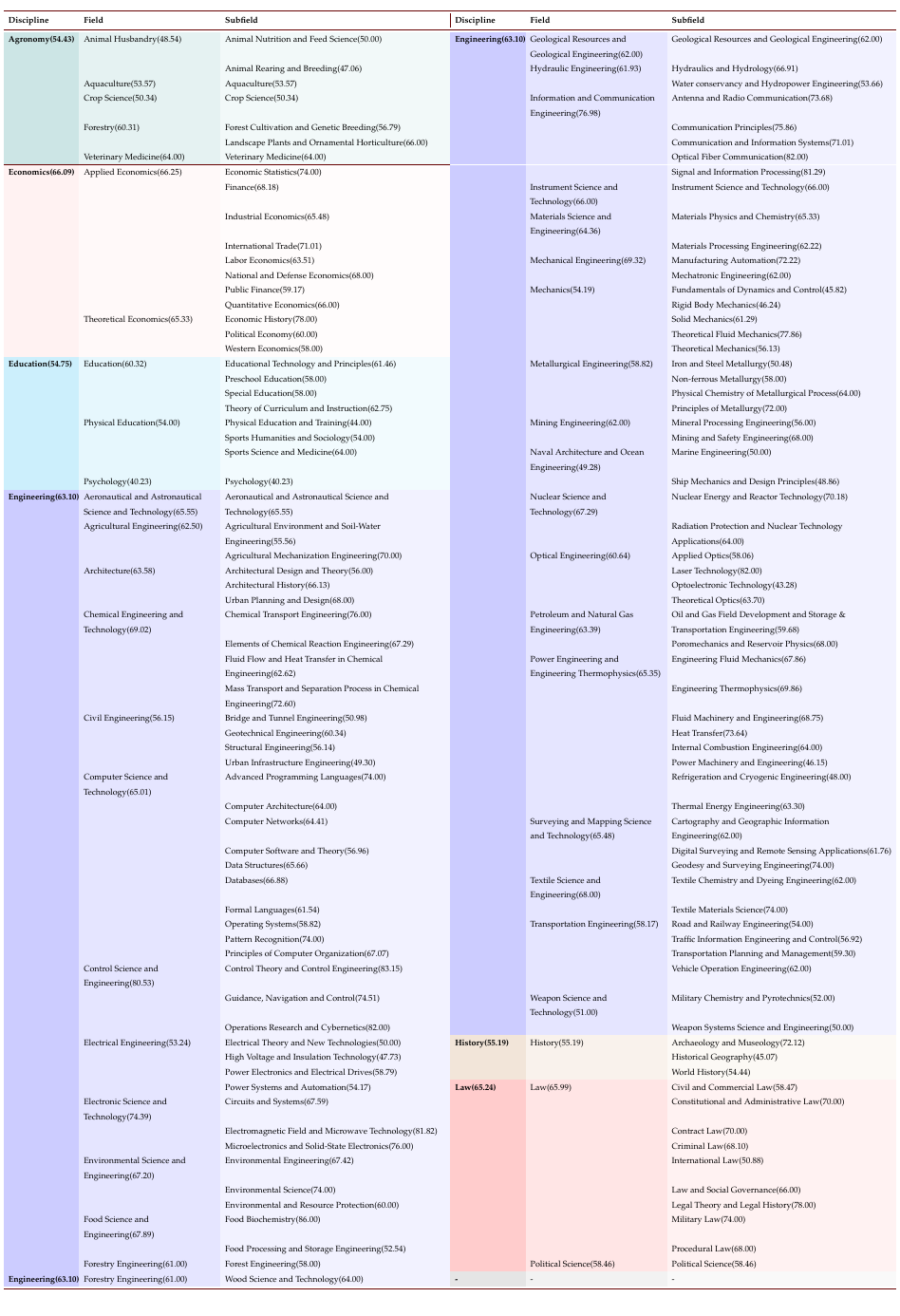} 
\end{subtable}
\vspace{-1.3cm}
\captionsetup{font=small}
\caption{Model Scores Across Three Levels of Disciplines: DeepSeek-R1.}
\label{tab:DeepSeek-R1}
\vspace{-0.5cm}
\centeredlinks{listofmodels}{Back to List of Models}{toc}{Back to Table of Contents}{blue}
\end{table}
\clearpage

\newpage
\afterpage{
    \begin{table}[t]
    \centering
    \ContinuedFloat 
    \begin{subtable}[t]{\textwidth}
    \centering
    \includegraphics[width=\textwidth]{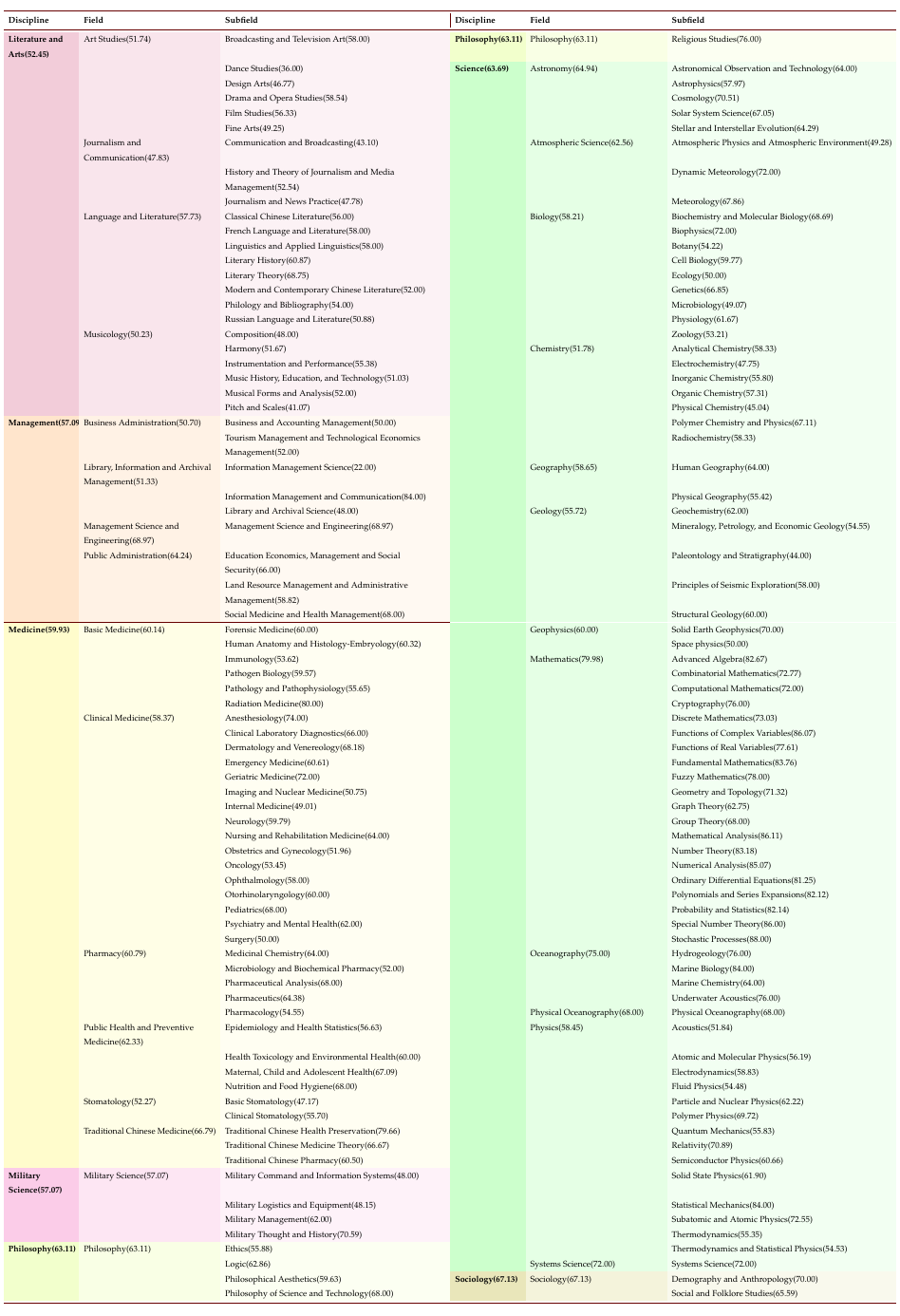} 
    \end{subtable}
    \vspace{-1.1cm}
    \captionsetup{font=small}
    \caption{Continued: Model Scores Across Three Levels of Disciplines: DeepSeek-R1.}
    \vspace{-0.6cm}
    \centeredlinks{listofmodels}{Back to List of Models}{toc}{Back to Table of Contents}{blue}
    \end{table}
}
\clearpage

\newpage
\vspace{-0.5cm}
\begin{table}[t]
\refstepcounter{models}%
\addcontentsline{csf}{models}{\protect\numberline{\themodels}DeepSeek-R1-Zero}
\centering
\begin{subtable}[t]{1\textwidth}
\centering
\includegraphics[width=\textwidth]{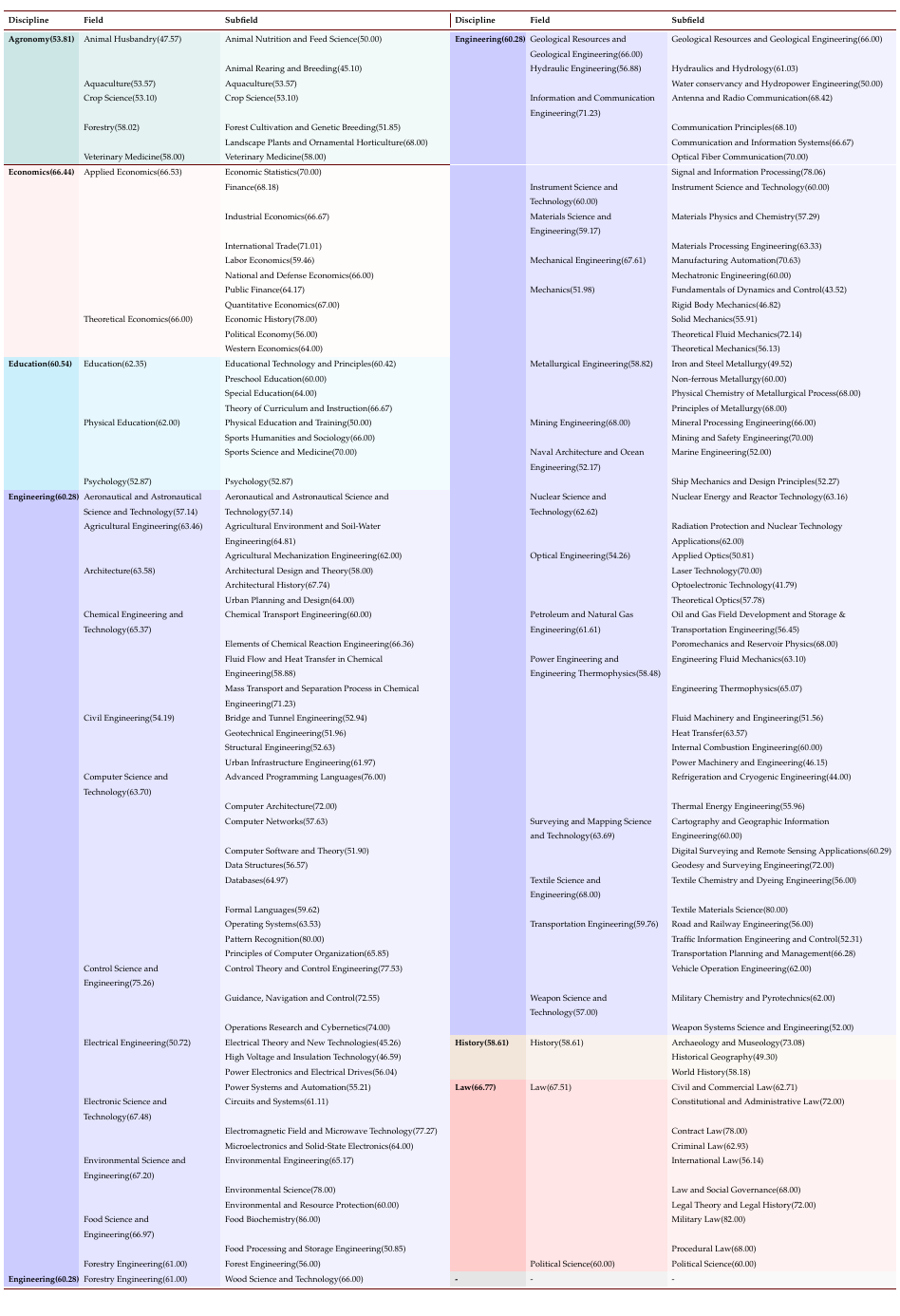} 
\end{subtable}
\vspace{-1.3cm}
\captionsetup{font=small}
\caption{Model Scores Across Three Levels of Disciplines: DeepSeek-R1-Zero.}
\label{tab:DeepSeek-R1-Zero}
\vspace{-0.5cm}
\centeredlinks{listofmodels}{Back to List of Models}{toc}{Back to Table of Contents}{blue}
\end{table}
\clearpage

\newpage
\afterpage{
    \begin{table}[t]
    \centering
    \ContinuedFloat 
    \begin{subtable}[t]{\textwidth}
    \centering
    \includegraphics[width=\textwidth]{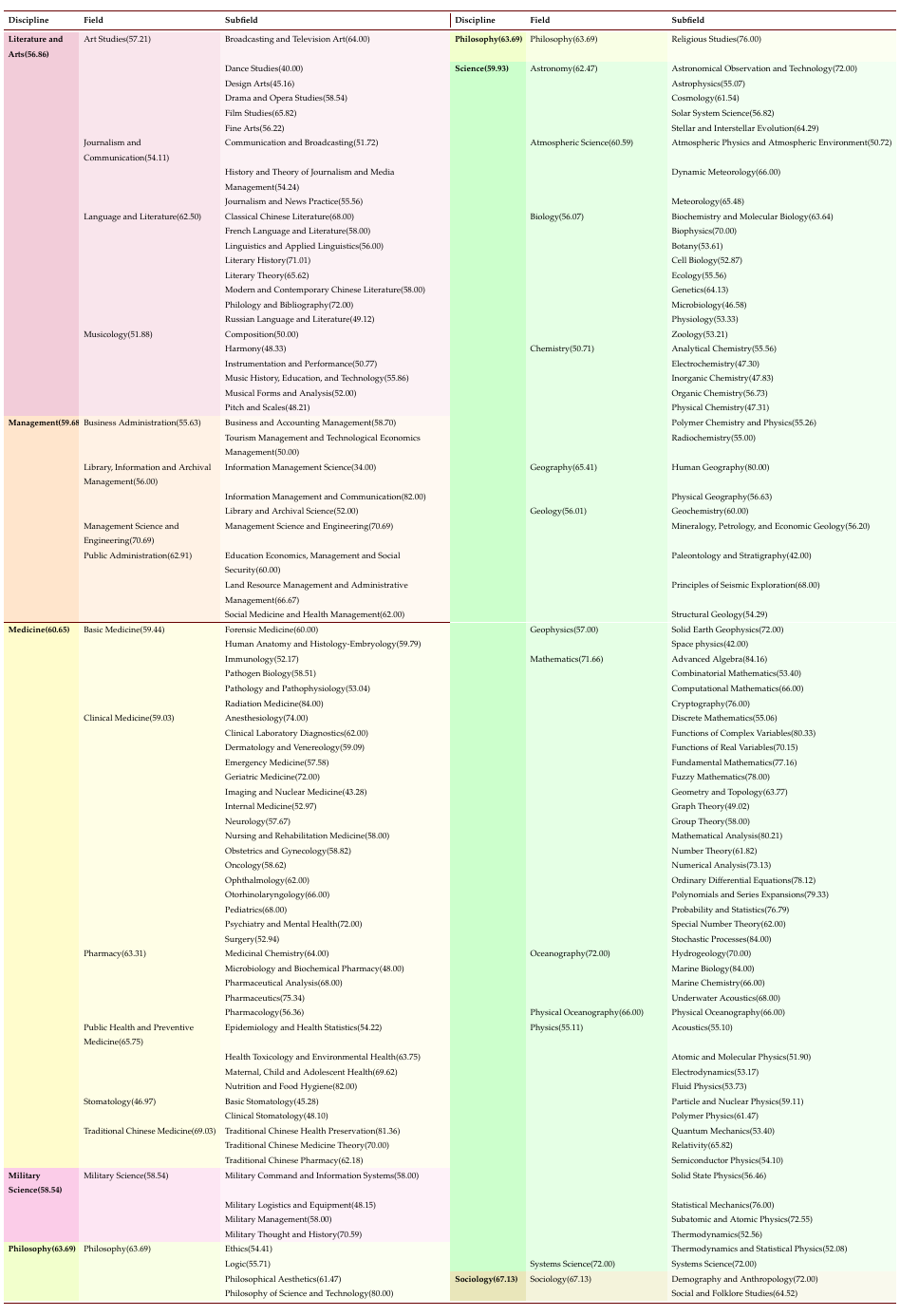} 
    \end{subtable}
    \vspace{-1.1cm}
    \captionsetup{font=small}
    \caption{Continued: Model Scores Across Three Levels of Disciplines: DeepSeek-R1-Zero.}
    \vspace{-0.6cm}
    \centeredlinks{listofmodels}{Back to List of Models}{toc}{Back to Table of Contents}{blue}
    \end{table}
}
\clearpage

\newpage
\vspace{-0.5cm}
\begin{table}[t]
\refstepcounter{models}%
\addcontentsline{csf}{models}{\protect\numberline{\themodels}o1-2024-12-17}
\centering
\begin{subtable}[t]{1\textwidth}
\centering
\includegraphics[width=\textwidth]{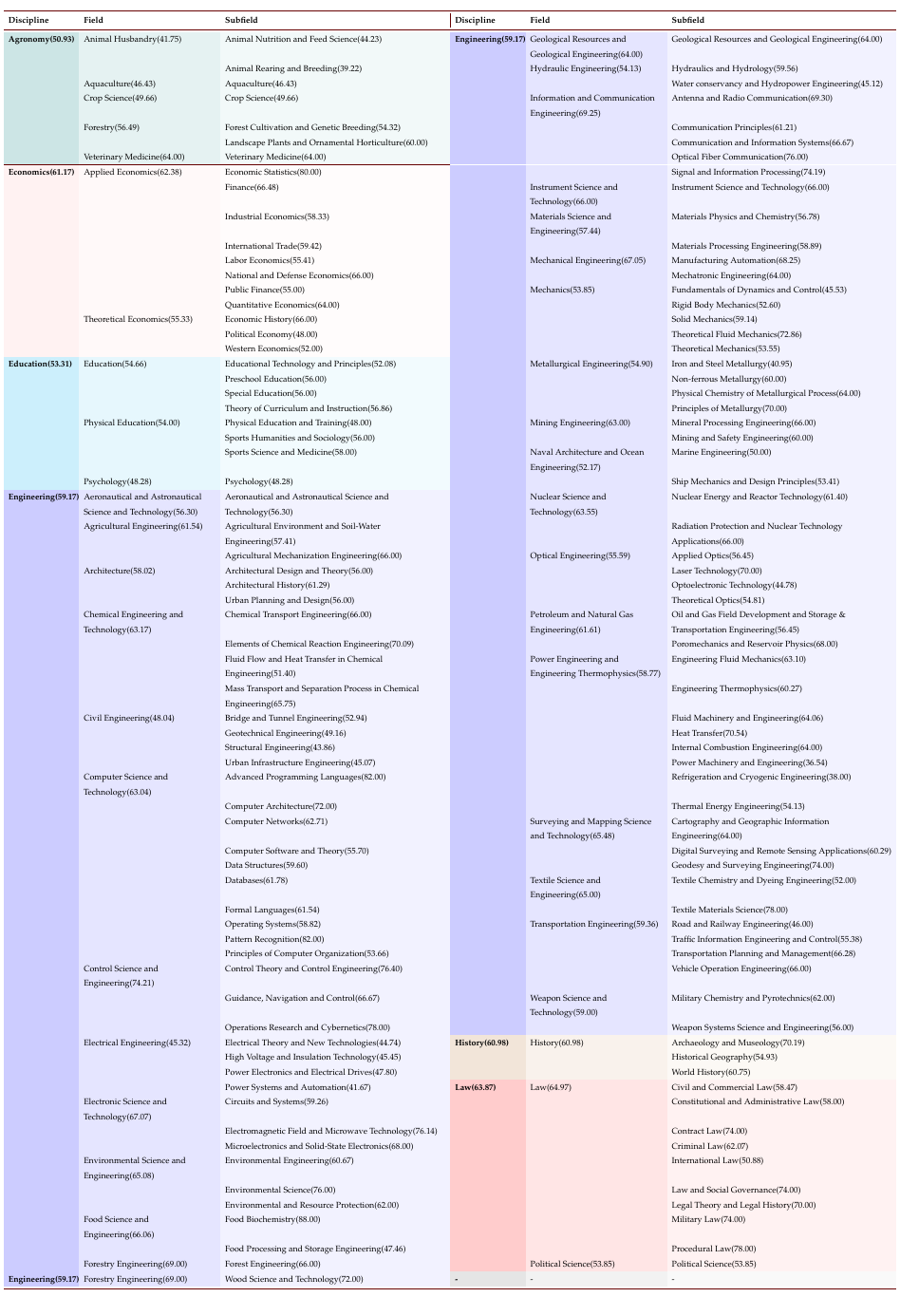} 
\end{subtable}
\vspace{-1.3cm}
\captionsetup{font=small}
\caption{Model Scores Across Three Levels of Disciplines: o1-2024-12-17.}
\label{tab:o1-2024-12-17}
\vspace{-0.5cm}
\centeredlinks{listofmodels}{Back to List of Models}{toc}{Back to Table of Contents}{blue}
\end{table}
\clearpage

\newpage
\afterpage{
    \begin{table}[t]
    \centering
    \ContinuedFloat 
    \begin{subtable}[t]{\textwidth}
    \centering
    \includegraphics[width=\textwidth]{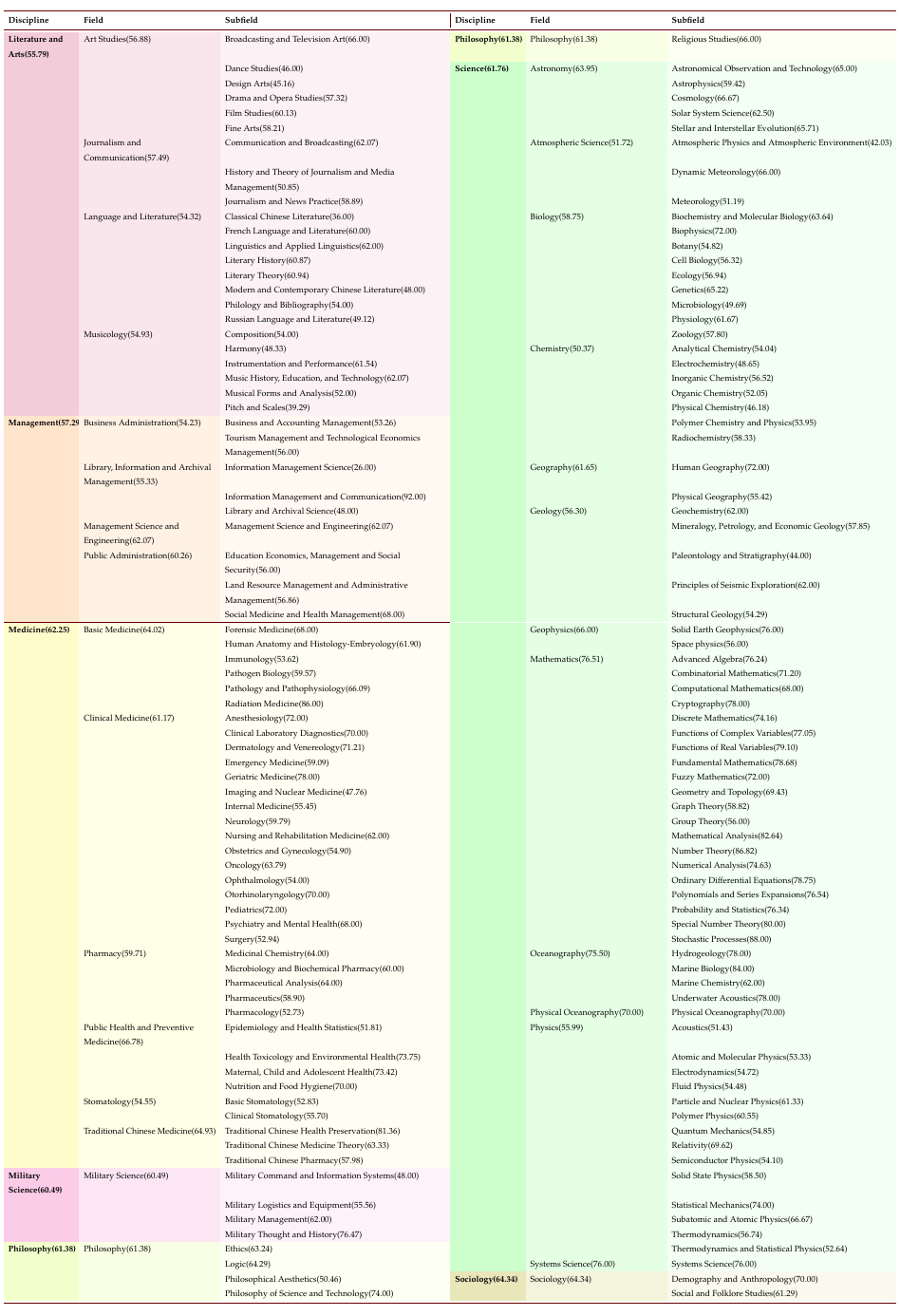} 
    \end{subtable}
    \vspace{-1.1cm}
    \captionsetup{font=small}
    \caption{Continued: Model Scores Across Three Levels of Disciplines: o1-2024-12-17.}
    \vspace{-0.6cm}
    \centeredlinks{listofmodels}{Back to List of Models}{toc}{Back to Table of Contents}{blue}
    \end{table}
}
\clearpage

\newpage
\vspace{-0.5cm}
\begin{table}[t]
\refstepcounter{models}%
\addcontentsline{csf}{models}{\protect\numberline{\themodels}o3-mini-2025-01-31-high}
\centering
\begin{subtable}[t]{1\textwidth}
\centering
\includegraphics[width=\textwidth]{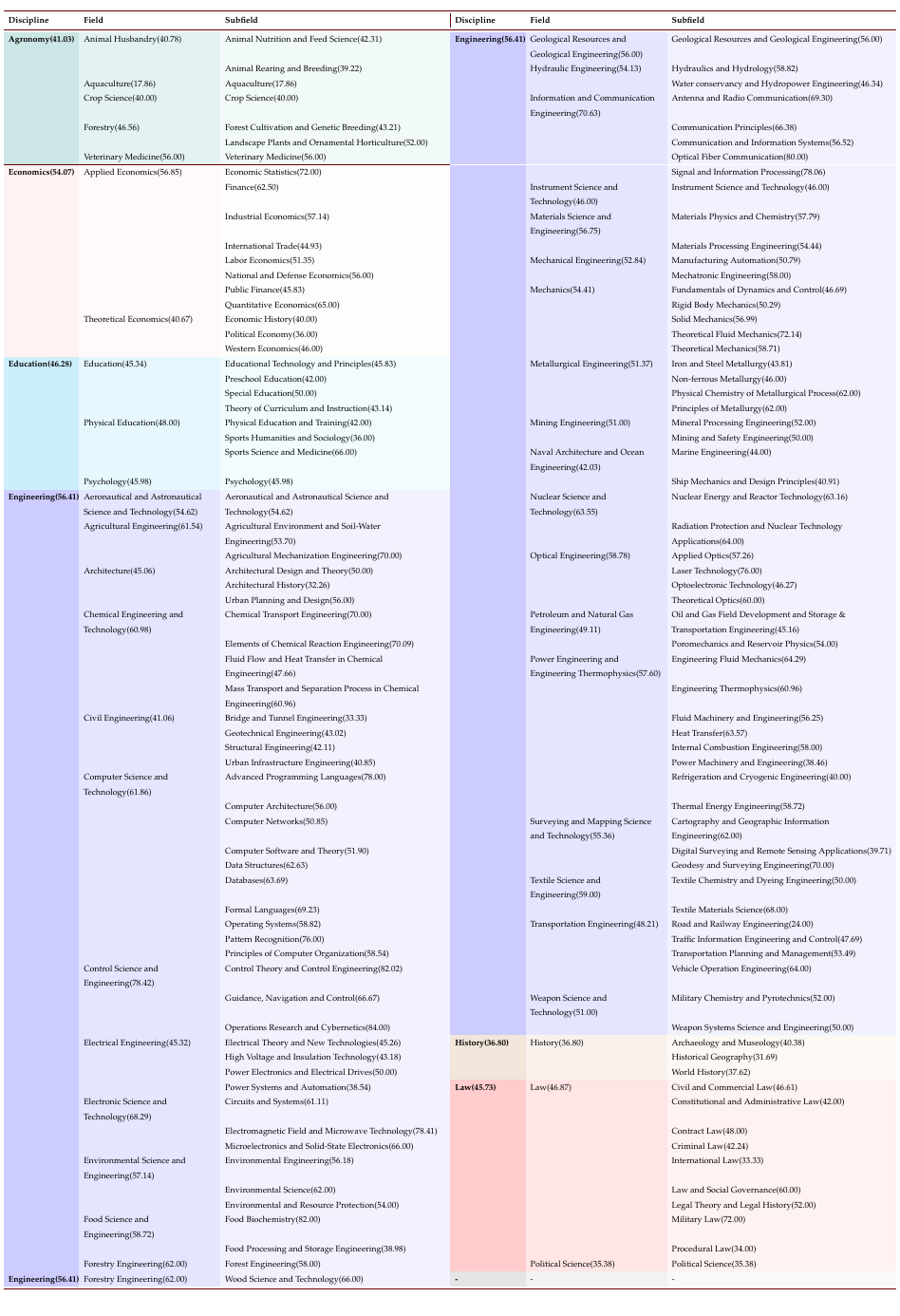} 
\end{subtable}
\vspace{-1.3cm}
\captionsetup{font=small}
\caption{Model Scores Across Three Levels of Disciplines: o3-mini-2025-01-31-high.}
\label{tab:o3-mini-2025-01-31-high}
\vspace{-0.5cm}
\centeredlinks{listofmodels}{Back to List of Models}{toc}{Back to Table of Contents}{blue}
\end{table}
\clearpage

\newpage
\afterpage{
    \begin{table}[t]
    \centering
    \ContinuedFloat 
    \begin{subtable}[t]{\textwidth}
    \centering
    \includegraphics[width=\textwidth]{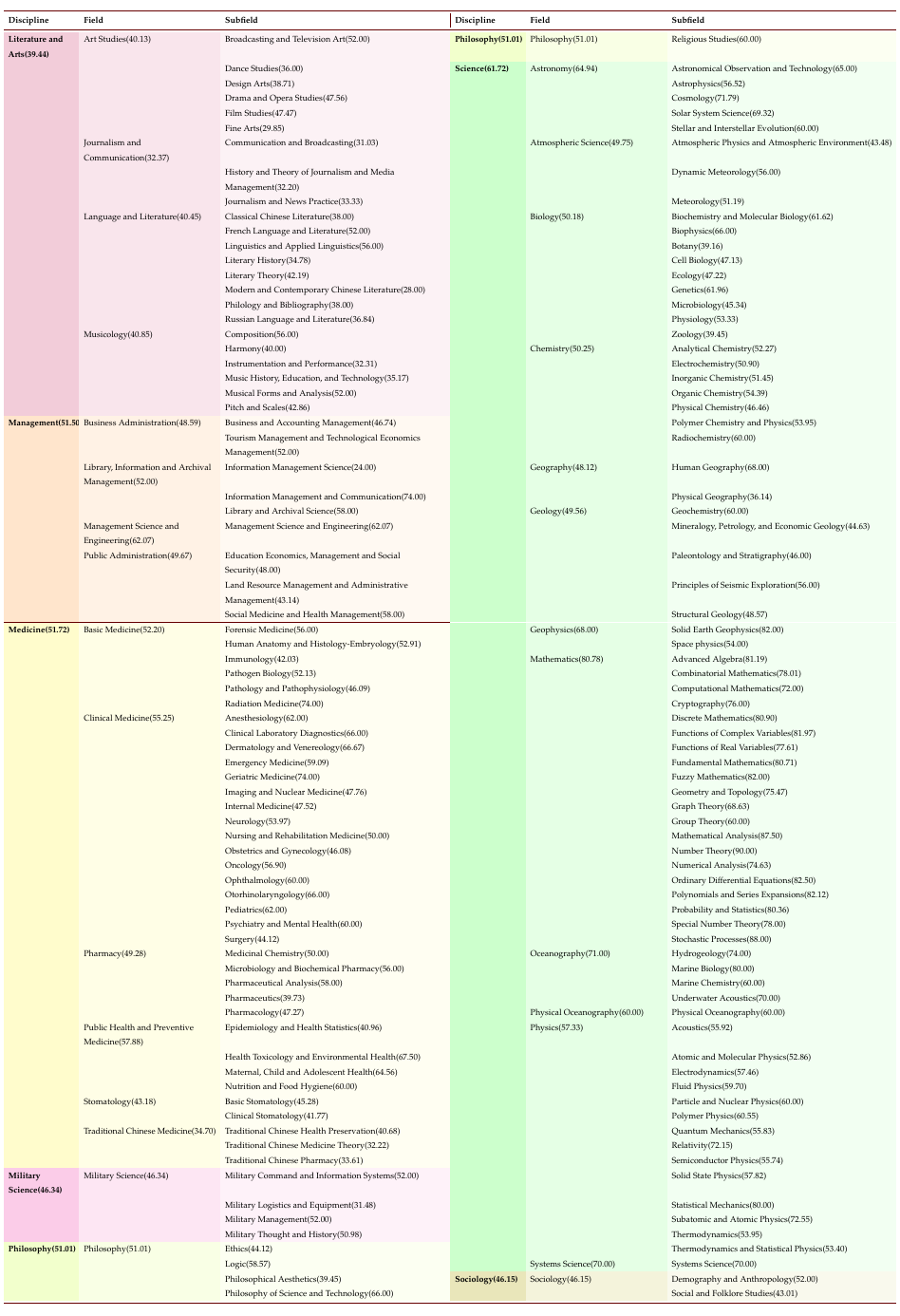} 
    \end{subtable}
    \vspace{-1.1cm}
    \captionsetup{font=small}
    \caption{Continued: Model Scores Across Three Levels of Disciplines: o3-mini-2025-01-31-high.}
    \vspace{-0.6cm}
    \centeredlinks{listofmodels}{Back to List of Models}{toc}{Back to Table of Contents}{blue}
    \end{table}
}
\clearpage

\newpage
\vspace{-0.5cm}
\begin{table}[t]
\refstepcounter{models}%
\addcontentsline{csf}{models}{\protect\numberline{\themodels}Doubao-1.5-pro-32k-250115}
\centering
\begin{subtable}[t]{1\textwidth}
\centering
\includegraphics[width=\textwidth]{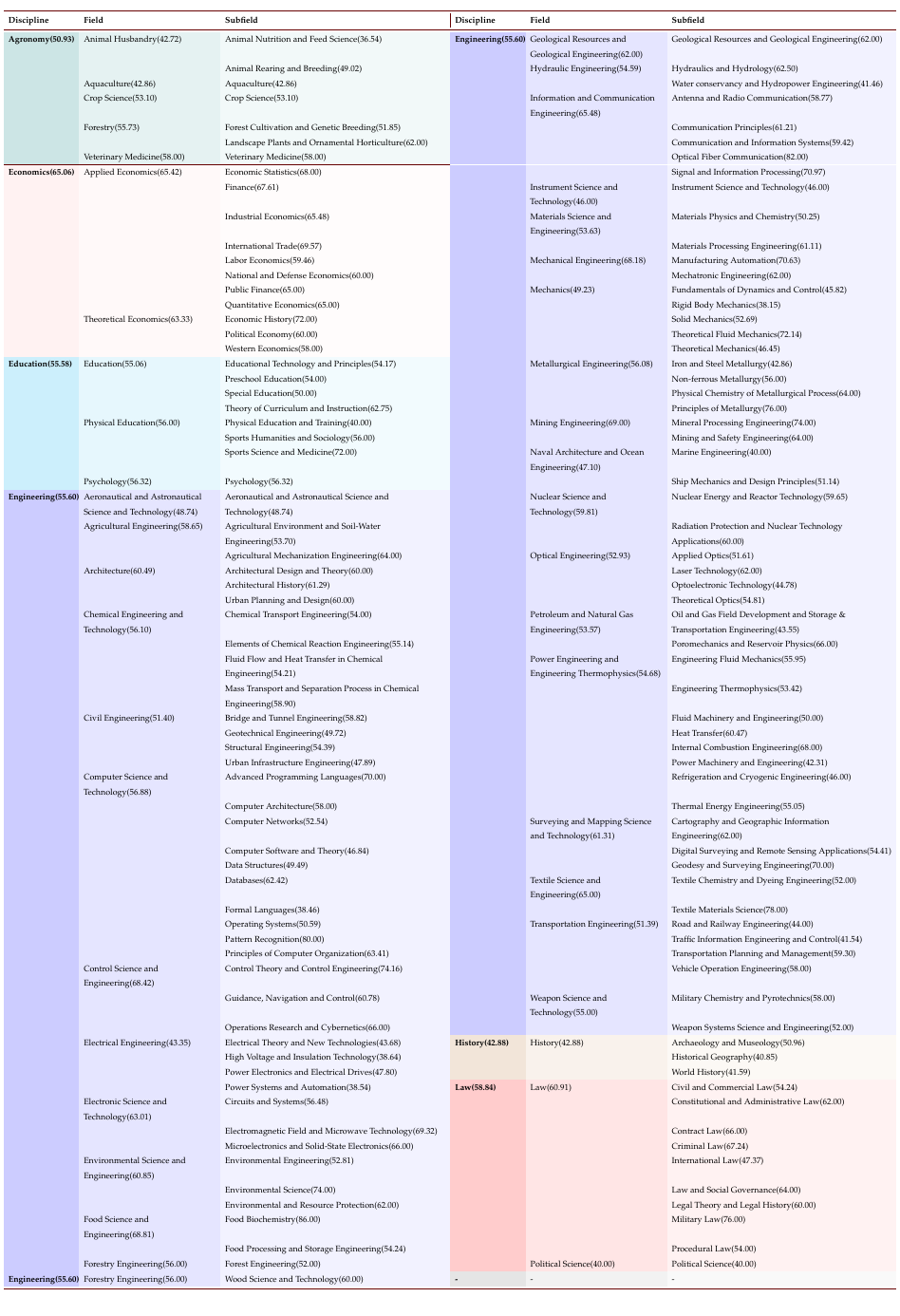} 
\end{subtable}
\vspace{-1.3cm}
\captionsetup{font=small}
\caption{Model Scores Across Three Levels of Disciplines: Doubao-1.5-pro-32k-250115.}
\label{tab:Doubao-1.5-pro-32k-250115}
\vspace{-0.5cm}
\centeredlinks{listofmodels}{Back to List of Models}{toc}{Back to Table of Contents}{blue}
\end{table}
\clearpage

\newpage
\afterpage{
    \begin{table}[t]
    \centering
    \ContinuedFloat 
    \begin{subtable}[t]{\textwidth}
    \centering
    \includegraphics[width=\textwidth]{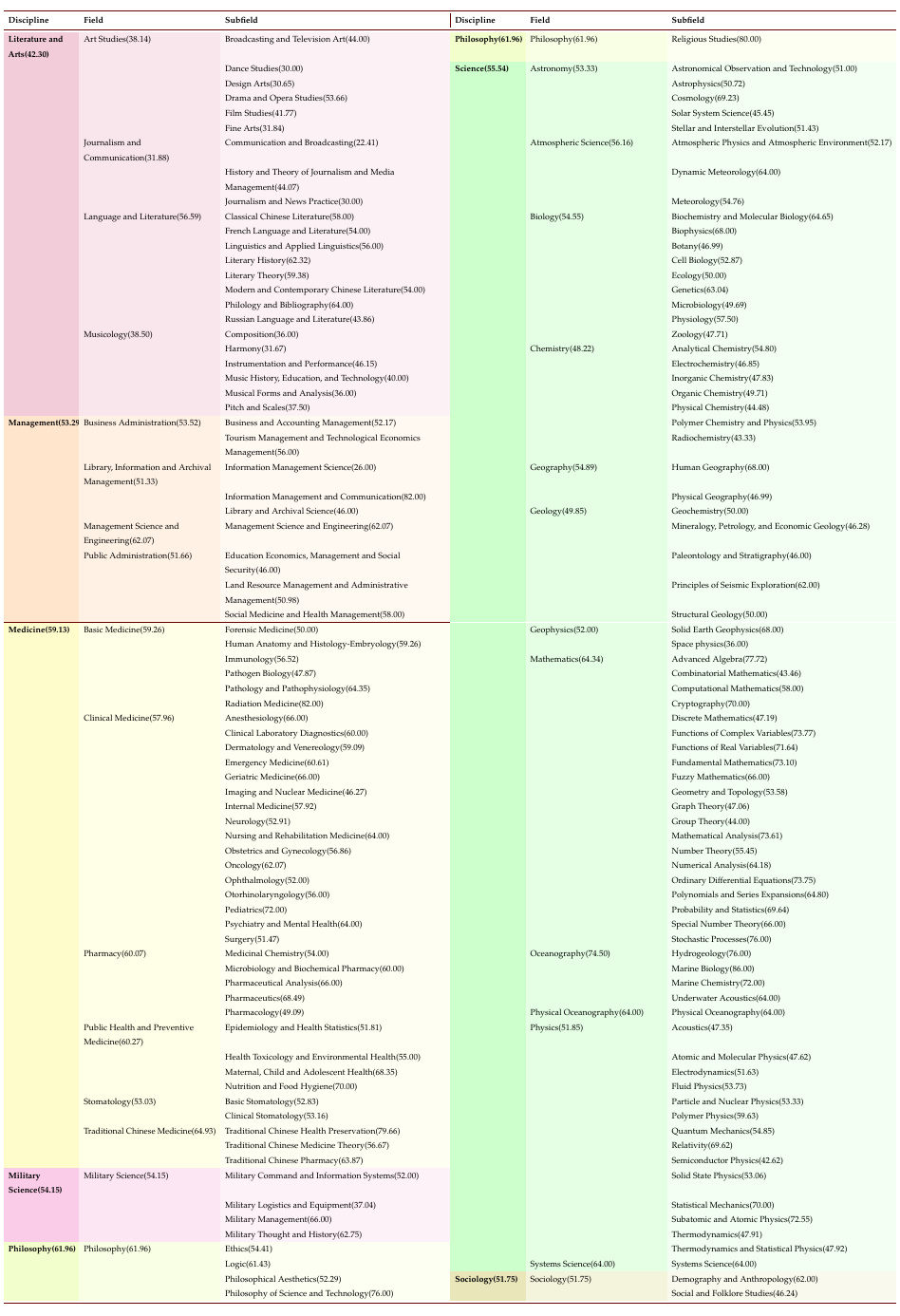} 
    \end{subtable}
    \vspace{-1.1cm}
    \captionsetup{font=small}
    \caption{Continued: Model Scores Across Three Levels of Disciplines: Doubao-1.5-pro-32k-250115.}
    \vspace{-0.6cm}
    \centeredlinks{listofmodels}{Back to List of Models}{toc}{Back to Table of Contents}{blue}
    \end{table}
}
\clearpage

\newpage
\vspace{-0.5cm}
\begin{table}[t]
\refstepcounter{models}%
\addcontentsline{csf}{models}{\protect\numberline{\themodels}o3-mini-2025-01-31-medium}
\centering
\begin{subtable}[t]{1\textwidth}
\centering
\includegraphics[width=\textwidth]{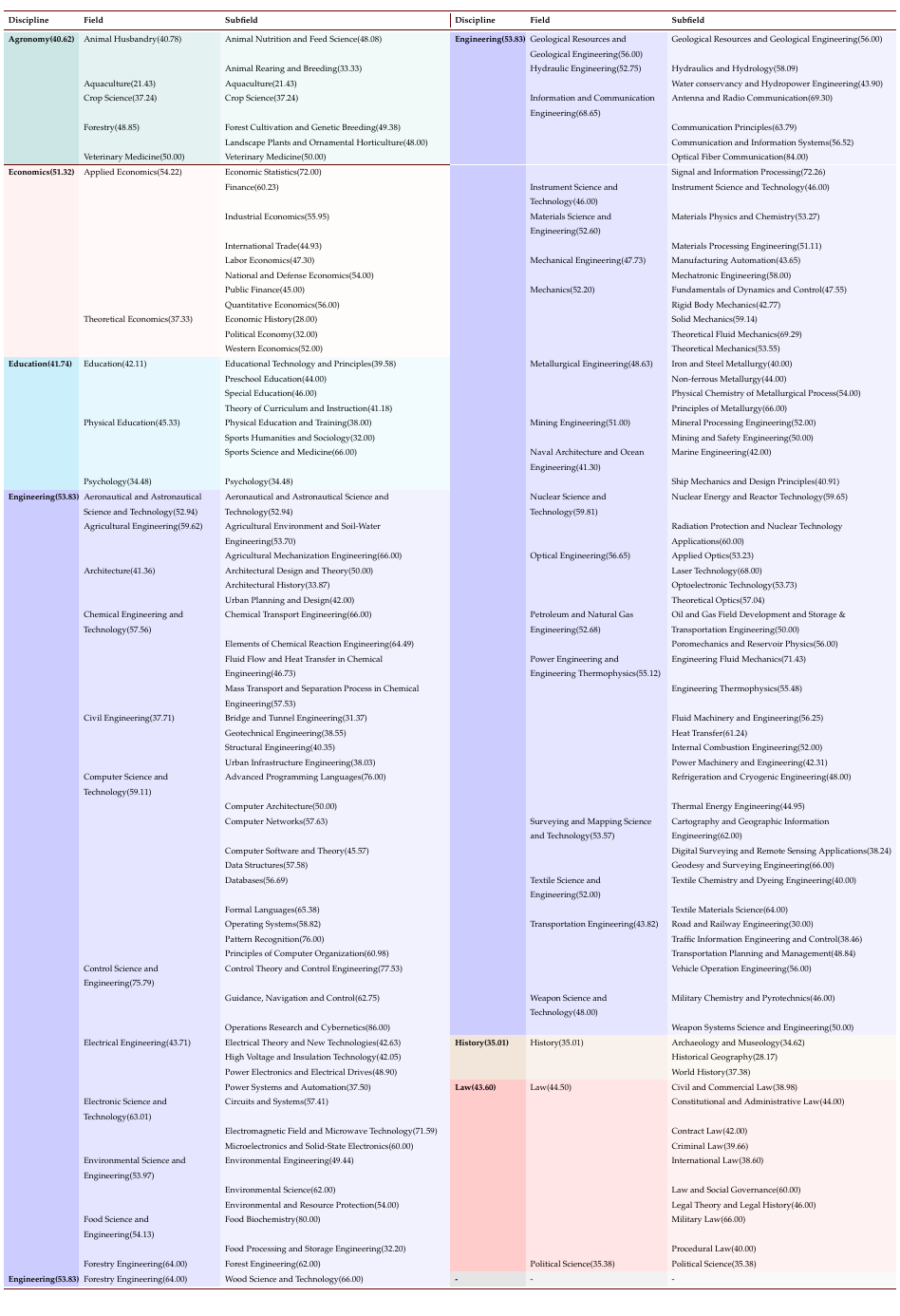} 
\end{subtable}
\vspace{-1.3cm}
\captionsetup{font=small}
\caption{Model Scores Across Three Levels of Disciplines: o3-mini-2025-01-31-medium.}
\label{tab:o3-mini-2025-01-31-medium}
\vspace{-0.5cm}
\centeredlinks{listofmodels}{Back to List of Models}{toc}{Back to Table of Contents}{blue}
\end{table}
\clearpage

\newpage
\afterpage{
    \begin{table}[t]
    \centering
    \ContinuedFloat 
    \begin{subtable}[t]{\textwidth}
    \centering
    \includegraphics[width=\textwidth]{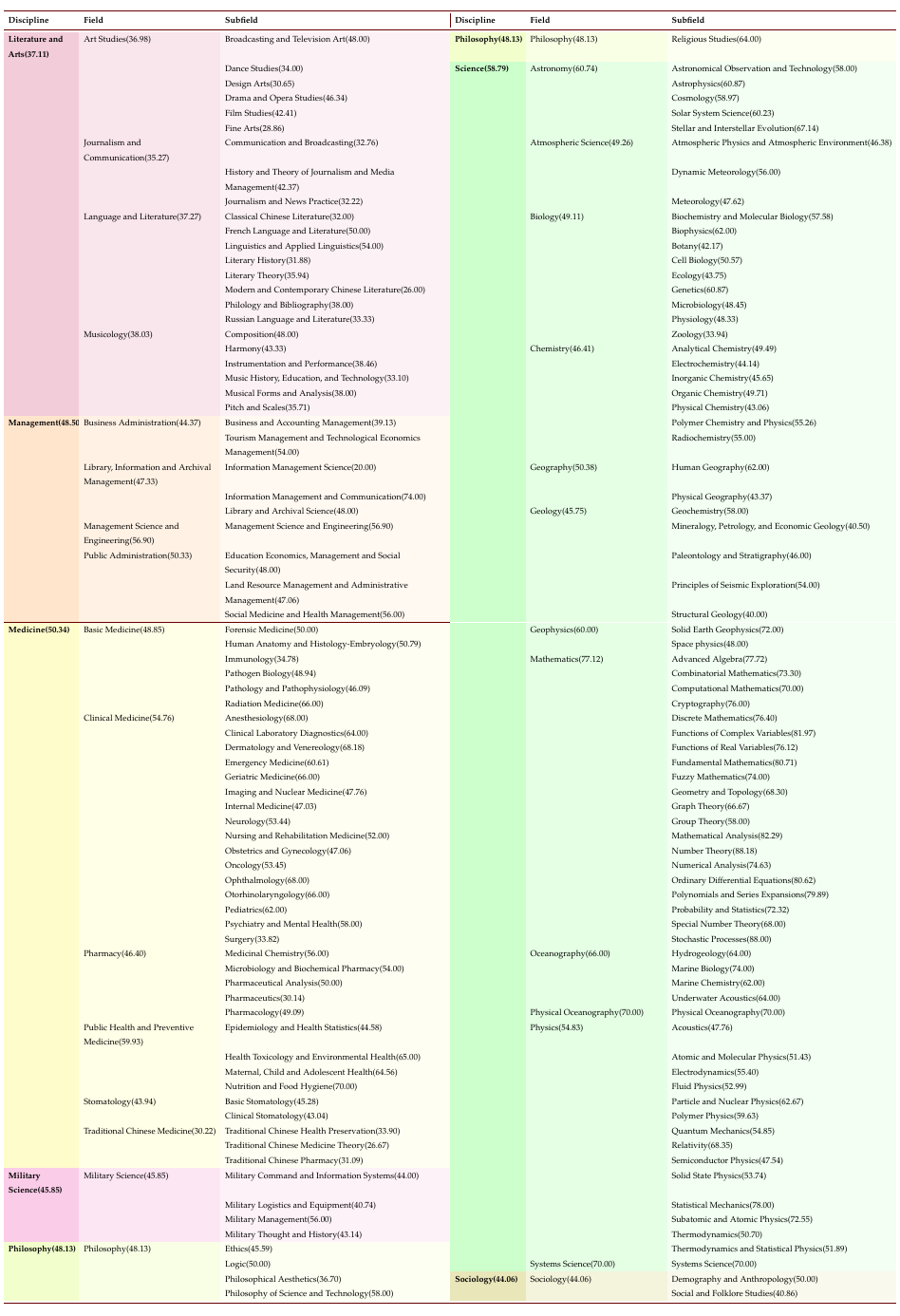} 
    \end{subtable}
    \vspace{-1.1cm}
    \captionsetup{font=small}
    \caption{Continued: Model Scores Across Three Levels of Disciplines: o3-mini-2025-01-31-medium.}
    \vspace{-0.6cm}
    \centeredlinks{listofmodels}{Back to List of Models}{toc}{Back to Table of Contents}{blue}
    \end{table}
}
\clearpage

\newpage
\vspace{-0.5cm}
\begin{table}[t]
\refstepcounter{models}%
\addcontentsline{csf}{models}{\protect\numberline{\themodels}Doubao-1.5-pro-32k-241225}
\centering
\begin{subtable}[t]{1\textwidth}
\centering
\includegraphics[width=\textwidth]{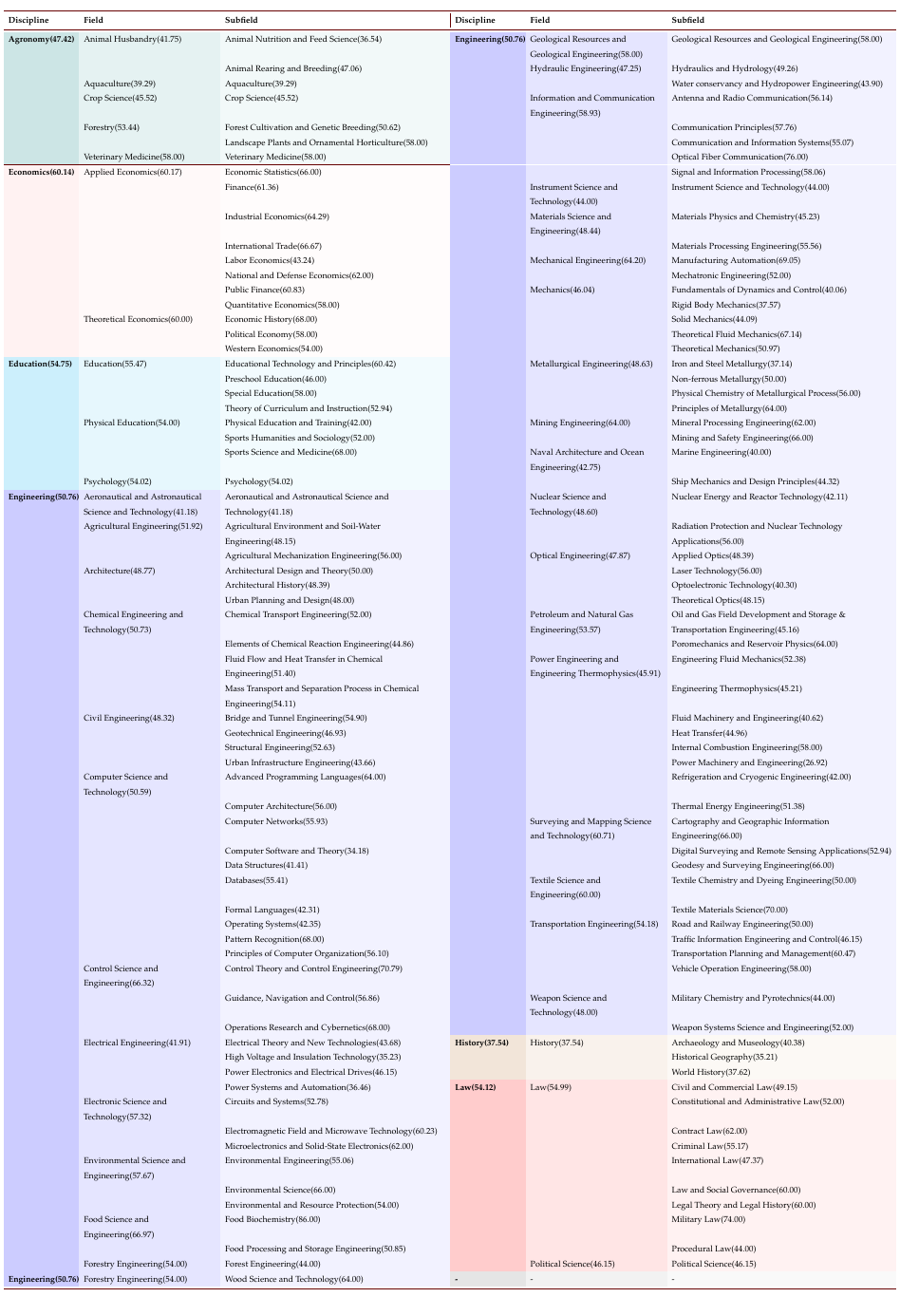} 
\end{subtable}
\vspace{-1.3cm}
\captionsetup{font=small}
\caption{Model Scores Across Three Levels of Disciplines: Doubao-1.5-pro-32k-241225.}
\label{tab:Doubao-1.5-pro-32k-241225}
\vspace{-0.5cm}
\centeredlinks{listofmodels}{Back to List of Models}{toc}{Back to Table of Contents}{blue}
\end{table}
\clearpage

\newpage
\afterpage{
    \begin{table}[t]
    \centering
    \ContinuedFloat 
    \begin{subtable}[t]{\textwidth}
    \centering
    \includegraphics[width=\textwidth]{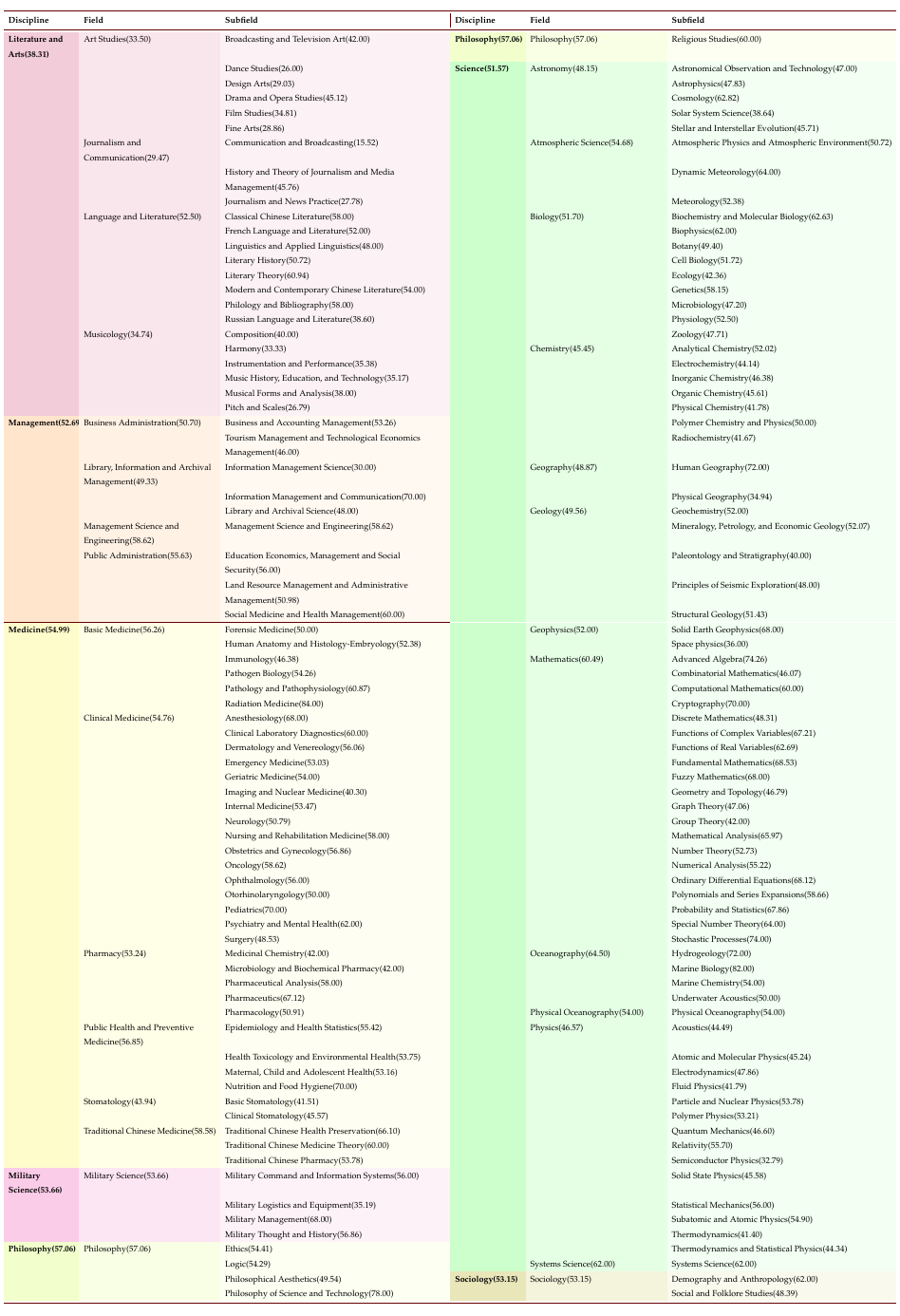} 
    \end{subtable}
    \vspace{-1.1cm}
    \captionsetup{font=small}
    \caption{Continued: Model Scores Across Three Levels of Disciplines: Doubao-1.5-pro-32k-241225.}
    \vspace{-0.6cm}
    \centeredlinks{listofmodels}{Back to List of Models}{toc}{Back to Table of Contents}{blue}
    \end{table}
}
\clearpage

\newpage
\vspace{-0.5cm}
\begin{table}[t]
\refstepcounter{models}%
\addcontentsline{csf}{models}{\protect\numberline{\themodels}qwen-max-2025-01-25}
\centering
\begin{subtable}[t]{1\textwidth}
\centering
\includegraphics[width=\textwidth]{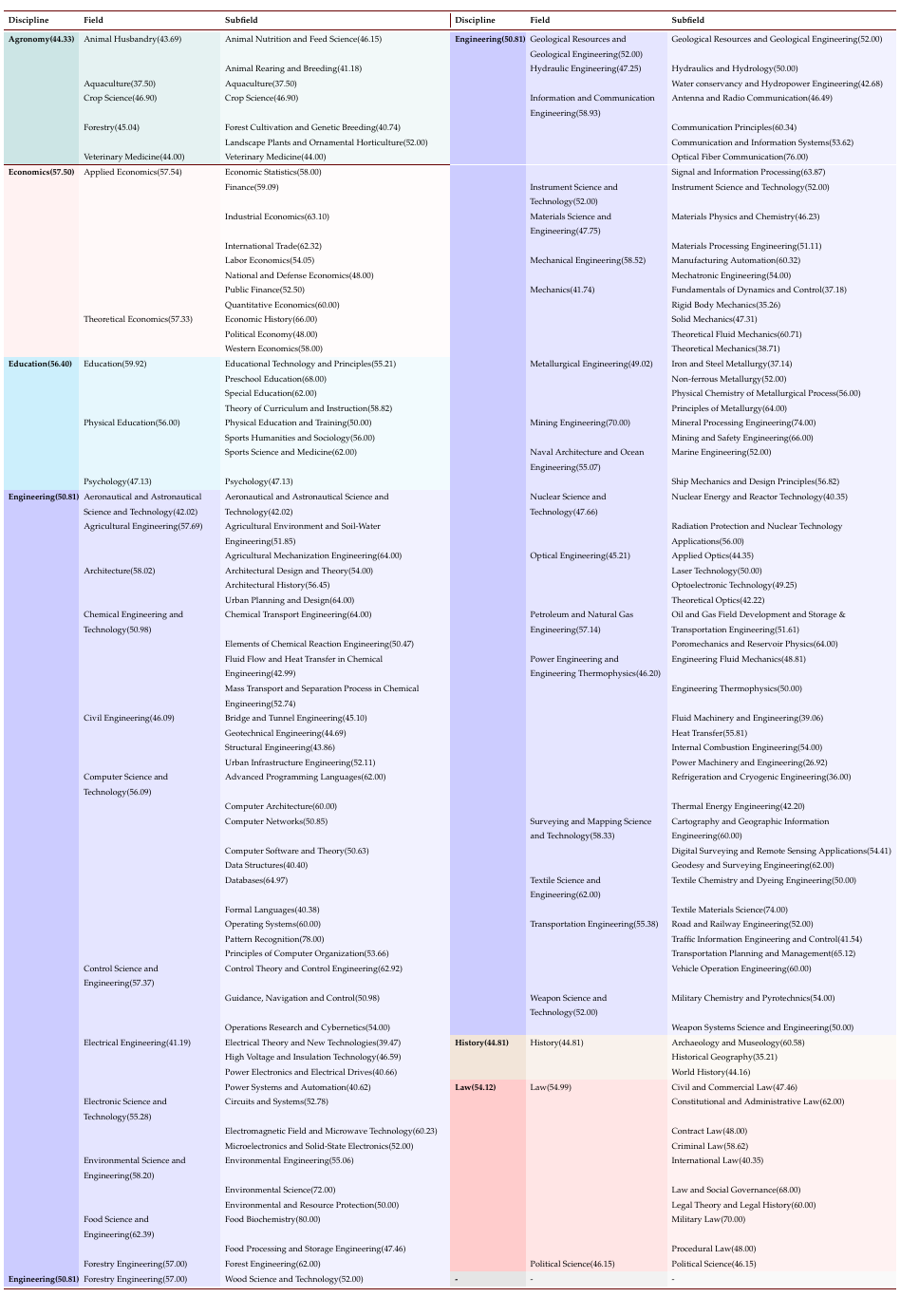} 
\end{subtable}
\vspace{-1.3cm}
\captionsetup{font=small}
\caption{Model Scores Across Three Levels of Disciplines: qwen-max-2025-01-25.}
\label{tab:qwen-max-2025-01-25}
\vspace{-0.5cm}
\centeredlinks{listofmodels}{Back to List of Models}{toc}{Back to Table of Contents}{blue}
\end{table}
\clearpage

\newpage
\afterpage{
    \begin{table}[t]
    \centering
    \ContinuedFloat 
    \begin{subtable}[t]{\textwidth}
    \centering
    \includegraphics[width=\textwidth]{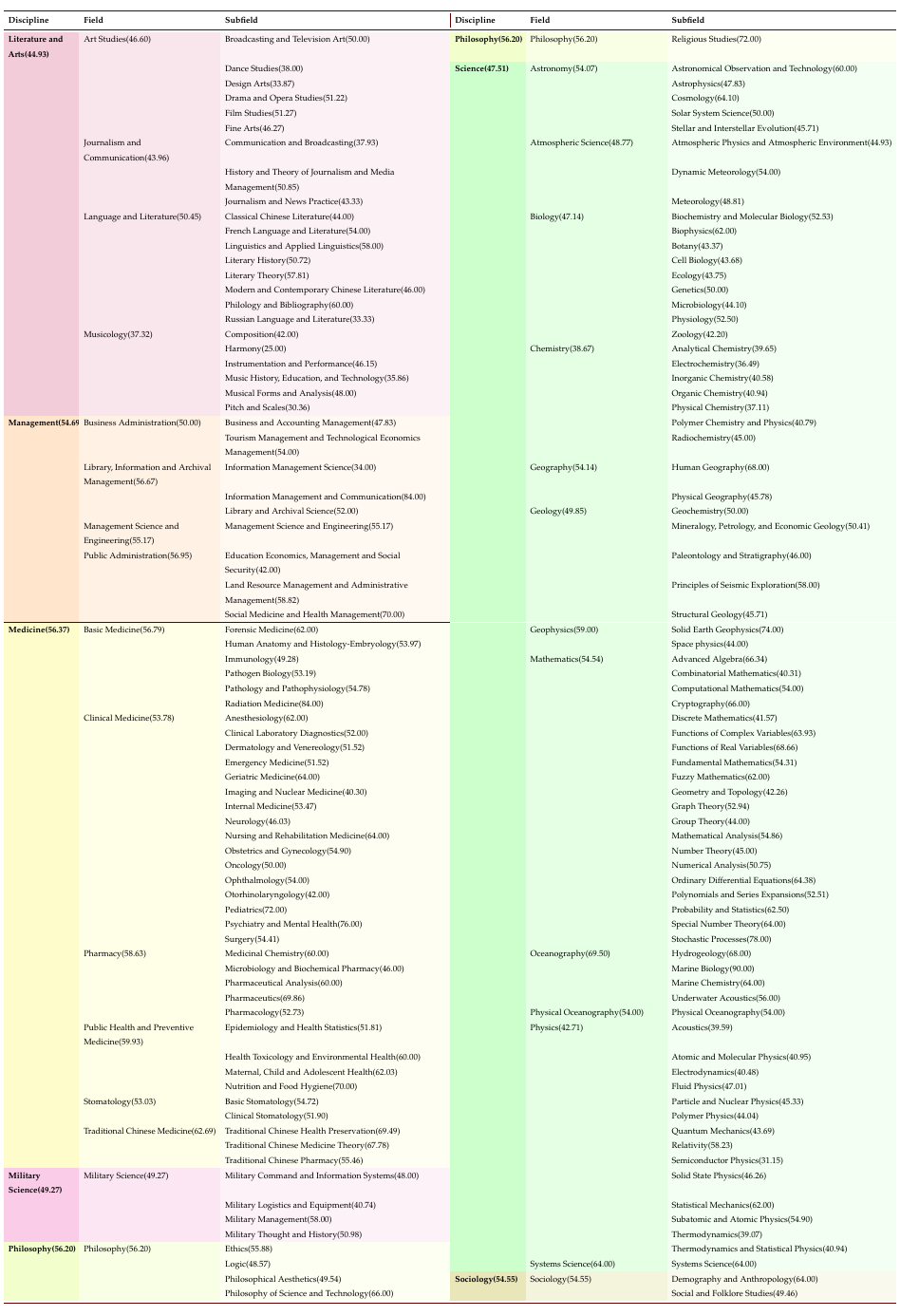} 
    \end{subtable}
    \vspace{-1.1cm}
    \captionsetup{font=small}
    \caption{Continued: Model Scores Across Three Levels of Disciplines: qwen-max-2025-01-25.}
    \vspace{-0.6cm}
    \centeredlinks{listofmodels}{Back to List of Models}{toc}{Back to Table of Contents}{blue}
    \end{table}
}
\clearpage

\newpage
\vspace{-0.5cm}
\begin{table}[t]
\refstepcounter{models}%
\addcontentsline{csf}{models}{\protect\numberline{\themodels}claude-3-5-sonnet-20241022}
\centering
\begin{subtable}[t]{1\textwidth}
\centering
\includegraphics[width=\textwidth]{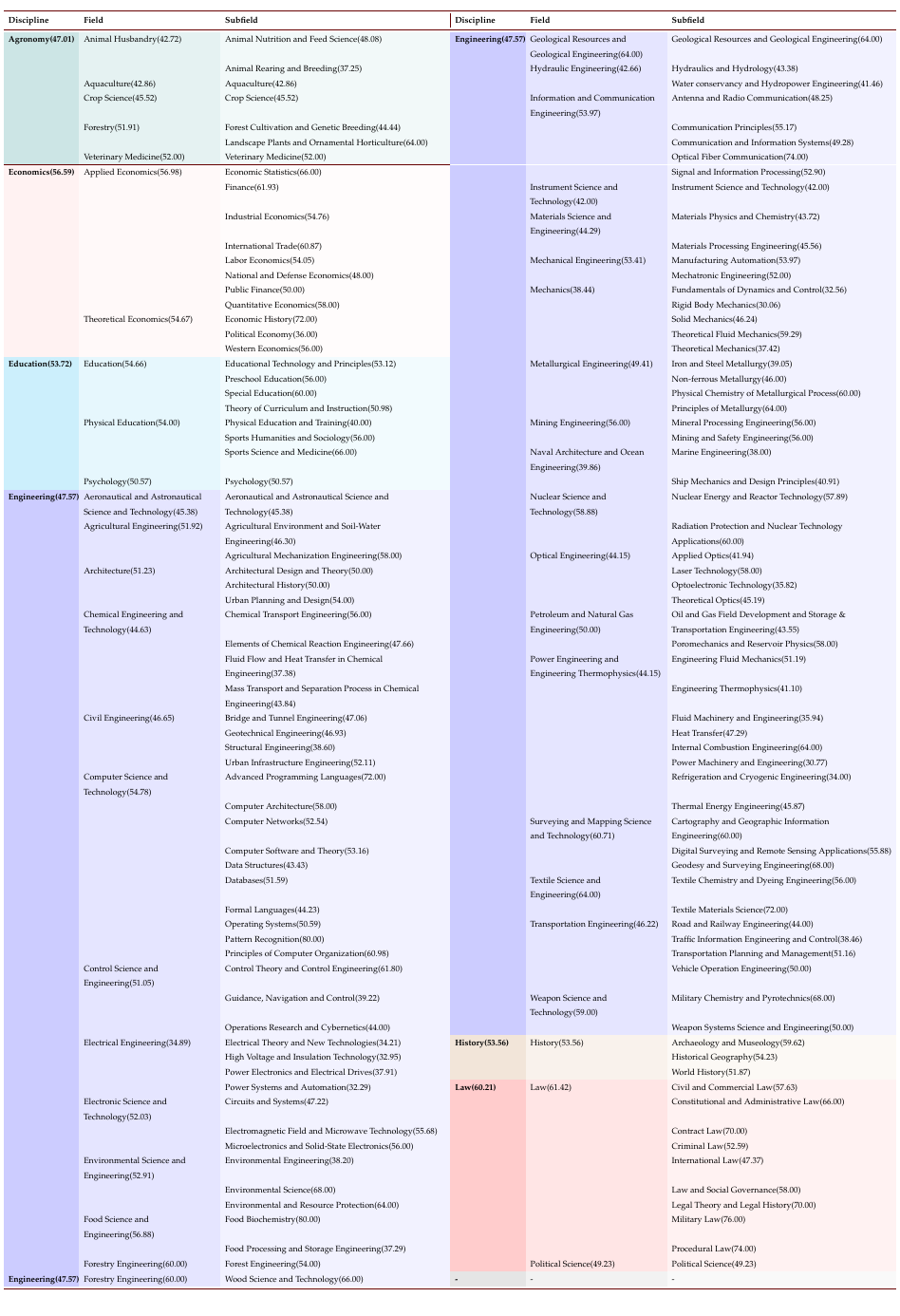} 
\end{subtable}
\vspace{-1.3cm}
\captionsetup{font=small}
\caption{Model Scores Across Three Levels of Disciplines: claude-3-5-sonnet-20241022.}
\label{tab:claude-3-5-sonnet-20241022}
\vspace{-0.5cm}
\centeredlinks{listofmodels}{Back to List of Models}{toc}{Back to Table of Contents}{blue}
\end{table}
\clearpage

\newpage
\afterpage{
    \begin{table}[t]
    \centering
    \ContinuedFloat 
    \begin{subtable}[t]{\textwidth}
    \centering
    \includegraphics[width=\textwidth]{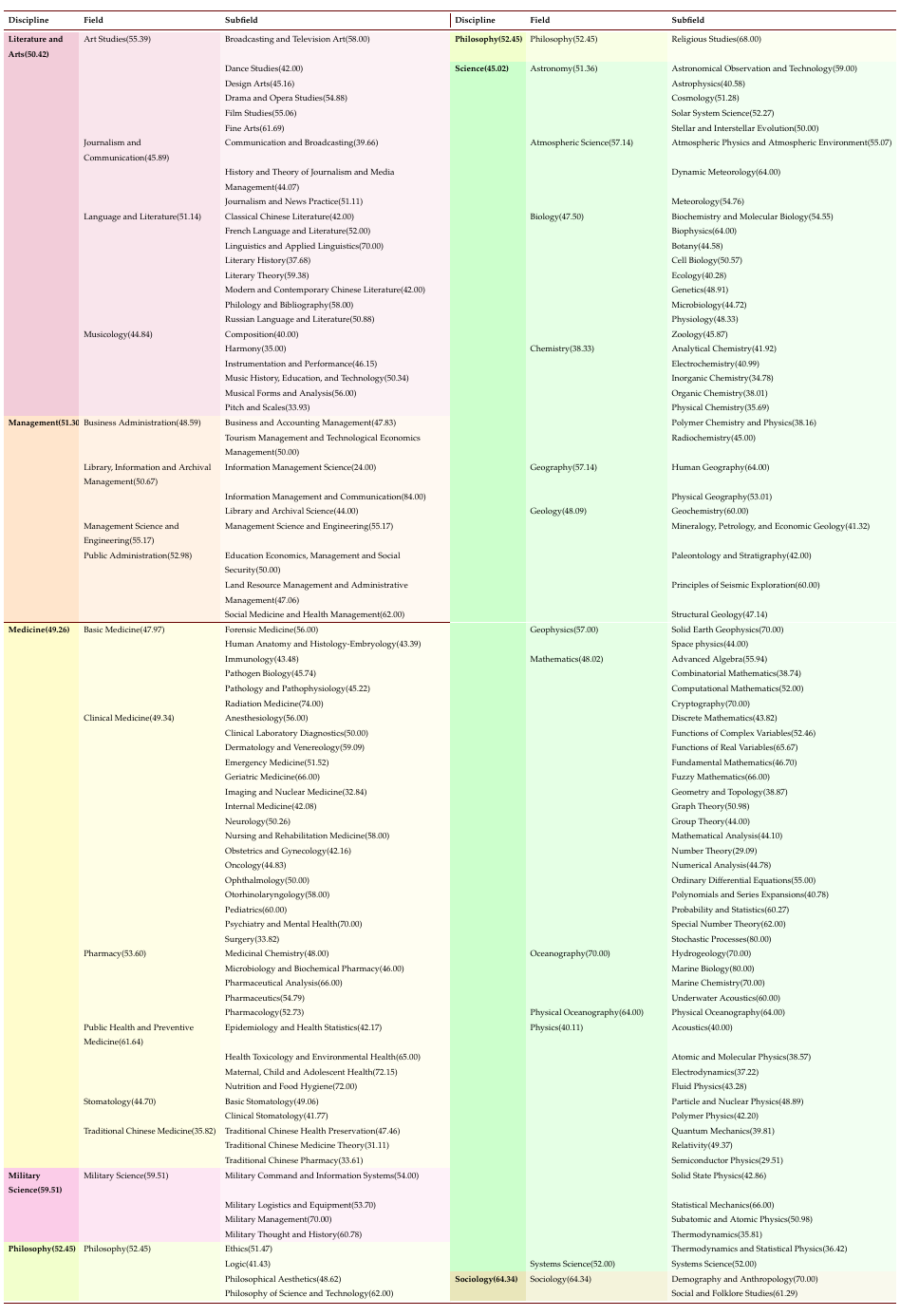} 
    \end{subtable}
    \vspace{-1.1cm}
    \captionsetup{font=small}
    \caption{Continued: Model Scores Across Three Levels of Disciplines: claude-3-5-sonnet-20241022.}
    \vspace{-0.6cm}
    \centeredlinks{listofmodels}{Back to List of Models}{toc}{Back to Table of Contents}{blue}
    \end{table}
}
\clearpage

\newpage
\vspace{-0.5cm}
\begin{table}[t]
\refstepcounter{models}%
\addcontentsline{csf}{models}{\protect\numberline{\themodels}o3-mini-2025-01-31-low}
\centering
\begin{subtable}[t]{1\textwidth}
\centering
\includegraphics[width=\textwidth]{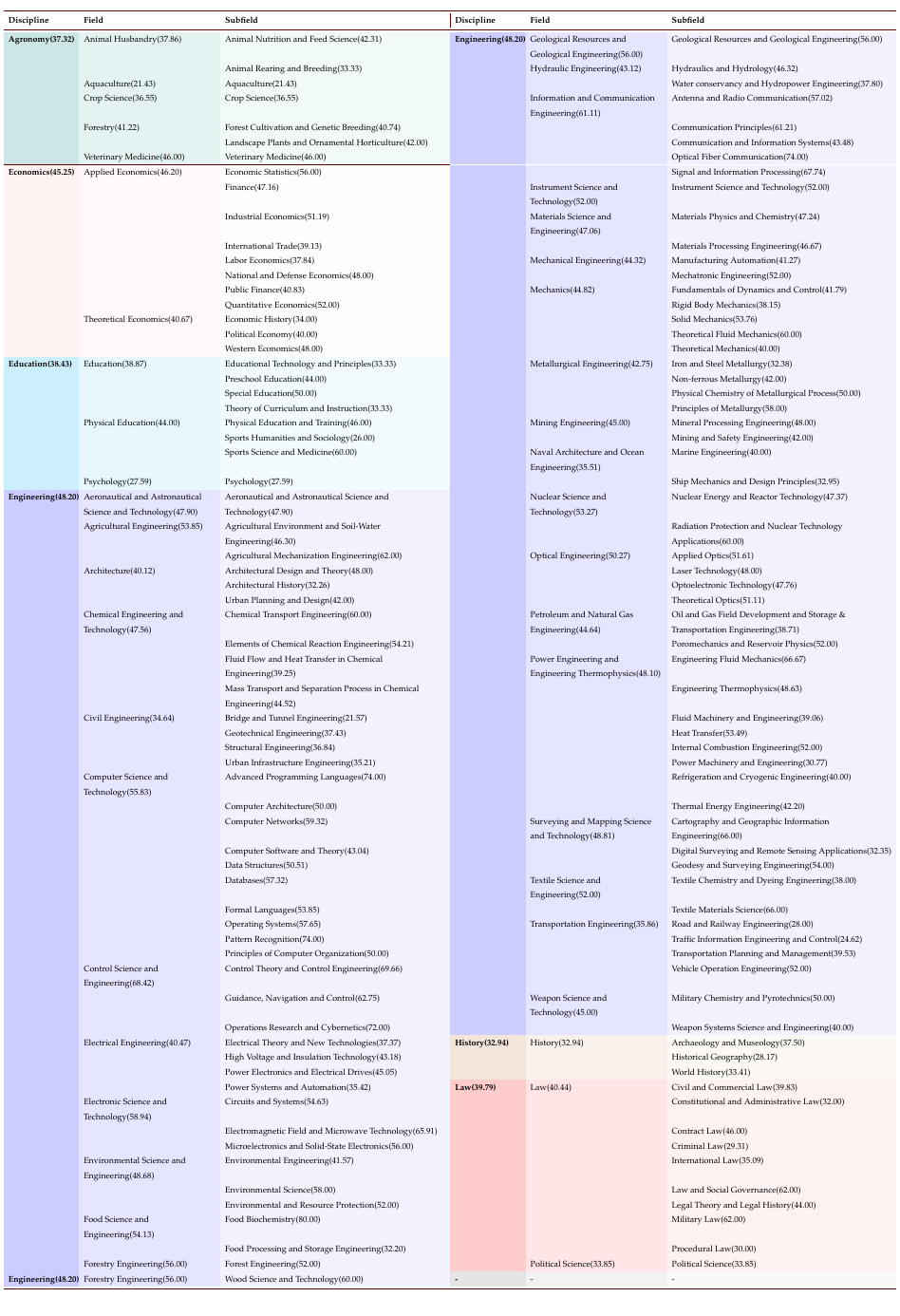} 
\end{subtable}
\vspace{-1.3cm}
\captionsetup{font=small}
\caption{Model Scores Across Three Levels of Disciplines: o3-mini-2025-01-31-low.}
\label{tab:o3-mini-2025-01-31-low}
\vspace{-0.5cm}
\centeredlinks{listofmodels}{Back to List of Models}{toc}{Back to Table of Contents}{blue}
\end{table}
\clearpage

\newpage
\afterpage{
    \begin{table}[t]
    \centering
    \ContinuedFloat 
    \begin{subtable}[t]{\textwidth}
    \centering
    \includegraphics[width=\textwidth]{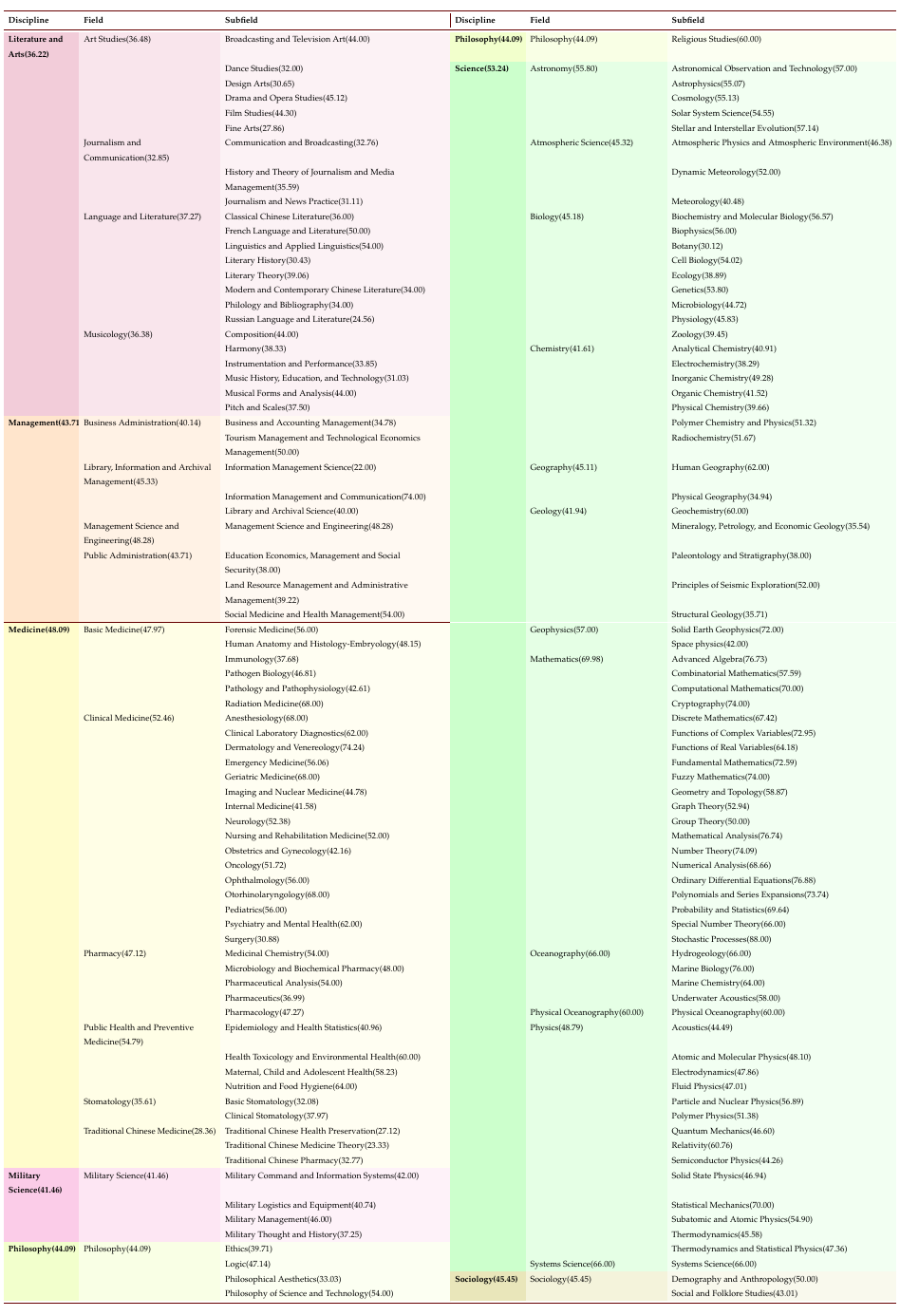} 
    \end{subtable}
    \vspace{-1.1cm}
    \captionsetup{font=small}
    \caption{Continued: Model Scores Across Three Levels of Disciplines: o3-mini-2025-01-31-low.}
    \vspace{-0.6cm}
    \centeredlinks{listofmodels}{Back to List of Models}{toc}{Back to Table of Contents}{blue}
    \end{table}
}
\clearpage

\newpage
\vspace{-0.5cm}
\begin{table}[t]
\refstepcounter{models}%
\addcontentsline{csf}{models}{\protect\numberline{\themodels}gemini-2.0-flash}
\centering
\begin{subtable}[t]{1\textwidth}
\centering
\includegraphics[width=\textwidth]{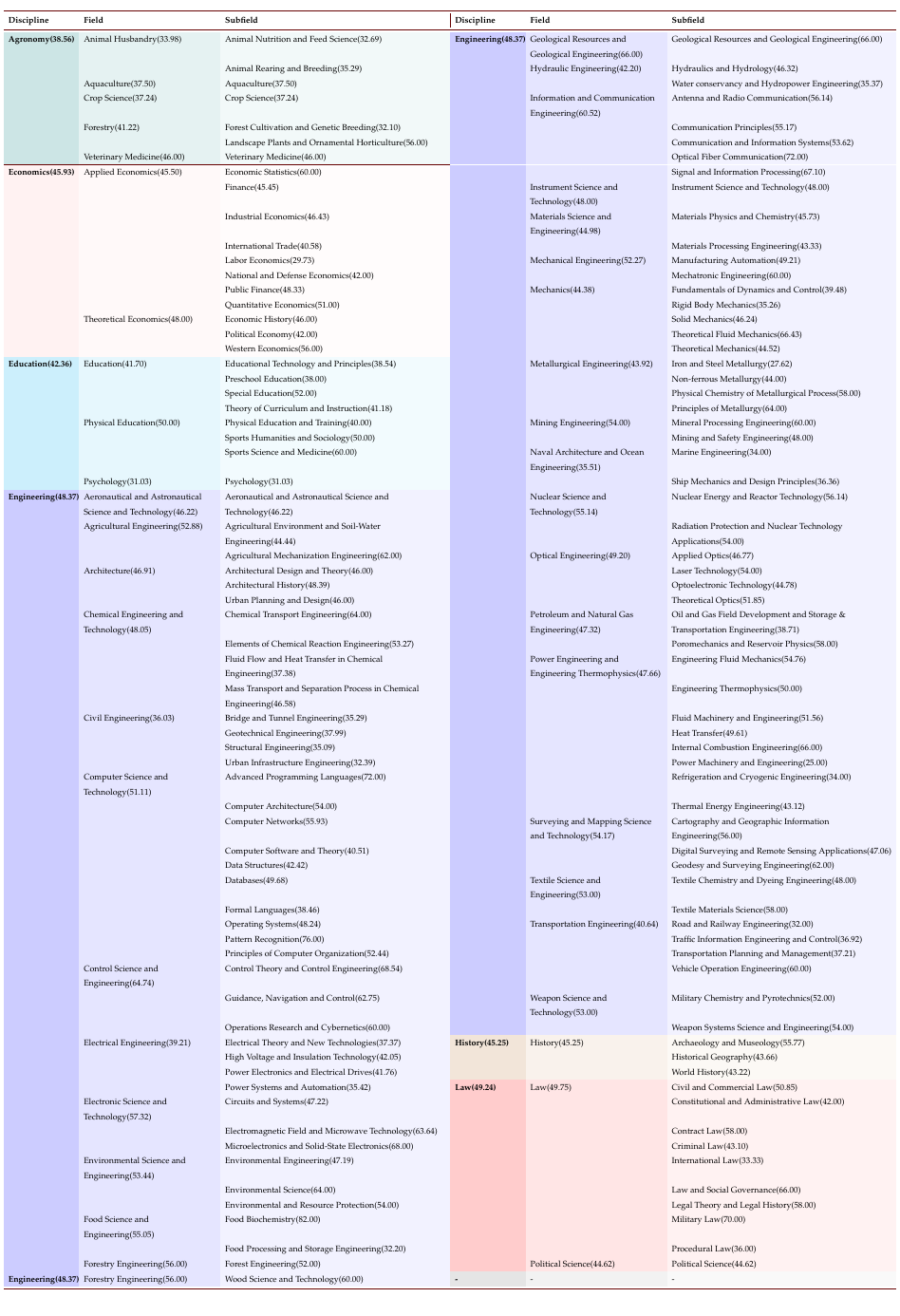} 
\end{subtable}
\vspace{-1.3cm}
\captionsetup{font=small}
\caption{Model Scores Across Three Levels of Disciplines: gemini-2.0-flash.}
\label{tab:gemini-2.0-flash}
\vspace{-0.5cm}
\centeredlinks{listofmodels}{Back to List of Models}{toc}{Back to Table of Contents}{blue}
\end{table}
\clearpage

\newpage
\afterpage{
    \begin{table}[t]
    \centering
    \ContinuedFloat 
    \begin{subtable}[t]{\textwidth}
    \centering
    \includegraphics[width=\textwidth]{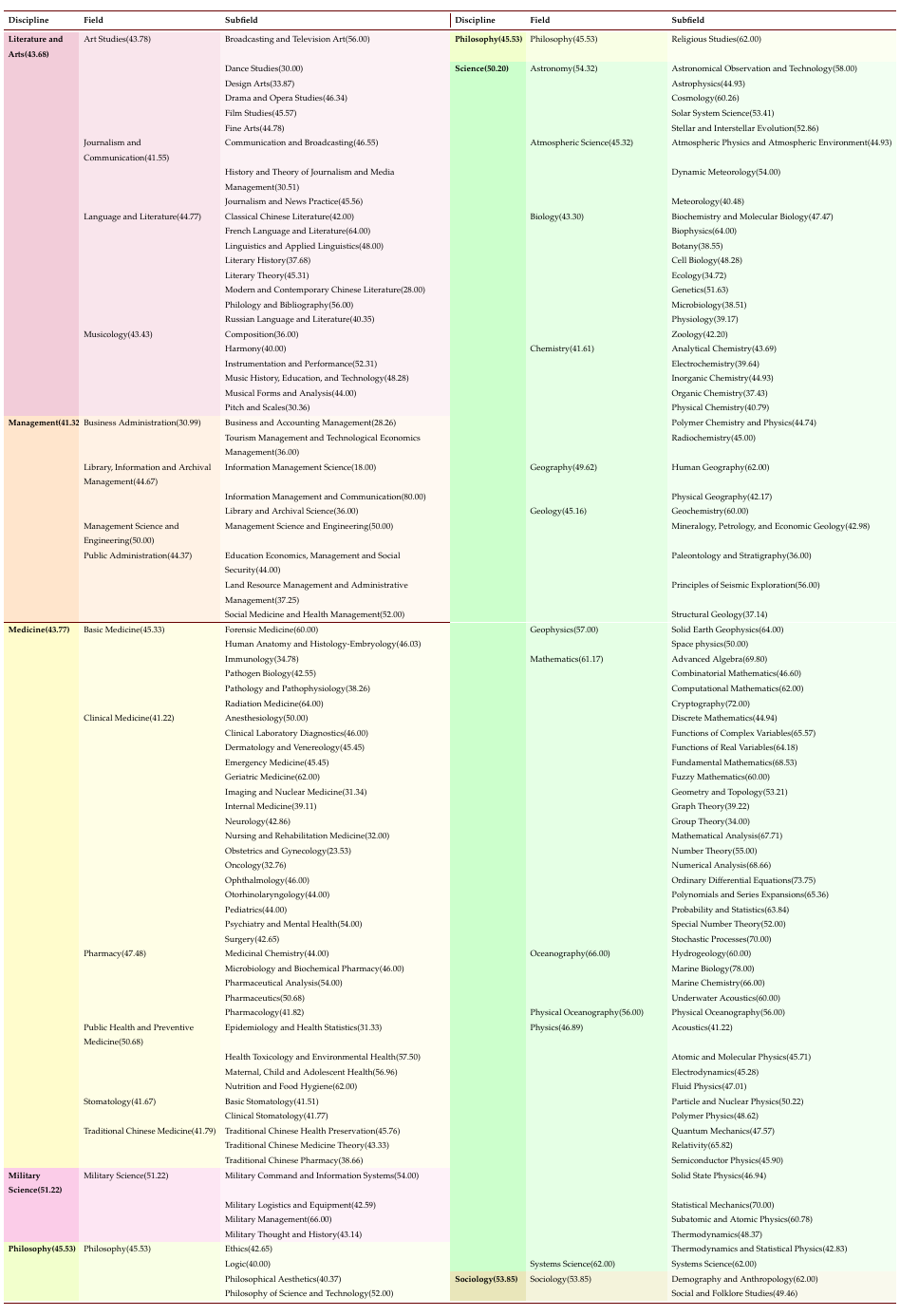} 
    \end{subtable}
    \vspace{-1.1cm}
    \captionsetup{font=small}
    \caption{Continued: Model Scores Across Three Levels of Disciplines: gemini-2.0-flash.}
    \vspace{-0.6cm}
    \centeredlinks{listofmodels}{Back to List of Models}{toc}{Back to Table of Contents}{blue}
    \end{table}
}
\clearpage

\newpage
\vspace{-0.5cm}
\begin{table}[t]
\refstepcounter{models}%
\addcontentsline{csf}{models}{\protect\numberline{\themodels}DeepSeek-V3}
\centering
\begin{subtable}[t]{1\textwidth}
\centering
\includegraphics[width=\textwidth]{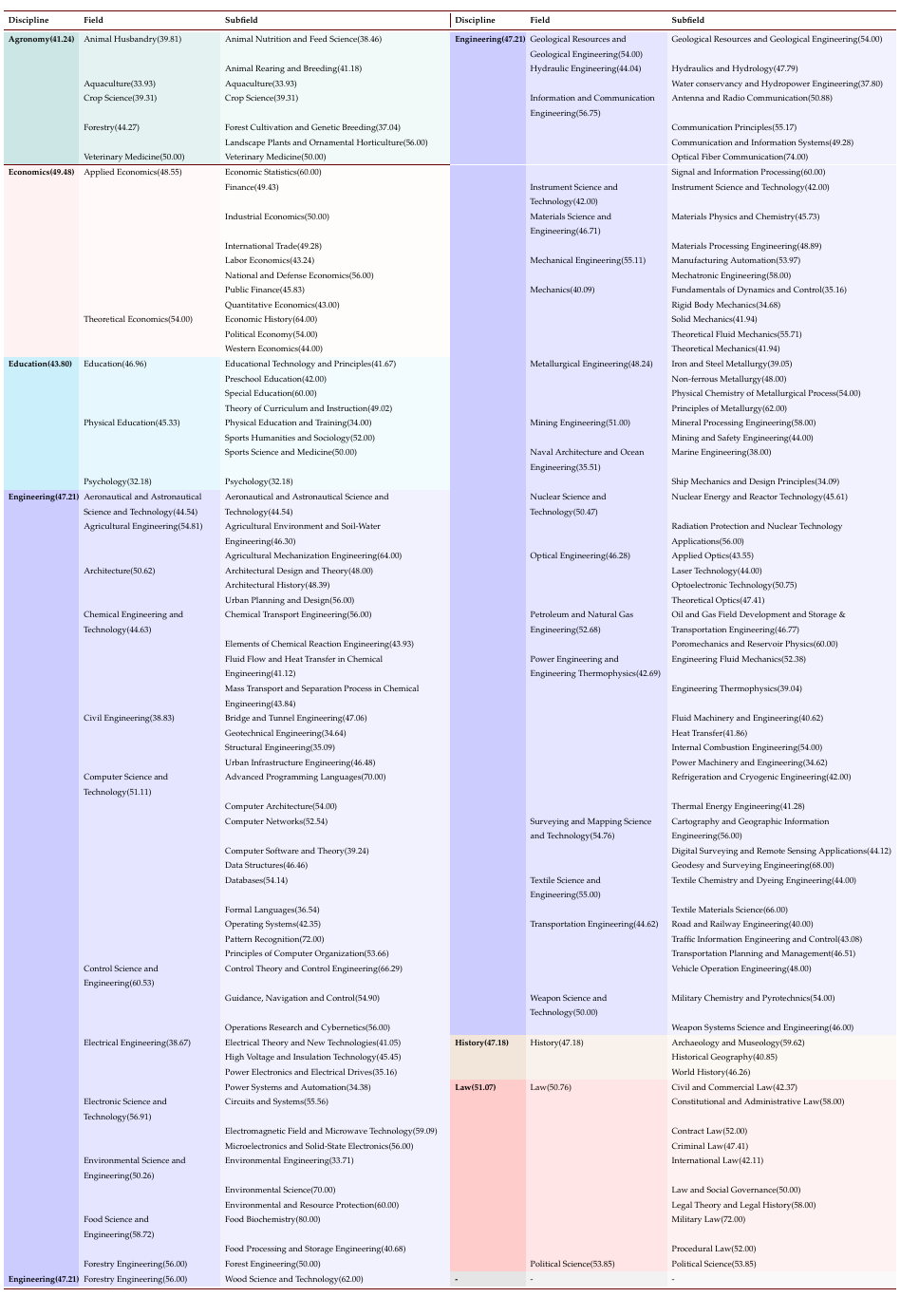} 
\end{subtable}
\vspace{-1.3cm}
\captionsetup{font=small}
\caption{Model Scores Across Three Levels of Disciplines: DeepSeek-V3.}
\label{tab:DeepSeek-V3}
\vspace{-0.5cm}
\centeredlinks{listofmodels}{Back to List of Models}{toc}{Back to Table of Contents}{blue}
\end{table}
\clearpage

\newpage
\afterpage{
    \begin{table}[t]
    \centering
    \ContinuedFloat 
    \begin{subtable}[t]{\textwidth}
    \centering
    \includegraphics[width=\textwidth]{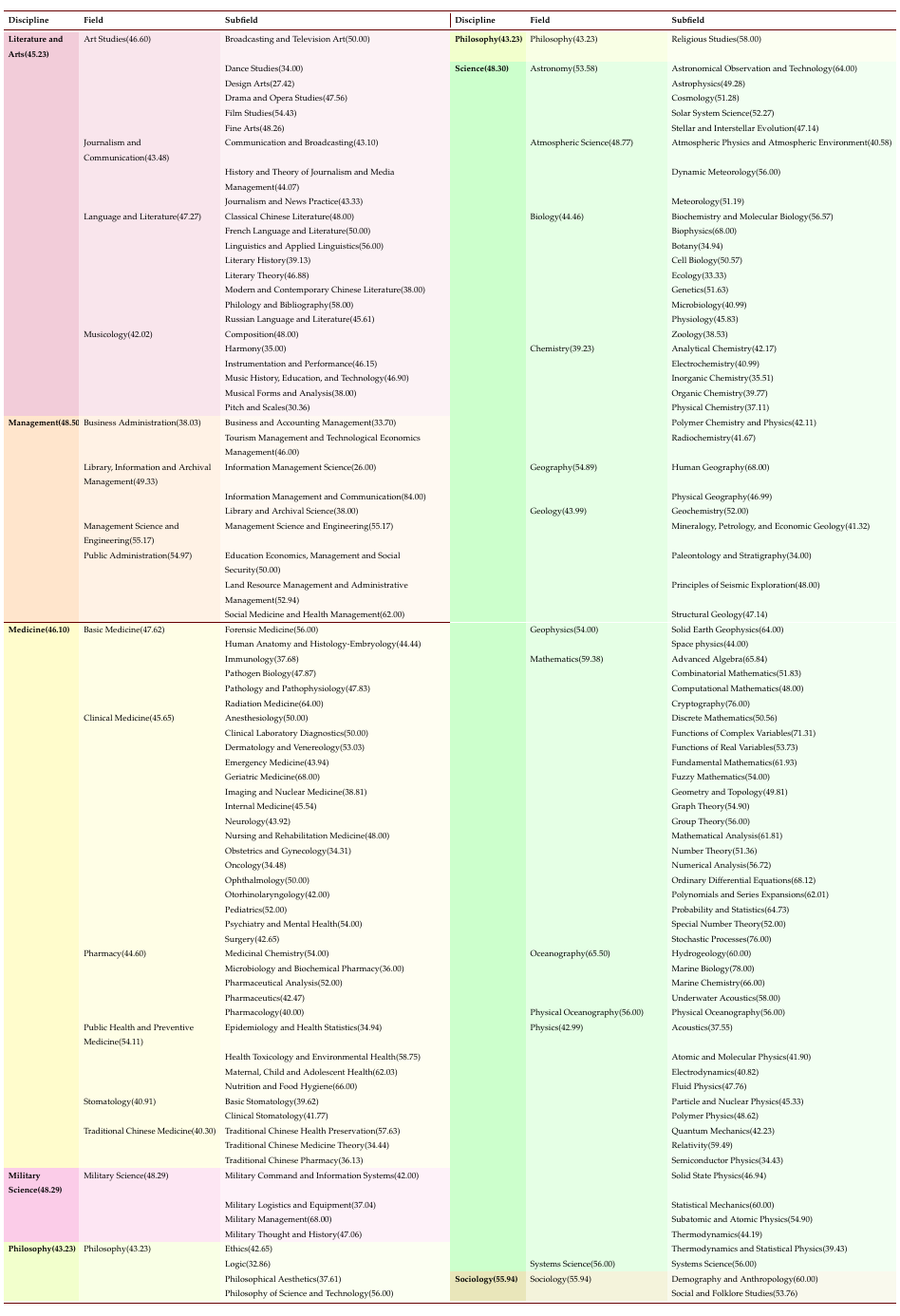} 
    \end{subtable}
    \vspace{-1.1cm}
    \captionsetup{font=small}
    \caption{Continued: Model Scores Across Three Levels of Disciplines: DeepSeek-V3.}
    \vspace{-0.6cm}
    \centeredlinks{listofmodels}{Back to List of Models}{toc}{Back to Table of Contents}{blue}
    \end{table}
}
\clearpage

\newpage
\vspace{-0.5cm}
\begin{table}[t]
\refstepcounter{models}%
\addcontentsline{csf}{models}{\protect\numberline{\themodels}o1-mini-2024-09-12}
\centering
\begin{subtable}[t]{1\textwidth}
\centering
\includegraphics[width=\textwidth]{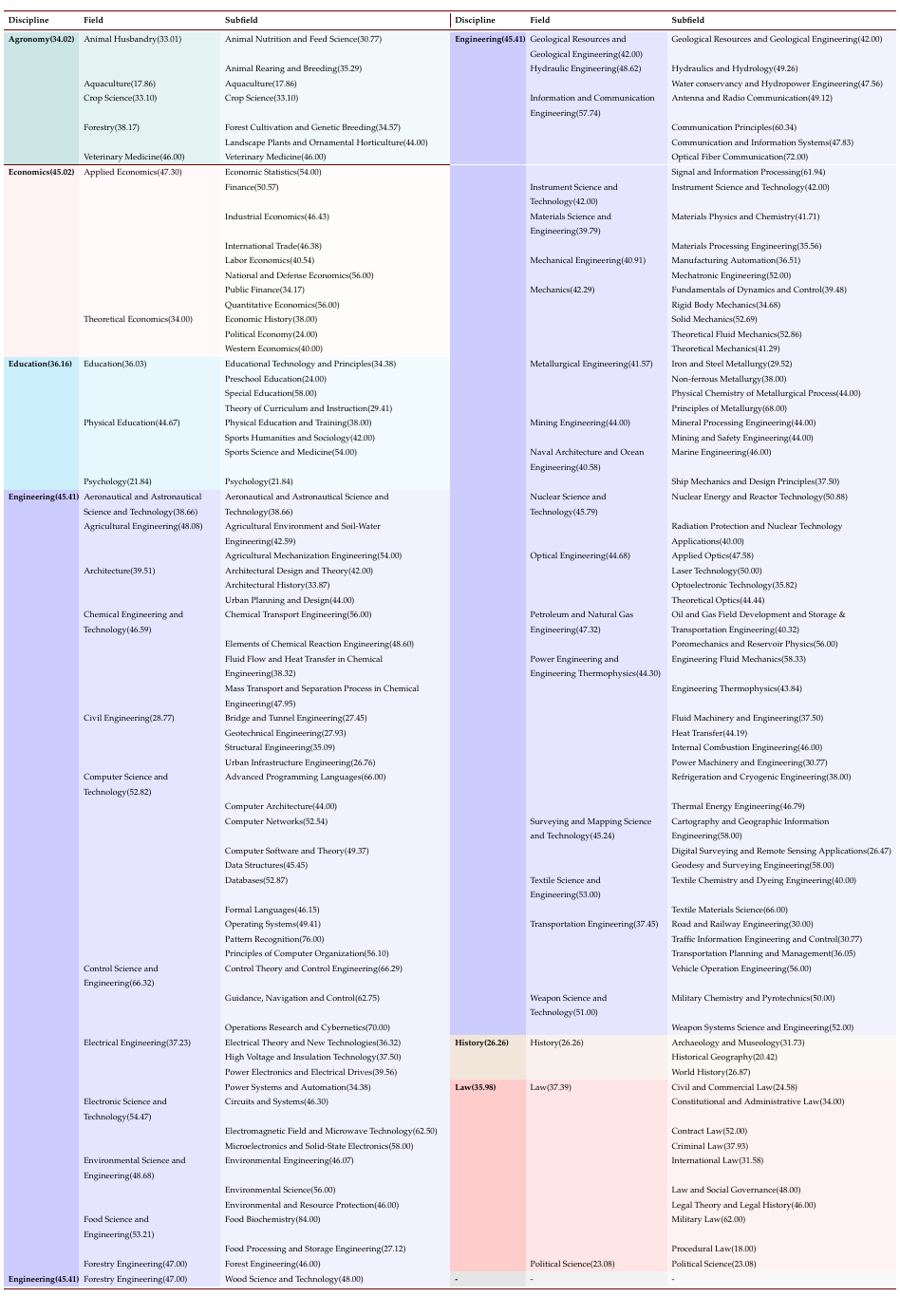} 
\end{subtable}
\vspace{-1.3cm}
\captionsetup{font=small}
\caption{Model Scores Across Three Levels of Disciplines: o1-mini-2024-09-12.}
\label{tab:o1-mini-2024-09-12}
\vspace{-0.5cm}
\centeredlinks{listofmodels}{Back to List of Models}{toc}{Back to Table of Contents}{blue}
\end{table}
\clearpage

\newpage
\afterpage{
    \begin{table}[t]
    \centering
    \ContinuedFloat 
    \begin{subtable}[t]{\textwidth}
    \centering
    \includegraphics[width=\textwidth]{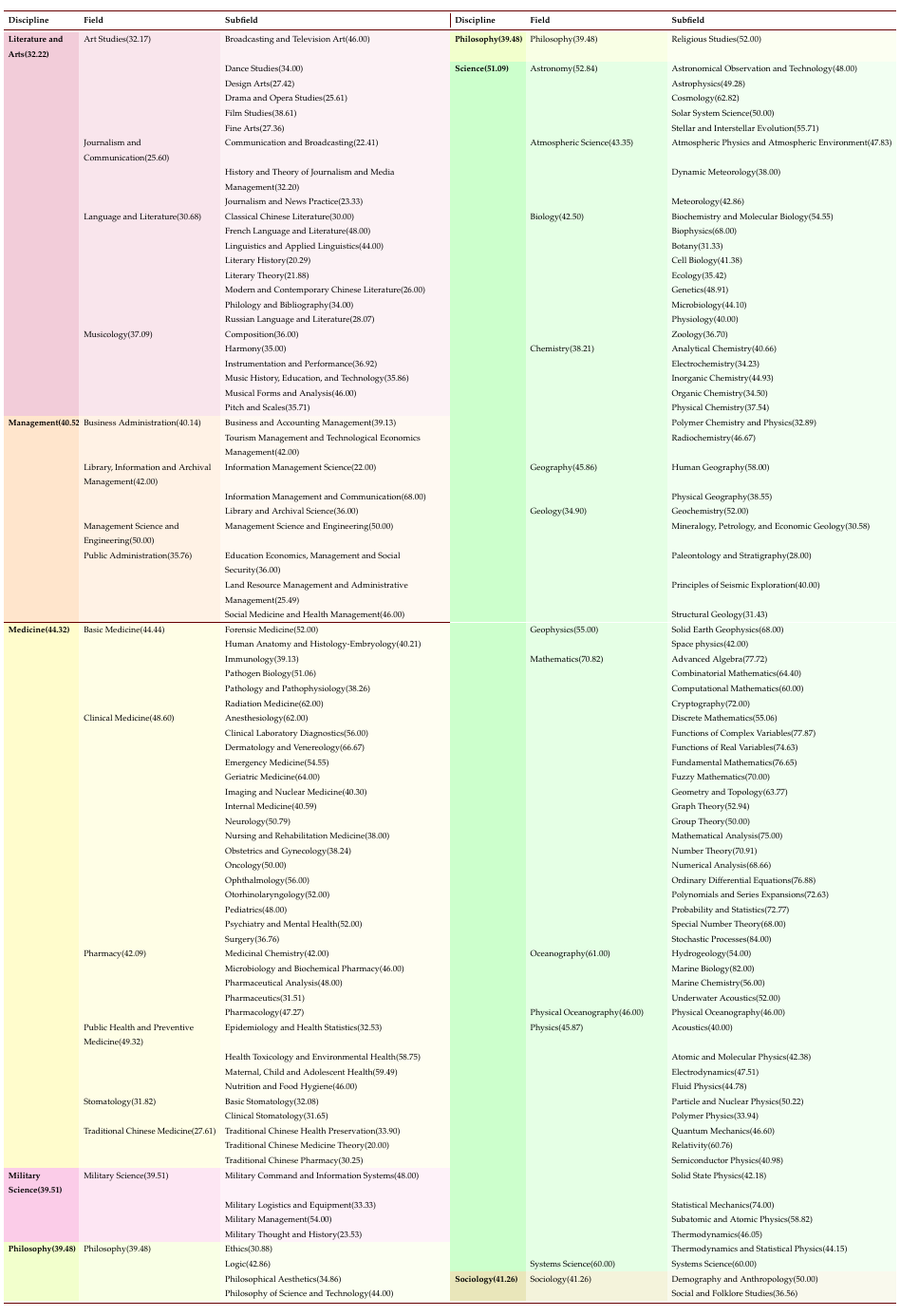} 
    \end{subtable}
    \vspace{-1.1cm}
    \captionsetup{font=small}
    \caption{Continued: Model Scores Across Three Levels of Disciplines: o1-mini-2024-09-12.}
    \vspace{-0.6cm}
    \centeredlinks{listofmodels}{Back to List of Models}{toc}{Back to Table of Contents}{blue}
    \end{table}
}
\clearpage

\newpage
\vspace{-0.5cm}
\begin{table}[t]
\refstepcounter{models}%
\addcontentsline{csf}{models}{\protect\numberline{\themodels}MiniMax-Text-01}
\centering
\begin{subtable}[t]{1\textwidth}
\centering
\includegraphics[width=\textwidth]{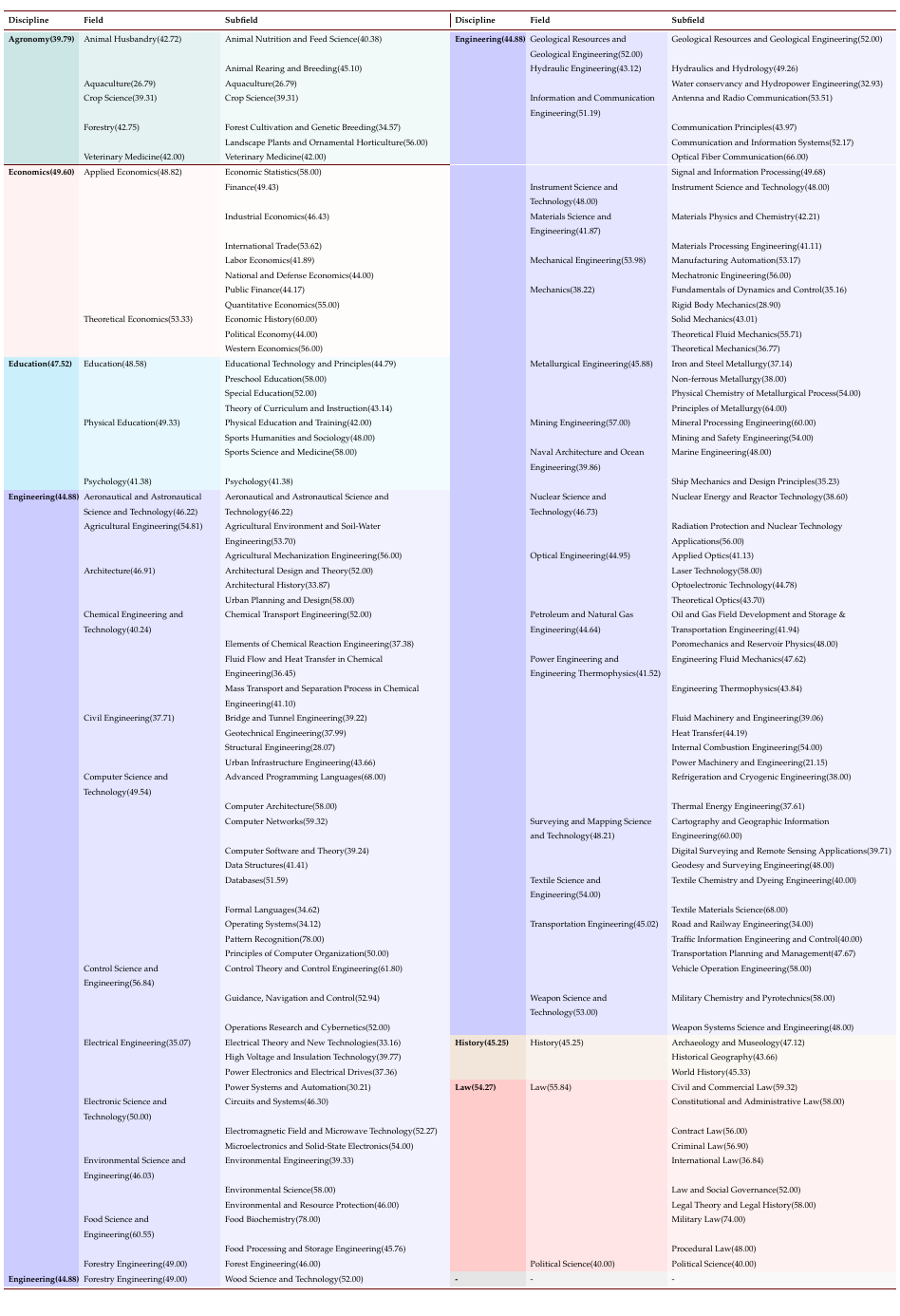} 
\end{subtable}
\vspace{-1.3cm}
\captionsetup{font=small}
\caption{Model Scores Across Three Levels of Disciplines: MiniMax-Text-01.}
\label{tab:MiniMax-Text-01}
\vspace{-0.5cm}
\centeredlinks{listofmodels}{Back to List of Models}{toc}{Back to Table of Contents}{blue}
\end{table}
\clearpage

\newpage
\afterpage{
    \begin{table}[t]
    \centering
    \ContinuedFloat 
    \begin{subtable}[t]{\textwidth}
    \centering
    \includegraphics[width=\textwidth]{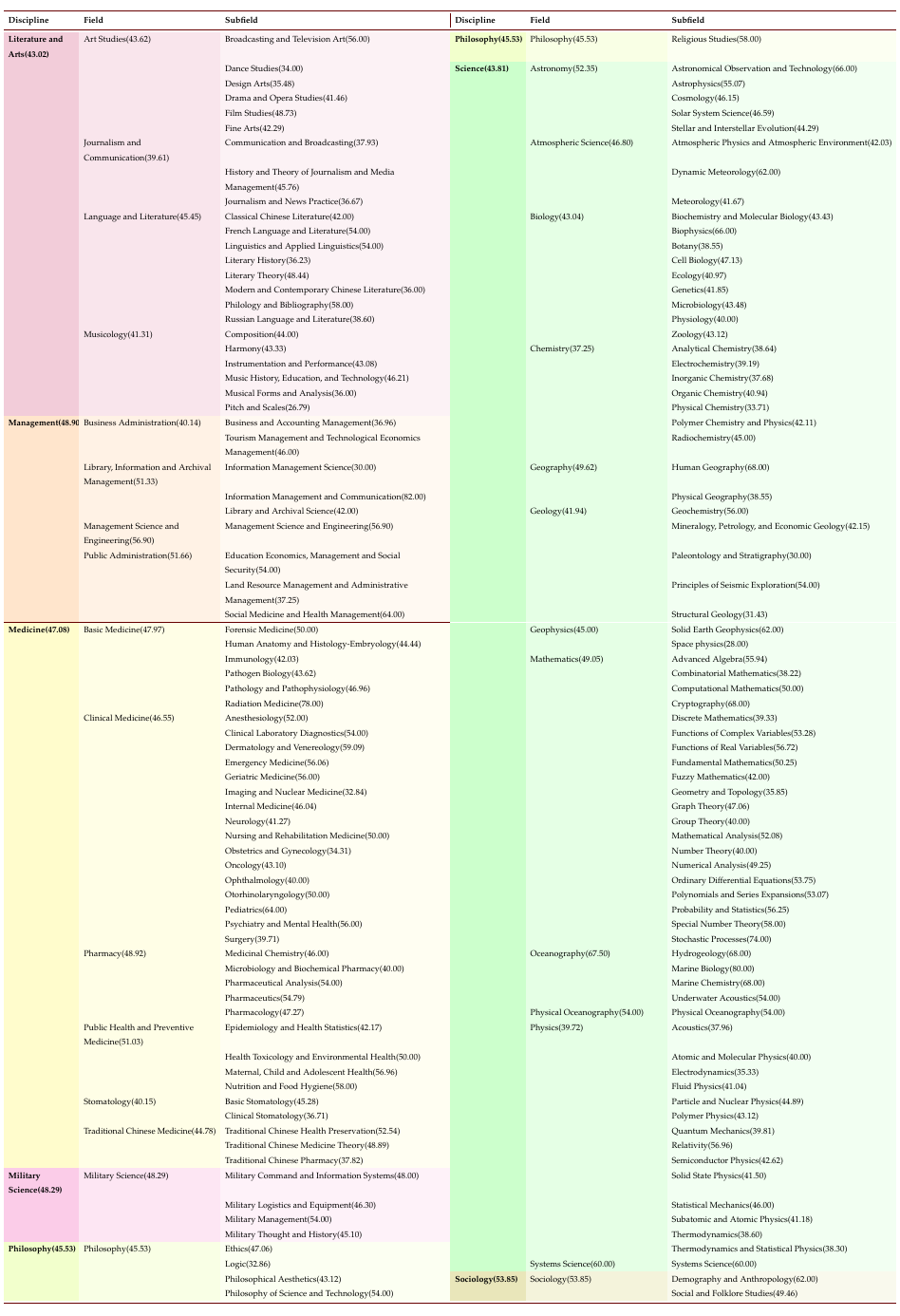} 
    \end{subtable}
    \vspace{-1.1cm}
    \captionsetup{font=small}
    \caption{Continued: Model Scores Across Three Levels of Disciplines: MiniMax-Text-01.}
    \vspace{-0.6cm}
    \centeredlinks{listofmodels}{Back to List of Models}{toc}{Back to Table of Contents}{blue}
    \end{table}
}
\clearpage

\newpage
\vspace{-0.5cm}
\begin{table}[t]
\refstepcounter{models}%
\addcontentsline{csf}{models}{\protect\numberline{\themodels}gpt-4o-2024-11-20}
\centering
\begin{subtable}[t]{1\textwidth}
\centering
\includegraphics[width=\textwidth]{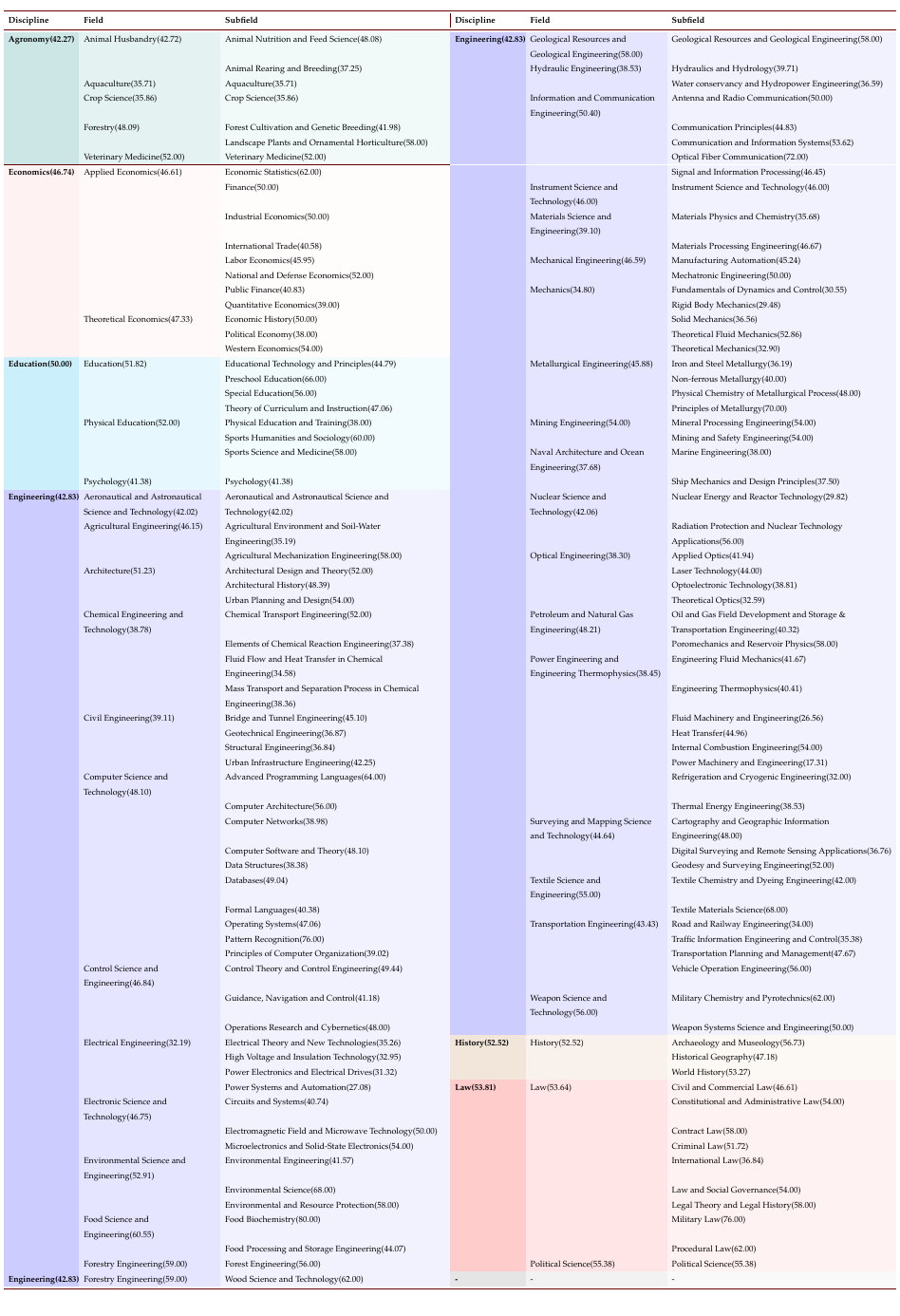} 
\end{subtable}
\vspace{-1.3cm}
\captionsetup{font=small}
\caption{Model Scores Across Three Levels of Disciplines: gpt-4o-2024-11-20.}
\label{tab:gpt-4o-2024-11-20}
\vspace{-0.5cm}
\centeredlinks{listofmodels}{Back to List of Models}{toc}{Back to Table of Contents}{blue}
\end{table}
\clearpage

\newpage
\afterpage{
    \begin{table}[t]
    \centering
    \ContinuedFloat 
    \begin{subtable}[t]{\textwidth}
    \centering
    \includegraphics[width=\textwidth]{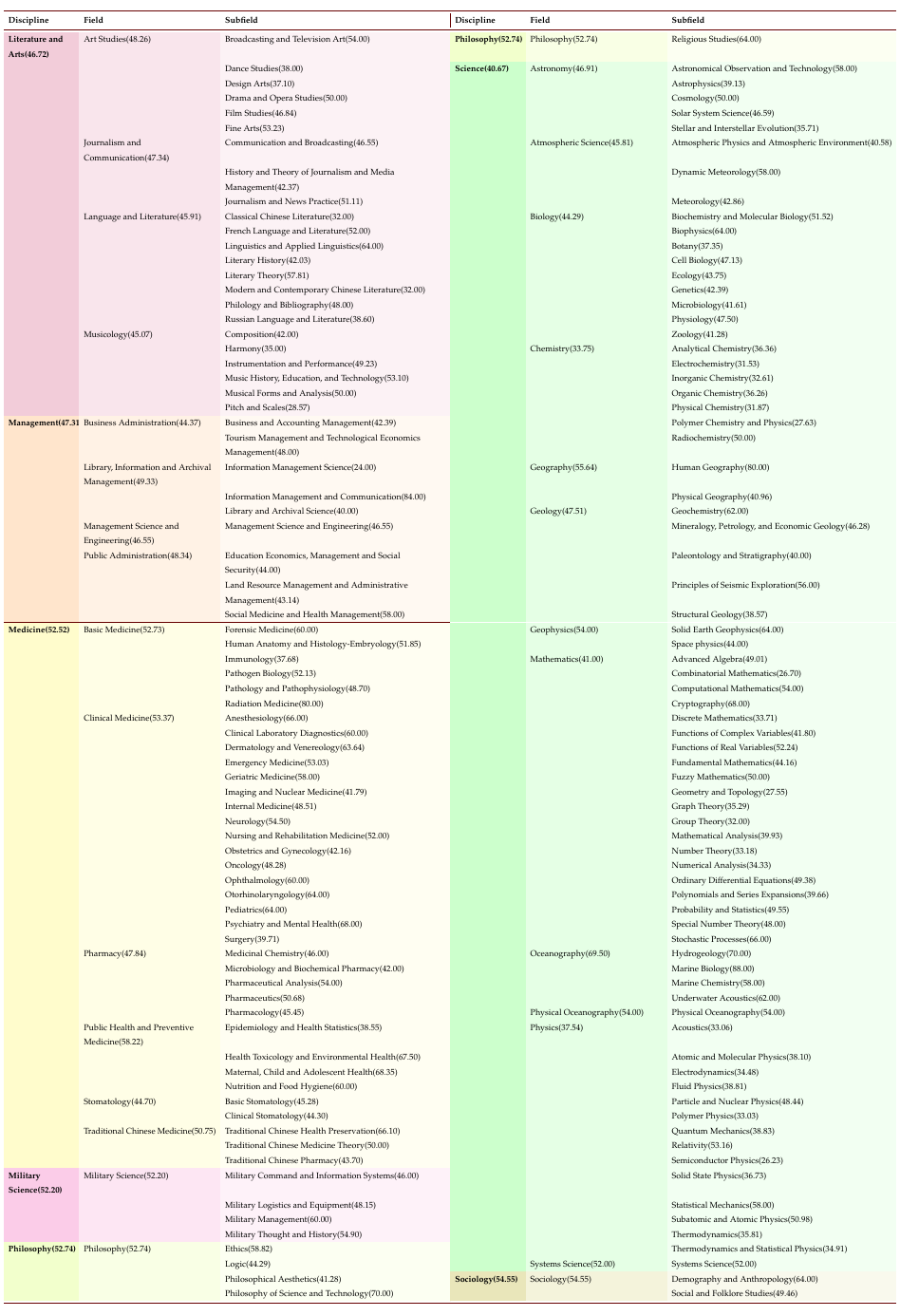} 
    \end{subtable}
    \vspace{-1.1cm}
    \captionsetup{font=small}
    \caption{Continued: Model Scores Across Three Levels of Disciplines: gpt-4o-2024-11-20.}
    \vspace{-0.6cm}
    \centeredlinks{listofmodels}{Back to List of Models}{toc}{Back to Table of Contents}{blue}
    \end{table}
}
\clearpage

\newpage
\vspace{-0.5cm}
\begin{table}[t]
\refstepcounter{models}%
\addcontentsline{csf}{models}{\protect\numberline{\themodels}QwQ}
\centering
\begin{subtable}[t]{1\textwidth}
\centering
\includegraphics[width=\textwidth]{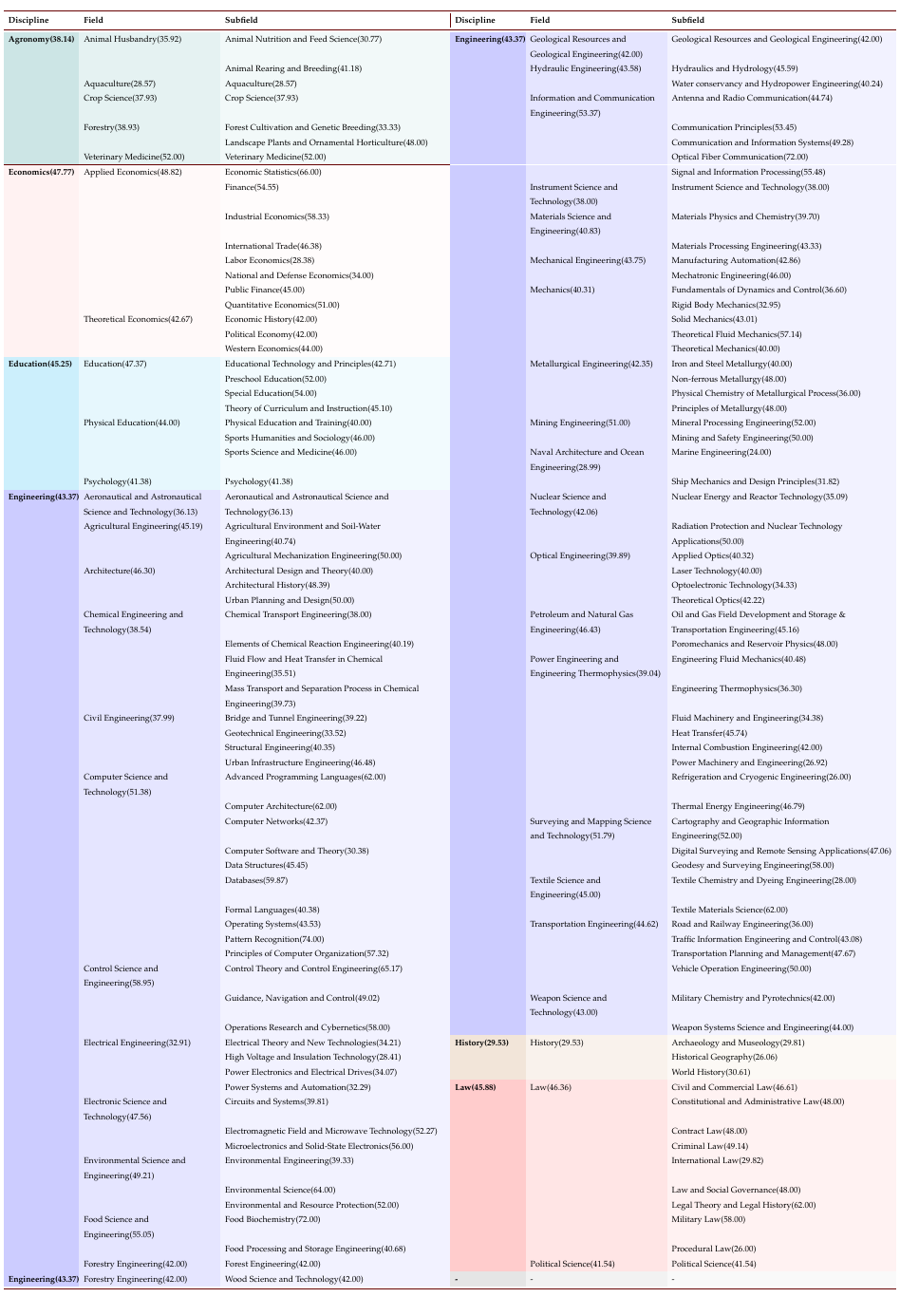} 
\end{subtable}
\vspace{-1.3cm}
\captionsetup{font=small}
\caption{Model Scores Across Three Levels of Disciplines: QwQ.}
\label{tab:QwQ}
\vspace{-0.5cm}
\centeredlinks{listofmodels}{Back to List of Models}{toc}{Back to Table of Contents}{blue}
\end{table}
\clearpage

\newpage
\afterpage{
    \begin{table}[t]
    \centering
    \ContinuedFloat 
    \begin{subtable}[t]{\textwidth}
    \centering
    \includegraphics[width=\textwidth]{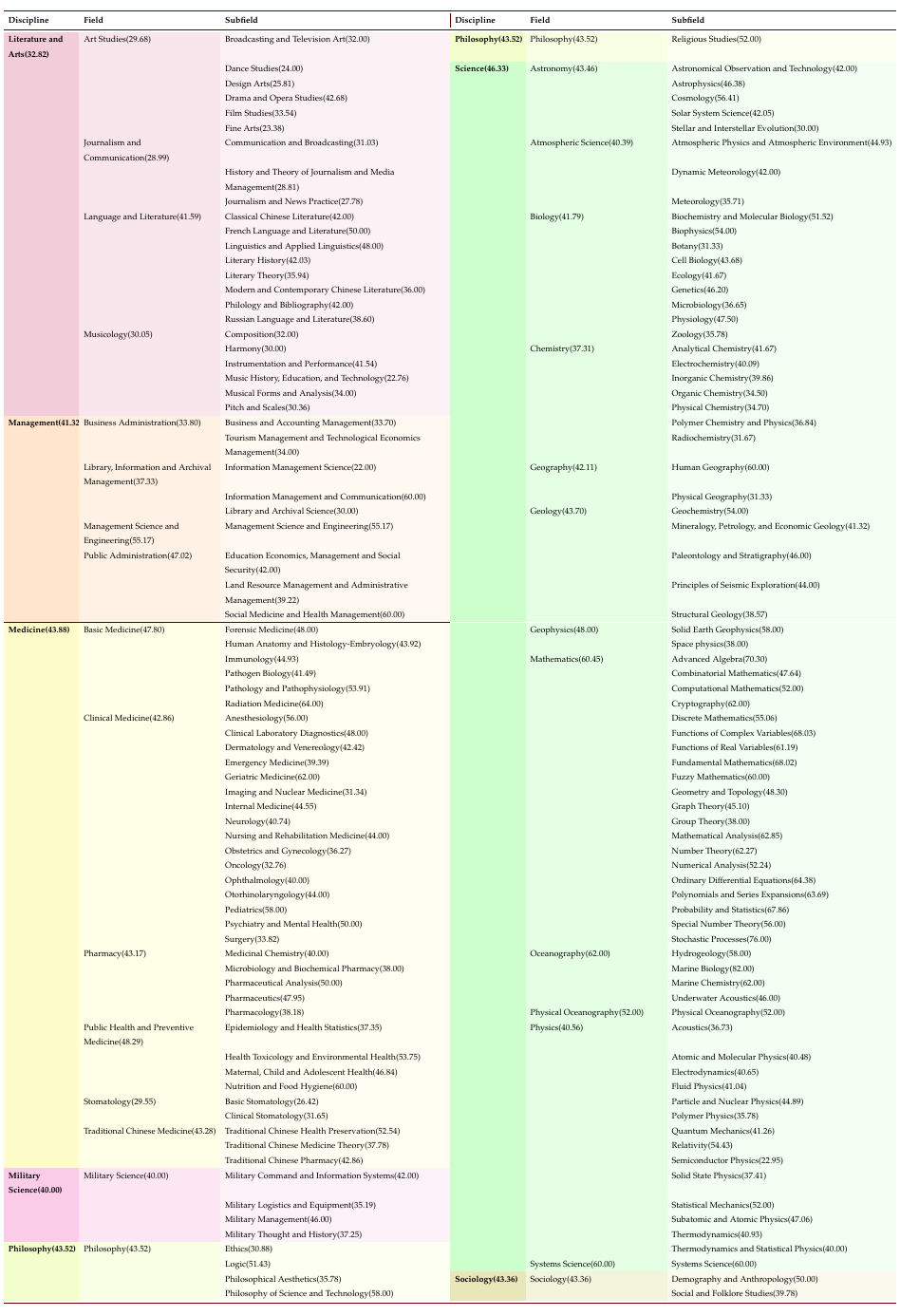} 
    \end{subtable}
    \vspace{-1.1cm}
    \captionsetup{font=small}
    \caption{Continued: Model Scores Across Three Levels of Disciplines: QwQ.}
    \vspace{-0.6cm}
    \centeredlinks{listofmodels}{Back to List of Models}{toc}{Back to Table of Contents}{blue}
    \end{table}
}
\clearpage

\newpage
\vspace{-0.5cm}
\begin{table}[t]
\refstepcounter{models}%
\addcontentsline{csf}{models}{\protect\numberline{\themodels}Llama-3.1-405B-Instruct}
\centering
\begin{subtable}[t]{1\textwidth}
\centering
\includegraphics[width=\textwidth]{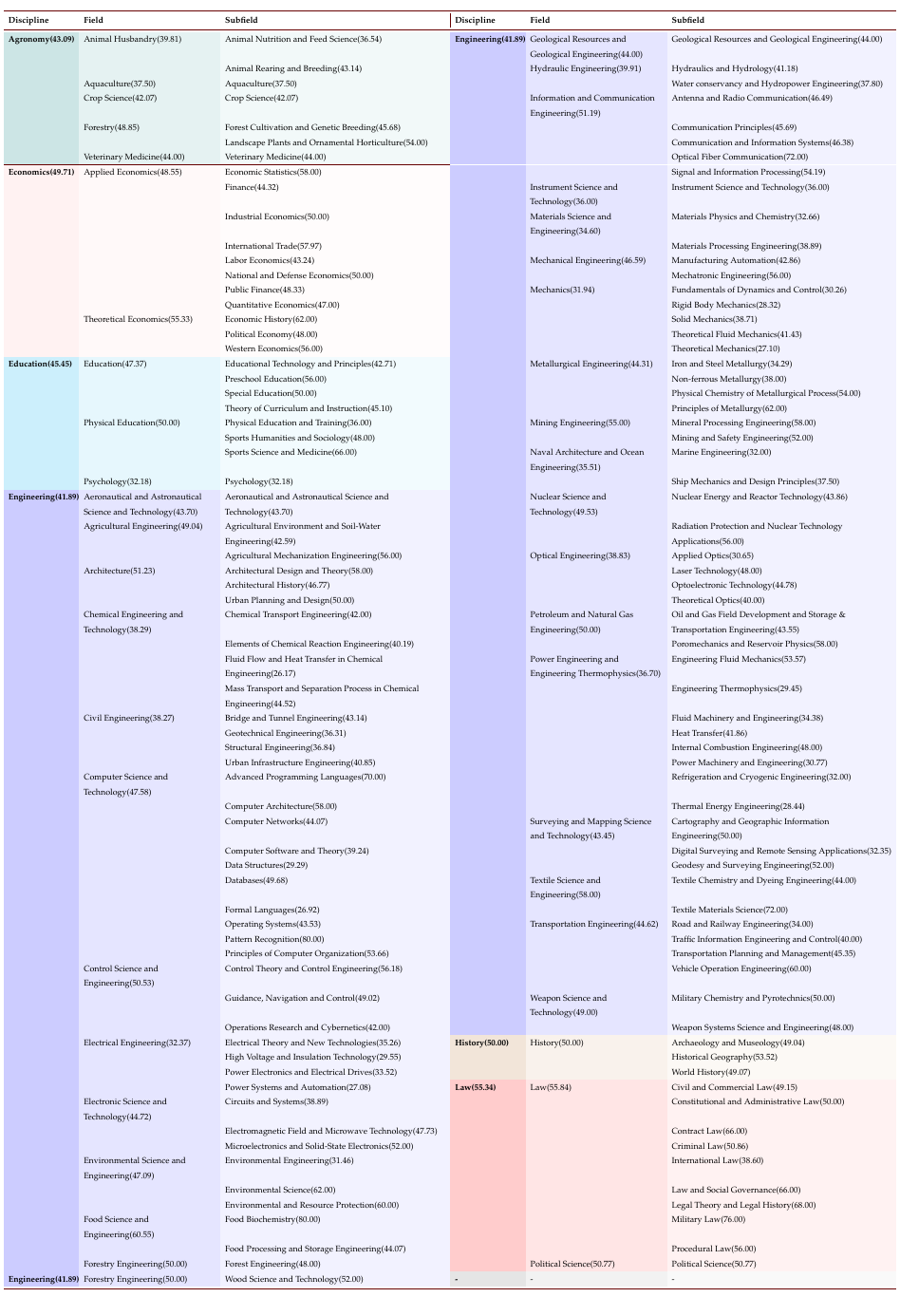} 
\end{subtable}
\vspace{-1.3cm}
\captionsetup{font=small}
\caption{Model Scores Across Three Levels of Disciplines: Llama-3.1-405B-Instruct.}
\label{tab:Llama-3.1-405B-Instruct}
\vspace{-0.5cm}
\centeredlinks{listofmodels}{Back to List of Models}{toc}{Back to Table of Contents}{blue}
\end{table}
\clearpage

\newpage
\afterpage{
    \begin{table}[t]
    \centering
    \ContinuedFloat 
    \begin{subtable}[t]{\textwidth}
    \centering
    \includegraphics[width=\textwidth]{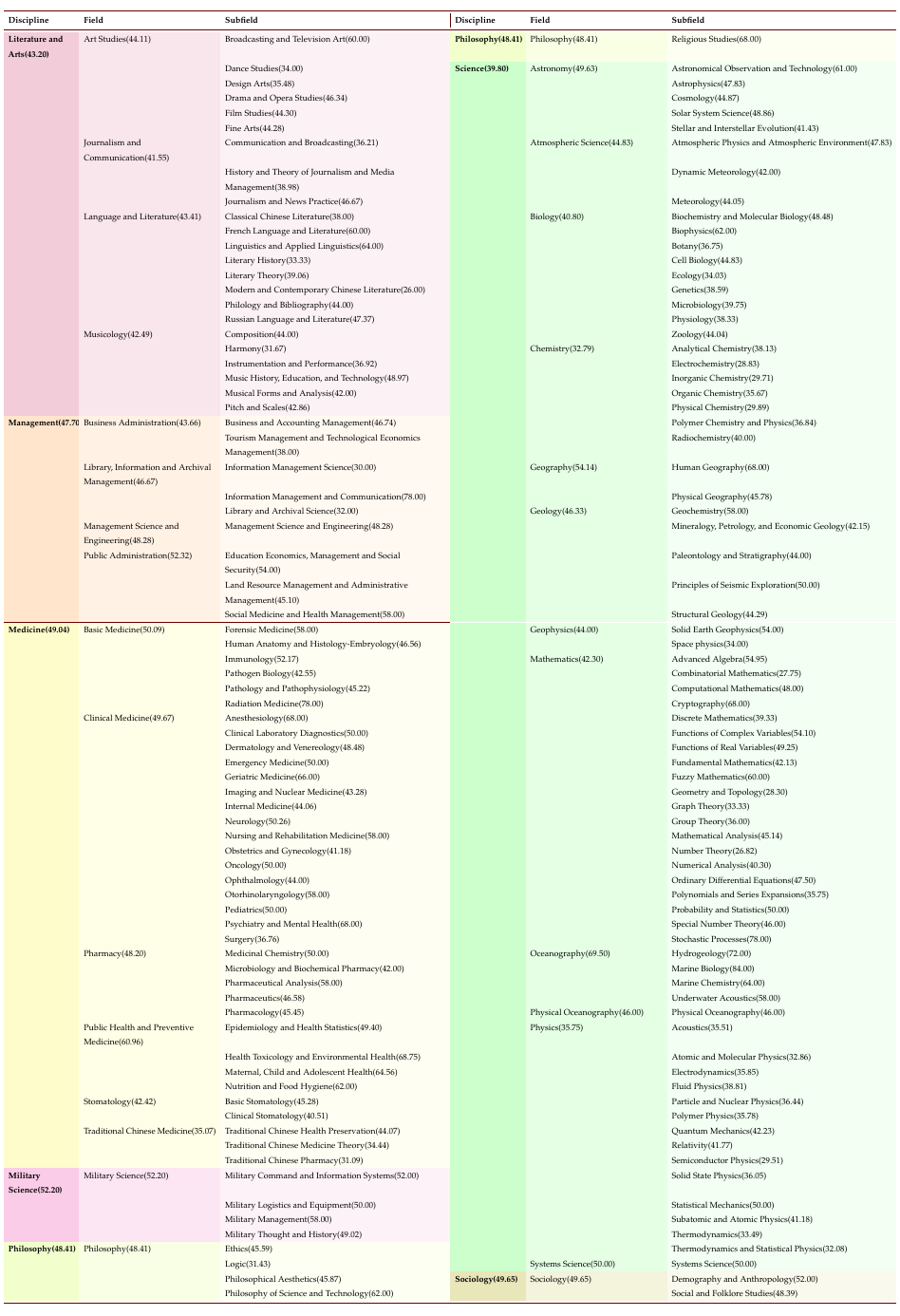} 
    \end{subtable}
    \vspace{-1.1cm}
    \captionsetup{font=small}
    \caption{Continued: Model Scores Across Three Levels of Disciplines: Llama-3.1-405B-Instruct.}
    \vspace{-0.6cm}
    \centeredlinks{listofmodels}{Back to List of Models}{toc}{Back to Table of Contents}{blue}
    \end{table}
}
\clearpage

\newpage
\vspace{-0.5cm}
\begin{table}[t]
\refstepcounter{models}%
\addcontentsline{csf}{models}{\protect\numberline{\themodels}gpt-4o-2024-08-06}
\centering
\begin{subtable}[t]{1\textwidth}
\centering
\includegraphics[width=\textwidth]{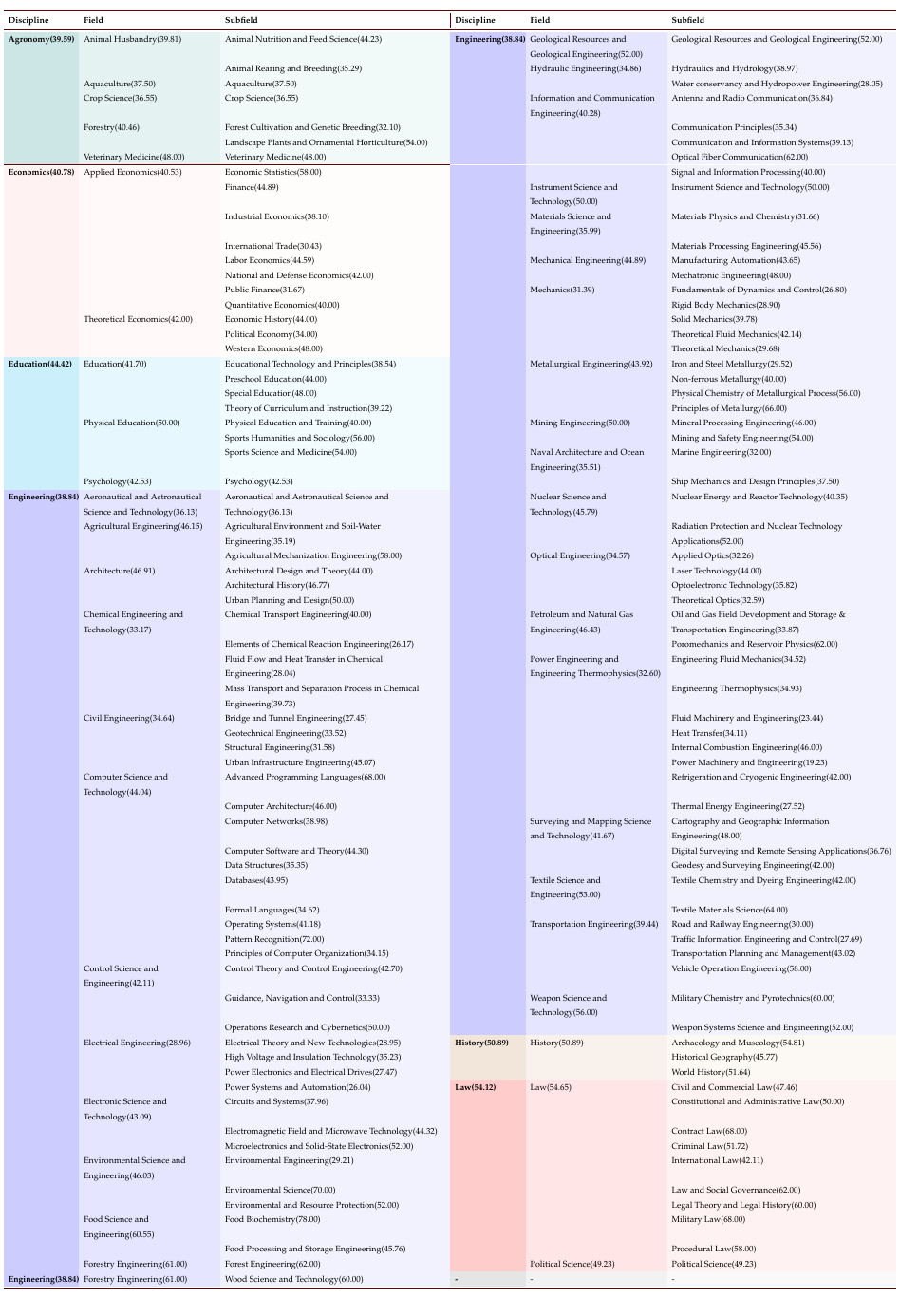} 
\end{subtable}
\vspace{-1.3cm}
\captionsetup{font=small}
\caption{Model Scores Across Three Levels of Disciplines: gpt-4o-2024-08-06.}
\label{tab:gpt-4o-2024-08-06}
\vspace{-0.5cm}
\centeredlinks{listofmodels}{Back to List of Models}{toc}{Back to Table of Contents}{blue}
\end{table}
\clearpage

\newpage
\afterpage{
    \begin{table}[t]
    \centering
    \ContinuedFloat 
    \begin{subtable}[t]{\textwidth}
    \centering
    \includegraphics[width=\textwidth]{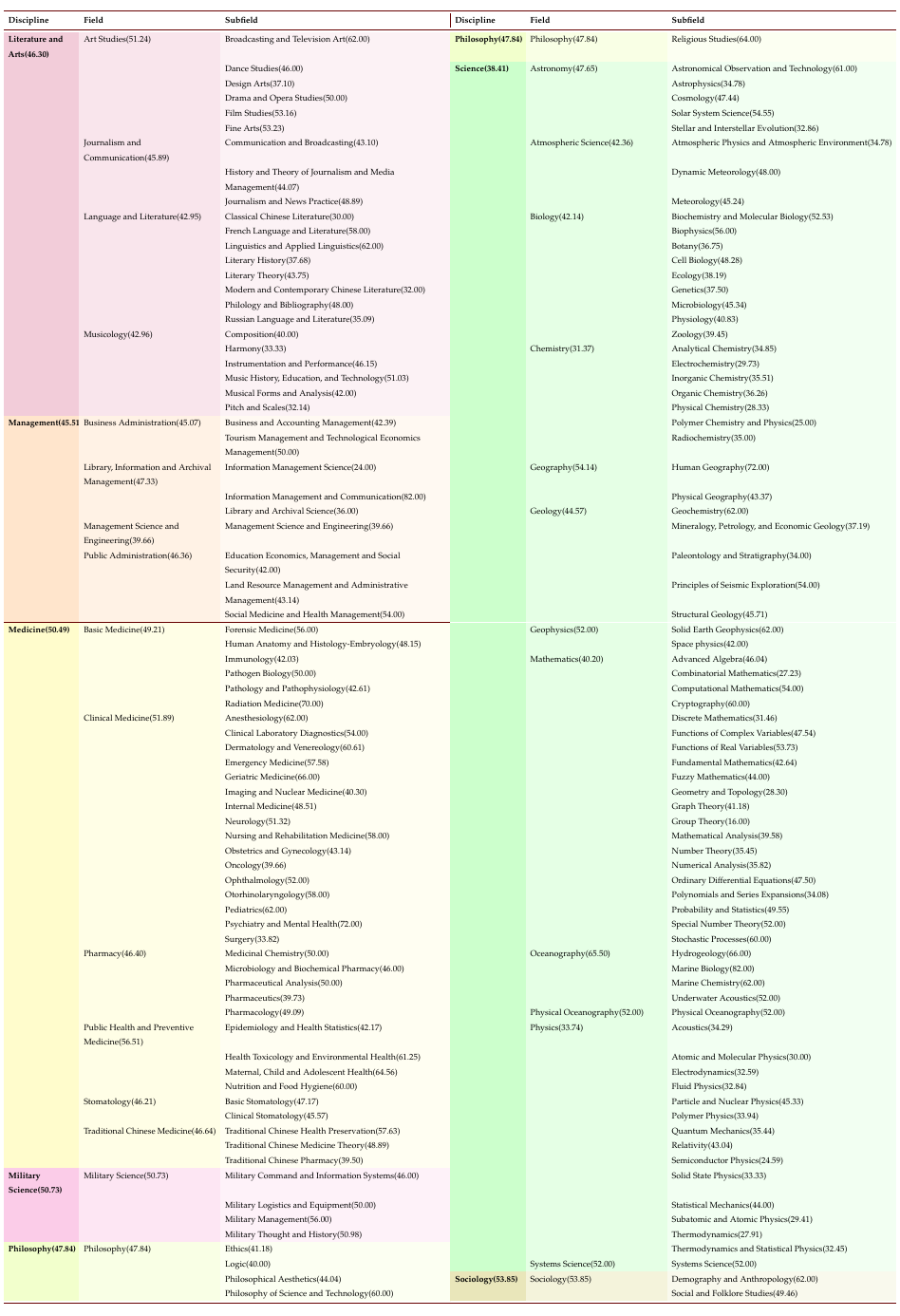} 
    \end{subtable}
    \vspace{-1.1cm}
    \captionsetup{font=small}
    \caption{Continued: Model Scores Across Three Levels of Disciplines: gpt-4o-2024-08-06.}
    \vspace{-0.6cm}
    \centeredlinks{listofmodels}{Back to List of Models}{toc}{Back to Table of Contents}{blue}
    \end{table}
}
\clearpage

\newpage
\vspace{-0.5cm}
\begin{table}[t]
\refstepcounter{models}%
\addcontentsline{csf}{models}{\protect\numberline{\themodels}Qwen2.5-72B-Instruct}
\centering
\begin{subtable}[t]{1\textwidth}
\centering
\includegraphics[width=\textwidth]{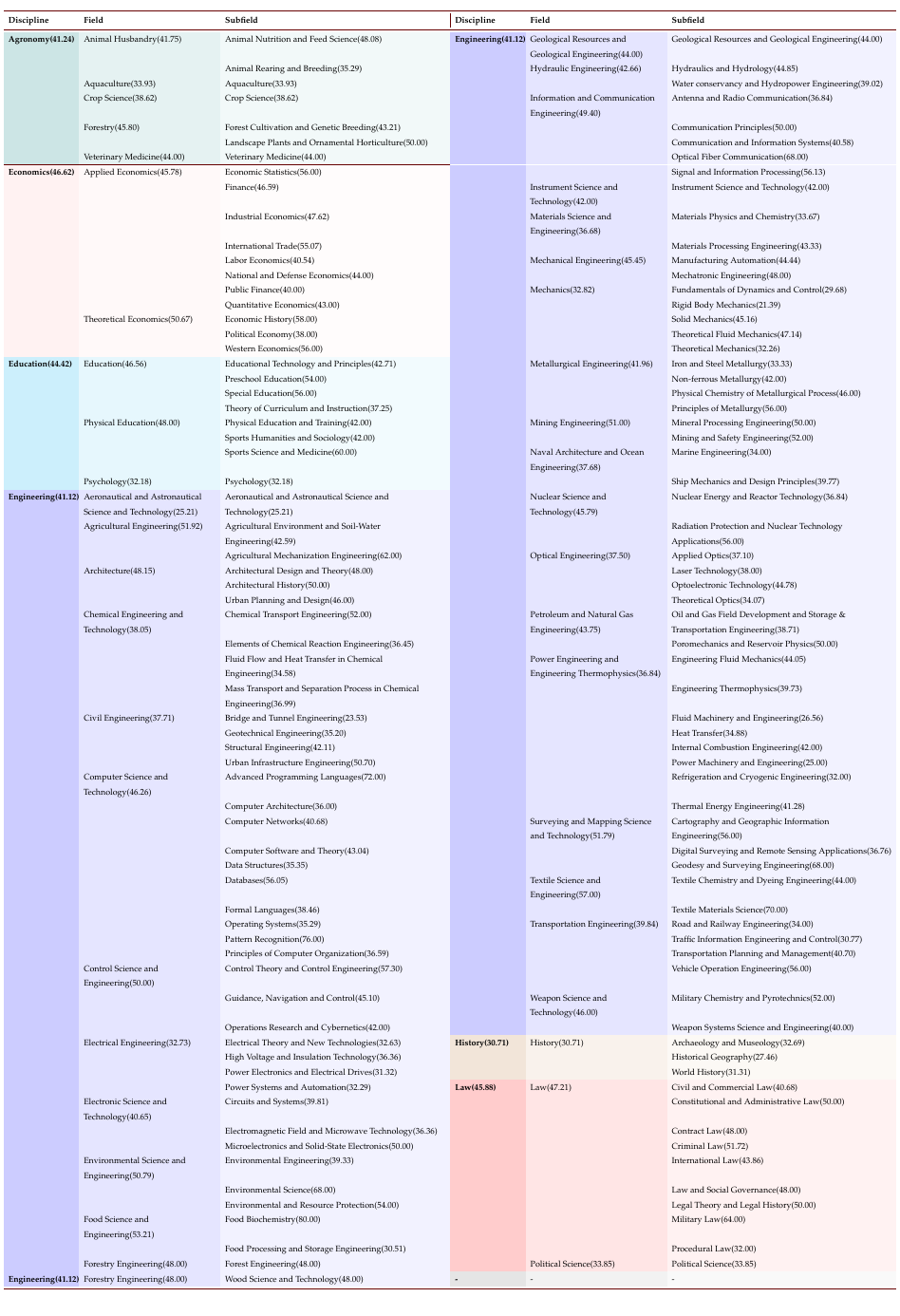} 
\end{subtable}
\vspace{-1.3cm}
\captionsetup{font=small}
\caption{Model Scores Across Three Levels of Disciplines: Qwen2.5-72B-Instruct.}
\label{tab:Qwen2.5-72B-Instruct}
\vspace{-0.5cm}
\centeredlinks{listofmodels}{Back to List of Models}{toc}{Back to Table of Contents}{blue}
\end{table}
\clearpage

\newpage
\afterpage{
    \begin{table}[t]
    \centering
    \ContinuedFloat 
    \begin{subtable}[t]{\textwidth}
    \centering
    \includegraphics[width=\textwidth]{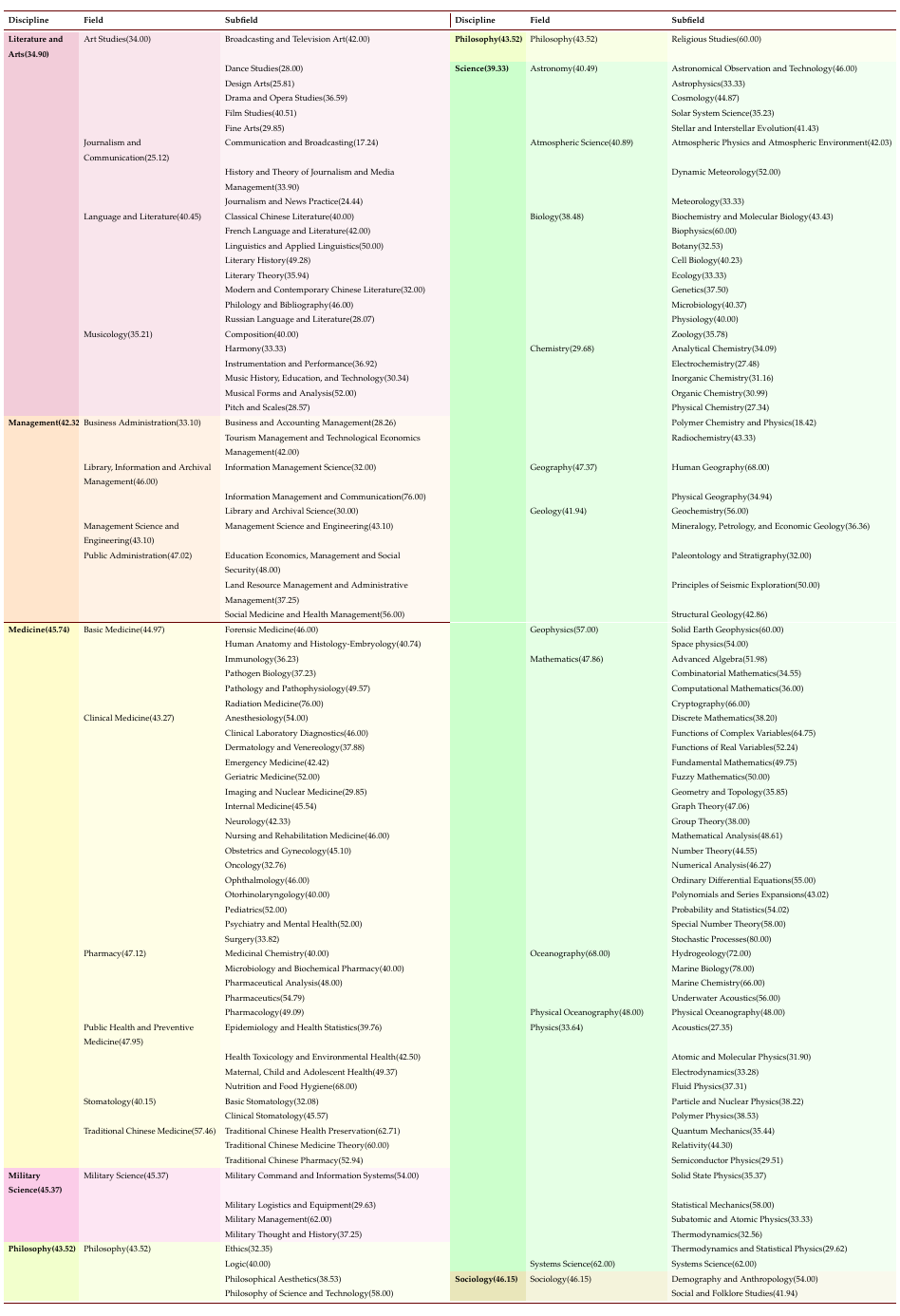} 
    \end{subtable}
    \vspace{-1.1cm}
    \captionsetup{font=small}
    \caption{Continued: Model Scores Across Three Levels of Disciplines: Qwen2.5-72B-Instruct.}
    \vspace{-0.6cm}
    \centeredlinks{listofmodels}{Back to List of Models}{toc}{Back to Table of Contents}{blue}
    \end{table}
}
\clearpage

\newpage
\vspace{-0.5cm}
\begin{table}[t]
\refstepcounter{models}%
\addcontentsline{csf}{models}{\protect\numberline{\themodels}Mistral-Large-Instruct-2411}
\centering
\begin{subtable}[t]{1\textwidth}
\centering
\includegraphics[width=\textwidth]{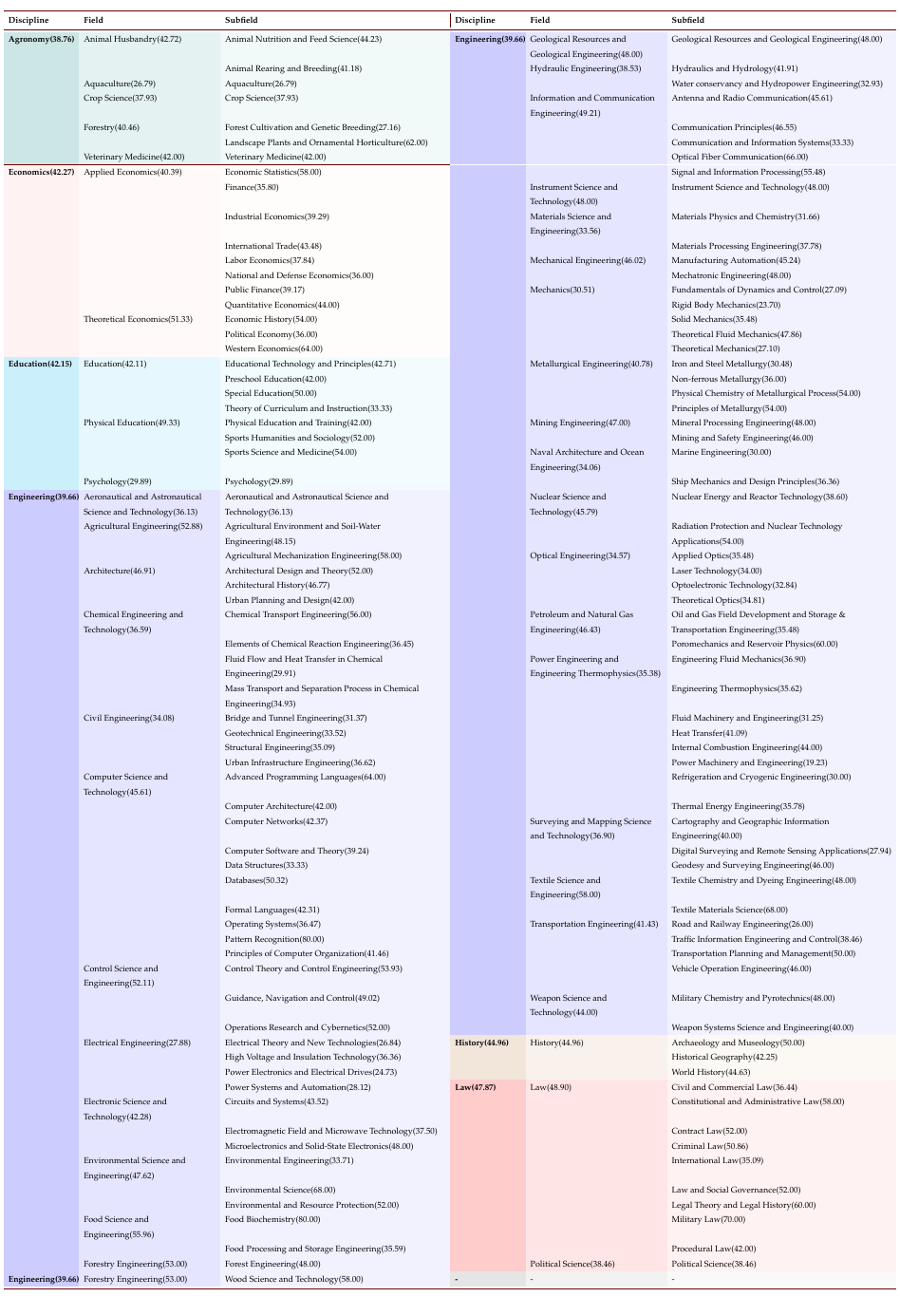} 
\end{subtable}
\vspace{-1.3cm}
\captionsetup{font=small}
\caption{Model Scores Across Three Levels of Disciplines: Mistral-Large-Instruct-2411.}
\label{tab:Mistral-Large-Instruct-2411}
\vspace{-0.5cm}
\centeredlinks{listofmodels}{Back to List of Models}{toc}{Back to Table of Contents}{blue}
\end{table}
\clearpage

\newpage
\afterpage{
    \begin{table}[t]
    \centering
    \ContinuedFloat 
    \begin{subtable}[t]{\textwidth}
    \centering
    \includegraphics[width=\textwidth]{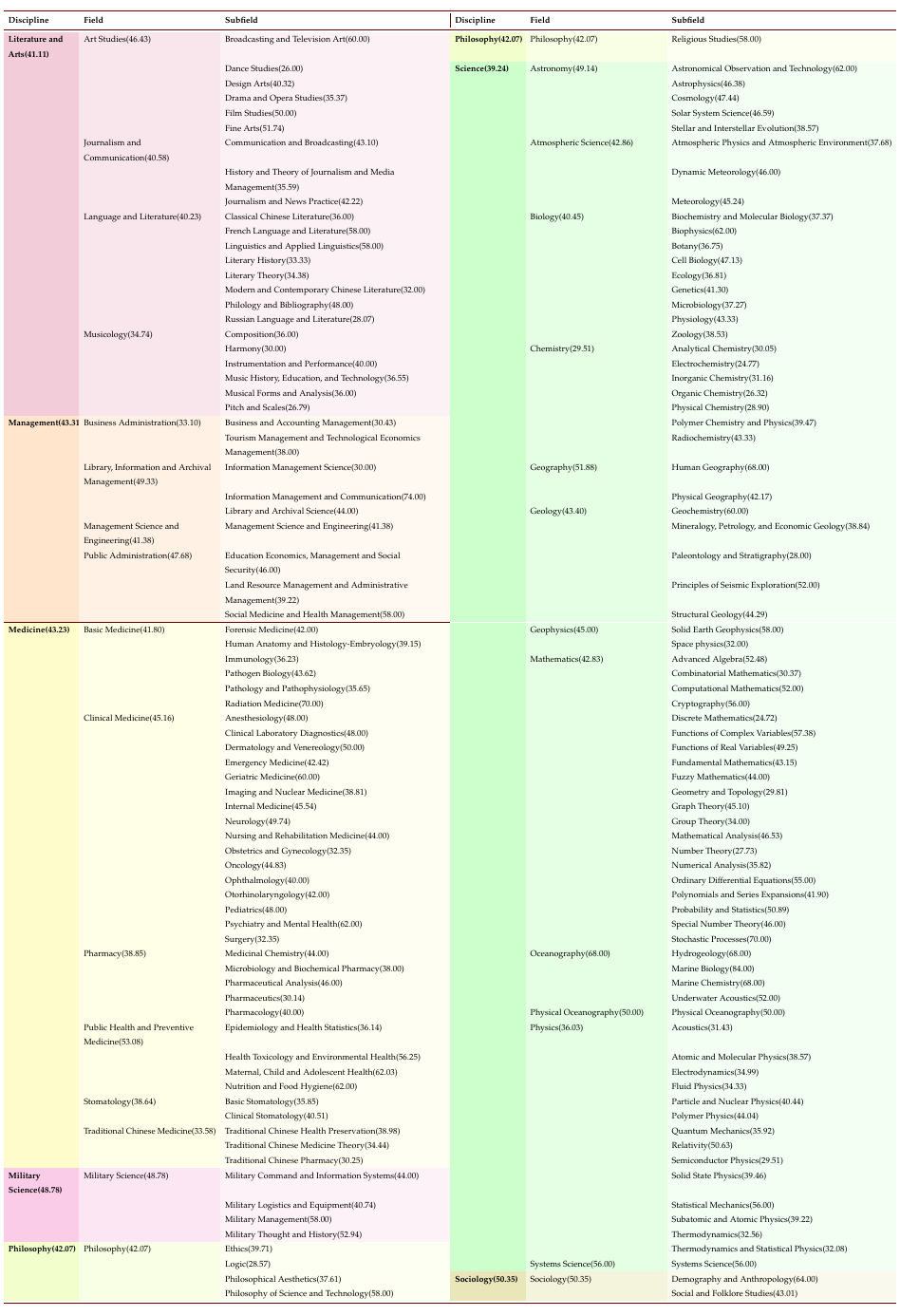} 
    \end{subtable}
    \vspace{-1.1cm}
    \captionsetup{font=small}
    \caption{Continued: Model Scores Across Three Levels of Disciplines: Mistral-Large-Instruct-2411.}
    \vspace{-0.6cm}
    \centeredlinks{listofmodels}{Back to List of Models}{toc}{Back to Table of Contents}{blue}
    \end{table}
}
\clearpage

\newpage
\vspace{-0.5cm}
\begin{table}[t]
\refstepcounter{models}%
\addcontentsline{csf}{models}{\protect\numberline{\themodels}qwen-max-2024-09-19}
\centering
\begin{subtable}[t]{1\textwidth}
\centering
\includegraphics[width=\textwidth]{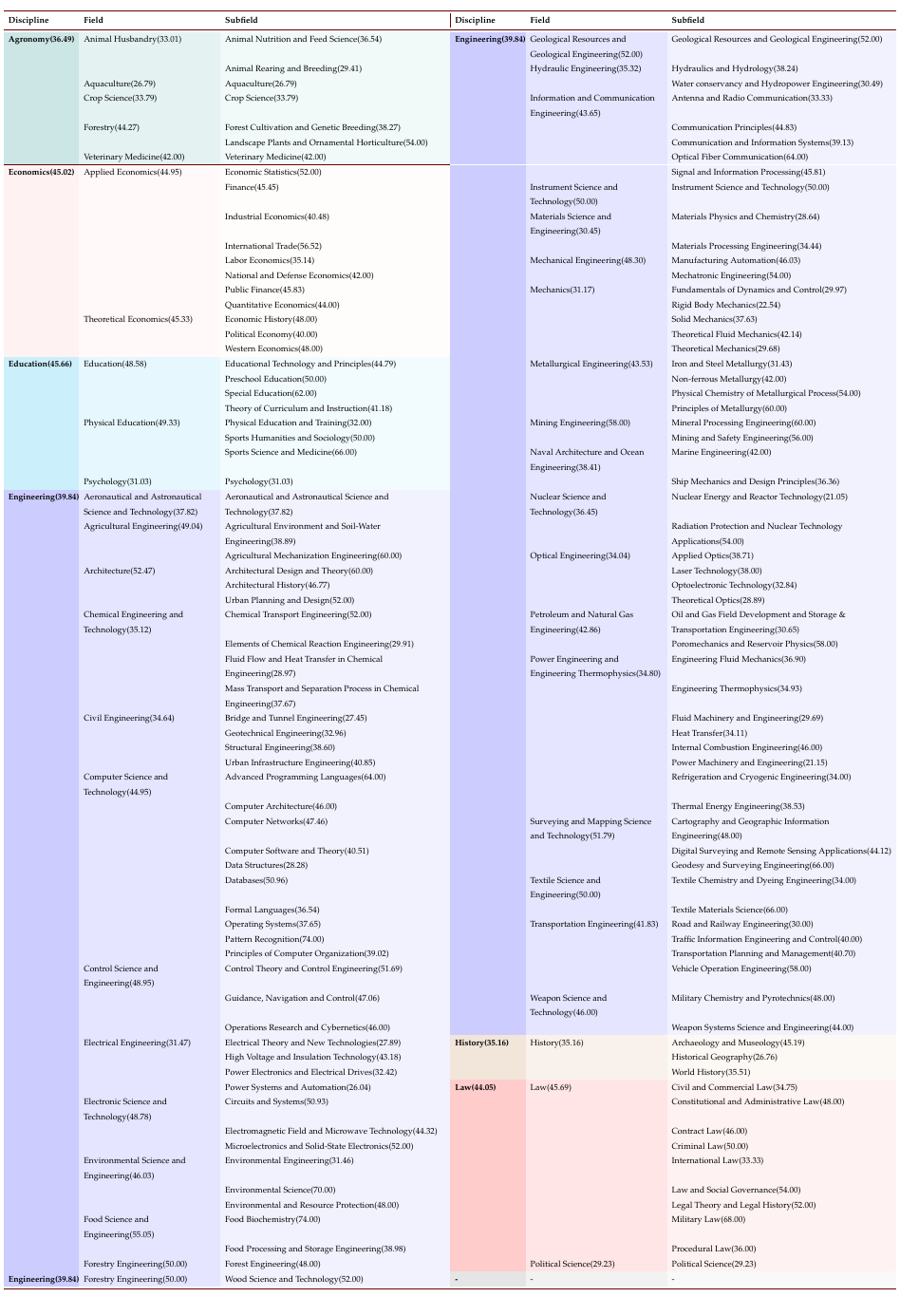} 
\end{subtable}
\vspace{-1.3cm}
\captionsetup{font=small}
\caption{Model Scores Across Three Levels of Disciplines: qwen-max-2024-09-19.}
\label{tab:qwen-max-2024-09-19}
\vspace{-0.5cm}
\centeredlinks{listofmodels}{Back to List of Models}{toc}{Back to Table of Contents}{blue}
\end{table}
\clearpage

\newpage
\afterpage{
    \begin{table}[t]
    \centering
    \ContinuedFloat 
    \begin{subtable}[t]{\textwidth}
    \centering
    \includegraphics[width=\textwidth]{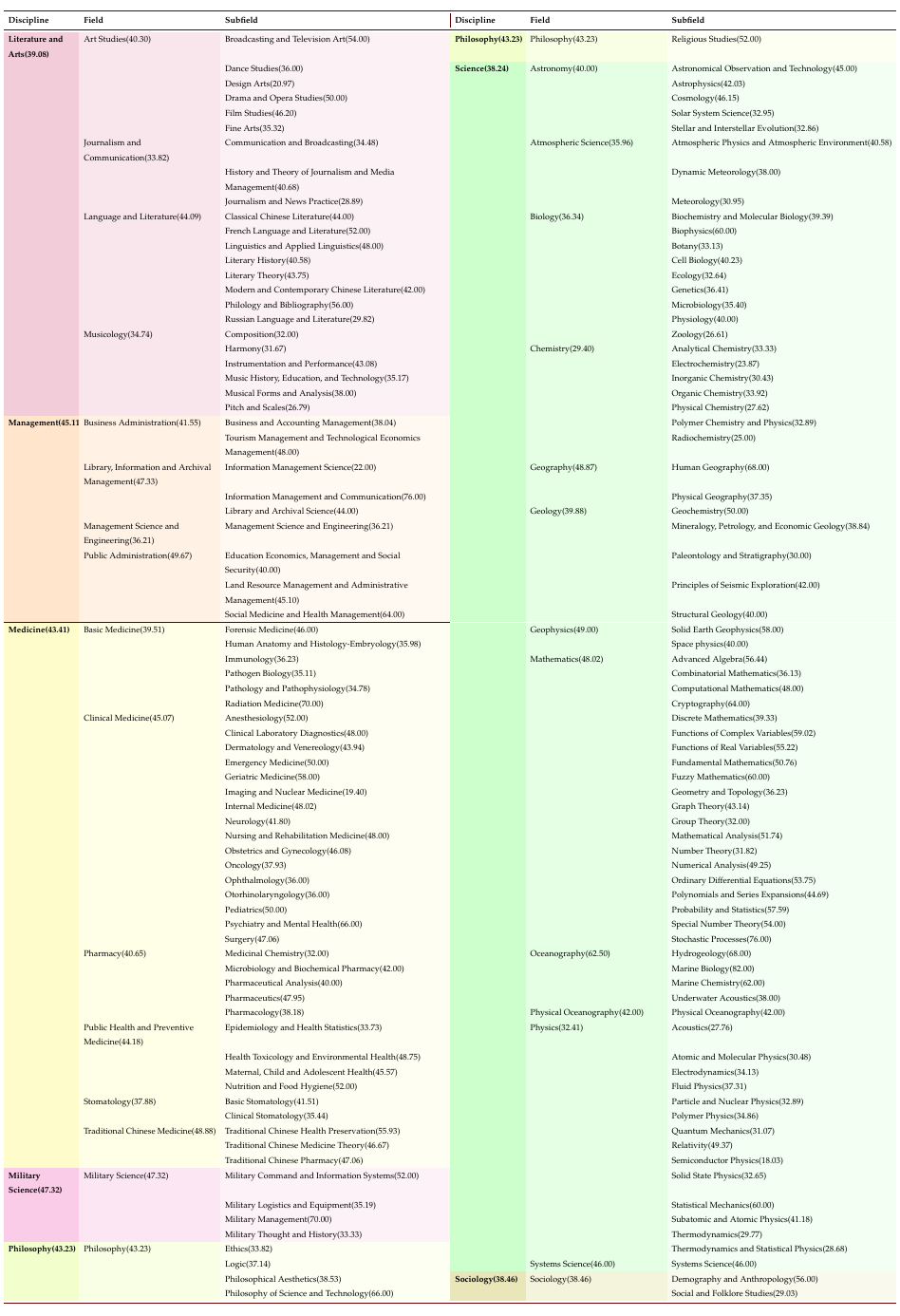} 
    \end{subtable}
    \vspace{-1.1cm}
    \captionsetup{font=small}
    \caption{Continued: Model Scores Across Three Levels of Disciplines: qwen-max-2024-09-19.}
    \vspace{-0.6cm}
    \centeredlinks{listofmodels}{Back to List of Models}{toc}{Back to Table of Contents}{blue}
    \end{table}
}
\clearpage

\newpage
\vspace{-0.5cm}
\begin{table}[t]
\refstepcounter{models}%
\addcontentsline{csf}{models}{\protect\numberline{\themodels}gpt-4o-2024-05-13}
\centering
\begin{subtable}[t]{1\textwidth}
\centering
\includegraphics[width=\textwidth]{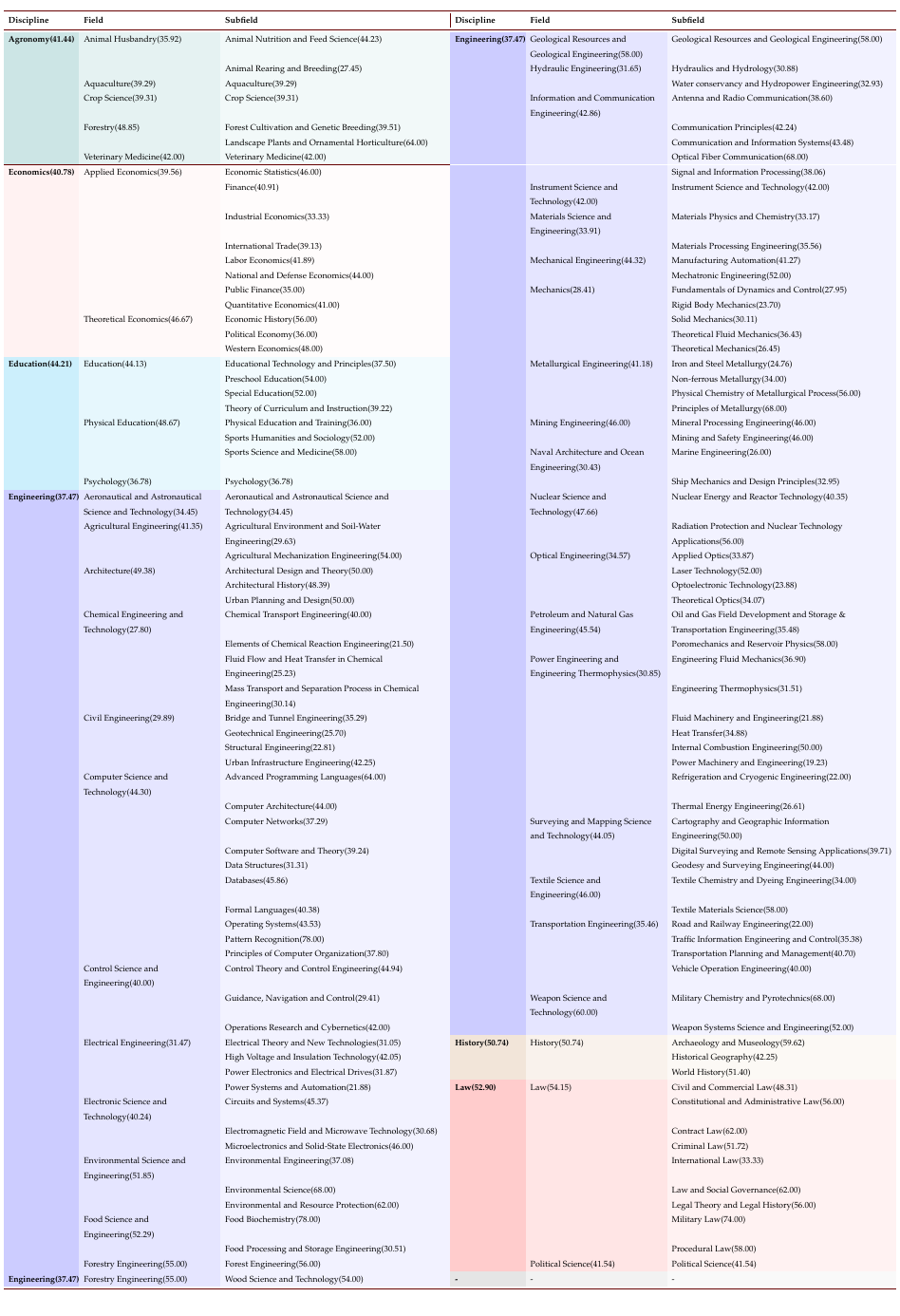} 
\end{subtable}
\vspace{-1.3cm}
\captionsetup{font=small}
\caption{Model Scores Across Three Levels of Disciplines: gpt-4o-2024-05-13.}
\label{tab:gpt-4o-2024-05-13}
\vspace{-0.5cm}
\centeredlinks{listofmodels}{Back to List of Models}{toc}{Back to Table of Contents}{blue}
\end{table}
\clearpage

\newpage
\afterpage{
    \begin{table}[t]
    \centering
    \ContinuedFloat 
    \begin{subtable}[t]{\textwidth}
    \centering
    \includegraphics[width=\textwidth]{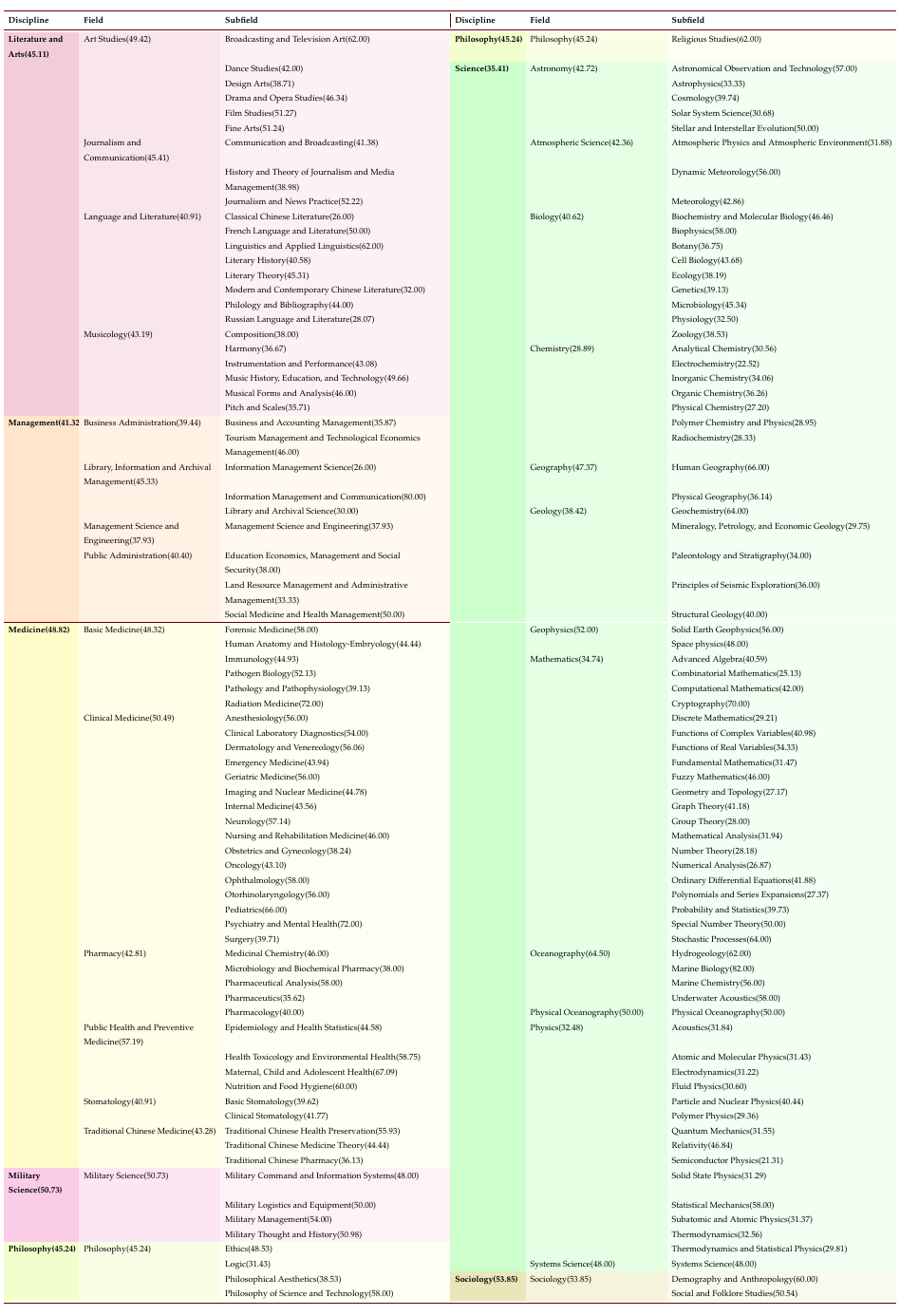} 
    \end{subtable}
    \vspace{-1.1cm}
    \captionsetup{font=small}
    \caption{Continued: Model Scores Across Three Levels of Disciplines: gpt-4o-2024-05-13.}
    \vspace{-0.6cm}
    \centeredlinks{listofmodels}{Back to List of Models}{toc}{Back to Table of Contents}{blue}
    \end{table}
}
\clearpage

\newpage
\vspace{-0.5cm}
\begin{table}[t]
\refstepcounter{models}%
\addcontentsline{csf}{models}{\protect\numberline{\themodels}Qwen2.5-32B-Instruct}
\centering
\begin{subtable}[t]{1\textwidth}
\centering
\includegraphics[width=\textwidth]{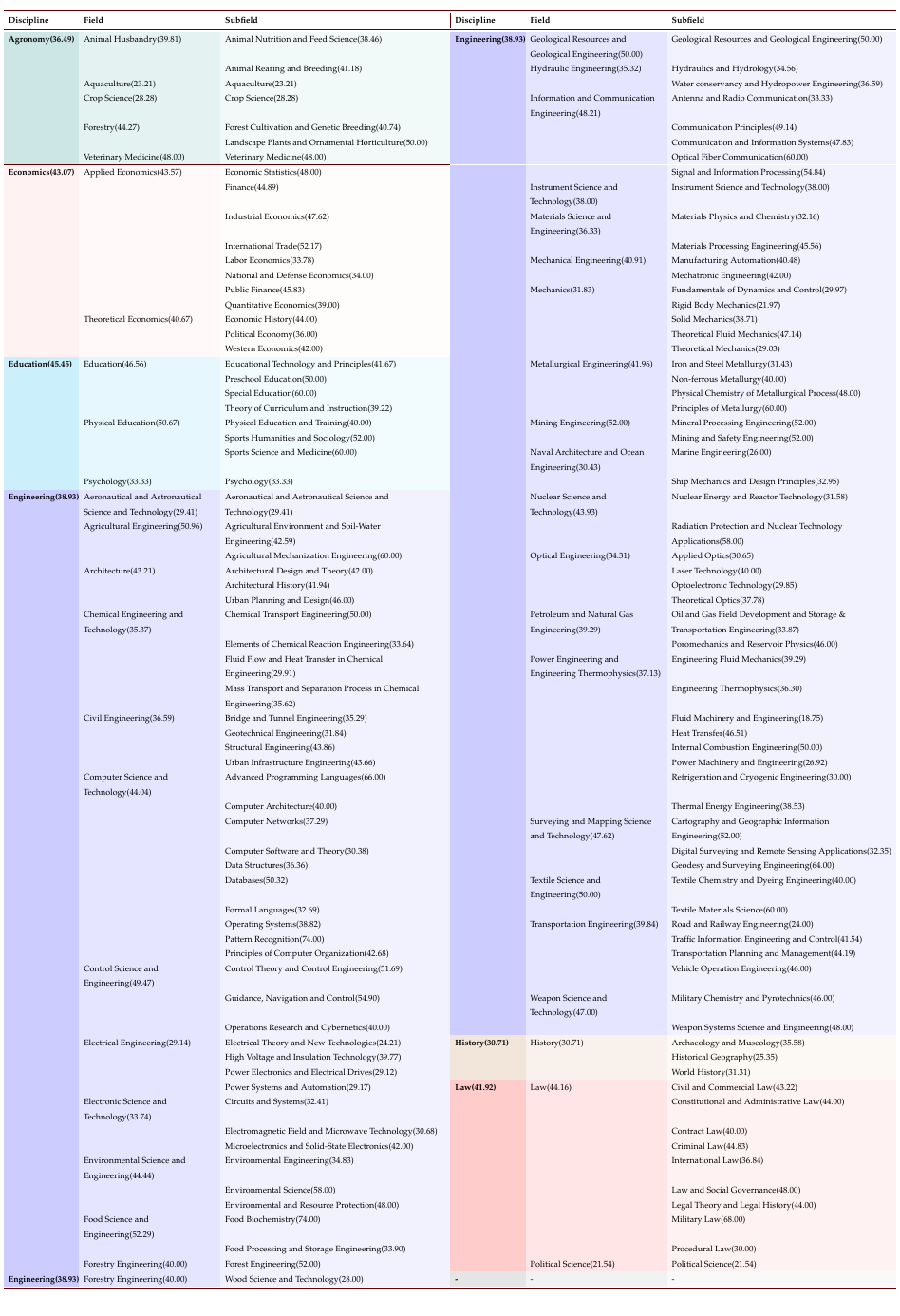} 
\end{subtable}
\vspace{-1.3cm}
\captionsetup{font=small}
\caption{Model Scores Across Three Levels of Disciplines: Qwen2.5-32B-Instruct.}
\label{tab:Qwen2.5-32B-Instruct}
\vspace{-0.5cm}
\centeredlinks{listofmodels}{Back to List of Models}{toc}{Back to Table of Contents}{blue}
\end{table}
\clearpage

\newpage
\afterpage{
    \begin{table}[t]
    \centering
    \ContinuedFloat 
    \begin{subtable}[t]{\textwidth}
    \centering
    \includegraphics[width=\textwidth]{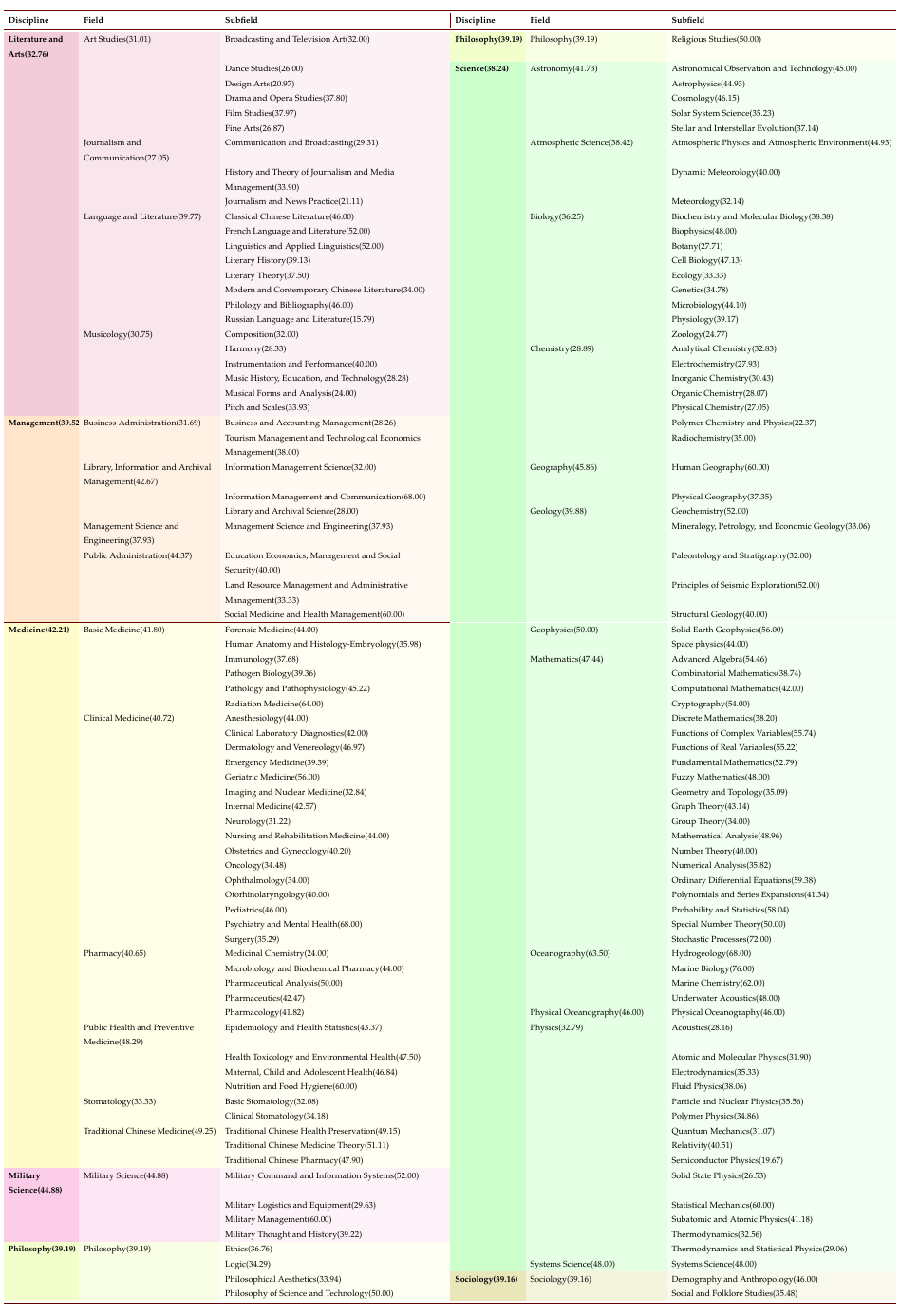} 
    \end{subtable}
    \vspace{-1.1cm}
    \captionsetup{font=small}
    \caption{Continued: Model Scores Across Three Levels of Disciplines: Qwen2.5-32B-Instruct.}
    \vspace{-0.6cm}
    \centeredlinks{listofmodels}{Back to List of Models}{toc}{Back to Table of Contents}{blue}
    \end{table}
}
\clearpage

\newpage
\vspace{-0.5cm}
\begin{table}[t]
\refstepcounter{models}%
\addcontentsline{csf}{models}{\protect\numberline{\themodels}Llama-3.3-70B-Instruct}
\centering
\begin{subtable}[t]{1\textwidth}
\centering
\includegraphics[width=\textwidth]{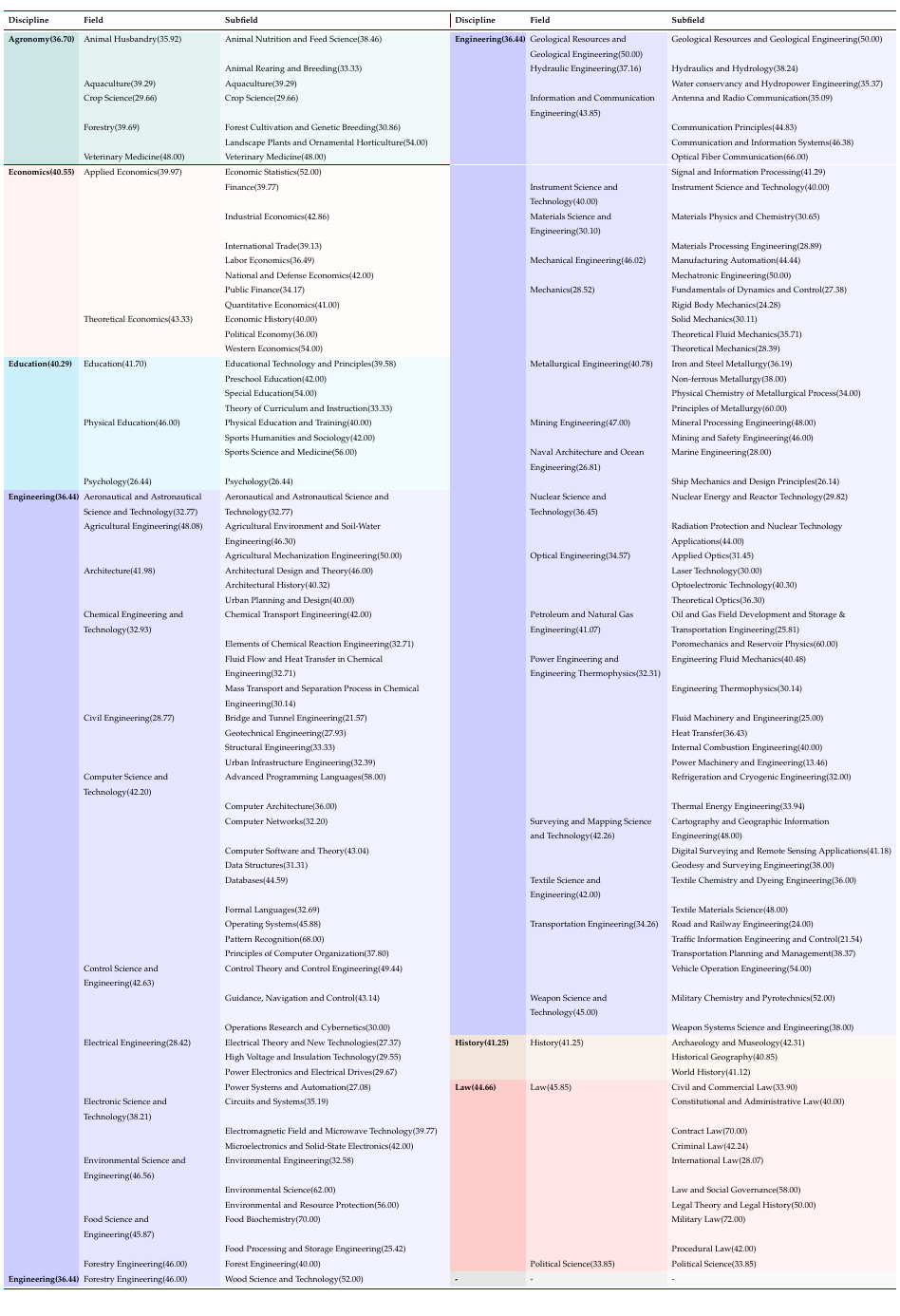} 
\end{subtable}
\vspace{-1.3cm}
\captionsetup{font=small}
\caption{Model Scores Across Three Levels of Disciplines: Llama-3.3-70B-Instruct.}
\label{tab:Llama-3.3-70B-Instruct}
\vspace{-0.5cm}
\centeredlinks{listofmodels}{Back to List of Models}{toc}{Back to Table of Contents}{blue}
\end{table}
\clearpage

\newpage
\afterpage{
    \begin{table}[t]
    \centering
    \ContinuedFloat 
    \begin{subtable}[t]{\textwidth}
    \centering
    \includegraphics[width=\textwidth]{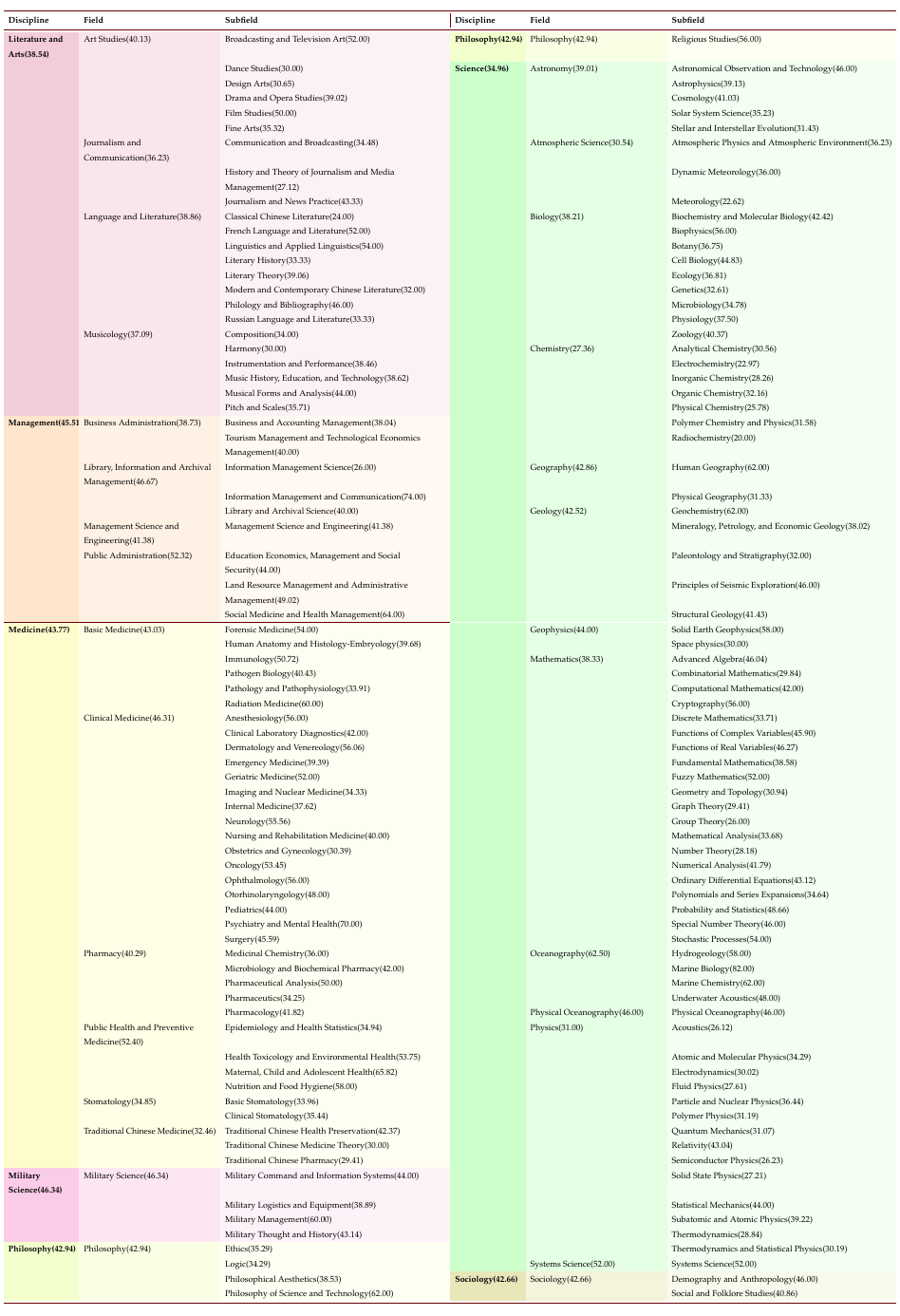} 
    \end{subtable}
    \vspace{-1.1cm}
    \captionsetup{font=small}
    \caption{Continued: Model Scores Across Three Levels of Disciplines: Llama-3.3-70B-Instruct.}
    \vspace{-0.6cm}
    \centeredlinks{listofmodels}{Back to List of Models}{toc}{Back to Table of Contents}{blue}
    \end{table}
}
\clearpage

\newpage
\vspace{-0.5cm}
\begin{table}[t]
\refstepcounter{models}%
\addcontentsline{csf}{models}{\protect\numberline{\themodels}phi-4}
\centering
\begin{subtable}[t]{1\textwidth}
\centering
\includegraphics[width=\textwidth]{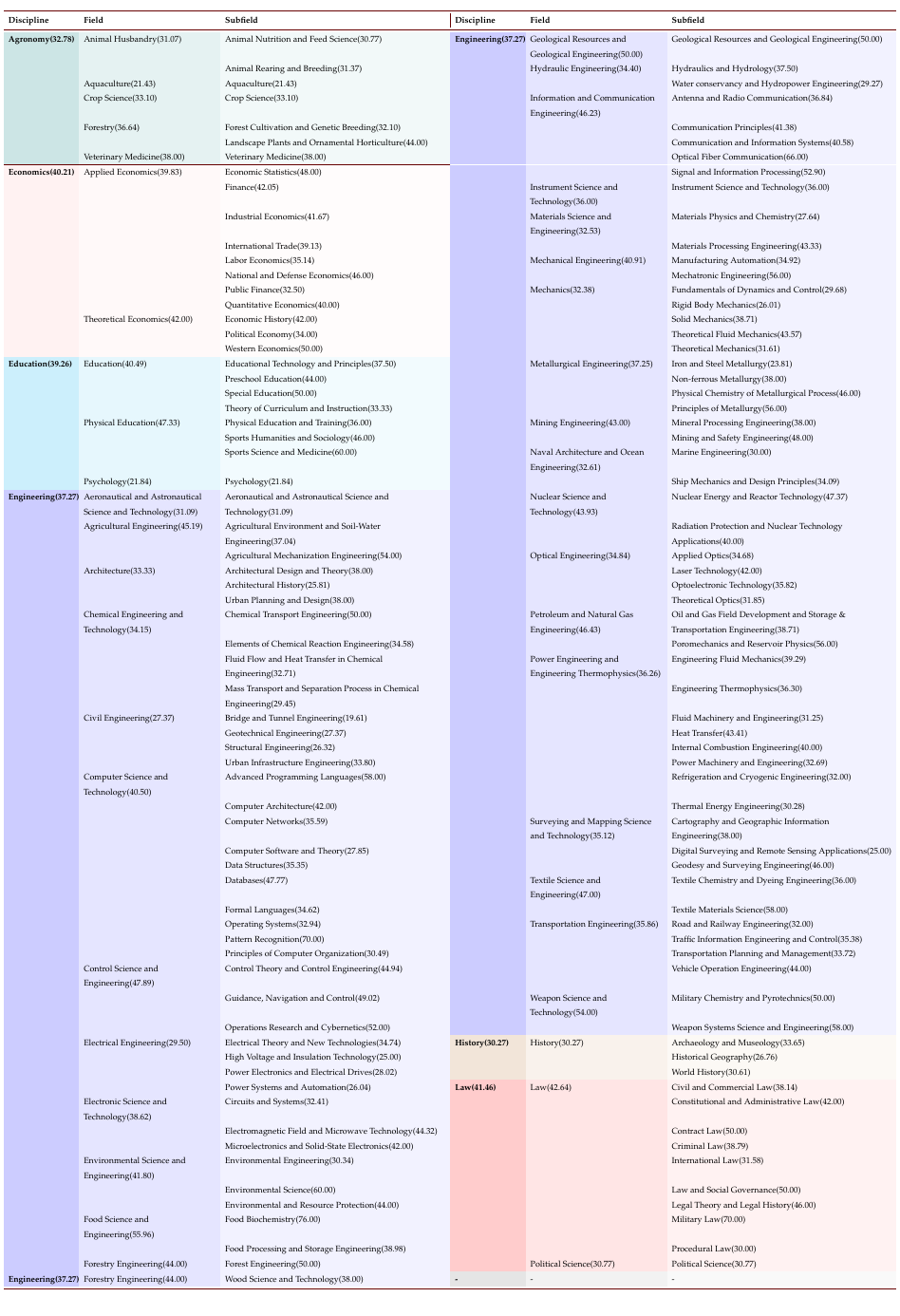} 
\end{subtable}
\vspace{-1.3cm}
\captionsetup{font=small}
\caption{Model Scores Across Three Levels of Disciplines: phi-4.}
\label{tab:phi-4}
\vspace{-0.5cm}
\centeredlinks{listofmodels}{Back to List of Models}{toc}{Back to Table of Contents}{blue}
\end{table}
\clearpage

\newpage
\afterpage{
    \begin{table}[t]
    \centering
    \ContinuedFloat 
    \begin{subtable}[t]{\textwidth}
    \centering
    \includegraphics[width=\textwidth]{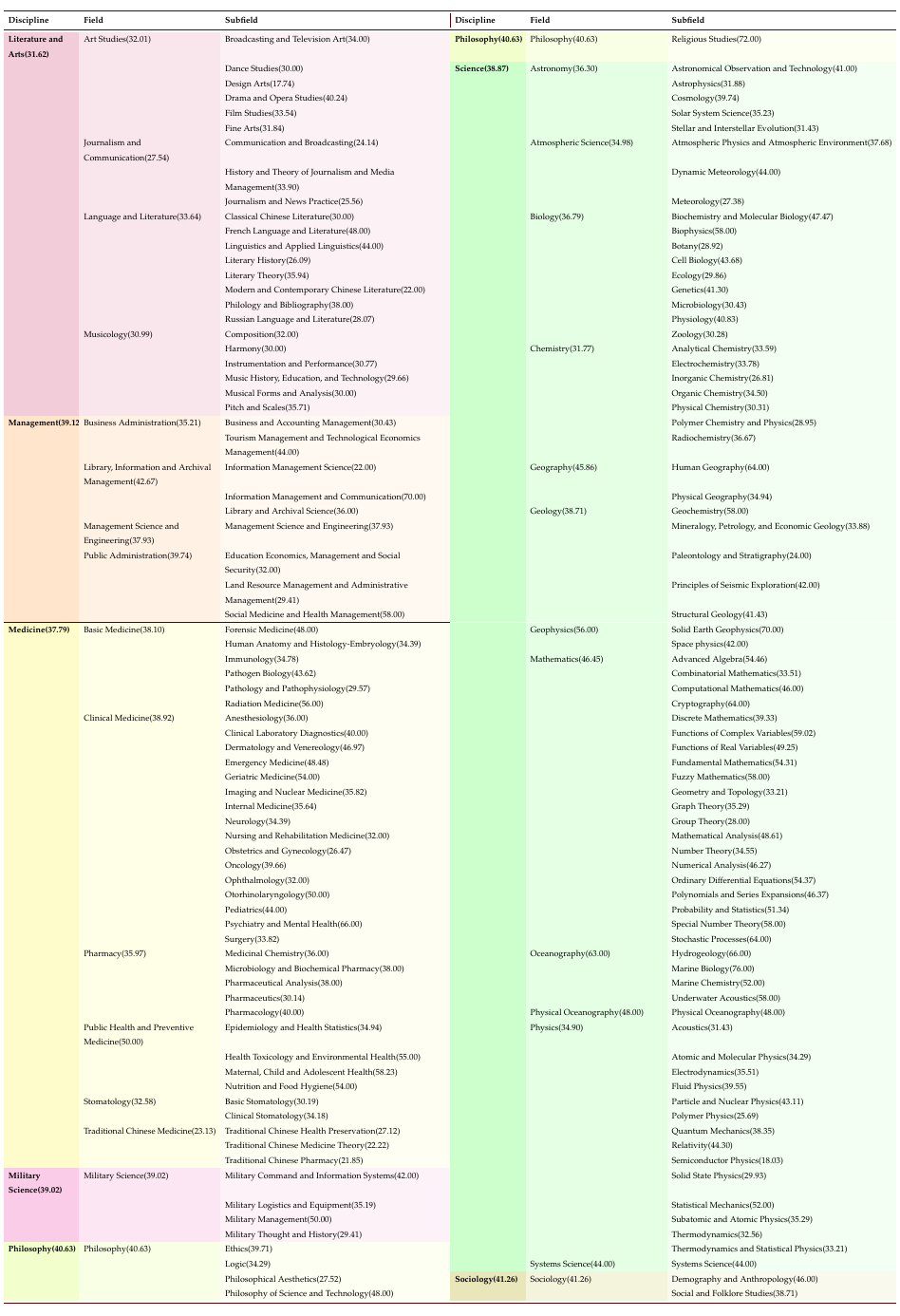} 
    \end{subtable}
    \vspace{-1.1cm}
    \captionsetup{font=small}
    \caption{Continued: Model Scores Across Three Levels of Disciplines: phi-4.}
    \vspace{-0.6cm}
    \centeredlinks{listofmodels}{Back to List of Models}{toc}{Back to Table of Contents}{blue}
    \end{table}
}
\clearpage

\newpage
\vspace{-0.5cm}
\begin{table}[t]
\refstepcounter{models}%
\addcontentsline{csf}{models}{\protect\numberline{\themodels}Qwen2.5-14B-Instruct}
\centering
\begin{subtable}[t]{1\textwidth}
\centering
\includegraphics[width=\textwidth]{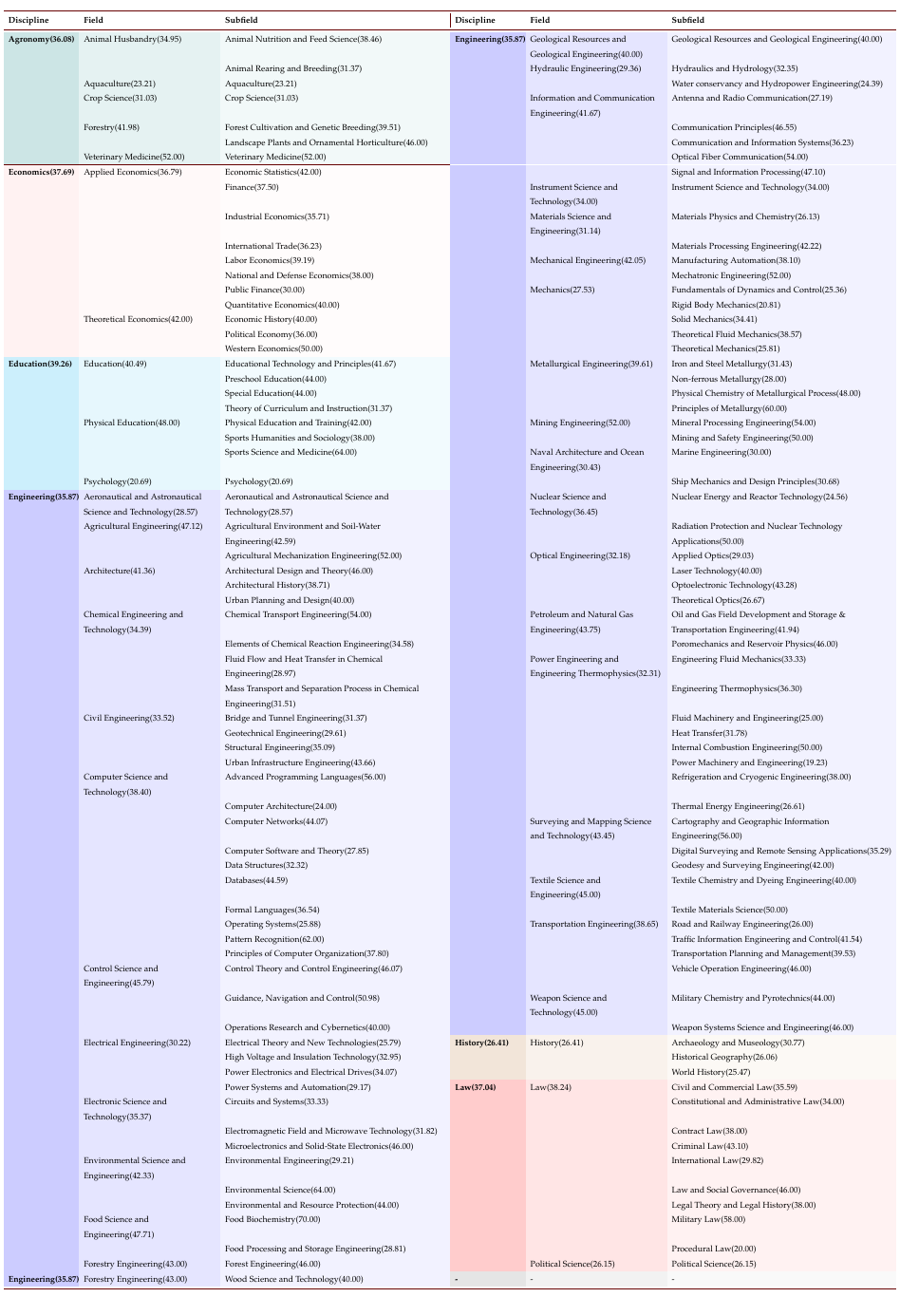} 
\end{subtable}
\vspace{-1.3cm}
\captionsetup{font=small}
\caption{Model Scores Across Three Levels of Disciplines: Qwen2.5-14B-Instruct.}
\label{tab:Qwen2.5-14B-Instruct}
\vspace{-0.5cm}
\centeredlinks{listofmodels}{Back to List of Models}{toc}{Back to Table of Contents}{blue}
\end{table}
\clearpage

\newpage
\afterpage{
    \begin{table}[t]
    \centering
    \ContinuedFloat 
    \begin{subtable}[t]{\textwidth}
    \centering
    \includegraphics[width=\textwidth]{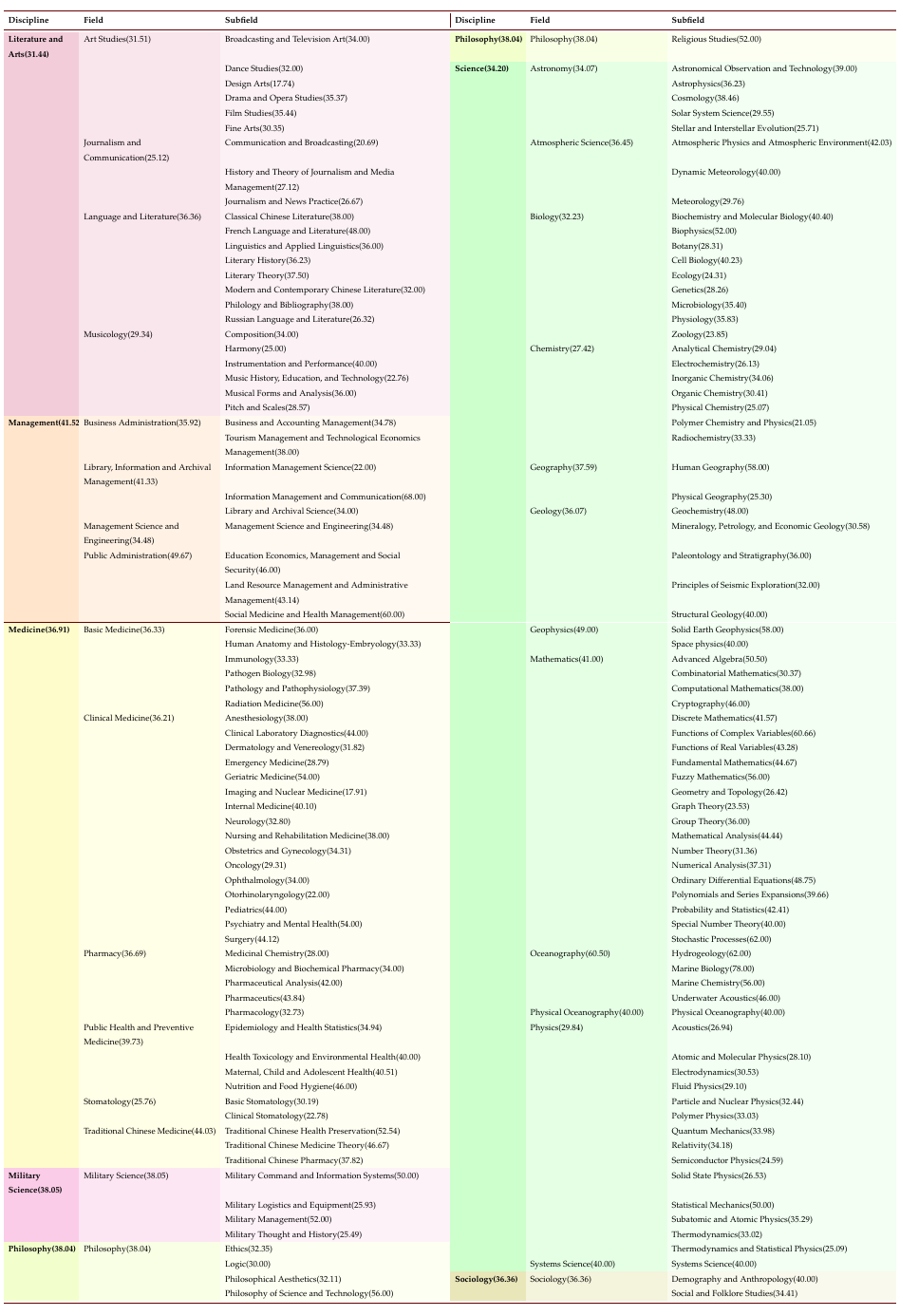} 
    \end{subtable}
    \vspace{-1.1cm}
    \captionsetup{font=small}
    \caption{Continued: Model Scores Across Three Levels of Disciplines: Qwen2.5-14B-Instruct.}
    \vspace{-0.6cm}
    \centeredlinks{listofmodels}{Back to List of Models}{toc}{Back to Table of Contents}{blue}
    \end{table}
}
\clearpage

\newpage
\vspace{-0.5cm}
\begin{table}[t]
\refstepcounter{models}%
\addcontentsline{csf}{models}{\protect\numberline{\themodels}Llama-3.1-70B-Instruct}
\centering
\begin{subtable}[t]{1\textwidth}
\centering
\includegraphics[width=\textwidth]{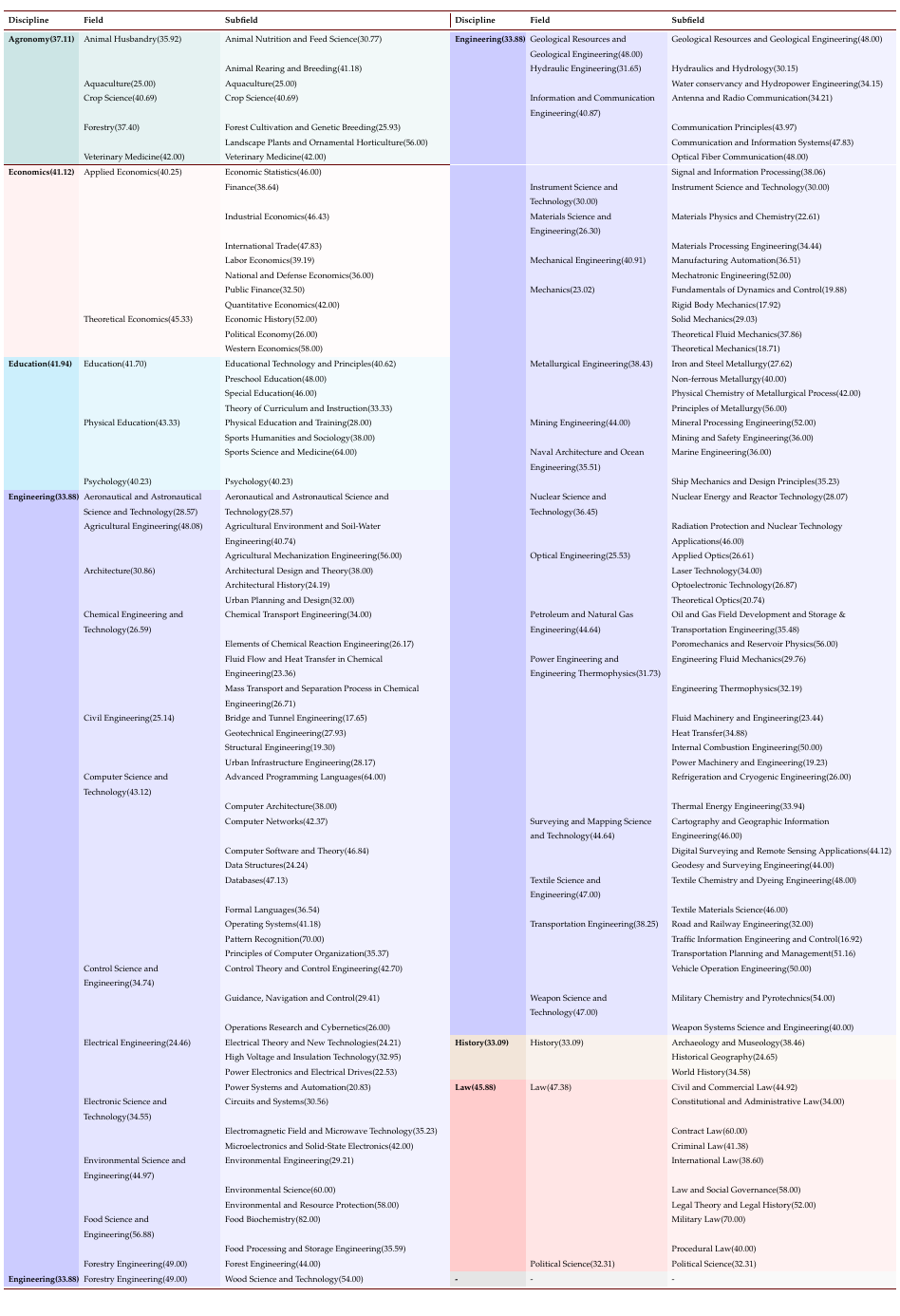} 
\end{subtable}
\vspace{-1.3cm}
\captionsetup{font=small}
\caption{Model Scores Across Three Levels of Disciplines: Llama-3.1-70B-Instruct.}
\label{tab:Llama-3.1-70B-Instruct}
\vspace{-0.5cm}
\centeredlinks{listofmodels}{Back to List of Models}{toc}{Back to Table of Contents}{blue}
\end{table}
\clearpage

\newpage
\afterpage{
    \begin{table}[t]
    \centering
    \ContinuedFloat 
    \begin{subtable}[t]{\textwidth}
    \centering
    \includegraphics[width=\textwidth]{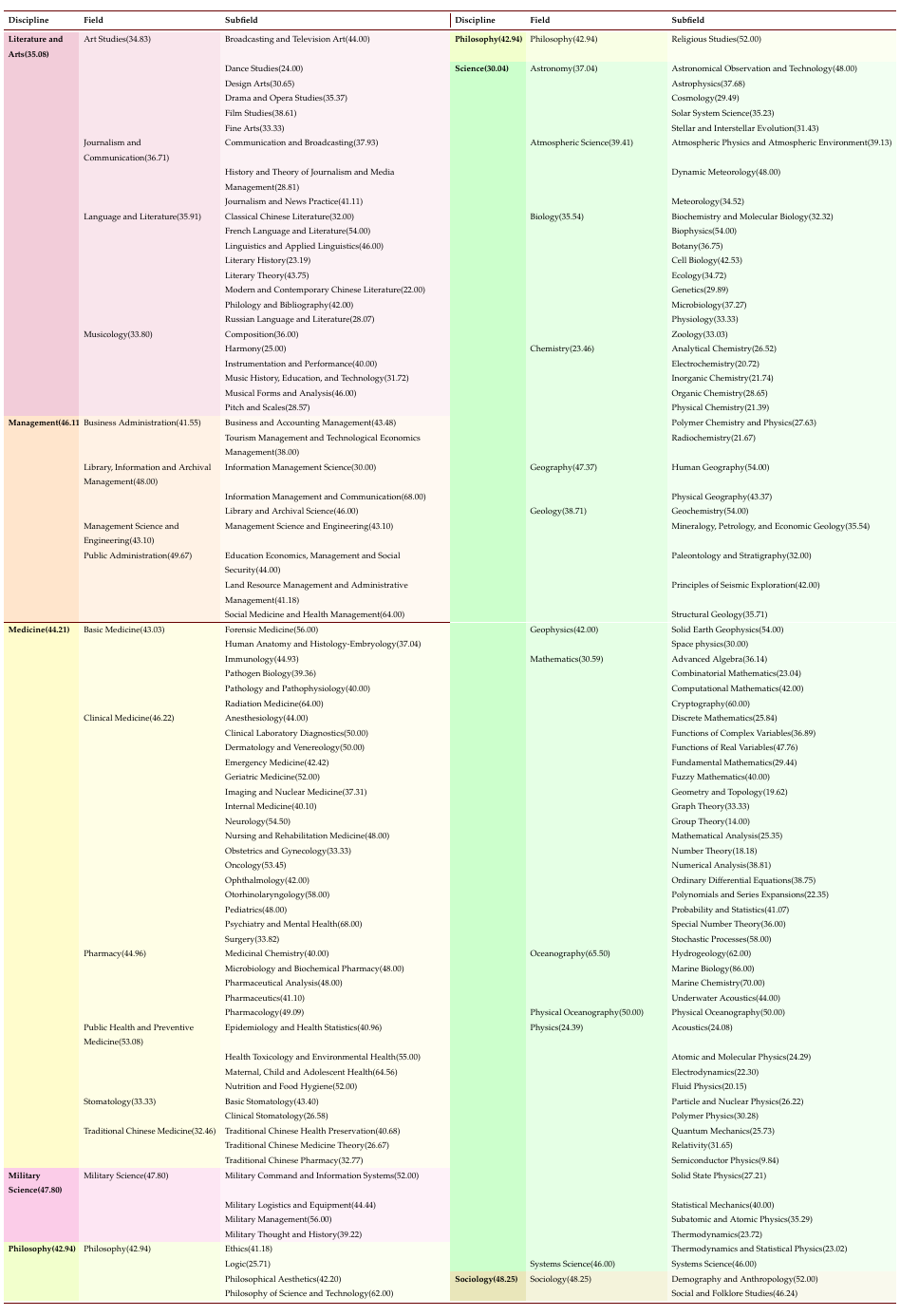} 
    \end{subtable}
    \vspace{-1.1cm}
    \captionsetup{font=small}
    \caption{Continued: Model Scores Across Three Levels of Disciplines: Llama-3.1-70B-Instruct.}
    \vspace{-0.6cm}
    \centeredlinks{listofmodels}{Back to List of Models}{toc}{Back to Table of Contents}{blue}
    \end{table}
}
\clearpage

\newpage
\vspace{-0.5cm}
\begin{table}[t]
\refstepcounter{models}%
\addcontentsline{csf}{models}{\protect\numberline{\themodels}Qwen2.5-72B}
\centering
\begin{subtable}[t]{1\textwidth}
\centering
\includegraphics[width=\textwidth]{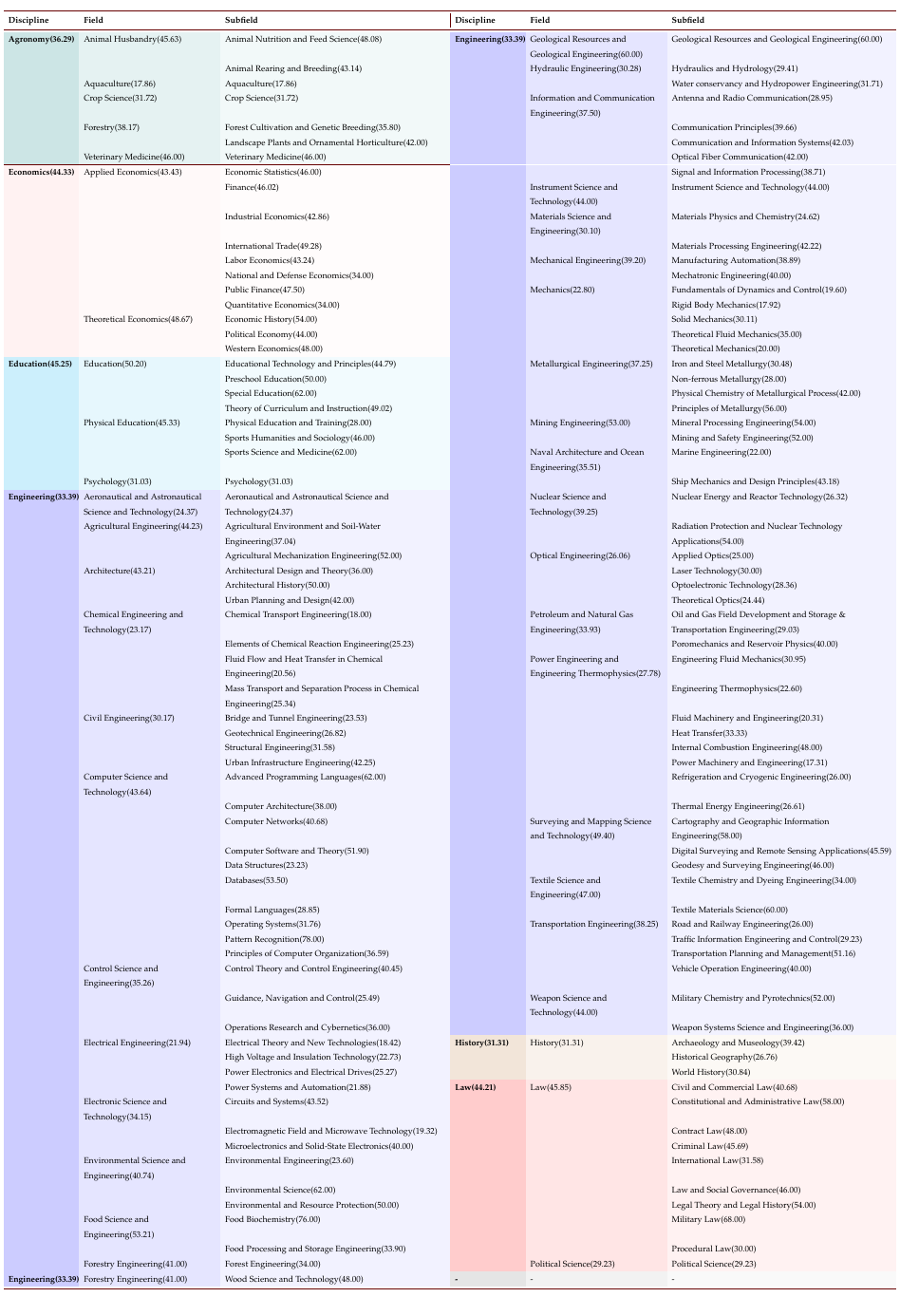} 
\end{subtable}
\vspace{-1.3cm}
\captionsetup{font=small}
\caption{Model Scores Across Three Levels of Disciplines: Qwen2.5-72B.}
\label{tab:Qwen2.5-72B}
\vspace{-0.5cm}
\centeredlinks{listofmodels}{Back to List of Models}{toc}{Back to Table of Contents}{blue}
\end{table}
\clearpage

\newpage
\afterpage{
    \begin{table}[t]
    \centering
    \ContinuedFloat 
    \begin{subtable}[t]{\textwidth}
    \centering
    \includegraphics[width=\textwidth]{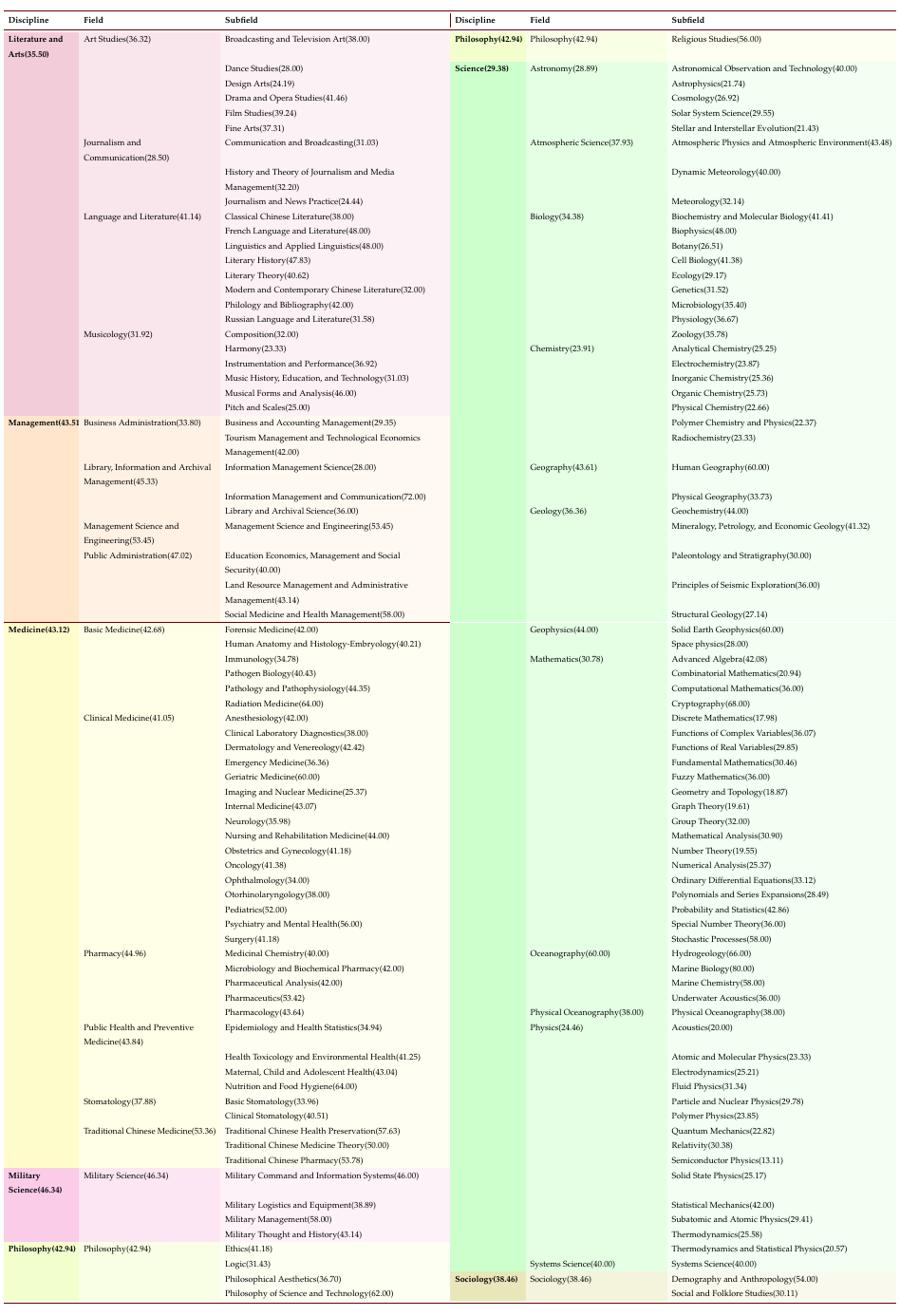} 
    \end{subtable}
    \vspace{-1.1cm}
    \captionsetup{font=small}
    \caption{Continued: Model Scores Across Three Levels of Disciplines: Qwen2.5-72B.}
    \vspace{-0.6cm}
    \centeredlinks{listofmodels}{Back to List of Models}{toc}{Back to Table of Contents}{blue}
    \end{table}
}
\clearpage

\newpage
\vspace{-0.5cm}
\begin{table}[t]
\refstepcounter{models}%
\addcontentsline{csf}{models}{\protect\numberline{\themodels}Yi-Lighting}
\centering
\begin{subtable}[t]{1\textwidth}
\centering
\includegraphics[width=\textwidth]{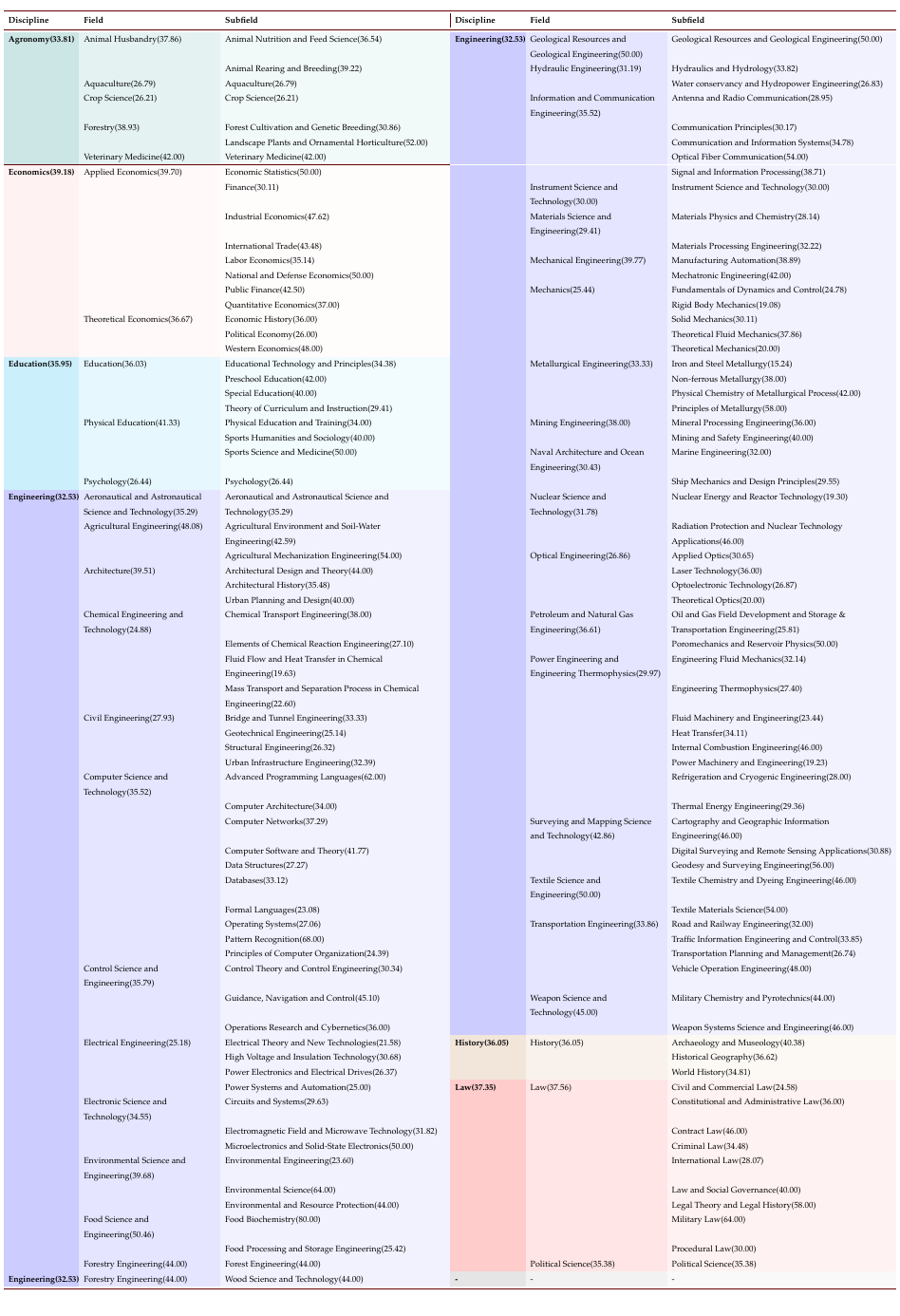} 
\end{subtable}
\vspace{-1.3cm}
\captionsetup{font=small}
\caption{Model Scores Across Three Levels of Disciplines: Yi-Lighting.}
\label{tab:Yi-Lighting}
\vspace{-0.5cm}
\centeredlinks{listofmodels}{Back to List of Models}{toc}{Back to Table of Contents}{blue}
\end{table}
\clearpage

\newpage
\afterpage{
    \begin{table}[t]
    \centering
    \ContinuedFloat 
    \begin{subtable}[t]{\textwidth}
    \centering
    \includegraphics[width=\textwidth]{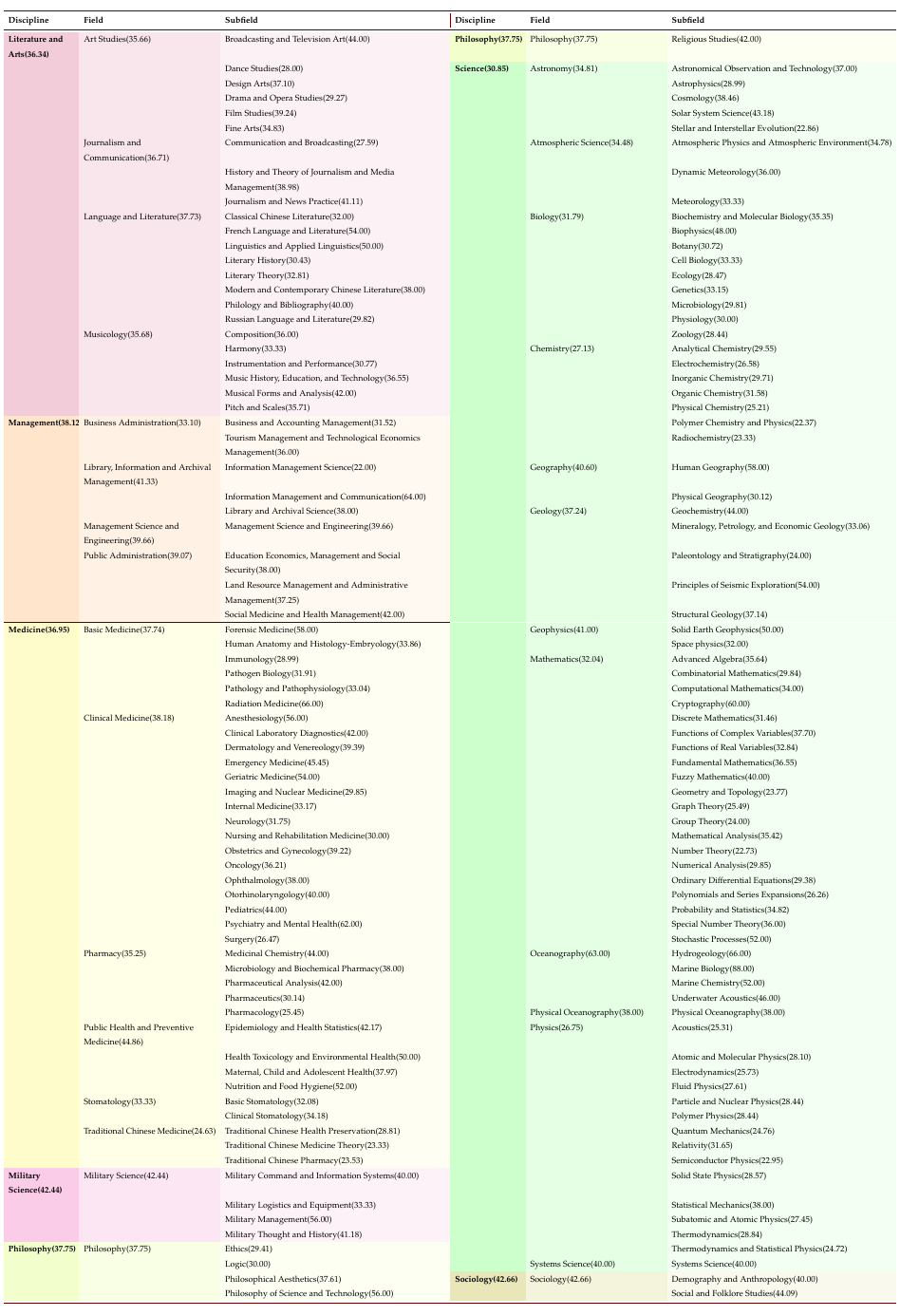} 
    \end{subtable}
    \vspace{-1.1cm}
    \captionsetup{font=small}
    \caption{Continued: Model Scores Across Three Levels of Disciplines: Yi-Lighting.}
    \vspace{-0.6cm}
    \centeredlinks{listofmodels}{Back to List of Models}{toc}{Back to Table of Contents}{blue}
    \end{table}
}
\clearpage

\newpage
\vspace{-0.5cm}
\begin{table}[t]
\refstepcounter{models}%
\addcontentsline{csf}{models}{\protect\numberline{\themodels}Qwen2.5-32B}
\centering
\begin{subtable}[t]{1\textwidth}
\centering
\includegraphics[width=\textwidth]{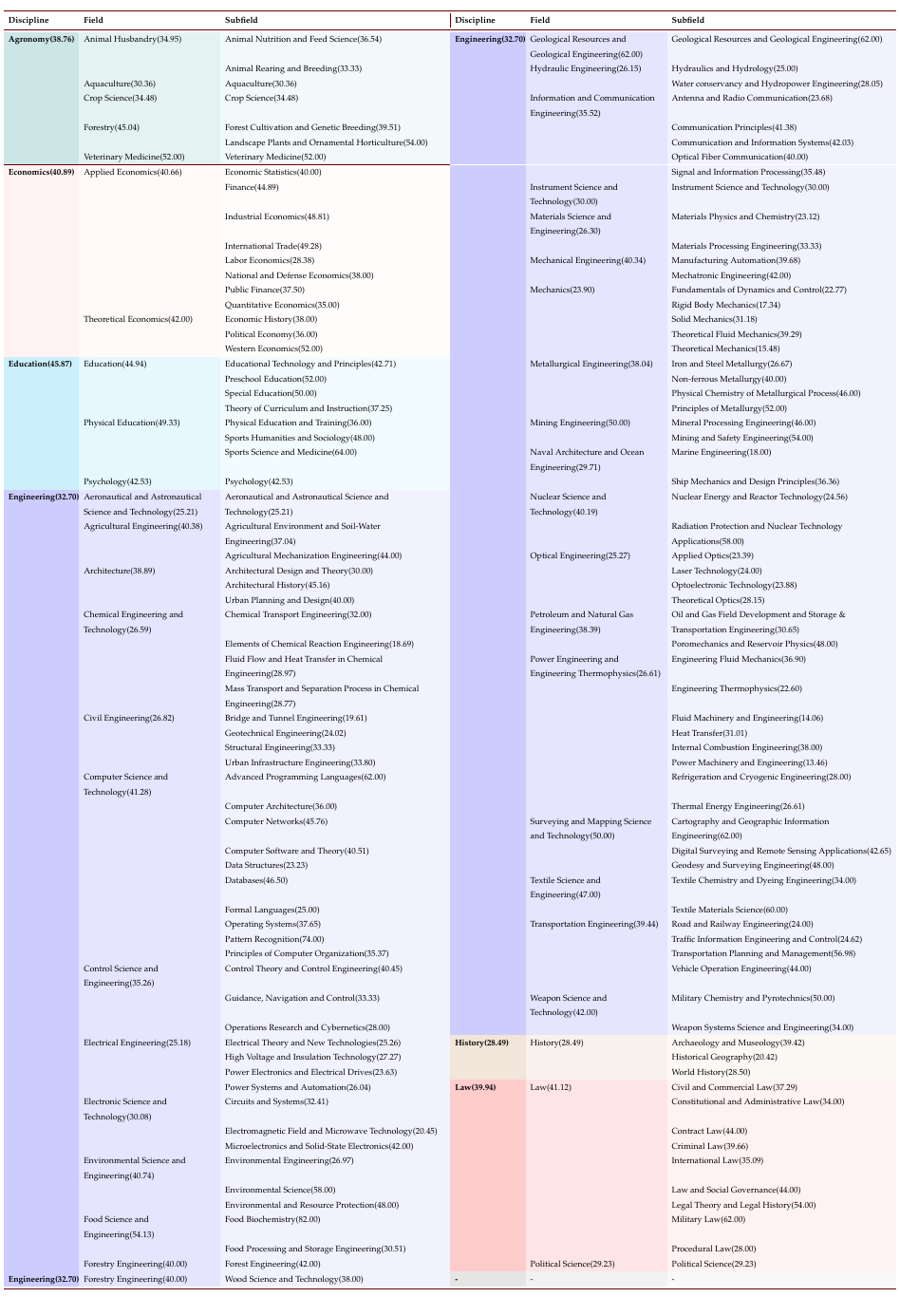} 
\end{subtable}
\vspace{-1.3cm}
\captionsetup{font=small}
\caption{Model Scores Across Three Levels of Disciplines: Qwen2.5-32B.}
\label{tab:Qwen2.5-32B}
\vspace{-0.5cm}
\centeredlinks{listofmodels}{Back to List of Models}{toc}{Back to Table of Contents}{blue}
\end{table}
\clearpage

\newpage
\afterpage{
    \begin{table}[t]
    \centering
    \ContinuedFloat 
    \begin{subtable}[t]{\textwidth}
    \centering
    \includegraphics[width=\textwidth]{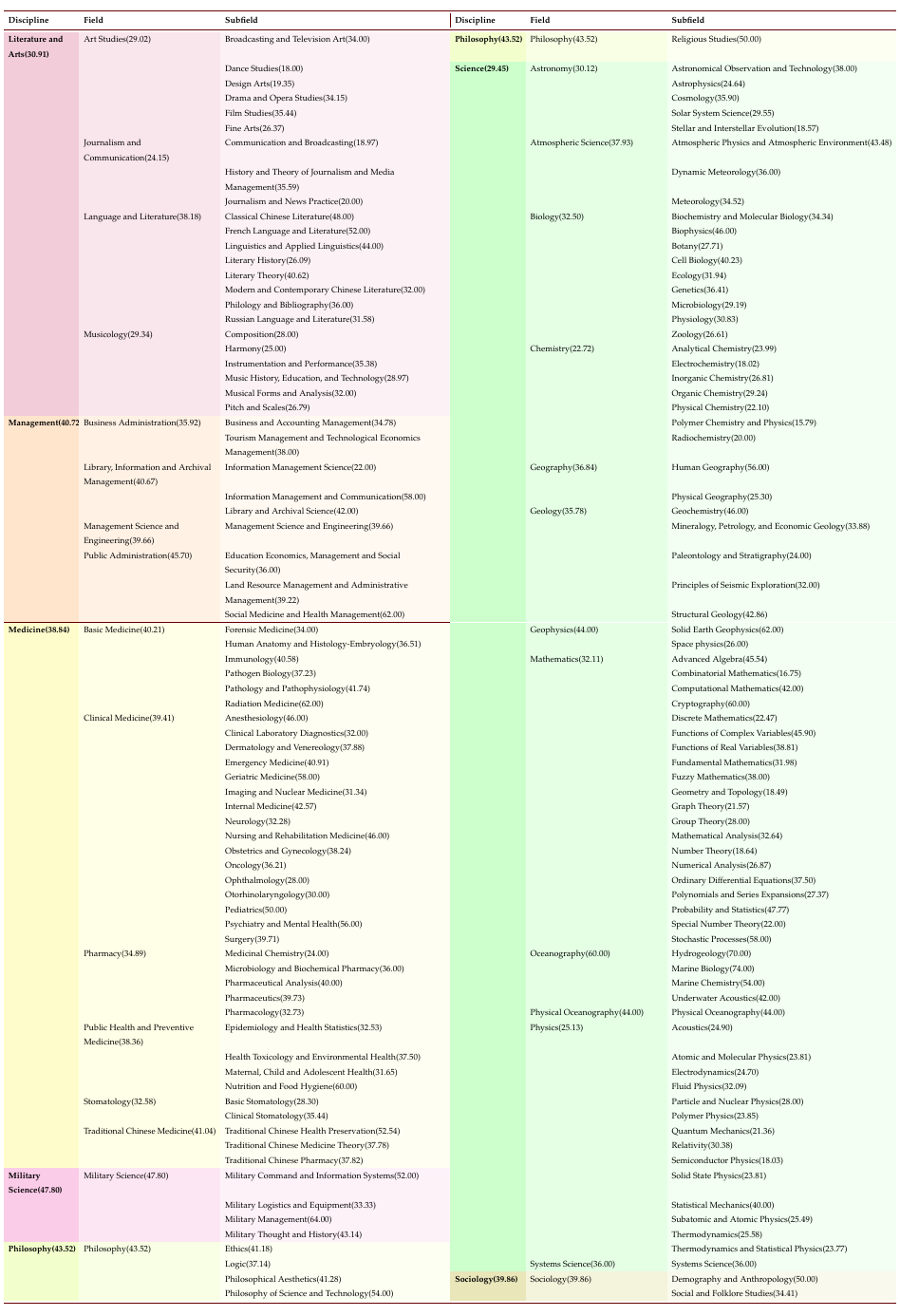} 
    \end{subtable}
    \vspace{-1.1cm}
    \captionsetup{font=small}
    \caption{Continued: Model Scores Across Three Levels of Disciplines: Qwen2.5-32B.}
    \vspace{-0.6cm}
    \centeredlinks{listofmodels}{Back to List of Models}{toc}{Back to Table of Contents}{blue}
    \end{table}
}
\clearpage

\newpage
\vspace{-0.5cm}
\begin{table}[t]
\refstepcounter{models}%
\addcontentsline{csf}{models}{\protect\numberline{\themodels}DeepSeek-V3-Base}
\centering
\begin{subtable}[t]{1\textwidth}
\centering
\includegraphics[width=\textwidth]{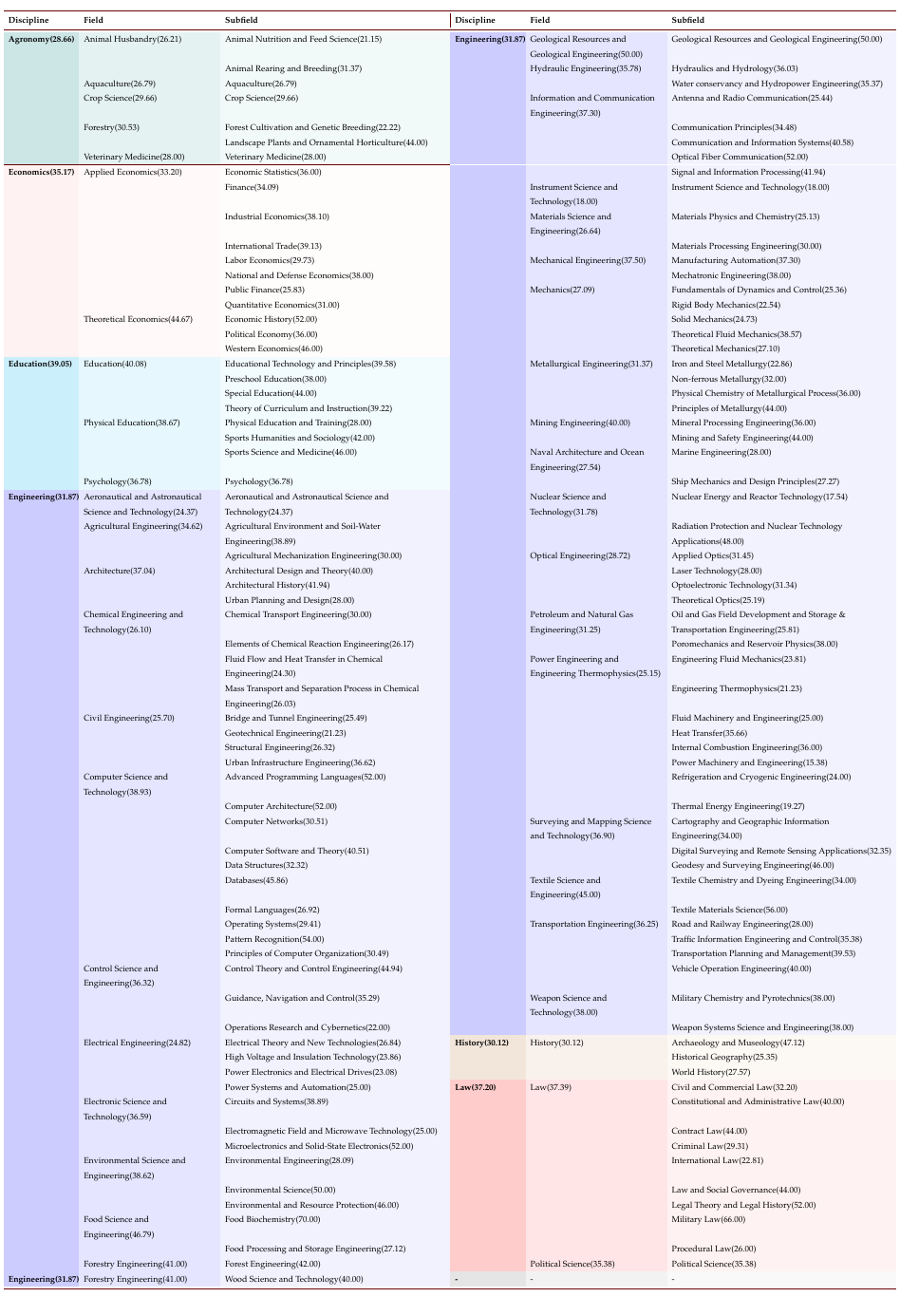} 
\end{subtable}
\vspace{-1.3cm}
\captionsetup{font=small}
\caption{Model Scores Across Three Levels of Disciplines: DeepSeek-V3-Base.}
\label{tab:DeepSeek-V3-Base}
\vspace{-0.5cm}
\centeredlinks{listofmodels}{Back to List of Models}{toc}{Back to Table of Contents}{blue}
\end{table}
\clearpage

\newpage
\afterpage{
    \begin{table}[t]
    \centering
    \ContinuedFloat 
    \begin{subtable}[t]{\textwidth}
    \centering
    \includegraphics[width=\textwidth]{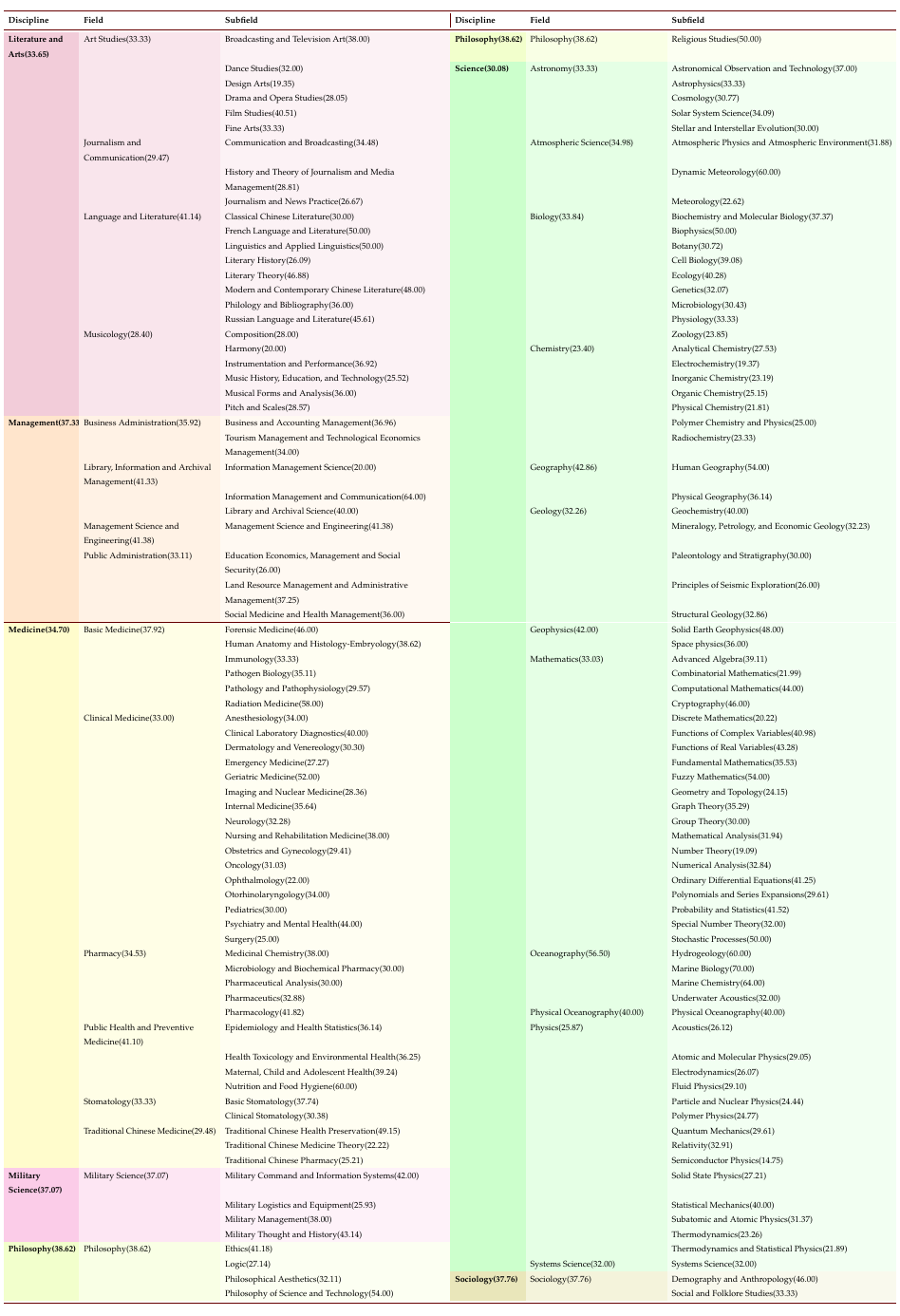} 
    \end{subtable}
    \vspace{-1.1cm}
    \captionsetup{font=small}
    \caption{Continued: Model Scores Across Three Levels of Disciplines: DeepSeek-V3-Base.}
    \vspace{-0.6cm}
    \centeredlinks{listofmodels}{Back to List of Models}{toc}{Back to Table of Contents}{blue}
    \end{table}
}
\clearpage

\newpage
\vspace{-0.5cm}
\begin{table}[t]
\refstepcounter{models}%
\addcontentsline{csf}{models}{\protect\numberline{\themodels}Qwen2.5-14B}
\centering
\begin{subtable}[t]{1\textwidth}
\centering
\includegraphics[width=\textwidth]{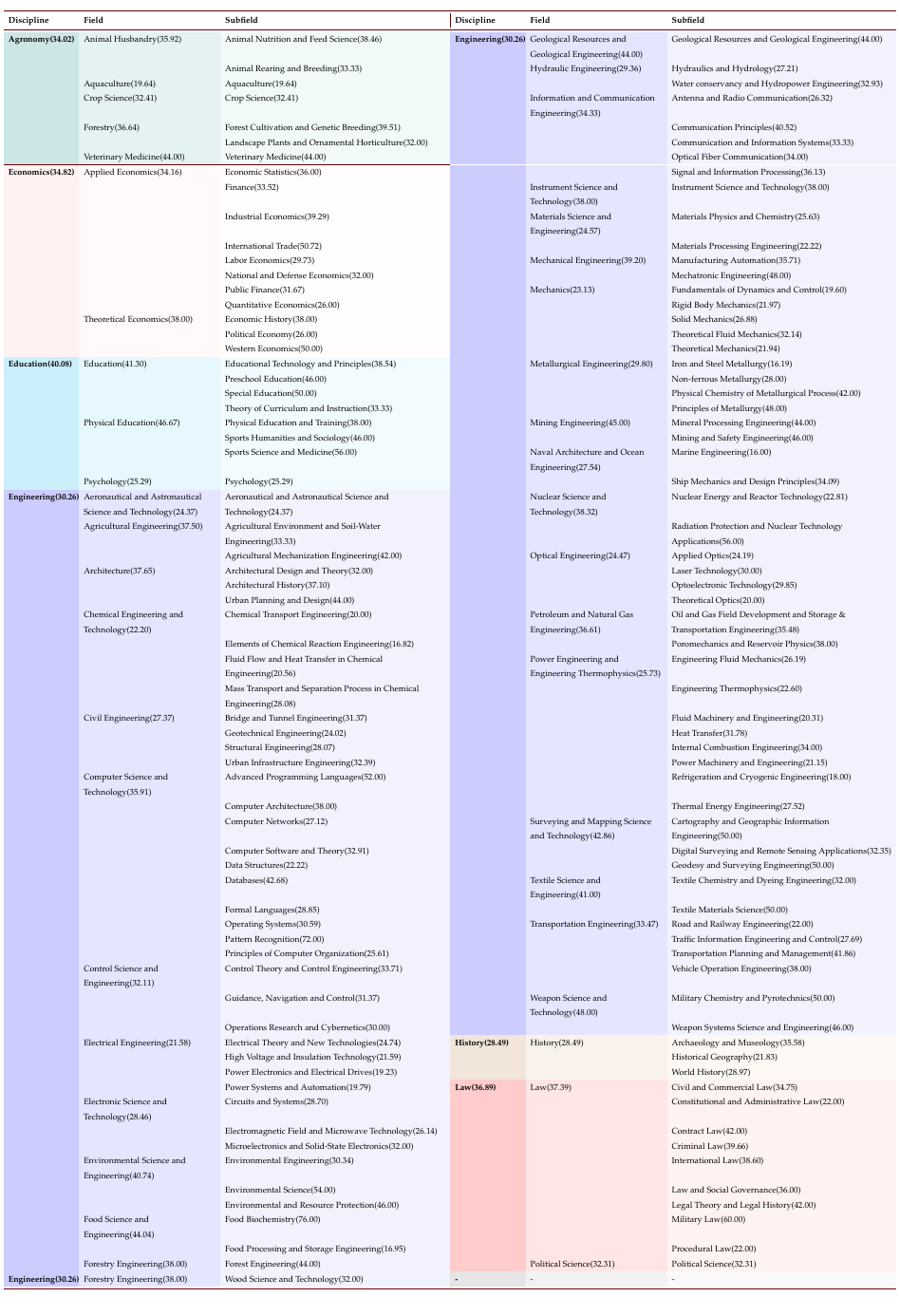} 
\end{subtable}
\vspace{-1.3cm}
\captionsetup{font=small}
\caption{Model Scores Across Three Levels of Disciplines: Qwen2.5-14B.}
\label{tab:Qwen2.5-14B}
\vspace{-0.5cm}
\centeredlinks{listofmodels}{Back to List of Models}{toc}{Back to Table of Contents}{blue}
\end{table}
\clearpage

\newpage
\afterpage{
    \begin{table}[t]
    \centering
    \ContinuedFloat 
    \begin{subtable}[t]{\textwidth}
    \centering
    \includegraphics[width=\textwidth]{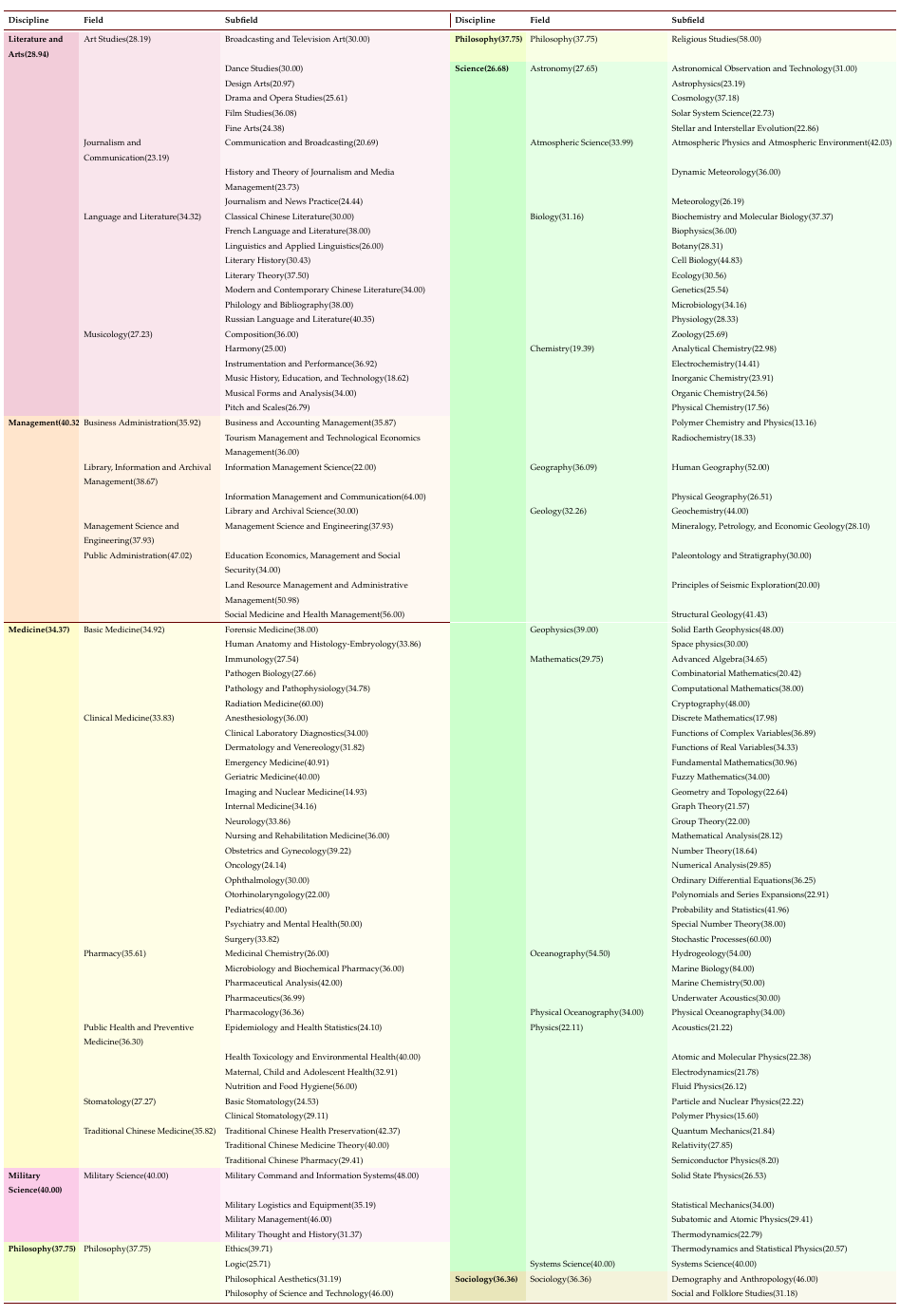} 
    \end{subtable}
    \vspace{-1.1cm}
    \captionsetup{font=small}
    \caption{Continued: Model Scores Across Three Levels of Disciplines: Qwen2.5-14B.}
    \vspace{-0.6cm}
    \centeredlinks{listofmodels}{Back to List of Models}{toc}{Back to Table of Contents}{blue}
    \end{table}
}
\clearpage

\newpage
\vspace{-0.5cm}
\begin{table}[t]
\refstepcounter{models}%
\addcontentsline{csf}{models}{\protect\numberline{\themodels}Mixtral-8x22B-Instruct-v0.1}
\centering
\begin{subtable}[t]{1\textwidth}
\centering
\includegraphics[width=\textwidth]{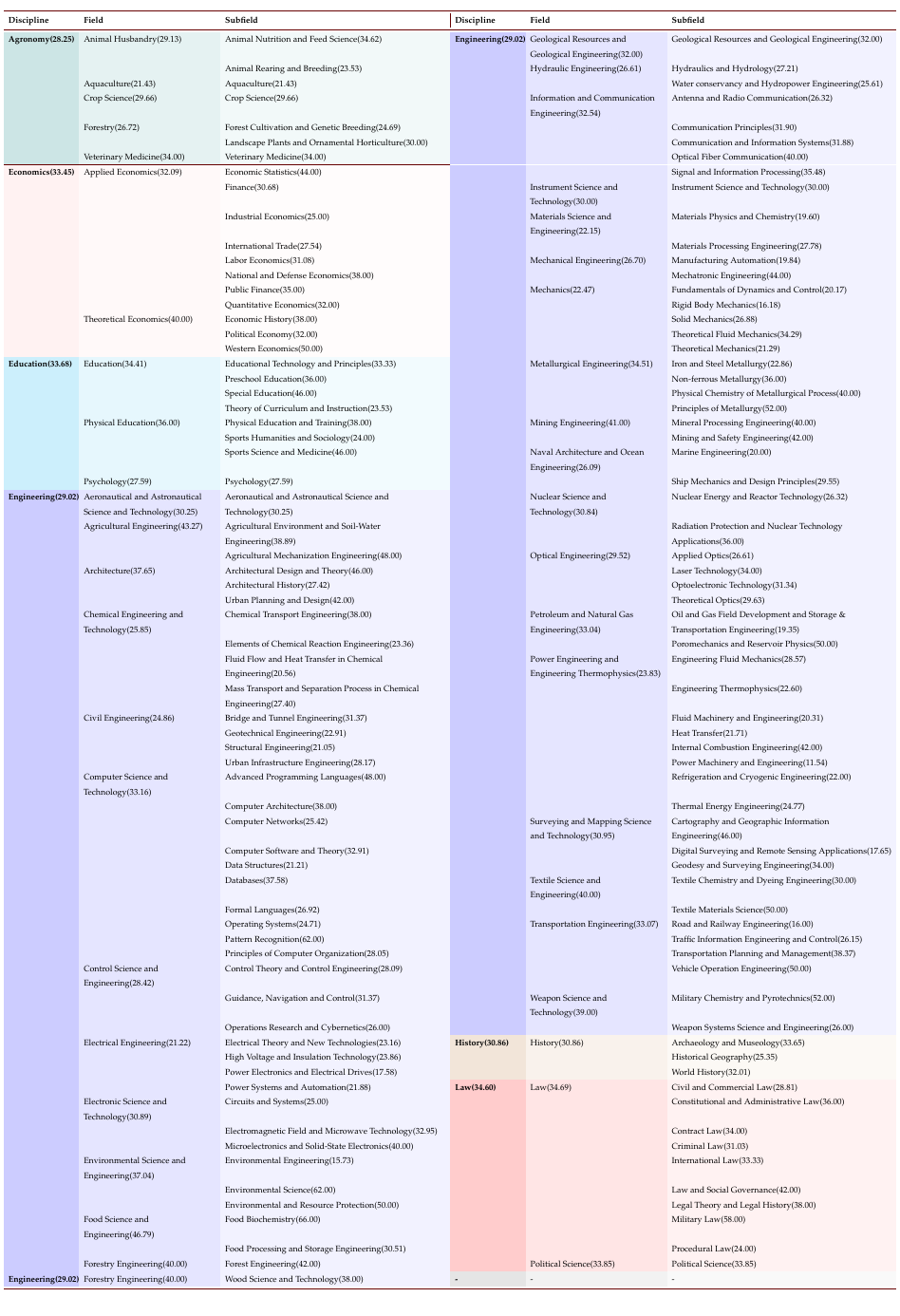} 
\end{subtable}
\vspace{-1.3cm}
\captionsetup{font=small}
\caption{Model Scores Across Three Levels of Disciplines: Mixtral-8x22B-Instruct-v0.1.}
\label{tab:Mixtral-8x22B-Instruct-v0.1}
\vspace{-0.5cm}
\centeredlinks{listofmodels}{Back to List of Models}{toc}{Back to Table of Contents}{blue}
\end{table}
\clearpage

\newpage
\afterpage{
    \begin{table}[t]
    \centering
    \ContinuedFloat 
    \begin{subtable}[t]{\textwidth}
    \centering
    \includegraphics[width=\textwidth]{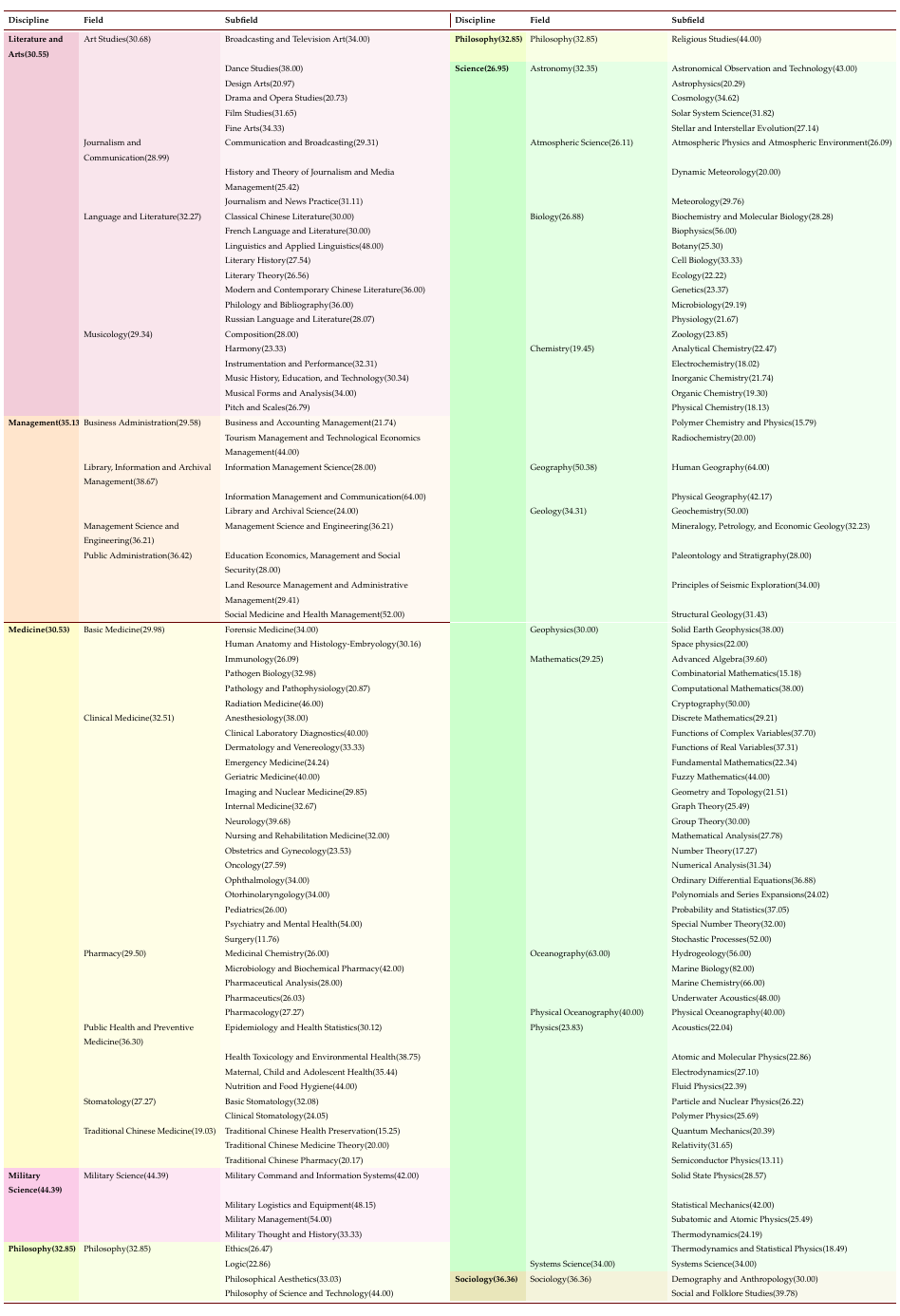} 
    \end{subtable}
    \vspace{-1.1cm}
    \captionsetup{font=small}
    \caption{Continued: Model Scores Across Three Levels of Disciplines: Mixtral-8x22B-Instruct-v0.1.}
    \vspace{-0.6cm}
    \centeredlinks{listofmodels}{Back to List of Models}{toc}{Back to Table of Contents}{blue}
    \end{table}
}
\clearpage

\newpage
\vspace{-0.5cm}
\begin{table}[t]
\refstepcounter{models}%
\addcontentsline{csf}{models}{\protect\numberline{\themodels}Qwen2.5-7B-Instruct}
\centering
\begin{subtable}[t]{1\textwidth}
\centering
\includegraphics[width=\textwidth]{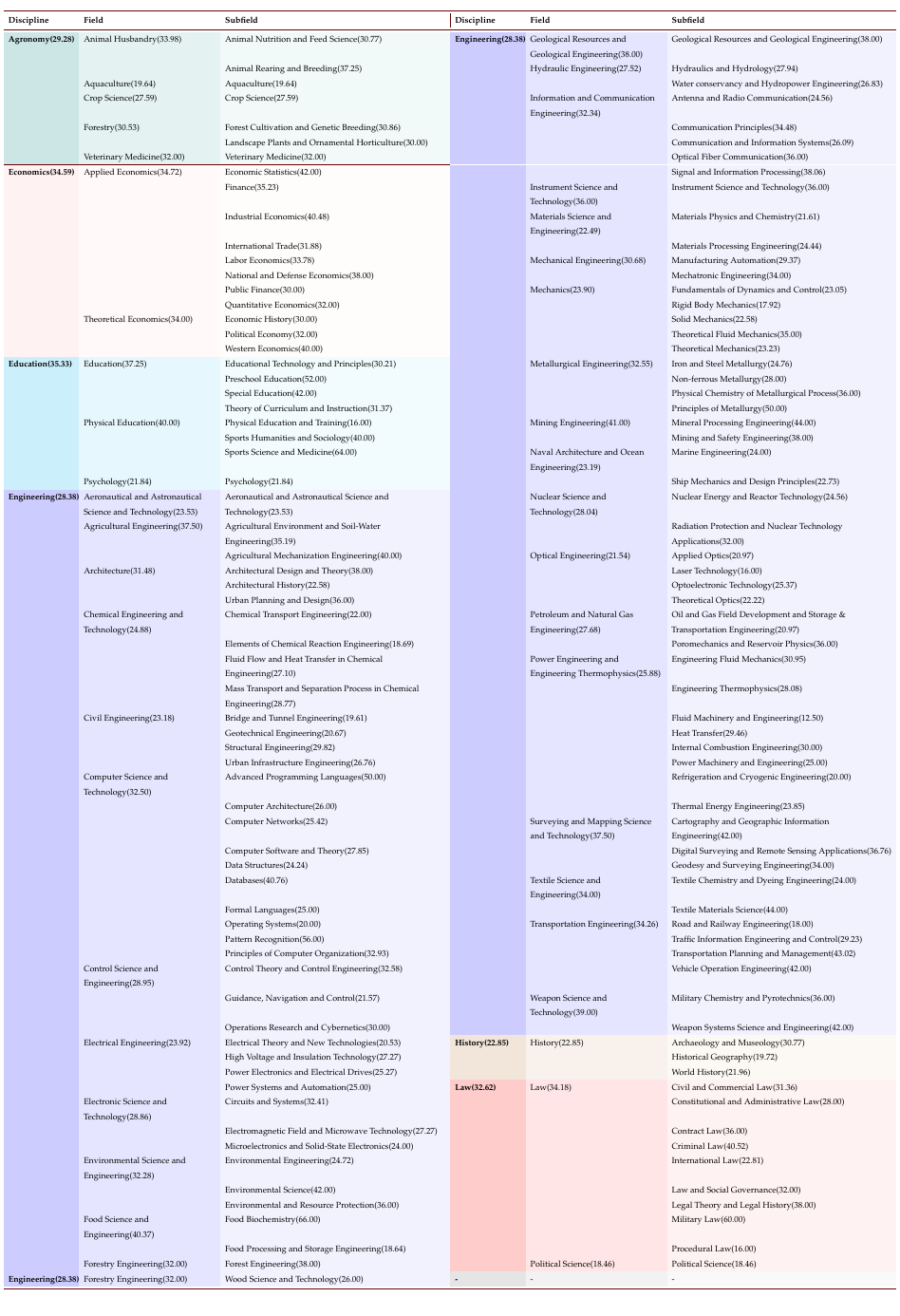} 
\end{subtable}
\vspace{-1.3cm}
\captionsetup{font=small}
\caption{Model Scores Across Three Levels of Disciplines: Qwen2.5-7B-Instruct.}
\label{tab:Qwen2.5-7B-Instruct}
\vspace{-0.5cm}
\centeredlinks{listofmodels}{Back to List of Models}{toc}{Back to Table of Contents}{blue}
\end{table}
\clearpage

\newpage
\afterpage{
    \begin{table}[t]
    \centering
    \ContinuedFloat 
    \begin{subtable}[t]{\textwidth}
    \centering
    \includegraphics[width=\textwidth]{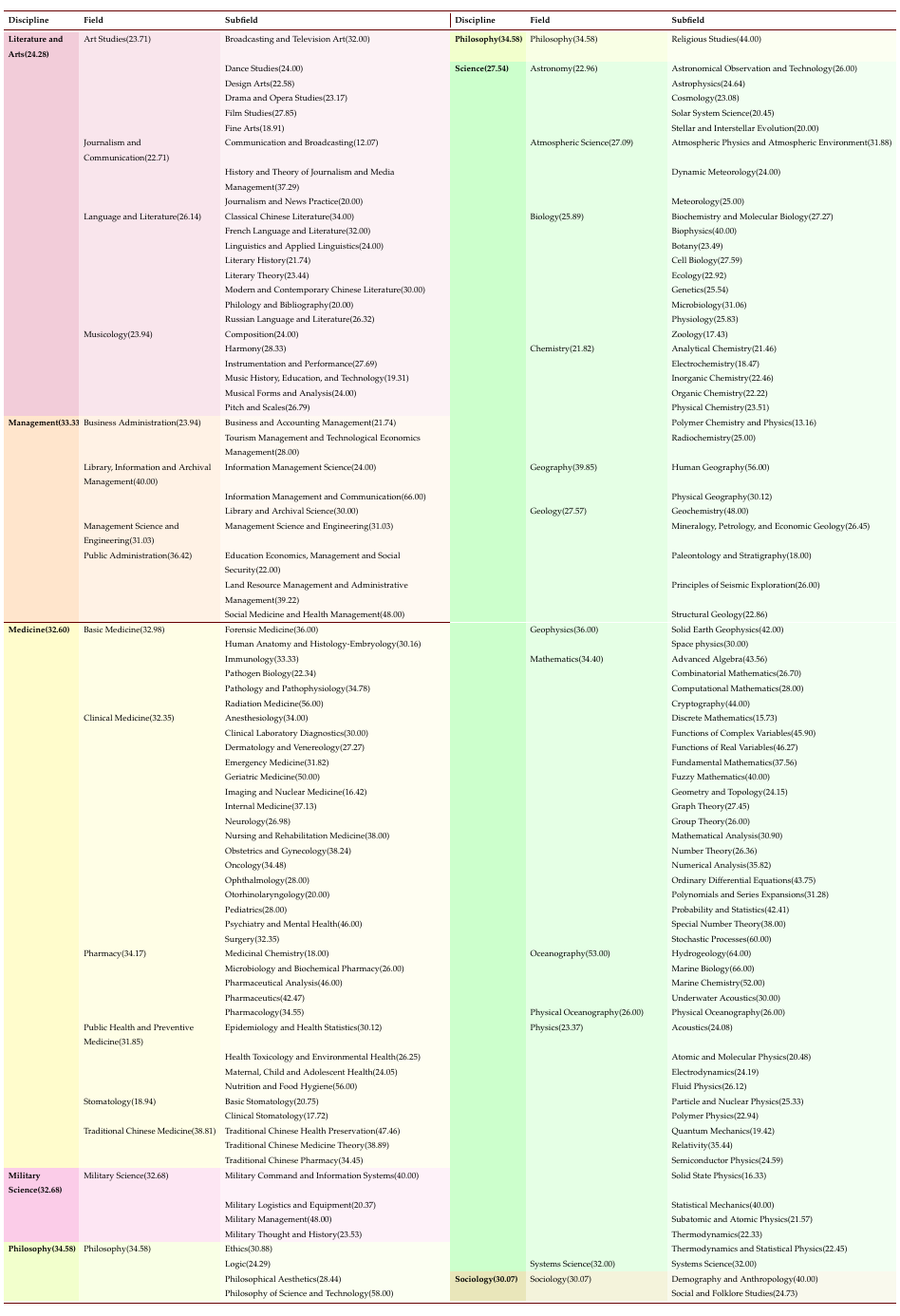} 
    \end{subtable}
    \vspace{-1.1cm}
    \captionsetup{font=small}
    \caption{Continued: Model Scores Across Three Levels of Disciplines: Qwen2.5-7B-Instruct.}
    \vspace{-0.6cm}
    \centeredlinks{listofmodels}{Back to List of Models}{toc}{Back to Table of Contents}{blue}
    \end{table}
}
\clearpage

\newpage
\vspace{-0.5cm}
\begin{table}[t]
\refstepcounter{models}%
\addcontentsline{csf}{models}{\protect\numberline{\themodels}Yi-1.5-34B}
\centering
\begin{subtable}[t]{1\textwidth}
\centering
\includegraphics[width=\textwidth]{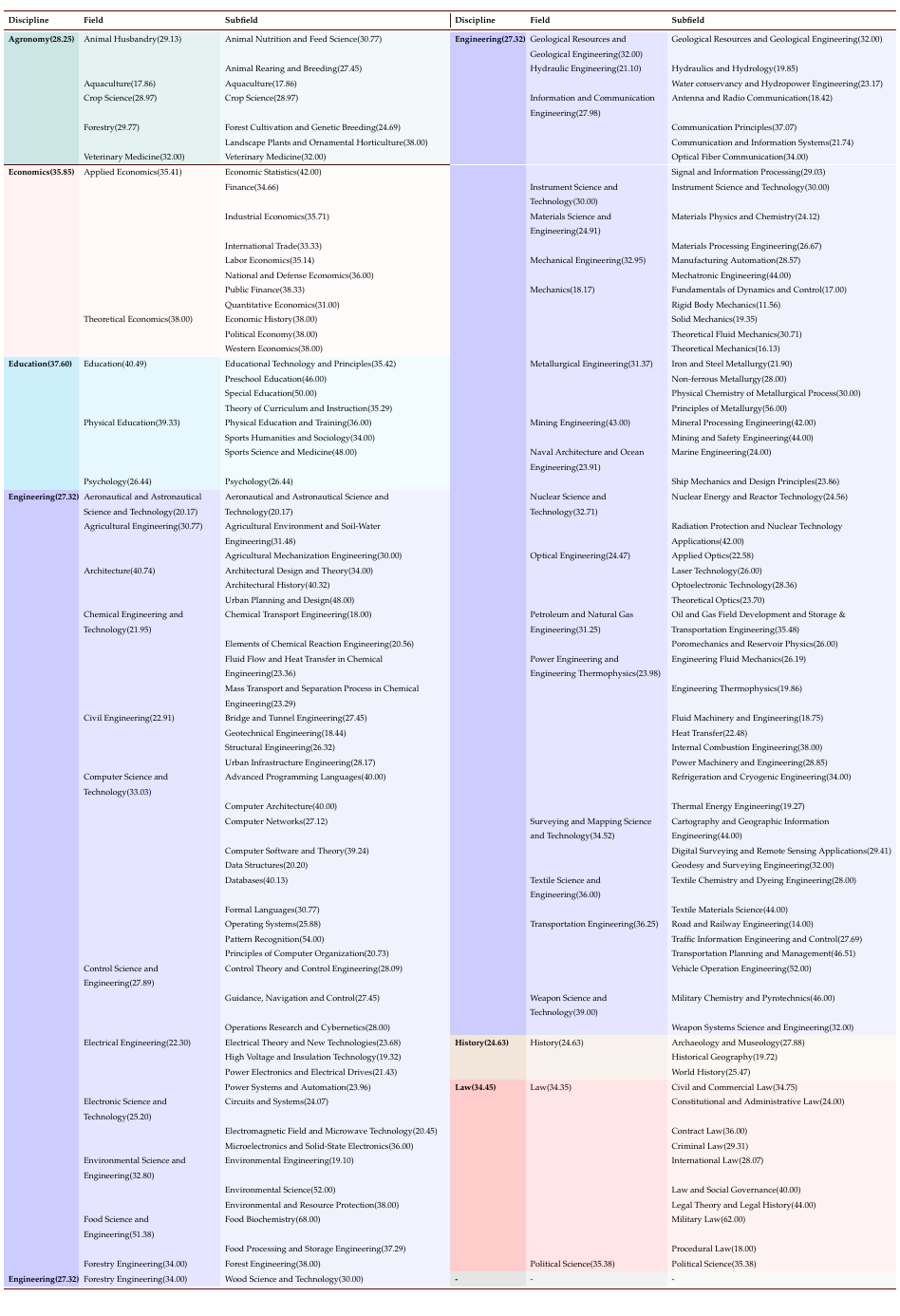} 
\end{subtable}
\vspace{-1.3cm}
\captionsetup{font=small}
\caption{Model Scores Across Three Levels of Disciplines: Yi-1.5-34B.}
\label{tab:Yi-1.5-34B}
\vspace{-0.5cm}
\centeredlinks{listofmodels}{Back to List of Models}{toc}{Back to Table of Contents}{blue}
\end{table}
\clearpage

\newpage
\afterpage{
    \begin{table}[t]
    \centering
    \ContinuedFloat 
    \begin{subtable}[t]{\textwidth}
    \centering
    \includegraphics[width=\textwidth]{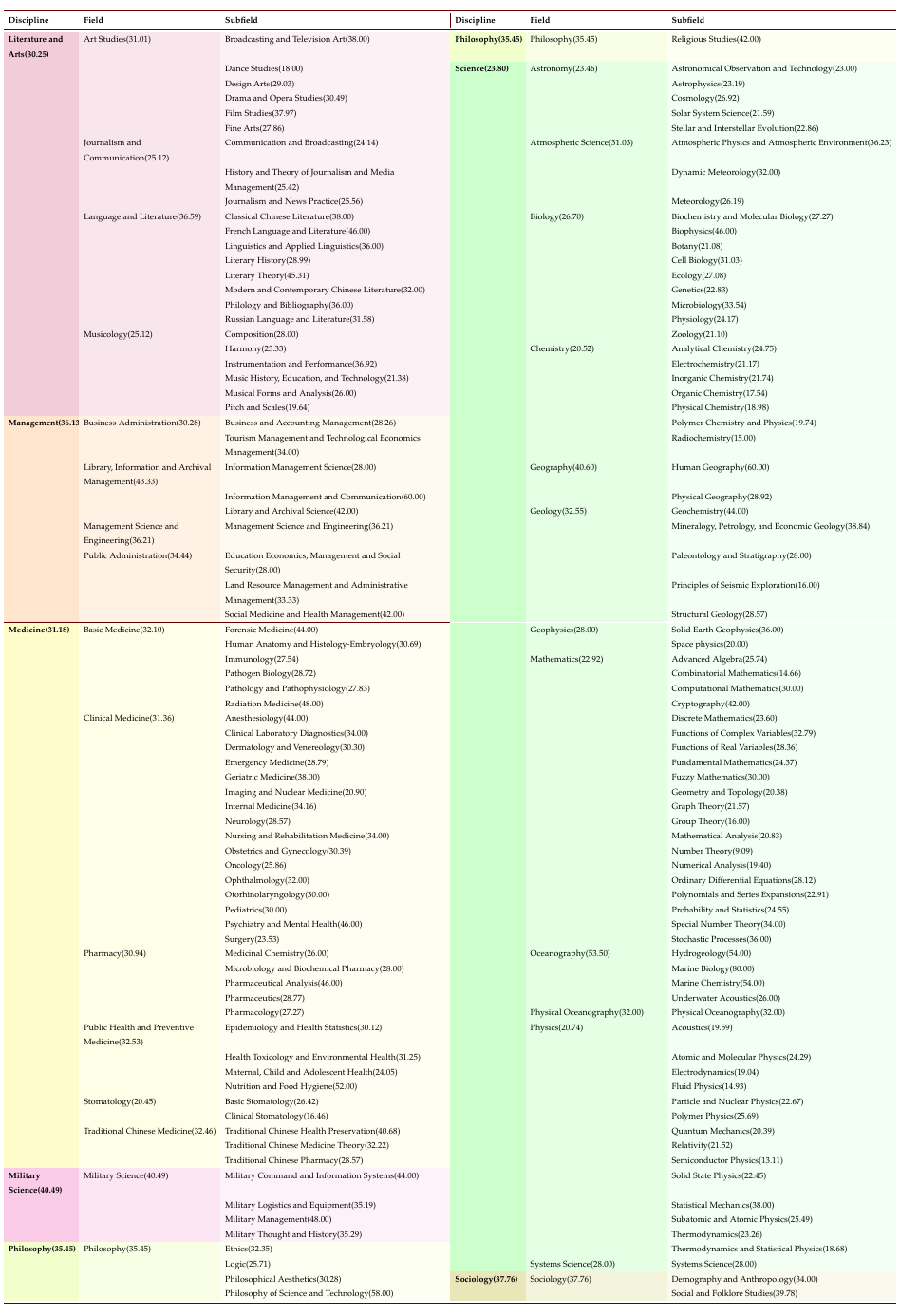} 
    \end{subtable}
    \vspace{-1.1cm}
    \captionsetup{font=small}
    \caption{Continued: Model Scores Across Three Levels of Disciplines: Yi-1.5-34B.}
    \vspace{-0.6cm}
    \centeredlinks{listofmodels}{Back to List of Models}{toc}{Back to Table of Contents}{blue}
    \end{table}
}
\clearpage

\newpage
\vspace{-0.5cm}
\begin{table}[t]
\refstepcounter{models}%
\addcontentsline{csf}{models}{\protect\numberline{\themodels}gemma-2-27b-it}
\centering
\begin{subtable}[t]{1\textwidth}
\centering
\includegraphics[width=\textwidth]{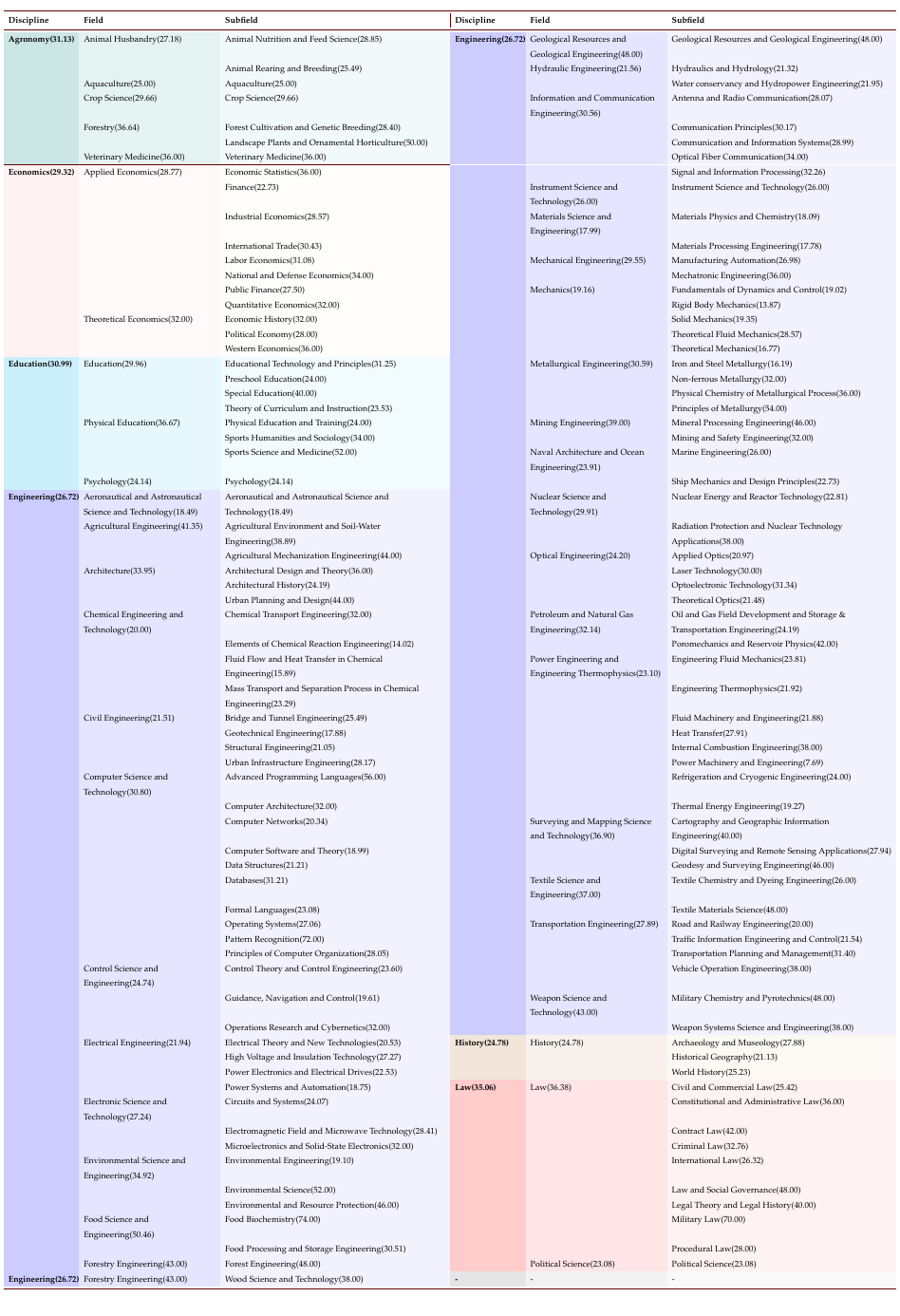} 
\end{subtable}
\vspace{-1.3cm}
\captionsetup{font=small}
\caption{Model Scores Across Three Levels of Disciplines: gemma-2-27b-it.}
\label{tab:gemma-2-27b-it}
\vspace{-0.5cm}
\centeredlinks{listofmodels}{Back to List of Models}{toc}{Back to Table of Contents}{blue}
\end{table}
\clearpage

\newpage
\afterpage{
    \begin{table}[t]
    \centering
    \ContinuedFloat 
    \begin{subtable}[t]{\textwidth}
    \centering
    \includegraphics[width=\textwidth]{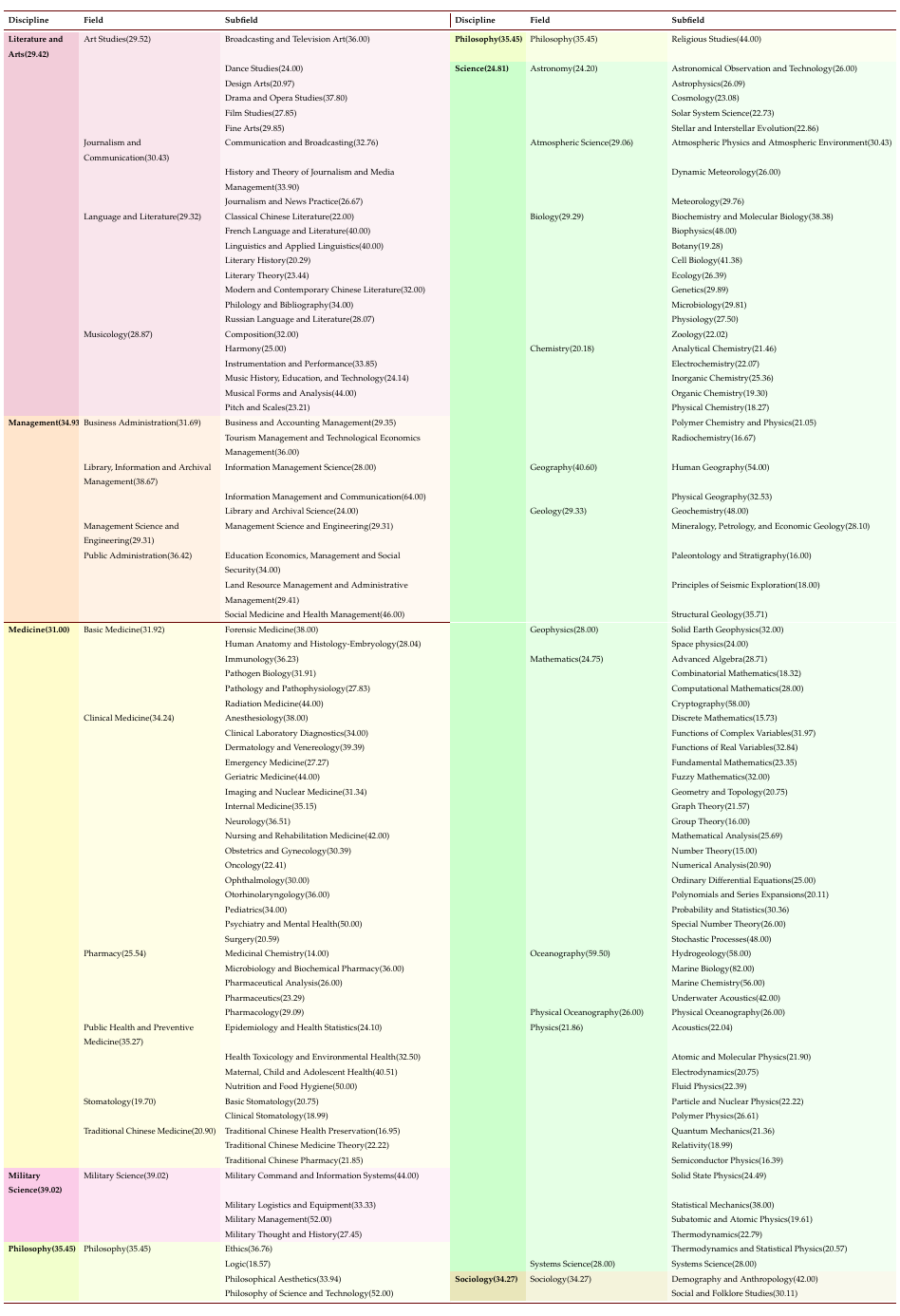} 
    \end{subtable}
    \vspace{-1.1cm}
    \captionsetup{font=small}
    \caption{Continued: Model Scores Across Three Levels of Disciplines: gemma-2-27b-it.}
    \vspace{-0.6cm}
    \centeredlinks{listofmodels}{Back to List of Models}{toc}{Back to Table of Contents}{blue}
    \end{table}
}
\clearpage

\newpage
\vspace{-0.5cm}
\begin{table}[t]
\refstepcounter{models}%
\addcontentsline{csf}{models}{\protect\numberline{\themodels}Llama-3.1-70B}
\centering
\begin{subtable}[t]{1\textwidth}
\centering
\includegraphics[width=\textwidth]{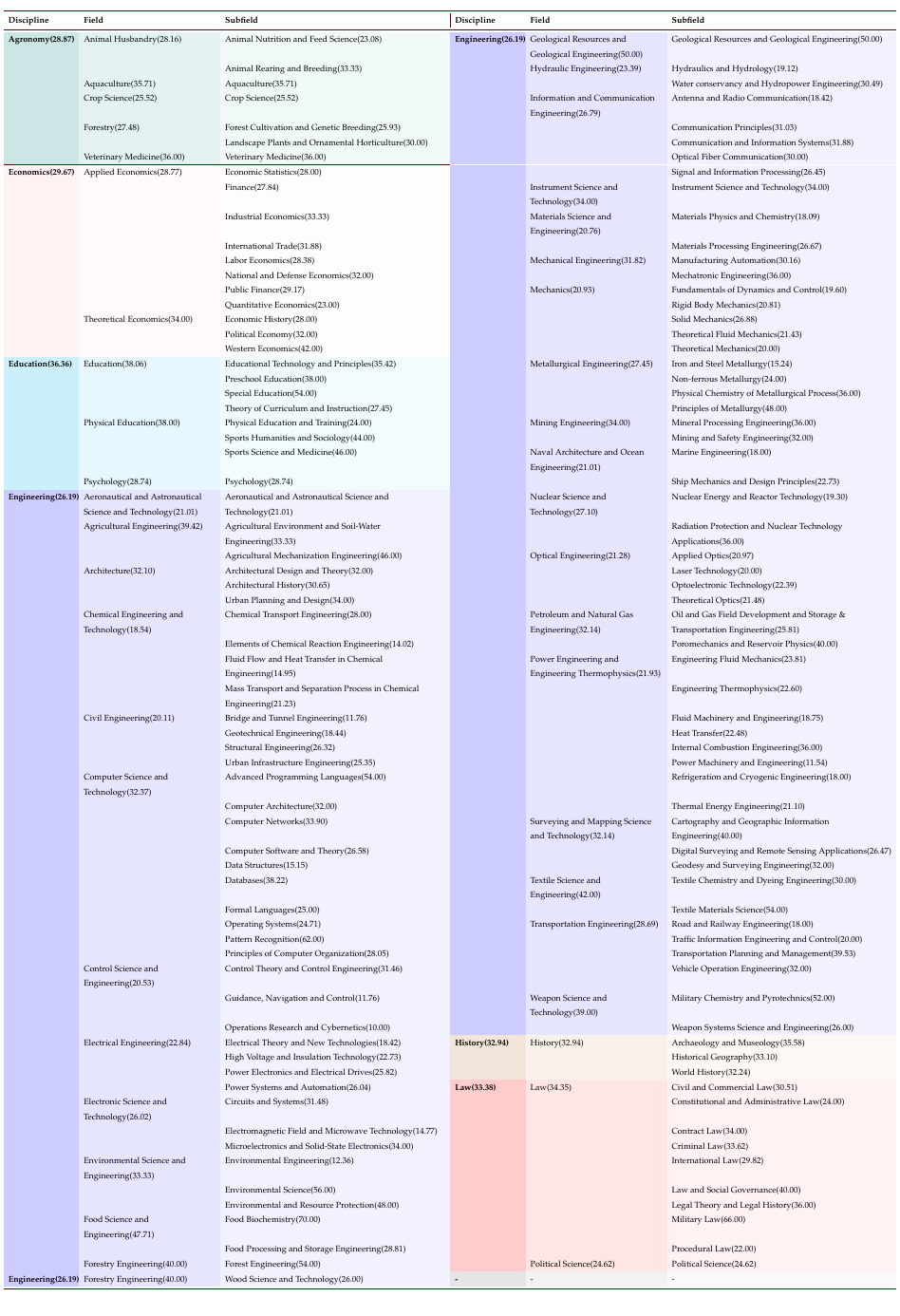} 
\end{subtable}
\vspace{-1.3cm}
\captionsetup{font=small}
\caption{Model Scores Across Three Levels of Disciplines: Llama-3.1-70B.}
\label{tab:Llama-3.1-70B}
\vspace{-0.5cm}
\centeredlinks{listofmodels}{Back to List of Models}{toc}{Back to Table of Contents}{blue}
\end{table}
\clearpage

\newpage
\afterpage{
    \begin{table}[t]
    \centering
    \ContinuedFloat 
    \begin{subtable}[t]{\textwidth}
    \centering
    \includegraphics[width=\textwidth]{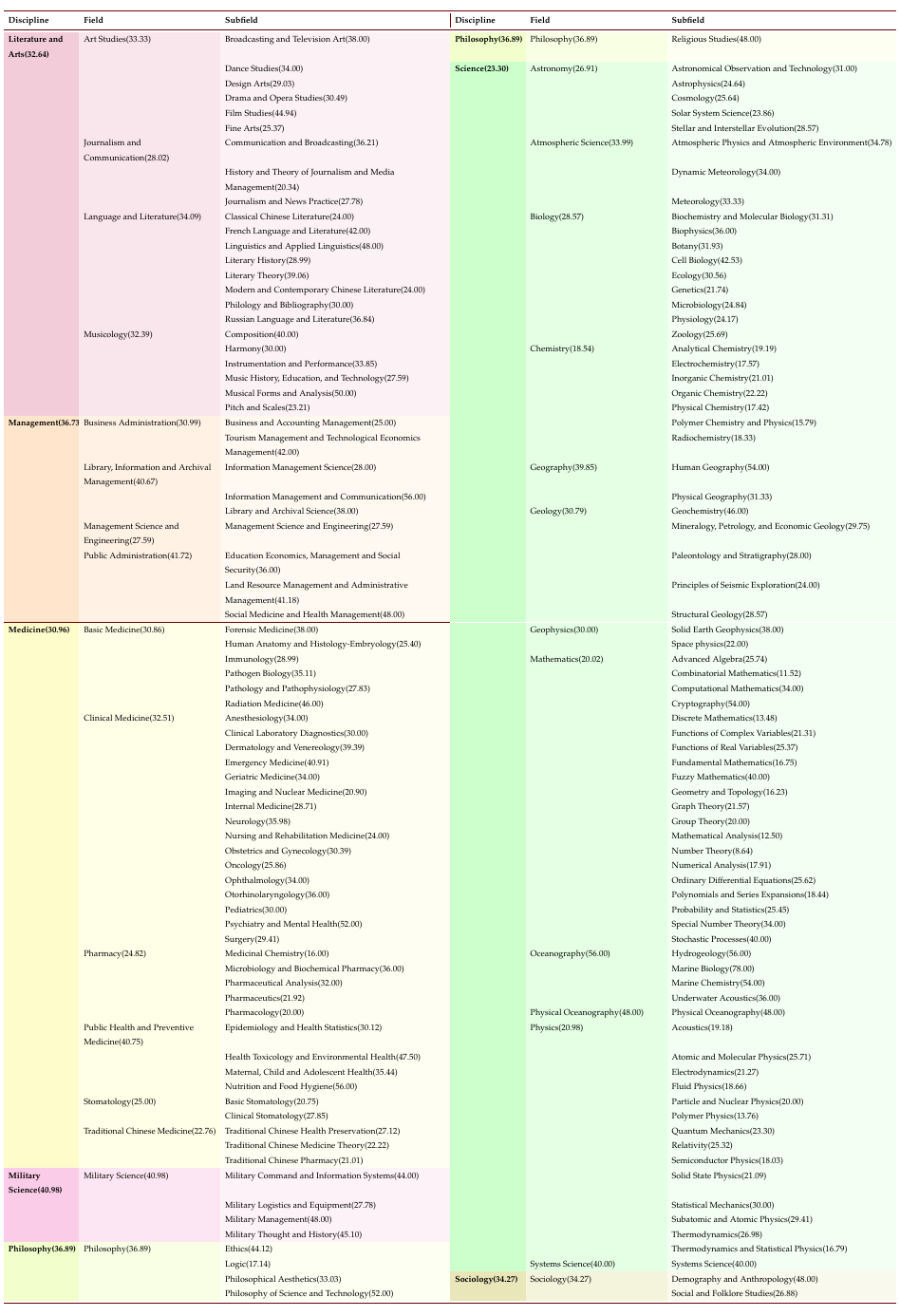} 
    \end{subtable}
    \vspace{-1.1cm}
    \captionsetup{font=small}
    \caption{Continued: Model Scores Across Three Levels of Disciplines: Llama-3.1-70B.}
    \vspace{-0.6cm}
    \centeredlinks{listofmodels}{Back to List of Models}{toc}{Back to Table of Contents}{blue}
    \end{table}
}
\clearpage

\newpage
\vspace{-0.5cm}
\begin{table}[t]
\refstepcounter{models}%
\addcontentsline{csf}{models}{\protect\numberline{\themodels}Yi-1.5-34B-Chat}
\centering
\begin{subtable}[t]{1\textwidth}
\centering
\includegraphics[width=\textwidth]{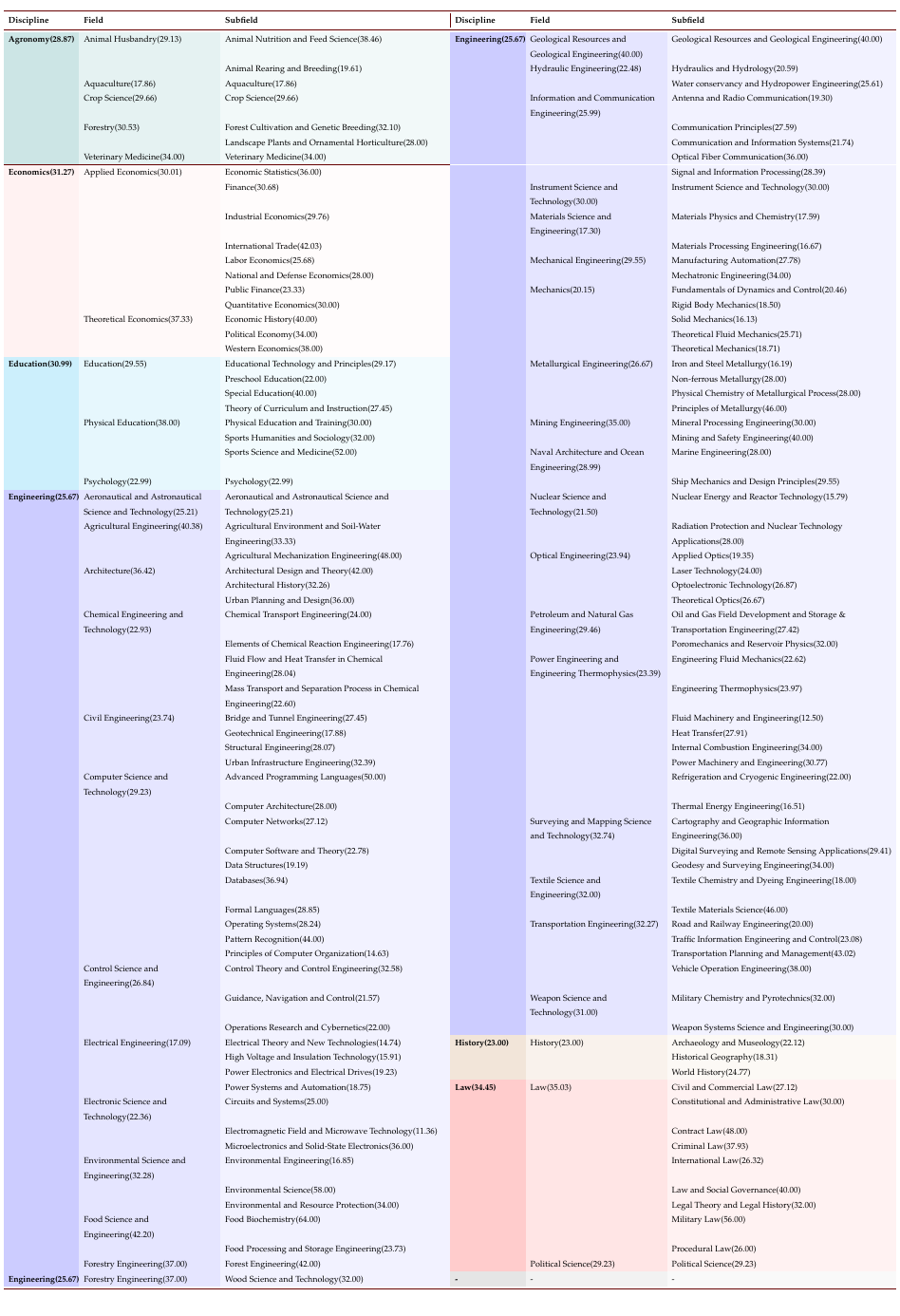} 
\end{subtable}
\vspace{-1.3cm}
\captionsetup{font=small}
\caption{Model Scores Across Three Levels of Disciplines: Yi-1.5-34B-Chat.}
\label{tab:Yi-1.5-34B-Chat}
\vspace{-0.5cm}
\centeredlinks{listofmodels}{Back to List of Models}{toc}{Back to Table of Contents}{blue}
\end{table}
\clearpage

\newpage
\afterpage{
    \begin{table}[t]
    \centering
    \ContinuedFloat 
    \begin{subtable}[t]{\textwidth}
    \centering
    \includegraphics[width=\textwidth]{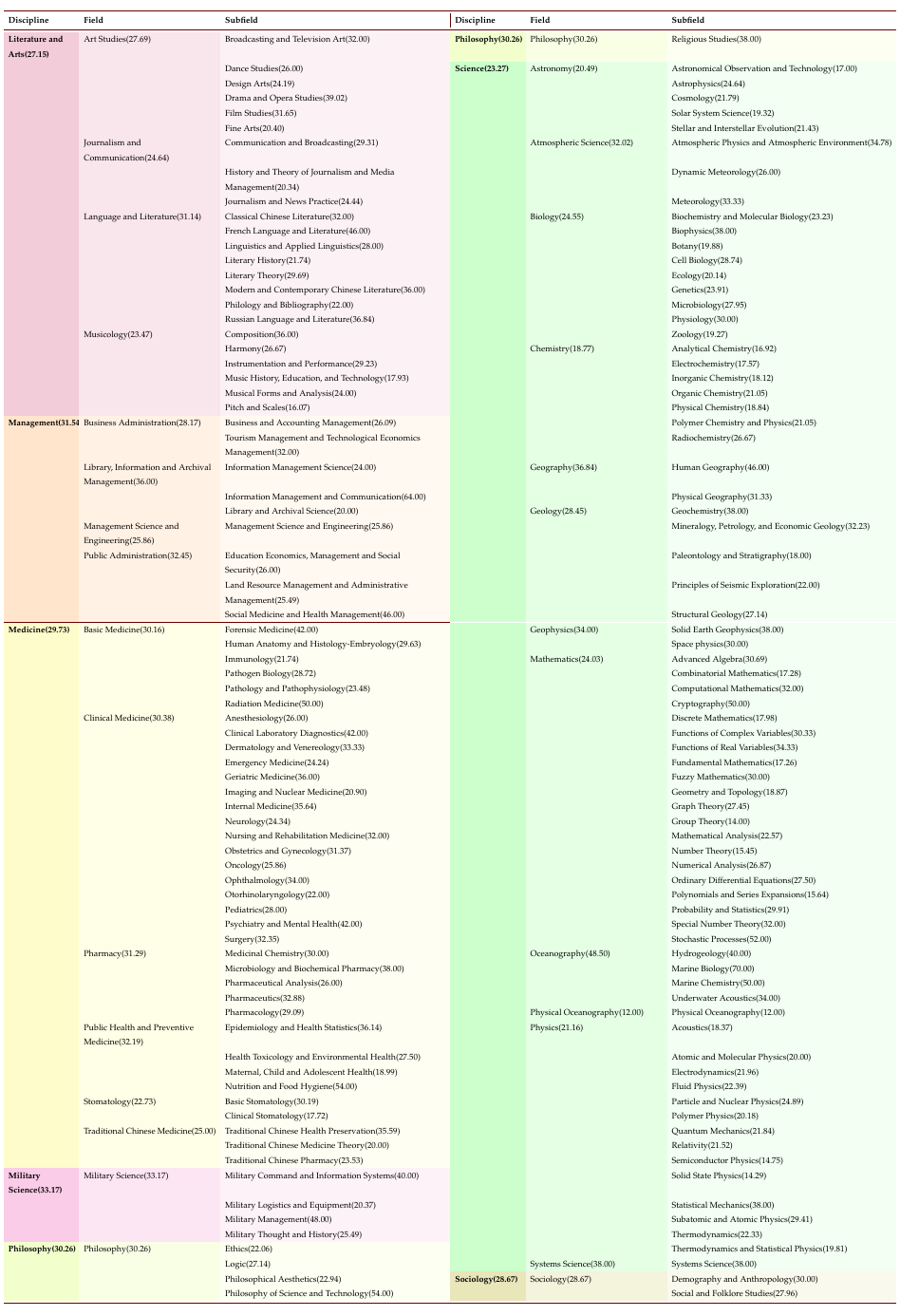} 
    \end{subtable}
    \vspace{-1.1cm}
    \captionsetup{font=small}
    \caption{Continued: Model Scores Across Three Levels of Disciplines: Yi-1.5-34B-Chat.}
    \vspace{-0.6cm}
    \centeredlinks{listofmodels}{Back to List of Models}{toc}{Back to Table of Contents}{blue}
    \end{table}
}
\clearpage

\newpage
\vspace{-0.5cm}
\begin{table}[t]
\refstepcounter{models}%
\addcontentsline{csf}{models}{\protect\numberline{\themodels}Mistral-Small-Instruct-2409}
\centering
\begin{subtable}[t]{1\textwidth}
\centering
\includegraphics[width=\textwidth]{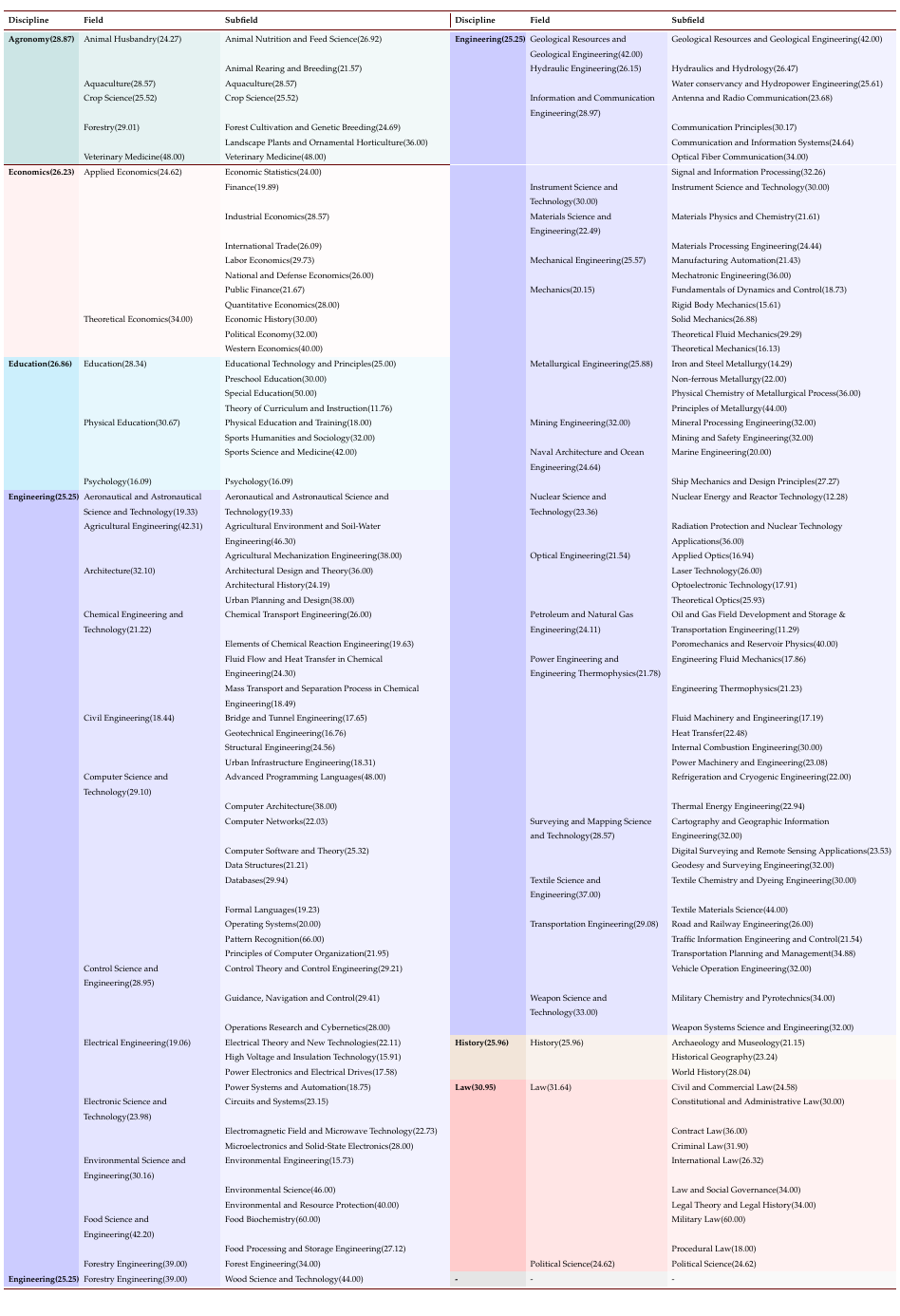} 
\end{subtable}
\vspace{-1.3cm}
\captionsetup{font=small}
\caption{Model Scores Across Three Levels of Disciplines: Mistral-Small-Instruct-2409.}
\label{tab:Mistral-Small-Instruct-2409}
\vspace{-0.5cm}
\centeredlinks{listofmodels}{Back to List of Models}{toc}{Back to Table of Contents}{blue}
\end{table}
\clearpage

\newpage
\afterpage{
    \begin{table}[t]
    \centering
    \ContinuedFloat 
    \begin{subtable}[t]{\textwidth}
    \centering
    \includegraphics[width=\textwidth]{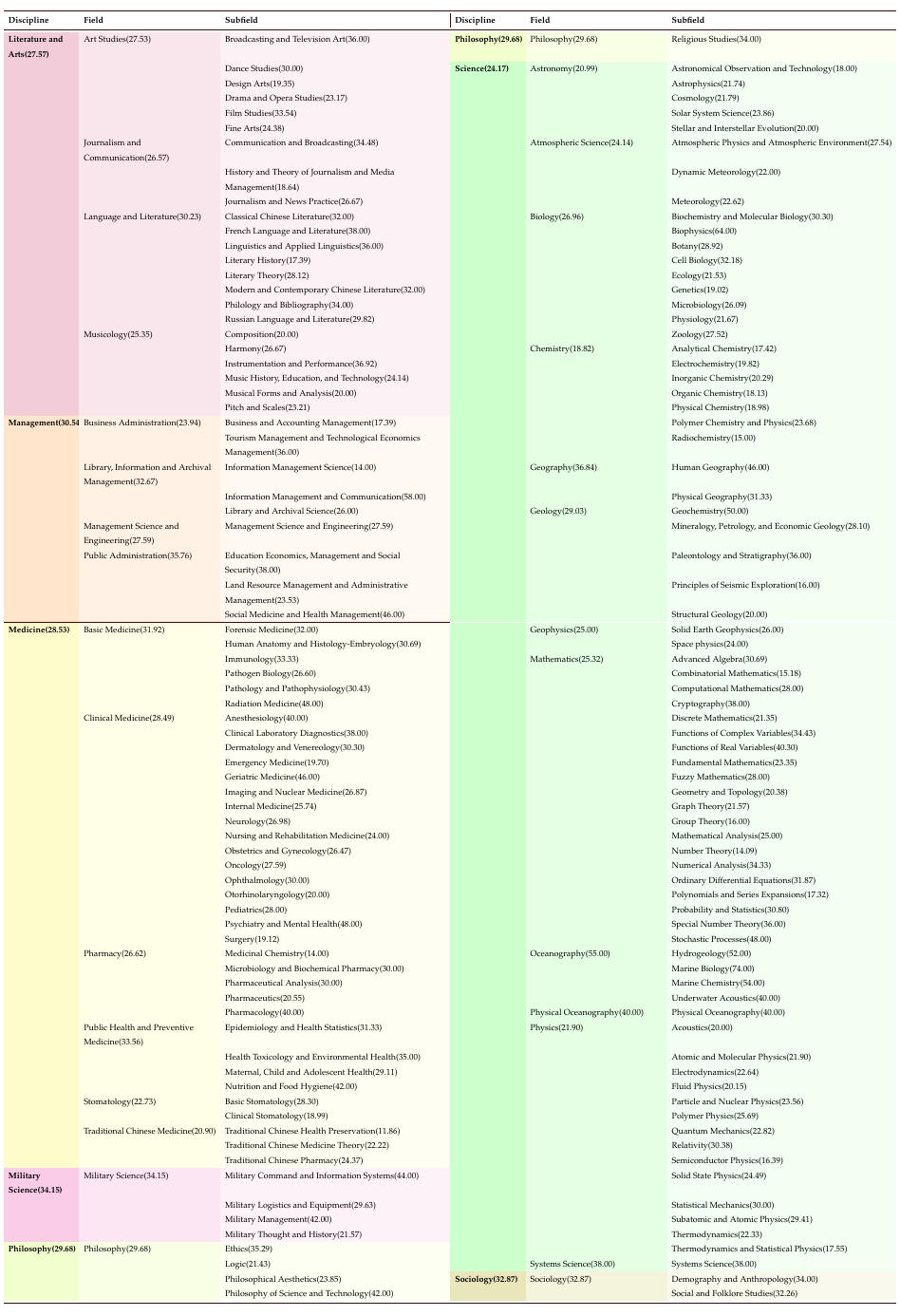} 
    \end{subtable}
    \vspace{-1.1cm}
    \captionsetup{font=small}
    \caption{Continued: Model Scores Across Three Levels of Disciplines: Mistral-Small-Instruct-2409.}
    \vspace{-0.6cm}
    \centeredlinks{listofmodels}{Back to List of Models}{toc}{Back to Table of Contents}{blue}
    \end{table}
}
\clearpage

\newpage
\vspace{-0.5cm}
\begin{table}[t]
\refstepcounter{models}%
\addcontentsline{csf}{models}{\protect\numberline{\themodels}Qwen2.5-7B}
\centering
\begin{subtable}[t]{1\textwidth}
\centering
\includegraphics[width=\textwidth]{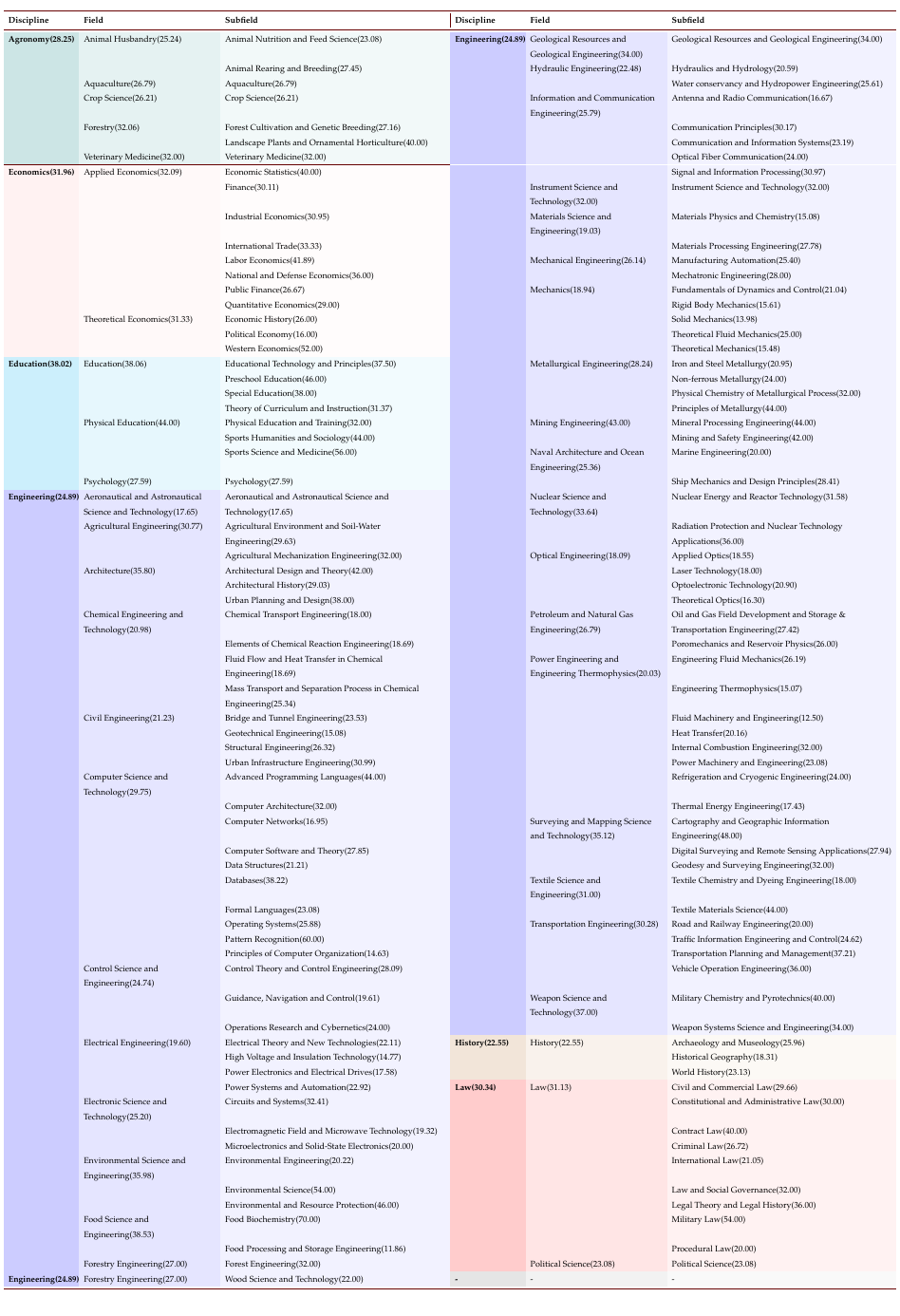} 
\end{subtable}
\vspace{-1.3cm}
\captionsetup{font=small}
\caption{Model Scores Across Three Levels of Disciplines: Qwen2.5-7B.}
\label{tab:Qwen2.5-7B}
\vspace{-0.5cm}
\centeredlinks{listofmodels}{Back to List of Models}{toc}{Back to Table of Contents}{blue}
\end{table}
\clearpage

\newpage
\afterpage{
    \begin{table}[t]
    \centering
    \ContinuedFloat 
    \begin{subtable}[t]{\textwidth}
    \centering
    \includegraphics[width=\textwidth]{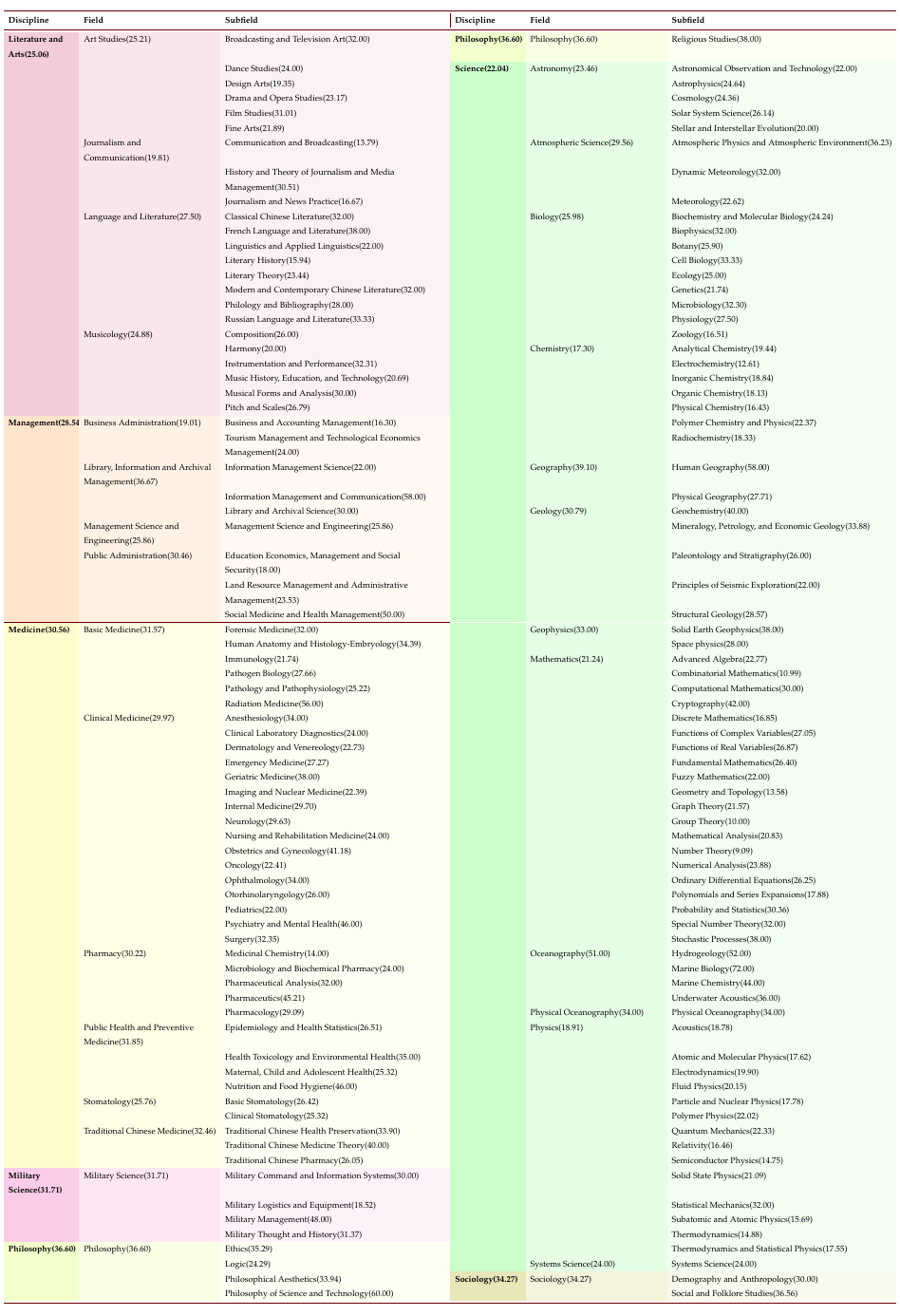} 
    \end{subtable}
    \vspace{-1.1cm}
    \captionsetup{font=small}
    \caption{Continued: Model Scores Across Three Levels of Disciplines: Qwen2.5-7B.}
    \vspace{-0.6cm}
    \centeredlinks{listofmodels}{Back to List of Models}{toc}{Back to Table of Contents}{blue}
    \end{table}
}
\clearpage

\newpage
\vspace{-0.5cm}
\begin{table}[t]
\refstepcounter{models}%
\addcontentsline{csf}{models}{\protect\numberline{\themodels}Llama-3.1-405B}
\centering
\begin{subtable}[t]{1\textwidth}
\centering
\includegraphics[width=\textwidth]{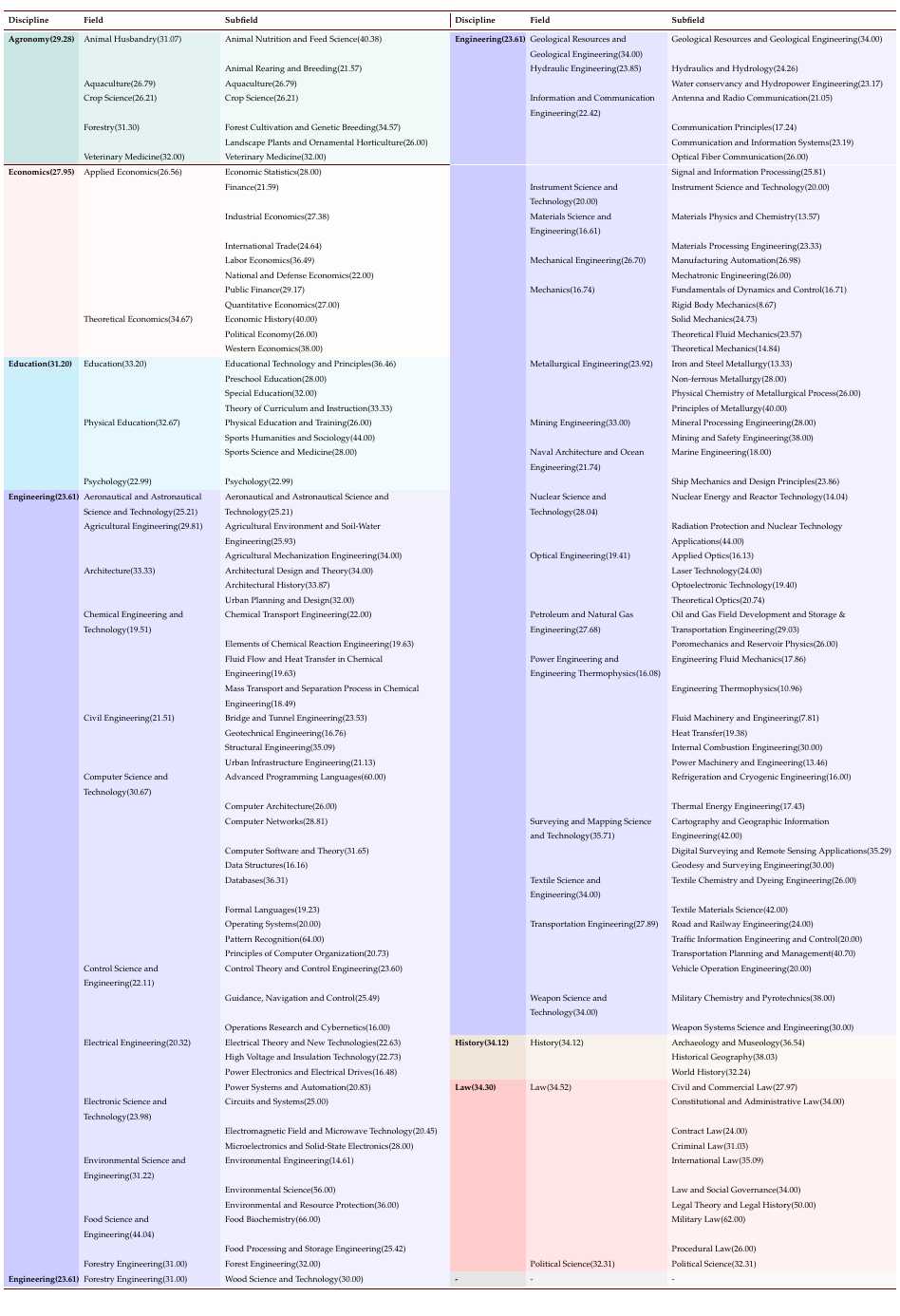} 
\end{subtable}
\vspace{-1.3cm}
\captionsetup{font=small}
\caption{Model Scores Across Three Levels of Disciplines: Llama-3.1-405B.}
\label{tab:Llama-3.1-405B}
\vspace{-0.5cm}
\centeredlinks{listofmodels}{Back to List of Models}{toc}{Back to Table of Contents}{blue}
\end{table}
\clearpage

\newpage
\afterpage{
    \begin{table}[t]
    \centering
    \ContinuedFloat 
    \begin{subtable}[t]{\textwidth}
    \centering
    \includegraphics[width=\textwidth]{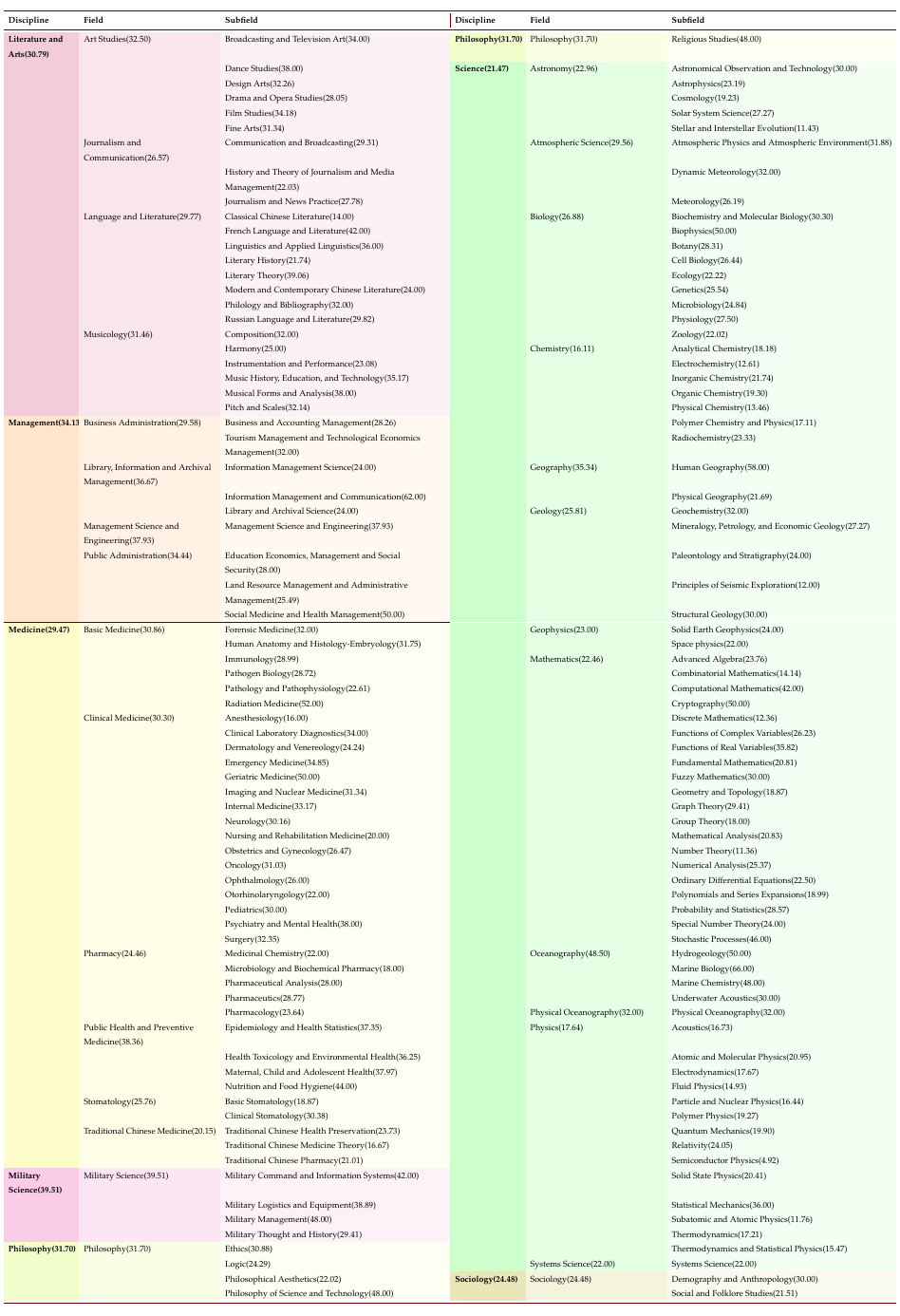} 
    \end{subtable}
    \vspace{-1.1cm}
    \captionsetup{font=small}
    \caption{Continued: Model Scores Across Three Levels of Disciplines: Llama-3.1-405B.}
    \vspace{-0.6cm}
    \centeredlinks{listofmodels}{Back to List of Models}{toc}{Back to Table of Contents}{blue}
    \end{table}
}
\clearpage

\newpage
\vspace{-0.5cm}
\begin{table}[t]
\refstepcounter{models}%
\addcontentsline{csf}{models}{\protect\numberline{\themodels}gemma-2-27b}
\centering
\begin{subtable}[t]{1\textwidth}
\centering
\includegraphics[width=\textwidth]{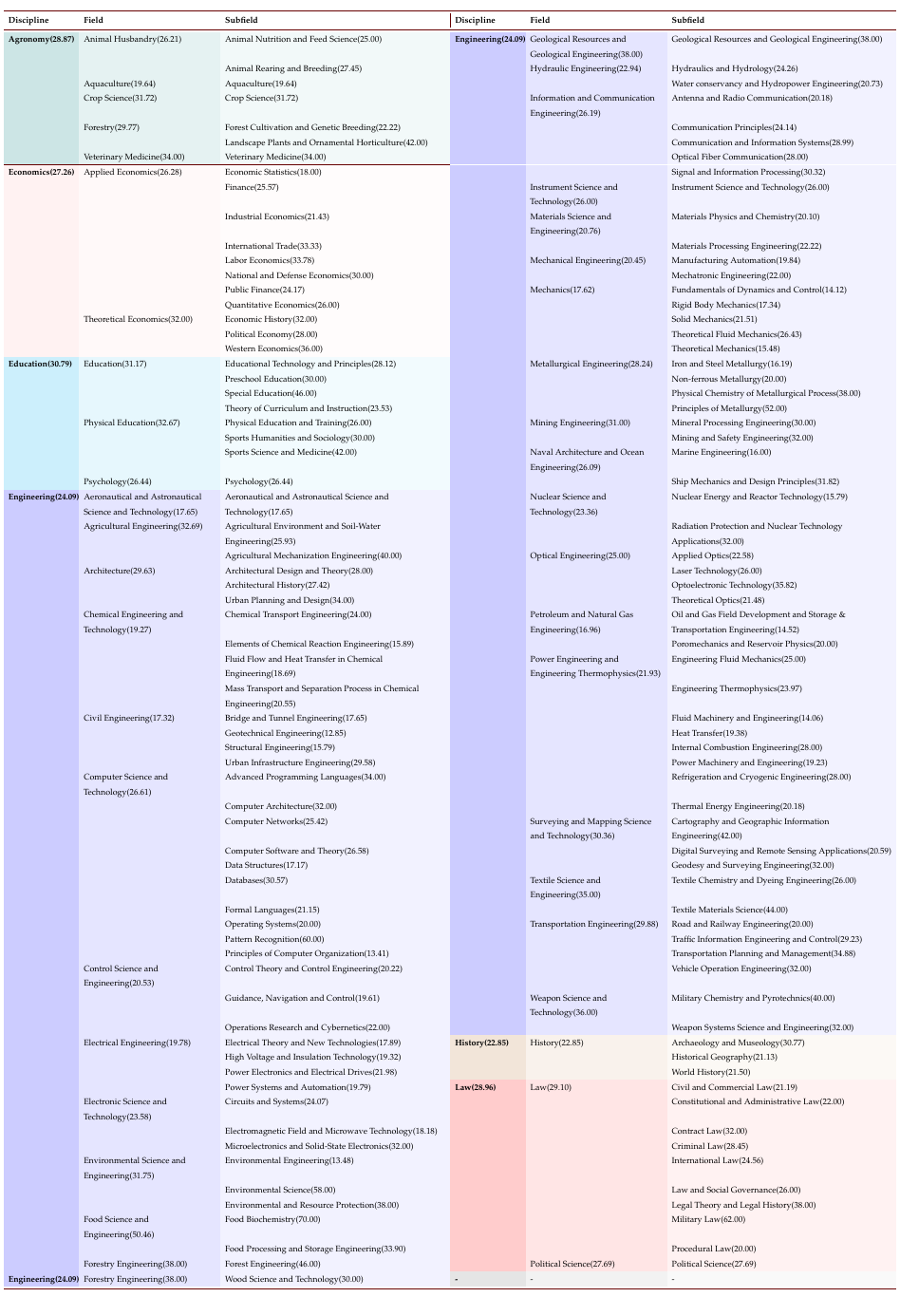} 
\end{subtable}
\vspace{-1.3cm}
\captionsetup{font=small}
\caption{Model Scores Across Three Levels of Disciplines: gemma-2-27b.}
\label{tab:gemma-2-27b}
\vspace{-0.5cm}
\centeredlinks{listofmodels}{Back to List of Models}{toc}{Back to Table of Contents}{blue}
\end{table}
\clearpage

\newpage
\afterpage{
    \begin{table}[t]
    \centering
    \ContinuedFloat 
    \begin{subtable}[t]{\textwidth}
    \centering
    \includegraphics[width=\textwidth]{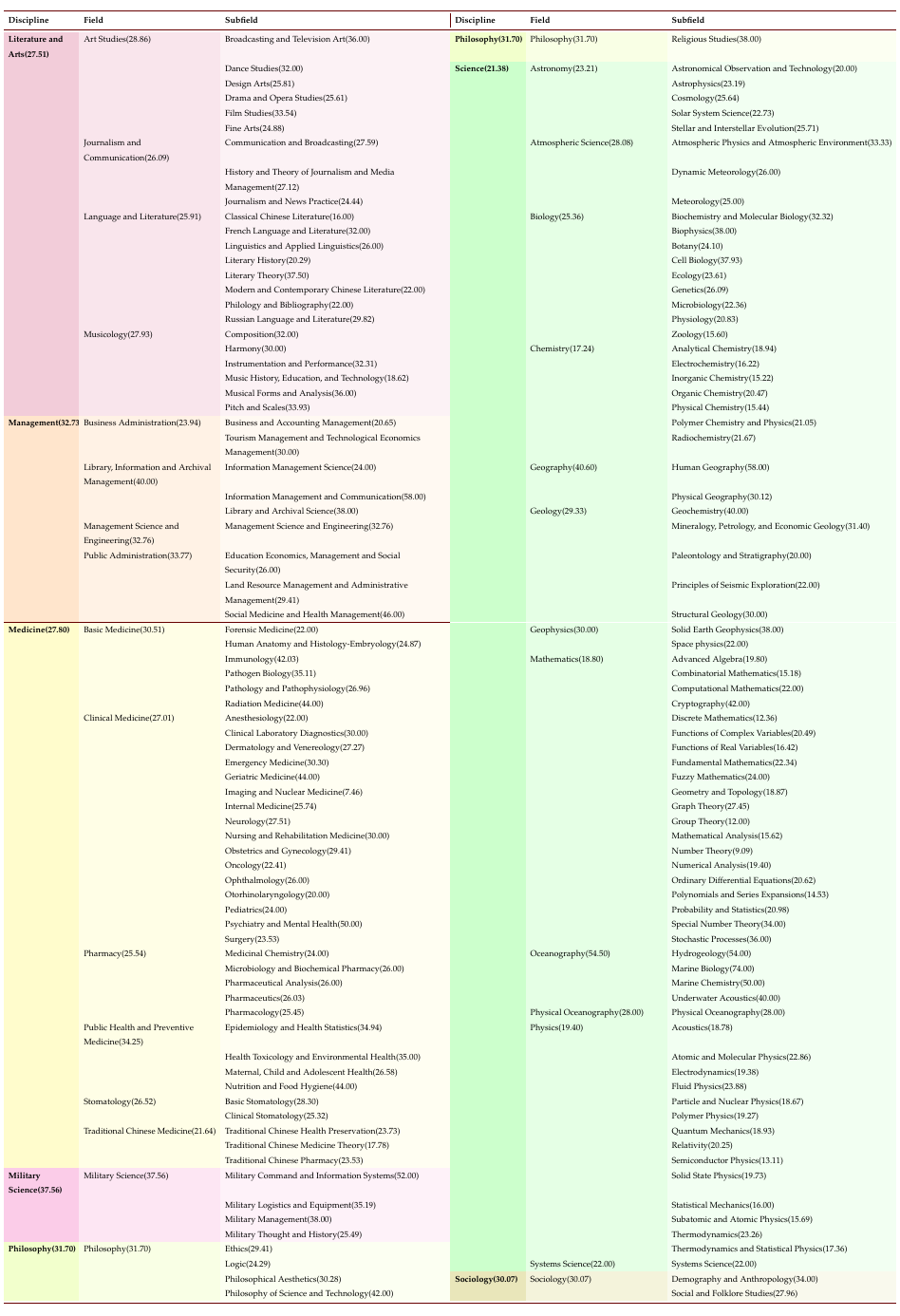} 
    \end{subtable}
    \vspace{-1.1cm}
    \captionsetup{font=small}
    \caption{Continued: Model Scores Across Three Levels of Disciplines: gemma-2-27b.}
    \vspace{-0.6cm}
    \centeredlinks{listofmodels}{Back to List of Models}{toc}{Back to Table of Contents}{blue}
    \end{table}
}
\clearpage

\newpage
\vspace{-0.5cm}
\begin{table}[t]
\refstepcounter{models}%
\addcontentsline{csf}{models}{\protect\numberline{\themodels}gemma-2-9b-it}
\centering
\begin{subtable}[t]{1\textwidth}
\centering
\includegraphics[width=\textwidth]{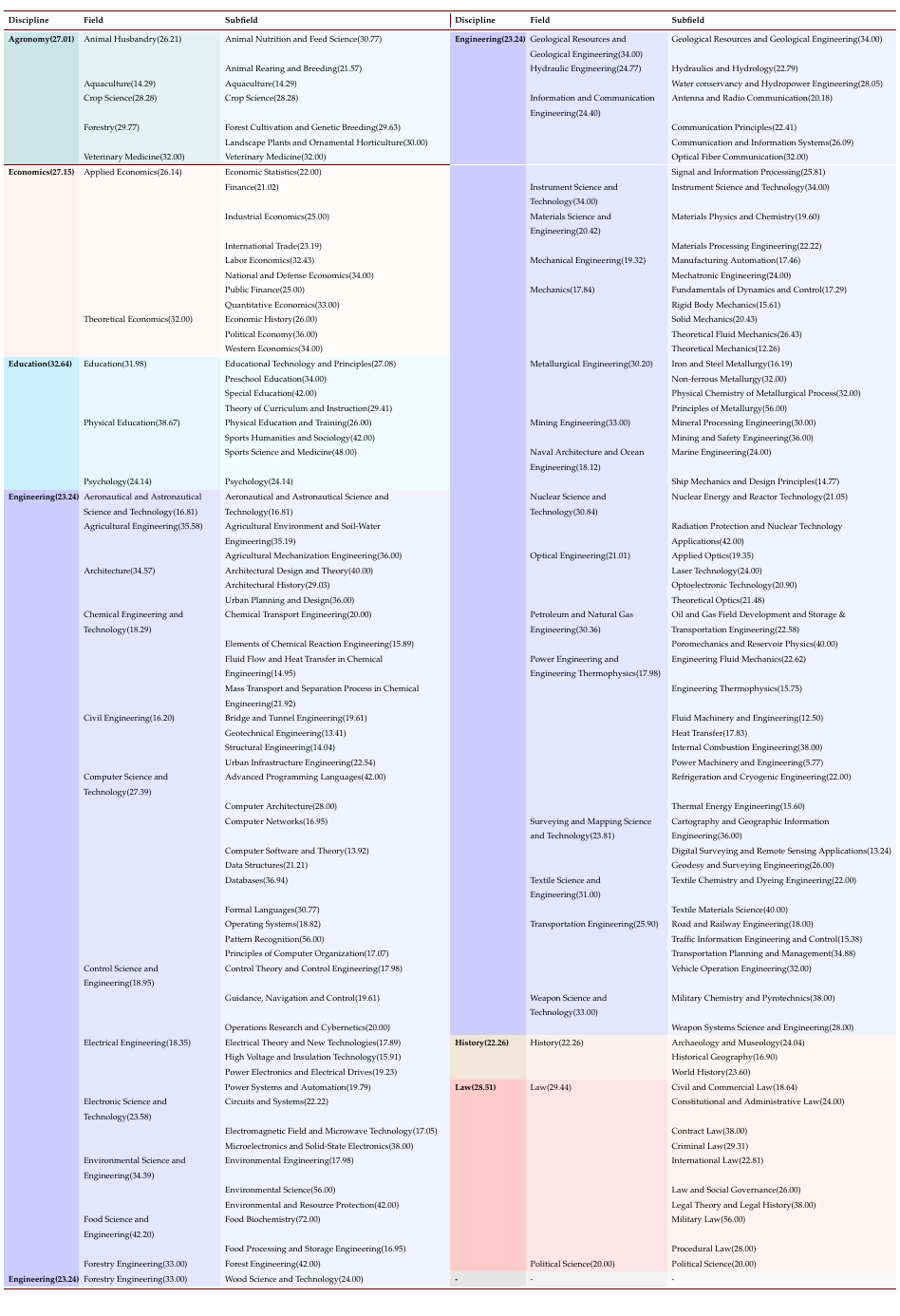} 
\end{subtable}
\vspace{-1.3cm}
\captionsetup{font=small}
\caption{Model Scores Across Three Levels of Disciplines: gemma-2-9b-it.}
\label{tab:gemma-2-9b-it}
\vspace{-0.5cm}
\centeredlinks{listofmodels}{Back to List of Models}{toc}{Back to Table of Contents}{blue}
\end{table}
\clearpage

\newpage
\afterpage{
    \begin{table}[t]
    \centering
    \ContinuedFloat 
    \begin{subtable}[t]{\textwidth}
    \centering
    \includegraphics[width=\textwidth]{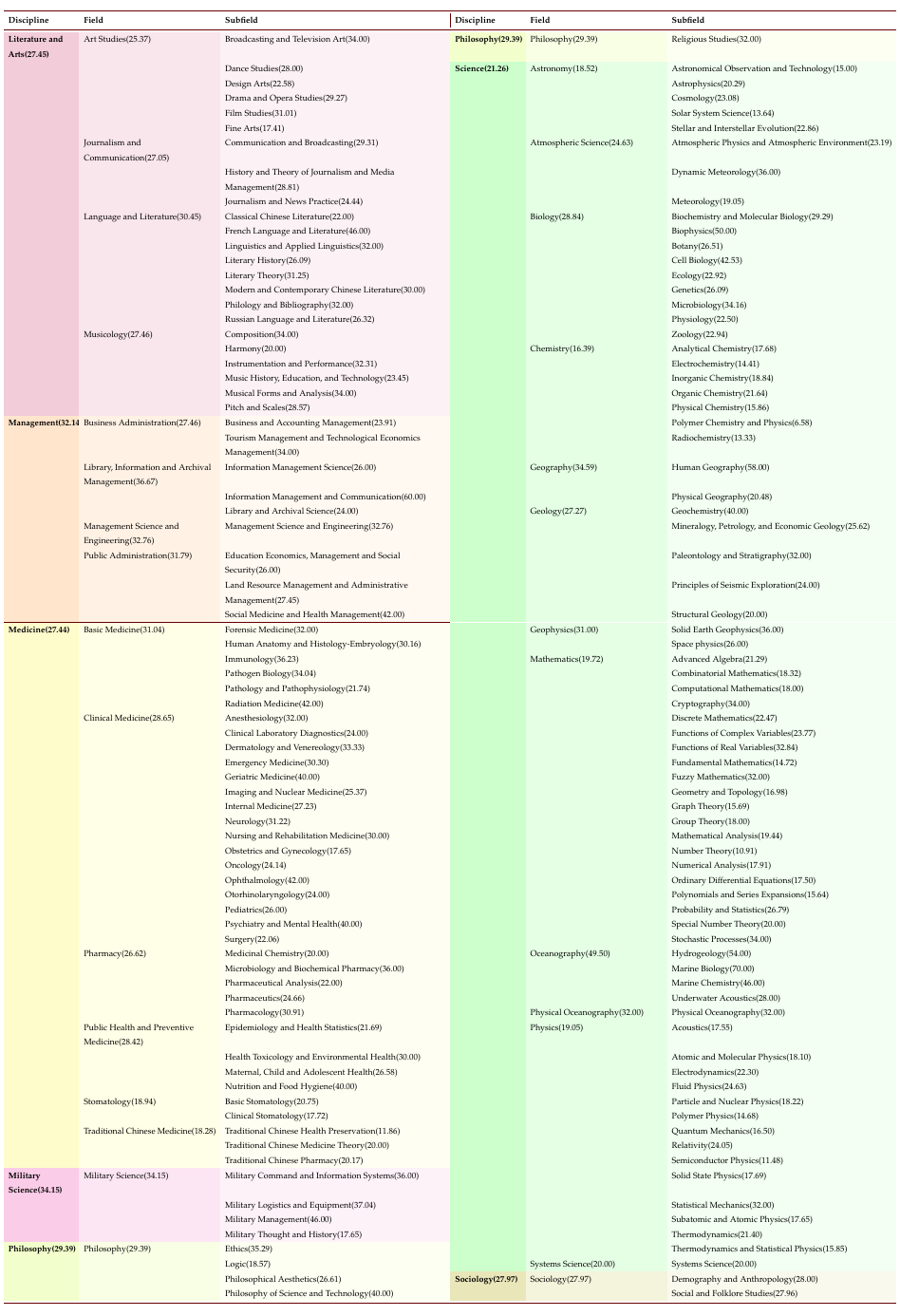} 
    \end{subtable}
    \vspace{-1.1cm}
    \captionsetup{font=small}
    \caption{Continued: Model Scores Across Three Levels of Disciplines: gemma-2-9b-it.}
    \vspace{-0.6cm}
    \centeredlinks{listofmodels}{Back to List of Models}{toc}{Back to Table of Contents}{blue}
    \end{table}
}
\clearpage

\newpage
\vspace{-0.5cm}
\begin{table}[t]
\refstepcounter{models}%
\addcontentsline{csf}{models}{\protect\numberline{\themodels}Qwen2.5-3B-Instruct}
\centering
\begin{subtable}[t]{1\textwidth}
\centering
\includegraphics[width=\textwidth]{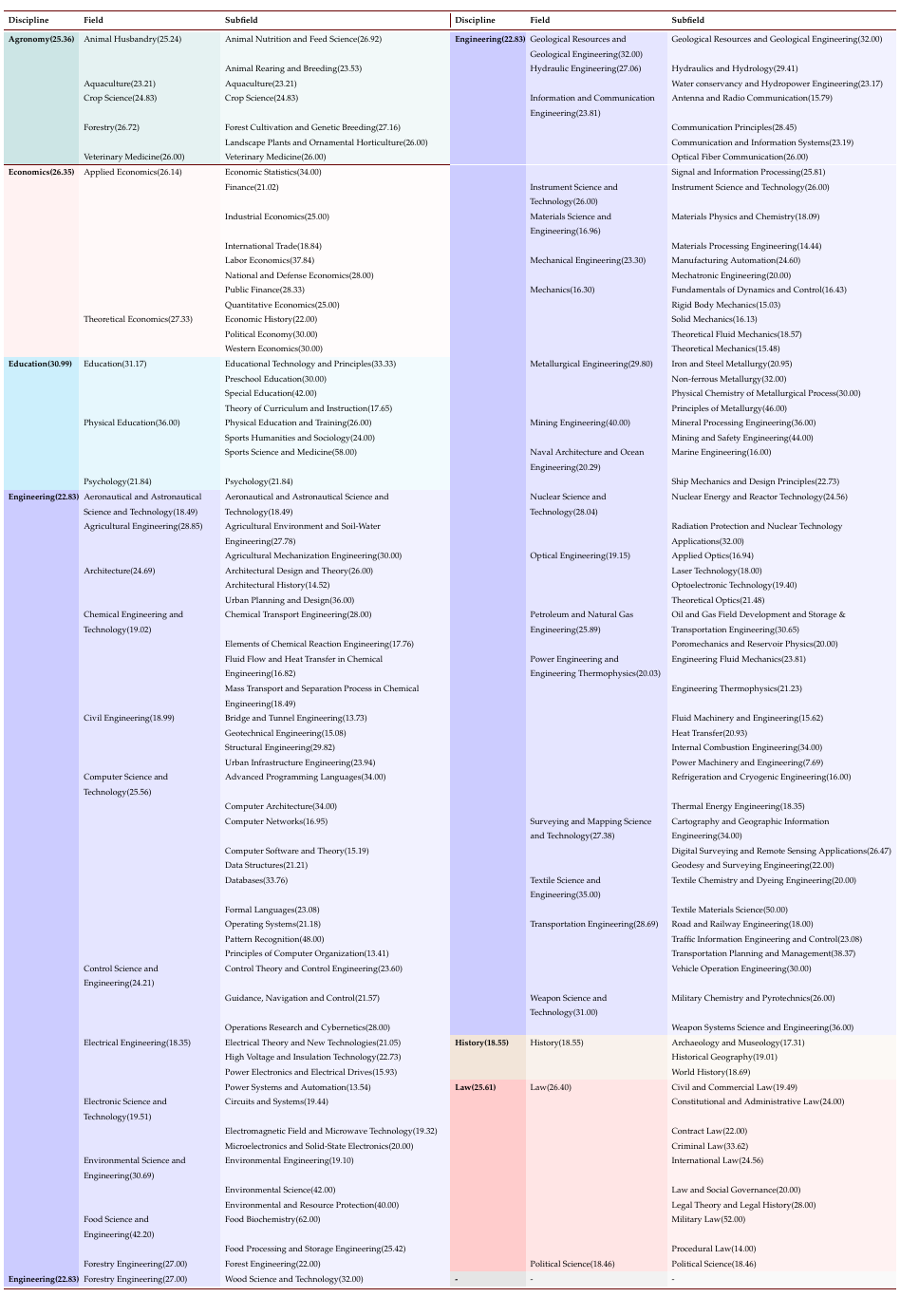} 
\end{subtable}
\vspace{-1.3cm}
\captionsetup{font=small}
\caption{Model Scores Across Three Levels of Disciplines: Qwen2.5-3B-Instruct.}
\label{tab:Qwen2.5-3B-Instruct}
\vspace{-0.5cm}
\centeredlinks{listofmodels}{Back to List of Models}{toc}{Back to Table of Contents}{blue}
\end{table}
\clearpage

\newpage
\afterpage{
    \begin{table}[t]
    \centering
    \ContinuedFloat 
    \begin{subtable}[t]{\textwidth}
    \centering
    \includegraphics[width=\textwidth]{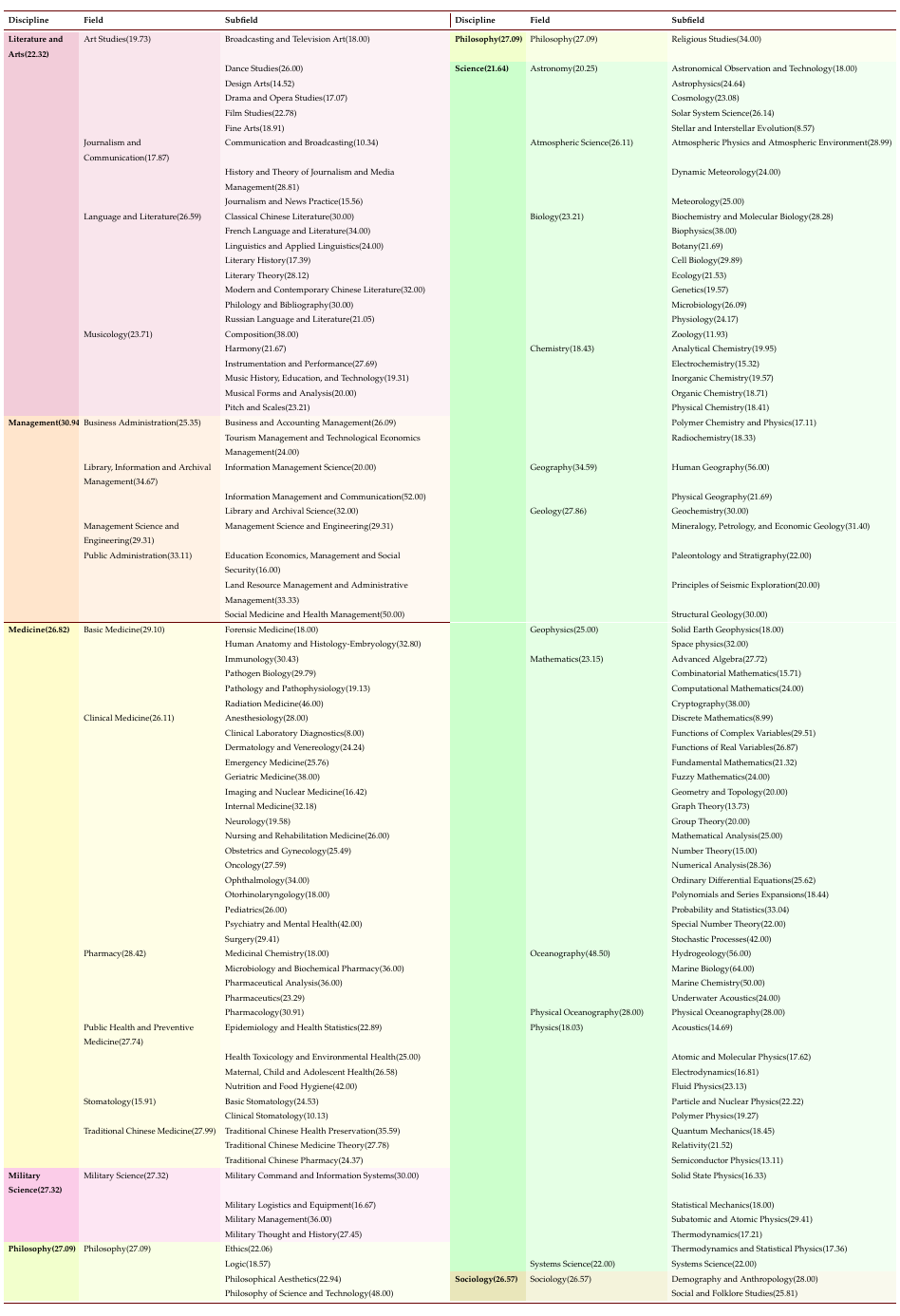} 
    \end{subtable}
    \vspace{-1.1cm}
    \captionsetup{font=small}
    \caption{Continued: Model Scores Across Three Levels of Disciplines: Qwen2.5-3B-Instruct.}
    \vspace{-0.6cm}
    \centeredlinks{listofmodels}{Back to List of Models}{toc}{Back to Table of Contents}{blue}
    \end{table}
}
\clearpage

\newpage
\vspace{-0.5cm}
\begin{table}[t]
\refstepcounter{models}%
\addcontentsline{csf}{models}{\protect\numberline{\themodels}Yi-1.5-9B-Chat}
\centering
\begin{subtable}[t]{1\textwidth}
\centering
\includegraphics[width=\textwidth]{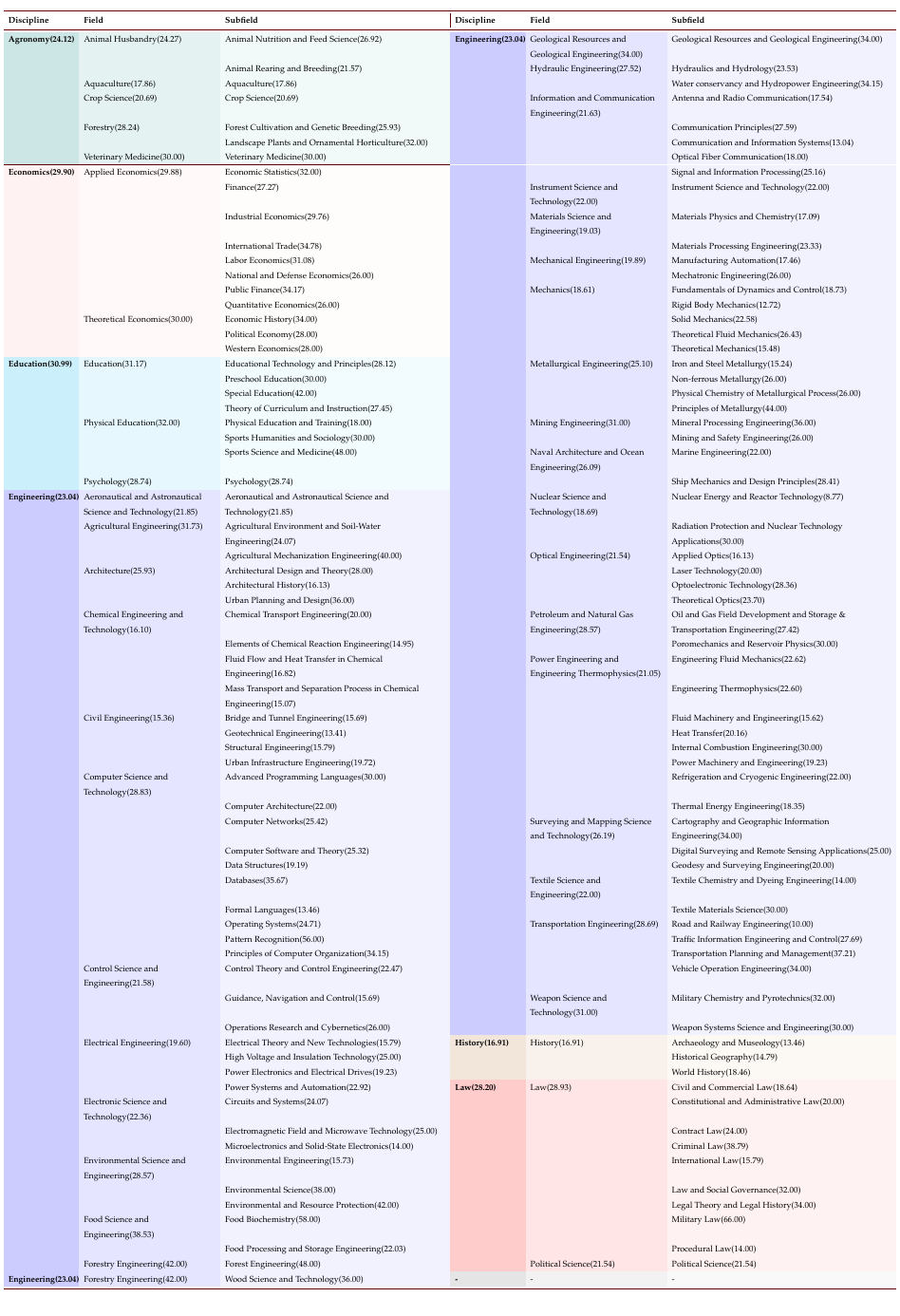} 
\end{subtable}
\vspace{-1.3cm}
\captionsetup{font=small}
\caption{Model Scores Across Three Levels of Disciplines: Yi-1.5-9B-Chat.}
\label{tab:Yi-1.5-9B-Chat}
\vspace{-0.5cm}
\centeredlinks{listofmodels}{Back to List of Models}{toc}{Back to Table of Contents}{blue}
\end{table}
\clearpage

\newpage
\afterpage{
    \begin{table}[t]
    \centering
    \ContinuedFloat 
    \begin{subtable}[t]{\textwidth}
    \centering
    \includegraphics[width=\textwidth]{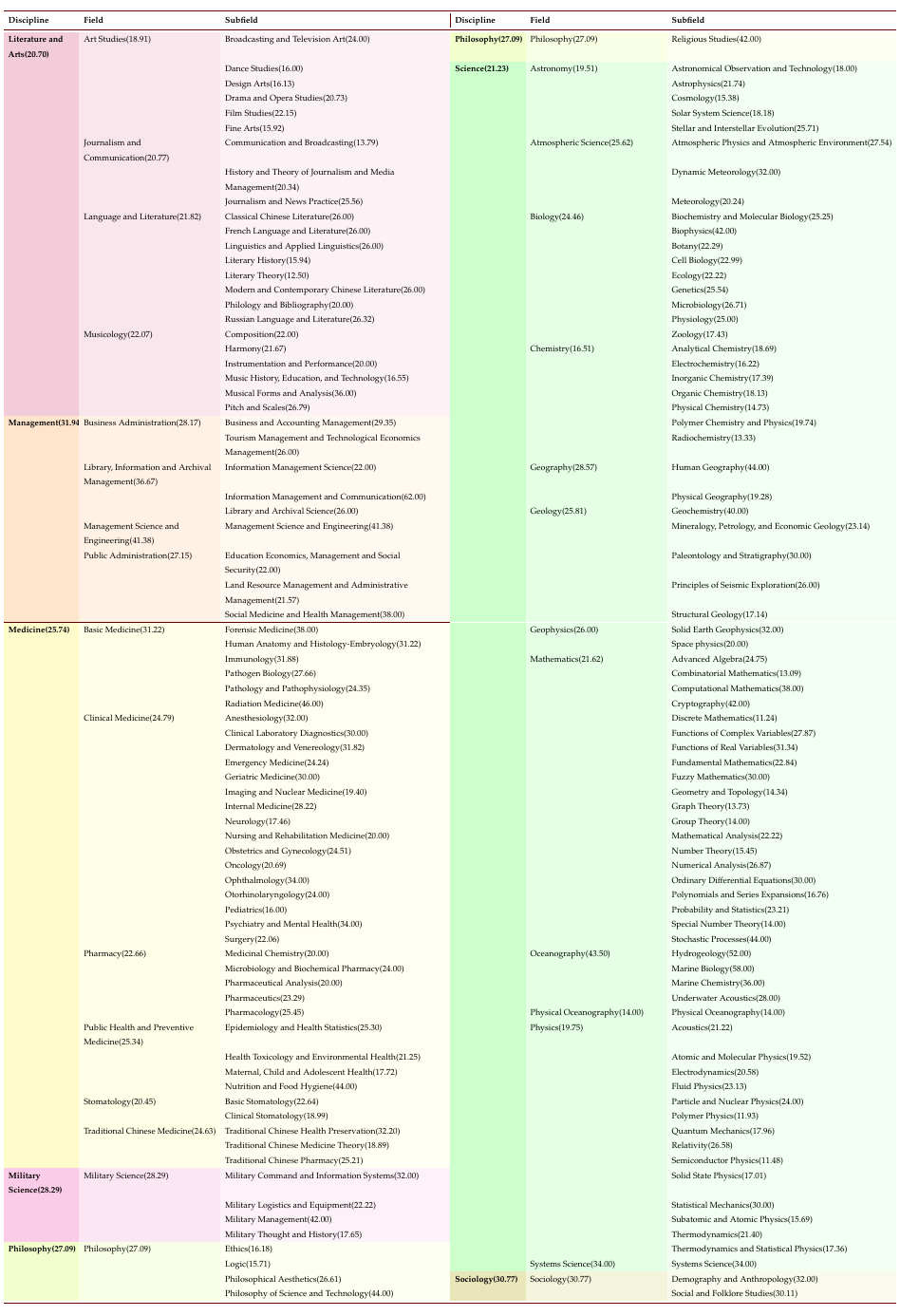} 
    \end{subtable}
    \vspace{-1.1cm}
    \captionsetup{font=small}
    \caption{Continued: Model Scores Across Three Levels of Disciplines: Yi-1.5-9B-Chat.}
    \vspace{-0.6cm}
    \centeredlinks{listofmodels}{Back to List of Models}{toc}{Back to Table of Contents}{blue}
    \end{table}
}
\clearpage

\newpage
\vspace{-0.5cm}
\begin{table}[t]
\refstepcounter{models}%
\addcontentsline{csf}{models}{\protect\numberline{\themodels}Yi-1.5-9B}
\centering
\begin{subtable}[t]{1\textwidth}
\centering
\includegraphics[width=\textwidth]{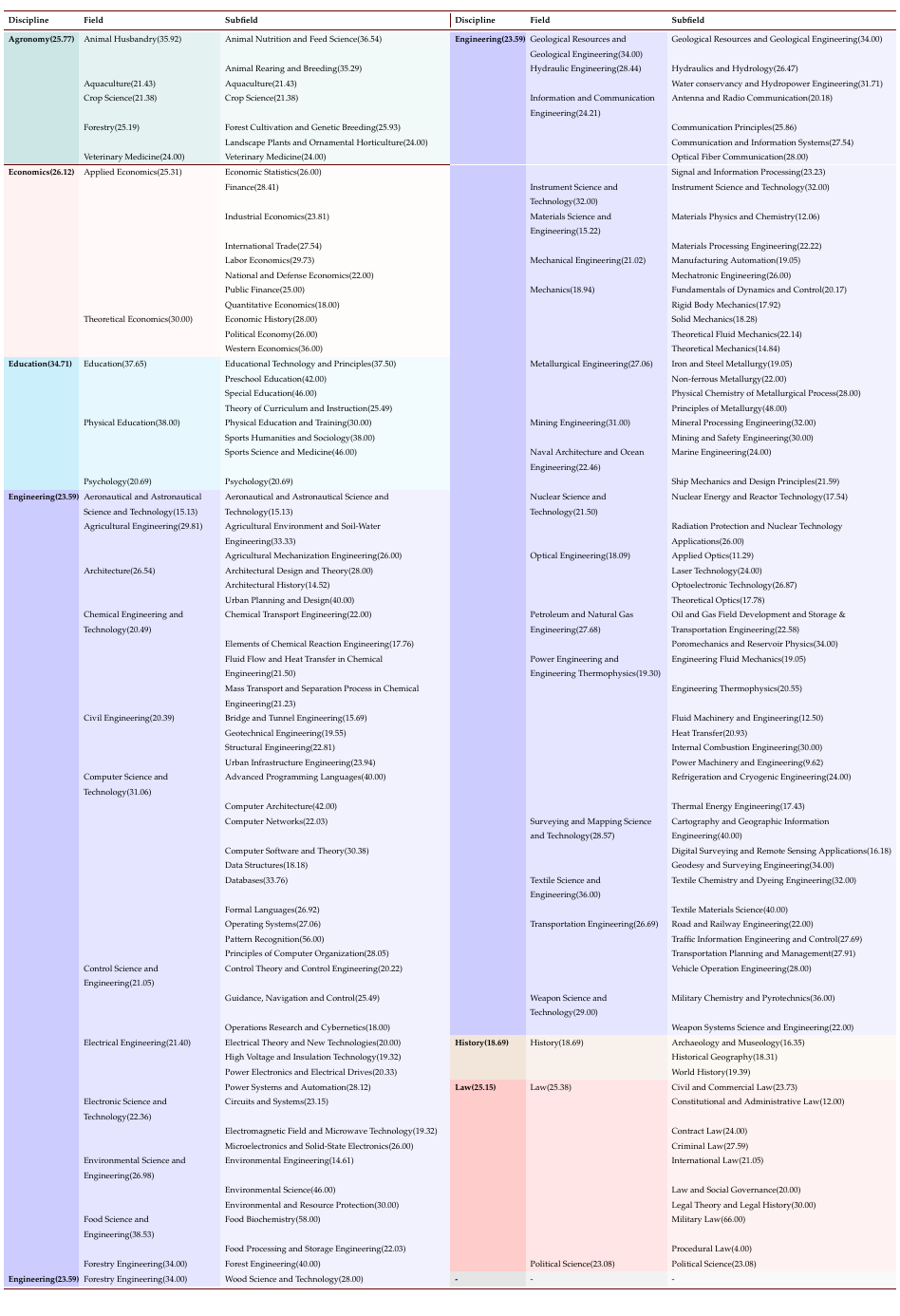} 
\end{subtable}
\vspace{-1.3cm}
\captionsetup{font=small}
\caption{Model Scores Across Three Levels of Disciplines: Yi-1.5-9B.}
\label{tab:Yi-1.5-9B}
\vspace{-0.5cm}
\centeredlinks{listofmodels}{Back to List of Models}{toc}{Back to Table of Contents}{blue}
\end{table}
\clearpage

\newpage
\afterpage{
    \begin{table}[t]
    \centering
    \ContinuedFloat 
    \begin{subtable}[t]{\textwidth}
    \centering
    \includegraphics[width=\textwidth]{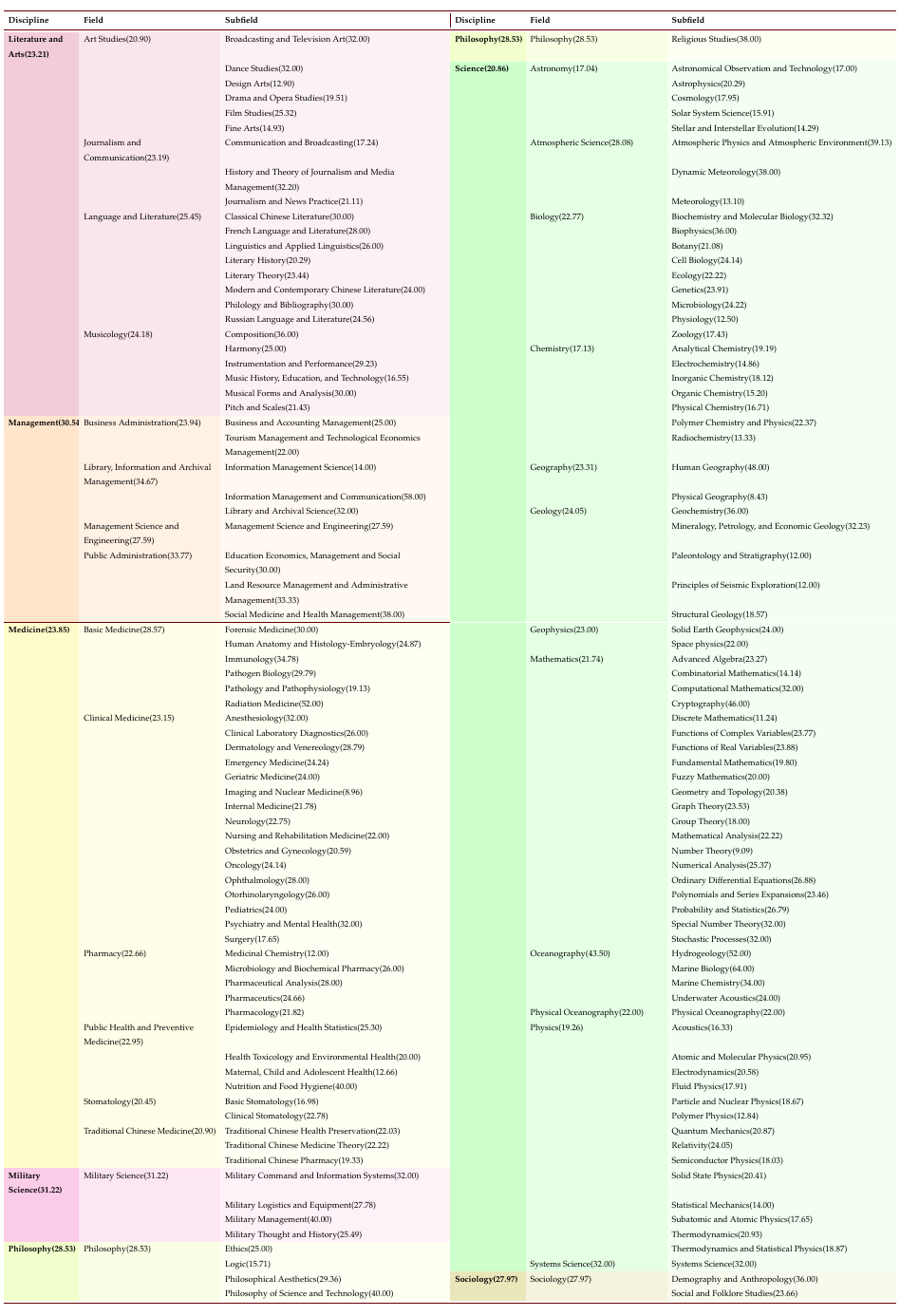} 
    \end{subtable}
    \vspace{-1.1cm}
    \captionsetup{font=small}
    \caption{Continued: Model Scores Across Three Levels of Disciplines: Yi-1.5-9B.}
    \vspace{-0.6cm}
    \centeredlinks{listofmodels}{Back to List of Models}{toc}{Back to Table of Contents}{blue}
    \end{table}
}
\clearpage

\newpage
\vspace{-0.5cm}
\begin{table}[t]
\refstepcounter{models}%
\addcontentsline{csf}{models}{\protect\numberline{\themodels}gemma-2-9b}
\centering
\begin{subtable}[t]{1\textwidth}
\centering
\includegraphics[width=\textwidth]{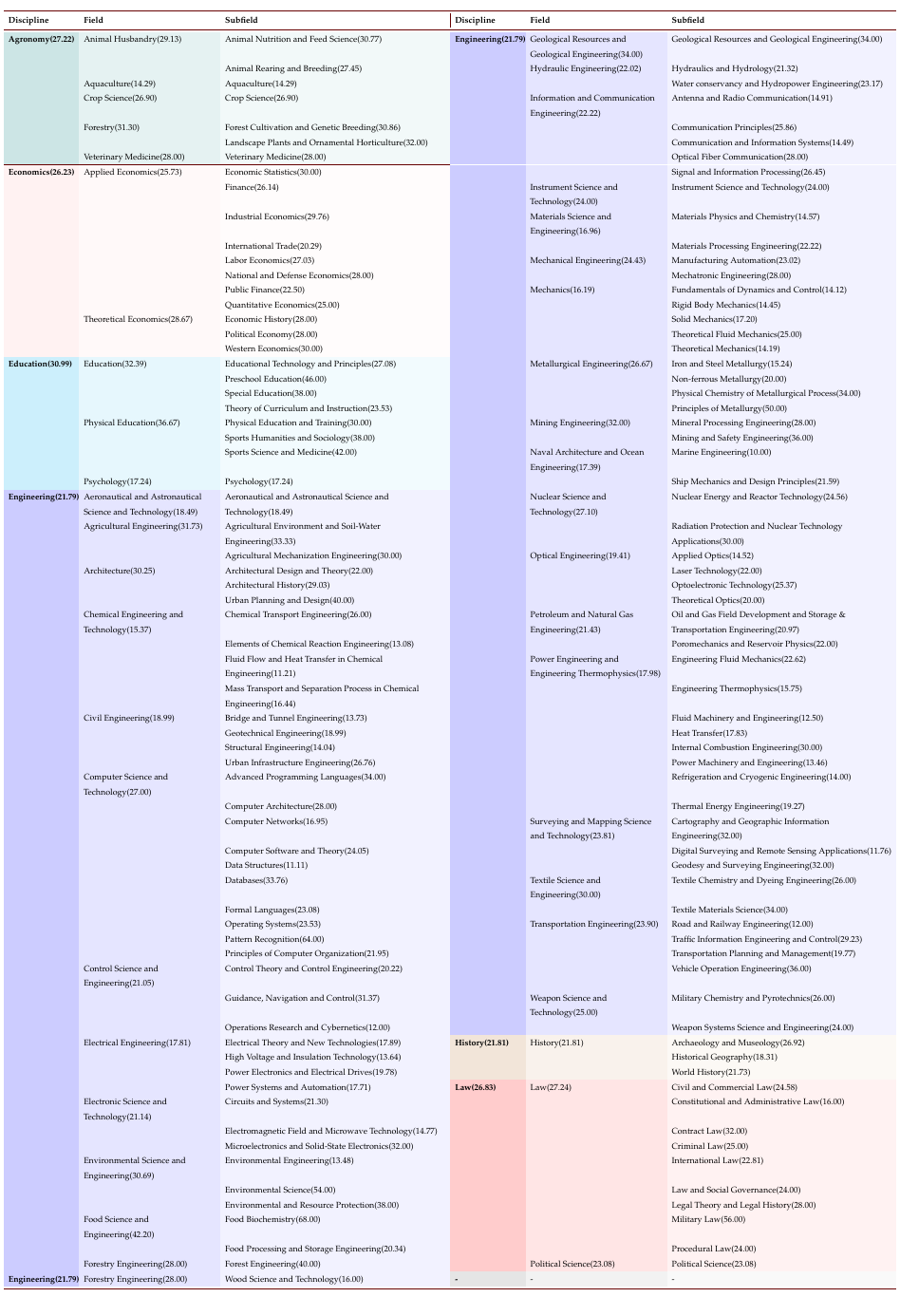} 
\end{subtable}
\vspace{-1.3cm}
\captionsetup{font=small}
\caption{Model Scores Across Three Levels of Disciplines: gemma-2-9b.}
\label{tab:gemma-2-9b}
\vspace{-0.5cm}
\centeredlinks{listofmodels}{Back to List of Models}{toc}{Back to Table of Contents}{blue}
\end{table}
\clearpage

\newpage
\afterpage{
    \begin{table}[t]
    \centering
    \ContinuedFloat 
    \begin{subtable}[t]{\textwidth}
    \centering
    \includegraphics[width=\textwidth]{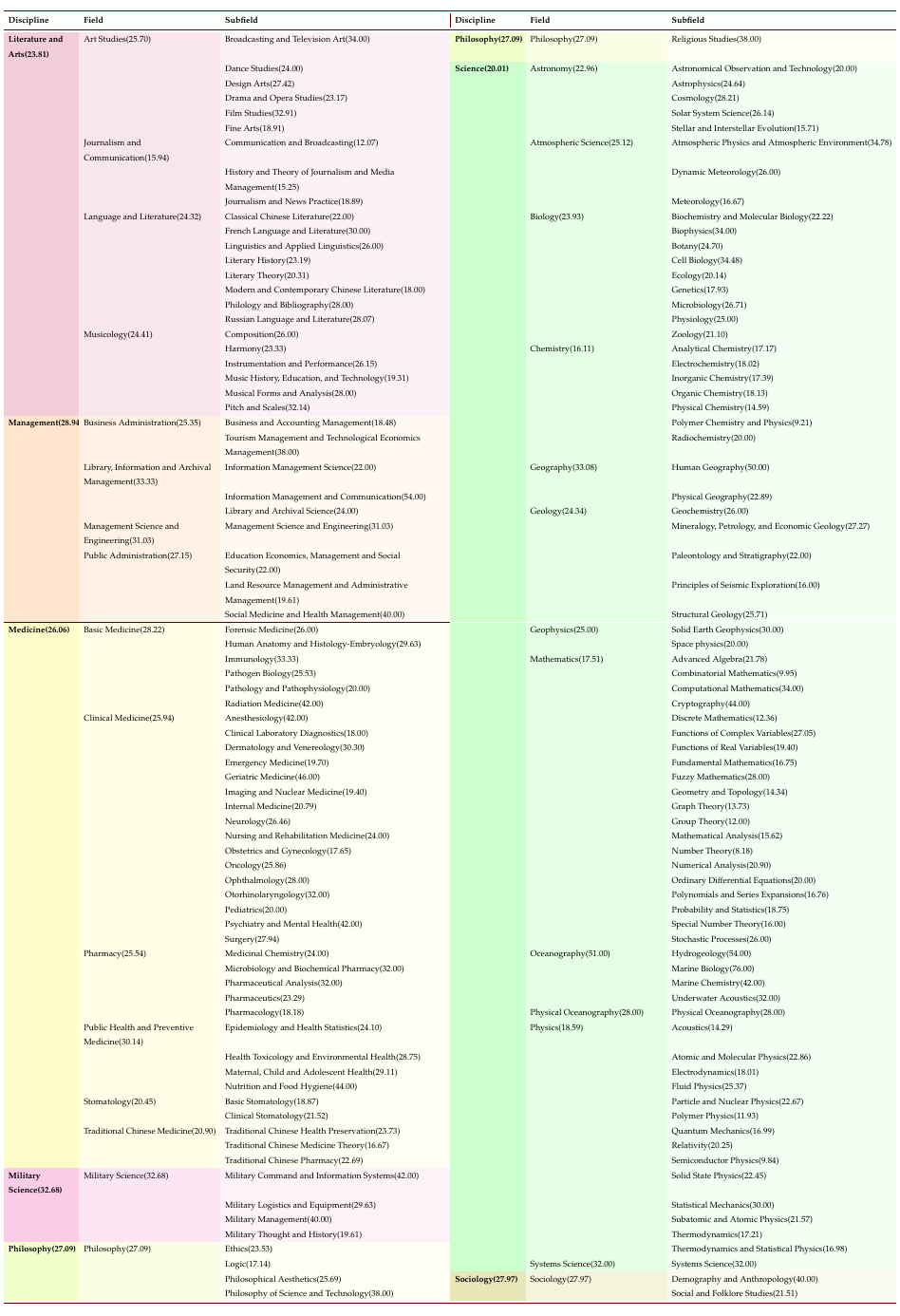} 
    \end{subtable}
    \vspace{-1.1cm}
    \captionsetup{font=small}
    \caption{Continued: Model Scores Across Three Levels of Disciplines: gemma-2-9b.}
    \vspace{-0.6cm}
    \centeredlinks{listofmodels}{Back to List of Models}{toc}{Back to Table of Contents}{blue}
    \end{table}
}
\clearpage

\newpage
\vspace{-0.5cm}
\begin{table}[t]
\refstepcounter{models}%
\addcontentsline{csf}{models}{\protect\numberline{\themodels}K2-Chat}
\centering
\begin{subtable}[t]{1\textwidth}
\centering
\includegraphics[width=\textwidth]{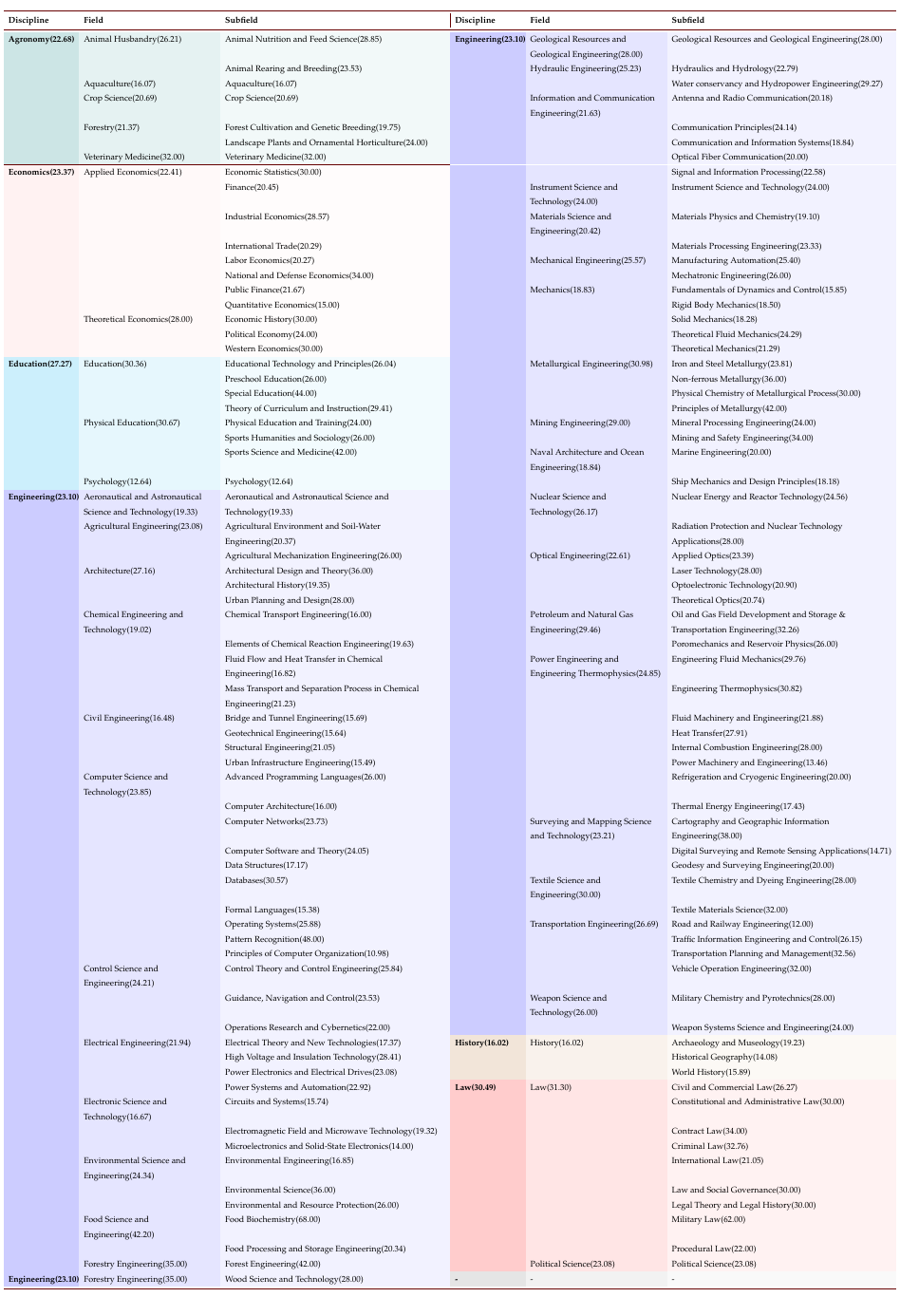} 
\end{subtable}
\vspace{-1.3cm}
\captionsetup{font=small}
\caption{Model Scores Across Three Levels of Disciplines: K2-Chat.}
\label{tab:K2-Chat}
\vspace{-0.5cm}
\centeredlinks{listofmodels}{Back to List of Models}{toc}{Back to Table of Contents}{blue}
\end{table}
\clearpage

\newpage
\afterpage{
    \begin{table}[t]
    \centering
    \ContinuedFloat 
    \begin{subtable}[t]{\textwidth}
    \centering
    \includegraphics[width=\textwidth]{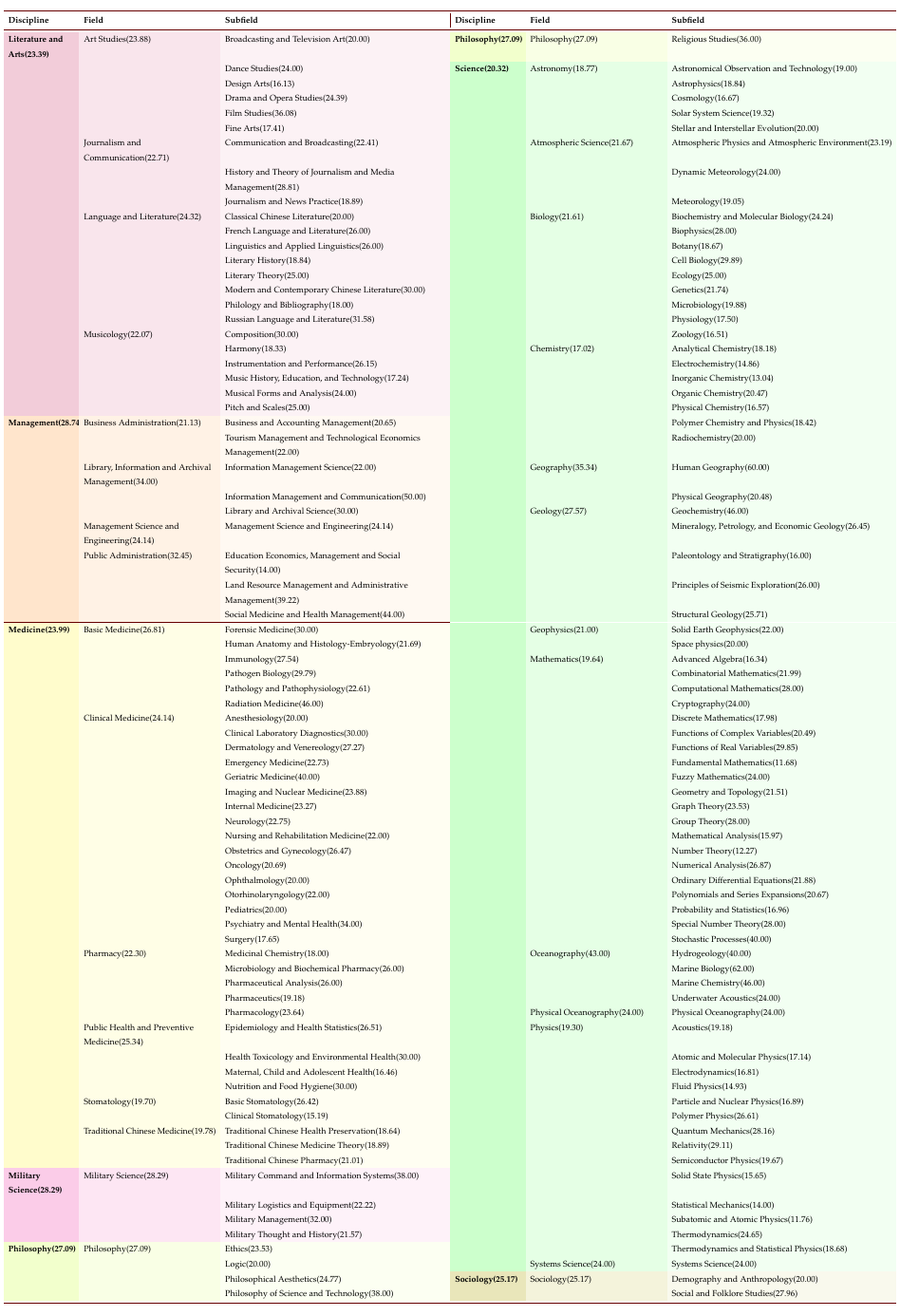} 
    \end{subtable}
    \vspace{-1.1cm}
    \captionsetup{font=small}
    \caption{Continued: Model Scores Across Three Levels of Disciplines: K2-Chat.}
    \vspace{-0.6cm}
    \centeredlinks{listofmodels}{Back to List of Models}{toc}{Back to Table of Contents}{blue}
    \end{table}
}
\clearpage

\newpage
\vspace{-0.5cm}
\begin{table}[t]
\refstepcounter{models}%
\addcontentsline{csf}{models}{\protect\numberline{\themodels}Mixtral-8x22B-v0.1}
\centering
\begin{subtable}[t]{1\textwidth}
\centering
\includegraphics[width=\textwidth]{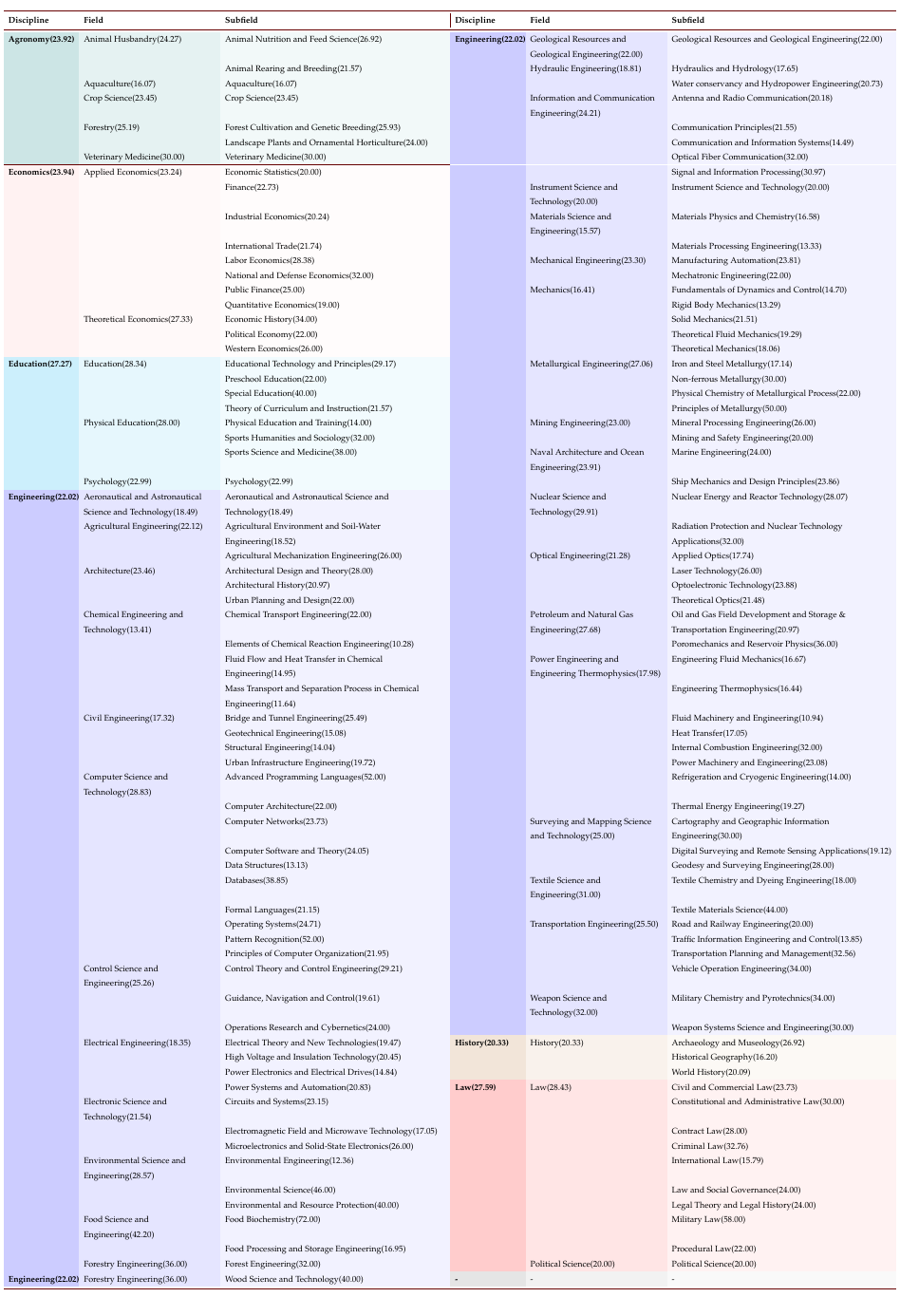} 
\end{subtable}
\vspace{-1.3cm}
\captionsetup{font=small}
\caption{Model Scores Across Three Levels of Disciplines: Mixtral-8x22B-v0.1.}
\label{tab:Mixtral-8x22B-v0.1}
\vspace{-0.5cm}
\centeredlinks{listofmodels}{Back to List of Models}{toc}{Back to Table of Contents}{blue}
\end{table}
\clearpage

\newpage
\afterpage{
    \begin{table}[t]
    \centering
    \ContinuedFloat 
    \begin{subtable}[t]{\textwidth}
    \centering
    \includegraphics[width=\textwidth]{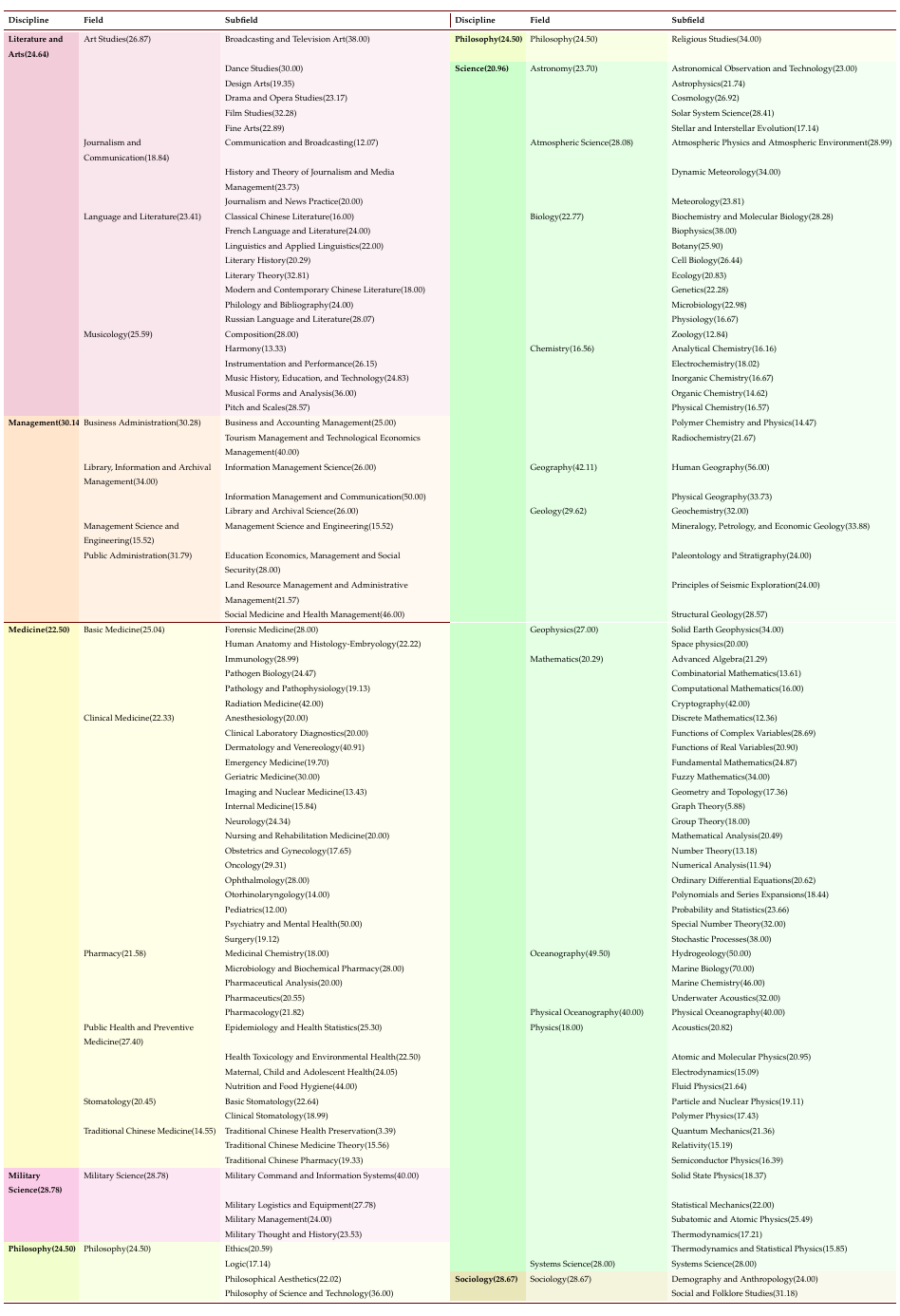} 
    \end{subtable}
    \vspace{-1.1cm}
    \captionsetup{font=small}
    \caption{Continued: Model Scores Across Three Levels of Disciplines: Mixtral-8x22B-v0.1.}
    \vspace{-0.6cm}
    \centeredlinks{listofmodels}{Back to List of Models}{toc}{Back to Table of Contents}{blue}
    \end{table}
}
\clearpage

\newpage
\vspace{-0.5cm}
\begin{table}[t]
\refstepcounter{models}%
\addcontentsline{csf}{models}{\protect\numberline{\themodels}Mixtral-8x7B-Instruct-v0.1}
\centering
\begin{subtable}[t]{1\textwidth}
\centering
\includegraphics[width=\textwidth]{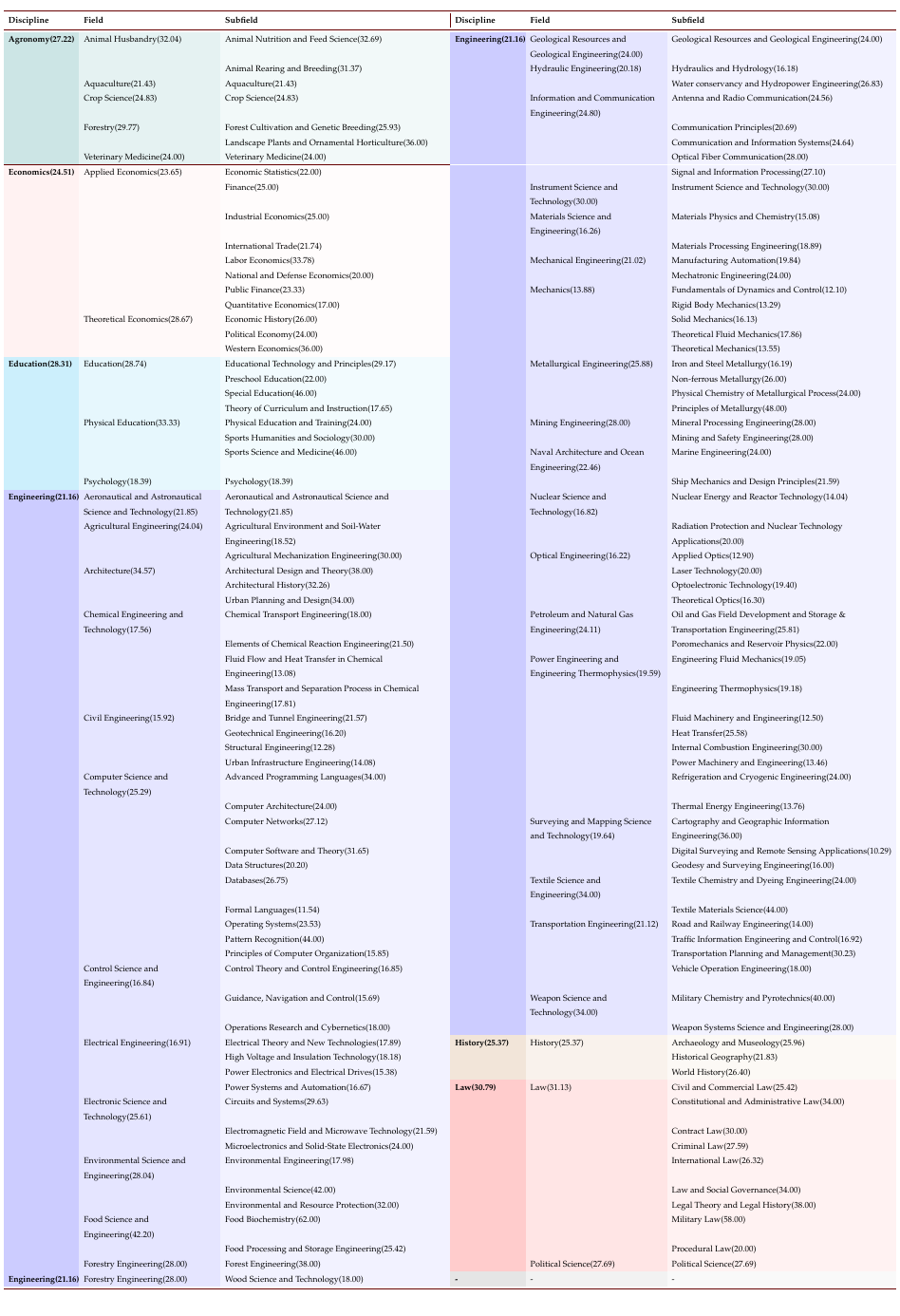} 
\end{subtable}
\vspace{-1.3cm}
\captionsetup{font=small}
\caption{Model Scores Across Three Levels of Disciplines: Mixtral-8x7B-Instruct-v0.1.}
\label{tab:Mixtral-8x7B-Instruct-v0.1}
\vspace{-0.5cm}
\centeredlinks{listofmodels}{Back to List of Models}{toc}{Back to Table of Contents}{blue}
\end{table}
\clearpage

\newpage
\afterpage{
    \begin{table}[t]
    \centering
    \ContinuedFloat 
    \begin{subtable}[t]{\textwidth}
    \centering
    \includegraphics[width=\textwidth]{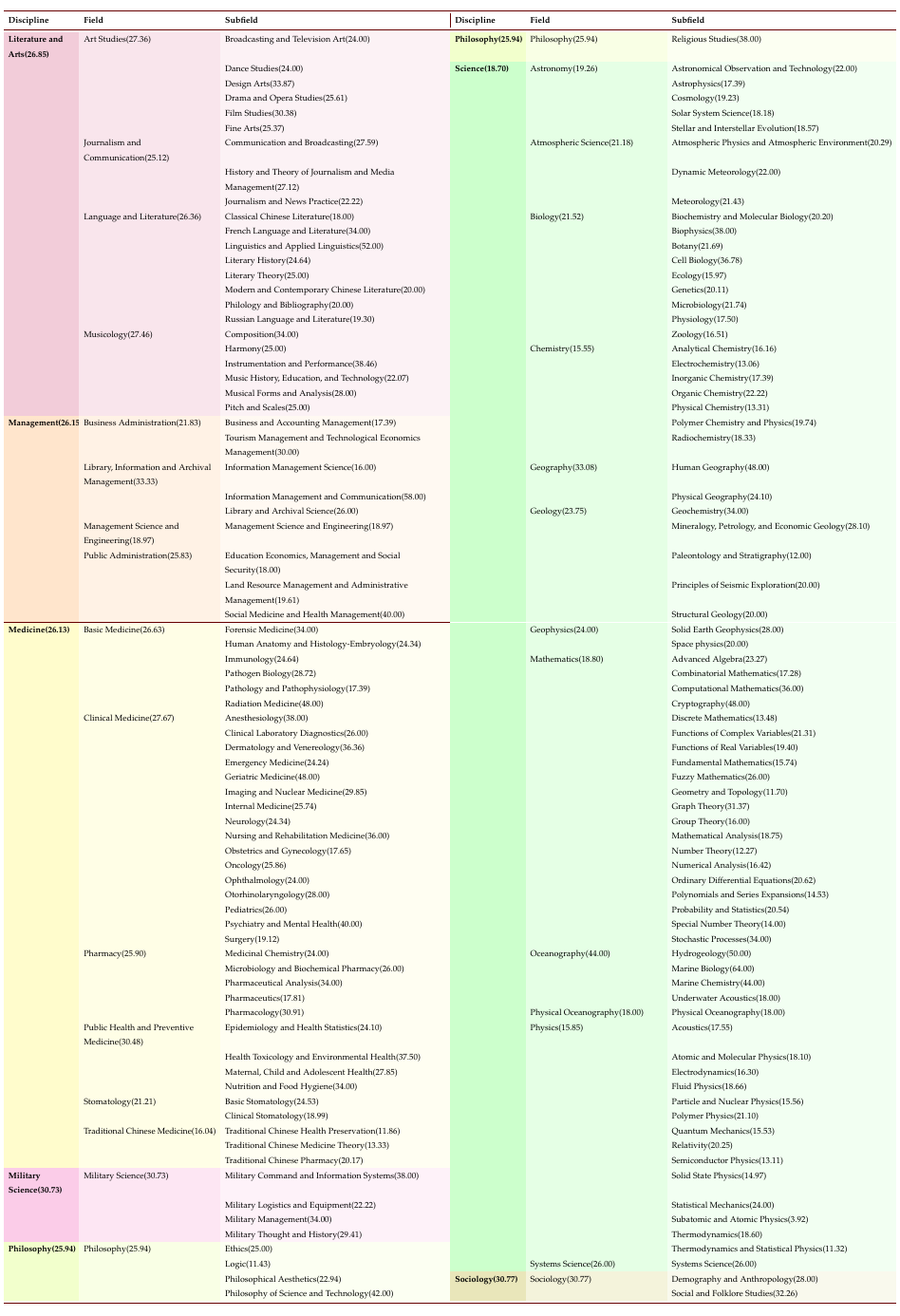} 
    \end{subtable}
    \vspace{-1.1cm}
    \captionsetup{font=small}
    \caption{Continued: Model Scores Across Three Levels of Disciplines: Mixtral-8x7B-Instruct-v0.1.}
    \vspace{-0.6cm}
    \centeredlinks{listofmodels}{Back to List of Models}{toc}{Back to Table of Contents}{blue}
    \end{table}
}
\clearpage

\newpage
\vspace{-0.5cm}
\begin{table}[t]
\refstepcounter{models}%
\addcontentsline{csf}{models}{\protect\numberline{\themodels}Mixtral-8x7B-v0.1}
\centering
\begin{subtable}[t]{1\textwidth}
\centering
\includegraphics[width=\textwidth]{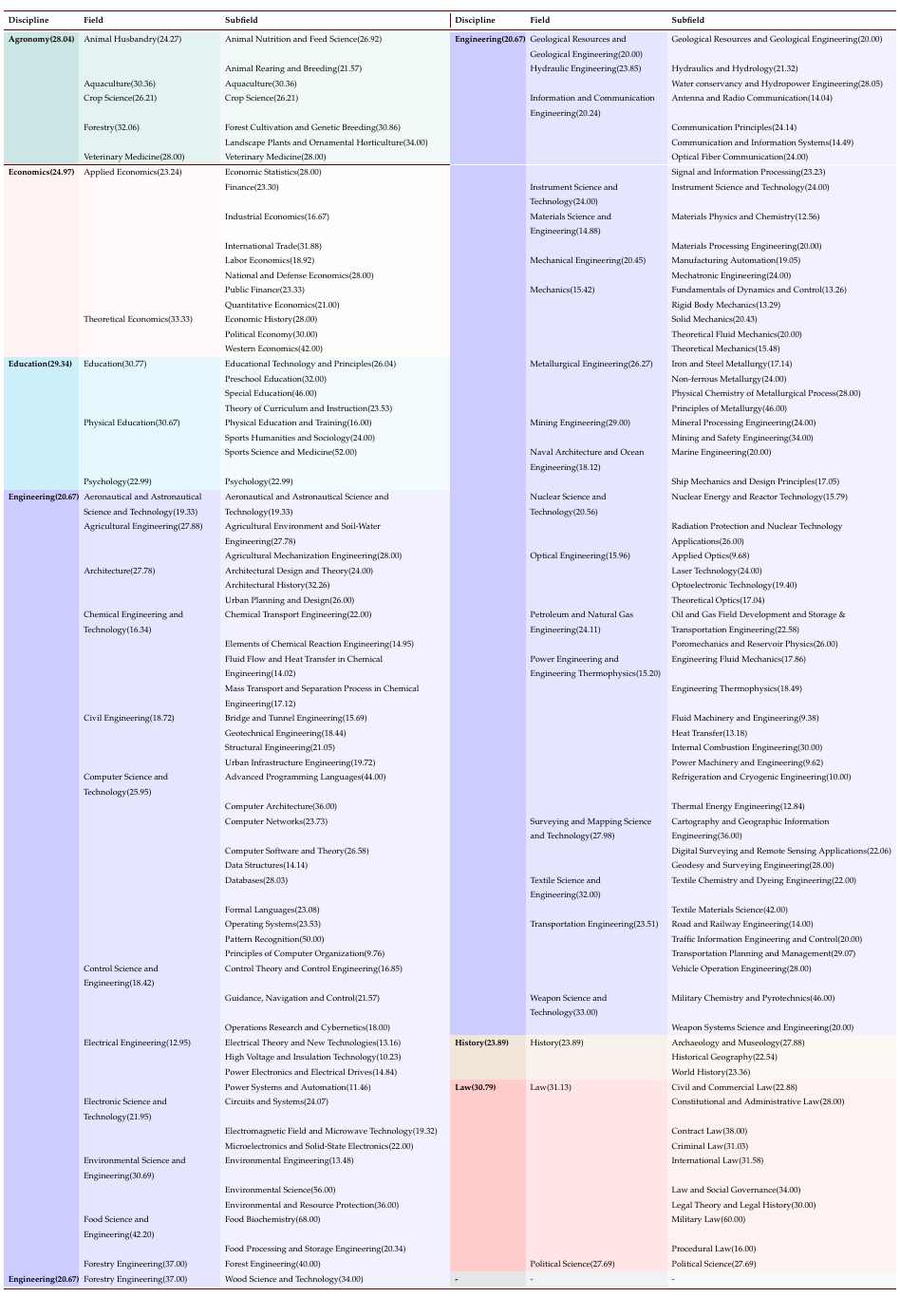} 
\end{subtable}
\vspace{-1.3cm}
\captionsetup{font=small}
\caption{Model Scores Across Three Levels of Disciplines: Mixtral-8x7B-v0.1.}
\label{tab:Mixtral-8x7B-v0.1}
\vspace{-0.5cm}
\centeredlinks{listofmodels}{Back to List of Models}{toc}{Back to Table of Contents}{blue}
\end{table}
\clearpage

\newpage
\afterpage{
    \begin{table}[t]
    \centering
    \ContinuedFloat 
    \begin{subtable}[t]{\textwidth}
    \centering
    \includegraphics[width=\textwidth]{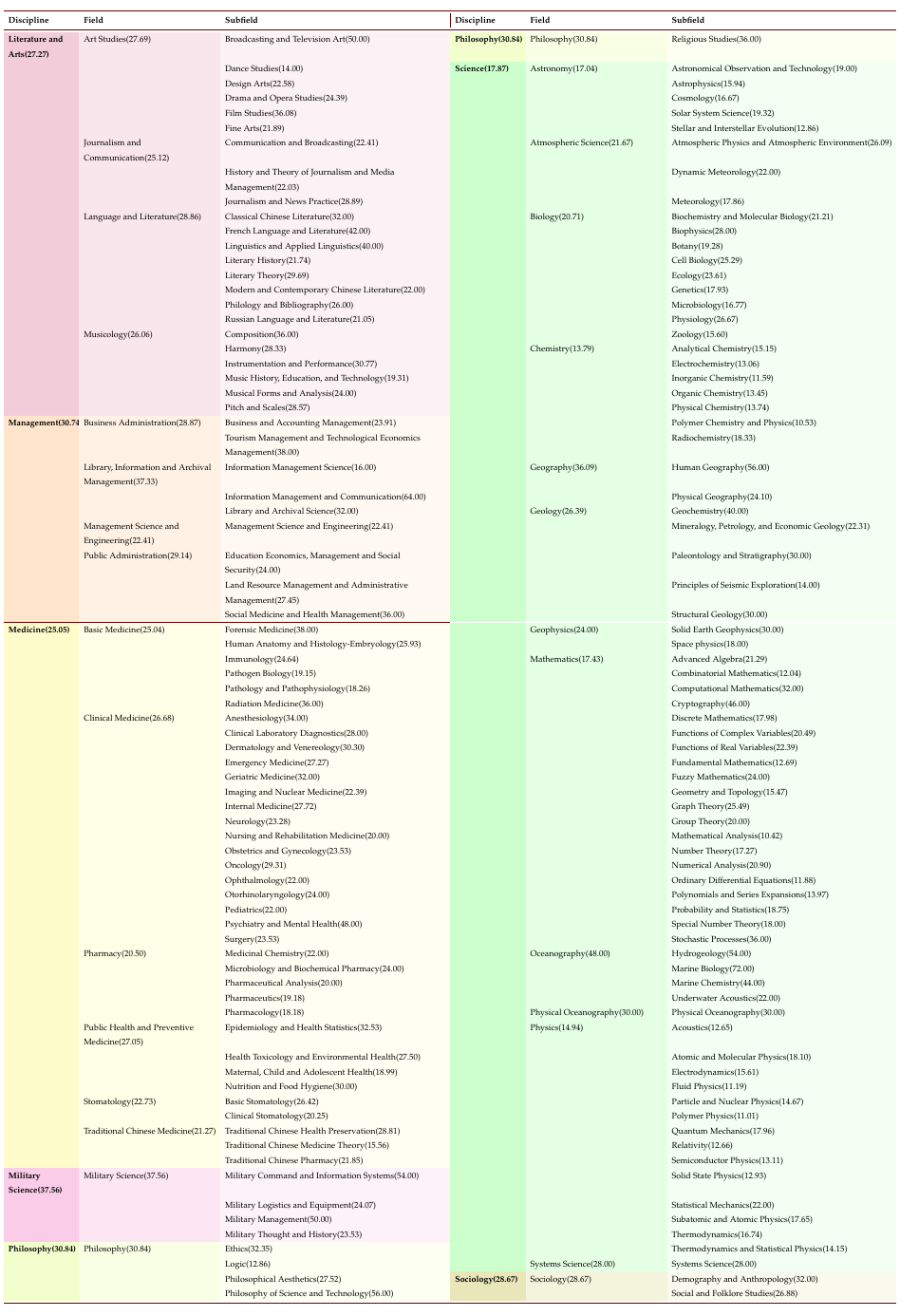} 
    \end{subtable}
    \vspace{-1.1cm}
    \captionsetup{font=small}
    \caption{Continued: Model Scores Across Three Levels of Disciplines: Mixtral-8x7B-v0.1.}
    \vspace{-0.6cm}
    \centeredlinks{listofmodels}{Back to List of Models}{toc}{Back to Table of Contents}{blue}
    \end{table}
}
\clearpage

\newpage
\vspace{-0.5cm}
\begin{table}[t]
\refstepcounter{models}%
\addcontentsline{csf}{models}{\protect\numberline{\themodels}K2}
\centering
\begin{subtable}[t]{1\textwidth}
\centering
\includegraphics[width=\textwidth]{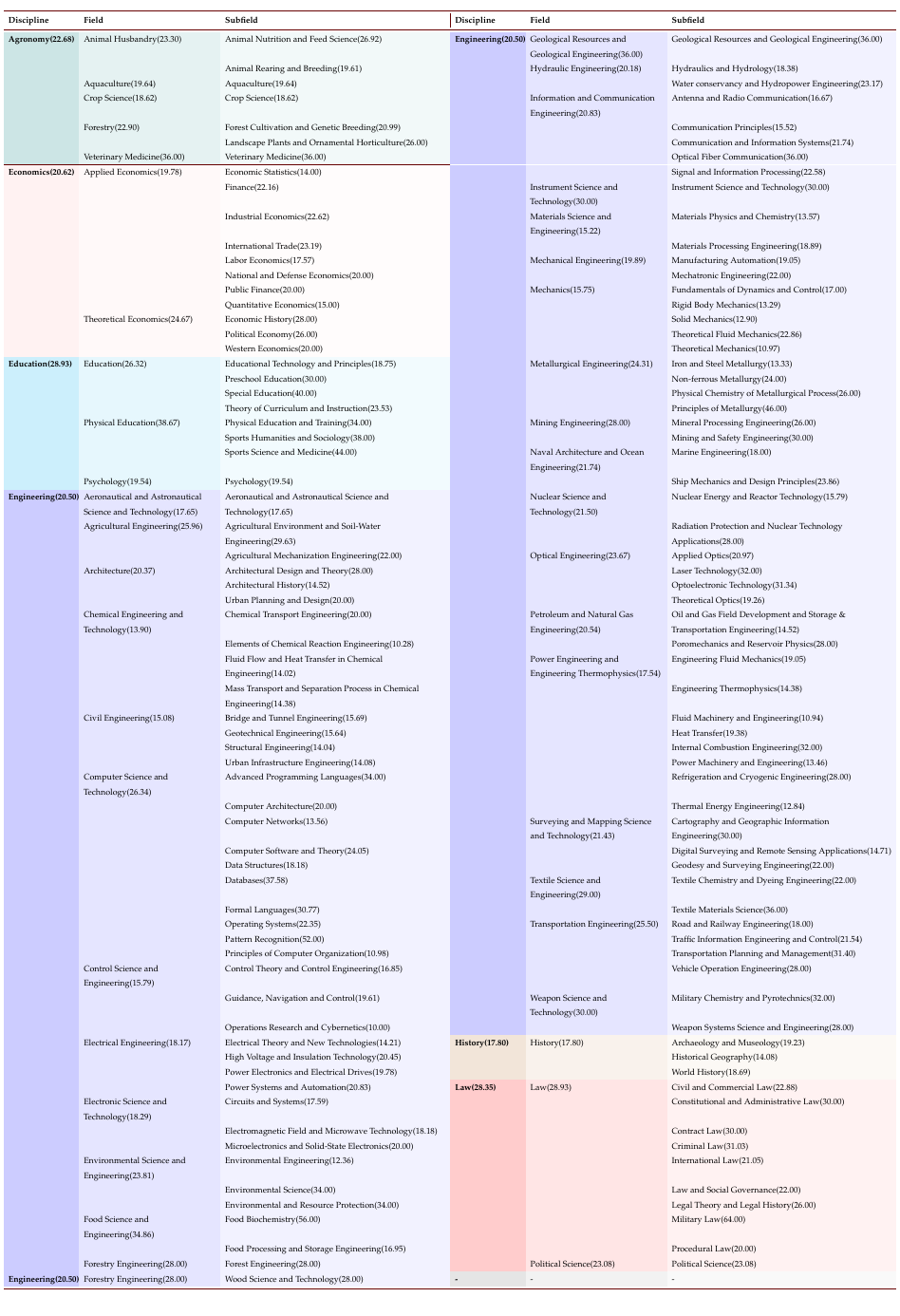} 
\end{subtable}
\vspace{-1.3cm}
\captionsetup{font=small}
\caption{Model Scores Across Three Levels of Disciplines: K2.}
\label{tab:K2}
\vspace{-0.5cm}
\centeredlinks{listofmodels}{Back to List of Models}{toc}{Back to Table of Contents}{blue}
\end{table}
\clearpage

\newpage
\afterpage{
    \begin{table}[t]
    \centering
    \ContinuedFloat 
    \begin{subtable}[t]{\textwidth}
    \centering
    \includegraphics[width=\textwidth]{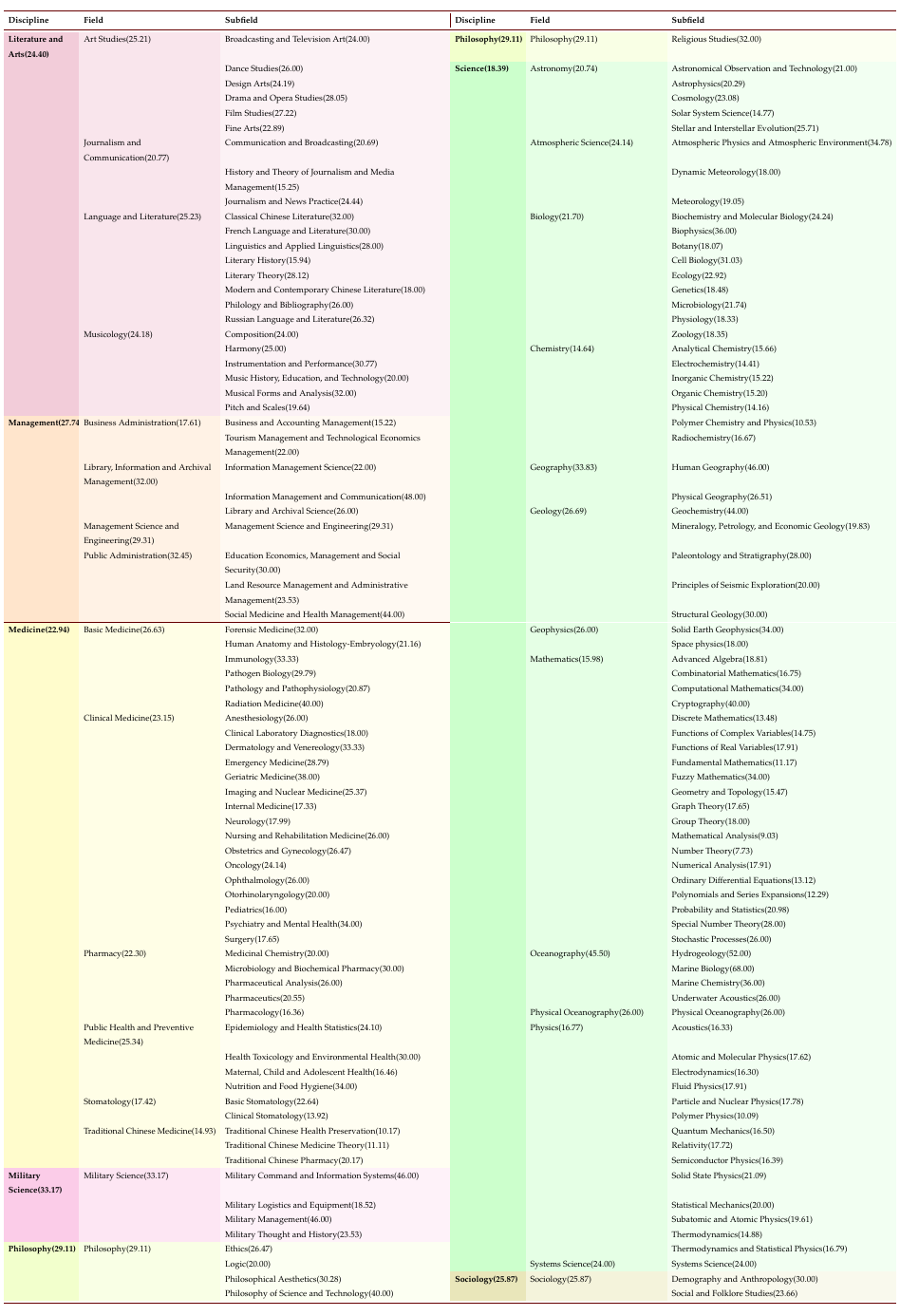} 
    \end{subtable}
    \vspace{-1.1cm}
    \captionsetup{font=small}
    \caption{Continued: Model Scores Across Three Levels of Disciplines: K2.}
    \vspace{-0.6cm}
    \centeredlinks{listofmodels}{Back to List of Models}{toc}{Back to Table of Contents}{blue}
    \end{table}
}
\clearpage

\newpage
\vspace{-0.5cm}
\begin{table}[t]
\refstepcounter{models}%
\addcontentsline{csf}{models}{\protect\numberline{\themodels}granite-3.1-8b-instruct}
\centering
\begin{subtable}[t]{1\textwidth}
\centering
\includegraphics[width=\textwidth]{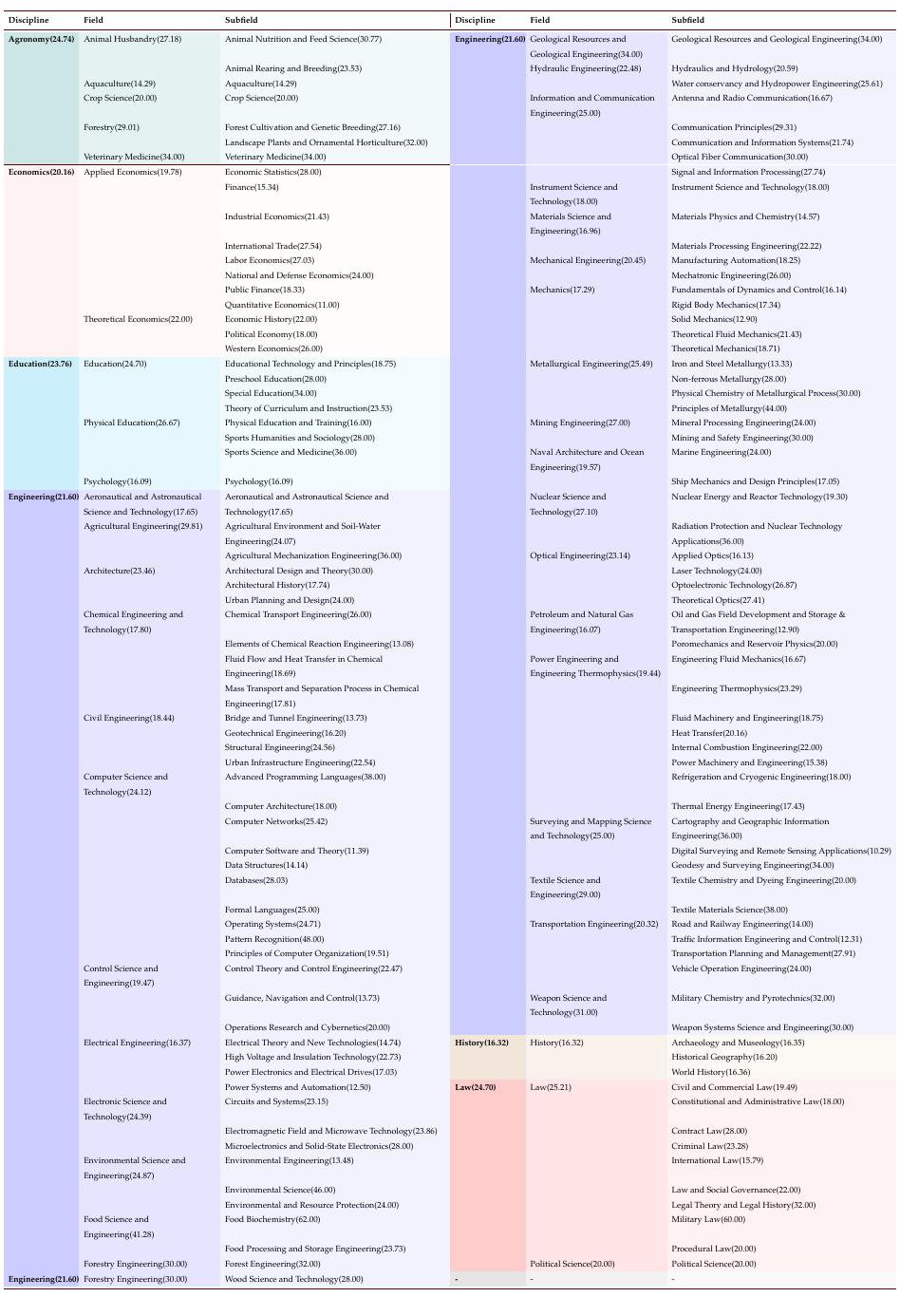} 
\end{subtable}
\vspace{-1.3cm}
\captionsetup{font=small}
\caption{Model Scores Across Three Levels of Disciplines: granite-3.1-8b-instruct.}
\label{tab:granite-3.1-8b-instruct}
\vspace{-0.5cm}
\centeredlinks{listofmodels}{Back to List of Models}{toc}{Back to Table of Contents}{blue}
\end{table}
\clearpage

\newpage
\afterpage{
    \begin{table}[t]
    \centering
    \ContinuedFloat 
    \begin{subtable}[t]{\textwidth}
    \centering
    \includegraphics[width=\textwidth]{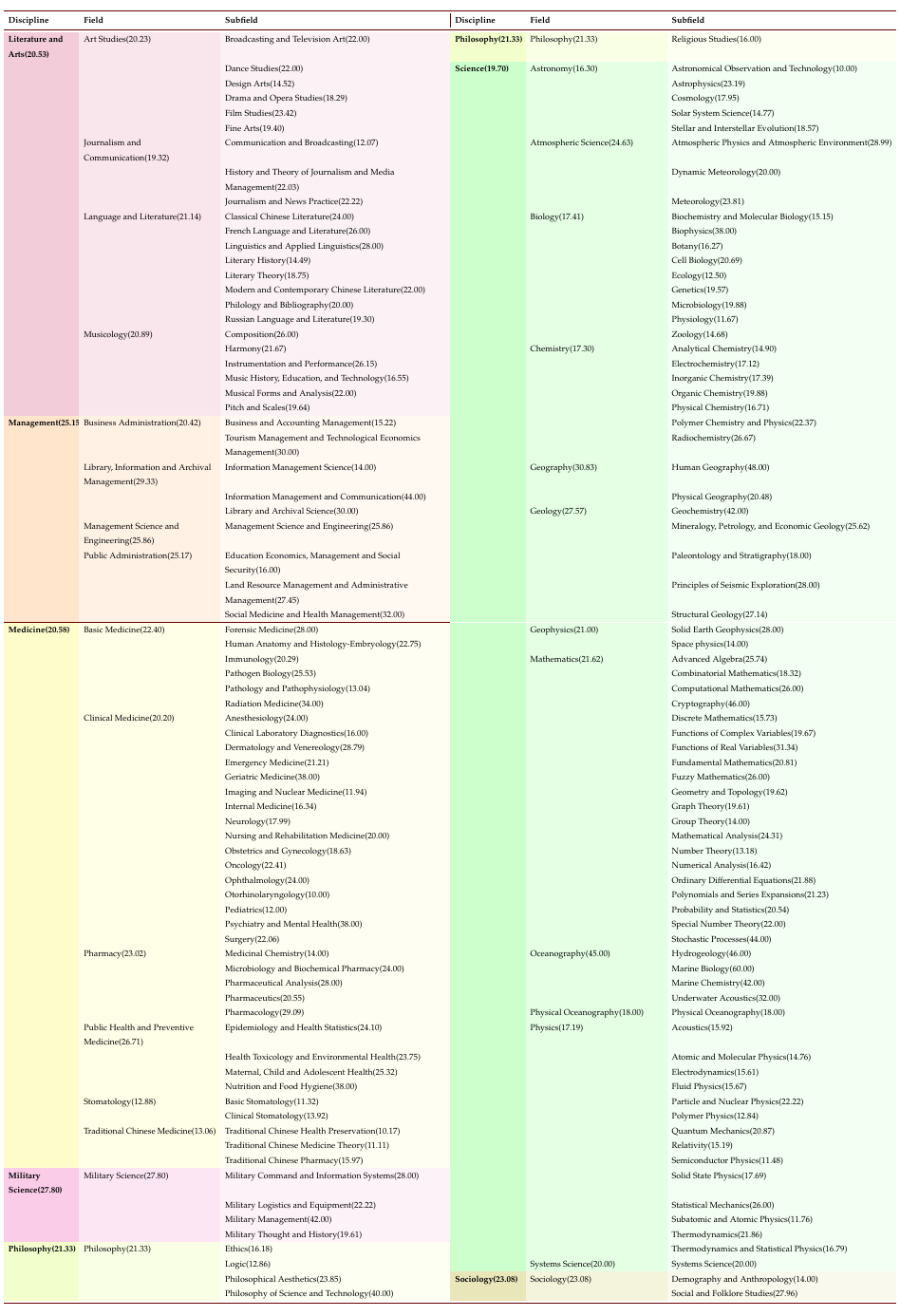} 
    \end{subtable}
    \vspace{-1.1cm}
    \captionsetup{font=small}
    \caption{Continued: Model Scores Across Three Levels of Disciplines: granite-3.1-8b-instruct.}
    \vspace{-0.6cm}
    \centeredlinks{listofmodels}{Back to List of Models}{toc}{Back to Table of Contents}{blue}
    \end{table}
}
\clearpage

\newpage
\vspace{-0.5cm}
\begin{table}[t]
\refstepcounter{models}%
\addcontentsline{csf}{models}{\protect\numberline{\themodels}Llama-3.1-8B-Instruct}
\centering
\begin{subtable}[t]{1\textwidth}
\centering
\includegraphics[width=\textwidth]{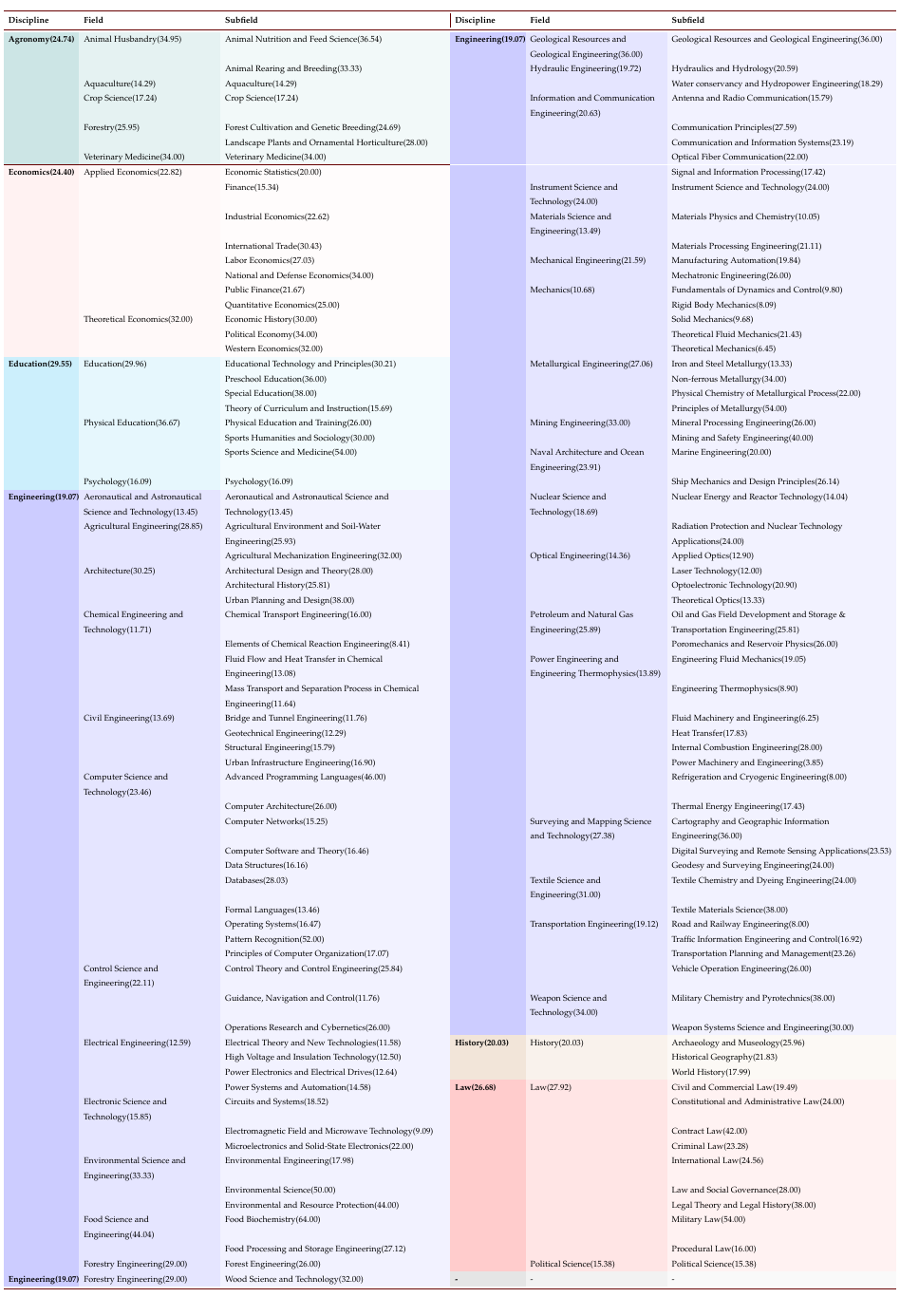} 
\end{subtable}
\vspace{-1.3cm}
\captionsetup{font=small}
\caption{Model Scores Across Three Levels of Disciplines: Llama-3.1-8B-Instruct.}
\label{tab:Llama-3.1-8B-Instruct}
\vspace{-0.5cm}
\centeredlinks{listofmodels}{Back to List of Models}{toc}{Back to Table of Contents}{blue}
\end{table}
\clearpage

\newpage
\afterpage{
    \begin{table}[t]
    \centering
    \ContinuedFloat 
    \begin{subtable}[t]{\textwidth}
    \centering
    \includegraphics[width=\textwidth]{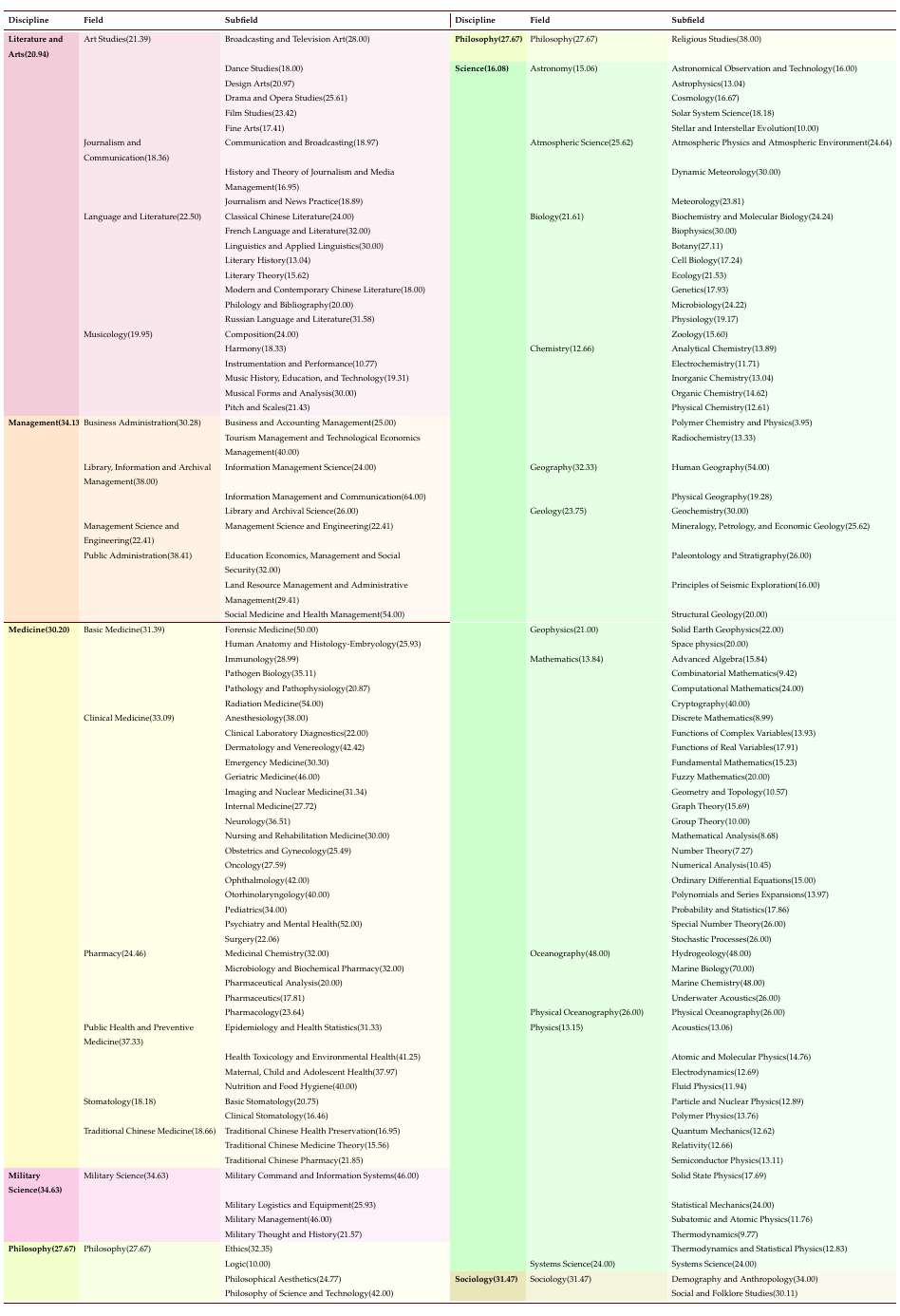} 
    \end{subtable}
    \vspace{-1.1cm}
    \captionsetup{font=small}
    \caption{Continued: Model Scores Across Three Levels of Disciplines: Llama-3.1-8B-Instruct.}
    \vspace{-0.6cm}
    \centeredlinks{listofmodels}{Back to List of Models}{toc}{Back to Table of Contents}{blue}
    \end{table}
}
\clearpage

\newpage
\vspace{-0.5cm}
\begin{table}[t]
\refstepcounter{models}%
\addcontentsline{csf}{models}{\protect\numberline{\themodels}Yi-1.5-6B}
\centering
\begin{subtable}[t]{1\textwidth}
\centering
\includegraphics[width=\textwidth]{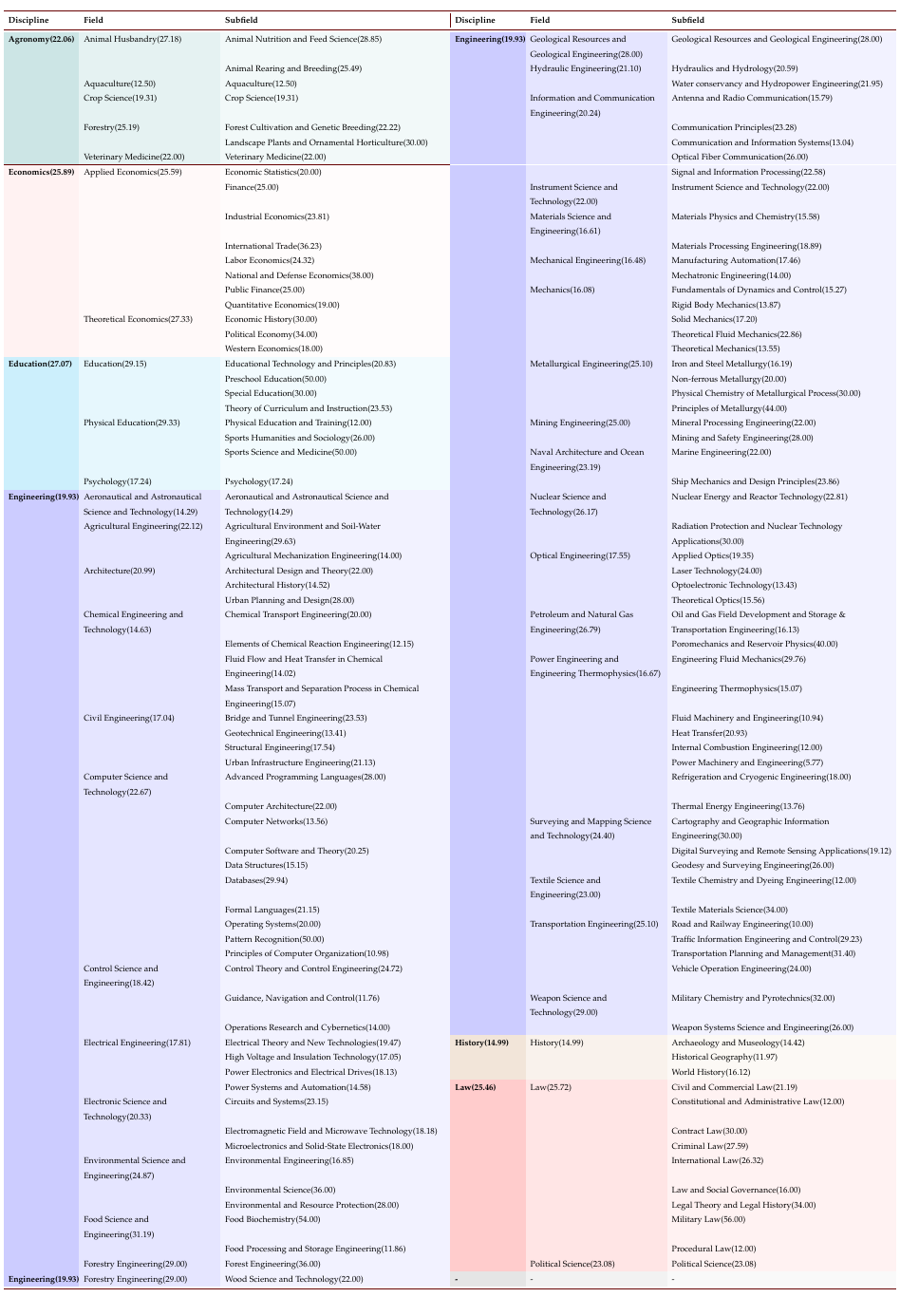} 
\end{subtable}
\vspace{-1.3cm}
\captionsetup{font=small}
\caption{Model Scores Across Three Levels of Disciplines: Yi-1.5-6B.}
\label{tab:Yi-1.5-6B}
\vspace{-0.5cm}
\centeredlinks{listofmodels}{Back to List of Models}{toc}{Back to Table of Contents}{blue}
\end{table}
\clearpage

\newpage
\afterpage{
    \begin{table}[t]
    \centering
    \ContinuedFloat 
    \begin{subtable}[t]{\textwidth}
    \centering
    \includegraphics[width=\textwidth]{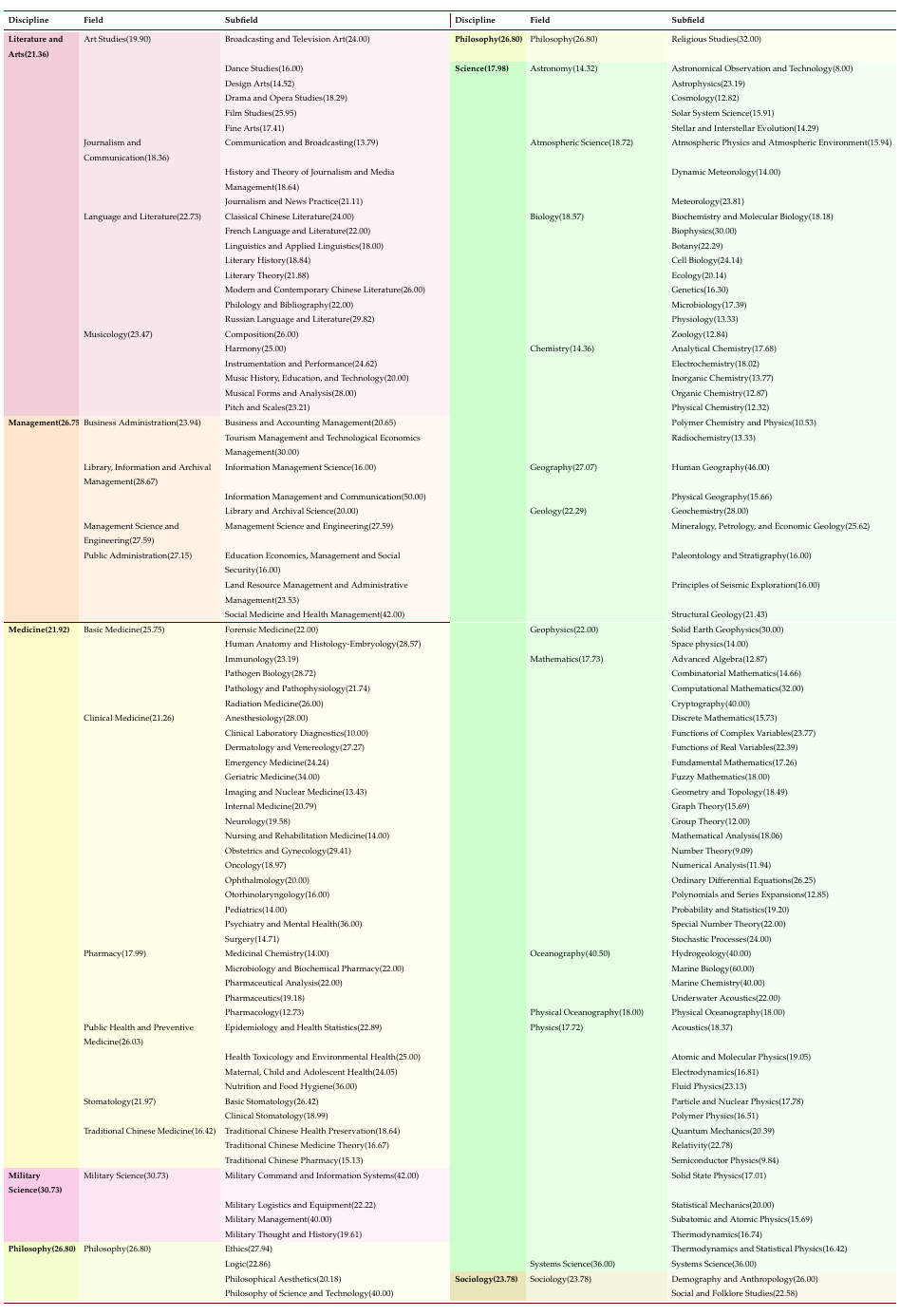} 
    \end{subtable}
    \vspace{-1.1cm}
    \captionsetup{font=small}
    \caption{Continued: Model Scores Across Three Levels of Disciplines: Yi-1.5-6B.}
    \vspace{-0.6cm}
    \centeredlinks{listofmodels}{Back to List of Models}{toc}{Back to Table of Contents}{blue}
    \end{table}
}
\clearpage

\newpage
\vspace{-0.5cm}
\begin{table}[t]
\refstepcounter{models}%
\addcontentsline{csf}{models}{\protect\numberline{\themodels}Qwen2.5-3B}
\centering
\begin{subtable}[t]{1\textwidth}
\centering
\includegraphics[width=\textwidth]{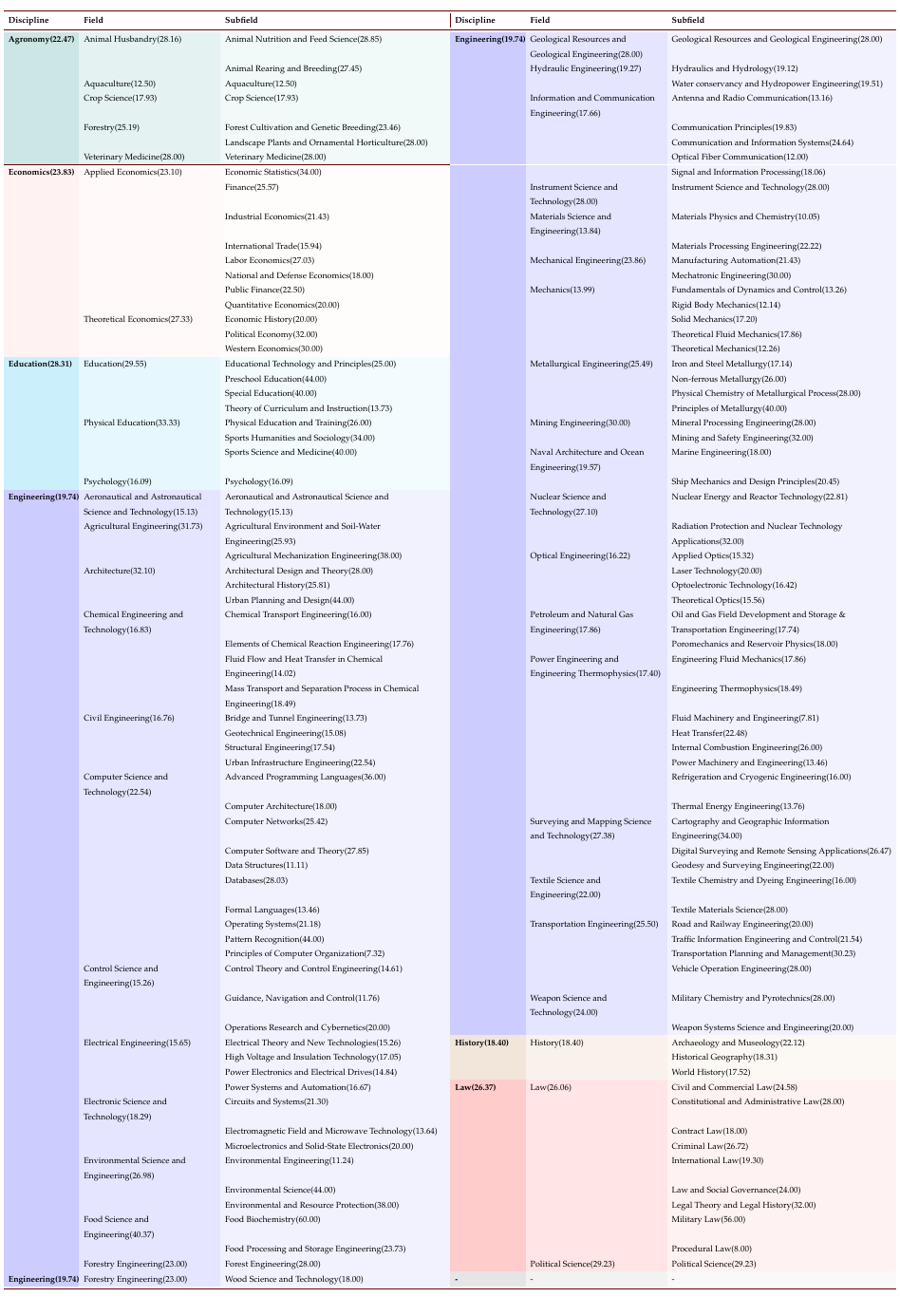} 
\end{subtable}
\vspace{-1.3cm}
\captionsetup{font=small}
\caption{Model Scores Across Three Levels of Disciplines: Qwen2.5-3B.}
\label{tab:Qwen2.5-3B}
\vspace{-0.5cm}
\centeredlinks{listofmodels}{Back to List of Models}{toc}{Back to Table of Contents}{blue}
\end{table}
\clearpage

\newpage
\afterpage{
    \begin{table}[t]
    \centering
    \ContinuedFloat 
    \begin{subtable}[t]{\textwidth}
    \centering
    \includegraphics[width=\textwidth]{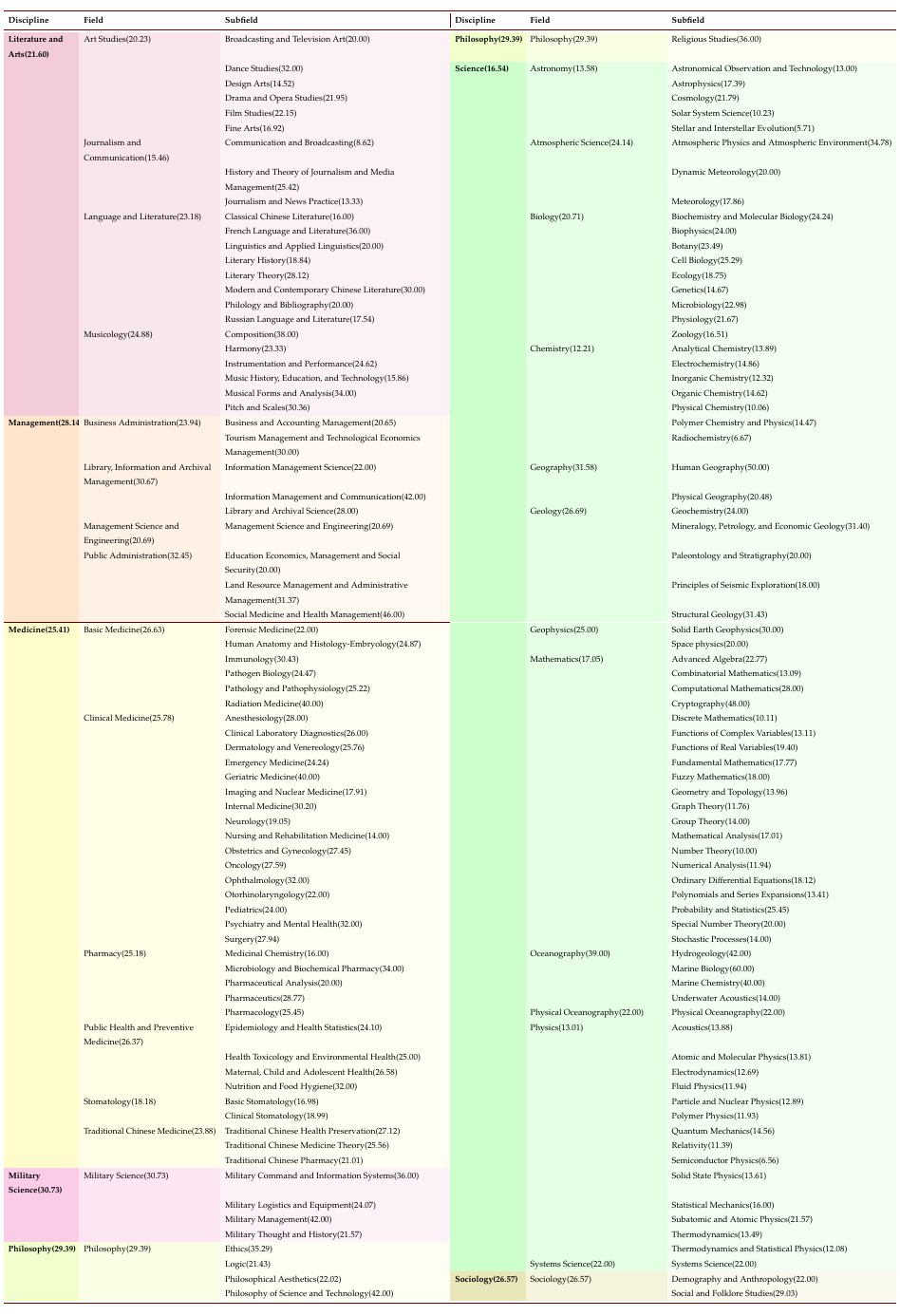} 
    \end{subtable}
    \vspace{-1.1cm}
    \captionsetup{font=small}
    \caption{Continued: Model Scores Across Three Levels of Disciplines: Qwen2.5-3B.}
    \vspace{-0.6cm}
    \centeredlinks{listofmodels}{Back to List of Models}{toc}{Back to Table of Contents}{blue}
    \end{table}
}
\clearpage

\newpage
\vspace{-0.5cm}
\begin{table}[t]
\refstepcounter{models}%
\addcontentsline{csf}{models}{\protect\numberline{\themodels}Llama-3.1-8B}
\centering
\begin{subtable}[t]{1\textwidth}
\centering
\includegraphics[width=\textwidth]{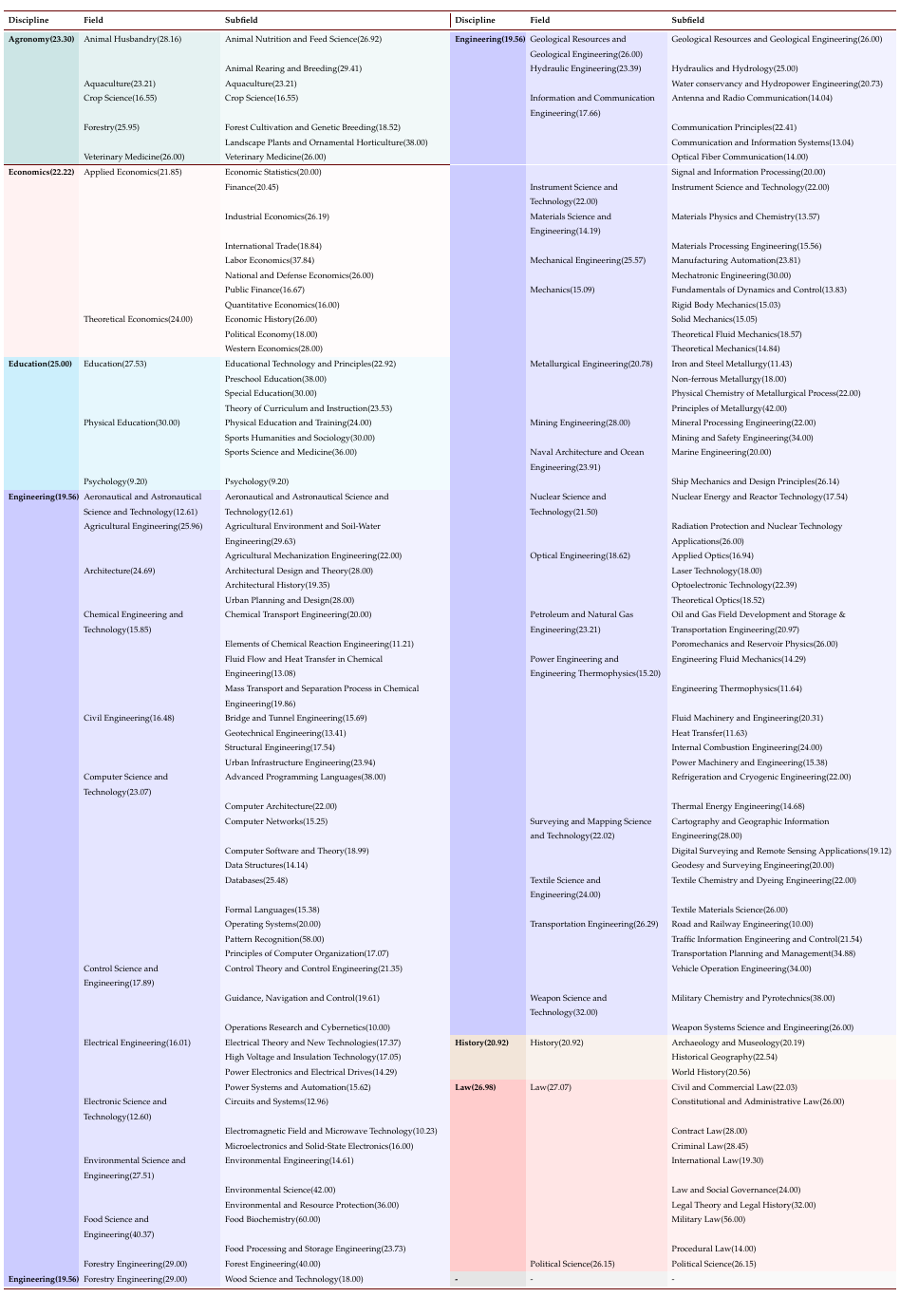} 
\end{subtable}
\vspace{-1.3cm}
\captionsetup{font=small}
\caption{Model Scores Across Three Levels of Disciplines: Llama-3.1-8B.}
\label{tab:Llama-3.1-8B}
\vspace{-0.5cm}
\centeredlinks{listofmodels}{Back to List of Models}{toc}{Back to Table of Contents}{blue}
\end{table}
\clearpage

\newpage
\afterpage{
    \begin{table}[t]
    \centering
    \ContinuedFloat 
    \begin{subtable}[t]{\textwidth}
    \centering
    \includegraphics[width=\textwidth]{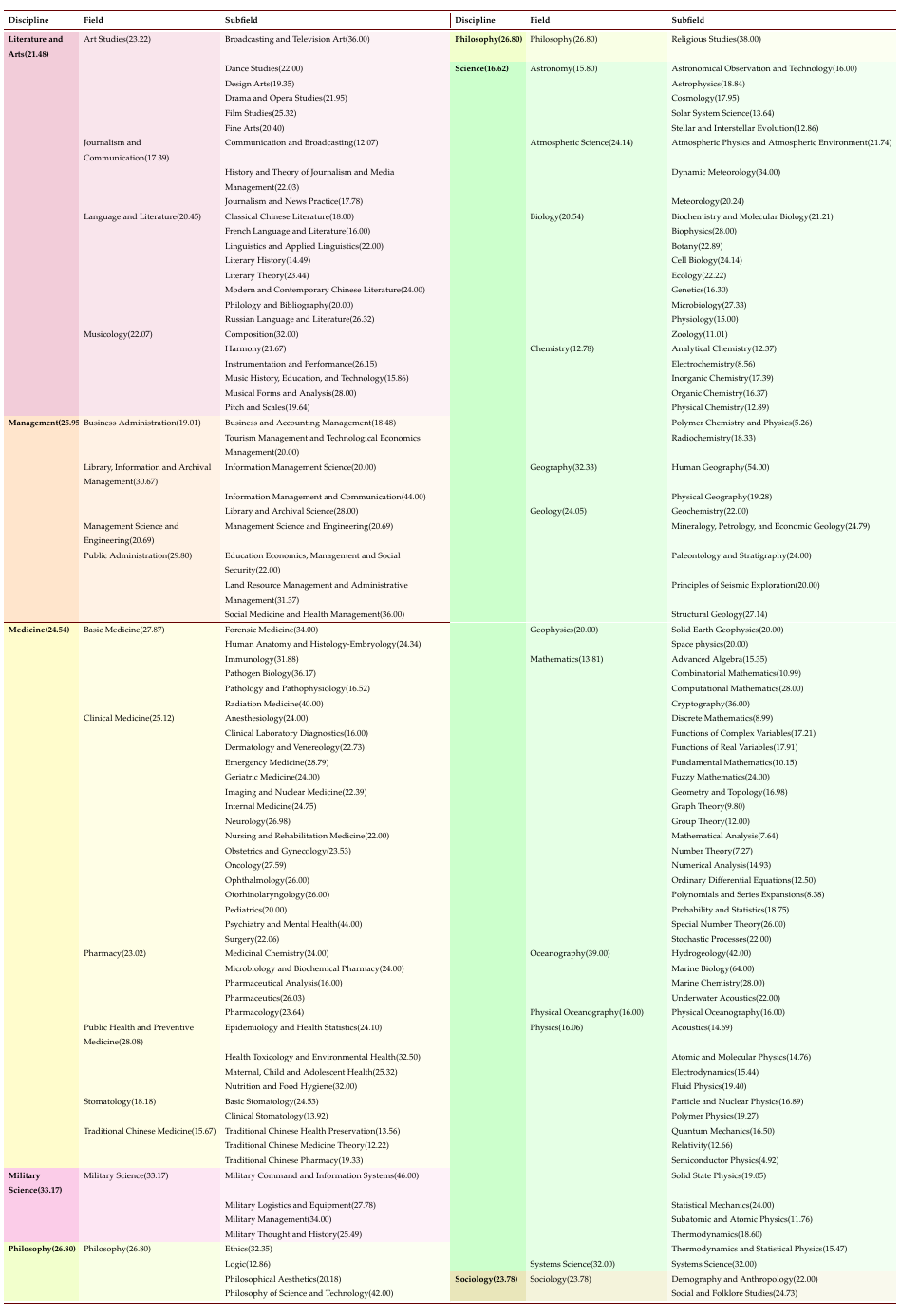} 
    \end{subtable}
    \vspace{-1.1cm}
    \captionsetup{font=small}
    \caption{Continued: Model Scores Across Three Levels of Disciplines: Llama-3.1-8B.}
    \vspace{-0.6cm}
    \centeredlinks{listofmodels}{Back to List of Models}{toc}{Back to Table of Contents}{blue}
    \end{table}
}
\clearpage

\newpage
\vspace{-0.5cm}
\begin{table}[t]
\refstepcounter{models}%
\addcontentsline{csf}{models}{\protect\numberline{\themodels}Mistral-7B-v0.3}
\centering
\begin{subtable}[t]{1\textwidth}
\centering
\includegraphics[width=\textwidth]{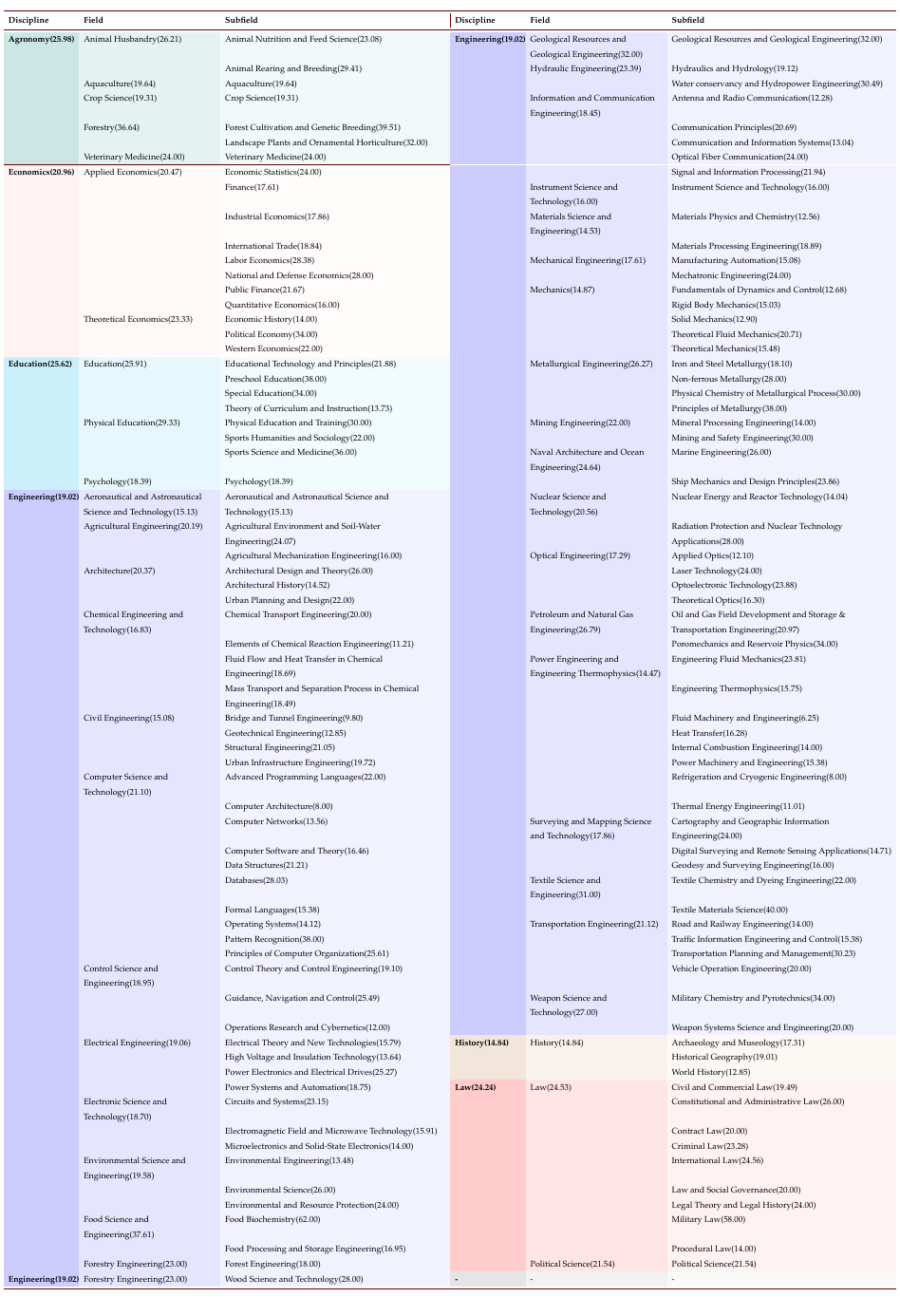} 
\end{subtable}
\vspace{-1.3cm}
\captionsetup{font=small}
\caption{Model Scores Across Three Levels of Disciplines: Mistral-7B-v0.3.}
\label{tab:Mistral-7B-v0.3}
\vspace{-0.5cm}
\centeredlinks{listofmodels}{Back to List of Models}{toc}{Back to Table of Contents}{blue}
\end{table}
\clearpage

\newpage
\afterpage{
    \begin{table}[t]
    \centering
    \ContinuedFloat 
    \begin{subtable}[t]{\textwidth}
    \centering
    \includegraphics[width=\textwidth]{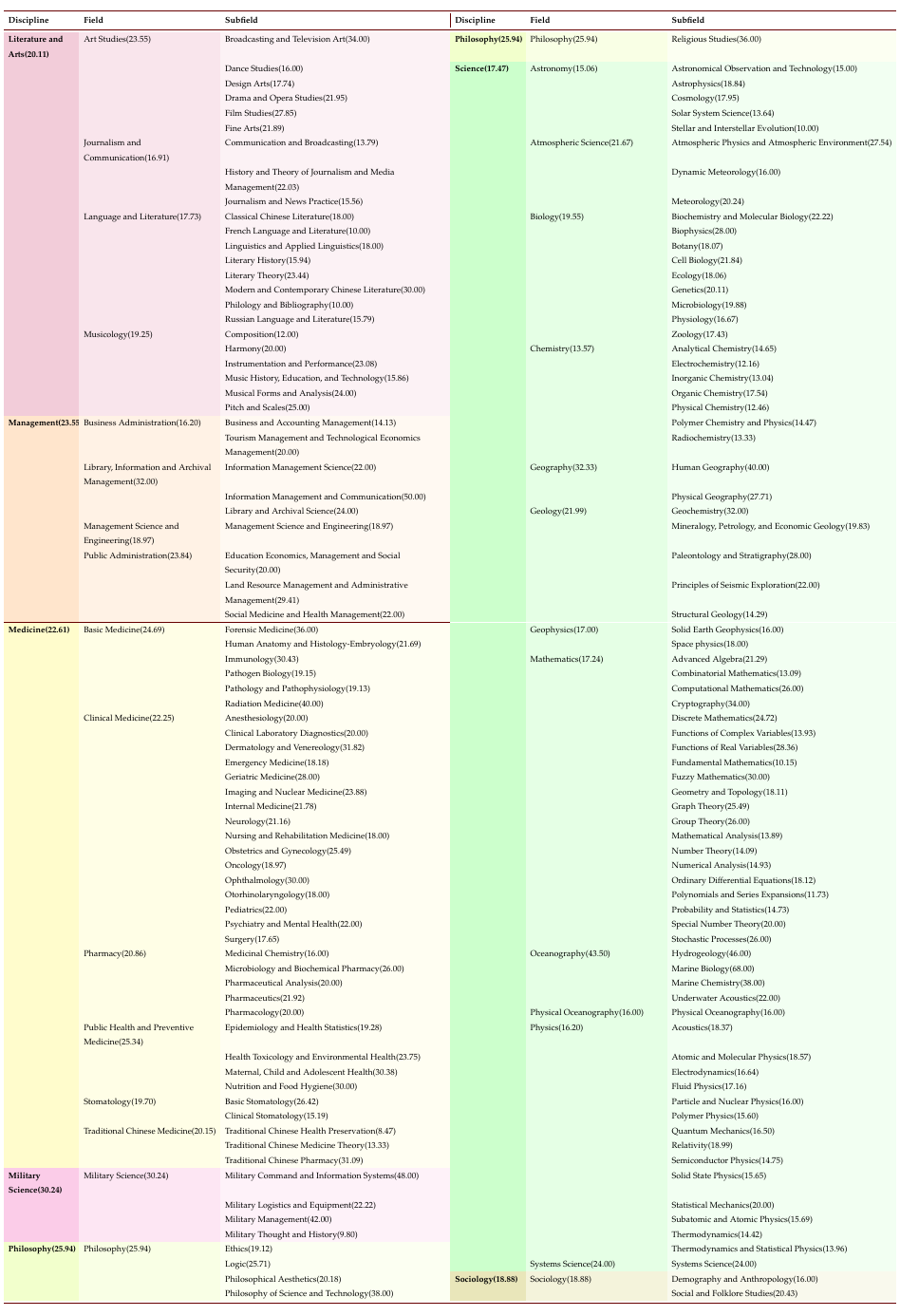} 
    \end{subtable}
    \vspace{-1.1cm}
    \captionsetup{font=small}
    \caption{Continued: Model Scores Across Three Levels of Disciplines: Mistral-7B-v0.3.}
    \vspace{-0.6cm}
    \centeredlinks{listofmodels}{Back to List of Models}{toc}{Back to Table of Contents}{blue}
    \end{table}
}
\clearpage

\newpage
\vspace{-0.5cm}
\begin{table}[t]
\refstepcounter{models}%
\addcontentsline{csf}{models}{\protect\numberline{\themodels}Yi-1.5-6B-Chat}
\centering
\begin{subtable}[t]{1\textwidth}
\centering
\includegraphics[width=\textwidth]{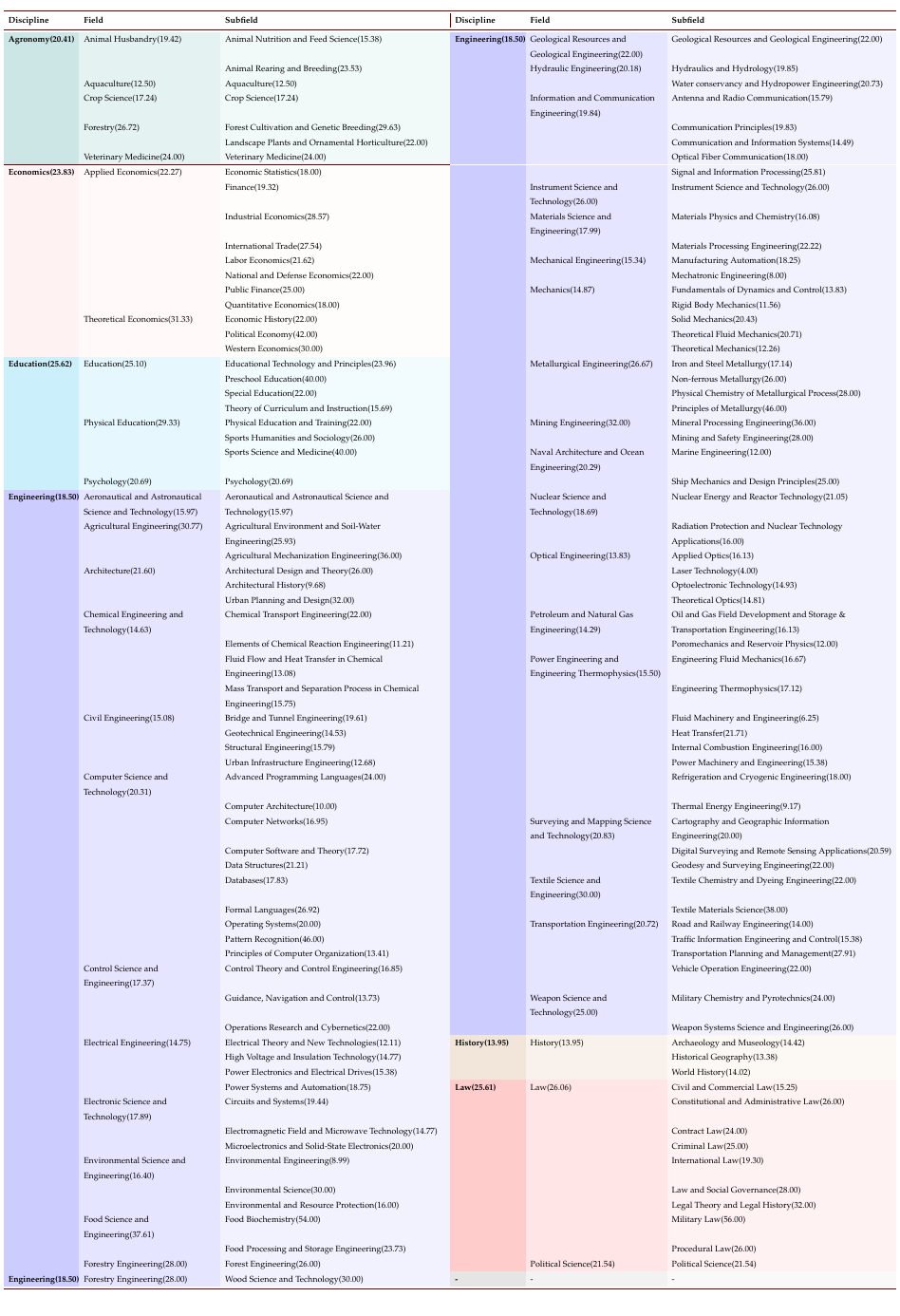} 
\end{subtable}
\vspace{-1.3cm}
\captionsetup{font=small}
\caption{Model Scores Across Three Levels of Disciplines: Yi-1.5-6B-Chat.}
\label{tab:Yi-1.5-6B-Chat}
\vspace{-0.5cm}
\centeredlinks{listofmodels}{Back to List of Models}{toc}{Back to Table of Contents}{blue}
\end{table}
\clearpage

\newpage
\afterpage{
    \begin{table}[t]
    \centering
    \ContinuedFloat 
    \begin{subtable}[t]{\textwidth}
    \centering
    \includegraphics[width=\textwidth]{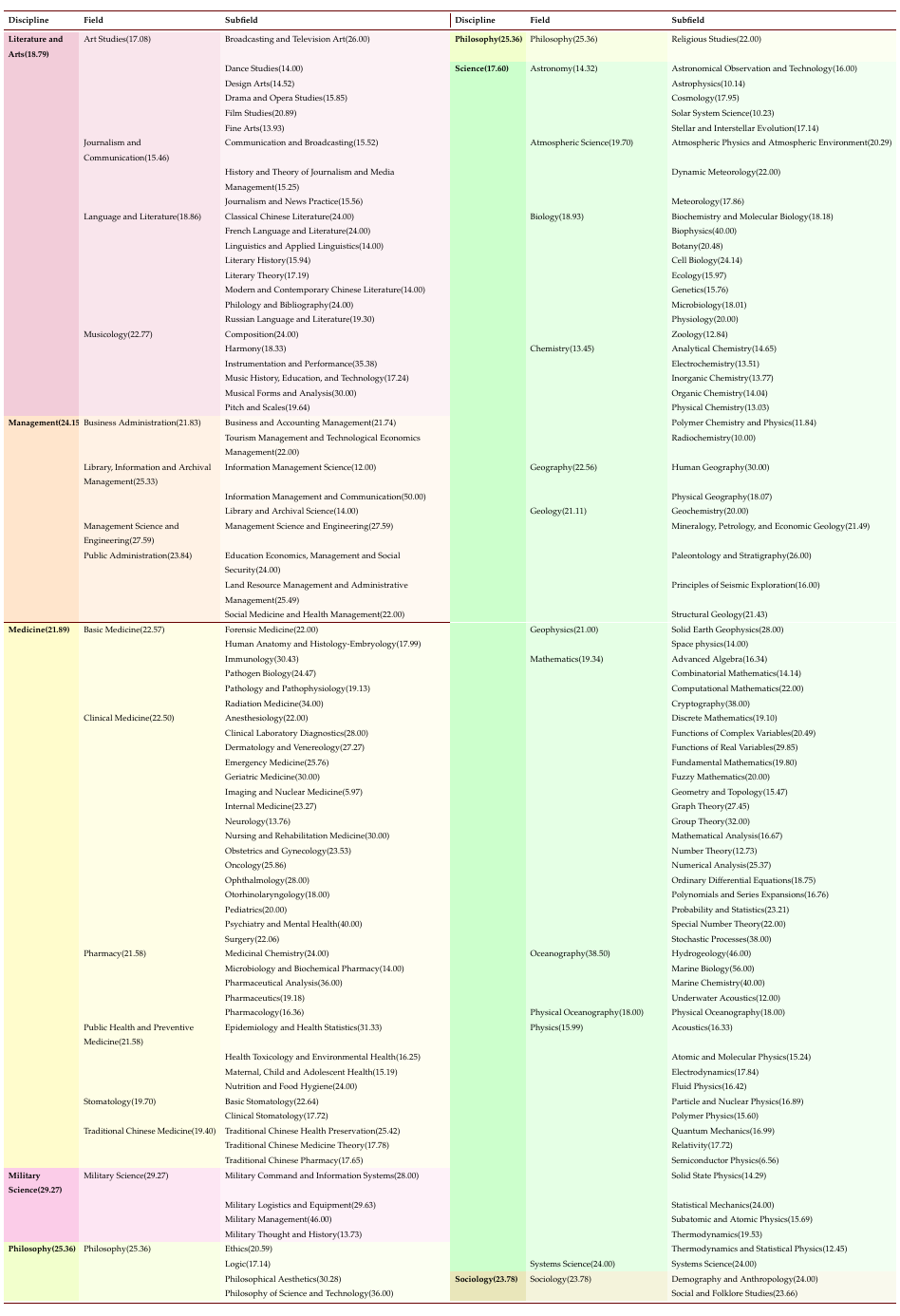} 
    \end{subtable}
    \vspace{-1.1cm}
    \captionsetup{font=small}
    \caption{Continued: Model Scores Across Three Levels of Disciplines: Yi-1.5-6B-Chat.}
    \vspace{-0.6cm}
    \centeredlinks{listofmodels}{Back to List of Models}{toc}{Back to Table of Contents}{blue}
    \end{table}
}
\clearpage

\newpage
\vspace{-0.5cm}
\begin{table}[t]
\refstepcounter{models}%
\addcontentsline{csf}{models}{\protect\numberline{\themodels}Qwen2.5-1.5B-Instruct}
\centering
\begin{subtable}[t]{1\textwidth}
\centering
\includegraphics[width=\textwidth]{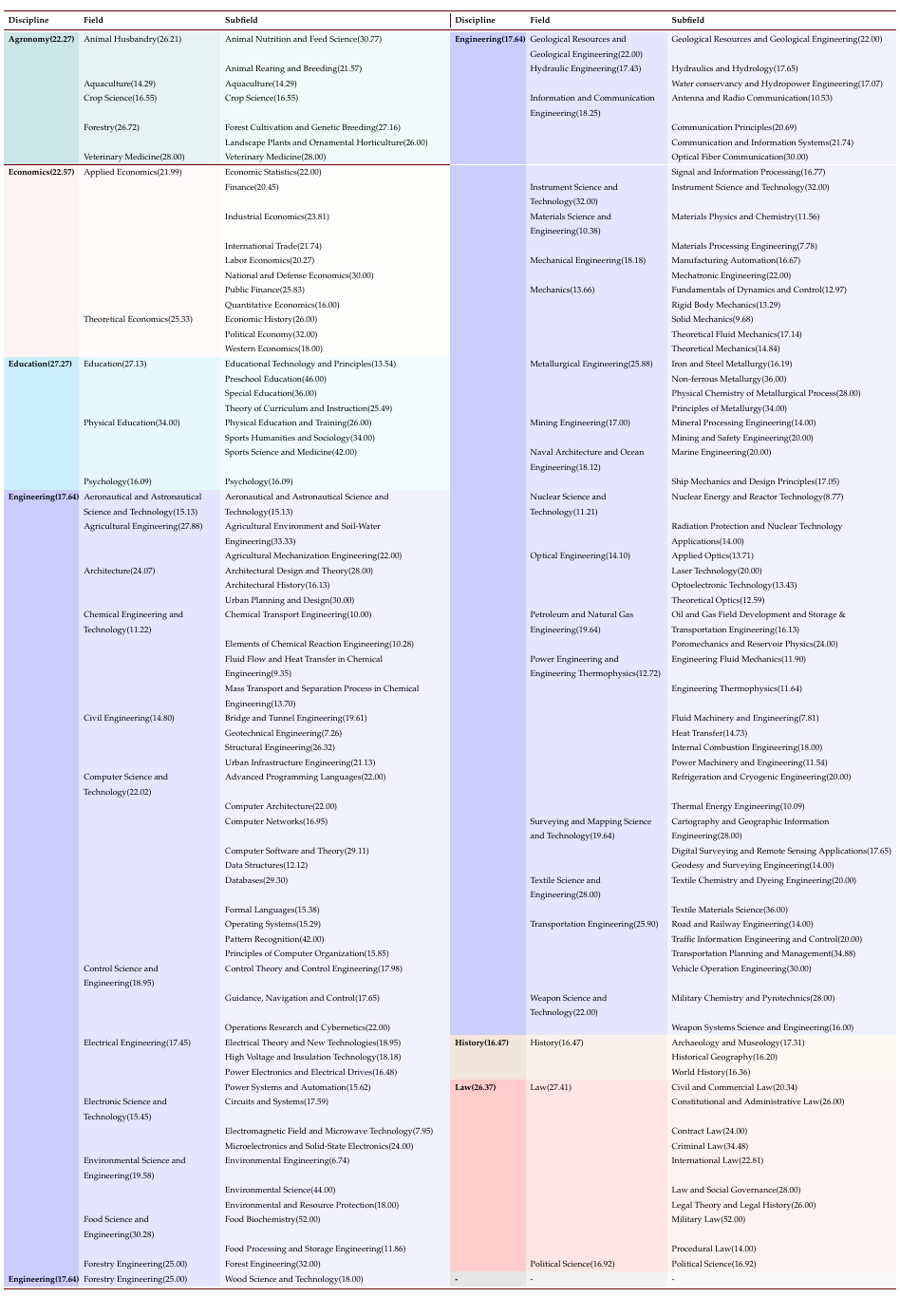} 
\end{subtable}
\vspace{-1.3cm}
\captionsetup{font=small}
\caption{Model Scores Across Three Levels of Disciplines: Qwen2.5-1.5B-Instruct.}
\label{tab:Qwen2.5-1.5B-Instruct}
\vspace{-0.5cm}
\centeredlinks{listofmodels}{Back to List of Models}{toc}{Back to Table of Contents}{blue}
\end{table}
\clearpage

\newpage
\afterpage{
    \begin{table}[t]
    \centering
    \ContinuedFloat 
    \begin{subtable}[t]{\textwidth}
    \centering
    \includegraphics[width=\textwidth]{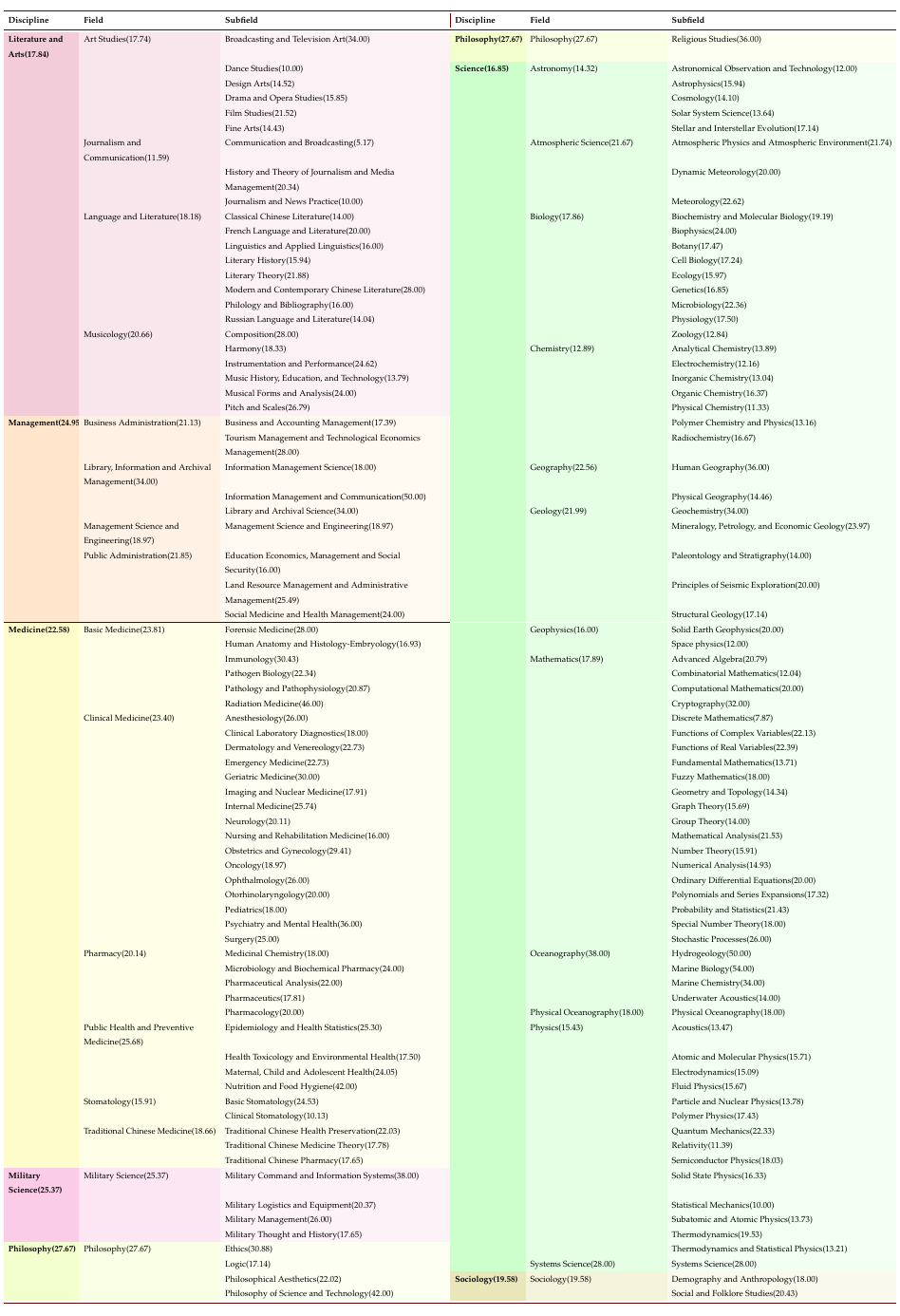} 
    \end{subtable}
    \vspace{-1.1cm}
    \captionsetup{font=small}
    \caption{Continued: Model Scores Across Three Levels of Disciplines: Qwen2.5-1.5B-Instruct.}
    \vspace{-0.6cm}
    \centeredlinks{listofmodels}{Back to List of Models}{toc}{Back to Table of Contents}{blue}
    \end{table}
}
\clearpage

\newpage
\vspace{-0.5cm}
\begin{table}[t]
\refstepcounter{models}%
\addcontentsline{csf}{models}{\protect\numberline{\themodels}OLMo-2-1124-13B-Instruct}
\centering
\begin{subtable}[t]{1\textwidth}
\centering
\includegraphics[width=\textwidth]{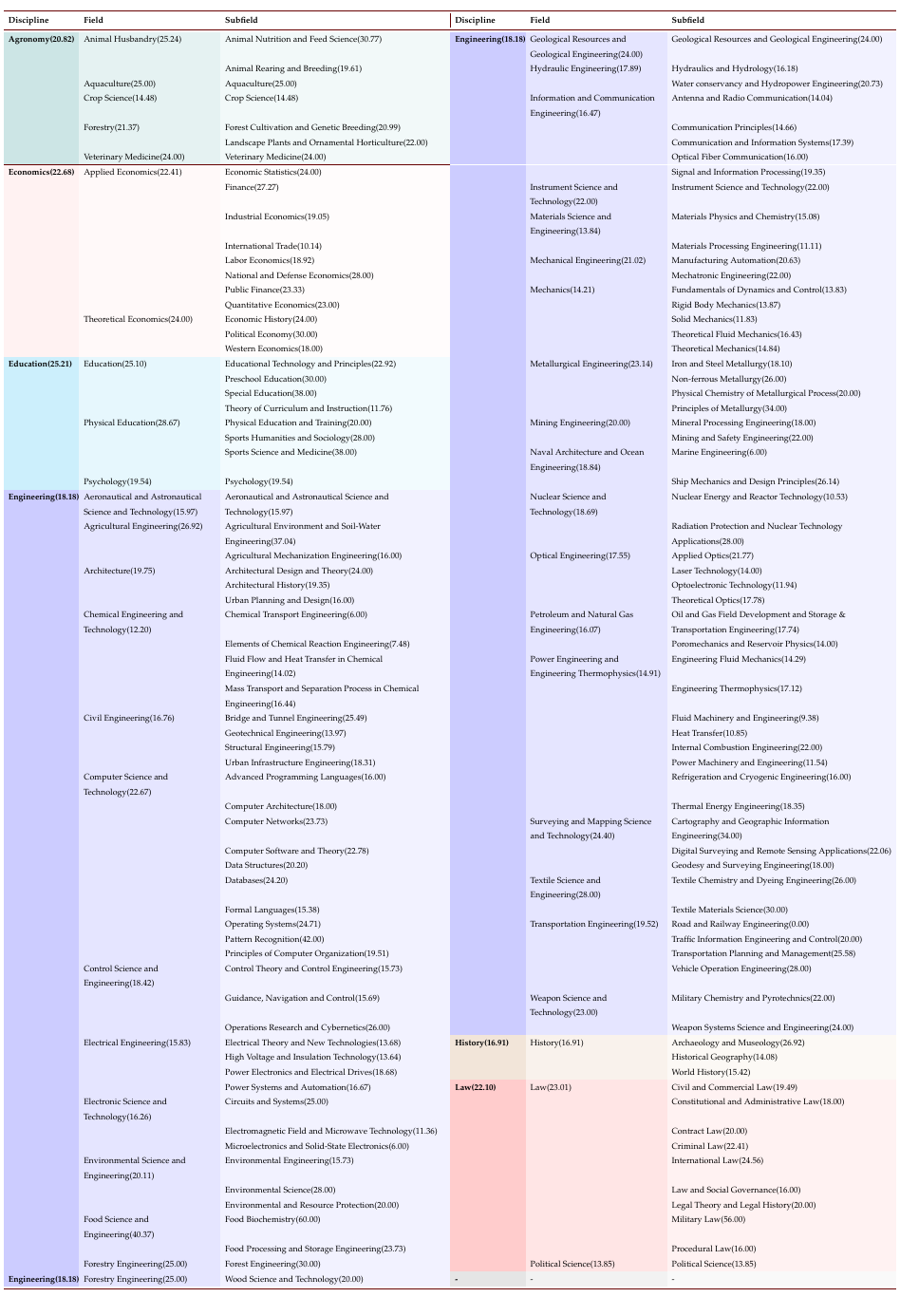} 
\end{subtable}
\vspace{-1.3cm}
\captionsetup{font=small}
\caption{Model Scores Across Three Levels of Disciplines: OLMo-2-1124-13B-Instruct.}
\label{tab:OLMo-2-1124-13B-Instruct}
\vspace{-0.5cm}
\centeredlinks{listofmodels}{Back to List of Models}{toc}{Back to Table of Contents}{blue}
\end{table}
\clearpage

\newpage
\afterpage{
    \begin{table}[t]
    \centering
    \ContinuedFloat 
    \begin{subtable}[t]{\textwidth}
    \centering
    \includegraphics[width=\textwidth]{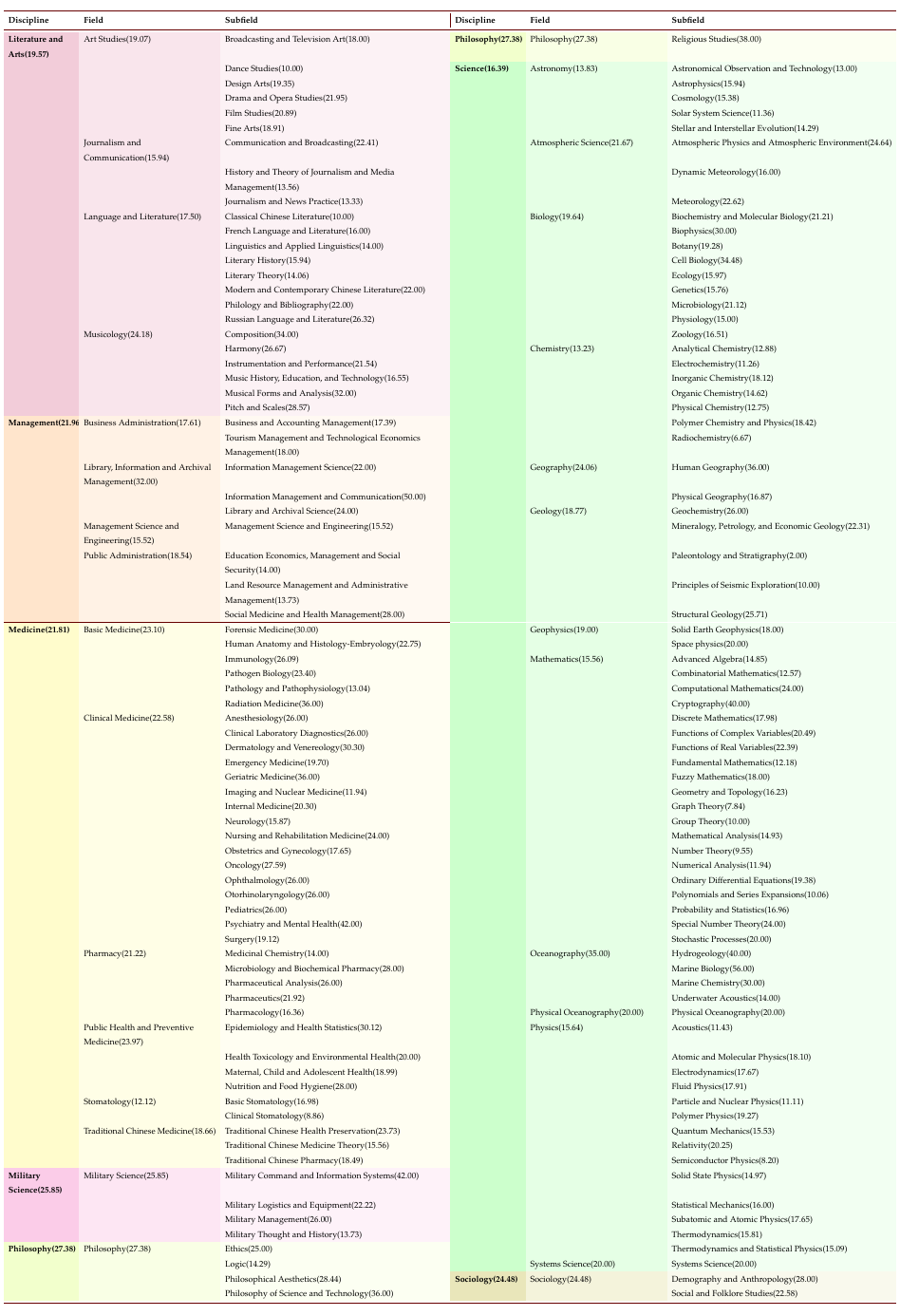} 
    \end{subtable}
    \vspace{-1.1cm}
    \captionsetup{font=small}
    \caption{Continued: Model Scores Across Three Levels of Disciplines: OLMo-2-1124-13B-Instruct.}
    \vspace{-0.6cm}
    \centeredlinks{listofmodels}{Back to List of Models}{toc}{Back to Table of Contents}{blue}
    \end{table}
}
\clearpage

\newpage
\vspace{-0.5cm}
\begin{table}[t]
\refstepcounter{models}%
\addcontentsline{csf}{models}{\protect\numberline{\themodels}gemma-2-2b-it}
\centering
\begin{subtable}[t]{1\textwidth}
\centering
\includegraphics[width=\textwidth]{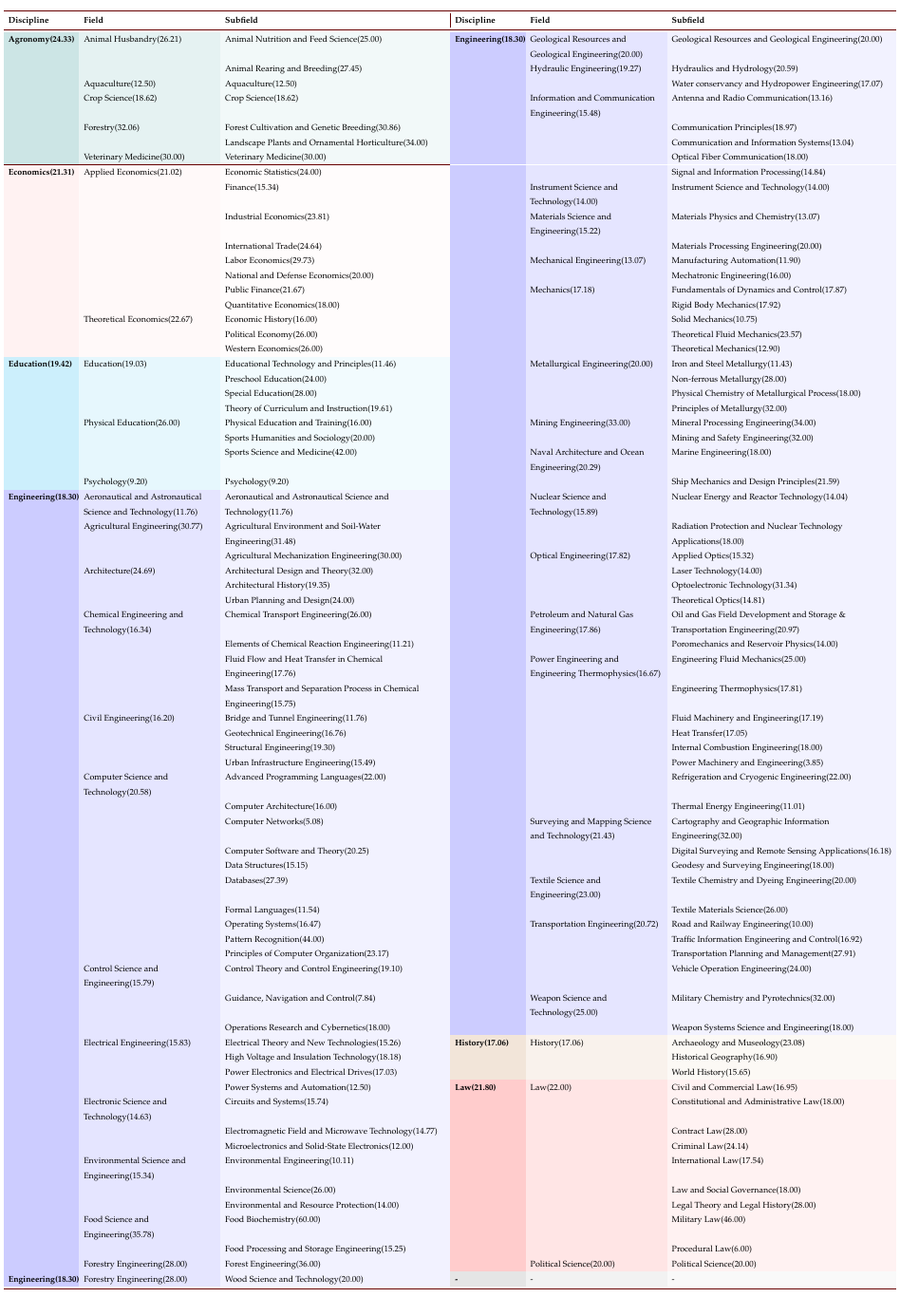} 
\end{subtable}
\vspace{-1.3cm}
\captionsetup{font=small}
\caption{Model Scores Across Three Levels of Disciplines: gemma-2-2b-it.}
\label{tab:gemma-2-2b-it}
\vspace{-0.5cm}
\centeredlinks{listofmodels}{Back to List of Models}{toc}{Back to Table of Contents}{blue}
\end{table}
\clearpage

\newpage
\afterpage{
    \begin{table}[t]
    \centering
    \ContinuedFloat 
    \begin{subtable}[t]{\textwidth}
    \centering
    \includegraphics[width=\textwidth]{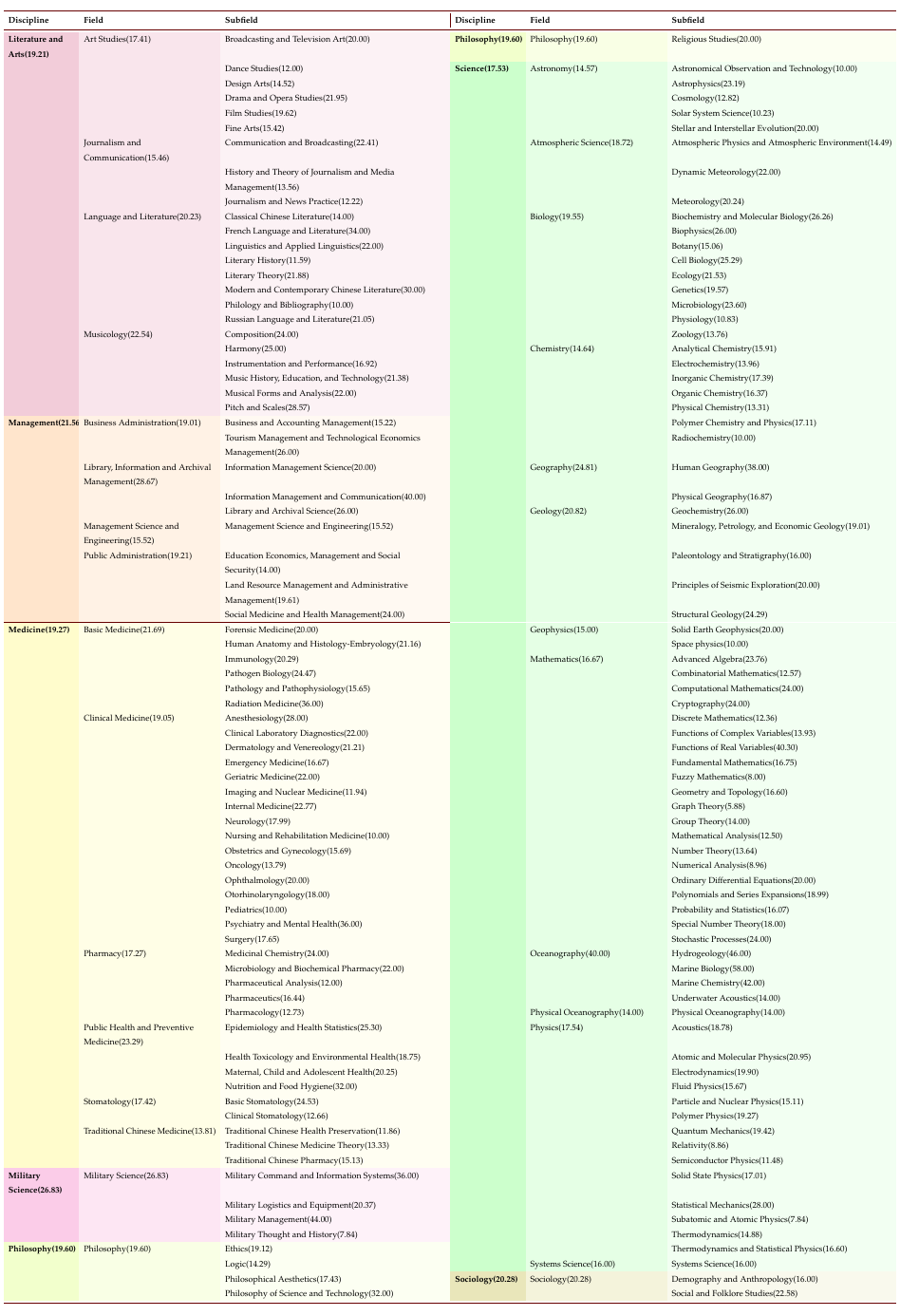} 
    \end{subtable}
    \vspace{-1.1cm}
    \captionsetup{font=small}
    \caption{Continued: Model Scores Across Three Levels of Disciplines: gemma-2-2b-it.}
    \vspace{-0.6cm}
    \centeredlinks{listofmodels}{Back to List of Models}{toc}{Back to Table of Contents}{blue}
    \end{table}
}
\clearpage

\newpage
\vspace{-0.5cm}
\begin{table}[t]
\refstepcounter{models}%
\addcontentsline{csf}{models}{\protect\numberline{\themodels}granite-3.1-2b-instruct}
\centering
\begin{subtable}[t]{1\textwidth}
\centering
\includegraphics[width=\textwidth]{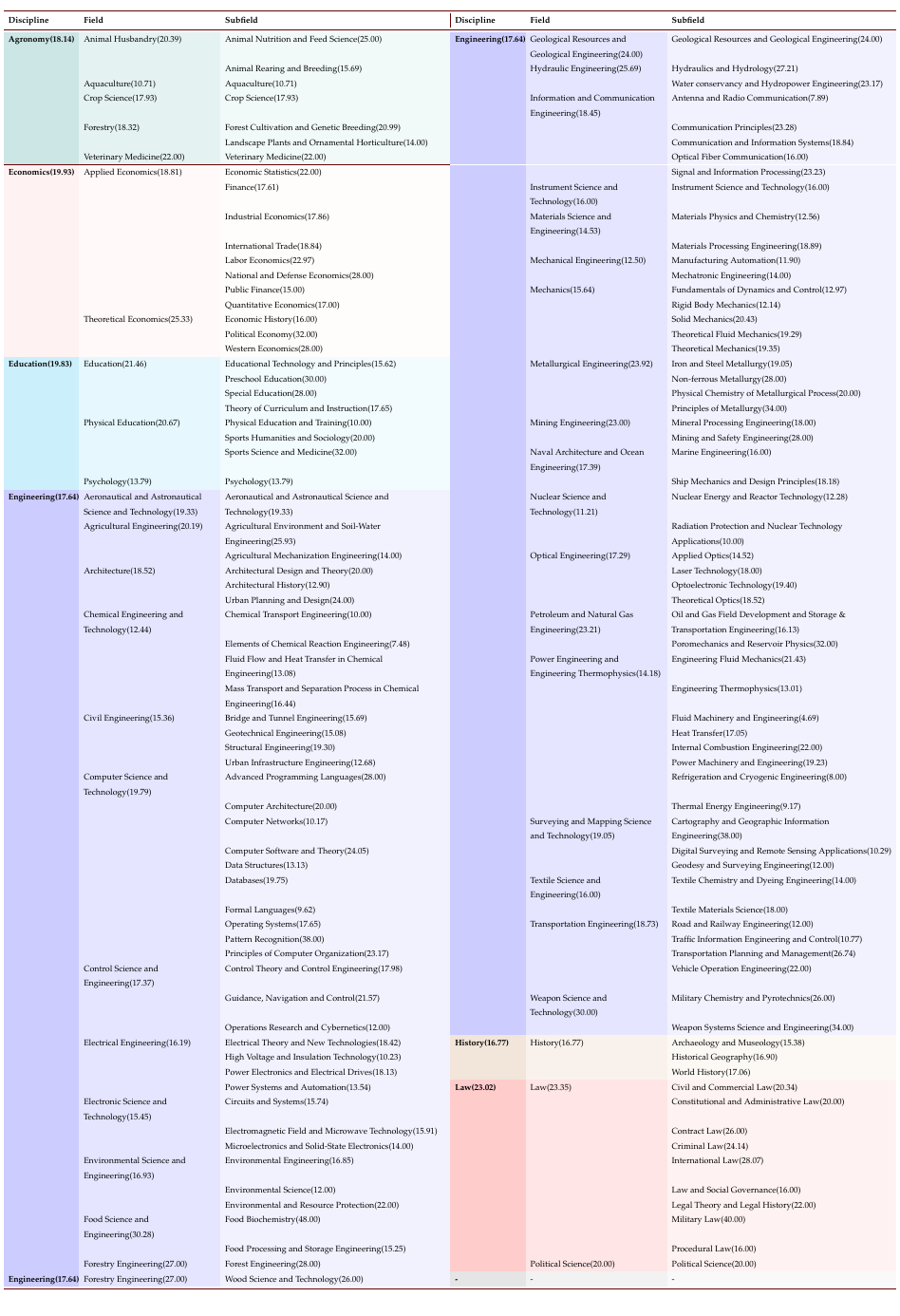} 
\end{subtable}
\vspace{-1.3cm}
\captionsetup{font=small}
\caption{Model Scores Across Three Levels of Disciplines: granite-3.1-2b-instruct.}
\label{tab:granite-3.1-2b-instruct}
\vspace{-0.5cm}
\centeredlinks{listofmodels}{Back to List of Models}{toc}{Back to Table of Contents}{blue}
\end{table}
\clearpage

\newpage
\afterpage{
    \begin{table}[t]
    \centering
    \ContinuedFloat 
    \begin{subtable}[t]{\textwidth}
    \centering
    \includegraphics[width=\textwidth]{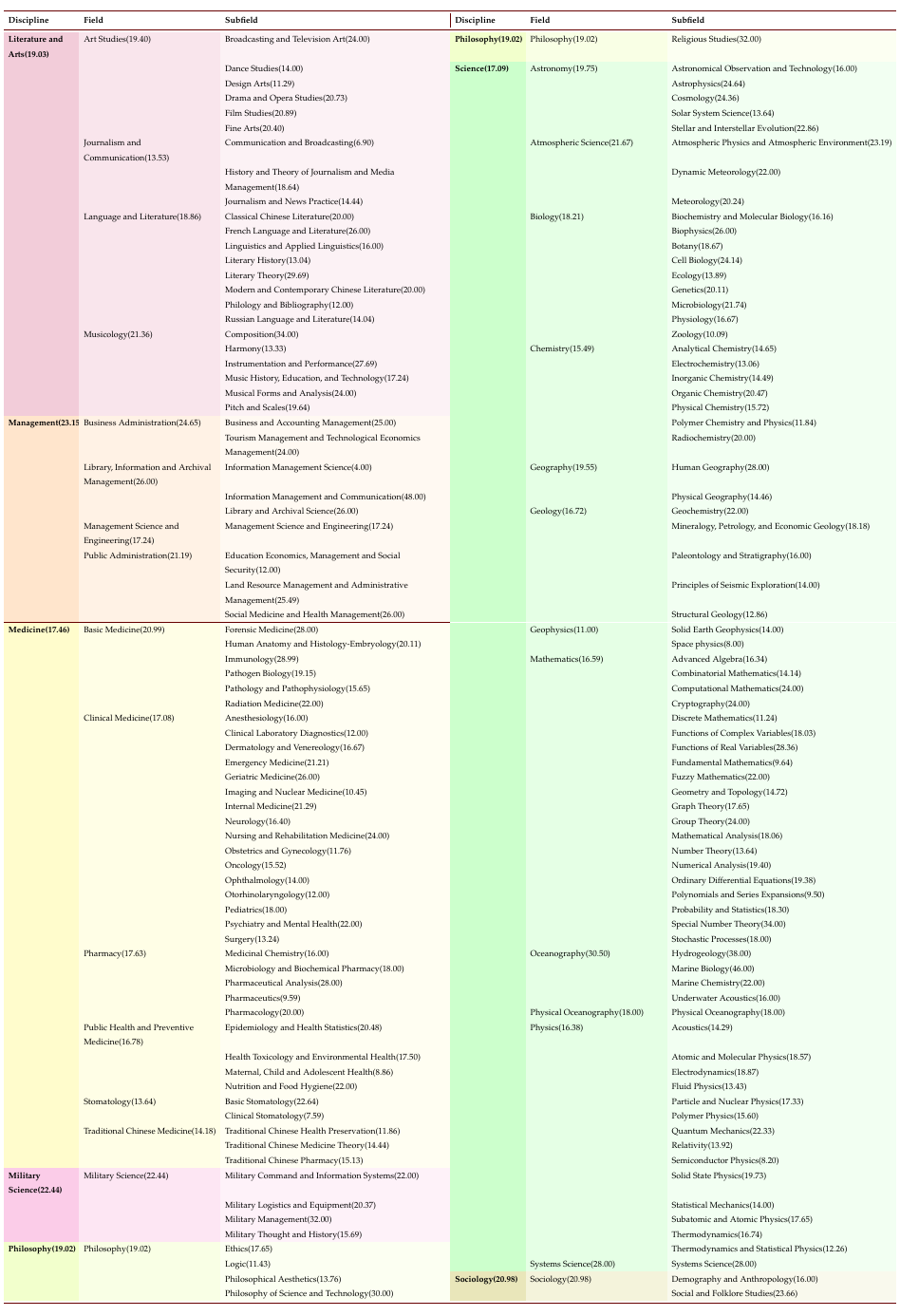} 
    \end{subtable}
    \vspace{-1.1cm}
    \captionsetup{font=small}
    \caption{Continued: Model Scores Across Three Levels of Disciplines: granite-3.1-2b-instruct.}
    \vspace{-0.6cm}
    \centeredlinks{listofmodels}{Back to List of Models}{toc}{Back to Table of Contents}{blue}
    \end{table}
}
\clearpage

\newpage
\vspace{-0.5cm}
\begin{table}[t]
\refstepcounter{models}%
\addcontentsline{csf}{models}{\protect\numberline{\themodels}Mistral-7B-Instruct-v0.3}
\centering
\begin{subtable}[t]{1\textwidth}
\centering
\includegraphics[width=\textwidth]{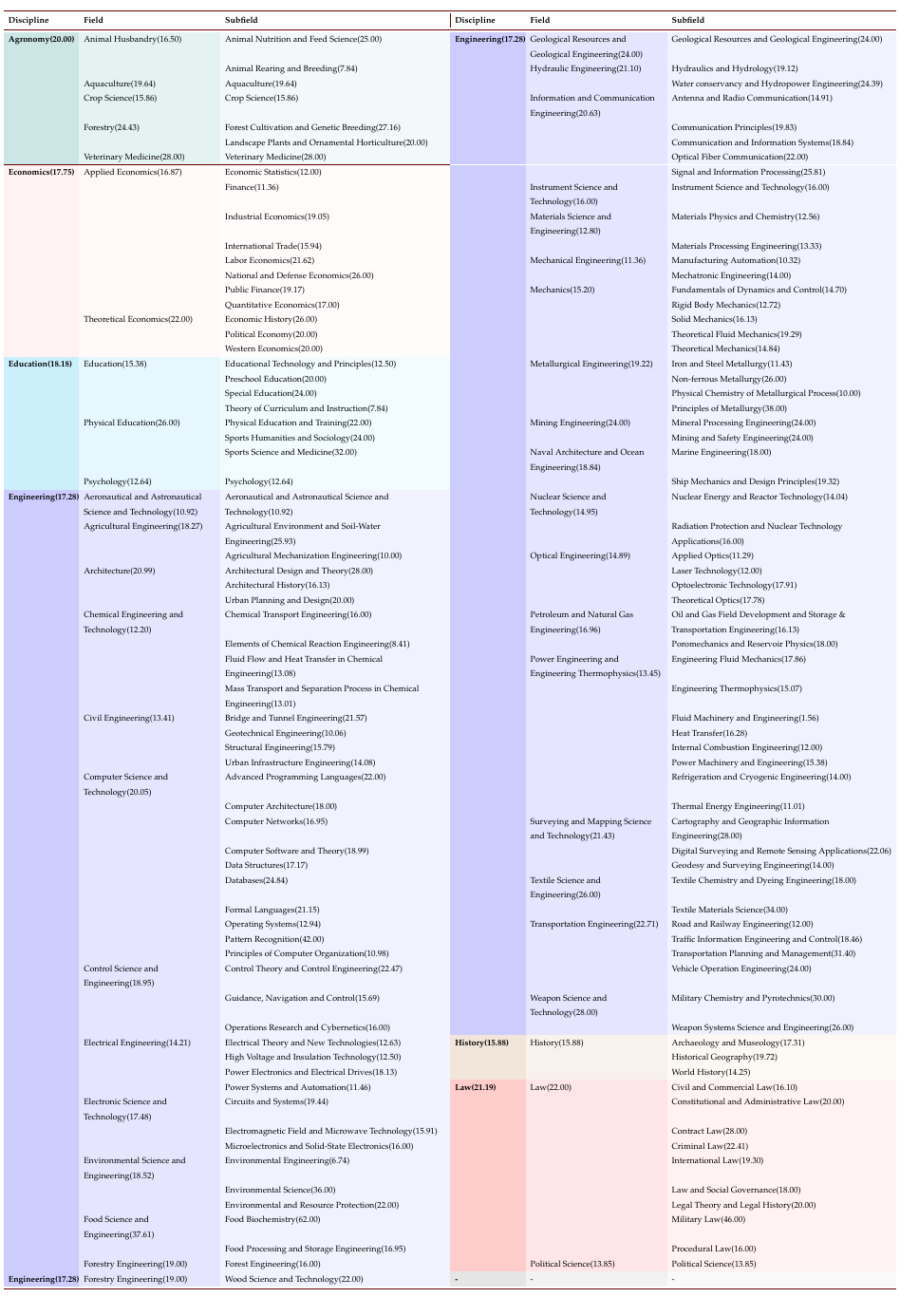} 
\end{subtable}
\vspace{-1.3cm}
\captionsetup{font=small}
\caption{Model Scores Across Three Levels of Disciplines: Mistral-7B-Instruct-v0.3.}
\label{tab:Mistral-7B-Instruct-v0.3}
\vspace{-0.5cm}
\centeredlinks{listofmodels}{Back to List of Models}{toc}{Back to Table of Contents}{blue}
\end{table}
\clearpage

\newpage
\afterpage{
    \begin{table}[t]
    \centering
    \ContinuedFloat 
    \begin{subtable}[t]{\textwidth}
    \centering
    \includegraphics[width=\textwidth]{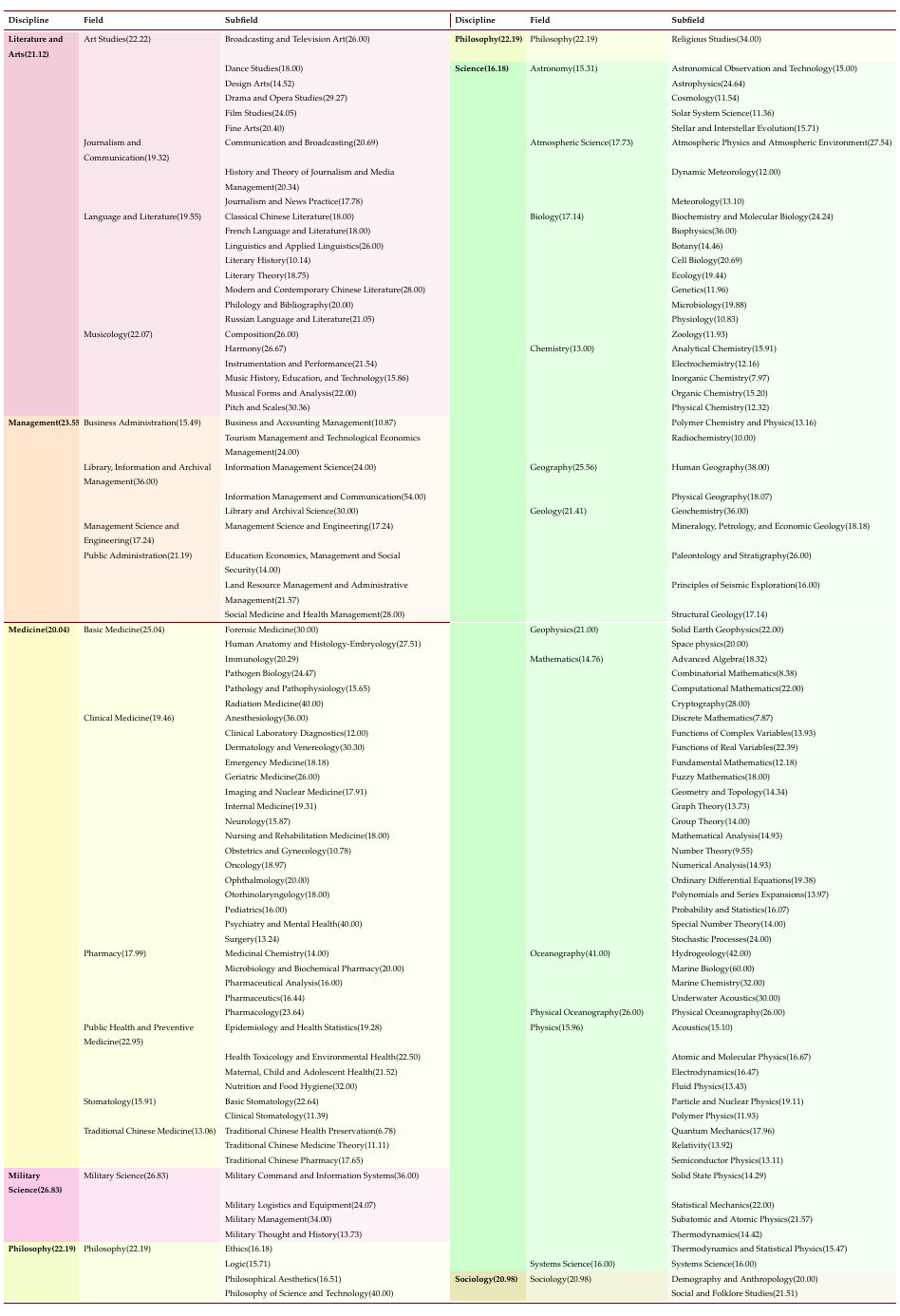} 
    \end{subtable}
    \vspace{-1.1cm}
    \captionsetup{font=small}
    \caption{Continued: Model Scores Across Three Levels of Disciplines: Mistral-7B-Instruct-v0.3.}
    \vspace{-0.6cm}
    \centeredlinks{listofmodels}{Back to List of Models}{toc}{Back to Table of Contents}{blue}
    \end{table}
}
\clearpage

\newpage
\vspace{-0.5cm}
\begin{table}[t]
\refstepcounter{models}%
\addcontentsline{csf}{models}{\protect\numberline{\themodels}Qwen2.5-1.5B}
\centering
\begin{subtable}[t]{1\textwidth}
\centering
\includegraphics[width=\textwidth]{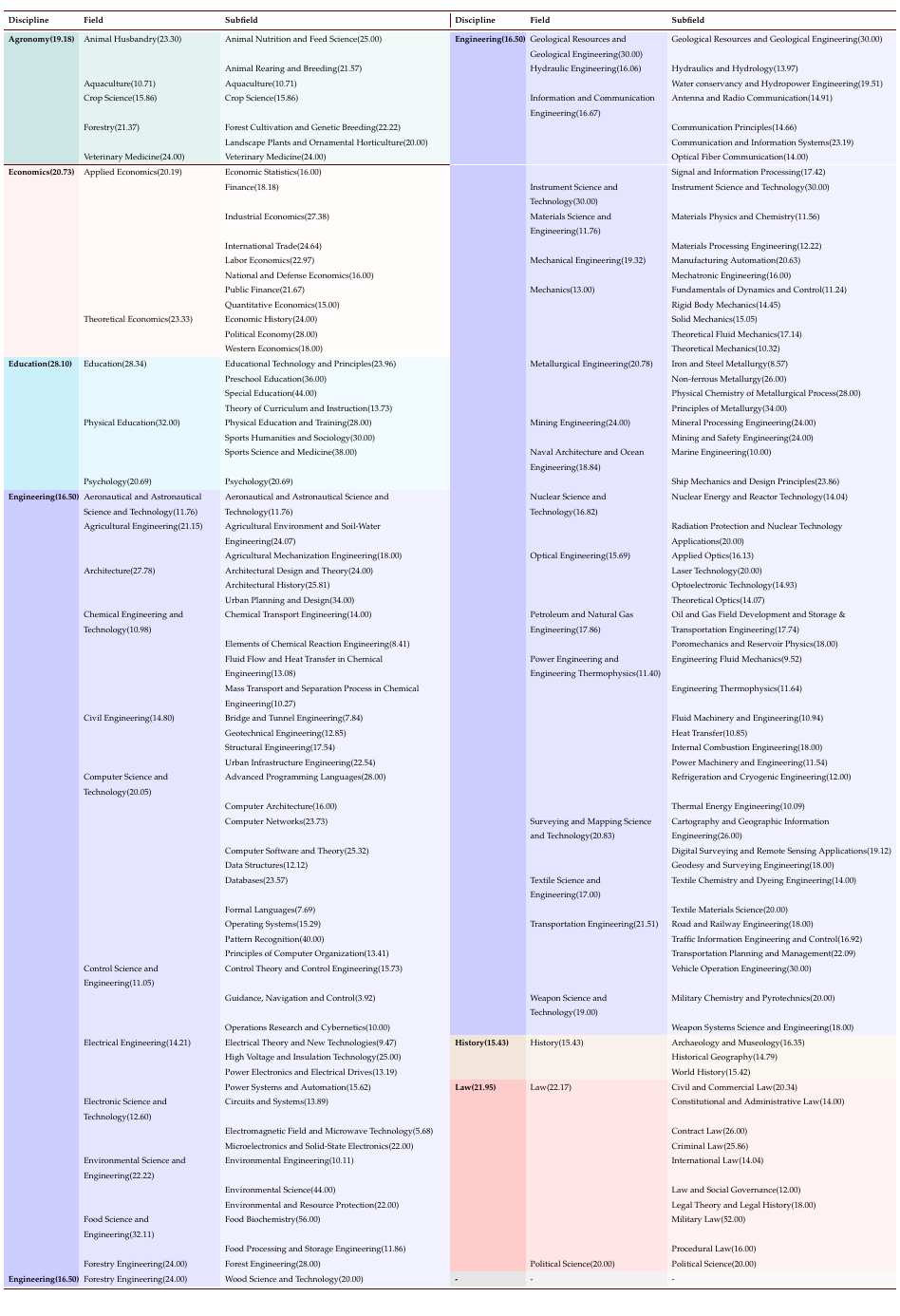} 
\end{subtable}
\vspace{-1.3cm}
\captionsetup{font=small}
\caption{Model Scores Across Three Levels of Disciplines: Qwen2.5-1.5B.}
\label{tab:Qwen2.5-1.5B}
\vspace{-0.5cm}
\centeredlinks{listofmodels}{Back to List of Models}{toc}{Back to Table of Contents}{blue}
\end{table}
\clearpage

\newpage
\afterpage{
    \begin{table}[t]
    \centering
    \ContinuedFloat 
    \begin{subtable}[t]{\textwidth}
    \centering
    \includegraphics[width=\textwidth]{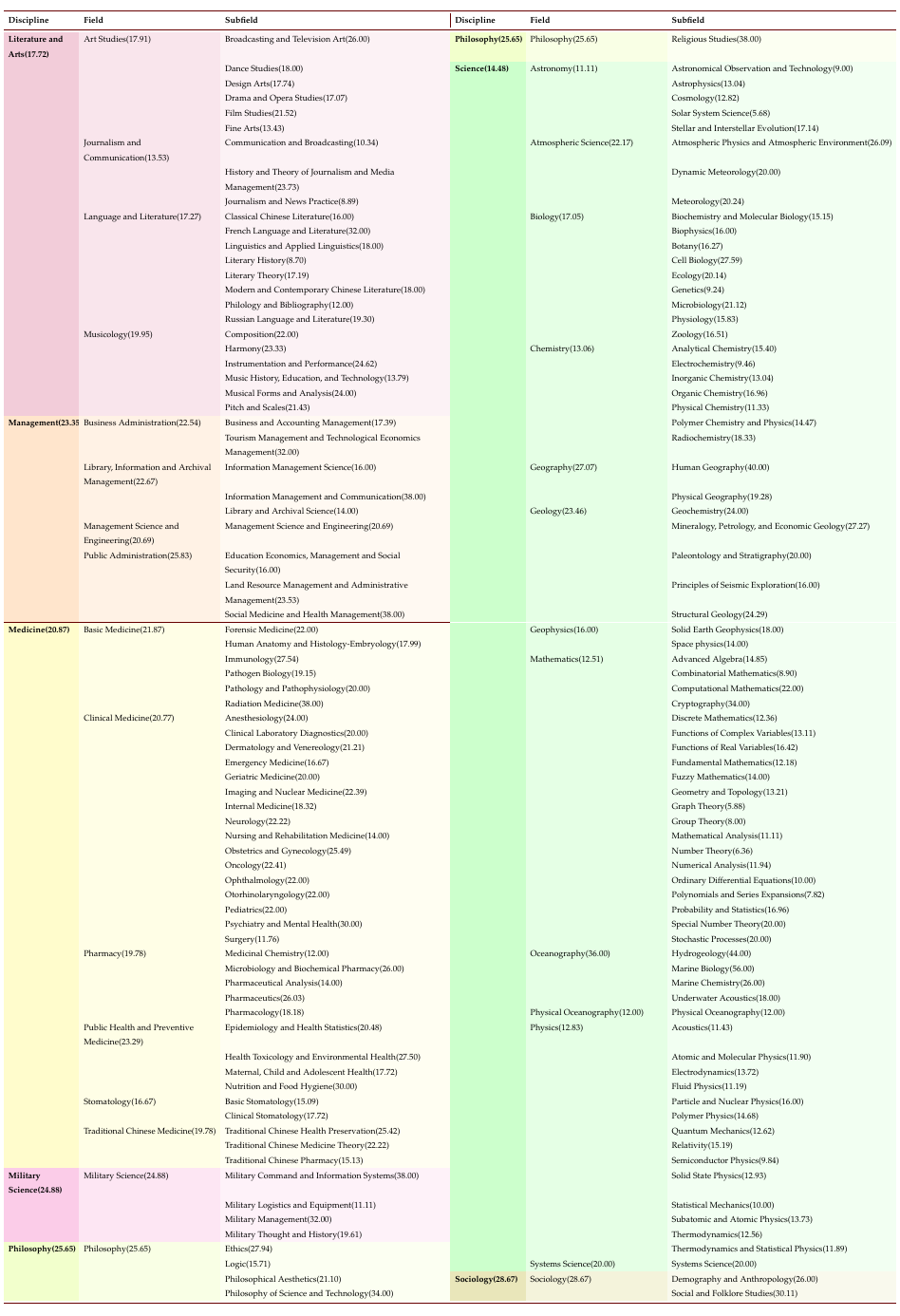} 
    \end{subtable}
    \vspace{-1.1cm}
    \captionsetup{font=small}
    \caption{Continued: Model Scores Across Three Levels of Disciplines: Qwen2.5-1.5B.}
    \vspace{-0.6cm}
    \centeredlinks{listofmodels}{Back to List of Models}{toc}{Back to Table of Contents}{blue}
    \end{table}
}
\clearpage

\newpage
\vspace{-0.5cm}
\begin{table}[t]
\refstepcounter{models}%
\addcontentsline{csf}{models}{\protect\numberline{\themodels}MAP-Neo-7B-Instruct-v0.1}
\centering
\begin{subtable}[t]{1\textwidth}
\centering
\includegraphics[width=\textwidth]{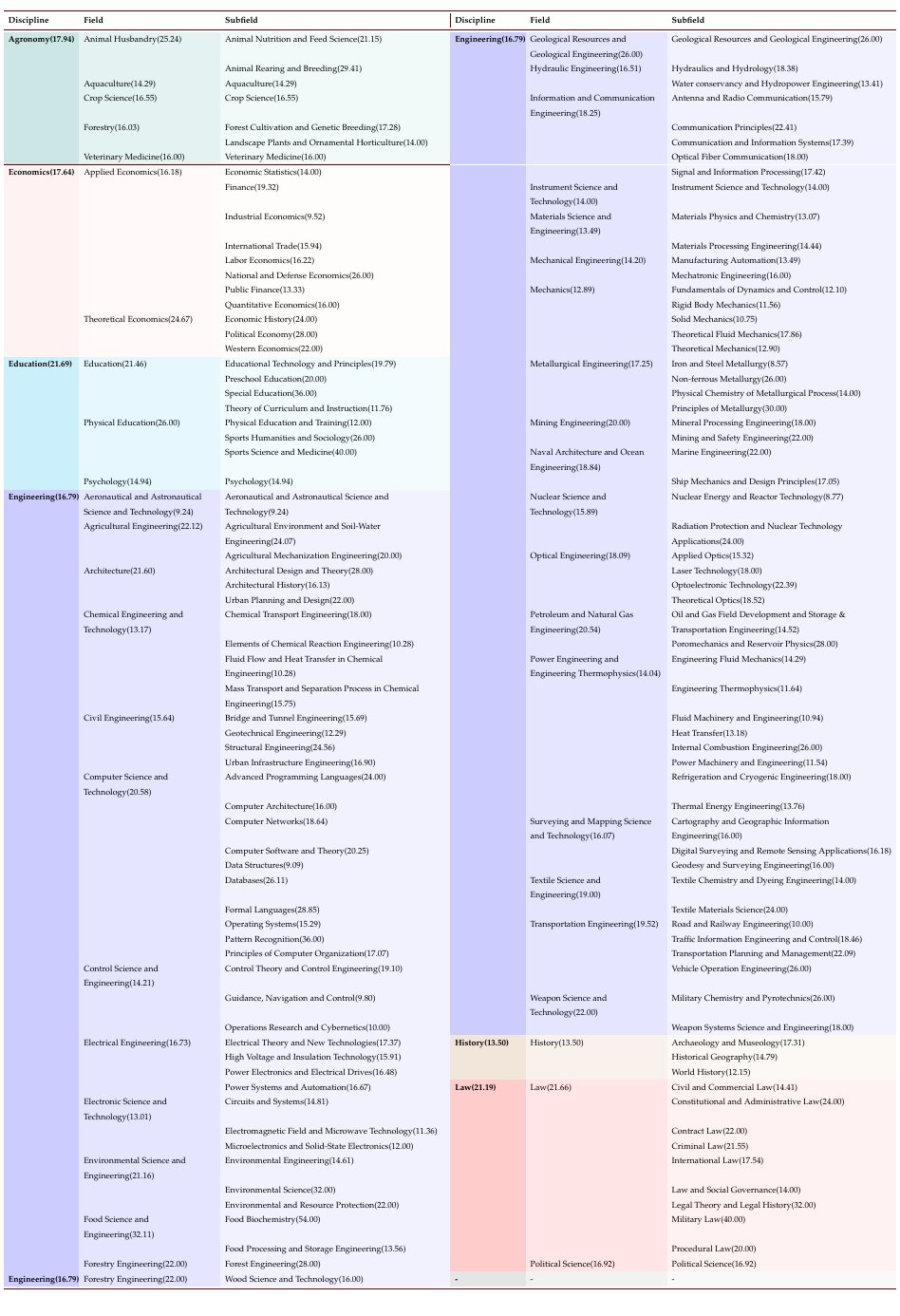} 
\end{subtable}
\vspace{-1.3cm}
\captionsetup{font=small}
\caption{Model Scores Across Three Levels of Disciplines: MAP-Neo-7B-Instruct-v0.1.}
\label{tab:MAP-Neo-7B-Instruct-v0.1}
\vspace{-0.5cm}
\centeredlinks{listofmodels}{Back to List of Models}{toc}{Back to Table of Contents}{blue}
\end{table}
\clearpage

\newpage
\afterpage{
    \begin{table}[t]
    \centering
    \ContinuedFloat 
    \begin{subtable}[t]{\textwidth}
    \centering
    \includegraphics[width=\textwidth]{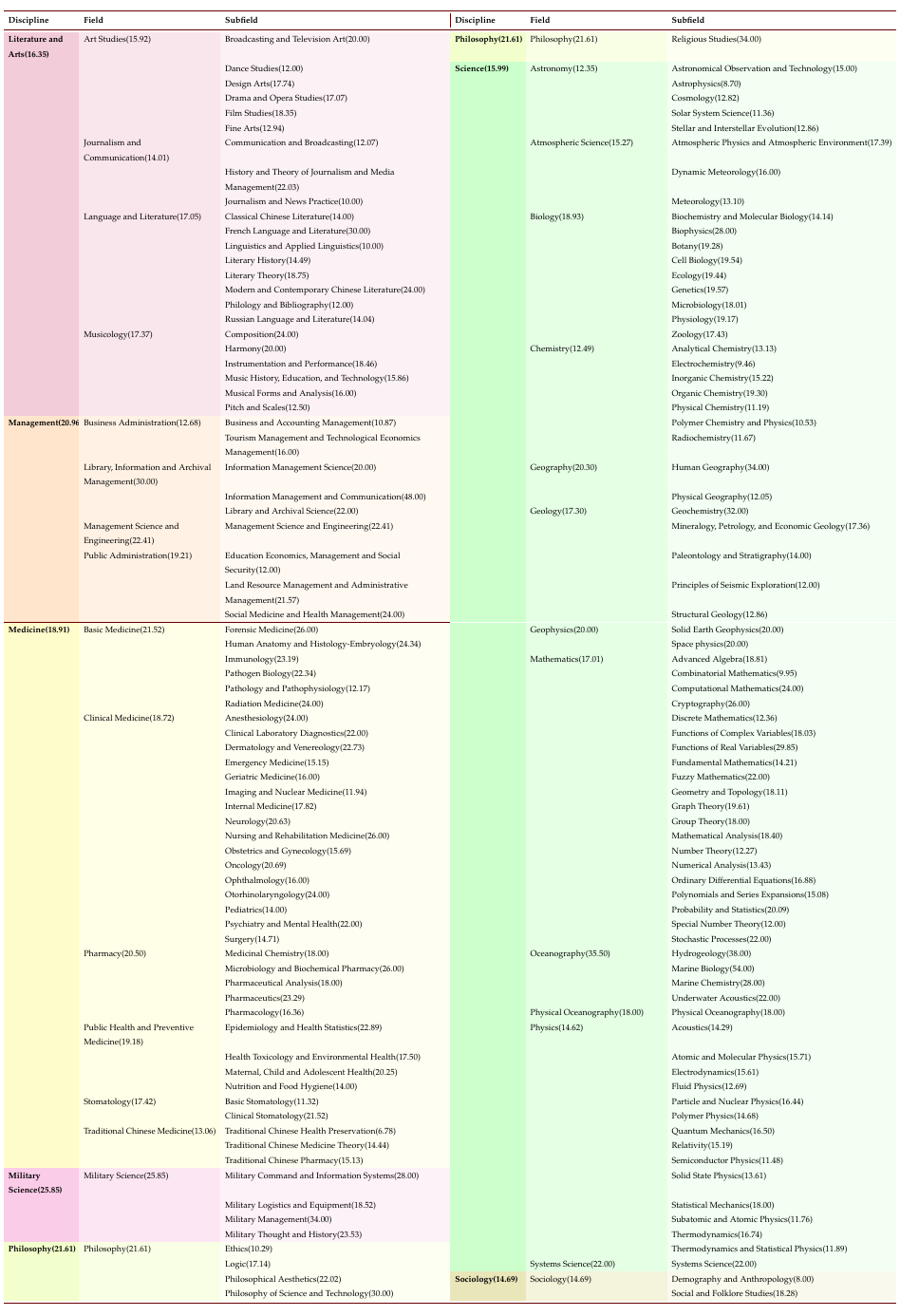} 
    \end{subtable}
    \vspace{-1.1cm}
    \captionsetup{font=small}
    \caption{Continued: Model Scores Across Three Levels of Disciplines: MAP-Neo-7B-Instruct-v0.1.}
    \vspace{-0.6cm}
    \centeredlinks{listofmodels}{Back to List of Models}{toc}{Back to Table of Contents}{blue}
    \end{table}
}
\clearpage

\newpage
\vspace{-0.5cm}
\begin{table}[t]
\refstepcounter{models}%
\addcontentsline{csf}{models}{\protect\numberline{\themodels}OLMo-2-1124-7B-Instruct}
\centering
\begin{subtable}[t]{1\textwidth}
\centering
\includegraphics[width=\textwidth]{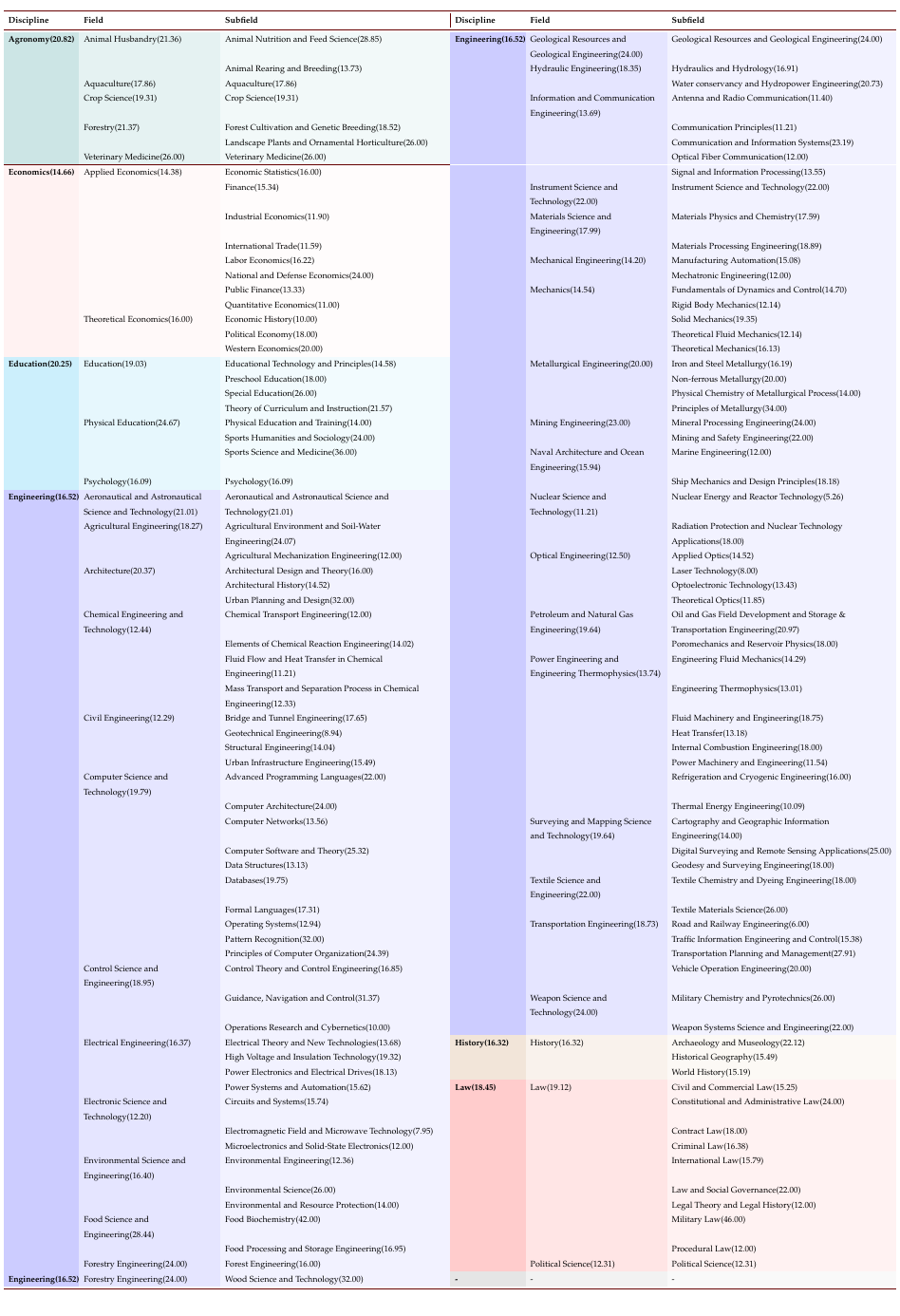} 
\end{subtable}
\vspace{-1.3cm}
\captionsetup{font=small}
\caption{Model Scores Across Three Levels of Disciplines: OLMo-2-1124-7B-Instruct.}
\label{tab:OLMo-2-1124-7B-Instruct}
\vspace{-0.5cm}
\centeredlinks{listofmodels}{Back to List of Models}{toc}{Back to Table of Contents}{blue}
\end{table}
\clearpage

\newpage
\afterpage{
    \begin{table}[t]
    \centering
    \ContinuedFloat 
    \begin{subtable}[t]{\textwidth}
    \centering
    \includegraphics[width=\textwidth]{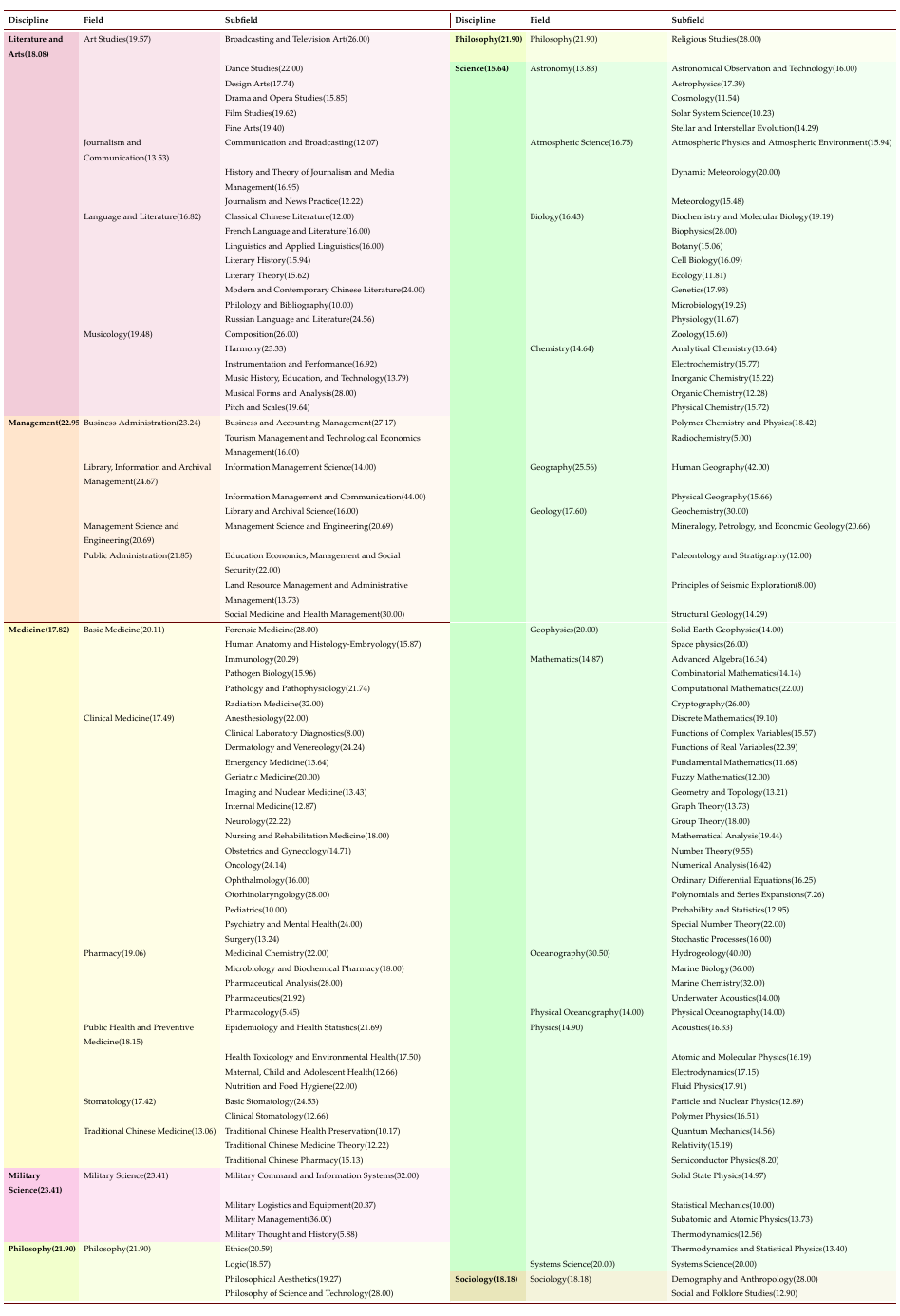} 
    \end{subtable}
    \vspace{-1.1cm}
    \captionsetup{font=small}
    \caption{Continued: Model Scores Across Three Levels of Disciplines: OLMo-2-1124-7B-Instruct.}
    \vspace{-0.6cm}
    \centeredlinks{listofmodels}{Back to List of Models}{toc}{Back to Table of Contents}{blue}
    \end{table}
}
\clearpage

\newpage
\vspace{-0.5cm}
\begin{table}[t]
\refstepcounter{models}%
\addcontentsline{csf}{models}{\protect\numberline{\themodels}granite-3.1-2b-base}
\centering
\begin{subtable}[t]{1\textwidth}
\centering
\includegraphics[width=\textwidth]{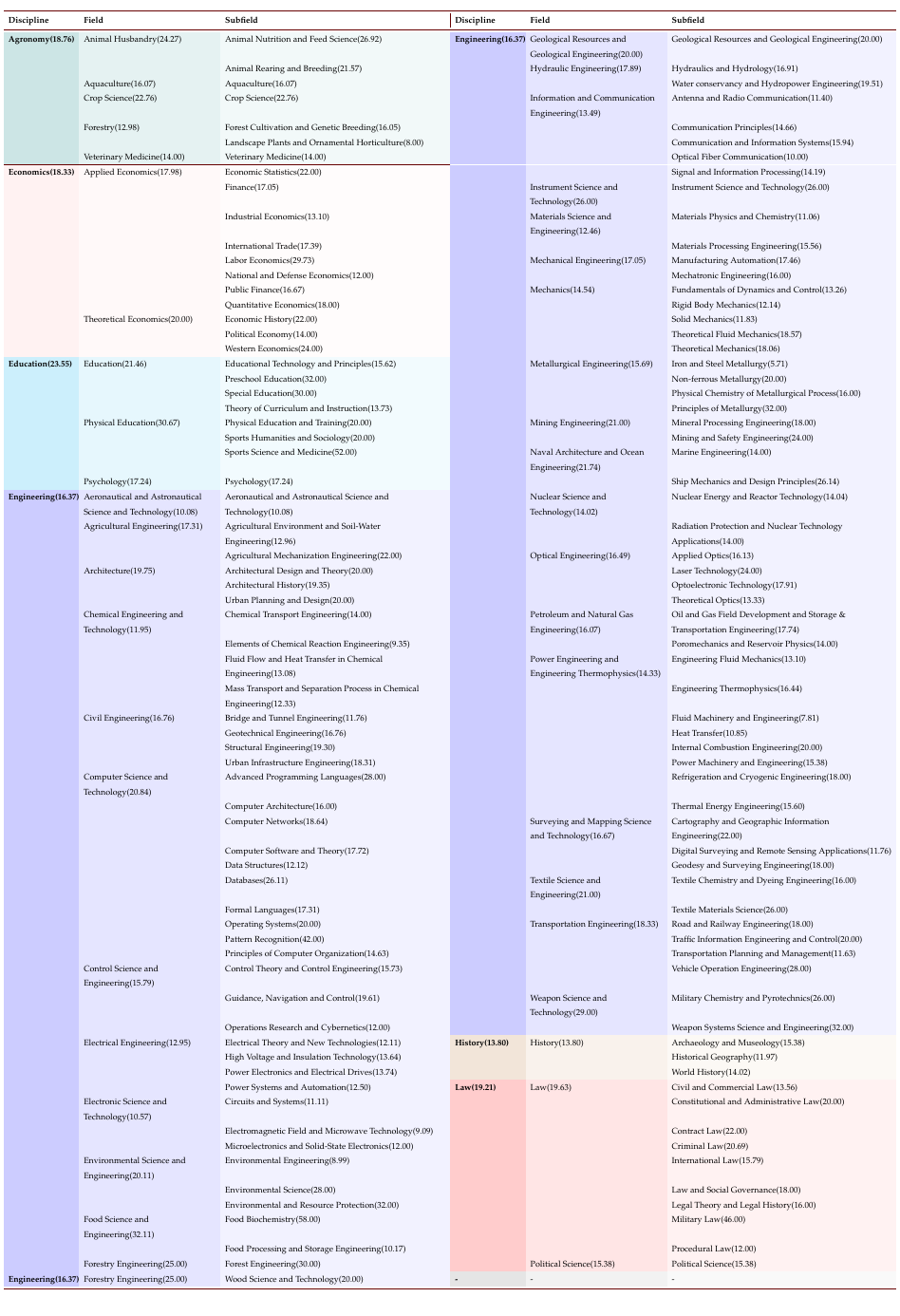} 
\end{subtable}
\vspace{-1.3cm}
\captionsetup{font=small}
\caption{Model Scores Across Three Levels of Disciplines: granite-3.1-2b-base.}
\label{tab:granite-3.1-2b-base}
\vspace{-0.5cm}
\centeredlinks{listofmodels}{Back to List of Models}{toc}{Back to Table of Contents}{blue}
\end{table}
\clearpage

\newpage
\afterpage{
    \begin{table}[t]
    \centering
    \ContinuedFloat 
    \begin{subtable}[t]{\textwidth}
    \centering
    \includegraphics[width=\textwidth]{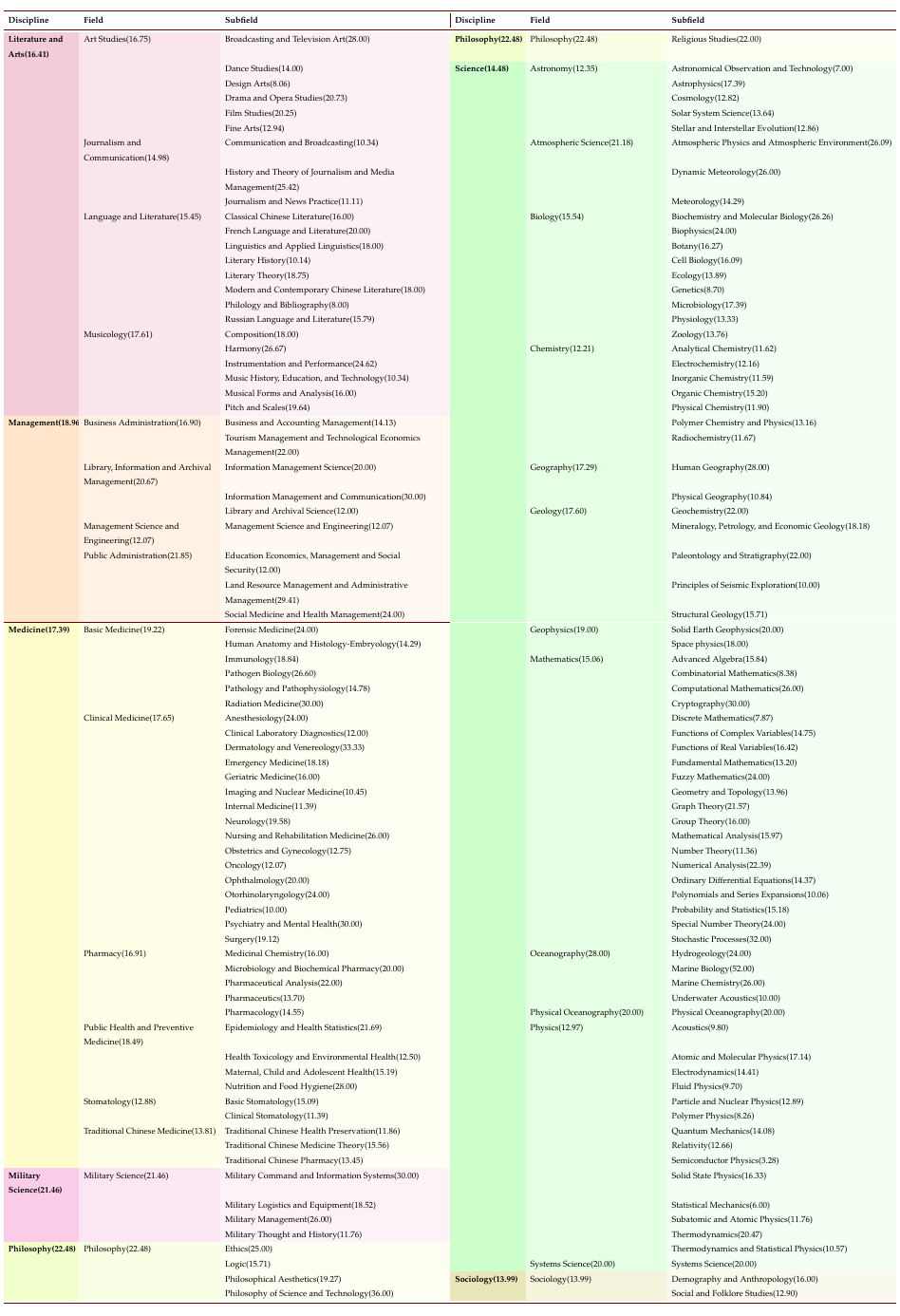} 
    \end{subtable}
    \vspace{-1.1cm}
    \captionsetup{font=small}
    \caption{Continued: Model Scores Across Three Levels of Disciplines: granite-3.1-2b-base.}
    \vspace{-0.6cm}
    \centeredlinks{listofmodels}{Back to List of Models}{toc}{Back to Table of Contents}{blue}
    \end{table}
}
\clearpage

\newpage
\vspace{-0.5cm}
\begin{table}[t]
\refstepcounter{models}%
\addcontentsline{csf}{models}{\protect\numberline{\themodels}OLMo-2-1124-13B}
\centering
\begin{subtable}[t]{1\textwidth}
\centering
\includegraphics[width=\textwidth]{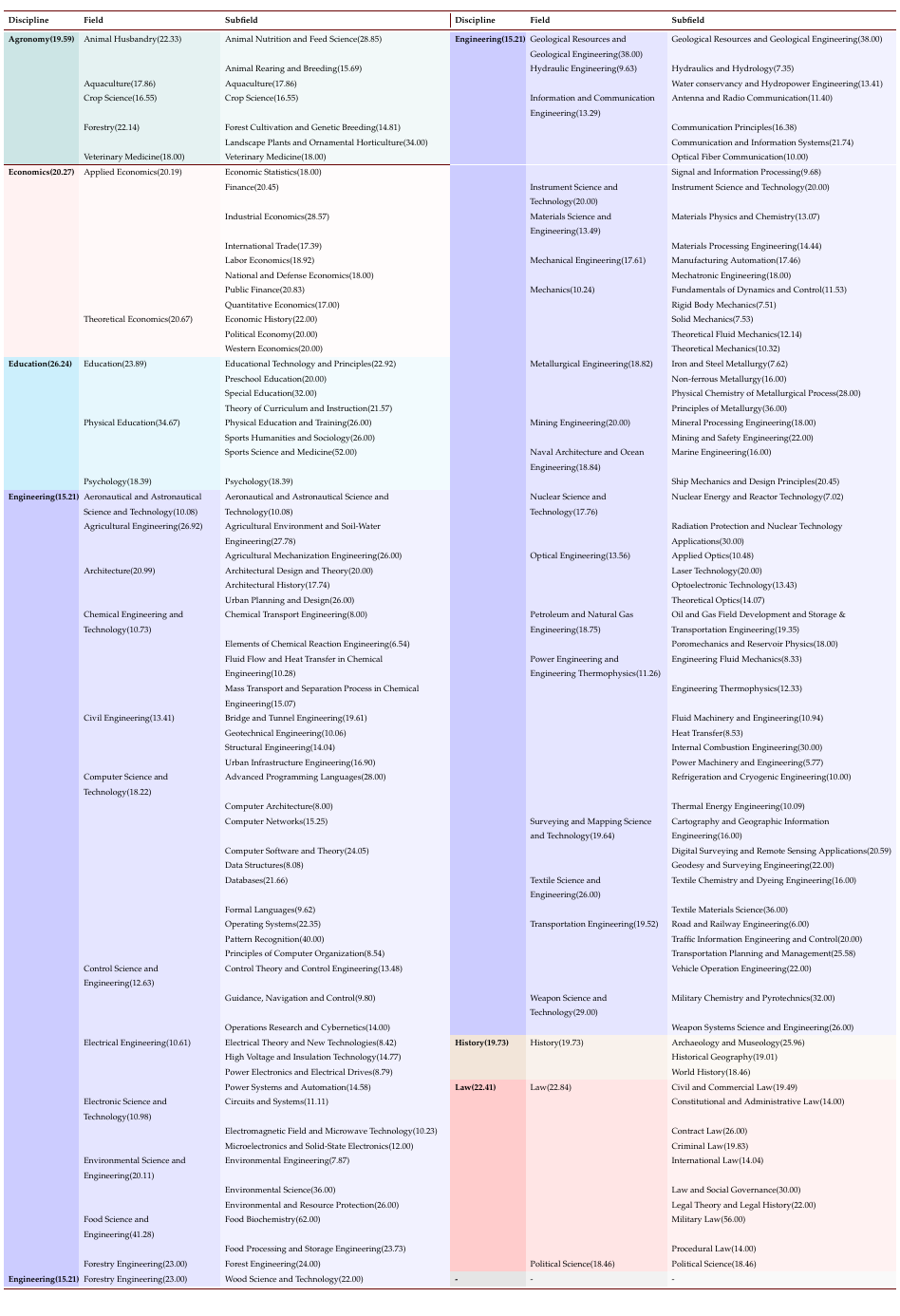} 
\end{subtable}
\vspace{-1.3cm}
\captionsetup{font=small}
\caption{Model Scores Across Three Levels of Disciplines: OLMo-2-1124-13B.}
\label{tab:OLMo-2-1124-13B}
\vspace{-0.5cm}
\centeredlinks{listofmodels}{Back to List of Models}{toc}{Back to Table of Contents}{blue}
\end{table}
\clearpage

\newpage
\afterpage{
    \begin{table}[t]
    \centering
    \ContinuedFloat 
    \begin{subtable}[t]{\textwidth}
    \centering
    \includegraphics[width=\textwidth]{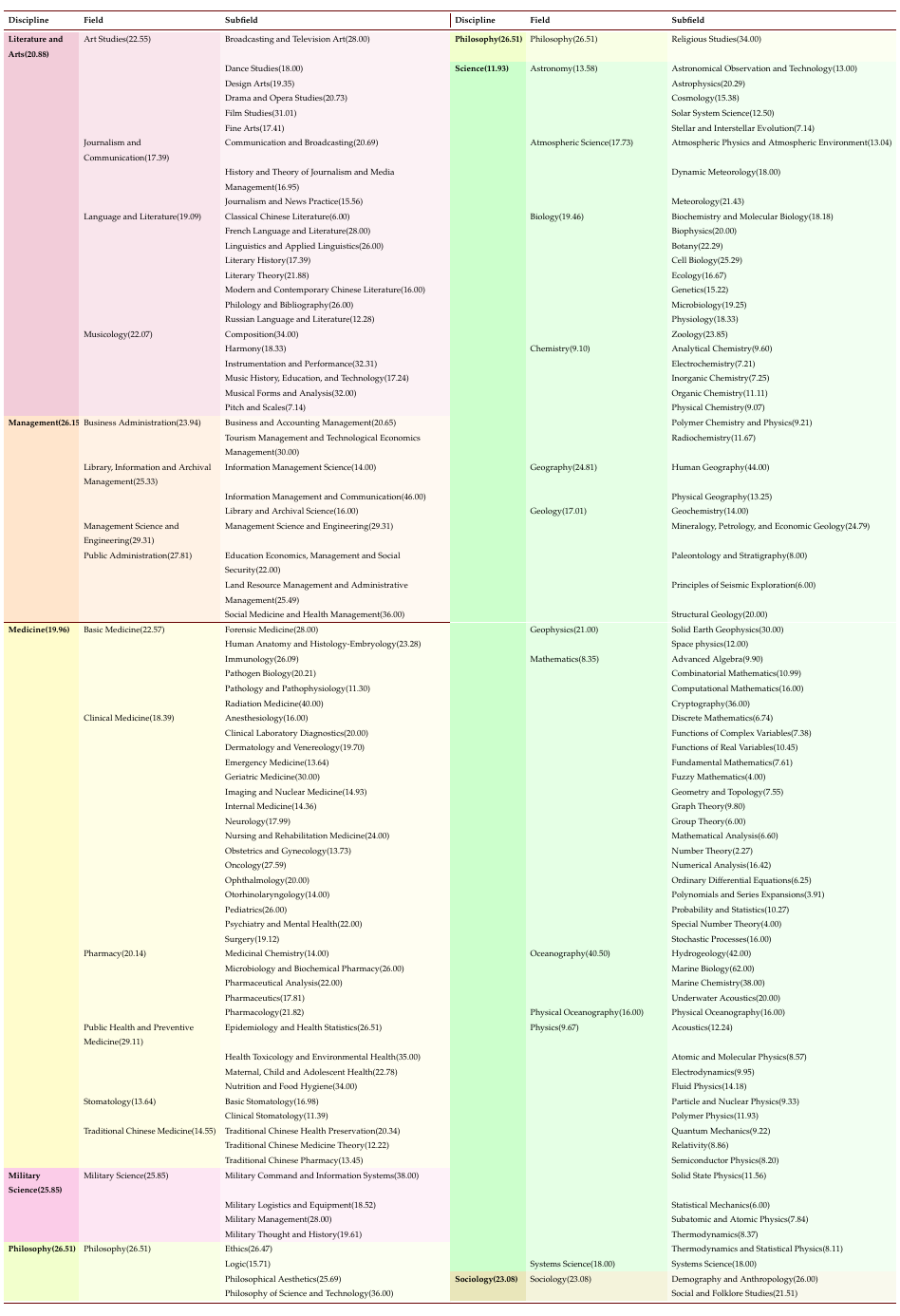} 
    \end{subtable}
    \vspace{-1.1cm}
    \captionsetup{font=small}
    \caption{Continued: Model Scores Across Three Levels of Disciplines: OLMo-2-1124-13B.}
    \vspace{-0.6cm}
    \centeredlinks{listofmodels}{Back to List of Models}{toc}{Back to Table of Contents}{blue}
    \end{table}
}
\clearpage

\newpage
\vspace{-0.5cm}
\begin{table}[t]
\refstepcounter{models}%
\addcontentsline{csf}{models}{\protect\numberline{\themodels}MAP-Neo-7B}
\centering
\begin{subtable}[t]{1\textwidth}
\centering
\includegraphics[width=\textwidth]{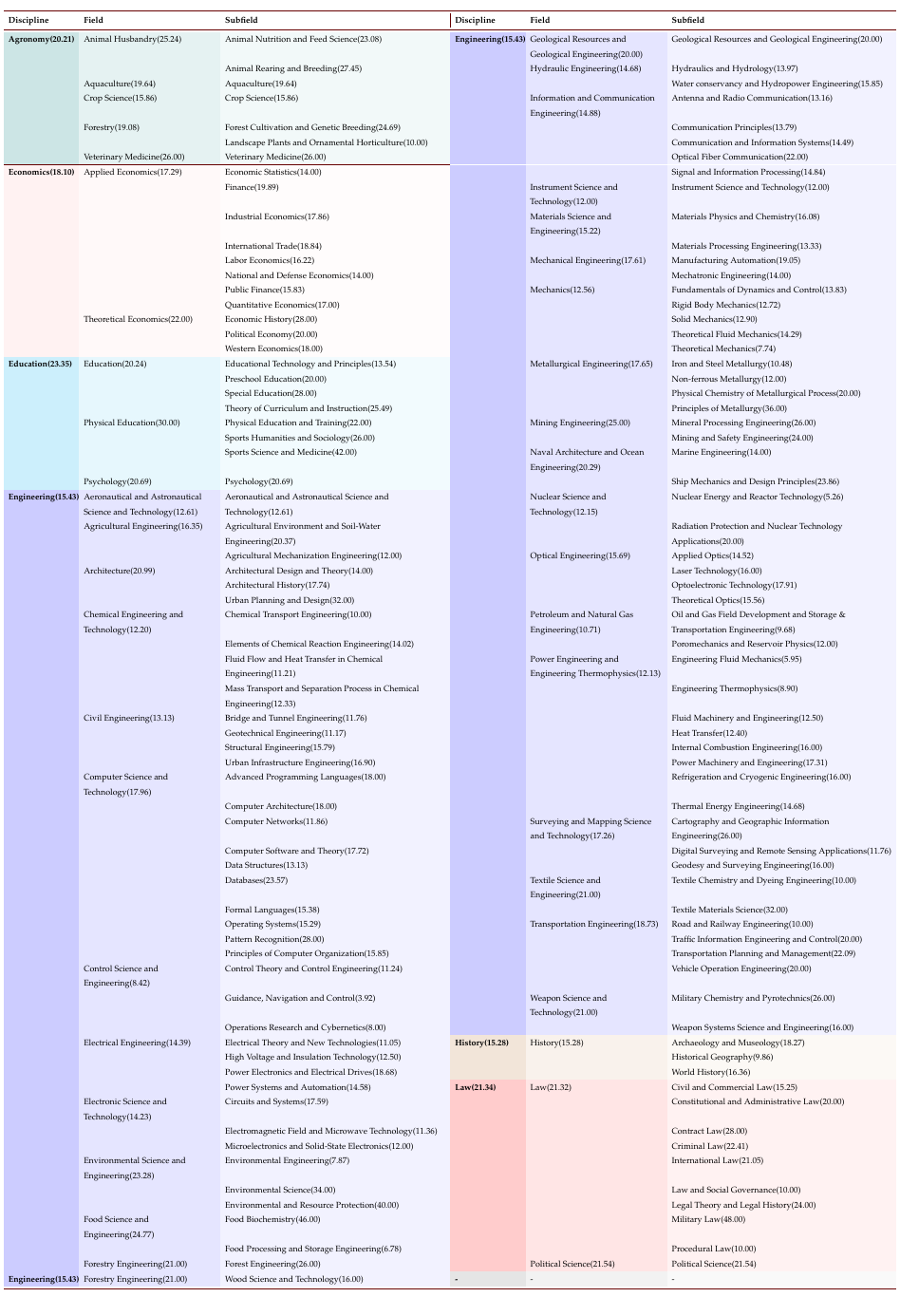} 
\end{subtable}
\vspace{-1.3cm}
\captionsetup{font=small}
\caption{Model Scores Across Three Levels of Disciplines: MAP-Neo-7B.}
\label{tab:MAP-Neo-7B}
\vspace{-0.5cm}
\centeredlinks{listofmodels}{Back to List of Models}{toc}{Back to Table of Contents}{blue}
\end{table}
\clearpage

\newpage
\afterpage{
    \begin{table}[t]
    \centering
    \ContinuedFloat 
    \begin{subtable}[t]{\textwidth}
    \centering
    \includegraphics[width=\textwidth]{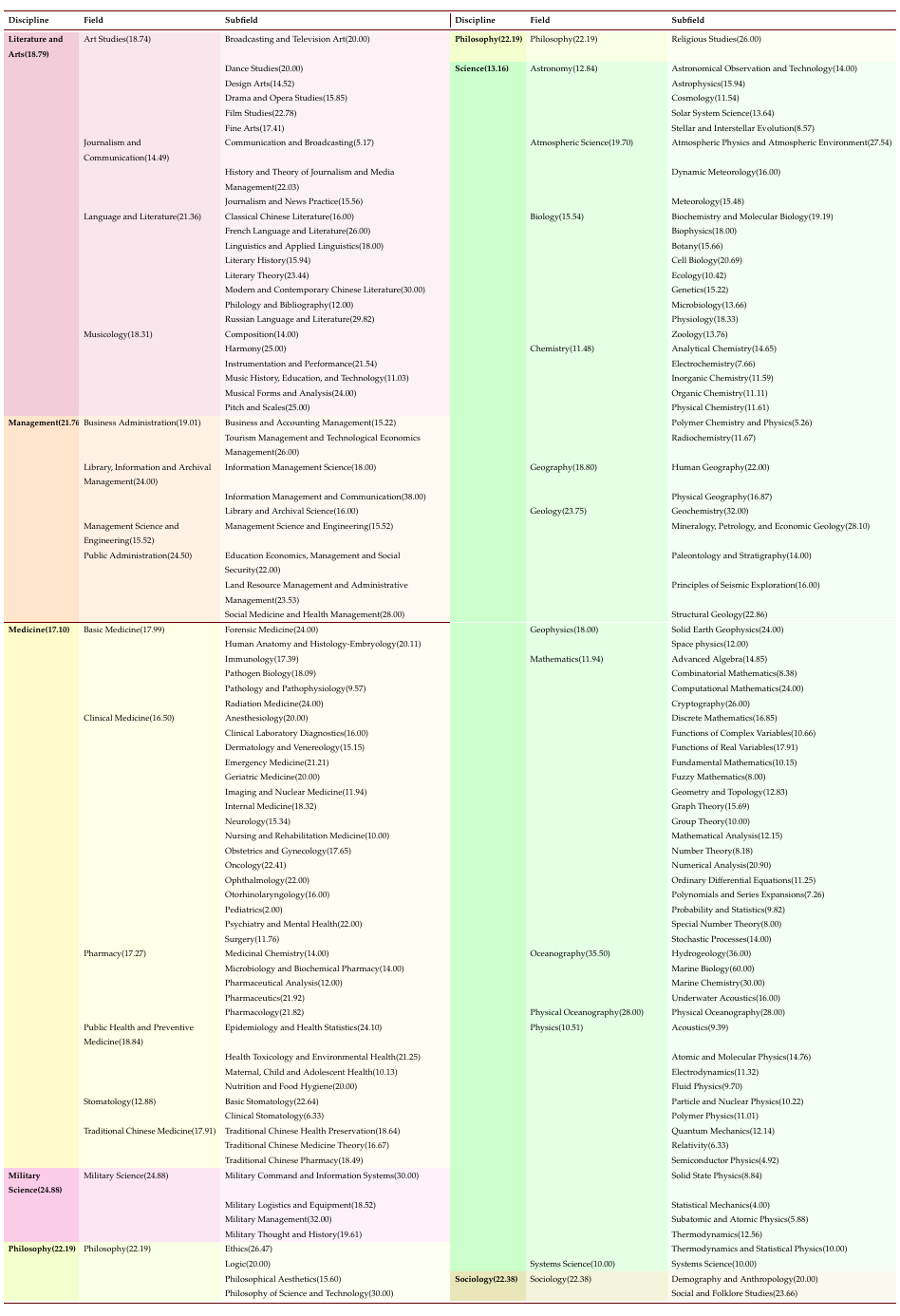} 
    \end{subtable}
    \vspace{-1.1cm}
    \captionsetup{font=small}
    \caption{Continued: Model Scores Across Three Levels of Disciplines: MAP-Neo-7B.}
    \vspace{-0.6cm}
    \centeredlinks{listofmodels}{Back to List of Models}{toc}{Back to Table of Contents}{blue}
    \end{table}
}
\clearpage

\newpage
\vspace{-0.5cm}
\begin{table}[t]
\refstepcounter{models}%
\addcontentsline{csf}{models}{\protect\numberline{\themodels}granite-3.1-8b-base}
\centering
\begin{subtable}[t]{1\textwidth}
\centering
\includegraphics[width=\textwidth]{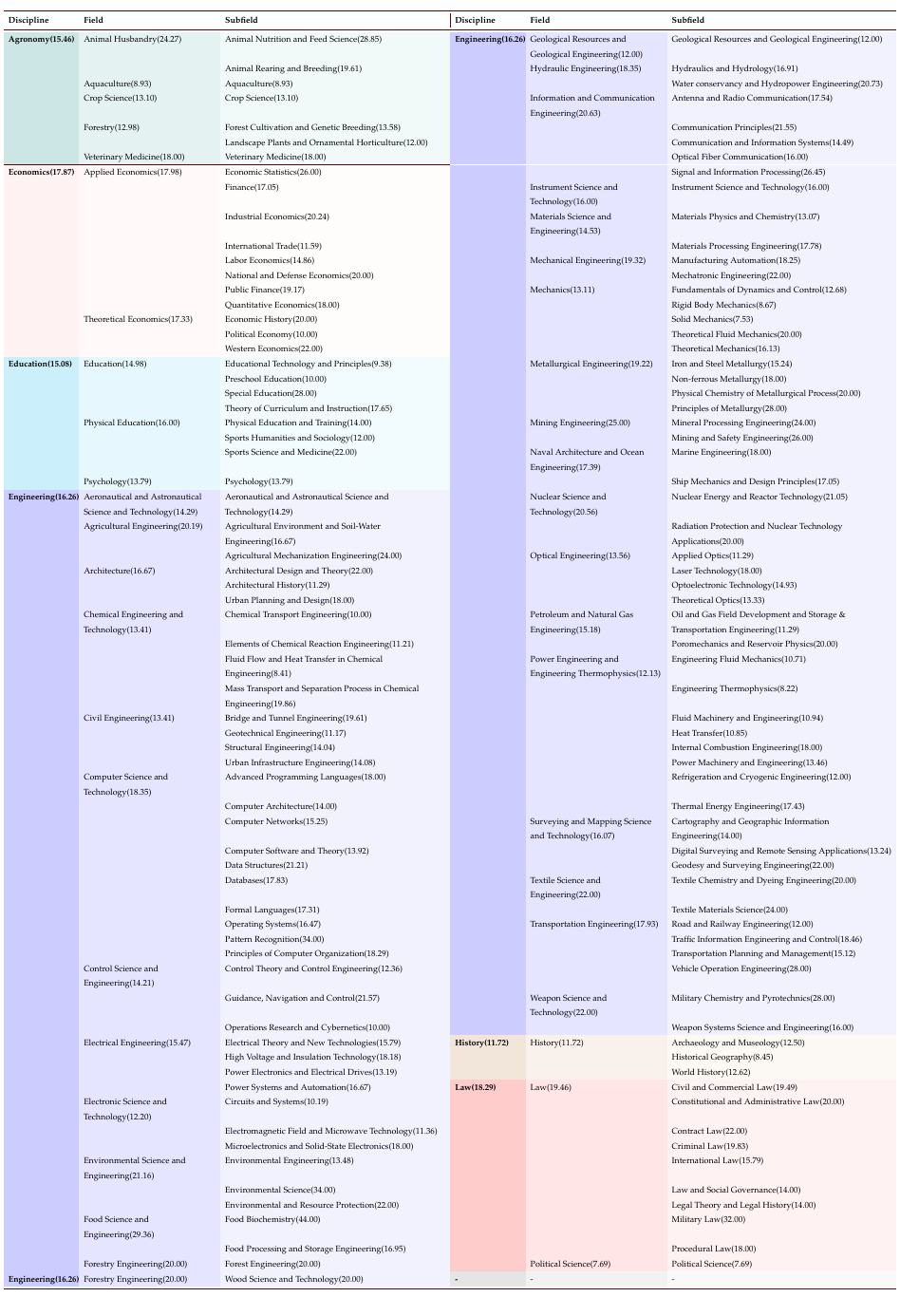} 
\end{subtable}
\vspace{-1.3cm}
\captionsetup{font=small}
\caption{Model Scores Across Three Levels of Disciplines: granite-3.1-8b-base.}
\label{tab:granite-3.1-8b-base}
\vspace{-0.5cm}
\centeredlinks{listofmodels}{Back to List of Models}{toc}{Back to Table of Contents}{blue}
\end{table}
\clearpage

\newpage
\afterpage{
    \begin{table}[t]
    \centering
    \ContinuedFloat 
    \begin{subtable}[t]{\textwidth}
    \centering
    \includegraphics[width=\textwidth]{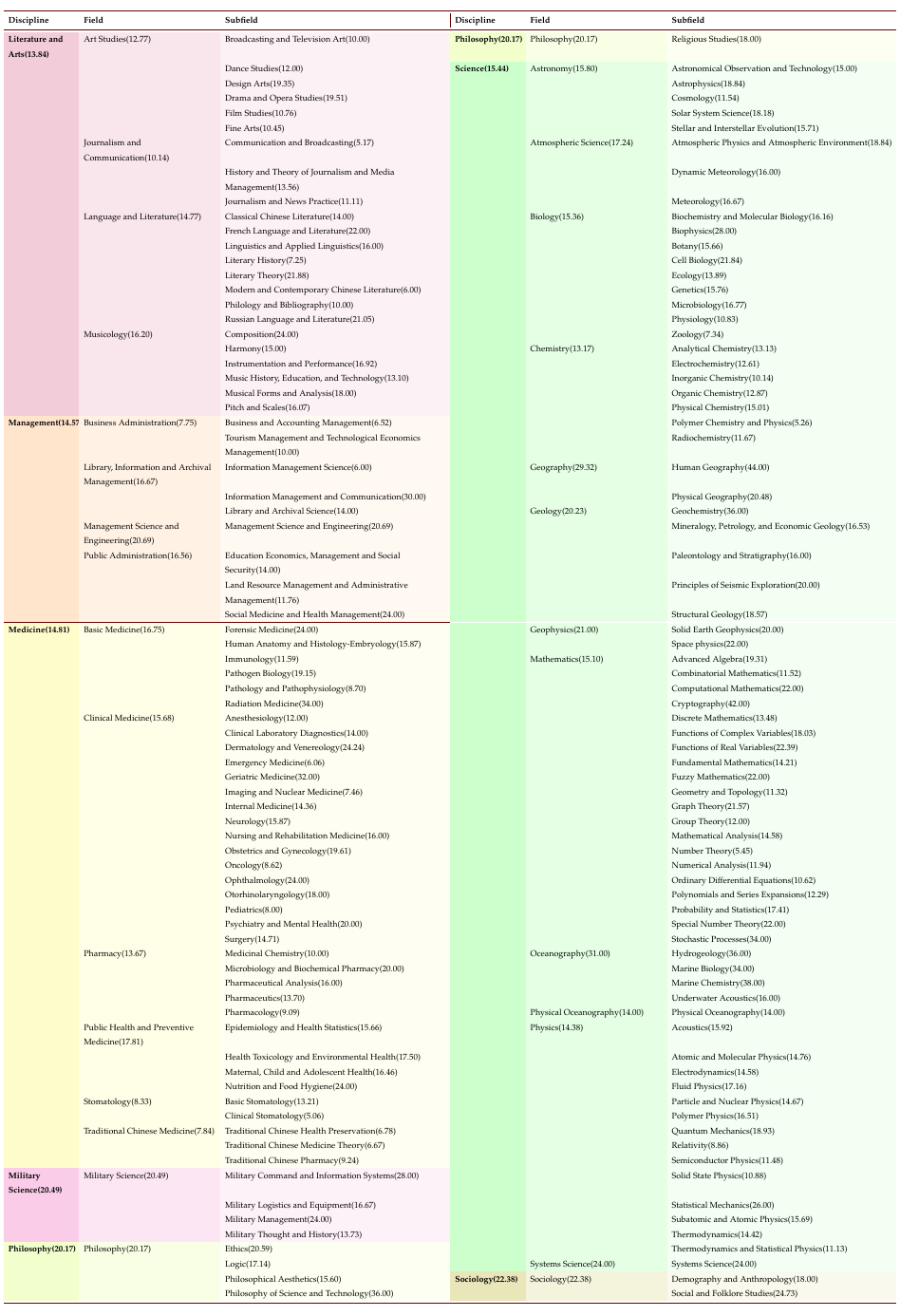} 
    \end{subtable}
    \vspace{-1.1cm}
    \captionsetup{font=small}
    \caption{Continued: Model Scores Across Three Levels of Disciplines: granite-3.1-8b-base.}
    \vspace{-0.6cm}
    \centeredlinks{listofmodels}{Back to List of Models}{toc}{Back to Table of Contents}{blue}
    \end{table}
}
\clearpage

\newpage
\vspace{-0.5cm}
\begin{table}[t]
\refstepcounter{models}%
\addcontentsline{csf}{models}{\protect\numberline{\themodels}OLMo-2-1124-7B}
\centering
\begin{subtable}[t]{1\textwidth}
\centering
\includegraphics[width=\textwidth]{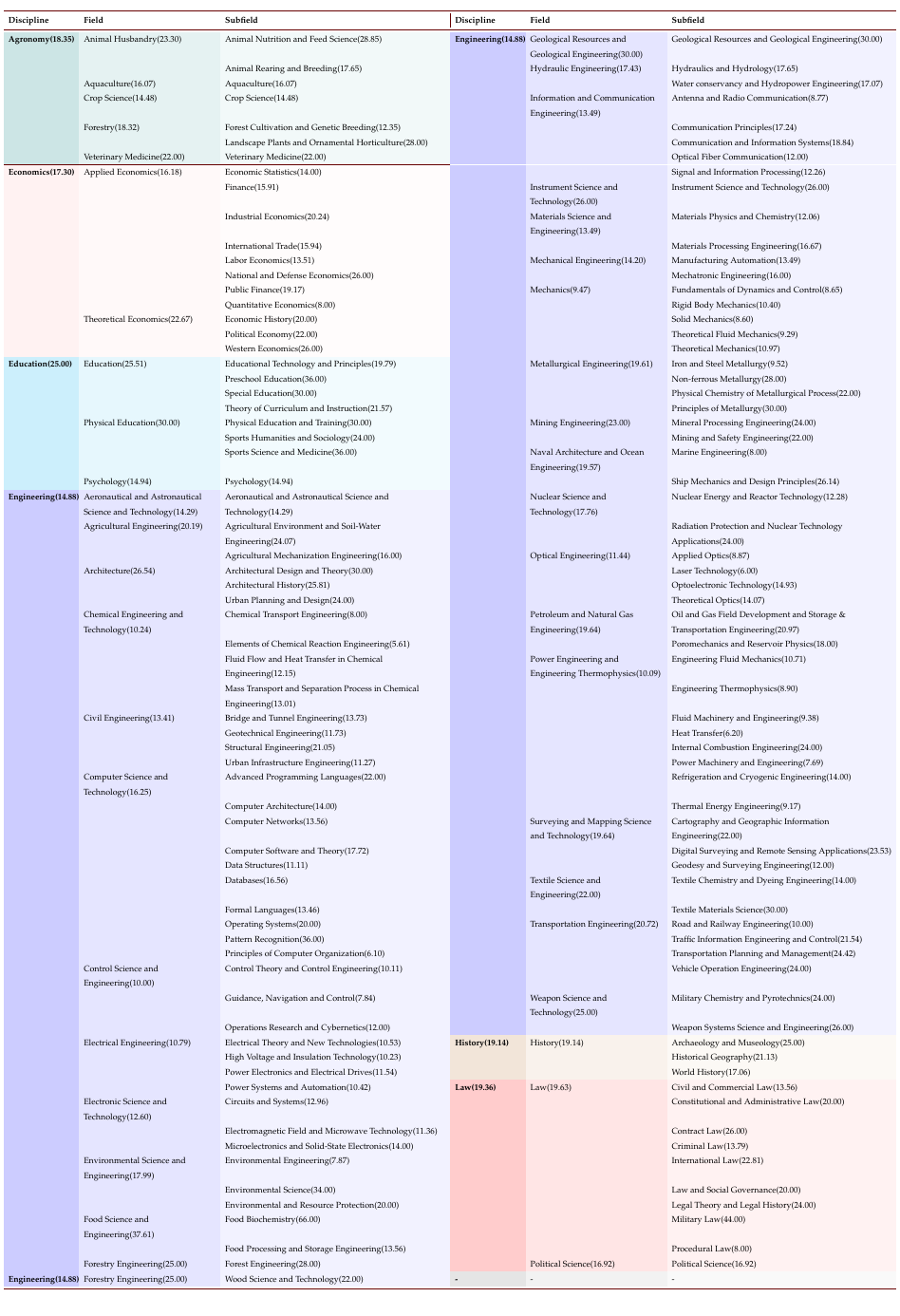} 
\end{subtable}
\vspace{-1.3cm}
\captionsetup{font=small}
\caption{Model Scores Across Three Levels of Disciplines: OLMo-2-1124-7B.}
\label{tab:OLMo-2-1124-7B}
\vspace{-0.5cm}
\centeredlinks{listofmodels}{Back to List of Models}{toc}{Back to Table of Contents}{blue}
\end{table}
\clearpage

\newpage
\afterpage{
    \begin{table}[t]
    \centering
    \ContinuedFloat 
    \begin{subtable}[t]{\textwidth}
    \centering
    \includegraphics[width=\textwidth]{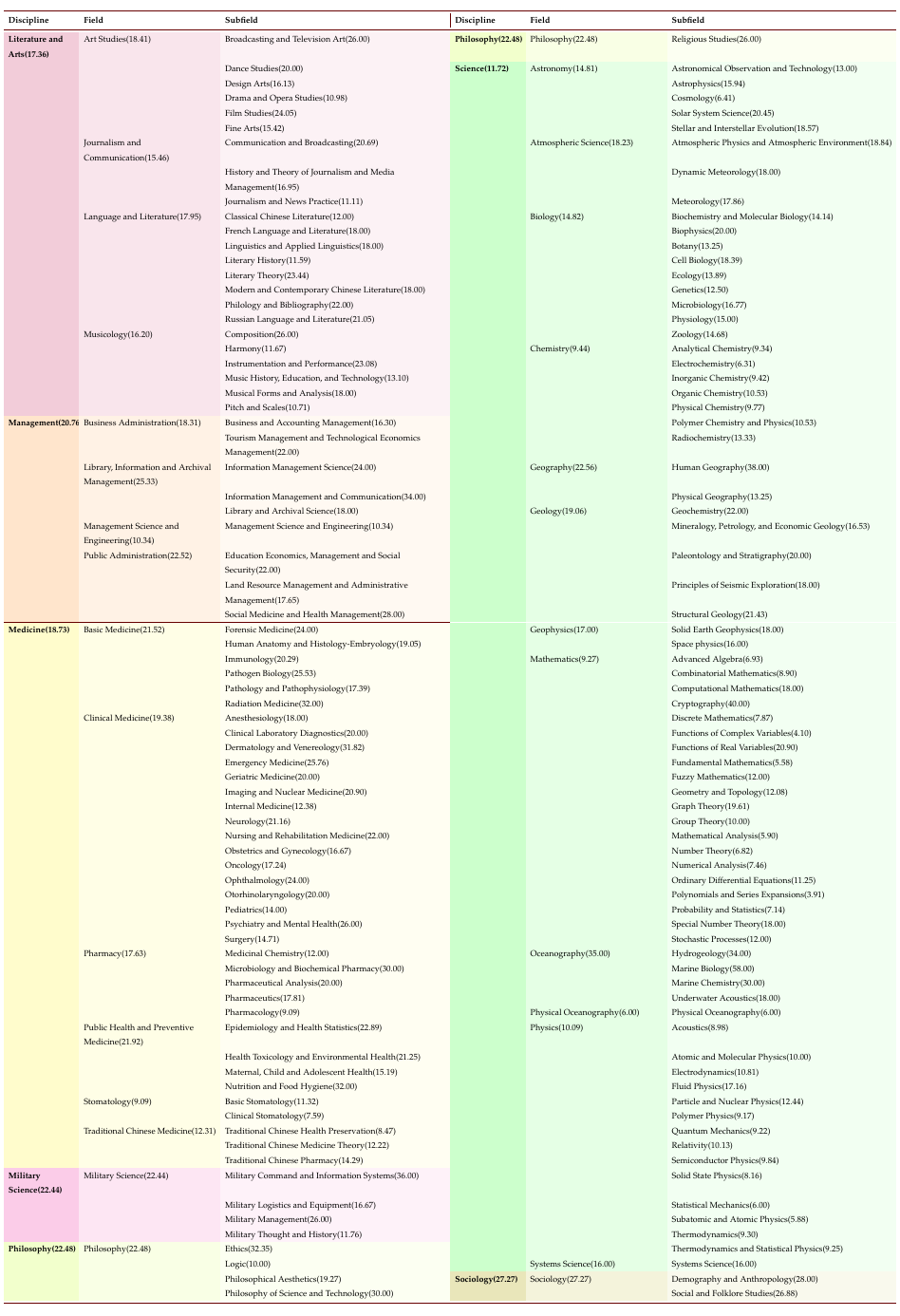} 
    \end{subtable}
    \vspace{-1.1cm}
    \captionsetup{font=small}
    \caption{Continued: Model Scores Across Three Levels of Disciplines: OLMo-2-1124-7B.}
    \vspace{-0.6cm}
    \centeredlinks{listofmodels}{Back to List of Models}{toc}{Back to Table of Contents}{blue}
    \end{table}
}
\clearpage

\newpage
\vspace{-0.5cm}
\begin{table}[t]
\refstepcounter{models}%
\addcontentsline{csf}{models}{\protect\numberline{\themodels}gemma-2-2b}
\centering
\begin{subtable}[t]{1\textwidth}
\centering
\includegraphics[width=\textwidth]{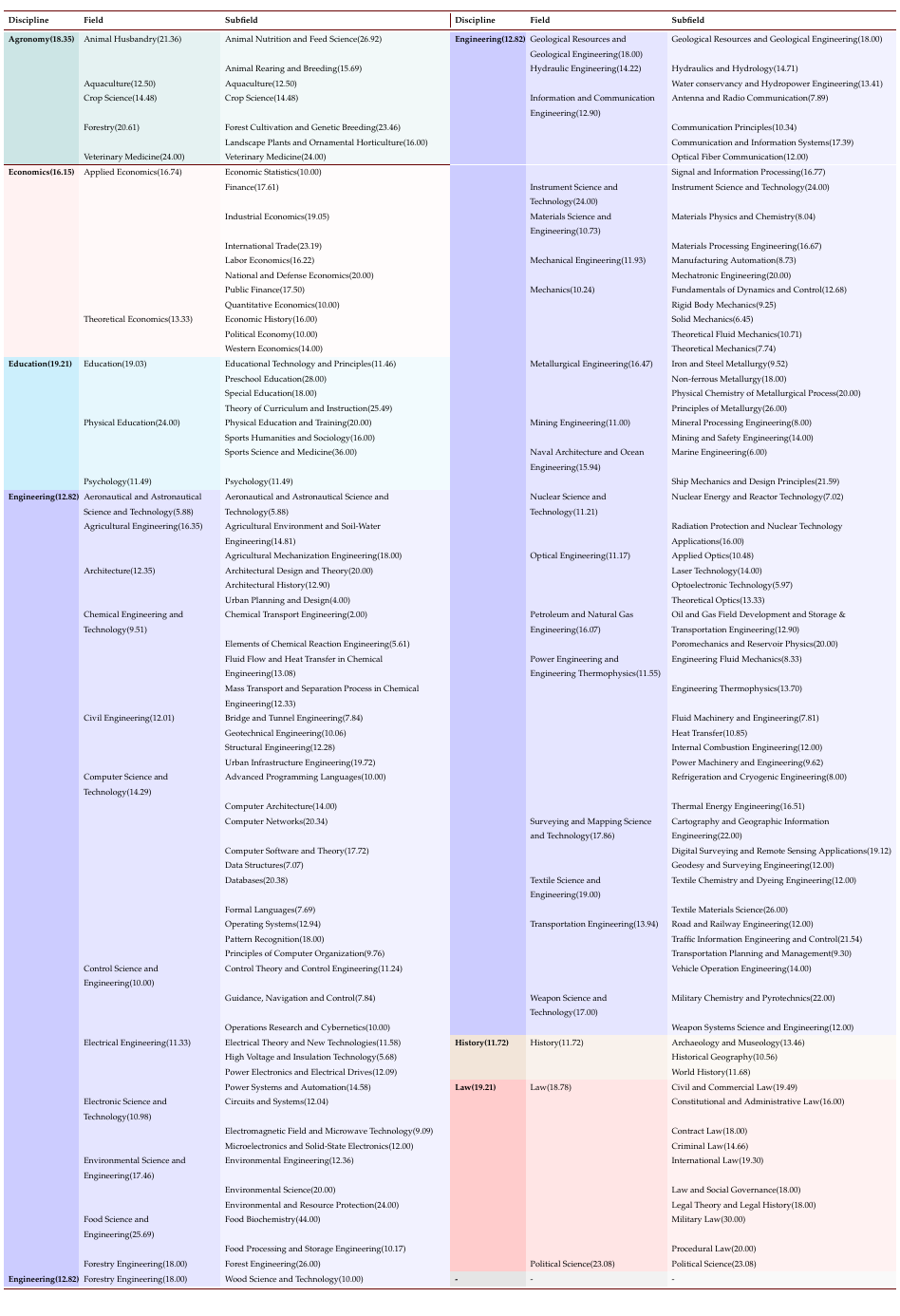} 
\end{subtable}
\vspace{-1.3cm}
\captionsetup{font=small}
\caption{Model Scores Across Three Levels of Disciplines: gemma-2-2b.}
\label{tab:gemma-2-2b}
\vspace{-0.5cm}
\centeredlinks{listofmodels}{Back to List of Models}{toc}{Back to Table of Contents}{blue}
\end{table}
\clearpage

\newpage
\afterpage{
    \begin{table}[t]
    \centering
    \ContinuedFloat 
    \begin{subtable}[t]{\textwidth}
    \centering
    \includegraphics[width=\textwidth]{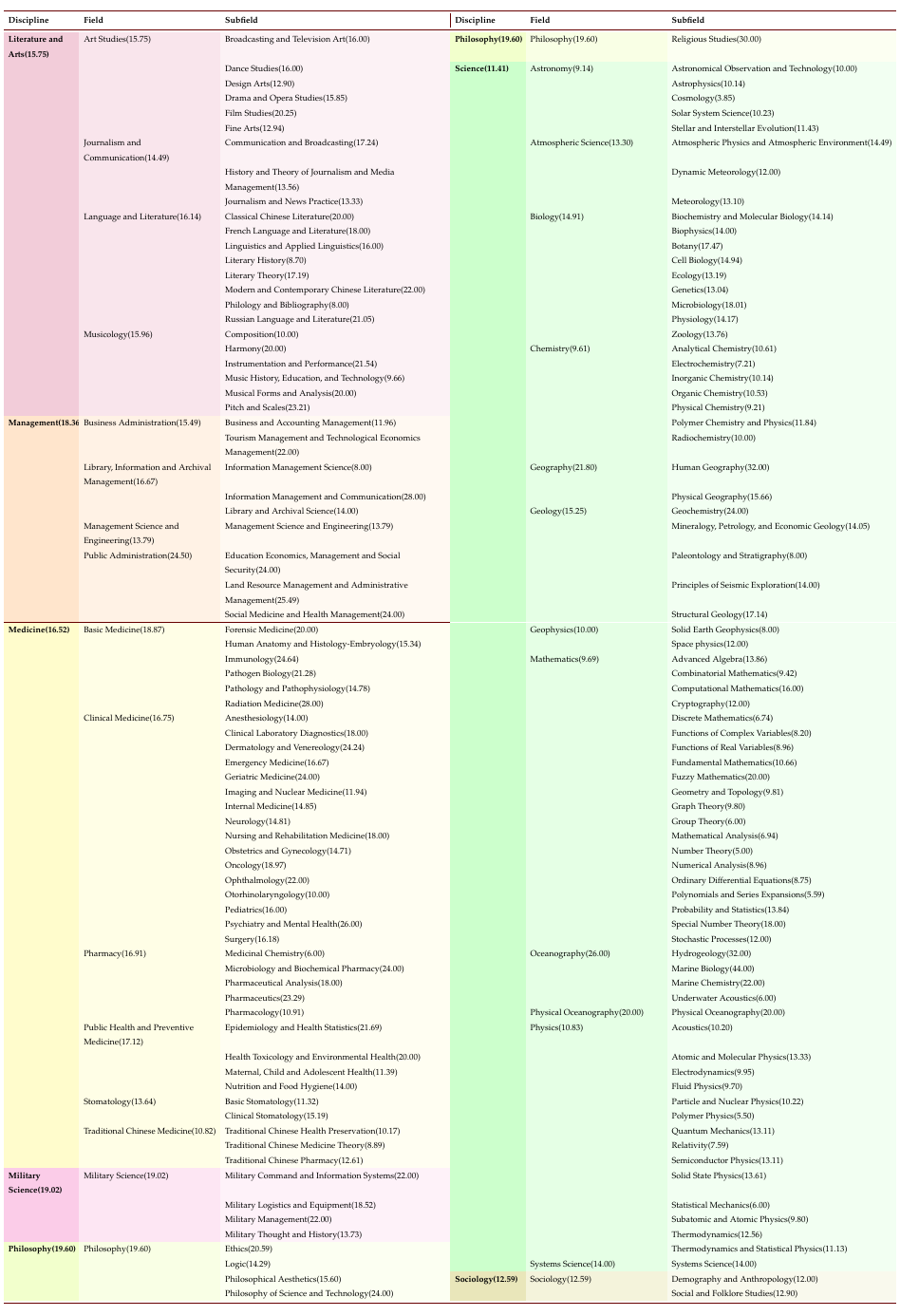} 
    \end{subtable}
    \vspace{-1.1cm}
    \captionsetup{font=small}
    \caption{Continued: Model Scores Across Three Levels of Disciplines: gemma-2-2b.}
    \vspace{-0.6cm}
    \centeredlinks{listofmodels}{Back to List of Models}{toc}{Back to Table of Contents}{blue}
    \end{table}
}
\clearpage

\newpage
\vspace{-0.5cm}
\begin{table}[t]
\refstepcounter{models}%
\addcontentsline{csf}{models}{\protect\numberline{\themodels}Qwen2.5-0.5B-Instruct}
\centering
\begin{subtable}[t]{1\textwidth}
\centering
\includegraphics[width=\textwidth]{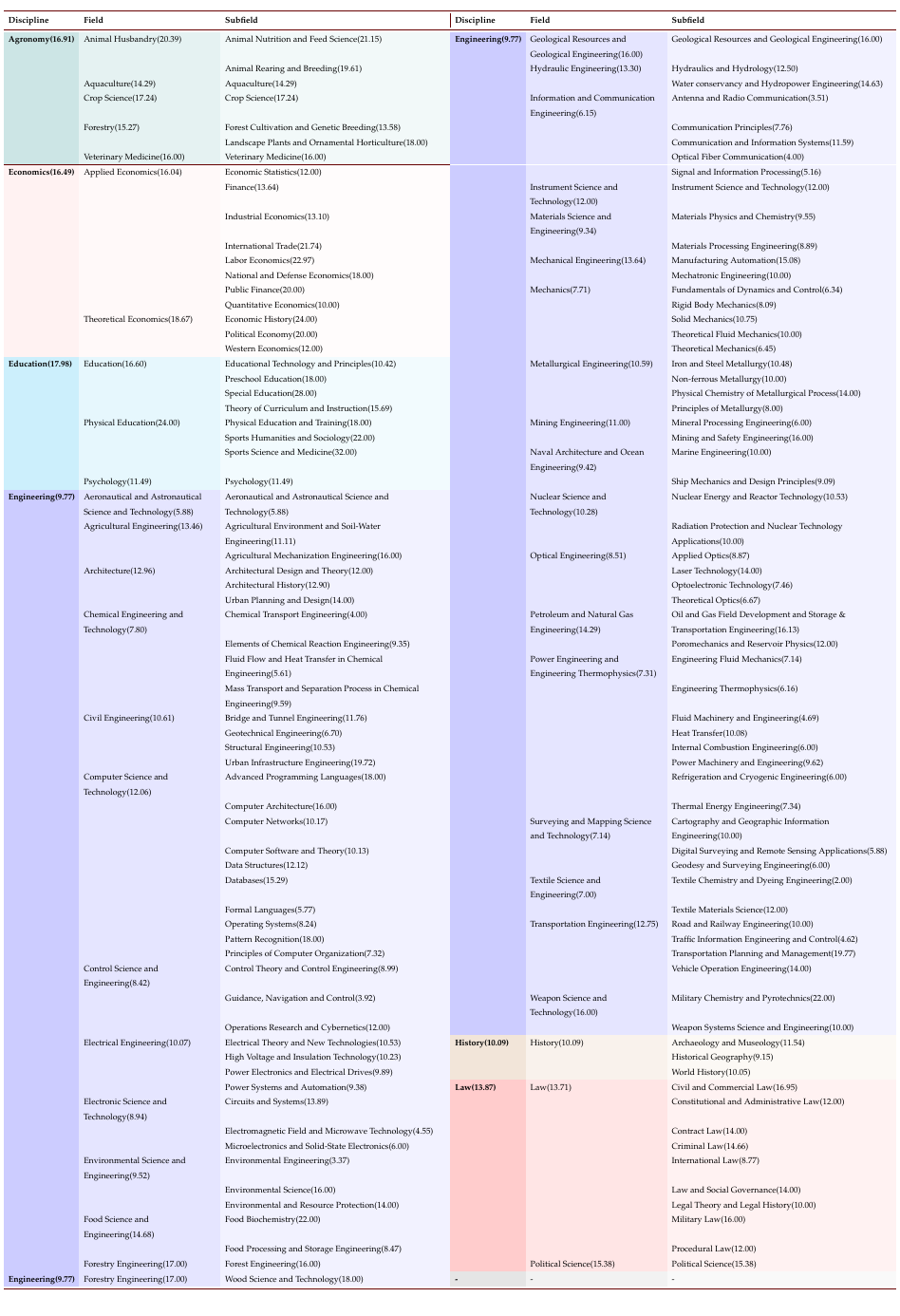} 
\end{subtable}
\vspace{-1.3cm}
\captionsetup{font=small}
\caption{Model Scores Across Three Levels of Disciplines: Qwen2.5-0.5B-Instruct.}
\label{tab:Qwen2.5-0.5B-Instruct}
\vspace{-0.5cm}
\centeredlinks{listofmodels}{Back to List of Models}{toc}{Back to Table of Contents}{blue}
\end{table}
\clearpage

\newpage
\afterpage{
    \begin{table}[t]
    \centering
    \ContinuedFloat 
    \begin{subtable}[t]{\textwidth}
    \centering
    \includegraphics[width=\textwidth]{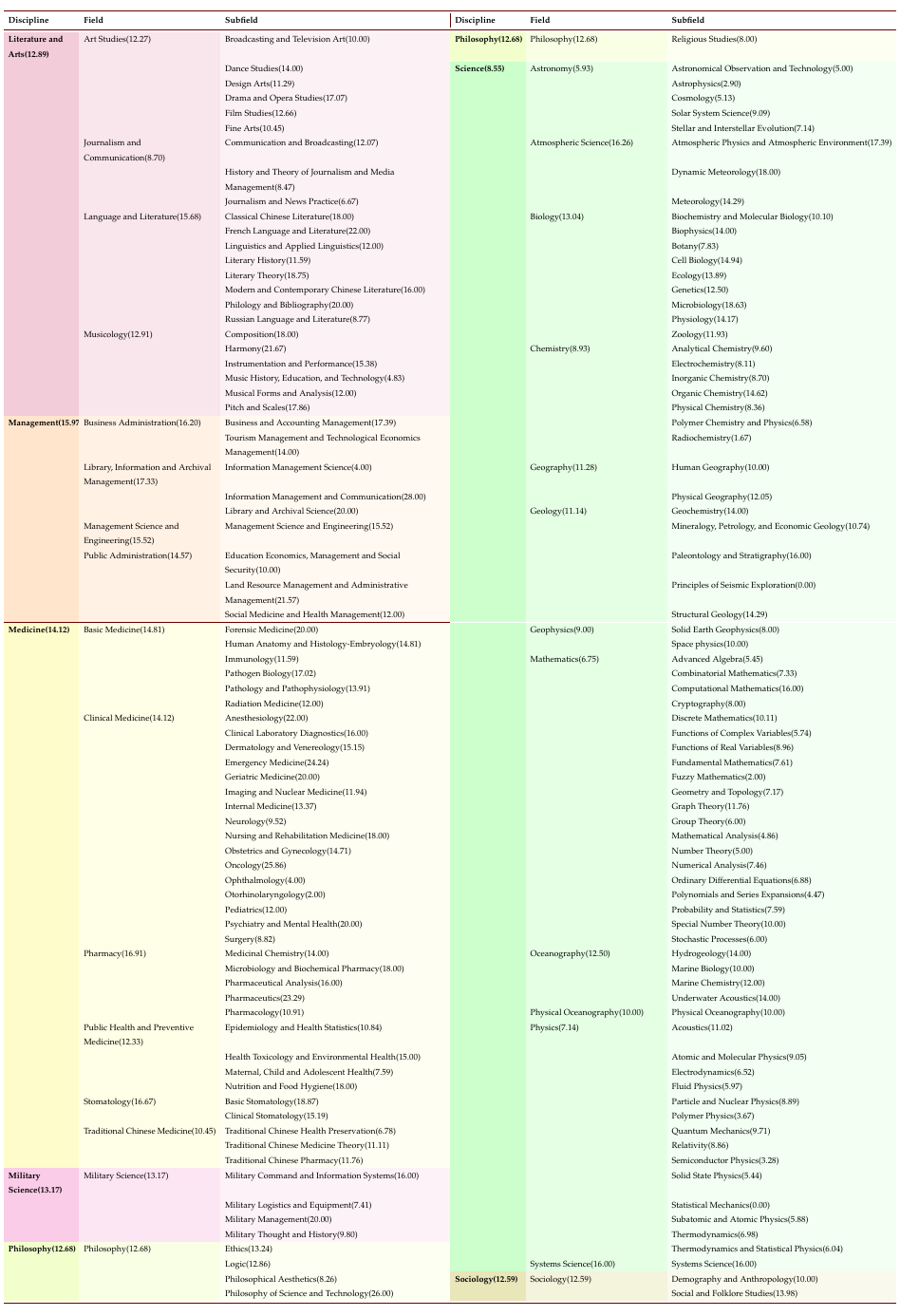} 
    \end{subtable}
    \vspace{-1.1cm}
    \captionsetup{font=small}
    \caption{Continued: Model Scores Across Three Levels of Disciplines: Qwen2.5-0.5B-Instruct.}
    \vspace{-0.6cm}
    \centeredlinks{listofmodels}{Back to List of Models}{toc}{Back to Table of Contents}{blue}
    \end{table}
}
\clearpage

\newpage
\vspace{-0.5cm}
\begin{table}[t]
\refstepcounter{models}%
\addcontentsline{csf}{models}{\protect\numberline{\themodels}Qwen2.5-0.5B}
\centering
\begin{subtable}[t]{1\textwidth}
\centering
\includegraphics[width=\textwidth]{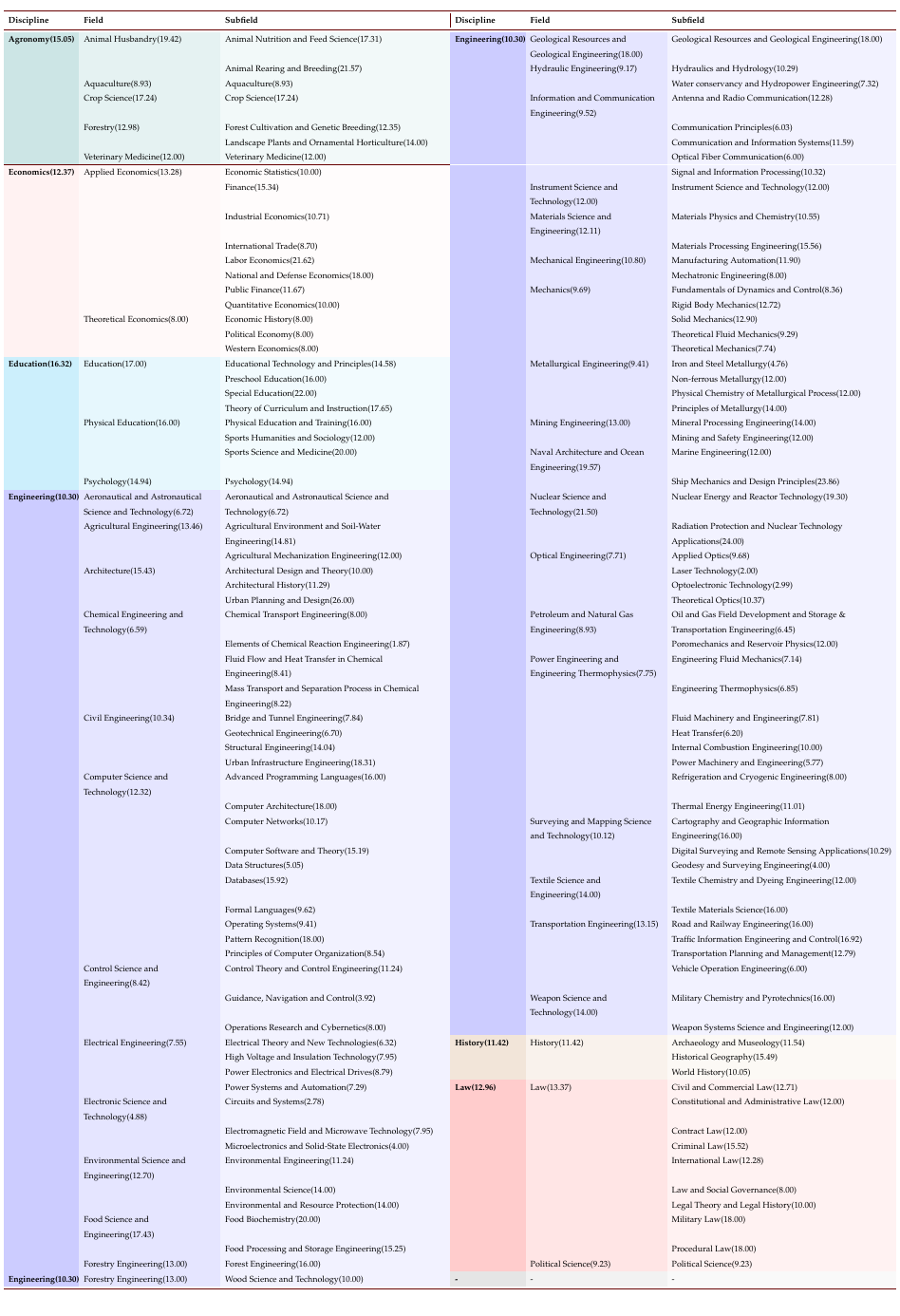} 
\end{subtable}
\vspace{-1.3cm}
\captionsetup{font=small}
\caption{Model Scores Across Three Levels of Disciplines: Qwen2.5-0.5B.}
\label{tab:Qwen2.5-0.5B}
\vspace{-0.5cm}
\centeredlinks{listofmodels}{Back to List of Models}{toc}{Back to Table of Contents}{blue}
\end{table}
\clearpage

\newpage
\afterpage{
    \begin{table}[t]
    \centering
    \ContinuedFloat 
    \begin{subtable}[t]{\textwidth}
    \centering
    \includegraphics[width=\textwidth]{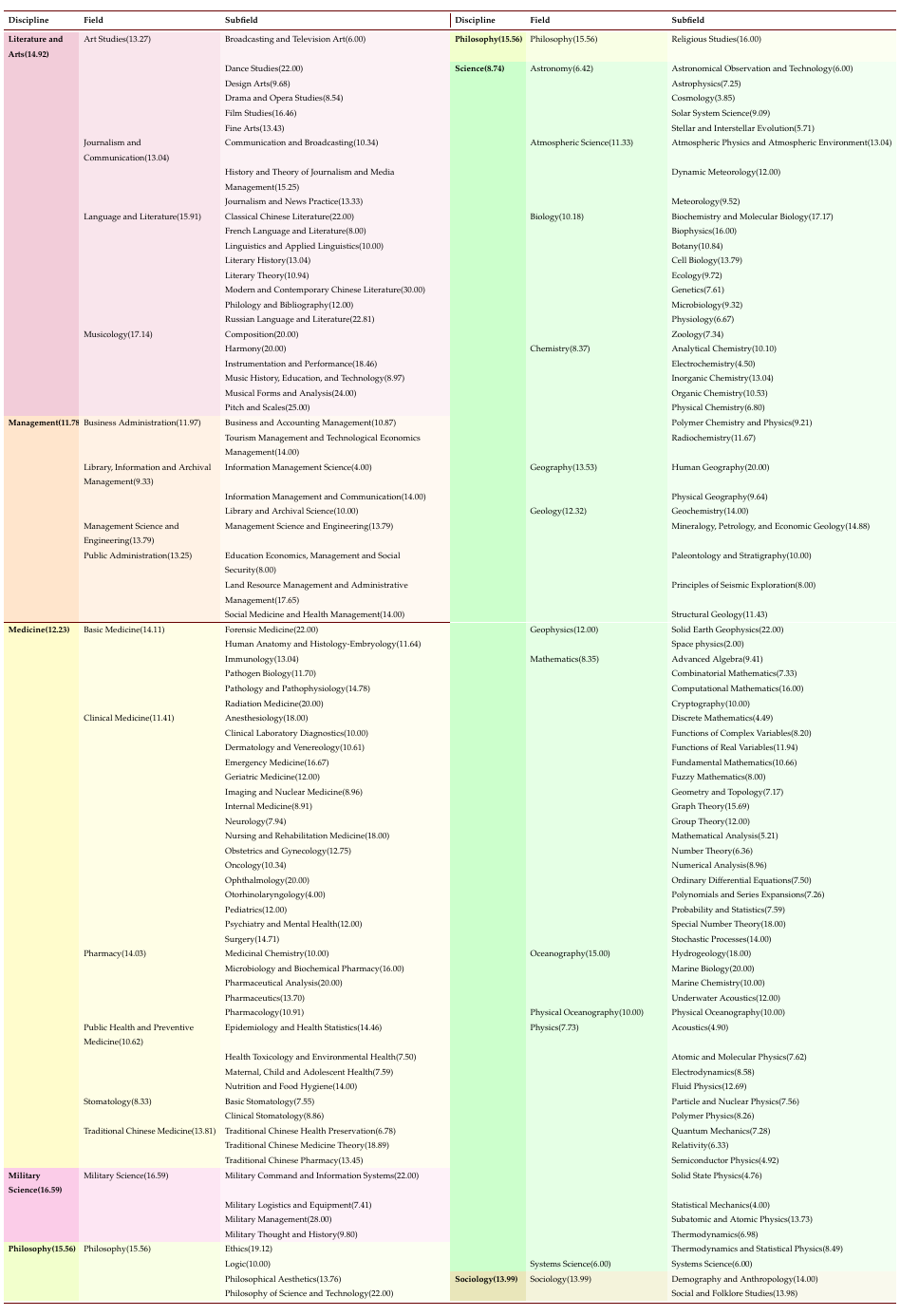} 
    \end{subtable}
    \vspace{-1.1cm}
    \captionsetup{font=small}
    \caption{Continued: Model Scores Across Three Levels of Disciplines: Qwen2.5-0.5B.}
    \vspace{-0.6cm}
    \centeredlinks{listofmodels}{Back to List of Models}{toc}{Back to Table of Contents}{blue}
    \end{table}
}
\clearpage

\end{CJK*}
\end{document}